%% file: main.tex
\newif\ifreview 
\newif\ifarxiv \newcommand{\arxiv}{\arxivtrue}
\newif\ifcamera \newcommand{\cameraready}{\cameratrue}
\newif\ifrebuttal 

\cameraready
\arxiv

\documentclass[10pt,twocolumn,letterpaper]{article}

\ifreview \usepackage[review]{cvpr} \fi
\ifarxiv \usepackage[pagenumbers]{cvpr} \fi
\ifrebuttal \usepackage[rebuttal]{cvpr} \fi
\ifcamera \usepackage{cvpr} \fi

\usepackage{graphicx}
\usepackage{amsmath}
\usepackage{amssymb}
\usepackage{booktabs}

\usepackage{listings}
\usepackage{numprint}
\usepackage{nicefrac}
\usepackage{hyphenat}
\usepackage{url}
\usepackage{multirow}
\usepackage{multicol}
\usepackage{gensymb}
\usepackage{diagbox}
\usepackage{multicol} 
\usepackage{rotating}
\usepackage{pifont}
\usepackage{relsize}
\usepackage[dvipsnames, svgnames]{xcolor}
\usepackage{colortbl}
\usepackage{multirow}
\usepackage{wrapfig}
\usepackage[labelfont=bf]{caption}
\usepackage{tabularx}
\usepackage{utfsym}

\definecolor{Yellow}{rgb}{1.0, 1.0, 0.6}
\definecolor{Orange}{rgb}{1.0, 0.8, 0.6}
\definecolor{Red}{rgb}{1.0, 0.6, 0.6}
\definecolor{brightpink}{rgb}{1.0, 0.0, 0.5}
\definecolor{jade}{rgb}{0.0, 0.66, 0.42}
\definecolor{pastelbrown}{rgb}{0.51, 0.41, 0.33}

\definecolor{brightpink}{rgb}{1.0, 0.0, 0.5}
\definecolor{brightgreen}{rgb}{0.4, 1.0, 0.0}
\definecolor{cadmiumyellow}{rgb}{1.0, 0.96, 0.0}
\definecolor{canaryyellow}{rgb}{1.0, 0.94, 0.0}
\definecolor{caribbeangreen}{rgb}{0.0, 0.8, 0.6}
\definecolor{citrine}{rgb}{0.89, 0.82, 0.04}
\definecolor{chromeyellow}{rgb}{1.0, 0.65, 0.0}
\definecolor{classicrose}{rgb}{0.98, 0.8, 0.91}
\definecolor{cherryblossompink}{rgb}{1.0, 0.72, 0.77}
\definecolor{carnationpink}{rgb}{1.0, 0.65, 0.79}
\definecolor{candypink}{rgb}{0.89, 0.44, 0.48}
\definecolor{applegreen}{rgb}{0.55, 0.71, 0.0}
\definecolor{ao}{rgb}{0.0, 0.5, 0.0}
\definecolor{brightgreen}{rgb}{0.4, 1.0, 0.0}
\definecolor{caribbeangreen}{rgb}{0.0, 0.8, 0.6}
\definecolor{caribbeangreen2}{rgb}{0.4, 0.8, 0.1}

\definecolor{lightgreen}{RGB}{197,224,180}
\definecolor{lightblue}{RGB}{222,235,247}
\definecolor{lightpurple}{RGB}{238,229,241}
\definecolor{lightorg}{RGB}{251,229,214}
\definecolor{lightorange}{rgb}{1.0, 0.8, 0.6}

\definecolor{keypoints}{HTML}{1864AB}
\definecolor{initial}{HTML}{ED7FA6}
\definecolor{latent}{HTML}{AA6CB9}
\definecolor{camera}{HTML}{087F5B}
\definecolor{virtual}{HTML}{6276D7}
\definecolor{render}{HTML}{0B7285}
\definecolor{projection}{HTML}{087F5B}
\definecolor{inversion}{HTML}{D893E9}
\definecolor{generator}{HTML}{A184F6}
\definecolor{openpose}{HTML}{64AEED}
\definecolor{modnet}{HTML}{5BC3D2}
\definecolor{expose}{HTML}{ED7FA6}
\definecolor{camcalib}{HTML}{59CDAA}

\lstdefinestyle{mystyle}{
    commentstyle=\color{codegreen},
    keywordstyle=\color{magenta},
    numberstyle=\tiny\color{codegray},
    stringstyle=\color{codepurple},
    basicstyle=\ttfamily\footnotesize,
    breakatwhitespace=false,         
    breaklines=true,                 
    captionpos=b,                    
    keepspaces=true,                 
    numbers=left,                    
    numbersep=5pt,                  
    showspaces=false,                
    showstringspaces=false,
    showtabs=false,                  
    tabsize=2
}

\newcommand{\first}[1]{\textbf{#1}\cellcolor{Red}}
\newcommand{\second}[1]{#1\cellcolor{Orange}}
\newcommand{\third}[1]{#1\cellcolor{Yellow}}
\newcommand{\up}[1]{#1$\uparrow$}
\newcommand{\down}[1]{#1$\downarrow$}

\usepackage[pagebackref,breaklinks,colorlinks]{hyperref}

\usepackage[capitalize]{cleveref}
\crefname{section}{Sec.}{Secs.}
\Crefname{section}{Section}{Sections}
\Crefname{table}{Table}{Tables}
\crefname{table}{Tab.}{Tabs.}

\def\cvprPaperID{5} 
\def\confName{CVPR}
\def\confYear{2023}

\newcommand{\pose}{\boldsymbol{\theta}}
\newcommand{\shape}{\boldsymbol{\beta}}
\newcommand{\transform}{\mathbf{T}}
\newcommand{\rotation}{\mathbf{R}}
\newcommand{\translation}{\mathbf{t}}
\newcommand{\image}{\mathbf{I}}
\newcommand{\field}{\mathbf{F}}
\newcommand{\boundary}{\mathbf{B}}
\newcommand{\keypoints}{\mathbf{k}}
\newcommand{\joints}{\mathbf{j}}
\newcommand{\barycentric}{\boldsymbol{b}}
\newcommand{\silhouette}{\mathbf{S}}
\newcommand{\rendered}{\hat{\mathbf{S}}}
\newcommand{\render}{\mathcal{R}}
\newcommand{\error}{\mathcal{E}}
\newcommand{\body}{\mathcal{H}}
\newcommand{\vertices}{\mathbf{V}}
\newcommand{\faces}{\mathbf{F}}
\newcommand{\generator}{\mathcal{G}}
\newcommand{\projection}{\boldsymbol{\mathcal{\pi}}}

\newcommand{\KBody}[2]{KBody}

\DeclareMathOperator*{\argmin}{\arg\!\min}

\usepackage{adjustbox}
\usepackage{soul}

\newcommand{\supp}{supplemental material\xspace}
\ifarxiv \renewcommand{\supp}{appendix\xspace} \fi

\newenvironment{packed_item}{
\begin{itemize}
  \setlength{\itemsep}{1pt}
  \setlength{\parskip}{2pt}
  \setlength{\parsep}{0pt}
}{\end{itemize}}

\usepackage[accsupp]{axessibility}  

\begin{document}

\title{\KBody{-.05}{.0225}: Towards general, robust, and aligned monocular whole-body estimation}



\author{Nikolaos Zioulis\textsuperscript{1}, James F. O'Brien\thanks{Corresponding author: \href{mailto:james@getklothed.com}{james@getklothed.com}} \textsuperscript{1,2}\\
\textsuperscript{1} Klothed Technologies Inc., \textsuperscript{2} UC Berkeley\\
\centering{\small\url{https://zokin.github.io/KBody}}
\vspace{-0.7em}
}

\maketitle

\begin{abstract}
KBody is a method for fitting a low-dimensional body model to an image.
It follows a predict-and-optimize approach, relying on data-driven model estimates for the constraints that will be used to solve for the body's parameters.
Acknowledging the importance of high quality correspondences, it leverages ``virtual joints" to improve fitting performance, disentangles the optimization between the pose and shape parameters, and integrates asymmetric distance fields to strike a balance in terms of pose and shape capturing capacity, as well as pixel alignment.
We also show that generative model inversion offers a strong appearance prior that can be used to complete partial human images and used as a building block for generalized and robust monocular body fitting.

\noindent \textup{Author's preprint version. Published in CVPR 2023 1st Workshop on Reconstruction of Human-Object Interactions (RHOBIN).}
Project page: \url{https://zokin.github.io/KBody}.
\end{abstract}

\section{Introduction}
\label{sec:intro}
\input{sections/intro.tex}

\section{Related Work}
\label{sec:related_work}
\input{sections/related.tex}

\section{Approach}
\label{sec:approach}
\input{sections/approach.tex}

\section{Results}
\label{sec:results}
\input{sections/results.tex}

\section{Conclusion}
\label{sec:conclusion}
\input{sections/conclusion.tex}

{\small
\bibliographystyle{ieee_fullname}
\bibliography{refs}
}

\ifarxiv \clearpage \input{appendix} \fi

\end{document}

%% file: sections/intro.tex
Machine perception of humans in images has seen remarkable progress recent years. 
This rapid advance has been the combined result of datasets like MS-COCO \cite{lin2014microsoft}, the evolution of data-driven methods \cite{he2016deep,sun2019deep}, and modern parametric human body representations that are compact and continuously differentiable \cite{loper2015smpl,xu2020ghum}.
Estimating the parameters of a dynamic human body is a cornerstone for human-centric applications such as virtual try-on based e-commerce \cite{lee2022high}, avatar creation for virtual presence \cite{li2022avatarcap}, and performance analysis for virtual coaching \cite{fieraru2021aifit}.
Multi-view configurations offer robust estimations in challenging conditions \cite{cheng2022generalizable}.
This is a result of strongly-constraining the problem, and the same cannot be said for the ill-posed monocular case, which is nonetheless, the foundation of many consumer-facing products.

\input{figures/teaser.tex}

Despite the significant progress, high-quality monocular human body estimation in-the-wild remains elusive due to the challenges arising from the problem formulation itself and the limitations of available constraints.
From an abstracted point of view, estimating the human body from a single image corresponds to estimating the articulation parameters $\pose \in \mathbb{SO}(3)^P$, the shape parameters $\shape \in \mathbb{R}^B$ and the global transformation $\transform = \left[ \begin{smallmatrix} \rotation & \translation\\ \mathbf{0}&1 \end{smallmatrix} \right]$.
These parameters reconstruct the human mesh $( \vertices, \faces ) = \body(\pose, \shape, \transform)$ via the body function $\body$.
Two dominant classes of approach exist.
The first fits the body by minimizing an objective \cite{bogo2016keep,pavlakos2019expressive}:
\begin{equation}
\label{eq:optimization}    
    \argmin_{\pose,\shape,\transform} \mathcal{E}_{data} + \mathcal{E}_{prior},
\end{equation}
that includes a data fitting term, $\mathcal{E}_{data}$, and $\mathcal{E}_{prior}$, an important prior regularization term to prevent degenerate solutions and provide additional constraints to alleviate the ill-posedness of the problem.
The constraints involved in the data term most typically include $2D$ keypoints \cite{pavlakos2019expressive}, that are typically inferred by a data-driven method \cite{openpose}. 
While the prior term helps, $2D$ keypoints usually lead to solutions that suffer from monocular ambiguity, producing poor results from a $3D$ accuracy perspective.
The second class of approach consists of data-driven methods that encode a learned prior in the parameters, $\chi$, of a neural network, $f$, and 
perform monocular inference:
\begin{equation}
\label{eq:datadriven}
    ( \pose, \shape, \transform, \projection ) = f_\chi(\image),
\end{equation}
with $\projection$ being the -- typically weak perspective / orthographic -- projection parameters that best explain the image content using the estimated parameters.
As the neural network function $f_\chi$ is supervised, it preserves $3D$ awareness but usually suffers from predictions with poor pixel alignment, and bias due to the long tailed distribution of data \cite{ren2022balanced}.

Another challenge that also hinders high-quality pixel alignment is the conflict between the pose $\pose$ and shape $\shape$, that are entangled though $\body$.
Early works \cite{pavlakos2019expressive,choutas2020monocular} focused on the difficult problem of pose capturing foremost, with proper shape being an unaccomplished side-objective.
Yet as progress was made, it became evident that inaccurate shape was hindering further advances, with more recent works \cite{corona2022learned,choutas2022accurate} focusing on higher quality shape capture, but seemingly, at the cost of poorer pose estimation.

In-the-wild images introduce additional challenges, some of which are only partly addressed by complex augmentation schemes \cite{wang2021human,weng2022domain}, and others, like missing information in partial human images, which is prevalent in some domains, are challenging to overcome.
For fitting approaches the prior terms are not sufficient to regularize the optimization process when keypoints are missing, while data-driven methods can only extrapolate up to the training data distribution's capacity.
Overall, achieving pixel-aligned estimates that are metrically correct (in world scale, not up to an unknown scale factor), and doing so robustly for a wide range of inputs remains a significant challenge.

In this work, we present a general framework for estimating whole-body human parameters from a single image.
Our goal is to deliver robust estimates, for a variety of inputs, while preserving pixel alignment and proper 3D estimations as much as possible, as well as to capture shape cues and pose information simultaneously, as seen in \cref{fig:teaser}.
More specifically, our contributions are the following:
\begin{packed_item}%
    \item We improve fitting quality by introducing virtual joints, adapted to fit the estimated data, and allowing for smooth interplay with silhouette constraints, expressed as an asymmetric distance field.
    We additionally show how disentangling the optimization process allows for improved joint shape and pose estimates.  
    \item We present an appearance prior based approach to handle images with missing information by completing them in a structurally plausible manner.
    Plausibility is enough for inferring constraints on the hallucinated parts which enable higher quality fits on partial images.
\end{packed_item}%

%% file: figures/teaser.tex
\begin{figure}[!tbp]
\setlength\tabcolsep{1.5pt} %
\begin{tabular}{@{}lcccc@{}}
\small
 & \begin{tabular}[c]{@{}c@{}}SMPLify-X\\ \cite{pavlakos2019expressive}\end{tabular} &
  \begin{tabular}[c]{@{}c@{}}PyMAF-X\\ \cite{pymafx2022}\end{tabular} &
  \begin{tabular}[c]{@{}c@{}}SHAPY\\ \cite{choutas2022accurate}\end{tabular} &
  \begin{tabular}[c]{@{}c@{}}\KBody{-.1}{.035}\\ (Ours) \end{tabular} \\ \midrule
Pose      & \textcolor{caribbeangreen}{\usym{1F5F8}} & \textcolor{caribbeangreen}{\textbf{\usym{1F5F8}\usym{1F5F8}}} & \textcolor{caribbeangreen}{\usym{1F5F8}} & \textcolor{caribbeangreen}{\textbf{\usym{1F5F8}\usym{1F5F8}}} \\
Shape     & \textcolor{caribbeangreen}{\textbf{\usym{1F5F8}}} & \textcolor{red}{\usym{2718}} & \textcolor{caribbeangreen}{\textbf{\usym{1F5F8}\usym{1F5F8}}} & \textcolor{caribbeangreen}{\textbf{\usym{1F5F8}\usym{1F5F8}}} \\
Pixel & \textcolor{caribbeangreen}{\textbf{\usym{1F5F8}}} & \textcolor{caribbeangreen}{\textbf{\usym{1F5F8}}} & \textcolor{red}{\usym{2718}} & \textcolor{caribbeangreen}{\textbf{\usym{1F5F8}\usym{1F5F8}}} \\ \midrule
\normalsize
 \rotatebox{90}{Full body}         & \includegraphics[width=0.22\linewidth]{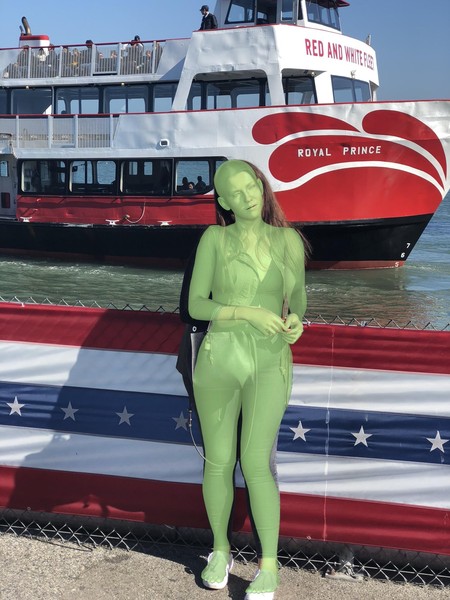} & \includegraphics[width=0.22\linewidth]{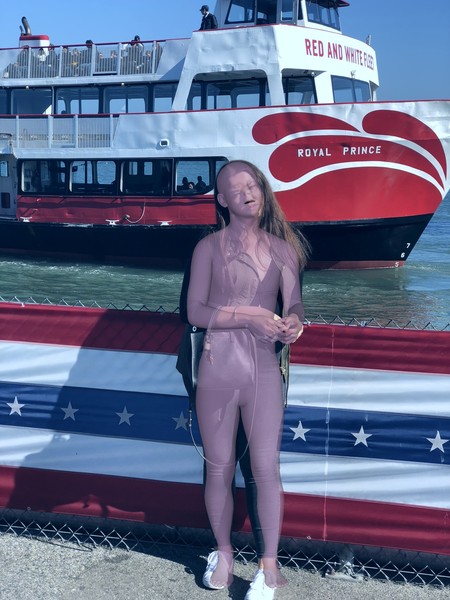}  & \includegraphics[width=0.22\linewidth]{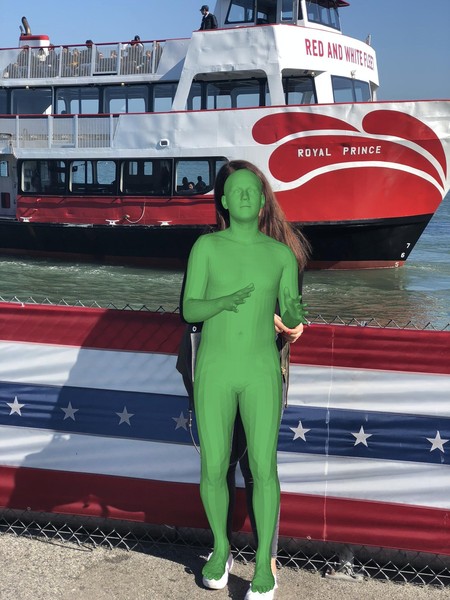} & \includegraphics[width=0.22\linewidth]{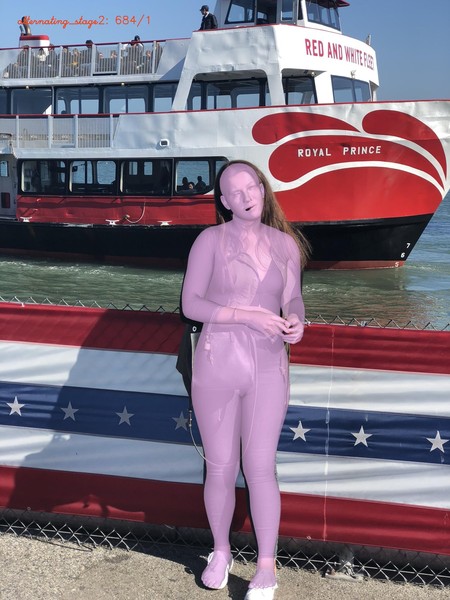} \\
       \rotatebox{90}{Partial body}   & \includegraphics[width=0.22\linewidth]{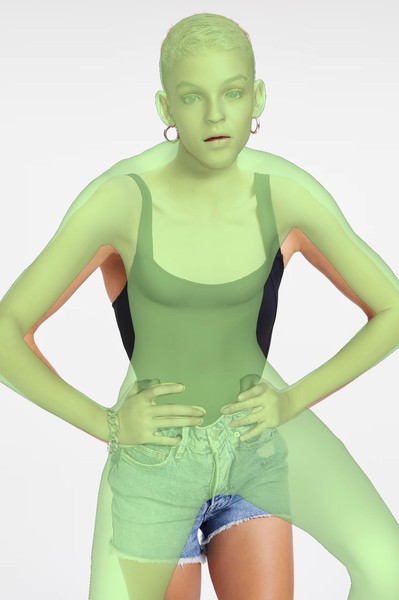} & \includegraphics[width=0.22\linewidth]{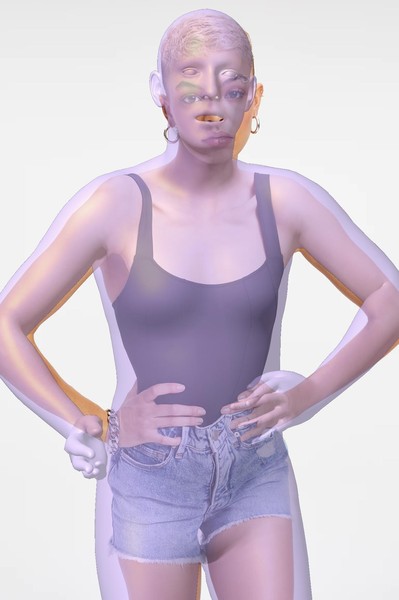}  & \includegraphics[width=0.22\linewidth]{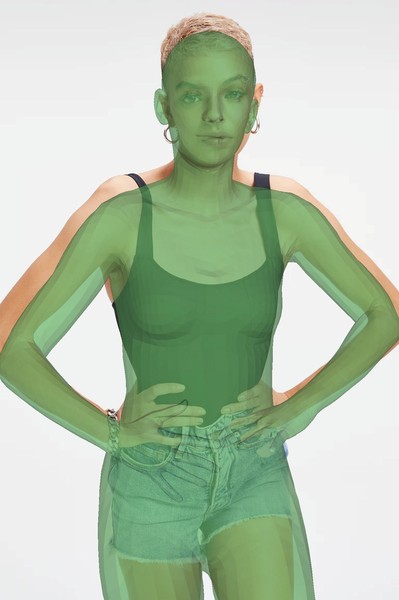} & \includegraphics[width=0.22\linewidth]{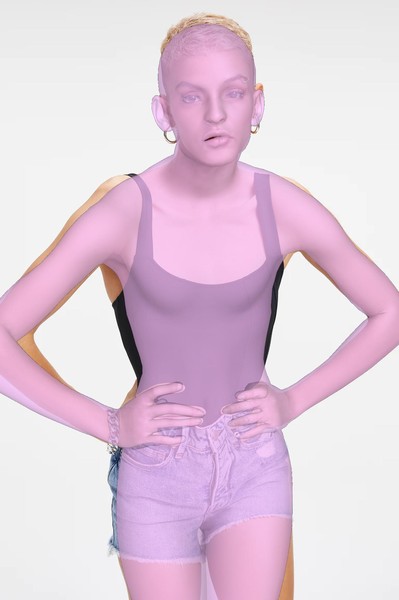} \\
\end{tabular}%
\caption{
Flexible, pixel aligned, accurate body pose and shape capture is the challenging, yet ultimate goal of monocular expressive body fitting.
\KBody{-.1}{.035} is a general approach that improves the balance between all 3 traits using a \textit{predict-and-optimize} approach while also gracefully handling partial images.
}
\label{fig:teaser}
\end{figure}

%% file: sections/related.tex
Estimating parametric human models from images is a rapidly evolving area forming a complex landscape of data, models, and training strategies, as discussed in a recent survey \cite{pang2022benchmarking} and benchmark \cite{tian2022recovering} papers.
Several parametric human body models, including STAR \cite{osman2020star}, GHUM \cite{xu2020ghum,alldieck2021imghum} and most recently SUPR \cite{osman2022supr} have been released, but we will focus on the expressive variant of SMPL, SMPL-X \cite{pavlakos2019expressive}.

\subsection{Single-shot estimation methods}
Pioneering the transition from keypoint estimation to full-body estimation involved the direct regression of low-dimensional body parameters from a single image \cite{kanazawa2018end}.
The method was supervised using keypoint annotations and thus, end-to-end training was achieved after also regressing the camera parameters that would project the articulated body joints to correct positions.
Regularization was applied in the form of a discriminator for the estimated pose and shape, so as to match a realistic distribution made available as a corpus of fit human scans.
Various extensions were later proposed, integrating inverse kinematics \cite{li2021hybrik}, topological priors \cite{mugaludi2021aligning}, and external camera estimation \cite{kocabas2021spec} to improve pose estimation performance.
While the latter two approaches use silhouettes in their training schemes, they remain an intermediate representation for skeletonization \cite{mugaludi2021aligning} or they include clothing layers \cite{kocabas2021spec}.

Initial efforts only regressed pure body parameters (\textit{i.e.}~SMPL), which unfortunately disregards details like hands and faces.
ExPose \cite{choutas2020monocular} included regressing parameters for the hands and face.
FrankMoCap \cite{rong2021frankmocap} built an efficient system, achieving real-time rates.
ExPose was extended to PIXIE \cite{feng2021collaborative}, which had separate experts for the body, hands and face that were optimally combined to improve results.
More recently, PyMAF-X \cite{pymafx2022} builds on the iterative nature of these models (\textit{e.g.}~\cite{kanazawa2018end,choutas2020monocular}) but instead of using global features at a single scale, PyMAF-X uses a pyramid of features, including finer-grained ones, achieving higher quality pixel alignment than other approaches.

Taking another direction, SHAPY \cite{choutas2022accurate} focuses on shape estimation using model agency annotations for shape measurements.
Having been trained with this supervision, it is capable of regressing metric-scale shapes.
SHAPY's pose estimation performance is not at the same level of PyMAF-X, but its capacity to output metric-scale shapes heavily compensates.

\subsection{Iterative optimization methods}
SMPLify \cite{bogo2016keep} was the seminal work that fit the SMPL body to a single image, showing the effectiveness of having priors for both the pose and shape alike.
SMPLify was later extended to use annotated silhouettes in its iterative optimization scheme, with the goal of improving dataset annotations \cite{lassner2017unite}.
Using an L1 silhouette objective allowed for capturing human performances in video \cite{guo2021human} using differentiable soft-rasterization \cite{liu2019soft,ravi2020accelerating}, and improved results when combined with a differentiable ray-tracer \cite{li2018differentiable} and part-based masks \cite{battogtokh2022simple}.
In a follow up work, it was extended to SMPLify-X \cite{pavlakos2019expressive}, adding details like hands and face, as well as a learned prior, VPoser \cite{pavlakos2019expressive}.
Similarly, to improve shape capturing for use within forensic contexts \cite{thakkar2022reliability}, an L2 mask loss was added into the optimization scheme through a differentiable renderer \cite{kato2018neural}.

While orthogonal improvements like better priors (\textit{e.g.}~Pose-NDF \cite{tiwari2022pose}) can improve fitting performance, results ultimately rely on the constraints $\keypoints$ and (optionally) $\silhouette$ \cite{guler2018densepose,ke2022modnet}.
Another important component is the initialization of the optimization which can significantly affect convergence due to the ill-posedness of monocular fit.
One solution \cite{iqbal2021kama} to this uses $3D$ keypoint estimates as constraints, and iterative refines the estimate via forward kinematics.

Finally, a relatively recent and novel direction combines data-driven models and optimization techniques.
HUND \cite{zanfir2021neural} learns a recurrent model that is learned to optimize an recurrent (initial) state and alignment errors
iteratively, which proves to be faster than traditional optimization approaches.
The same applies to LVD \cite{corona2022learned} that learns descent updates for each body vertex so as to predict the depicted human mesh.
Both approaches are limited by their training data compared to other optimization techniques.
Particularly so for LVD, which is trained on 3D human scans, a data category that is hard to acquire at scale.

Last, test-time optimization is a relatively new field that finetunes an entire model on a specific target sample using predicted constraints like keypoints and silhouettes.
Through this technique and the use of separate model and parameter steps, as well as silhouette constraints, a recent work \cite{li2021everybody} has demonstrated improved shape estimation.

%% file: sections/approach.tex
\input{figures/model.tex}

Our overall framework is presented in \cref{fig:model}, and thematically split in two distinct stages, both following a \textit{predict-then-optimize} approach.
First, there is an optional completion preprocessing step on the left for use with partial images of humans. 
Second, is a process for high quality monocular fitting using $2D$ constraints on the right.
While the illustration follows the processing order from left-to-right, we will first present the fitting on \cref{subsec:fitting}, and then the optional completion preprocessing step on \cref{subsec:completion}.

\subsection{Pixel-aligned Shape-aware 3D Body Fitting}
\label{subsec:fitting}
\input{subsections/fitting.tex}

\subsection{Structurally Plausible Human Completion}
\label{subsec:completion}
\input{subsections/completion.tex}

%% file: figures/model.tex
\begin{figure*}
\includegraphics[width=\textwidth]{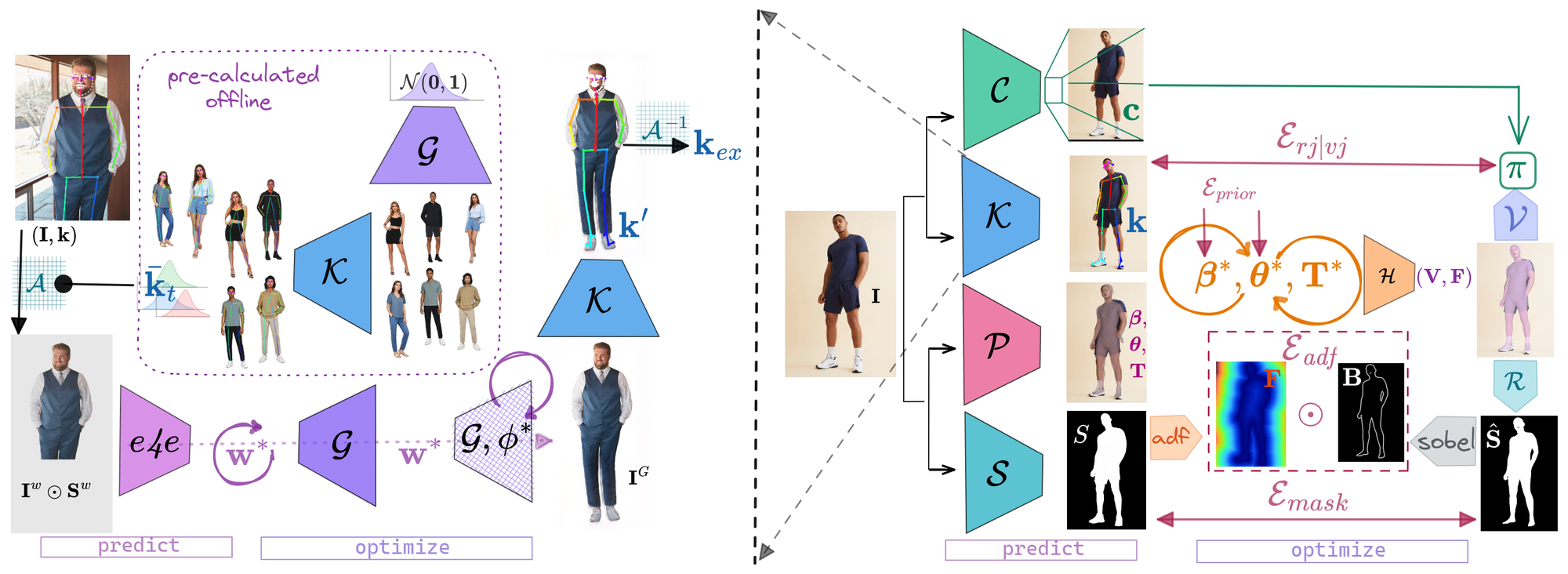}
\caption{
The \KBody{-.1}{.015} framework considers 2 stages, an optional image-based body completion on the left, and a general body fitting on the right.
Keypoints \textcolor{keypoints}{$\keypoints$}, silhouette $\silhouette$ and (optionally) camera \textcolor{camera}{$\mathbf{c}$} constraints are predicted from the respective models \textcolor{openpose}{$\mathcal{K}$}, \textcolor{modnet}{$\mathcal{S}$} and \textcolor{camcalib}{$\mathcal{C}$}.
Then, an initial state \textcolor{initial}{$\shape, \pose, \transform$} predicted by $\mathcal{P}$ is iteratively optimized to fit these constraints using the rendering \textcolor{render}{$\mathcal{R}$}, virtual joint \textcolor{virtual}{$\mathcal{V}$}, and camera-conditioned projection \textcolor{camera}{$\pi$} functions.
When identifying partial keypoints \textcolor{keypoints}{$\keypoints$}, the optional step on the left produces extrapolated keypoints \textcolor{keypoints}{$\keypoints_{ex}$} to improve fits on partial images.
After properly aligning the masked image $\image^w \odot \silhouette^w$ using \textcolor{keypoints}{$\keypoints$} and the distribution \textcolor{keypoints}{$\bar{\keypoints}_t$} expected by the generative model, an initial inversion vector \textcolor{latent}{$\mathbf{w}$}, estimated by a single-shot inversion model \textcolor{inversion}{$e4e$}, is iteratively refined twice, first on the \textcolor{latent}{$\mathcal{W}$} latent space and then on the manifold \textcolor{generator}{$\mathcal{G}_{\phi}$} using the warped masked partial image as constraint.
}
\label{fig:model}
\end{figure*}

%% file: subsections/fitting.tex
Similar to prior approaches, we fit a parametric body model to image-domain constraints by minimizing Eq.~\eqref{eq:optimization}, using the same prior terms as SMPLify-X~\cite{pavlakos2019expressive}, but with a disentangled optimization process (\cref{subsec:disentangled}), while also using virtual joints in the projected keypoints objective (\cref{subsec:virtual_joints}), and adding a silhouette-based objective (\cref{subsec:adt}):

\vspace{-5pt}
\begin{equation}
    \error_{data} = \underbrace{\lambda_{k} ( \error_{rj} + \error_{vj})}_\textrm{keypoints} + \underbrace{\lambda_{m} \error_{mask} + \lambda_{d} \error_{adf}}_\textrm{silhouette},
\end{equation}
where $\error_{rj|vj} = \varrho(\keypoints, \projection(\joints_{rj|vj}))$ is the Geman-McClure penalty function \cite{geman1987statistical} for the regular joints, $\joints_{rj}$, and virtual joints, $\joints_{vj}$, matching them to the corresponding keypoints $\keypoints$ via the projection function $\projection$ of given camera model.
The parameters of the camera can be known (when available in image metadata), estimated (\textit{e.g.}~using a camera parameters estimation model), or fixed (when no information is available).
$\error_{mask} = \sum^{\Omega} || \silhouette - \rendered ||_1$ is an L1 silhouette overlay term defined on the image domain $\Omega$, between an inferred silhouette $\silhouette$ and the body model's silhouette $\rendered = \render(\vertices, \faces)$ rendered via a differentiable rendering function $\render$.
The virtual joints calculation and the asymmetric distance field term, $\error_{adf}$, are described in the following subsections.

\subsubsection{Disentangled Optimization}
\label{subsec:disentangled}
Prior monocular human body fitting works perform a staged optimization of Eq.\eqref{eq:optimization}, where each stage adds a layer of complexity in the optimization (e.g. details like fingers), and also anneals the constraints' weights \cite{bogo2016keep,pavlakos2019expressive} across stages.
Initial estimates of global parameters $\transform$ have also been included as a first stage \cite{bogo2016keep,pavlakos2019expressive}, but sensitivity to localisation of the torso joints has led to alternatives \cite{lassner2017unite}.
To overcome sensitivity to initialization we use a data-driven initial estimate which serves as a good initial starting point.

However, all prior work up to now optimize both $\shape$ and $\pose$ simultaneously at each iteration $i$ of each stage $s$: $(\shape^{s}_{i+1}, \pose^{s}_{i+1}) = (\shape^{s}_{i} + \Delta \shape{s}_{i}, \pose^{s}_{i} + \Delta \pose^{s}_{i})$.
These two sets of parameters are entangled by the human body function $\body$ that allows for their joint optimization.
While this is effective with a $3D$ objective that is conditioned on the same domain where the function $\body$ exists, it is much less effective in the monocular $2D$ case that comes with inherent $3D$ ambiguities.
As a result, optimization is dominated by the pose updates $\Delta \pose$.
This imbalance is evident in both keypoint-only optimization approaches \cite{pavlakos2019expressive} as well as data-driven models trained with only keypoint losses \cite{kolotouros2019learning,choutas2020monocular,pymafx2022}.
Both tend to produce shape coefficients biased towards the zero mean vector.
More recent shape-aware approaches either optimize in $3D$ \cite{corona2022learned} or use $3D$ losses during training \cite{choutas2022accurate}.

Seeking to improve our optimization loop, we separate the parameter updates of the shape $\shape$ and pose $\pose$ components in an alternating fashion for stage $s$: $(\shape^{s}_{i}, \pose^{s}_{i+1}) = (\shape^{s}_{i-1} + \Delta \shape^{s}_{i-1}, \pose^{s}_{i} +  \Delta \pose^{s}_{i})$.
Similar to block coordinate optimization, the shape $\shape$ parameters are only updated in even iterations $i$, while the pose parameters $\pose$ are only updated in odd iterations $i+1$.
This method exhibits significantly better joint optimization of these parameters even in the highly ill-posed monocular case.
However, it is critical that the global pose is close to the minima, meaning that such a disentangled optimization stage can only be introduced later in the optimization process.
Alternatively, one could introduce scaling factors on the parameters and loss function so that $\theta$ and $\beta$ would be well balanced, but it is unclear how the scaling factors would be computed and then updated as the optimization progresses.

\subsubsection{Virtual Joints}
\label{subsec:virtual_joints}
An iterative fitting approach crucially relies on high quality correspondences.
Defining proper joint locations on the body to match the keypoint estimates has troubled past approaches, with the hip joints ignored from the optimization \cite{pavlakos2019expressive}, or regressed via empirically defined and manually created joint regressor functions \cite{kolotouros2019learning}.
However, the location of the keypoints $\keypoints$ are typically inferred from a data-driven model which aggregates numerous annotations and thus, includes their biases as well.
Recent works that acknowledge this have resorted to learning a joint regressor for a specific dataset \cite{hedlin2022simple} which comes with new challenges like properly constraining the joints' locations inside the human body.

Our approach also seeks to identify better matching locations, but not for a specific dataset, instead matching the inference distribution of a pre-trained $2D$ keypoint estimator. 
We introduce the concept of \textit{virtual joints} $\joints_{vj} = \mathcal{V}(\barycentric, \joints_s)$, by parameterizing joint locations as a linear combination of weights $\barycentric$ and pre-defined (empirically or anthropomorphically) joint subsets $\joints_s, s \in [1, \dots, S]$.
More specifically, we focus on the more ambiguous torso joints, which carry a two-fold importance, \textbf{i)} they are high in the kinematic chain, and thus, highly influential of the articulated body fit, and \textbf{ii)} they are highly dependent on human shape, and thus, are necessary to avoid cross data-term conflicts between the keypoint and the silhouette terms.

Virtual joint localisation is restricted to planes formed by joint triangles (\textit{i.e.} $S=3$), illustrated in \cref{fig:virtual_joints}, using a barycentric formulation for the virtual joints.
This allows for the reduction of the number of weights $\barycentric$ to $2$ (or $1$ for joints that need to lie on one of the triangle's altitudes), by exploiting $\sum_{b\in\barycentric} b = 1$.
While this relies on a non-holding rigidity assumption for the joints subset, albeit relaxed in the torso area, the goal is to better localize joints matching those inferred by a $2D$ estimation model, which itself exhibits limited expressivity at the torso. 
Finding the best matching locations is an one-off process that involves fitting a variety of pre-defined poses to inferred keypoints and identifying the best performing weights using a performance indicator.

\input{figures/vj.tex}

\subsubsection{Asymmetric Distance Fields}
\label{subsec:adt}
Human body fitting requires both pose and shape parameters that eventually get mapped to $3D$ or $2D$ joints, the latter being a function of the reconstructed mesh vertices.
Dense representations like silhouettes or distance maps have been used even in earlier parametric model fitting approaches \cite{sminchisescu2002human,balan2007detailed}, and lately in approaches involving both Eq.\eqref{eq:optimization} or Eq.\eqref{eq:datadriven} to offer a less domain sensitive (proxy) representation for synthetic data training \cite{biggs2018creatures,sengupta2021hierarchical,sengupta2021probabilistic}, topological objectives \cite{mugaludi2021aligning}, pose refinement \cite{guo2021human} and better shape (and pose) estimates \cite{lassner2017unite,huang2017towards,thakkar2022reliability}.

Optimization approaches typically rely on differentiable rendering \cite{ravi2020accelerating,kato2018neural} and a per-pixel L1/2 loss between a constraint input silhouette and the body rendered one.
This loss is inefficient, suffering from an irregular loss landscape and the lack of directional information for parameter updates \cite{mugaludi2021aligning}.
The approach is also highly susceptible to body and estimated silhouette inconsistencies, usually derived from  hair, clothing, background mixing or inference uncertainty.

Instead our silhouette term supplements the per-pixel mask alignment objective with a boundary-based distance-field objective by first extracting the boundary $\boundary = G \circledast \rendered$ of the rendered fit body mesh silhouette $\rendered$ using convolutional edge extraction kernel $G$ \cite{sobel}.
The result is used to derive a Chamfer distance objective using a distance field $\field$ via the Hadamard product $\error_{adf} = \sum^{\Omega} \boundary \odot \field$ summed over the pixel domain $\Omega$ which is minimized when the two silhouette boundaries align.
This is a more efficient alternative compared to nearest-neighbor queries \cite{lassner2017unite,huang2017towards} but lacks the symmetric component of Chamfer distance, \textit{i.e.} the distance from the inferred silhouette $\silhouette$ to the fit rendered one $\rendered$, which is nonetheless noisier \cite{lassner2017unite,huang2017towards,alldieck2017optical}.

Indeed, for generalized fitting where hair, clothing and silhouette estimation artifacts come into play, the silhouette-based term tends to become noisy and hinder optimization or produce unrealistic shape estimates.
To overcome this, we calculate an asymmetric distance field (ADF) defined over the entire image domain $\Omega$:
\begin{equation}
\label{eq:adf}
    \field = \lambda_{o} D(\silhouette) \odot \bar{\silhouette} + \lambda_{i} D(\bar{\silhouette}) \odot \silhouette,
\end{equation}
with $D(\cdot)$ being the distance field function and $\bar{\silhouette}$ denotes pixel-wise binary inversion.
While \cite{alldieck2017optical} downscales the noisier symmetric Chamfer objective, we completely disregard it and provide explicit control over pushing the body inwards or outwards with respect to the silhouette by respectively controlling the outer and inner distance field scaling factors $\lambda_{o}$ and $\lambda_{i}$.
For blendshape models such as \mbox{SMPL(-X)}, downweighting the inner field and/or upweighting the outer one, heavily restricts the body shape inside the silhouette while still allowing for greater freedom in not exactly matching the boundary in its entirety.

%% file: figures/vj.tex
\begin{figure*}
\begin{adjustbox}{minipage=\textwidth,bgcolor=darkgray}

\includegraphics[width=0.28\textwidth]{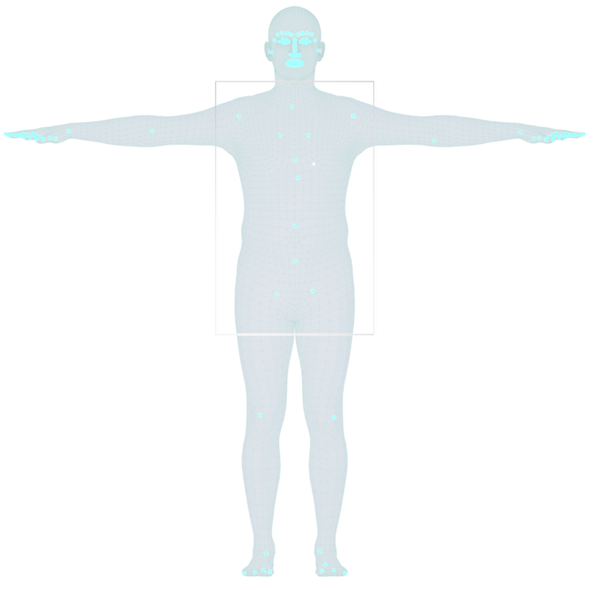}
\includegraphics[width=0.17\textwidth]{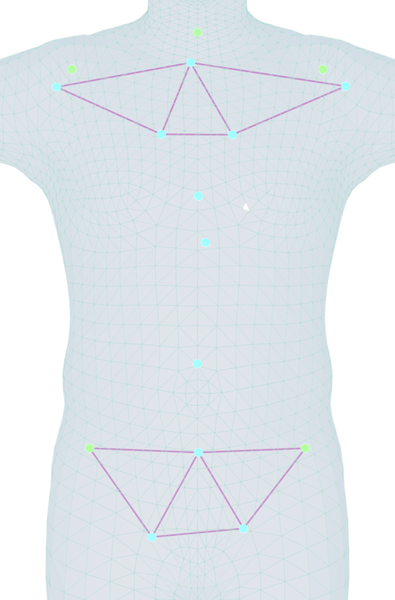}
\includegraphics[width=0.17\textwidth]{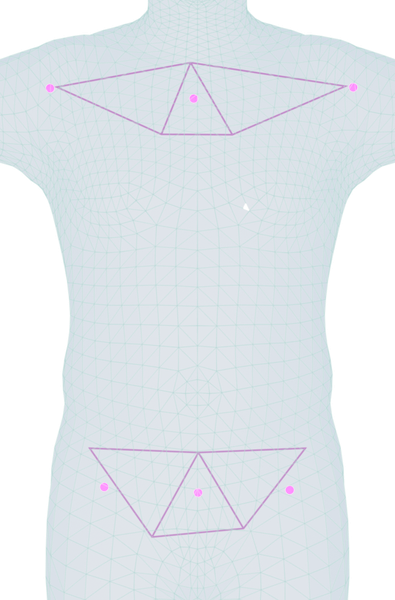}
\includegraphics[width=0.17\textwidth]{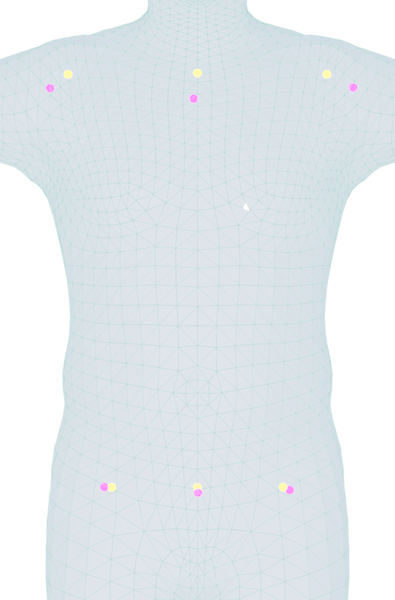}
\includegraphics[width=0.17\textwidth]{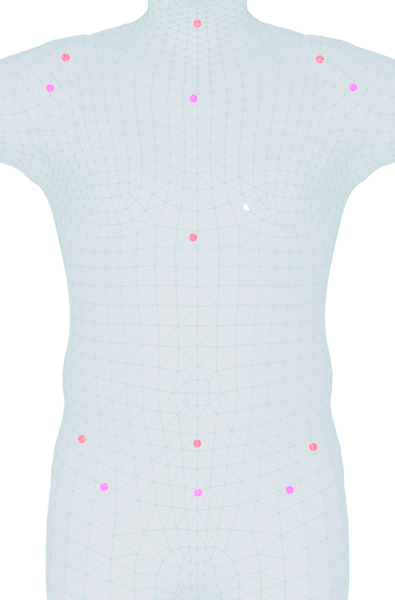}

 \end{adjustbox}
\caption{From left to right: \textbf{i)} the SMPL-X body \textcolor{cyan}{surface and joints}, \textbf{ii)} the inset torso with the barycentric parameterization comprising the \textcolor{violet}{triangles} formed by \textcolor{cyan}{raw} and \textcolor{green}{manually picked} \cite{kolotouros2019learning} joints, \textbf{iii)} our best-estimated \textcolor{magenta}{virtual joints}, and their comparison with \textbf{iv)} \textcolor{orange}{manually picked openpose} joints \cite{bhatnagar2020combining,bhatnagar2020loopreg} and \textbf{v)} the \textcolor{red}{learned regressor} joints fit to Human3.6M \cite{hedlin2022simple}.
As illustrated, the virtual joints can extrapolate to exterior triangle locations by using negative barycentric weights.}
\label{fig:virtual_joints}
\end{figure*}

%% file: subsections/completion.tex
Generalized human body estimation is greatly challenged by partial human images, be it either data-driven estimates or optimization-based approaches, as the partial context and lack of annotated data reduce the prediction accuracy of single-shot estimates and the quality of the constraints for iterative fitting.
We integrate partial human image completion in our fitting framework as an optional step which can be easily identified post keypoint estimation.
The goal is to complement the partial image inferred keypoints $\keypoints_{in}$ with high confidence estimates for the invisible keypoints $\keypoints_{ex} = \keypoints_{c} \setminus \keypoints_{in}$, $\keypoints_{c}$ being the inferred keypoints on the completed image.
This gracefully benefits the fitting process as the projection function $\projection$ is not confined in the original image domain $\Omega$, and can optimize for the combined set $\keypoints = \keypoints_{in} \uplus \keypoints_{ex}$.

While image inpainting could be a proper technique, we find that its generalized nature hurts the structural plausibility and quality of the results, mainly due to the extended nature of partial human image completions.
Human specific solutions either only focus on visible body part completion \cite{zhao2021prior} or rely on intermediate models producing high level extrapolated completions \cite{wu2019deep} as secondary inputs.
Instead, we invert the partial images to the latent space of a generative appearance prior \cite{Karras_2020_CVPR} learned using clothed humans \cite{fu2022stylegan}, carrying more appropriate human structural and shape-aware semantics, and producing high quality extended completions.

\subsubsection{Partial image alignment}
\label{subsec:alignment}
The StyleGAN variants \cite{karras2019style,Karras_2020_CVPR,karras2021alias} are a pure appearance-based family of generative models.
They have been shown to be highly sensitive to affine transformations \cite{abdal2019image2stylegan} even when considering faces whose articulation, and thus spatial variance, is limited compared to full body images.
Therefore, for all StyleGAN face inversion techniques, centralization is a necessary preprocessing step.
To overcome this challenge, which is pronounced when considering full bodies, StyleGAN-Human \cite{fu2022stylegan} investigated various alignment techniques and showed that mid-body alignment behaved better.
In our case we seek to align partial images, making mid-body alignment not an option, and necessitating the design of a different alignment strategy.

Given that most partial images are either missing heads or the bottom half, and typically include some torso joints, we seek to identify the likely positions of these joints on generated samples.
We generate $M=20000$ samples from a normal distribution $\mathcal{N}(\mathbf{0}, \mathbf{1})$ using a truncation factor \cite{brock2018large,kingma2018glow} $\psi = 0.5$ to enforce a proper distribution, and run the keypoint detector on each sample.
This derives the expected (\textit{i.e.}~mean) location of these joints $\bar{\keypoints}_{t}$.
We can then align the corresponding torso joints of the partial image with these expected locations by calculating an affine transformation $\mathbf{A} = \alpha(\bar{\keypoints}_{t}, \keypoints_{t})$, comprising a translation and scale.
This in turn drives an image warping operation $\mathbf{I}^{w} = \mathcal{A}(\image \odot \silhouette, \alpha)$ that aligns the masked image to the StyleGAN-Human resolution $(w,h)=(512,1024)$ and the partial image content to the model's expectation from an appearance reconstruction perspective.

\subsubsection{Generative appearance-based completion}
\label{subsec:inversion}
Using a pre-trained StyleGAN-Human model \cite{fu2022stylegan} our goal is to invert an aligned partial image $\image^{w}$ into its intermediate latent space $\mathbf{w} \in \mathcal{W}$ that reconstructs a full, completed image matching the partial image's appearance.
We initialize $\mathbf{w}$ with a single-shot estimate from a pre-trained inversion model \cite{tov2021designing} (\textit{e4e}).
As the input image is partial, this only serves as a rough initialization into $\mathcal{W}$, which we further refine following \cite{Karras_2020_CVPR} by minimizing:
\begin{equation}
    \argmin_{\mathbf{w}} \lambda_{L1} \mathcal{E}_{L1} + \lambda_{VGG} \mathcal{E}_{VGG} + \lambda_{reg} \mathcal{E}_{reg},
\end{equation}
where $\mathcal{E}_{L1}$ is a $L1$ reconstruction loss, $\mathcal{E}_{VGG}$ is a VGG \cite{simonyan2014very} based perceptual feature loss, and $\mathcal{E}_{reg}$ is a noise regularization term.
These terms are described in \cite{Karras_2020_CVPR} but in our implementation we adapt the pixel-based losses to focus only the original content of the aligned partial image.
This is achieved by using the warped silhouette image $\silhouette^{w}$ to mask the generated ($\image^{G} \odot \silhouette^{w}$) and target images ($\image^{w} \odot \silhouette^{w}$) prior to error term calculation.

Nonetheless this process is not sufficient to plausibly complete the image but is rather a way to quickly initialize the latent space.
Plausible completion is achieved by relying on generator fine-tuning with latent space regularization \cite{roich2022pivotal}.
Essentially, this process is a test-time optimization of the StyleGAN generator manifold around the latent space point $\mathbf{w}^{*}$.
We optimize the generator $\generator$ parameters $\phi$:
\begin{equation}
    \argmin_{\phi} \lambda_{L2} \mathcal{E}_{L2} + \lambda_{LPIPS} \mathcal{E}_{LPIPS} + \lambda_{R} \mathcal{E}_{R} + \lambda_{D} \mathcal{E}_{D},
\end{equation}
where we complement the LPIPS \cite{zhang2018unreasonable} and space regularization ($\mathcal{E}_{R}$) terms used in \cite{roich2022pivotal} with an $L2$ reconstruction term to accelerate convergence and a discriminator loss ($\mathcal{E}_{D}$) to improve the global coherence of our results.
Notably, the reconstruction and perceptual image domain losses are calculated only at the valid warped image $\image^{w}$ domain, denoted by the mask $\mathbf{M}^{c}$, including the partial image's background which we found to significantly improve convergence.
Thus, only the areas to be completed are masked to not participate in error and gradient calculations.
The high quality inversion capacity of the pivotal tuning technique ensures photometric convergence on the valid image regions.
While image editing is not our goal, we find that the space regularization and discriminator terms help in producing structurally and photometrically plausible extended completions, acting as global regularizers.

%% file: sections/results.tex
We refer to the approach depicted in \cref{fig:model} as \KBody{}{} and implement it using SMPL-X \cite{pavlakos2019expressive} as the body model $\body$, OpenPose \cite{openpose} as the 2D keypoint $\keypoints$ estimator model $\mathcal{K}$, MODNet \cite{ke2022modnet} as the silhouette $\silhouette$ estimator $\mathcal{S}$ after thresholding the estimated matte at $0.85$, ExPose \cite{choutas2020monocular} as the initial body parameters $(\shape, \pose, \transform)$ predictor $\mathcal{P}$, and the CamCalib model $\mathcal{C}$, presented in SPEC \cite{kocabas2021spec}, as the camera parameter $\mathbf{c}$ estimator.
For the subsequent optimization we rely on the limited memory BFGS optimizer \cite{wright1999numerical} with strong Wolfe line search and a budget of $30$ iterations.
Similar to prior work we perform annealed optimization with the early stages using stronger regularization to make the objective function more convex, and then progressively reduce the regularization term weights and increase the data terms of the details (hands, face).
For the differentiable mesh rendering we use a high-performance rasterization based implementation \cite{laine2020modular}.
For the pose prior and regularization terms we use the same as SMPLify-X \cite{pavlakos2019expressive}, but relax the latter's weights as the initialization and silhouette constraints provide extra prior knowledge about the pose and shape.

First we validate the effectiveness of the virtual joint localization by running only two stages of fitting to $\mathcal{K}$ after initializing with $\mathcal{P}$, with only the second stage optimizing the details, and without involving $\mathcal{S}$ or $\mathcal{C}$ for a fair comparison with other works.
We use the EHF \cite{pavlakos2019expressive} dataset as well as a manually collected set of plain background human photos whose foreground masks are estimated in high-quality using a background removal service \cite{removebg}.
We run a hierarchical and empirically defined search to identify the parameters $\barycentric$ by fitting to the keypoints estimated by $\mathcal{K}$.
Performance is measured using an indicator combining the keypoints' RMSE and the IoU using the service generated masks, defined as $(1 - IoU) \times RMSE$.
After estimating the barycentric coordinates $\barycentric$ resulting in the best fits, we conduct an experiment on EHF that is presented in \cref{tab:vj}.
Performance is assessed via procrustes-aligned vertex-to-vertex error on the SMPL-X body's vertices (PA-V2V-X) \cite{pavlakos2019expressive}.
As also shown in Pose-NDF \cite{tiwari2022pose} and the first 3 rows of \cref{tab:vj}, simply optimizing the initial estimates of a data-driven model does not necessarily lead to improved fits. 
Using better priors like GAN-S~\cite{davydov2022adversarial} and Pose-NDF~\cite{tiwari2022pose} slightly improves results over the baseline single-shot model ExPose \cite{choutas2020monocular}, while a manually selected joint regressor \cite{bhatnagar2020combining,bhatnagar2020loopreg} (\cref{fig:virtual_joints} \textbf{iv}) does not result in improved fits.
The virtual joints produce the most significant gain, showcasing the importance of higher quality correspondences between the estimated keypoints used as constraints and body's joints.
It should be noted that, apart from the last 2 rows, bad joint-to-keypoint correspondences (\textit{e.g.}~hips) are ignored during optimization.

\input{tables/vj}

Next we evaluate our approach when adding the silhouette constraints using $\mathcal{S}$ and the disentangled optimization for improved shape estimation, adding one such stage and then a final stage for detail (hands,face) capture.
We perform two experiments, first using EHF to assess performance for pose capture as a single subject is used, and second, using SSP3D \cite{sengupta2020synthetic} that includes higher shape variance.
For both experiments we employ pixel-based IoU and use the PA-V2V metric for general body pose estimation and the PVE-T-SC metric \cite{sengupta2020synthetic} for shape estimation.
Likewise we use the SMPL meshes instead of SMPL-X to reduce the effect of the densely sampled head, after converting SMPL-X fits to SMPL meshes using pre-calculated mesh-to-mesh vertex transfer maps \cite{osman2020star}.
Comparisons against optimization \cite{pavlakos2019expressive,corona2022learned} and single-shot \cite{choutas2020monocular,pymafx2022,choutas2022accurate} based approaches are given, with some of the latter focusing on shape \cite{choutas2022accurate}, and others on expressive pose \cite{pymafx2022,choutas2020monocular}.
\cref{tab:both} (left) shows that our \textit{predict-then-optimize} approach outperforms the other methods with respect to pose estimates. 
PyMAF-X is a robust pose estimator when considering an average shaped subject, while LVD suffers due to its limited training data. (accordingly, LVD is omitted from the remainder of the experiments).
On the contrary, on SSP3D the shape-aware SHAPY method offers better performance than PyMAF-X as presented in \cref{tab:both} (right).
Still, our approach produces the best results in terms of pixel alignment and shape estimation, while also showing the benefit of disentangled optimization on shape capturing performance.

\input{tables/ehf_spp3d}

We also present results on the validation set of the HBW dataset \cite{choutas2022accurate} that uses pre-scanned shapes and an assortment of in-the-wild images of the same persons to assess body shape estimation.
However, up to now all results were presented using an arbitary camera, in line with prior work for a fair comparison.
As shown in \cref{tab:hbw} SHAPY outperforms all methods but this is reasonable as it was trained with metric scale supervision, whereas all other approaches were not.
Still, our approach compares favorably to the remaining methods while offering higher quality pixel alignment than all alternatives.
We also ablate the effect of estimating the camera's parameters through $\mathcal{C}$, which naturally improves metric-scale performance.

\input{tables/hbw.tex}

\KBody{-.05}{.035}'s efficacy is qualitatively illustrated in \cref{fig:teaser} using images collected online.
For these representative examples, \KBody{-.05}{.035} provides more balanced solutions, capturing pose and shape in high-quality for both heavy and lighter subjects, while also achieving good pixel alignment.
With respect to partial images, we provide a qualitative evaluation in \cref{fig:partial} that shows how our inversion-based completion can handle missing head and/or lower body information.
Moreover, \cref{fig:adf} presents an ablation of the ADF objective and its benefits to clothed estimation.

Finally, an extended set of $112$ full, $78$ partial, and $32$ ADF ablations of randomly selected in-the-wild examples can be found in our \supp and \href{https://zokin.github.io/KBody}{project page}.

\input{figures/full.tex} %
\input{figures/partial.tex} %
\input{figures/adf.tex} %

%% file: tables/vj.tex
\begin{table}[!htbp]
\centering
\resizebox{\columnwidth}{!}{%
\begin{tabular}{@{}lllll@{}}
\toprule
Initialization               & Optimization                                                           & Joints                                                            & \multicolumn{1}{l}{Prior}                                                                                                                                          & \down{PA-V2V-X} \\ \midrule
\textcolor{red}{\usym{2718}} & SMPLify-X~\cite{pavlakos2019expressive} & $\joints_{rj}$                                                    & VPoser~\cite{pavlakos2019expressive}                                                                                             & 60.3 $mm$         \\
ExPose~\cite{choutas2020monocular}  & \textcolor{red}{\usym{2718}}                                           & $\joints_{rj}$                                                    & \textcolor{red}{\usym{2718}}                                                                                                                                       & 54.8 $mm$          \\
ExPose~\cite{choutas2020monocular}  & SMPLify-X~\cite{pavlakos2019expressive} & $\joints_{rj}$                                                    & VPoser~\cite{pavlakos2019expressive}                                                                                             & 67.2 $mm$          \\
\textcolor{red}{\usym{2718}} & SMPLify-X~\cite{pavlakos2019expressive} & $\joints_{rj}$                                                    & \multicolumn{1}{l}{PoseNDF~\cite{tiwari2022pose}} & 57.4 $mm$          \\
ExPose~\cite{choutas2020monocular}  & SMPLify-X~\cite{pavlakos2019expressive} & $\joints_{rj}$                                                    & \multicolumn{1}{l}{PoseNDF~\cite{tiwari2022pose}} & \second{53.8} $mm$ \\
ExPose~\cite{choutas2020monocular}  & SMPLify-X~\cite{pavlakos2019expressive} & $\joints_{rj}$                                                    & \multicolumn{1}{l}{GAN-S~\cite{davydov2022adversarial}}                                 & \third{54.1} $mm$  \\
ExPose~\cite{choutas2020monocular}  & SMPLify-X~\cite{pavlakos2019expressive} & $\joints_{op}$ \cite{bhatnagar2020combining,bhatnagar2020loopreg} & VPoser~\cite{pavlakos2019expressive}                                                                                             & 57.5 $mm$          \\
ExPose~\cite{choutas2020monocular}  & SMPLify-X~\cite{pavlakos2019expressive} & $\joints_{rj|\mathbf{vj}}$                                             & VPoser~\cite{pavlakos2019expressive}                                                                                             & \first{49.3} $mm$  \\ \bottomrule
\end{tabular}%
}
\caption{
Virtual joints improvement analysis on EHF \cite{pavlakos2019expressive}.
The columns indicate parameter initialization and optimization, which joints are optimized, and with which pose prior.
}
\label{tab:vj}
\end{table}

%% file: tables/ehf_spp3d.tex
\begin{table}[!htbp]
\resizebox{\columnwidth}{!}{%
\centering
\begin{tabular}{@{}lcc|ccc@{}}
\toprule
                          
 & \multicolumn{2}{c|}{EHF \cite{pavlakos2019expressive}} & \multicolumn{2}{c}{SSP3D \cite{sengupta2020synthetic}} \\
\multicolumn{1}{l}{Method} & \down{PA-V2V} & \up{IoU} & \down{PVE-T-SC} & \up{IoU}  \\ \midrule
ExPose \cite{choutas2020monocular}                  & 71.7 $mm$               & \third{84.72}\% & 33.0 $mm$                 & 71.00\%          \\
LVD \cite{corona2022learned}            & 131.7 $mm$              & -  & - & - \\
SMPLify-X \cite{pavlakos2019expressive}             & 95.9 $mm$               & 81.46\% & 33.9 $mm$                 & \third{76.60}\%  \\
PyMAF-X \cite{pymafx2022}                           & \second{66.6} $mm$      & \second{85.57}\% & 30.6 $mm$                 & 75.87\%          \\
SHAPY \cite{choutas2022accurate}                    & \third{71.1} $mm$       & 81.29\% & \third{29.3} $mm$         & 72.65\%          \\
\KBody{-.05}{.035} (w/o $\mathcal{C}$ \& \S \ref{subsec:disentangled}) & - & - & \second{28.1} $mm$        & \second{77.87}\% \\
\KBody{-.05}{.035} (w/o $\mathcal{C}$)                                  & \first{64.2} $mm$       & \first{87.72}\% & \first{25.6} $mm$         & \first{80.35}\%  \\ \bottomrule
\end{tabular}
}
\caption{Results on the the EHF \cite{pavlakos2019expressive} \& SSP3D \cite{sengupta2020synthetic} datasets.}
\label{tab:both}
\end{table}

%% file: tables/hbw.tex
\begin{table}[]
\centering
\resizebox{\columnwidth}{!}{%
\begin{tabular}{@{}lcccccc@{}}
\toprule
\multicolumn{1}{c}{Method}                  & \down{Height} & \down{Chest} & \down{Waist} & \down{Hips} & \down{P2P\textsubscript{20k}} & \up{IoU (\%)} \\ \midrule
ExPose \cite{choutas2020monocular}          & \second{75}            & 91           & 93           & 91          & 36                            & 80.50         \\
SMPLify-X \cite{pavlakos2019expressive}     & 121           & 133          & 150          & \third{62}          & 41                            & \third{83.68}         \\
PyMAF-X \cite{pymafx2022}                   & 100           & \third{74}           & \third{90}           & 64          & 34                            & 80.36         \\
SHAPY \cite{choutas2022accurate}            & \first{62}            & \first{58}           & \first{83}           & \second{63}          & \first{24}                            & 77.40         \\
\KBody{-.05}{.035} (w/o $\mathcal{C}$) &   79            &      81        &         96     &      70       &     \third{32}                          &      \second{84.40}         \\
\KBody{-.05}{.035}                          &     \third{78}          &    \second{70}          &          \second{88}    &     \first{61}        &               \second{30}                &      \first{85.19}         \\ \bottomrule
\end{tabular}%
}
\caption{Quantitative results on the HBW (val) \cite{choutas2022accurate} dataset.}
\label{tab:hbw}
\end{table}

%% file: figures/full.tex
\begin{figure}[!htbp]
\captionsetup[subfigure]{position=bottom,labelformat=empty}

\centering

\subfloat{\includegraphics[width=0.24\linewidth]{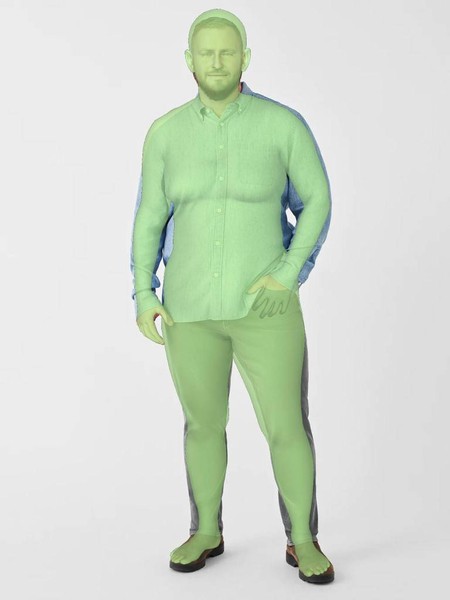}}
\subfloat{\includegraphics[width=0.24\linewidth]{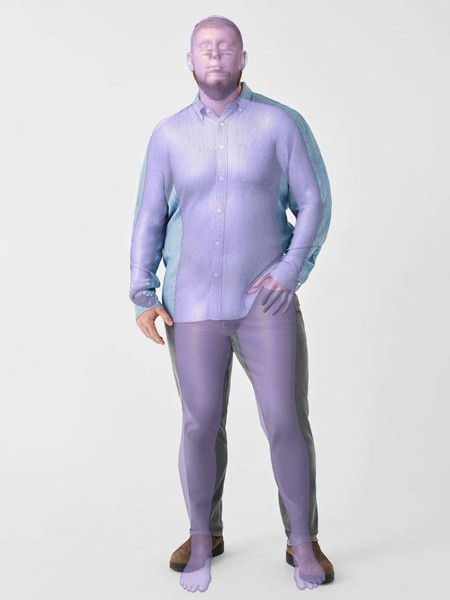}}
\subfloat{\includegraphics[width=0.24\linewidth]{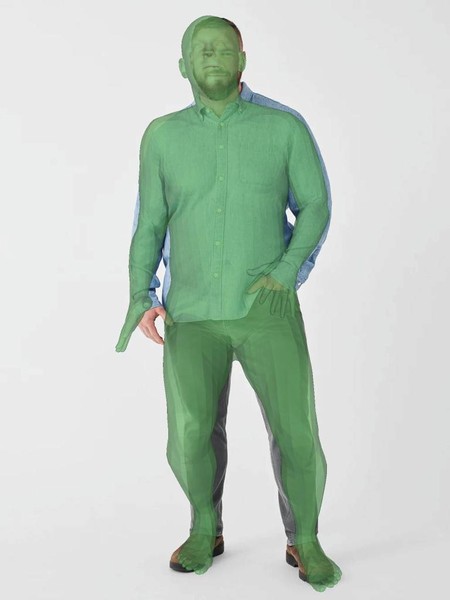}}
\subfloat{\includegraphics[width=0.24\linewidth]{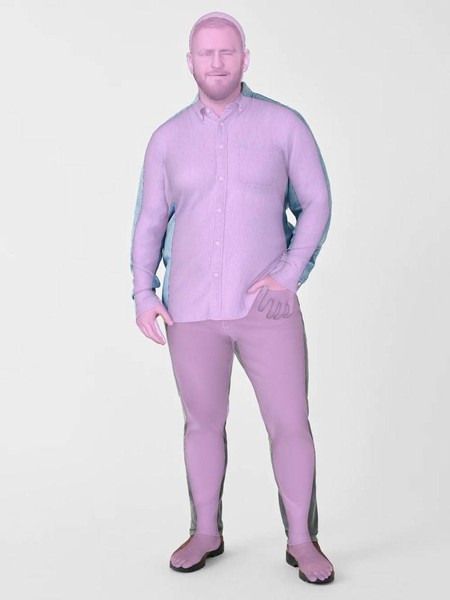}}

\subfloat{\includegraphics[width=0.24\linewidth]{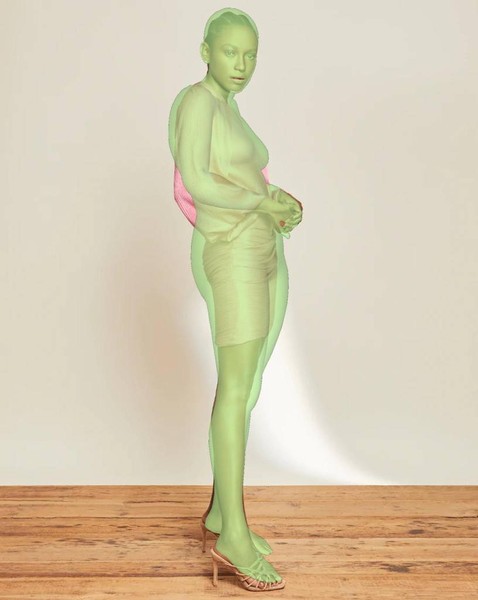}}
\subfloat{\includegraphics[width=0.24\linewidth]{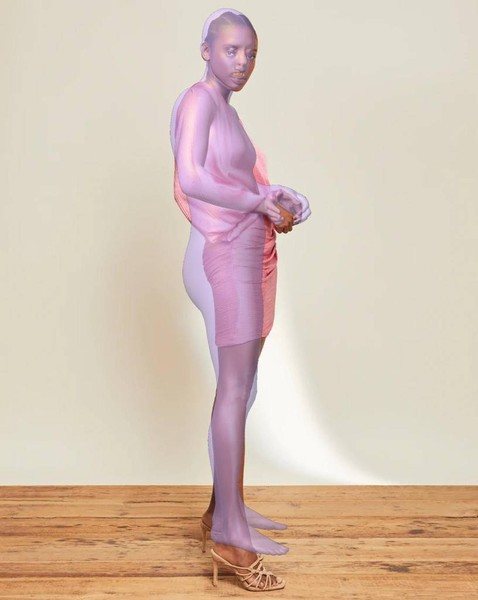}}
\subfloat{\includegraphics[width=0.24\linewidth]{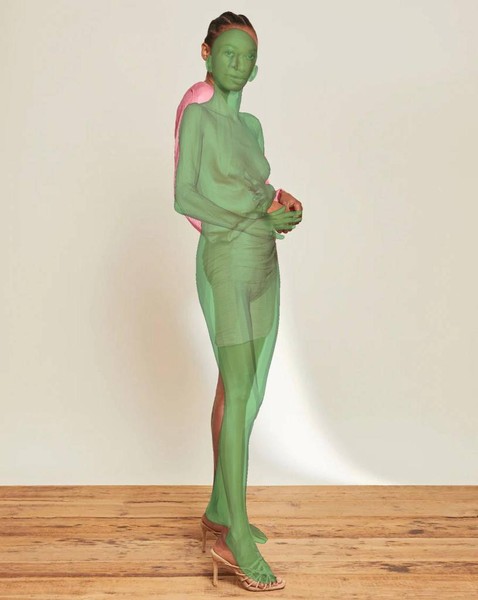}}
\subfloat{\includegraphics[width=0.24\linewidth]{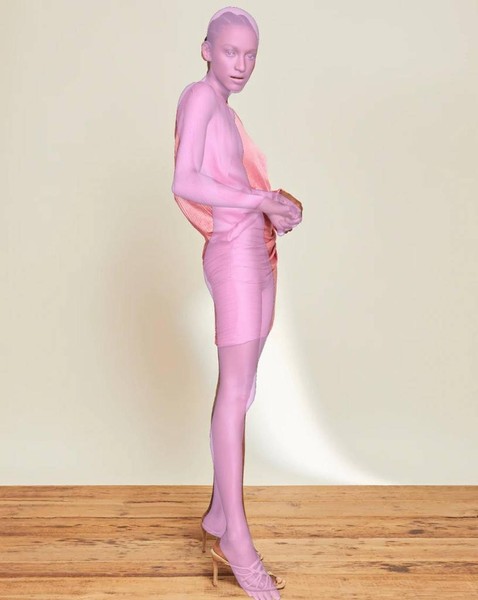}}

\subfloat{\includegraphics[width=0.24\linewidth]{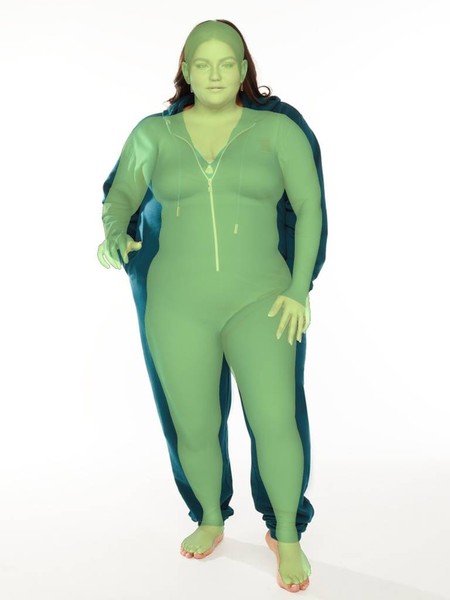}}
\subfloat{\includegraphics[width=0.24\linewidth]{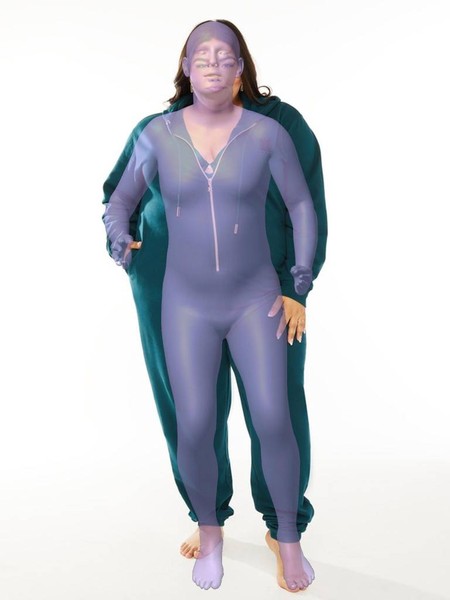}}
\subfloat{\includegraphics[width=0.24\linewidth]{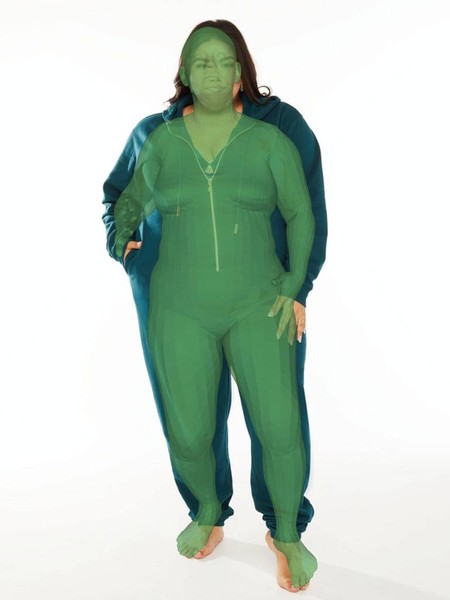}}
\subfloat{\includegraphics[width=0.24\linewidth]{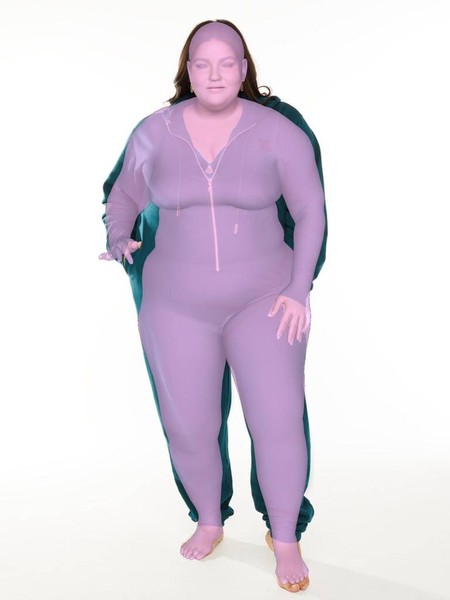}}

\subfloat[SMPLify-X \cite{pavlakos2019expressive}]{\includegraphics[width=0.24\linewidth]{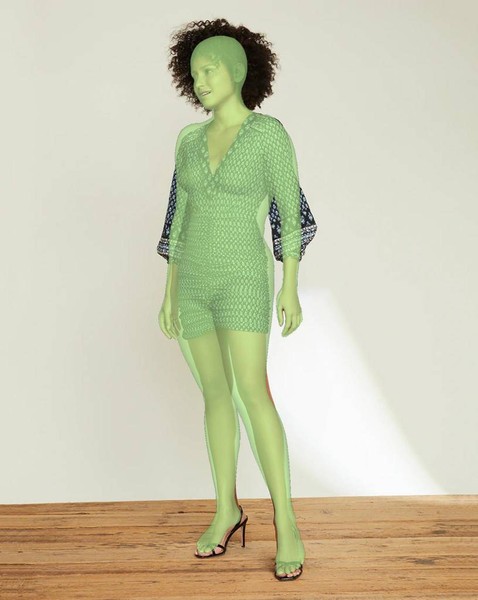}}
\subfloat[PyMAF-X \cite{pymafx2022}]{\includegraphics[width=0.24\linewidth]{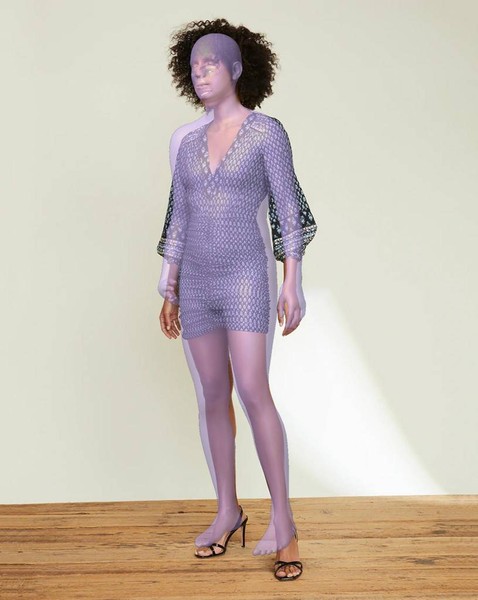}}
\subfloat[SHAPY \cite{choutas2022accurate}]{\includegraphics[width=0.24\linewidth]{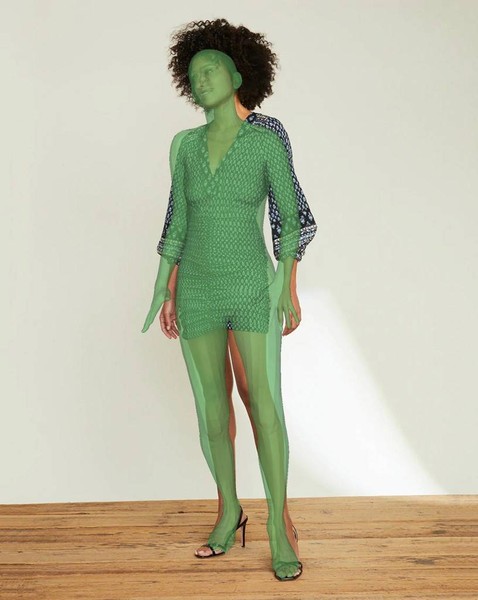}}
\subfloat[\KBody{-.1}{.035}]{\includegraphics[width=0.24\linewidth]{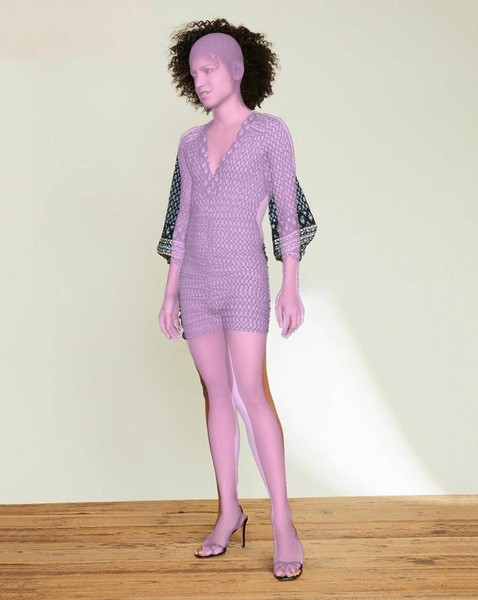}}

\caption{
Left-to-right: SMPLify-X \cite{pavlakos2019expressive} (\textcolor{caribbeangreen2}{light green}), PyMAF-X \cite{pymafx2022} (\textcolor{violet}{purple}), SHAPY \cite{choutas2022accurate} (\textcolor{jade}{green}) and KBody (\textcolor{candypink}{pink}).
}
\label{fig:full}
\end{figure}

%% file: figures/partial.tex
\begin{figure}[!htbp]
\captionsetup[subfigure]{position=bottom,labelformat=empty}

\centering

\subfloat[]{\includegraphics[width=0.24\linewidth]{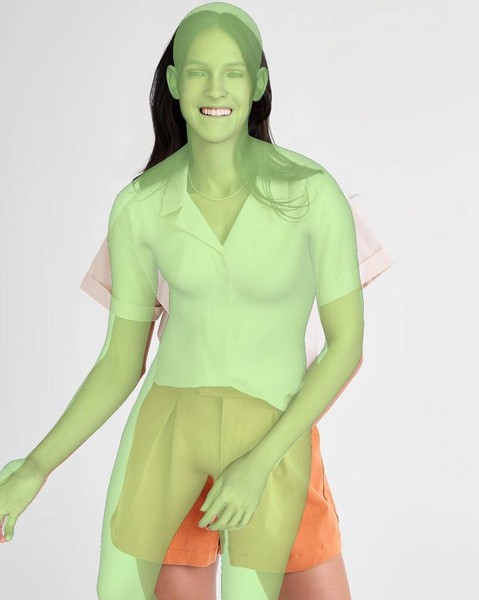}}
\subfloat[]{\includegraphics[width=0.24\linewidth]{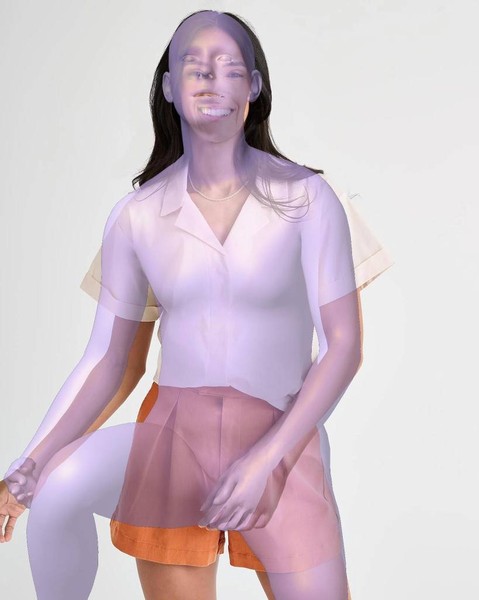}}
\subfloat[]{\includegraphics[width=0.24\linewidth]{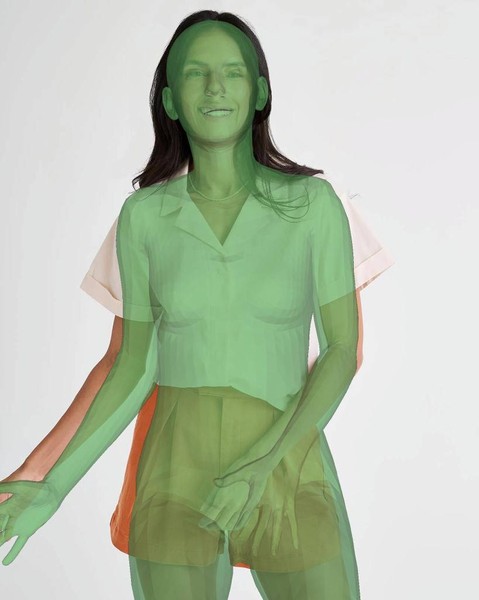}}
\subfloat[]{\includegraphics[width=0.24\linewidth]{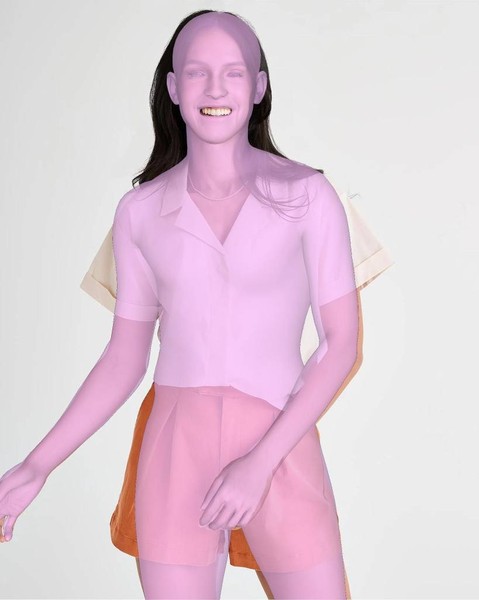}}\\

\vspace{-10pt}

\subfloat[]{\includegraphics[width=0.24\linewidth]{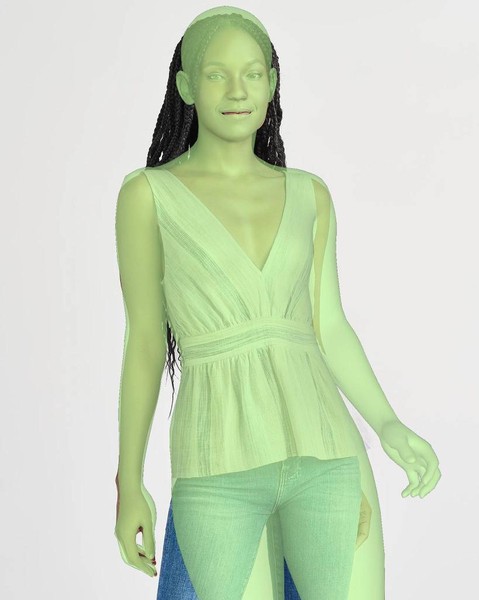}}
\subfloat[]{\includegraphics[width=0.24\linewidth]{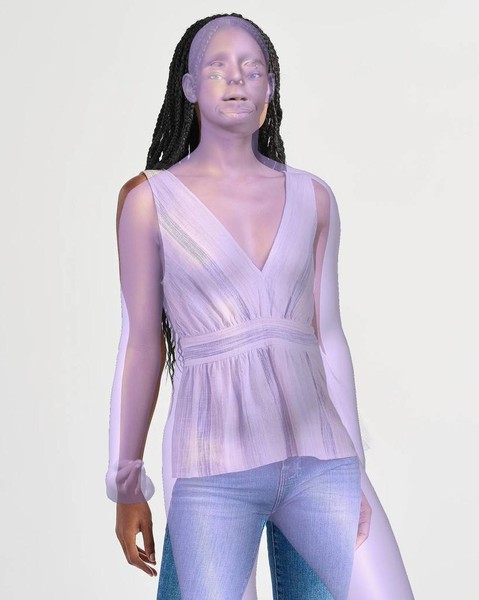}}
\subfloat[]{\includegraphics[width=0.24\linewidth]{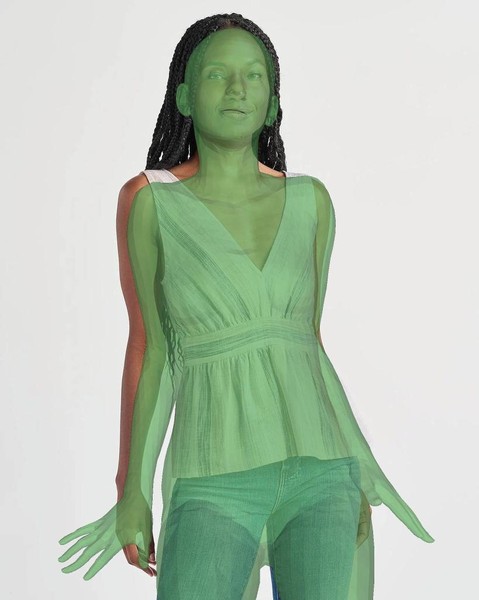}}
\subfloat[]{\includegraphics[width=0.24\linewidth]{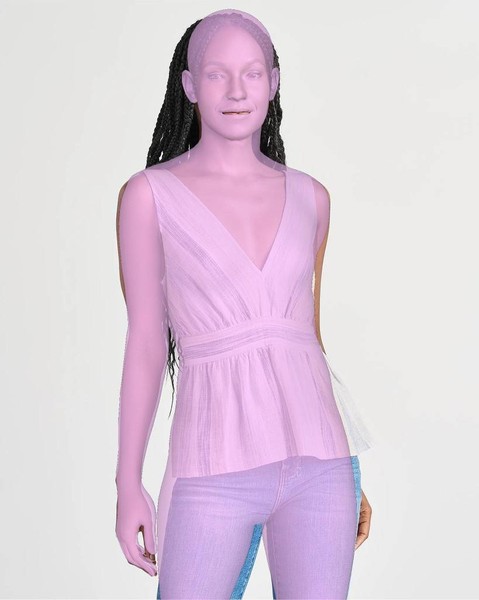}}\\

\vspace{-10pt}

\subfloat[]{\includegraphics[width=0.24\linewidth]{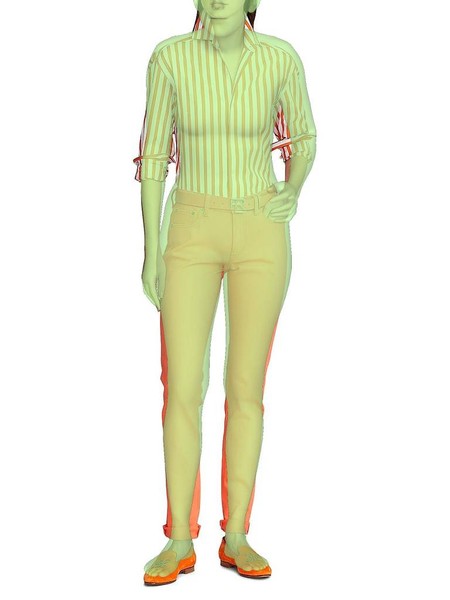}}
\subfloat[]{\includegraphics[width=0.24\linewidth]{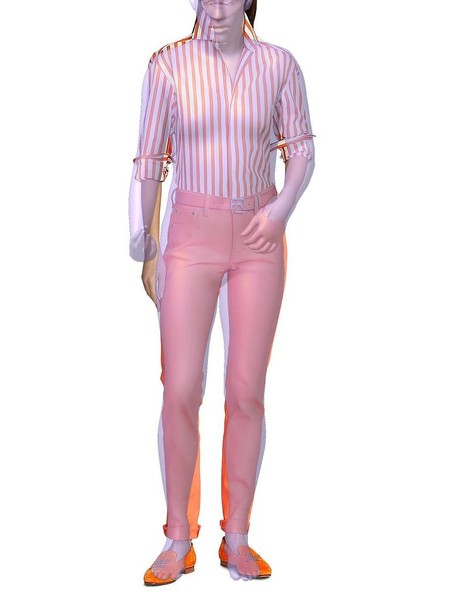}}
\subfloat[]{\includegraphics[width=0.24\linewidth]{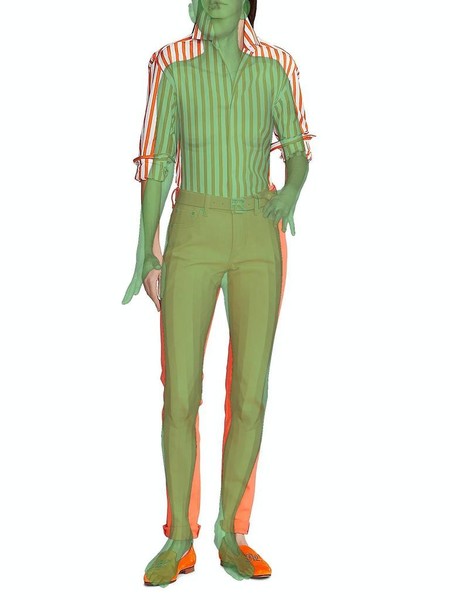}}
\subfloat[]{\includegraphics[width=0.24\linewidth]{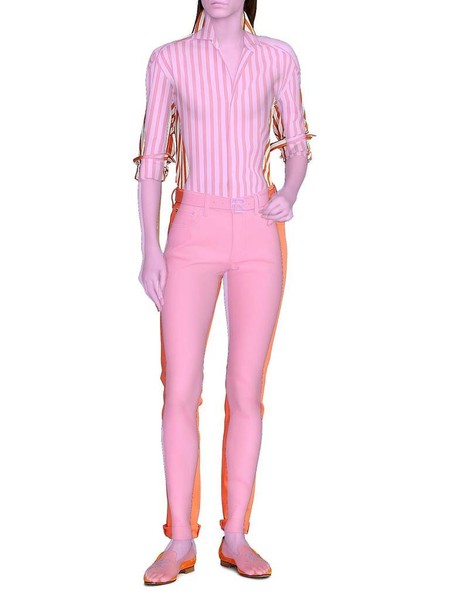}}\\

\vspace{-10pt}

\subfloat[SMPLify-X \cite{pavlakos2019expressive}]{\includegraphics[width=0.24\linewidth]{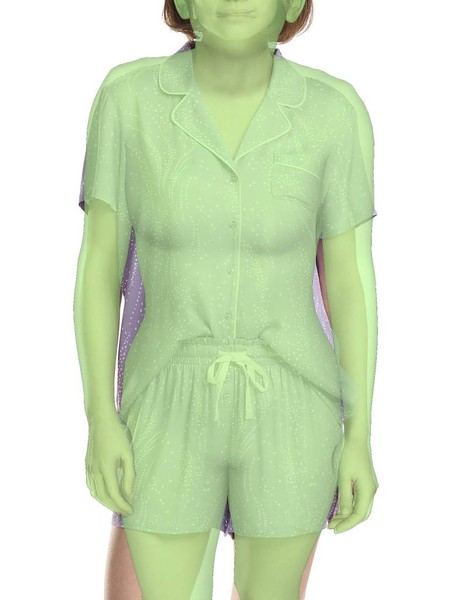}}
\subfloat[PyMAF-X \cite{pymafx2022}]{\includegraphics[width=0.24\linewidth]{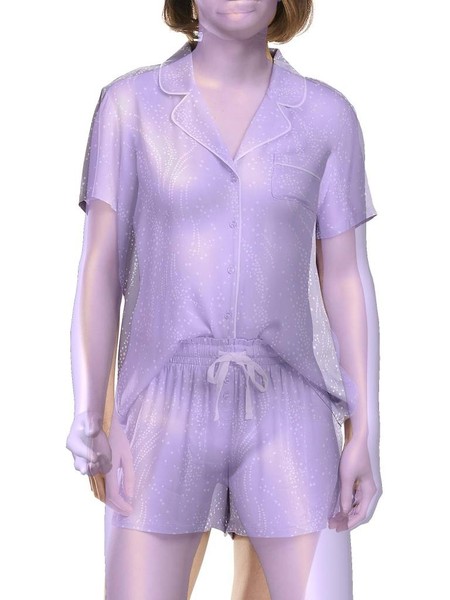}}
\subfloat[SHAPY \cite{choutas2022accurate}]{\includegraphics[width=0.24\linewidth]{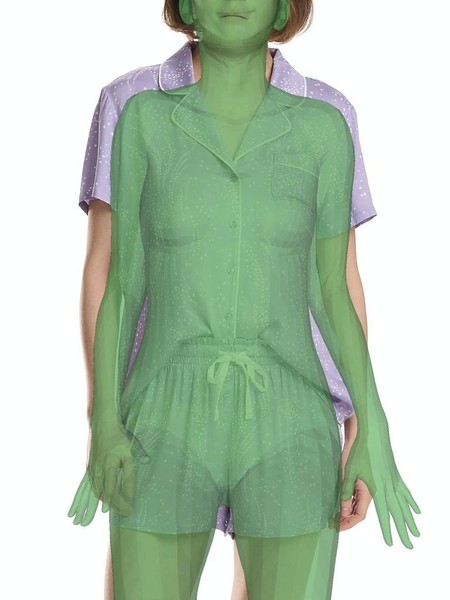}}
\subfloat[\KBody{-.1}{.035}]{\includegraphics[width=0.24\linewidth]{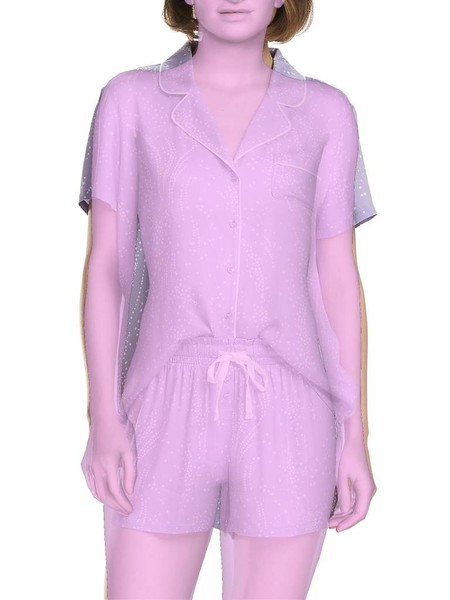}}\\

\caption{
Partial image qualitative results. Same scheme as \cref{fig:full}.
}
\label{fig:partial}
\end{figure}

%% file: figures/adf.tex
\begin{figure}[!htbp]
\captionsetup[subfigure]{position=bottom,labelformat=empty}

\centering

\raisebox{3.0\normalbaselineskip}[0pt][0pt]{\rotatebox[origin=c]{90}{\small \textcolor{green}{with} ADF (\S \ref{subsec:adt})}}
\subfloat[]{\includegraphics[width=0.26\linewidth]{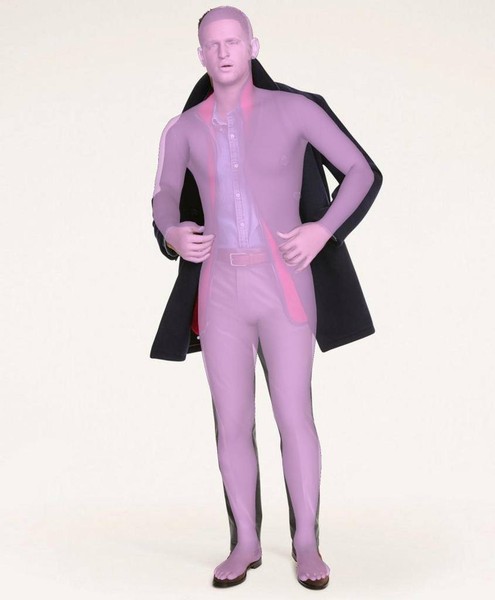}}
\subfloat[]{\includegraphics[width=0.225\linewidth]{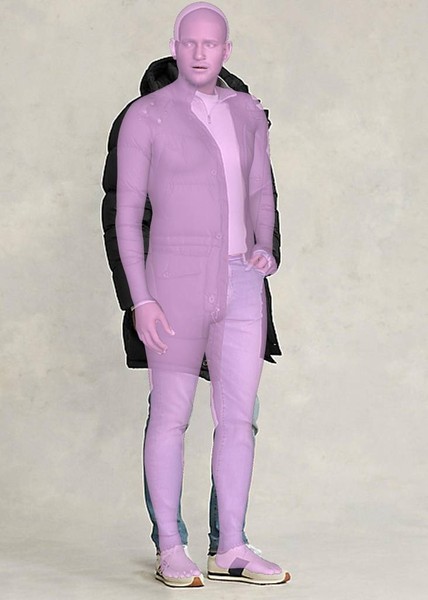}}
\subfloat[]{\includegraphics[width=0.2\linewidth]{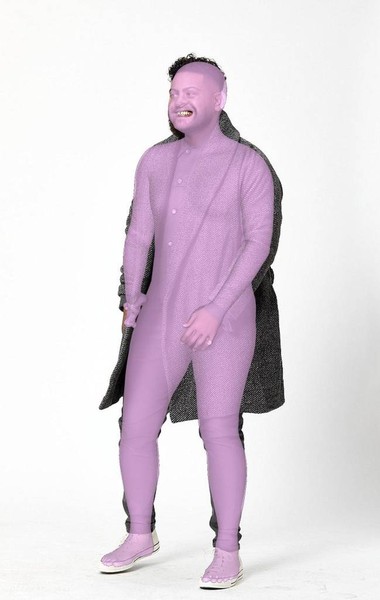}}
\subfloat[]{\includegraphics[width=0.2525\linewidth]{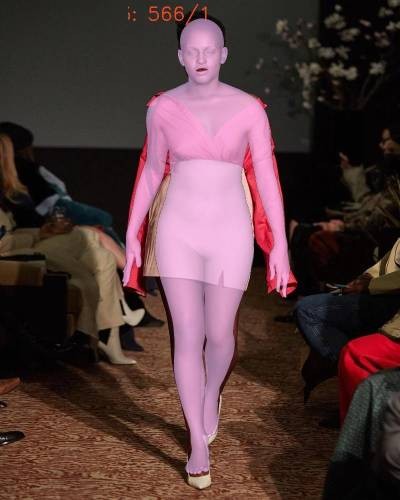}}

\vspace{-12.5pt}

\raisebox{3.0\normalbaselineskip}[0pt][0pt]{\rotatebox[origin=c]{90}{\small \textcolor{red}{w/o} ADF (\S \ref{subsec:adt})}}
\subfloat[]{\includegraphics[width=0.26\linewidth]{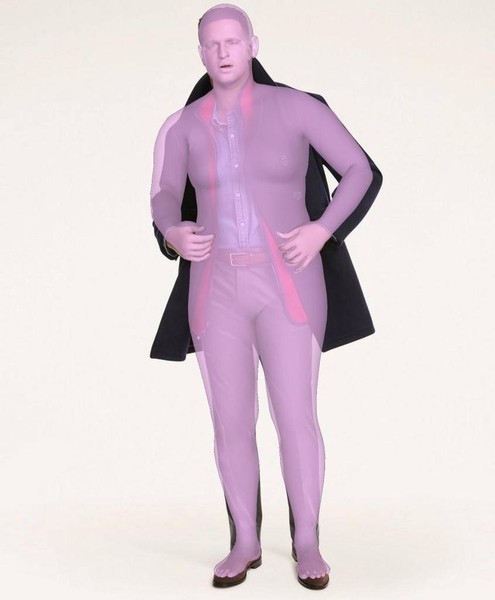}}
\subfloat[]{\includegraphics[width=0.225\linewidth]{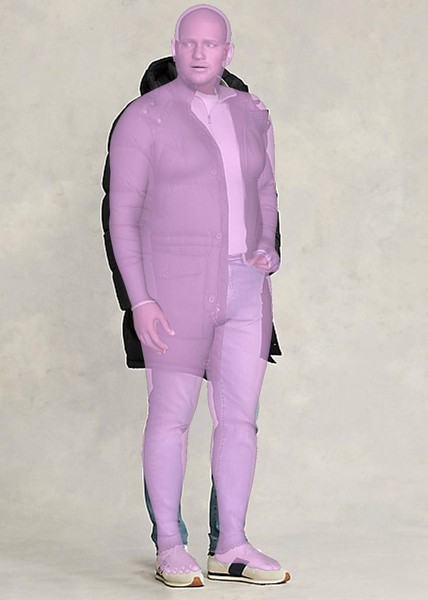}}
\subfloat[]{\includegraphics[width=0.2\linewidth]{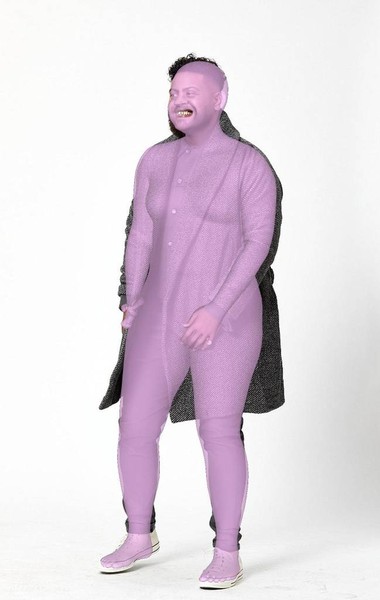}}
\subfloat[]{\includegraphics[width=0.2525\linewidth]{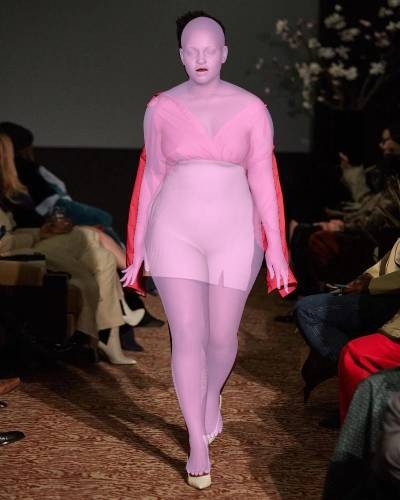}}

\vspace{-10pt}

\caption{
\KBody{}{} fitting results \textcolor{green}{with} (top) and \textcolor{red}{w/o} (bottom) ADF.
}
\label{fig:adf}
\end{figure}

%% file: sections/conclusion.tex
In this work we have presented \KBody{-.05}{.035} a general method for monocular body fitting.
\KBody{-.05}{.035} can handle partial images gracefully via a generative completion stage and employs a multi-constraint fitting approach that delivers high-quality fits, and a balanced performance across pose, shape and image alignment performance.
While the conflicts between pose and shape performance as well as world scale outputs and image alignment remain to be solved, we believe \KBody{-.05}{.035} is a step towards that direction as it shows that it is necessary for single-shot and iterative approaches to co-exist.
However, relying on externally estimated constraints limits applicability to situations where these models under-perform.
Novel solutions for (black-box) uncertainty estimation and multi-modal solving are required.
Still, improving $2D$ estimation models is more practical than acquiring $3D$ data for supervision \cite{corona2022learned} or a wide-range of images and corresponding measurements \cite{choutas2022accurate}.
Finally, relying on an image-based appearance prior for completion comes with limitations for non frontal facing images which can be addressed in the future with $3D$ aware generative models.

%% file: appendix.tex
\appendix
\label{sec:appendix}

\section{Complementary Discussion}
In the main paper, qualitative and quantitative comparisons are offered to help appreciate the gains of \KBody{}{}.
We select PyMAf-X \cite{pymafx2022} as a representative of a single-shot estimator that delivers high quality performance on pose capture, evident also by its quantitative and qualitative performance.
On the other end, we also select SHAPY \cite{choutas2022accurate} as a representative of a single-shot estimator that delivers high quality performance on shape capture, also evident by its quantiative and qualitative performance.
Further, as our approach is an iterative optimization based one, we also include the SMPLify-X \cite{pavlakos2019expressive} baseline and its data-driven successor ExPose \cite{choutas2020monocular} which is mostly used as a parameter initializer.
All aforementioned methods are used with their original -- best -- hyper-parameters and pre-trained models, receiving the same inputs, and, where necessary, cropped with an extended bounding box calculated from OpenPose's \cite{openpose} keypoints.
Performance is first evaluated on a 3D vertex level, with V2V evaluating pose and shape jointly, PVE-T-SC evaluating shape only, and the body specific measurement metrics presented in SHAPY \cite{choutas2022accurate}.
We also evaluate pixel alignment performance using the ground-truth rendered bodies when available (EHF \& SSP3D), and by extracting segmentation masks using a background subtraction service \cite{removebg} on the Lab images of HBW (val), which contain minimal clothing and plain backgrounds.

From a strict comparison perspective, PyMAF-X and SHAPY are single-shot regressors, while SMPLify-X and KBody are iterative optimizers.
Still, the latter are constrained by the estimations of third-party regressors like OpenPose \cite{openpose} for the 2D keypoints and MODNet \cite{ke2022modnet} for the silhouette.
While the \textit{predict-and-optimize} approach can be considered as a generic scheme and be applied to different regressors like PyMAF-X and SHAPY, there are conflicting arguments made in the literature.

Both EFT \cite{joo2021exemplar} and Pose-NDF \cite{tiwari2022pose} show that simply post-processing an initial regressor estimate with iterative optimization using third-party regressed constraints does not necessarily lead to improved results.
This is the case for ExPose \cite{choutas2020monocular} being refined via SMPLify-X \cite{pavlakos2019expressive} in the experiments conducted in Pose-NDF \cite{tiwari2022pose} and reported in Table~1 of the main paper, as well as the experiments reported in Table~5 of EFT \cite{joo2021exemplar}, where better performing initializations lead to inferior results post fine-tuning.

Instead the exemplar fine-tuning \cite{joo2021exemplar} that includes no prior terms and only exploits the prior learned by the model is shown to always improve results, albeit less in cases where the initialization is already of high quality.
However, EFT post-processing \cite{joo2021exemplar} was only applied to simple single-shot pose-dominant regressors, whereas the best performing models in each category, PyMAF-X \cite{pymafx2022} and SHAPY \cite{choutas2022accurate}, are modified versions of these regressors whose interplay with EFT post-processing is to be investigated.
PyMAF-X \cite{pymafx2022} already contains a feedback loop, essentially a mini iteration loop using its pyramidal features, to improve pose estimates, and its results interestingly show that this does not improve shape capture as it regresses towards a camera-scaled mean shape.
SHAPY \cite{choutas2022accurate} on the other hand was (partly) trained with shape annotations, a fact that improved the performance of the shape coefficient regression, but crucially reduced pose performance.
Original EFT post-processing would only use the OpenPose keypoints to fine-tune, with unclear consequences for the shape estimate.
Still, different schemes that combine the silhouette term, the original SHAPY shape measurements and a discarding of the optimized shape coefficients are potential options, but these are expected to require extensive fine-tuning of the parameters and the process itself, opening up an entirely new problem.

Our findings show that achieving holistic human capture across its different components (shape, pose, image alignment -- see Figure~1 of the main paper) requires the integration of predictions and optimization which, considering the rapid important advances in the former, needs more investigative effort on the post-processing/fine-tuning side to align with these developments.

Finally, robustness and applicability to in-the-wild images is an orthogonal goal.
These include challenging poses and a variety of body shapes, depicted in images with plain and complex backgrounds, either in full or partially.
Section \ref{supp:qual} of this \supp includes an extensive set (over $200$) of qualitative results comparison across full and partial images and an asymmetric distance field ablation.

\section{Runtime Performance}
The single-shot regressors (PyMAF-X \cite{pymafx2022}, SHAPY \cite{choutas2022accurate}) exhibit significantly better runtime performance as they produce their estimates an order of magnitude faster (around $0.1s$) than the \textit{unoptimized} iterative optimization approaches (SMPLify-X \cite{pavlakos2019expressive}, KBody) that need some seconds ($\sim \geq 17s$) to converge.
While KBody exploits the ExPose \cite{choutas2020monocular} initialization to skip the camera initialization stages of SMPLify-X \cite{pavlakos2019expressive}, the use of a differentiable renderer at its later stages incurs extra computational costs, especially for gradient computations.
This amounts to an increase of $40\%$ in runtime compared to SMPLify-X \cite{pavlakos2019expressive}, with all the measurements taken on the same computational infrastructure, an AWS $g4dn.2xlarge$ instance.

\section{Qualitative Results}
\label{supp:qual}
\cref{fig:hm1,fig:hm2,fig:hm3,fig:hm_splendid,fig:good,fig:fh_good,fig:fh1,fig:fh2,fig:fh3,fig:fh4,fig:ov1,fig:ov2,fig:ov3,fig:p1,fig:p2,fig:p3,fig:p4,fig:p5,fig:splendid,fig:v1,fig:v2,fig:v3,fig:w1,fig:w2,fig:w3} present $112$ qualitative result comparisons between the presented KBody method (rightmost - \textcolor{candypink}{pink}) the optimization-based SMPLify-X \cite{pavlakos2019expressive} (leftmost - \textcolor{caribbeangreen2}{light green}), and the single-shot models PyMAF-X \cite{pymafx2022} (middle left - \textcolor{violet}{purple}) and SHAPY \cite{choutas2022accurate} (middle right - \textcolor{jade}{green}), focusing on pose and shape capturing respectively.

In addition, \cref{fig:adf_hm1,fig:adf_hm2,fig:adf_hm_mens_big,fig:adf_mens_big1} present an extra $32$ qualitative results that demonstrate benefits of the asymmetric distance field (ADF), compared to a symmetric variant, when considering clothing robustness gains and unnatural shape captures.

Further, \cref{fig:partial_fh1,fig:partial_fh2,fig:partial_fh3,fig:partial_fh4,fig:partial_ll1,fig:partial_ll2,fig:partial_sp1,fig:partial_sp2,fig:partial_sp3,fig:partial_sp5,fig:partial_sp7,fig:partial_sp8,fig:partial_sp9,fig:partial_sp_good,fig:partial_good,fig:partial_rl1,fig:partial_rl2,fig:partial_rl3,fig:partial_rl4,fig:partial_rl5} present $78$ qualitative result comparisons between the presented KBody method (rightmost - \textcolor{candypink}{pink}), the optimization-based SMPLify-X \cite{pavlakos2019expressive} (leftmost - \textcolor{caribbeangreen2}{light green}), and the single-shot models PyMAF-X \cite{pymafx2022} (middle left - \textcolor{violet}{purple}) and SHAPY \cite{choutas2022accurate} (middle right - \textcolor{jade}{green}), focusing on pose and shape capturing respectively.
These examples focus on partial images with missing head and/or lower body information, and present a challenging scenario for high quality monocular body fitting.
Our generative inversion-based completion approach handles them gracefully and helps produce reasonable fits even in the absence of important information.
As illustrated by the examples, priors alone cannot handle this properly for optimization-based approaches like SMPLify-X \cite{pavlakos2019expressive}, while single-shot estimates \cite{pymafx2022,choutas2022accurate} exhibit reduced performance given the lack of necessary image context.

\clearpage

\input{figures/supp/faherty1.tex}
\input{figures/supp/faherty2.tex}
\input{figures/supp/faherty3.tex}
\input{figures/supp/faherty4.tex}
\input{figures/supp/faherty_good_american.tex}
\input{figures/supp/good_american.tex}
\input{figures/supp/hm1.tex}
\input{figures/supp/hm2.tex}
\input{figures/supp/hm3.tex}
\input{figures/supp/hm_splendid.tex}
\input{figures/supp/splendid.tex}
\input{figures/supp/problem1.tex}
\input{figures/supp/problem2.tex}
\input{figures/supp/problem3.tex}
\input{figures/supp/problem4.tex}
\input{figures/supp/problem5.tex}
\input{figures/supp/oversized1.tex}
\input{figures/supp/oversized2.tex}
\input{figures/supp/oversized3.tex}
\input{figures/supp/veronica1.tex}
\input{figures/supp/veronica2.tex}
\input{figures/supp/veronica3.tex}
\input{figures/supp/womens_plus_size1.tex}
\input{figures/supp/womens_plus_size2.tex}
\input{figures/supp/womens_plus_size3.tex}
\input{figures/supp/womens_plus_size4.tex}
\input{figures/supp/womens_plus_size5.tex}
\input{figures/supp/womens_plus_size6.tex}

\input{figures/supp/adf_hm1.tex}
\input{figures/supp/adf_hm2.tex}
\input{figures/supp/adf_hm_mens_big.tex}
\input{figures/supp/adf_mens_big1.tex}

\input{figures/supp/partial_faherty1.tex}
\input{figures/supp/partial_faherty2.tex}
\input{figures/supp/partial_faherty3.tex}
\input{figures/supp/partial_faherty4.tex}
\input{figures/supp/partial_lululemone1.tex}
\input{figures/supp/partial_lululemone2.tex}
\input{figures/supp/partial_splendid1.tex}
\input{figures/supp/partial_splendid2.tex}
\input{figures/supp/partial_splendid3.tex}
\input{figures/supp/partial_splendid5.tex}
\input{figures/supp/partial_splendid7.tex}
\input{figures/supp/partial_splendid8.tex}
\input{figures/supp/partial_splendid9.tex}
\input{figures/supp/partial_splendid_good_american.tex}
\input{figures/supp/partial_good_american.tex}
\input{figures/supp/partial_rl1.tex}
\input{figures/supp/partial_rl2.tex}
\input{figures/supp/partial_rl3.tex}
\input{figures/supp/partial_rl4.tex}
\input{figures/supp/partial_rl5.tex}

%% file: figures/supp/faherty1.tex
\begin{figure*}[!htbp]
\captionsetup[subfigure]{position=bottom,labelformat=empty}

\centering

\subfloat{\includegraphics[height=0.24\textheight]{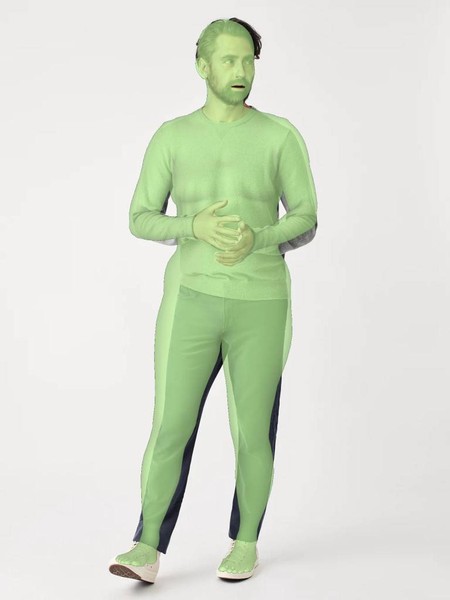}}
\subfloat{\includegraphics[height=0.24\textheight]{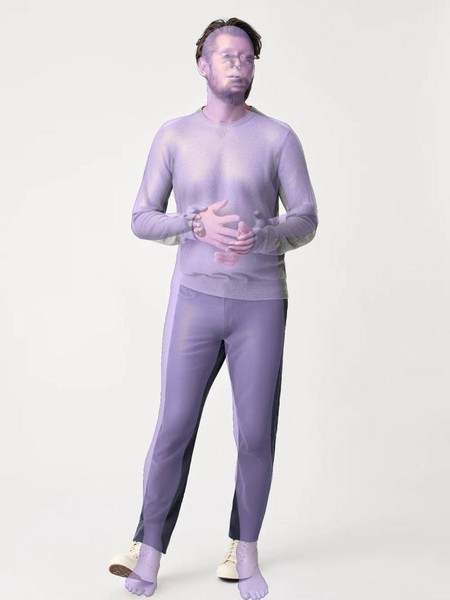}}
\subfloat{\includegraphics[height=0.24\textheight]{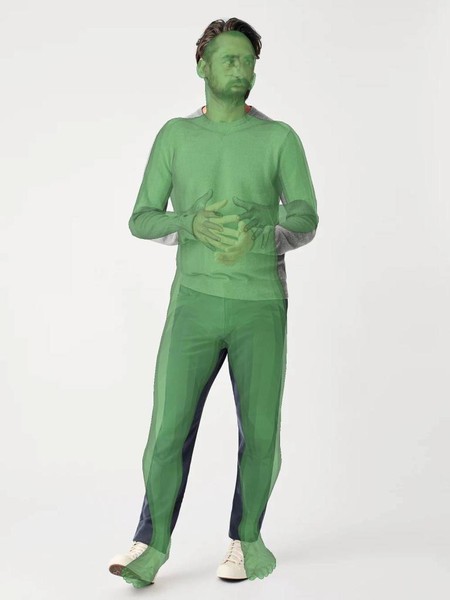}}
\subfloat{\includegraphics[height=0.24\textheight]{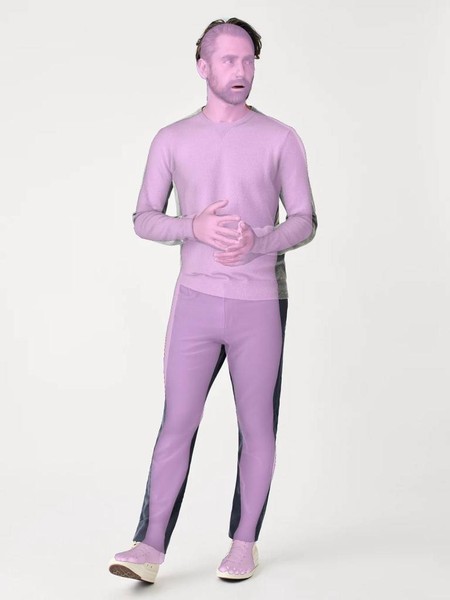}}

\subfloat{\includegraphics[height=0.24\textheight]{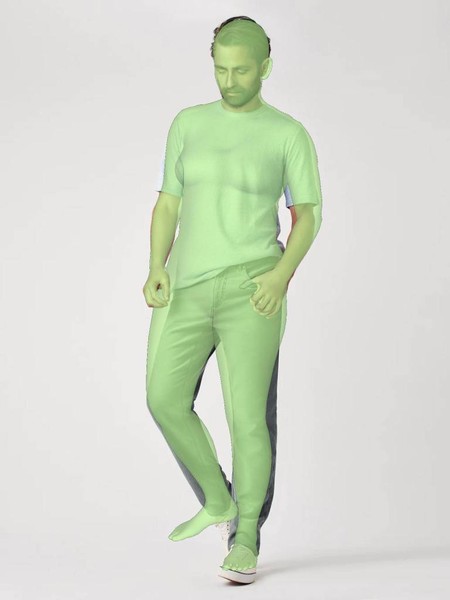}}
\subfloat{\includegraphics[height=0.24\textheight]{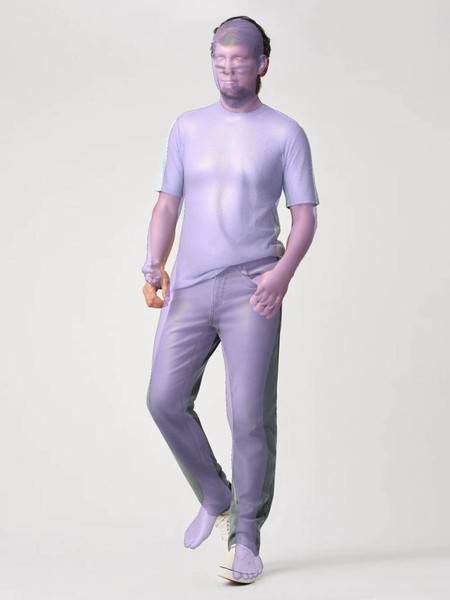}}
\subfloat{\includegraphics[height=0.24\textheight]{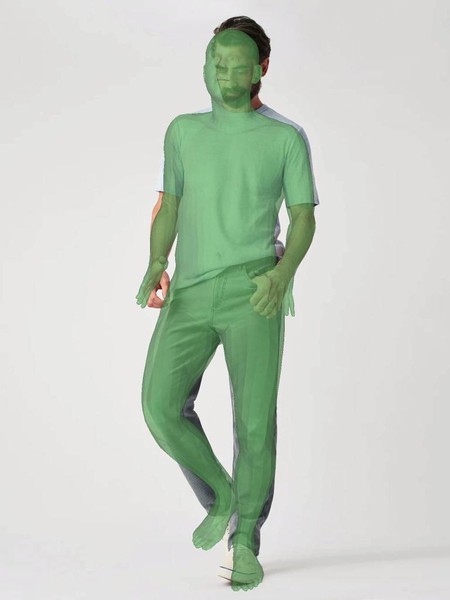}}
\subfloat{\includegraphics[height=0.24\textheight]{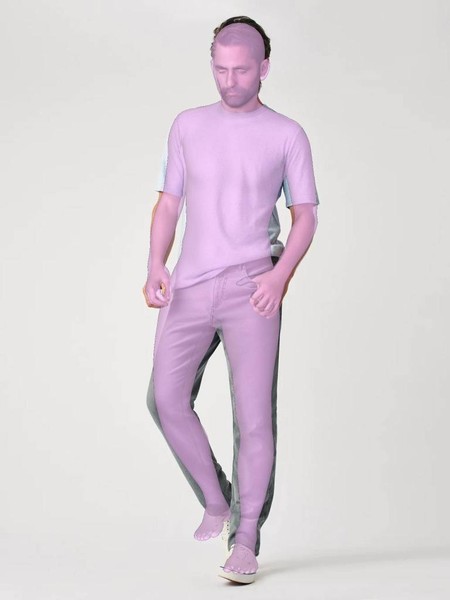}}

\subfloat{\includegraphics[height=0.24\textheight]{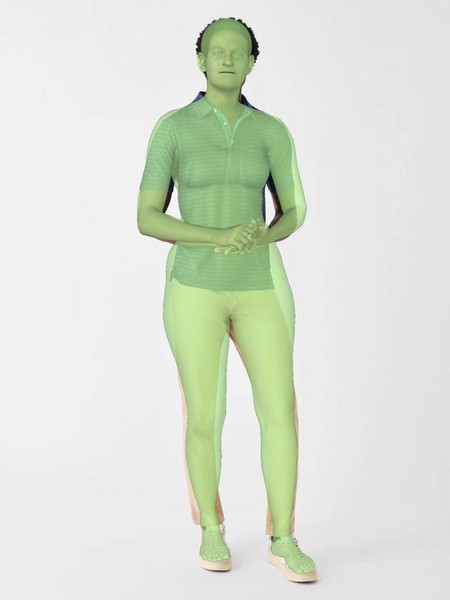}}
\subfloat{\includegraphics[height=0.24\textheight]{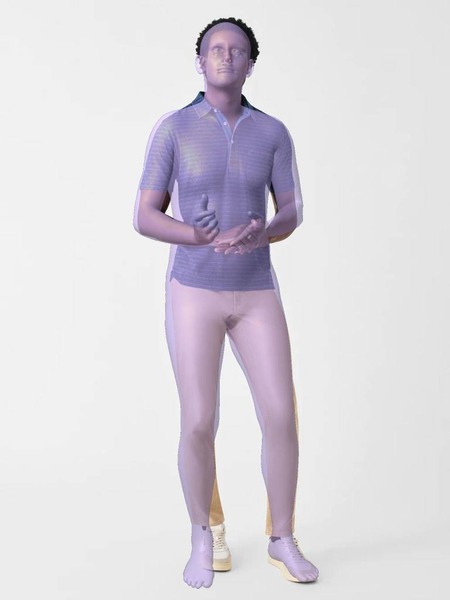}}
\subfloat{\includegraphics[height=0.24\textheight]{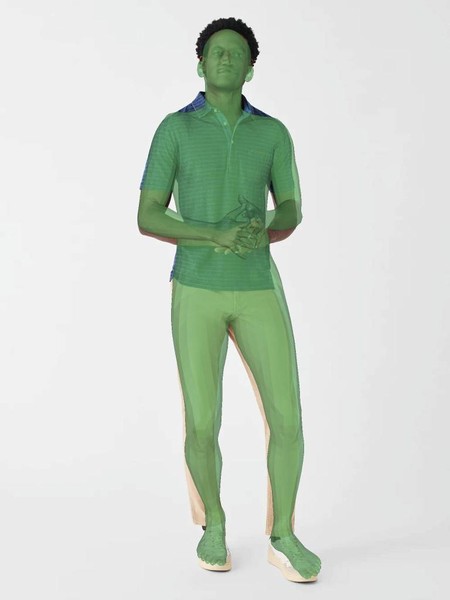}}
\subfloat{\includegraphics[height=0.24\textheight]{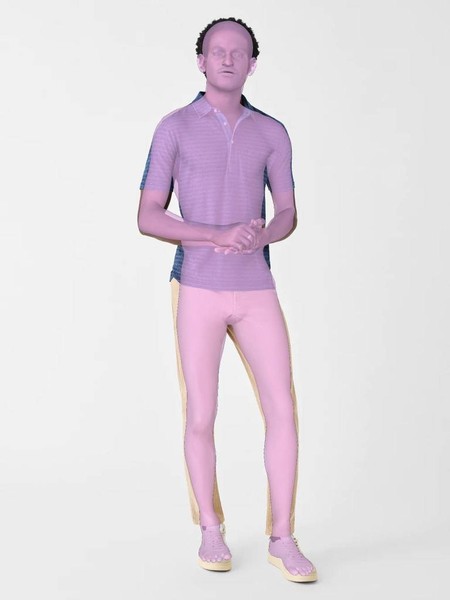}}

\subfloat[SMPLify-X \cite{pavlakos2019expressive}]
{\includegraphics[height=0.24\textheight]{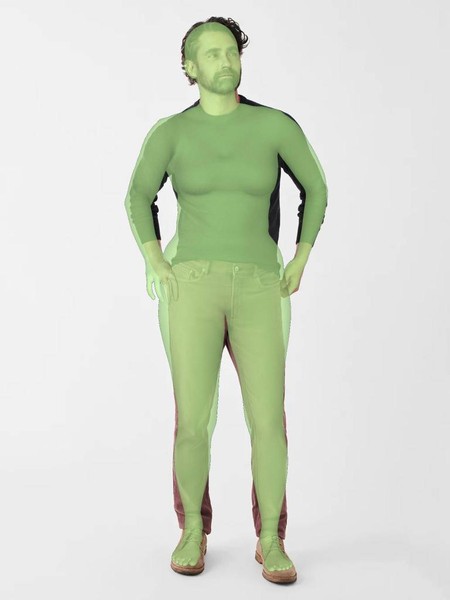}}
\subfloat[PyMAF-X \cite{pymafx2022}]{\includegraphics[height=0.24\textheight]{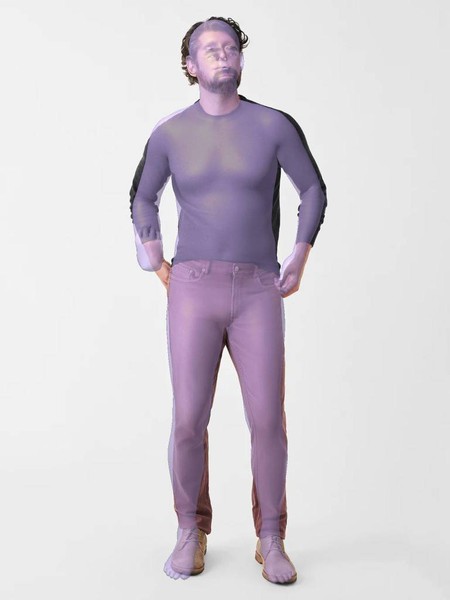}}
\subfloat[SHAPY \cite{choutas2022accurate}]
{\includegraphics[height=0.24\textheight]
{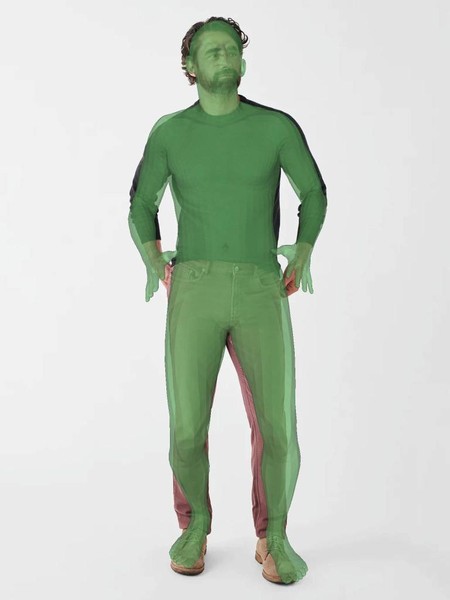}}
\subfloat[\KBody{-.1}{.035} (Ours)]{\includegraphics[height=0.24\textheight]{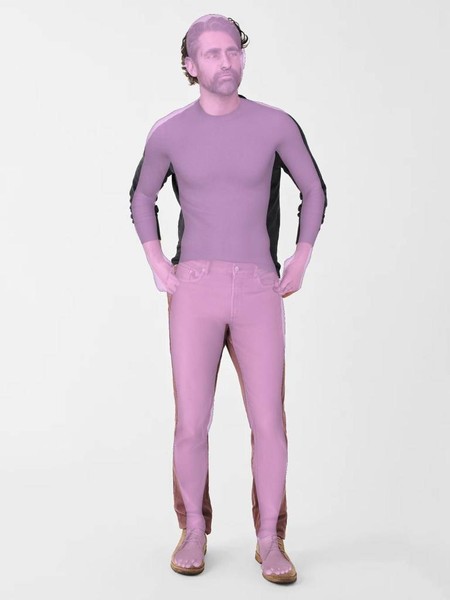}}

\caption{
Left-to-right: SMPLify-X \cite{pavlakos2019expressive} (\textcolor{caribbeangreen2}{light green}), PyMAF-X \cite{pymafx2022} (\textcolor{violet}{purple}), SHAPY \cite{choutas2022accurate} (\textcolor{jade}{green}) and KBody (\textcolor{candypink}{pink}).
}
\label{fig:fh1}
\end{figure*}

%% file: figures/supp/faherty2.tex
\begin{figure*}[!htbp]
\captionsetup[subfigure]{position=bottom,labelformat=empty}

\centering

\subfloat{\includegraphics[height=0.24\textheight]{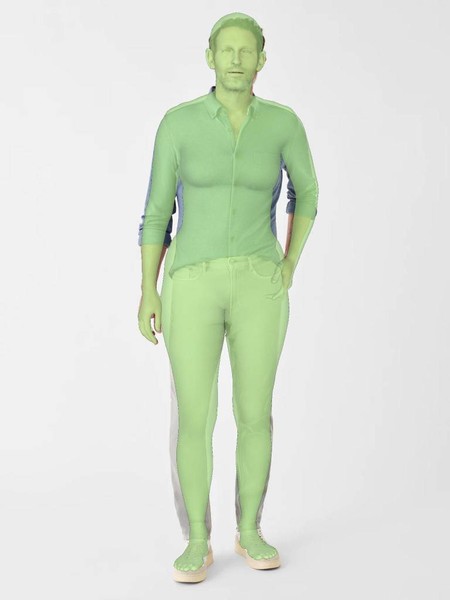}}
\subfloat{\includegraphics[height=0.24\textheight]{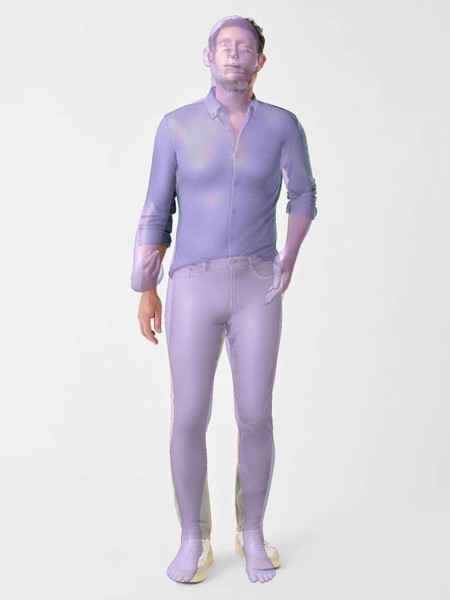}}
\subfloat{\includegraphics[height=0.24\textheight]{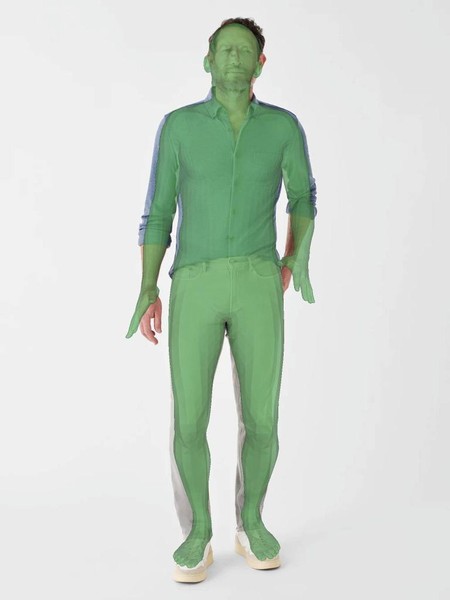}}
\subfloat{\includegraphics[height=0.24\textheight]{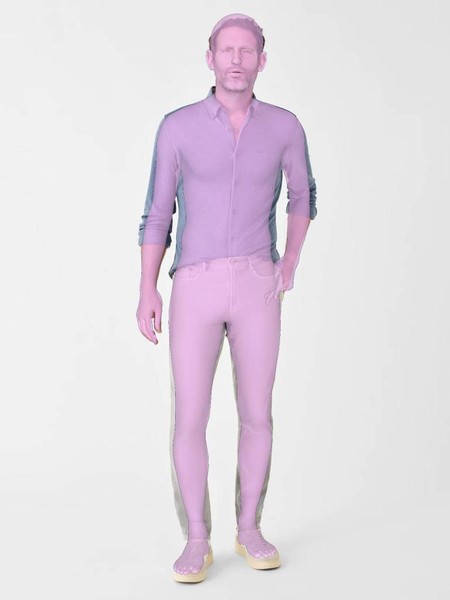}}

\subfloat{\includegraphics[height=0.24\textheight]{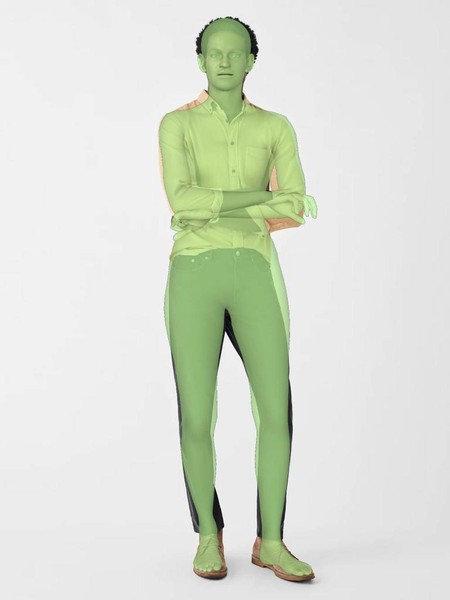}}
\subfloat{\includegraphics[height=0.24\textheight]{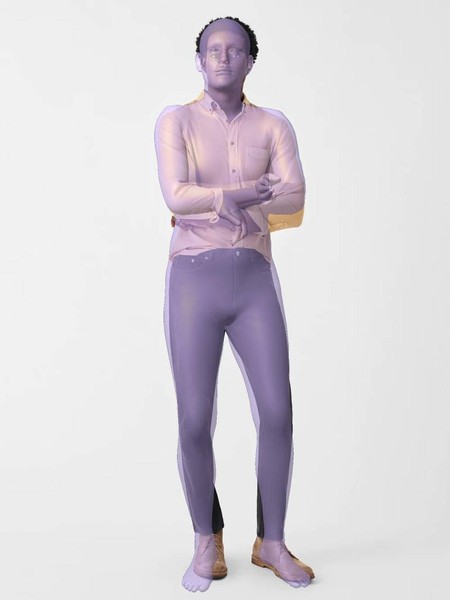}}
\subfloat{\includegraphics[height=0.24\textheight]{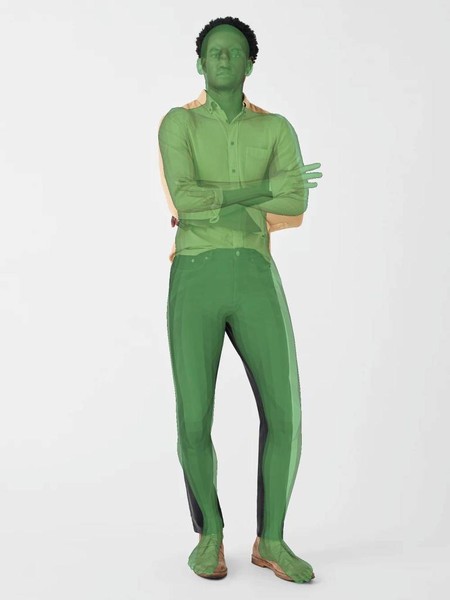}}
\subfloat{\includegraphics[height=0.24\textheight]{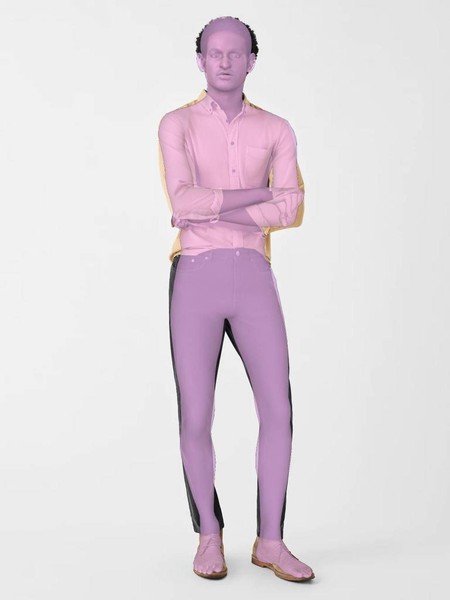}}

\subfloat{\includegraphics[height=0.24\textheight]{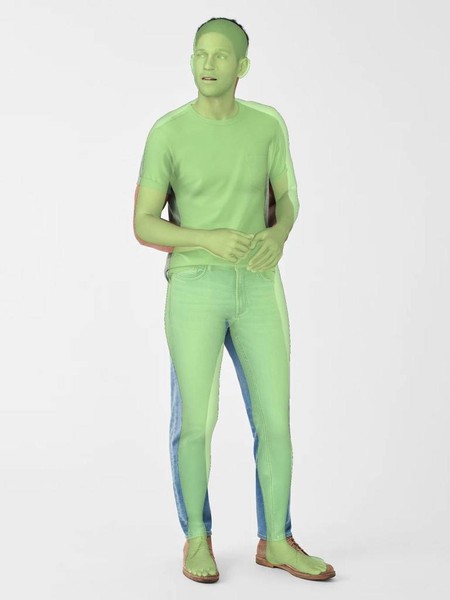}}
\subfloat{\includegraphics[height=0.24\textheight]{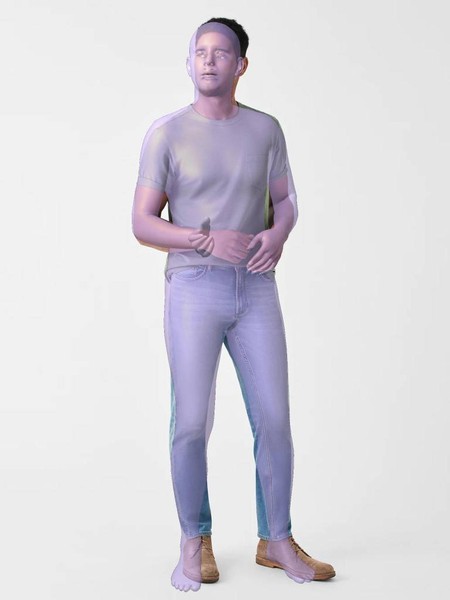}}
\subfloat{\includegraphics[height=0.24\textheight]{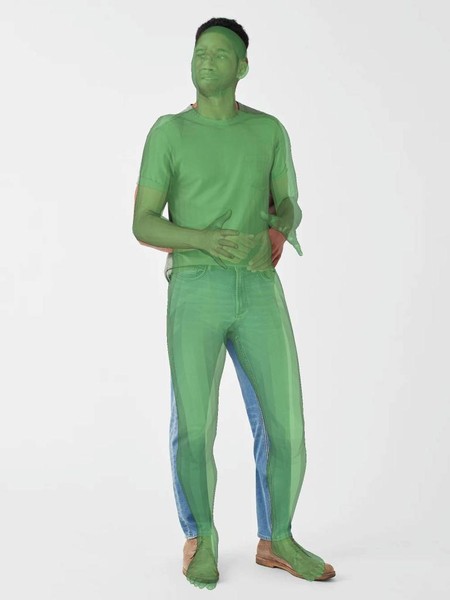}}
\subfloat{\includegraphics[height=0.24\textheight]{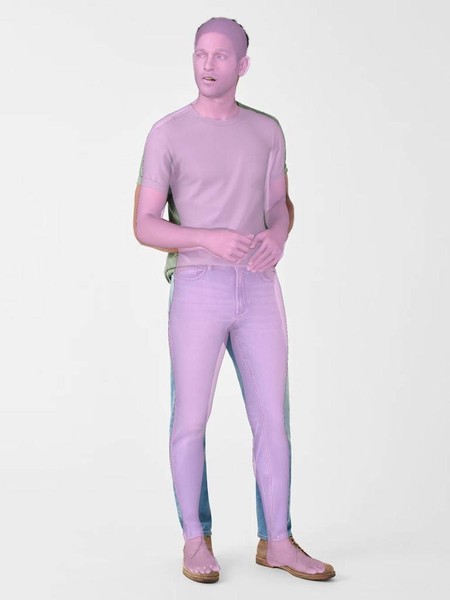}}

\subfloat[SMPLify-X \cite{pavlakos2019expressive}]
{\includegraphics[height=0.24\textheight]{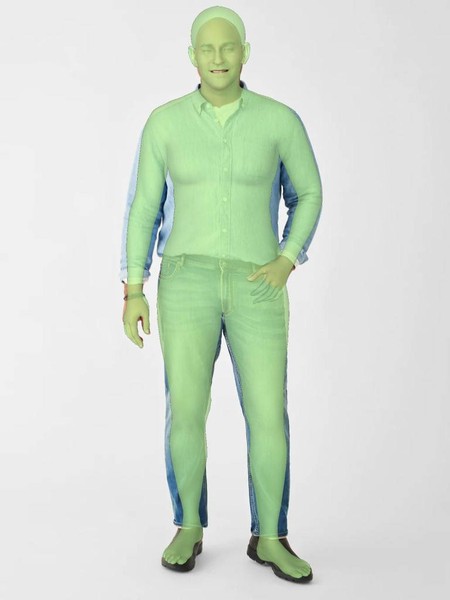}}
\subfloat[PyMAF-X \cite{pymafx2022}]{\includegraphics[height=0.24\textheight]{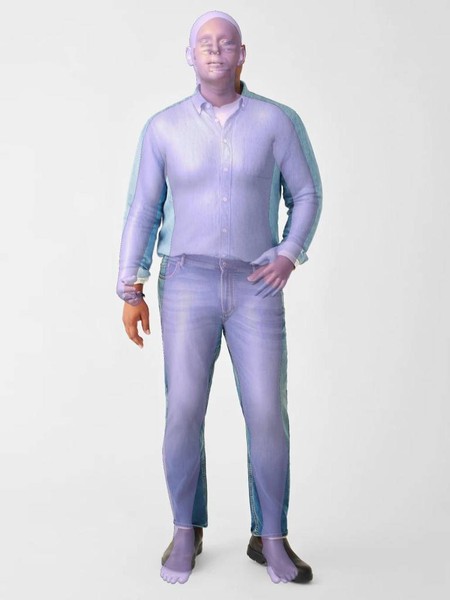}}
\subfloat[SHAPY \cite{choutas2022accurate}]
{\includegraphics[height=0.24\textheight]
{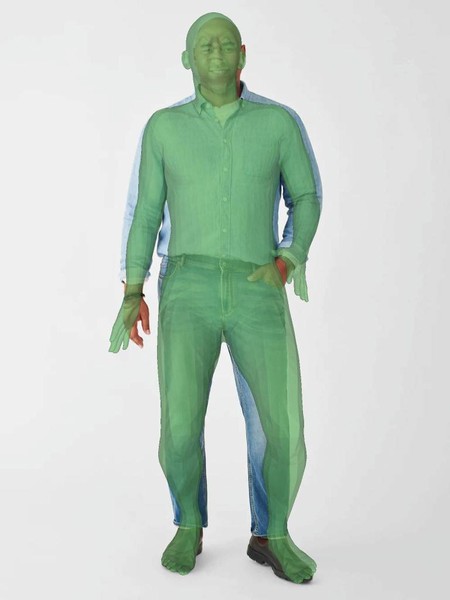}}
\subfloat[\KBody{-.1}{.035} (Ours)]{\includegraphics[height=0.24\textheight]{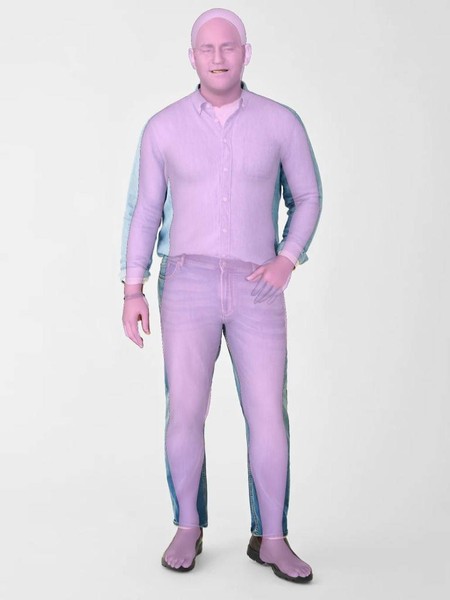}}

\caption{
Left-to-right: SMPLify-X \cite{pavlakos2019expressive} (\textcolor{caribbeangreen2}{light green}), PyMAF-X \cite{pymafx2022} (\textcolor{violet}{purple}), SHAPY \cite{choutas2022accurate} (\textcolor{jade}{green}) and KBody (\textcolor{candypink}{pink}).
}
\label{fig:fh2}
\end{figure*}

%% file: figures/supp/faherty3.tex
\begin{figure*}[!htbp]
\captionsetup[subfigure]{position=bottom,labelformat=empty}

\centering

\subfloat{\includegraphics[height=0.24\textheight]{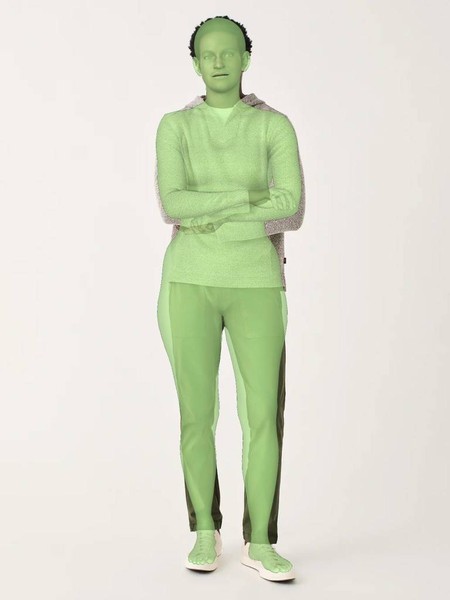}}
\subfloat{\includegraphics[height=0.24\textheight]{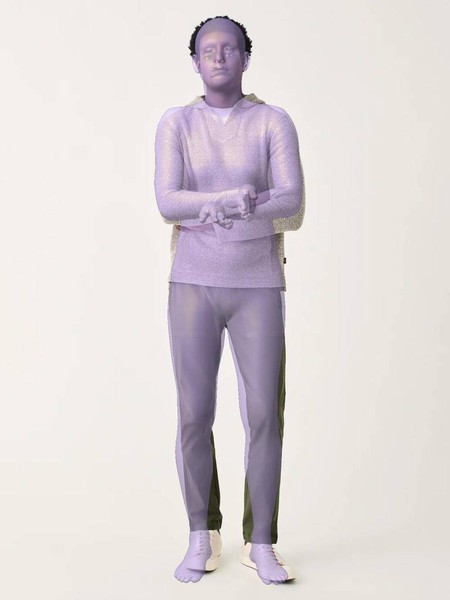}}
\subfloat{\includegraphics[height=0.24\textheight]{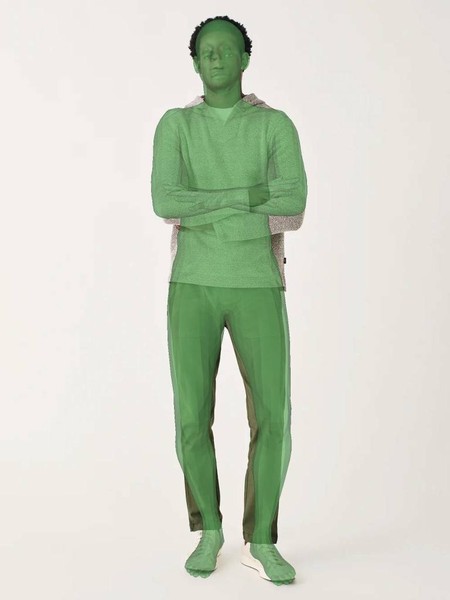}}
\subfloat{\includegraphics[height=0.24\textheight]{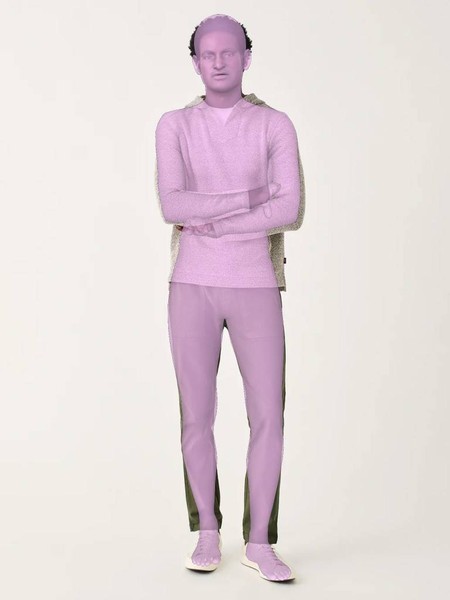}}

\subfloat{\includegraphics[height=0.24\textheight]{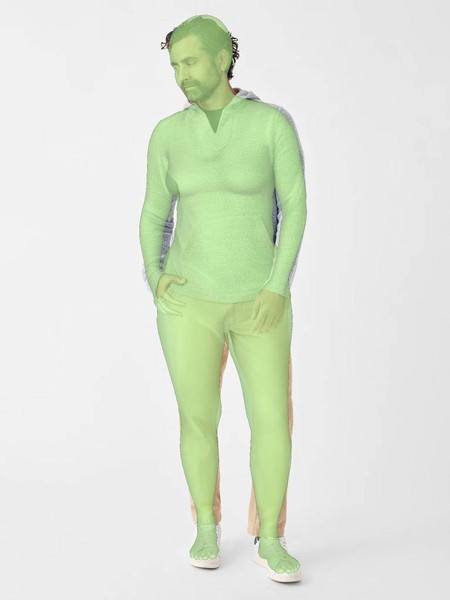}}
\subfloat{\includegraphics[height=0.24\textheight]{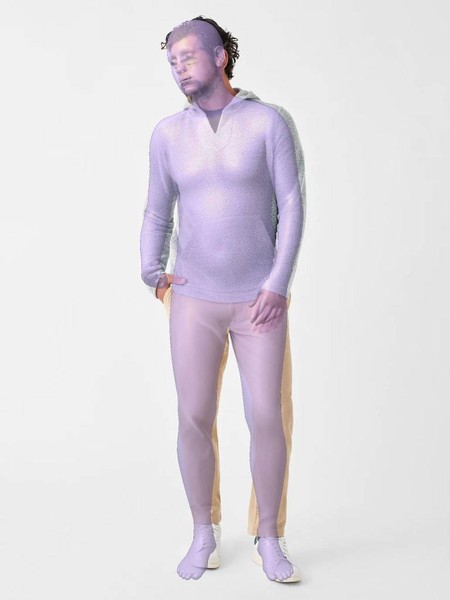}}
\subfloat{\includegraphics[height=0.24\textheight]{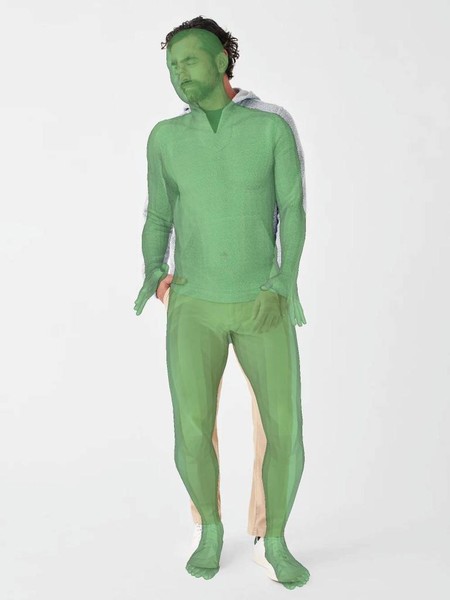}}
\subfloat{\includegraphics[height=0.24\textheight]{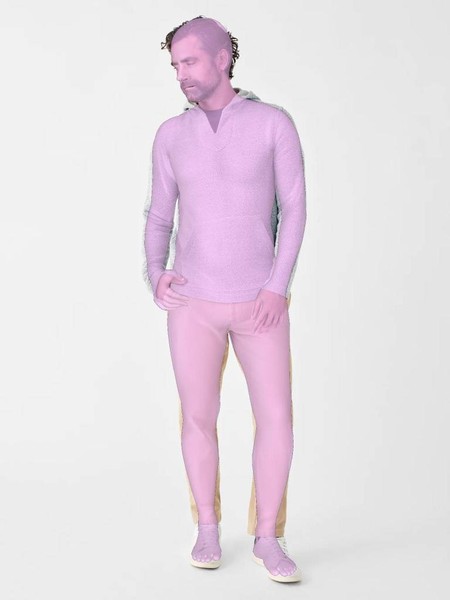}}

\subfloat{\includegraphics[height=0.24\textheight]{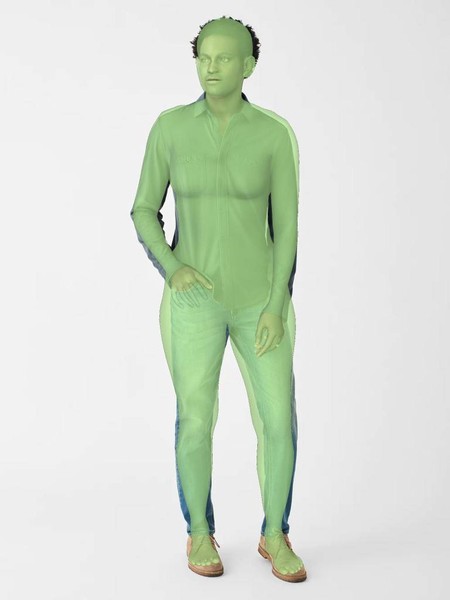}}
\subfloat{\includegraphics[height=0.24\textheight]{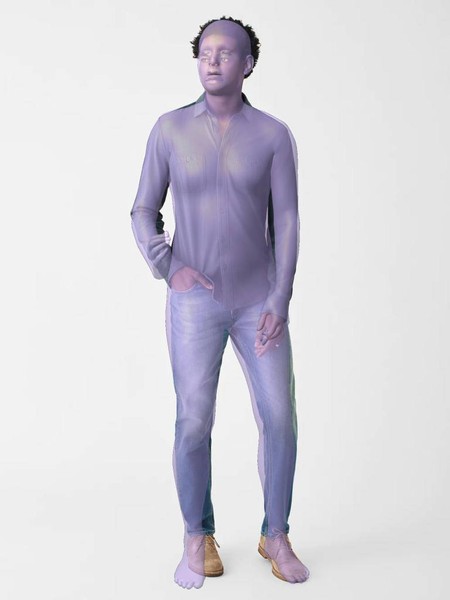}}
\subfloat{\includegraphics[height=0.24\textheight]{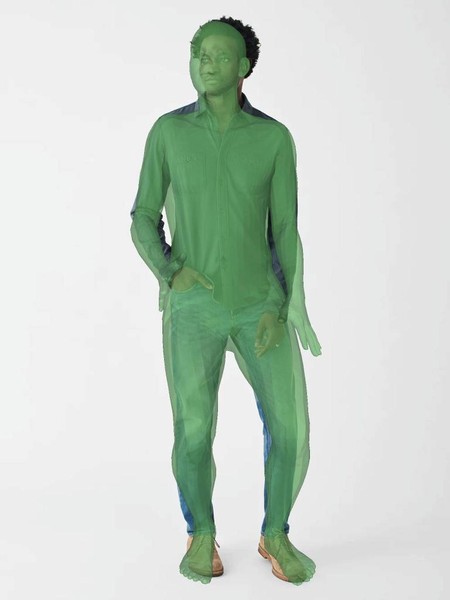}}
\subfloat{\includegraphics[height=0.24\textheight]{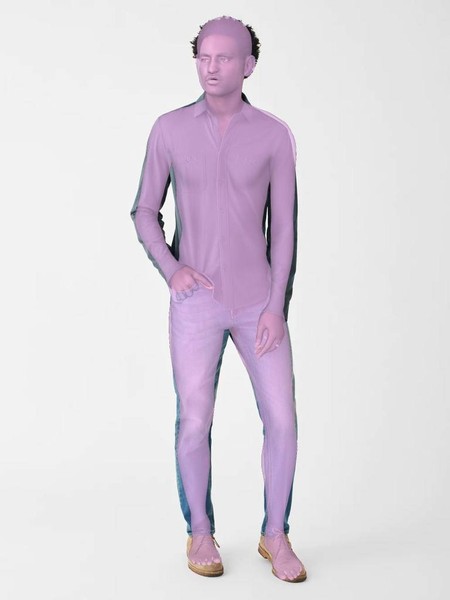}}

\subfloat[SMPLify-X \cite{pavlakos2019expressive}]
{\includegraphics[height=0.24\textheight]{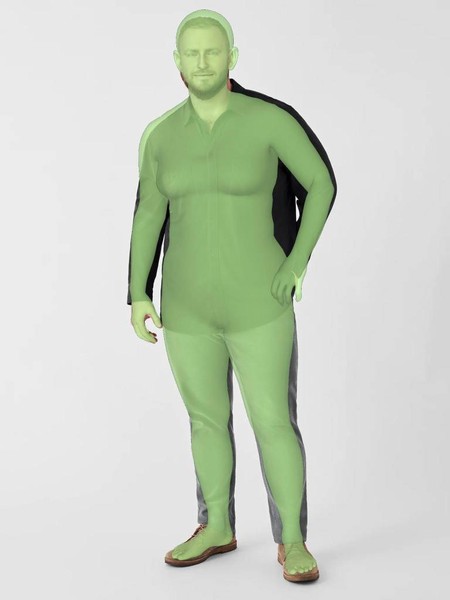}}
\subfloat[PyMAF-X \cite{pymafx2022}]{\includegraphics[height=0.24\textheight]{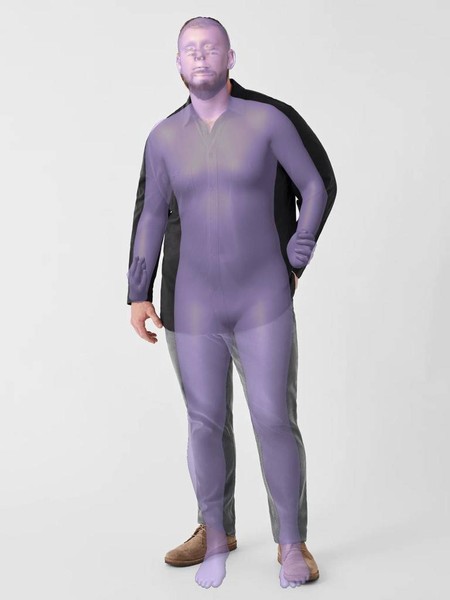}}
\subfloat[SHAPY \cite{choutas2022accurate}]
{\includegraphics[height=0.24\textheight]
{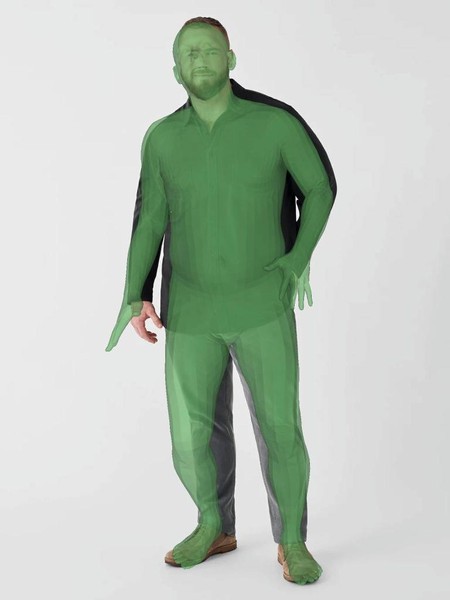}}
\subfloat[\KBody{-.1}{.035} (Ours)]{\includegraphics[height=0.24\textheight]{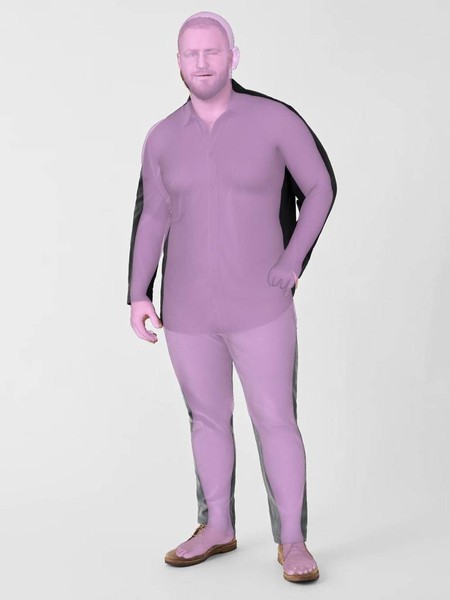}}

\caption{
Left-to-right: SMPLify-X \cite{pavlakos2019expressive} (\textcolor{caribbeangreen2}{light green}), PyMAF-X \cite{pymafx2022} (\textcolor{violet}{purple}), SHAPY \cite{choutas2022accurate} (\textcolor{jade}{green}) and KBody (\textcolor{candypink}{pink}).
}
\label{fig:fh3}
\end{figure*}

%% file: figures/supp/faherty4.tex
\begin{figure*}[!htbp]
\captionsetup[subfigure]{position=bottom,labelformat=empty}

\centering

\subfloat{\includegraphics[height=0.24\textheight]{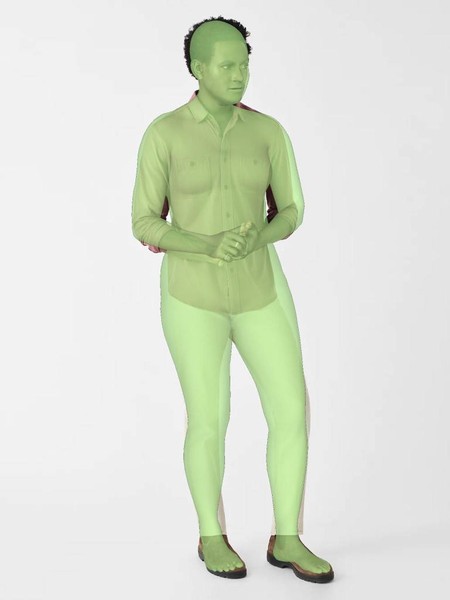}}
\subfloat{\includegraphics[height=0.24\textheight]{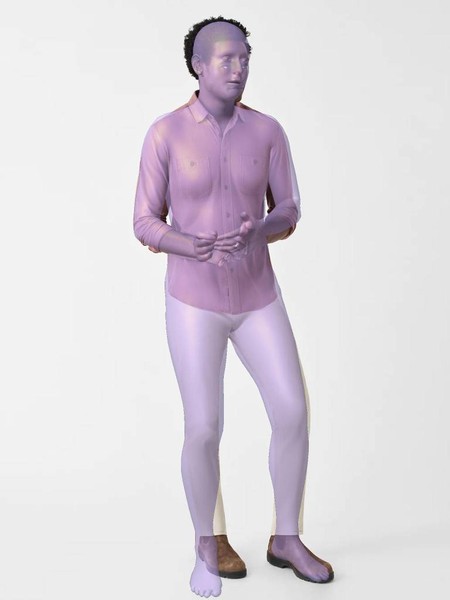}}
\subfloat{\includegraphics[height=0.24\textheight]{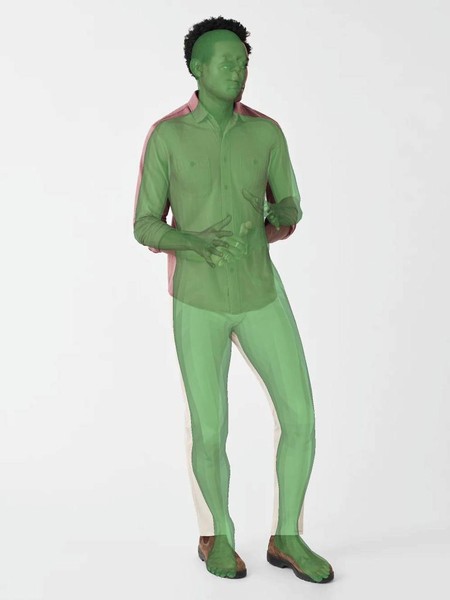}}
\subfloat{\includegraphics[height=0.24\textheight]{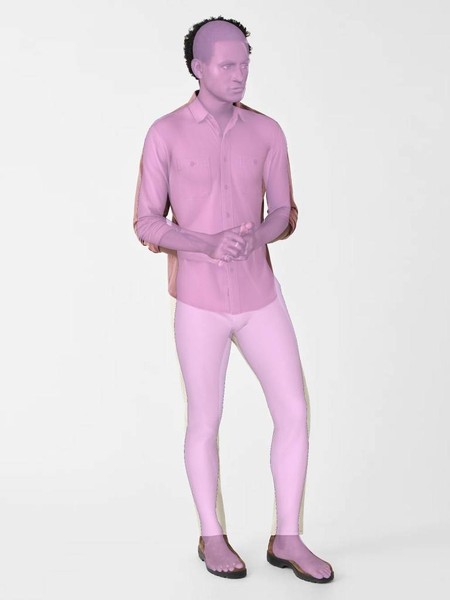}}

\subfloat{\includegraphics[height=0.24\textheight]{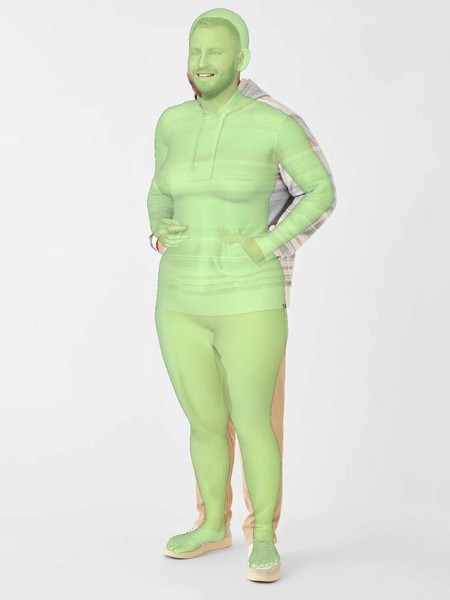}}
\subfloat{\includegraphics[height=0.24\textheight]{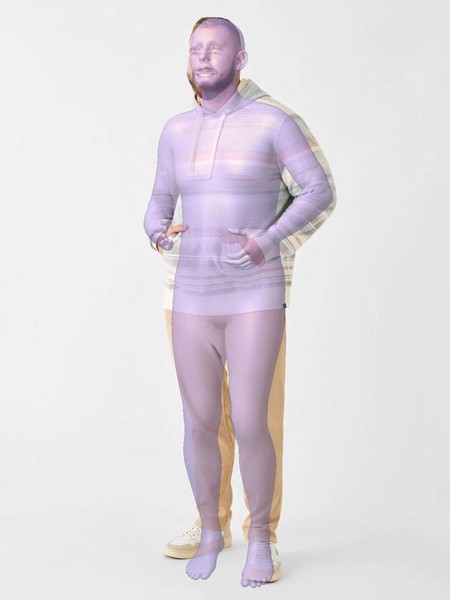}}
\subfloat{\includegraphics[height=0.24\textheight]{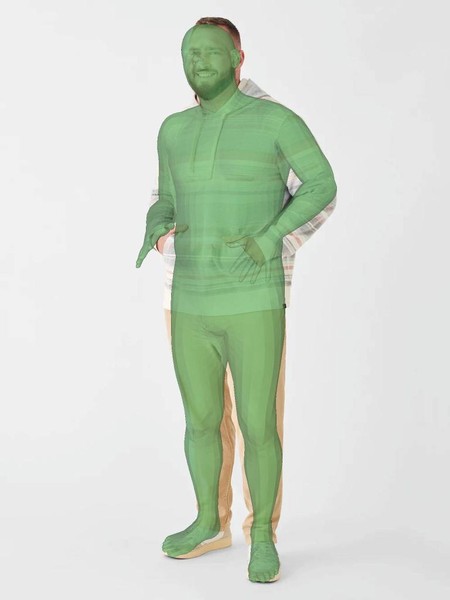}}
\subfloat{\includegraphics[height=0.24\textheight]{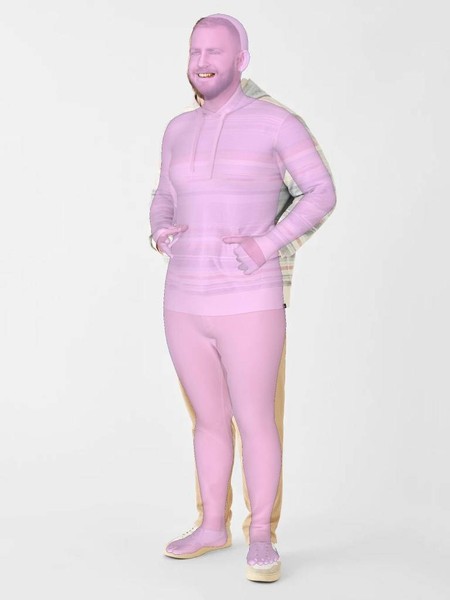}}

\subfloat{\includegraphics[height=0.24\textheight]{images/results/full/faherty/smplifyx_014.jpg}}
\subfloat{\includegraphics[height=0.24\textheight]{images/results/full/faherty/pymafx_014.jpg}}
\subfloat{\includegraphics[height=0.24\textheight]{images/results/full/faherty/shapy_014.jpg}}
\subfloat{\includegraphics[height=0.24\textheight]{images/results/full/faherty/kbody_014.jpg}}

\subfloat[SMPLify-X \cite{pavlakos2019expressive}]
{\includegraphics[height=0.24\textheight]{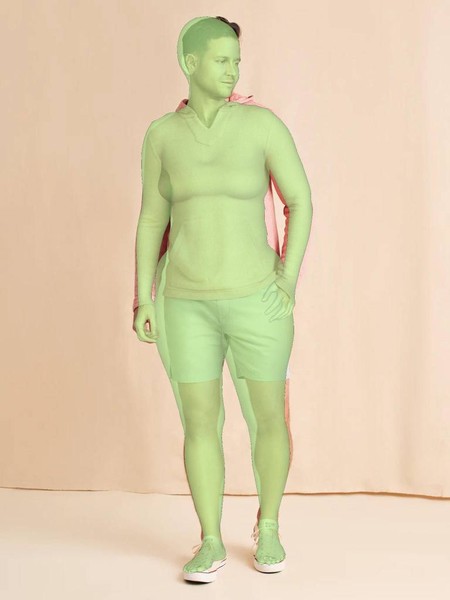}}
\subfloat[PyMAF-X \cite{pymafx2022}]{\includegraphics[height=0.24\textheight]{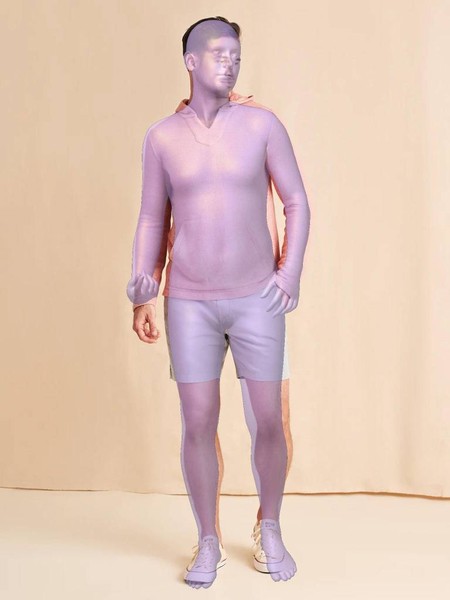}}
\subfloat[SHAPY \cite{choutas2022accurate}]
{\includegraphics[height=0.24\textheight]
{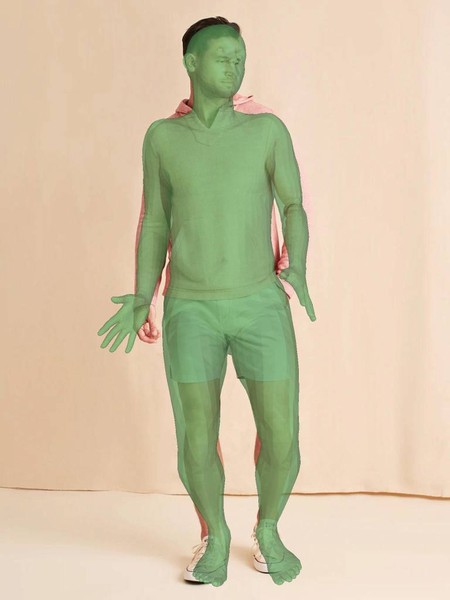}}
\subfloat[\KBody{-.1}{.035} (Ours)]{\includegraphics[height=0.24\textheight]{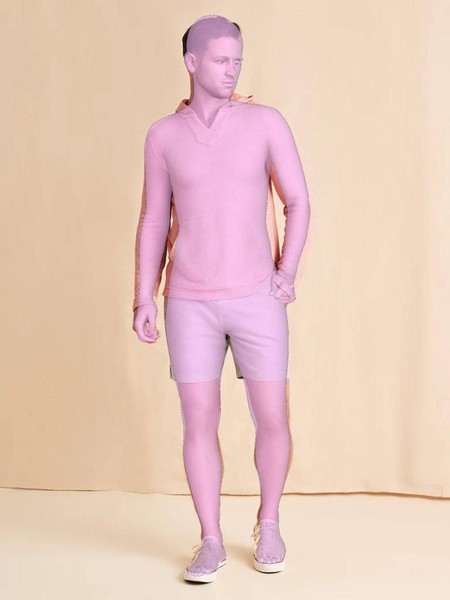}}

\caption{
Left-to-right: SMPLify-X \cite{pavlakos2019expressive} (\textcolor{caribbeangreen2}{light green}), PyMAF-X \cite{pymafx2022} (\textcolor{violet}{purple}), SHAPY \cite{choutas2022accurate} (\textcolor{jade}{green}) and KBody (\textcolor{candypink}{pink}).
}
\label{fig:fh4}
\end{figure*}

%% file: figures/supp/faherty_good_american.tex
\begin{figure*}[!htbp]
\captionsetup[subfigure]{position=bottom,labelformat=empty}

\centering

\subfloat{\includegraphics[width=0.205\textwidth]{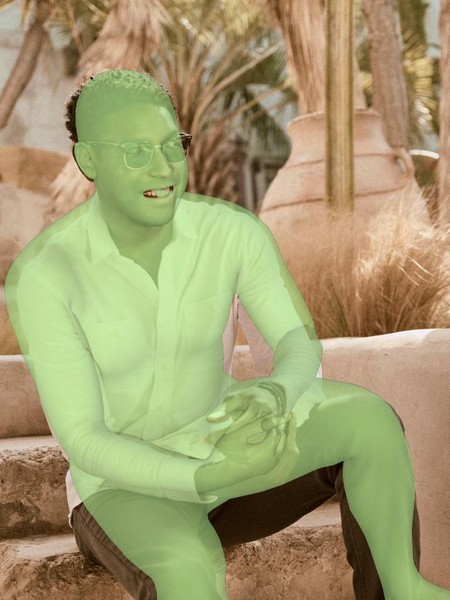}}
\subfloat{\includegraphics[width=0.205\textwidth]{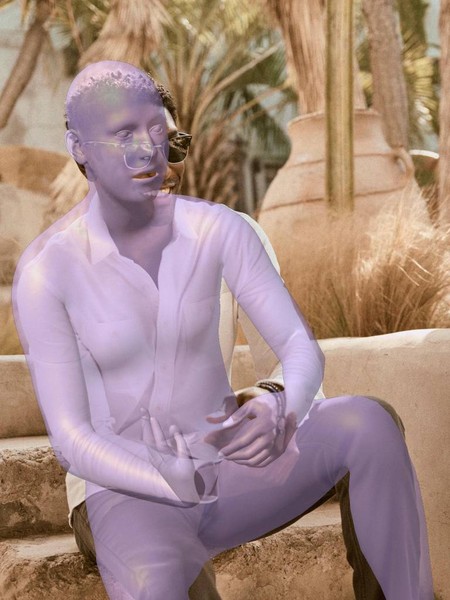}}
\subfloat{\includegraphics[width=0.205\textwidth]{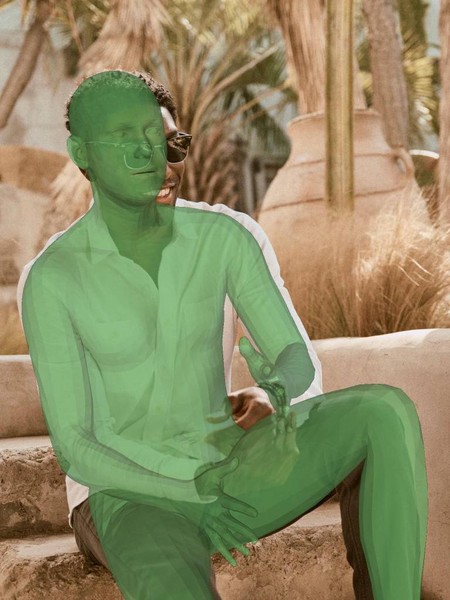}}
\subfloat{\includegraphics[width=0.205\textwidth]{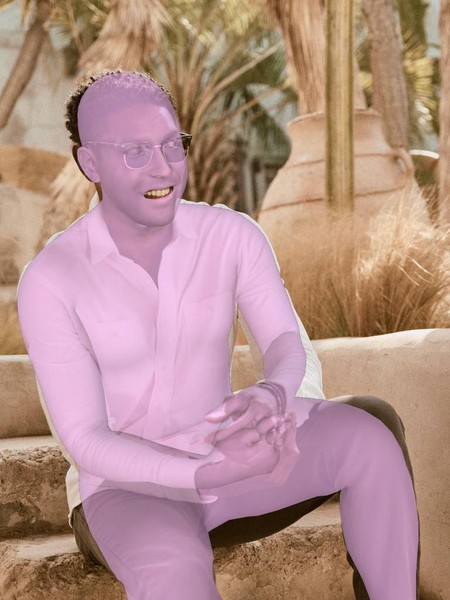}}

\subfloat{\includegraphics[width=0.205\textwidth]{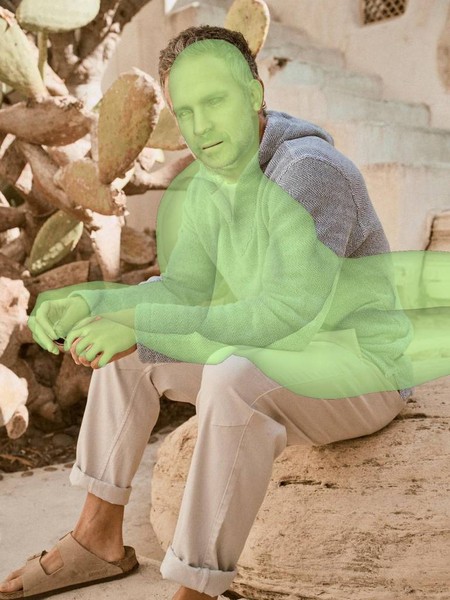}}
\subfloat{\includegraphics[width=0.205\textwidth]{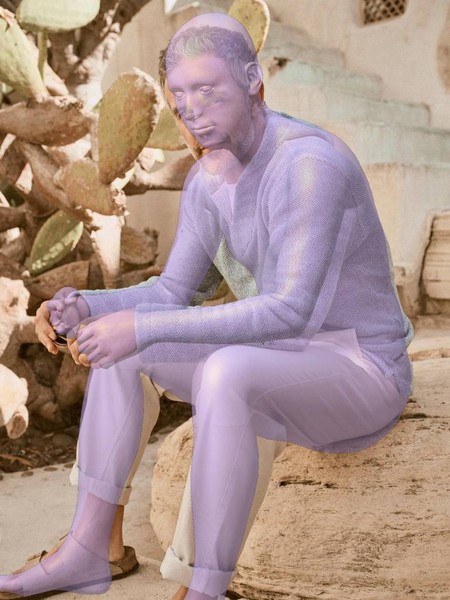}}
\subfloat{\includegraphics[width=0.205\textwidth]{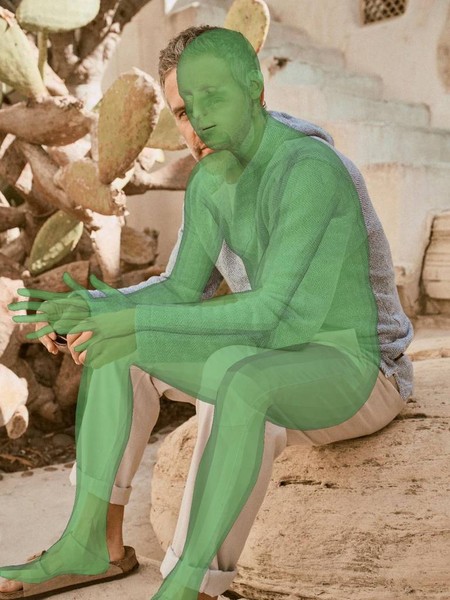}}
\subfloat{\includegraphics[width=0.205\textwidth]{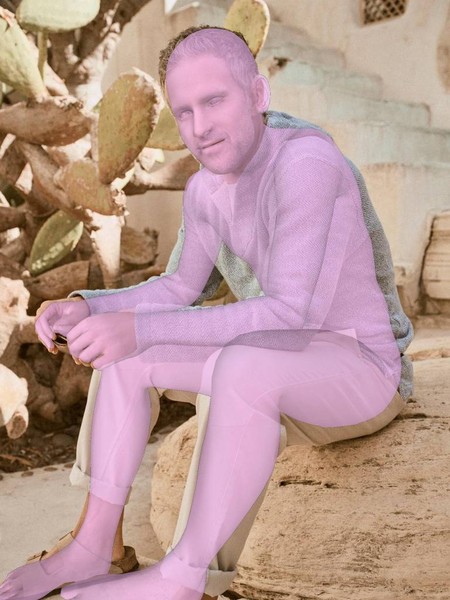}}

\subfloat{\includegraphics[height=0.24\textheight]{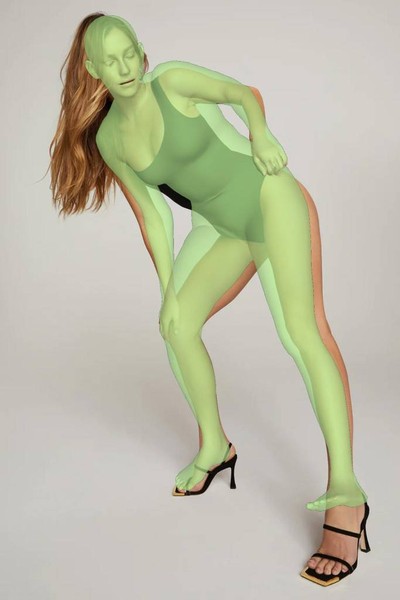}}
\subfloat{\includegraphics[height=0.24\textheight]{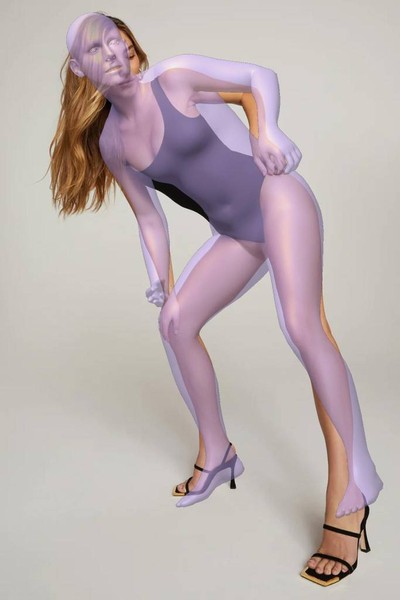}}
\subfloat{\includegraphics[height=0.24\textheight]{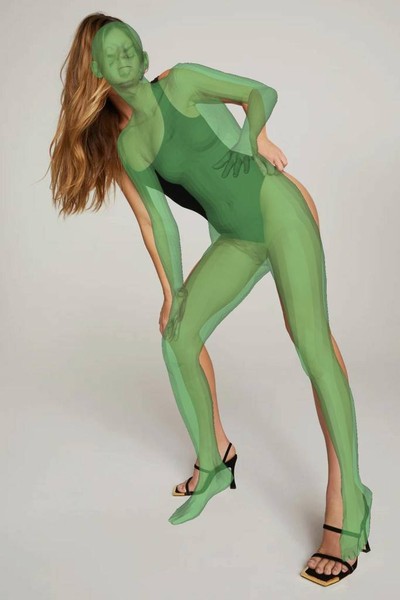}}
\subfloat{\includegraphics[height=0.24\textheight]{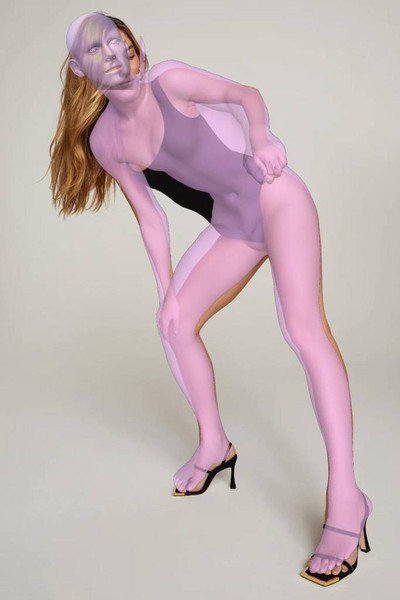}}

\subfloat[SMPLify-X \cite{pavlakos2019expressive}]
{\includegraphics[height=0.24\textheight]{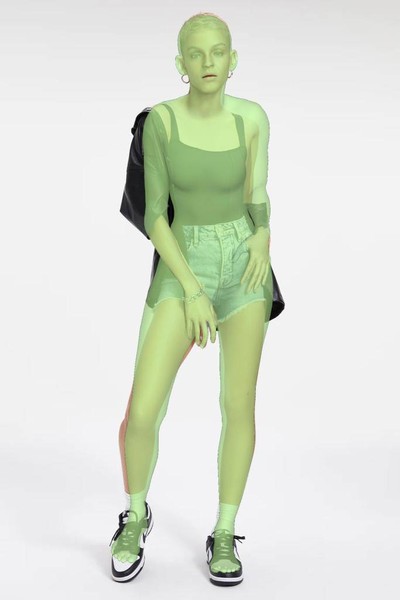}}
\subfloat[PyMAF-X \cite{pymafx2022}]{\includegraphics[height=0.24\textheight]{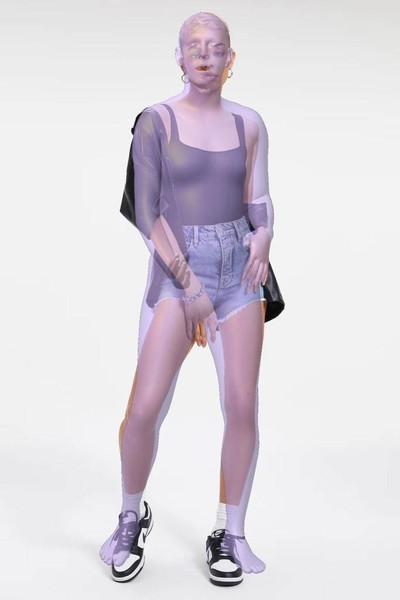}}
\subfloat[SHAPY \cite{choutas2022accurate}]
{\includegraphics[height=0.24\textheight]
{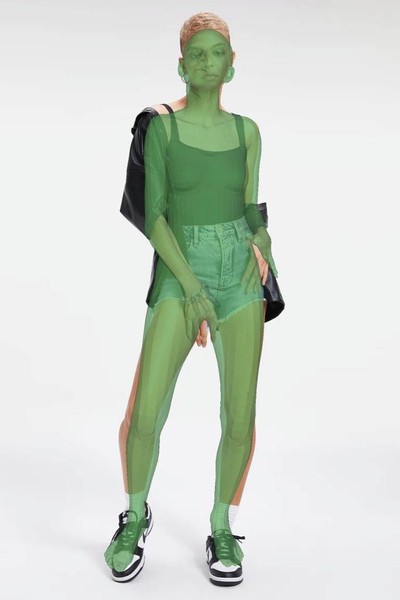}}
\subfloat[\KBody{-.1}{.035} (Ours)]{\includegraphics[height=0.24\textheight]{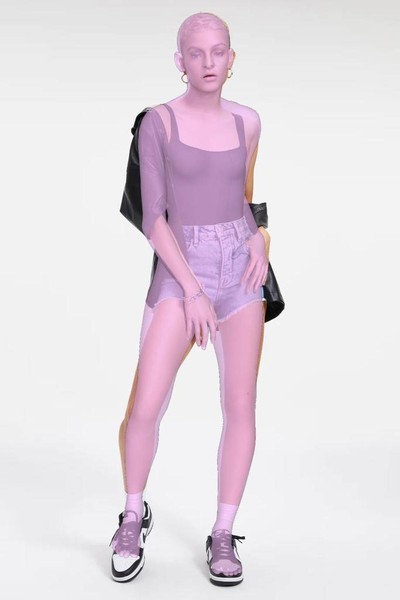}}

\caption{
Left-to-right: SMPLify-X \cite{pavlakos2019expressive} (\textcolor{caribbeangreen2}{light green}), PyMAF-X \cite{pymafx2022} (\textcolor{violet}{purple}), SHAPY \cite{choutas2022accurate} (\textcolor{jade}{green}) and KBody (\textcolor{candypink}{pink}).
}
\label{fig:fh_good}
\end{figure*}

%% file: figures/supp/good_american.tex
\begin{figure*}[!htbp]
\captionsetup[subfigure]{position=bottom,labelformat=empty}

\centering

\subfloat{\includegraphics[height=0.24\textheight]{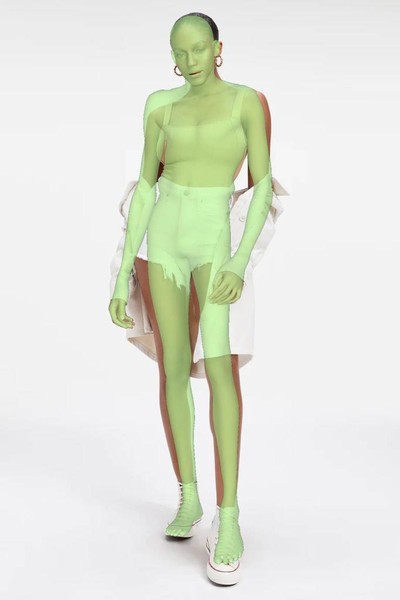}}
\subfloat{\includegraphics[height=0.24\textheight]{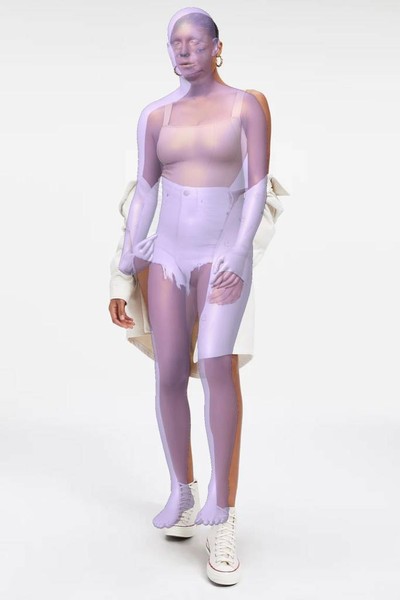}}
\subfloat{\includegraphics[height=0.24\textheight]{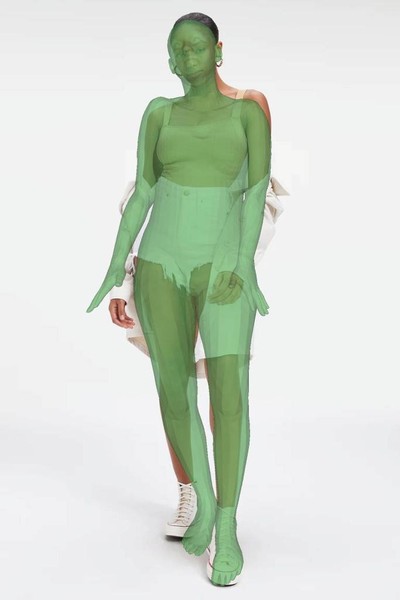}}
\subfloat{\includegraphics[height=0.24\textheight]{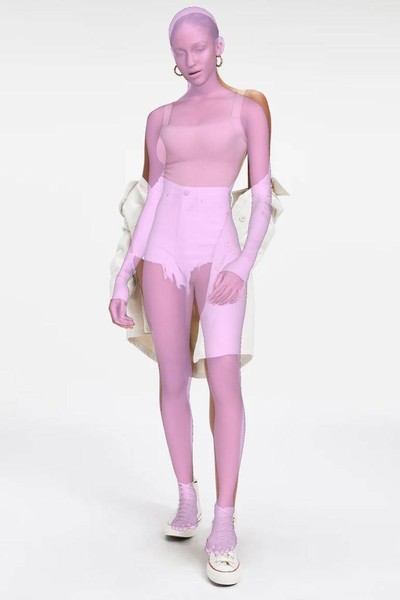}}

\subfloat{\includegraphics[height=0.24\textheight]{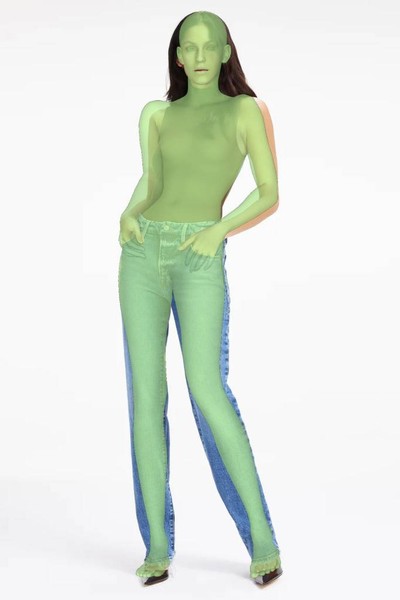}}
\subfloat{\includegraphics[height=0.24\textheight]{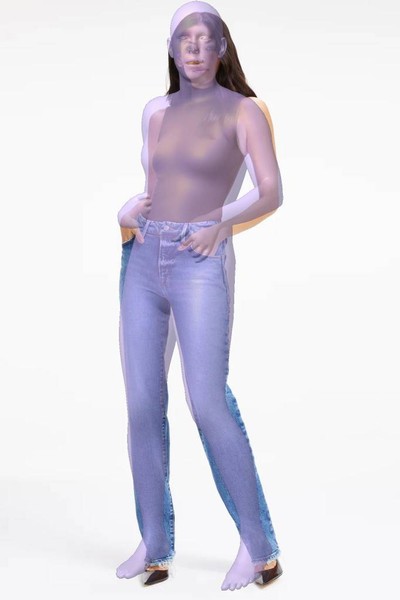}}
\subfloat{\includegraphics[height=0.24\textheight]{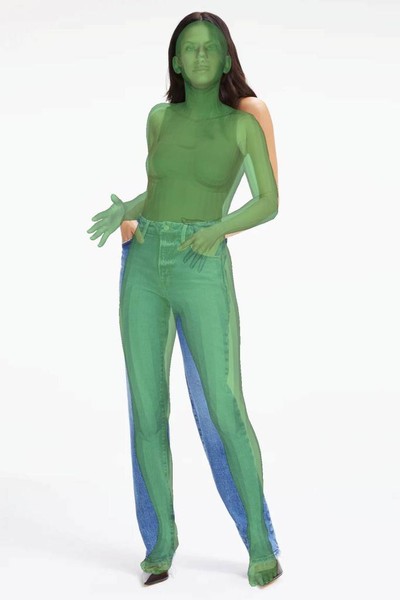}}
\subfloat{\includegraphics[height=0.24\textheight]{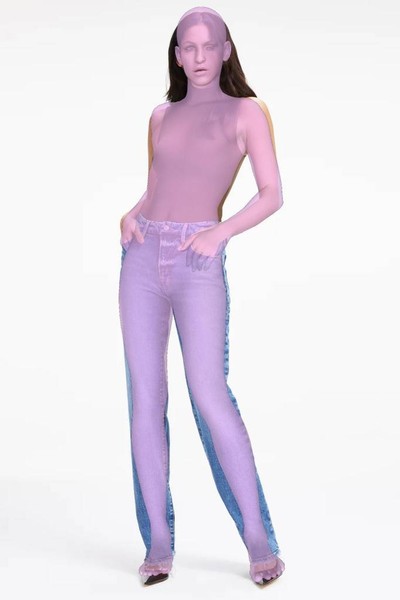}}

\subfloat{\includegraphics[height=0.24\textheight]{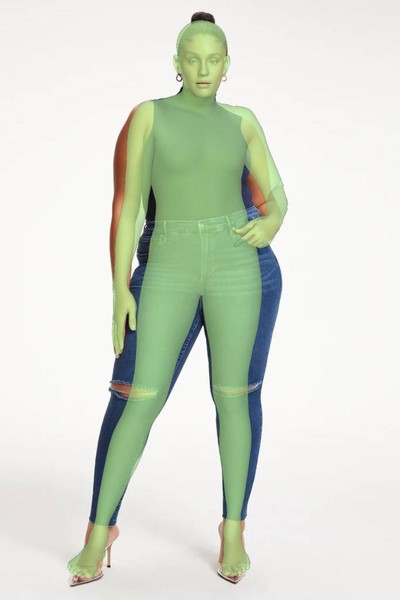}}
\subfloat{\includegraphics[height=0.24\textheight]{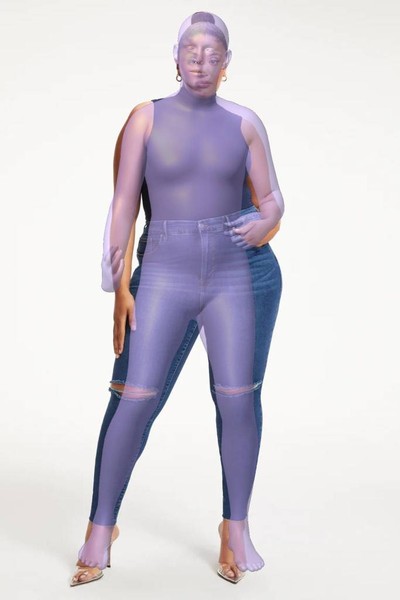}}
\subfloat{\includegraphics[height=0.24\textheight]{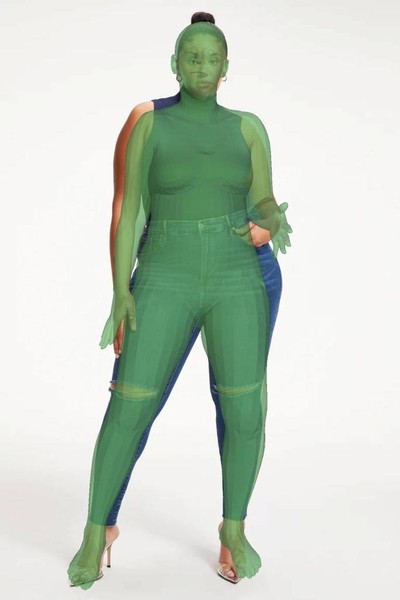}}
\subfloat{\includegraphics[height=0.24\textheight]{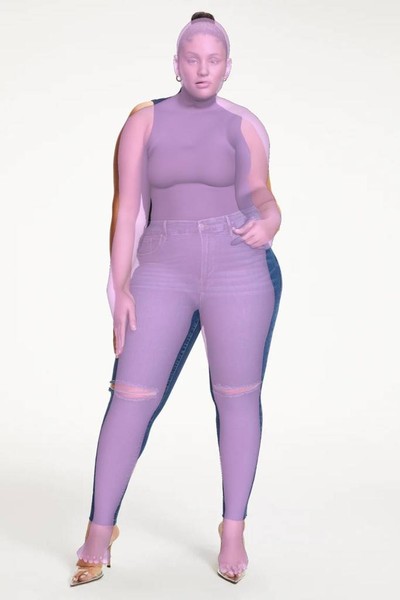}}

\subfloat[SMPLify-X \cite{pavlakos2019expressive}]
{\includegraphics[height=0.24\textheight]{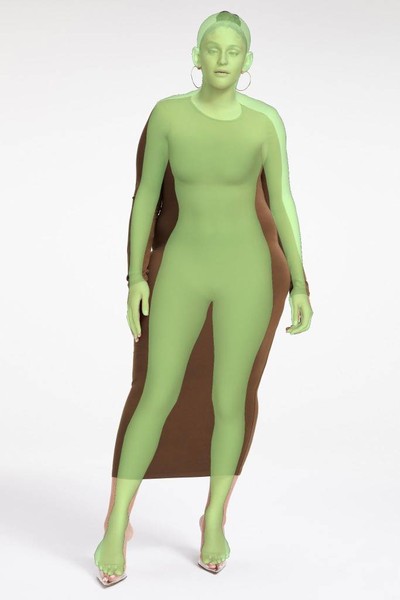}}
\subfloat[PyMAF-X \cite{pymafx2022}]{\includegraphics[height=0.24\textheight]{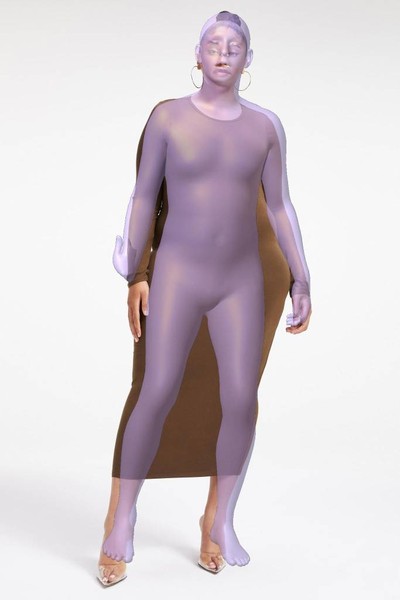}}
\subfloat[SHAPY \cite{choutas2022accurate}]
{\includegraphics[height=0.24\textheight]
{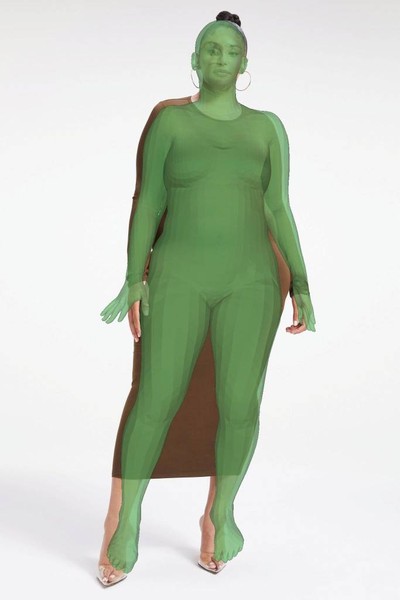}}
\subfloat[\KBody{-.1}{.035} (Ours)]{\includegraphics[height=0.24\textheight]{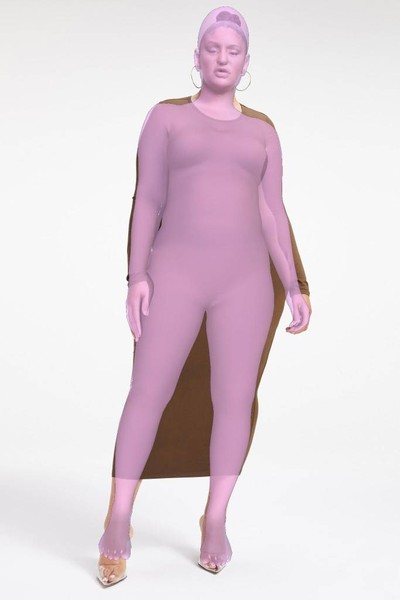}}

\caption{
Left-to-right: SMPLify-X \cite{pavlakos2019expressive} (\textcolor{caribbeangreen2}{light green}), PyMAF-X \cite{pymafx2022} (\textcolor{violet}{purple}), SHAPY \cite{choutas2022accurate} (\textcolor{jade}{green}) and KBody (\textcolor{candypink}{pink}).
}
\label{fig:good}
\end{figure*}

%% file: figures/supp/hm1.tex
\begin{figure*}[!htbp]
\captionsetup[subfigure]{position=bottom,labelformat=empty}

\centering

\subfloat{\includegraphics[height=0.24\textheight]{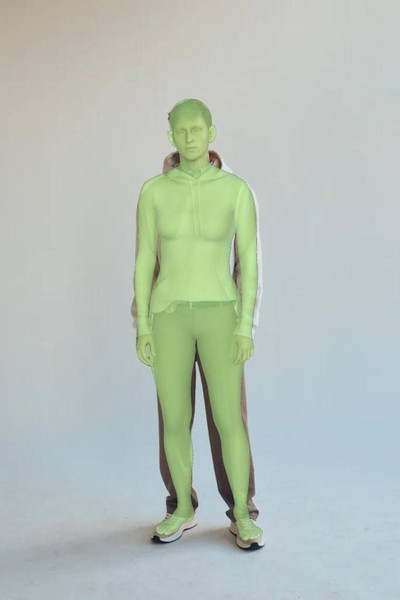}}
\subfloat{\includegraphics[height=0.24\textheight]{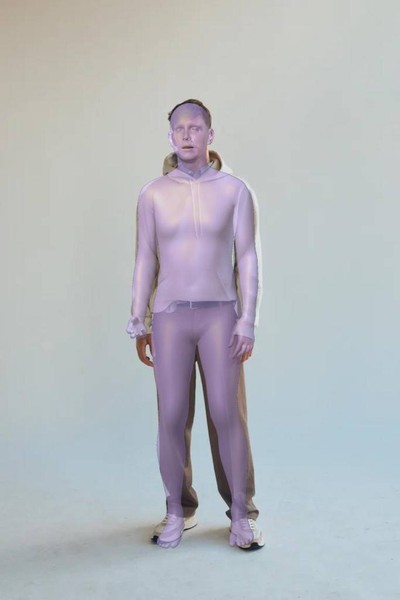}}
\subfloat{\includegraphics[height=0.24\textheight]{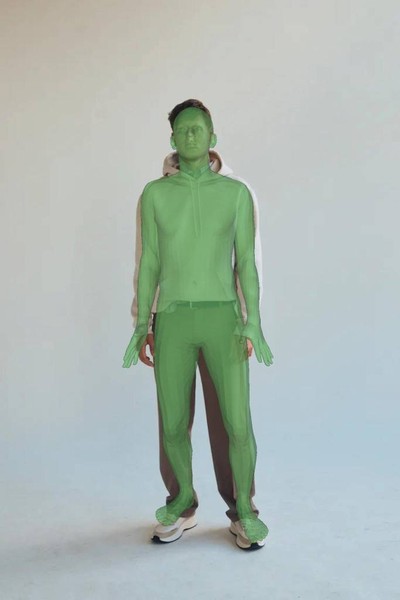}}
\subfloat{\includegraphics[height=0.24\textheight]{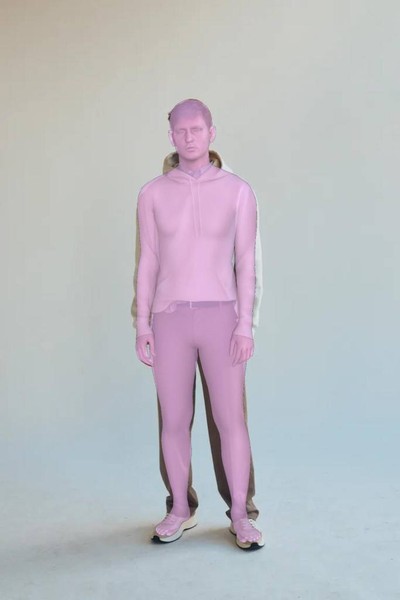}}

\subfloat{\includegraphics[height=0.24\textheight]{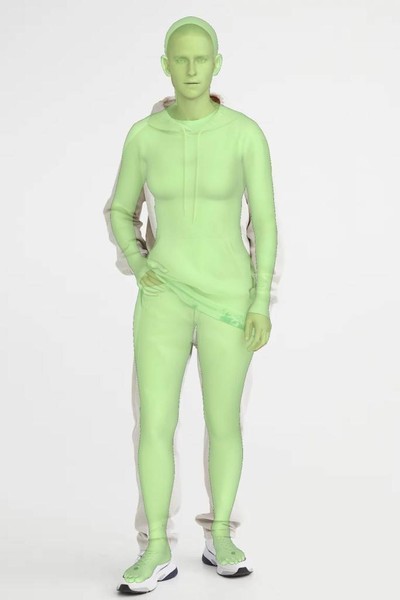}}
\subfloat{\includegraphics[height=0.24\textheight]{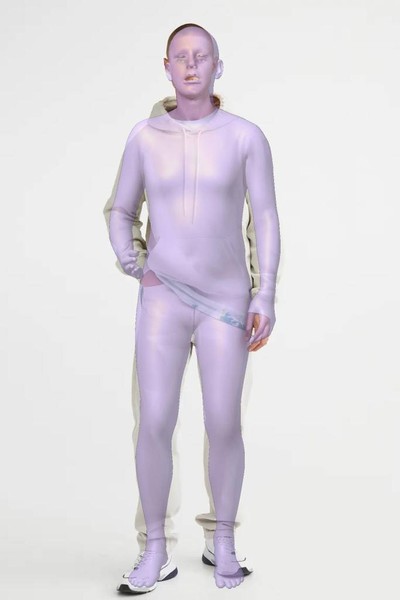}}
\subfloat{\includegraphics[height=0.24\textheight]{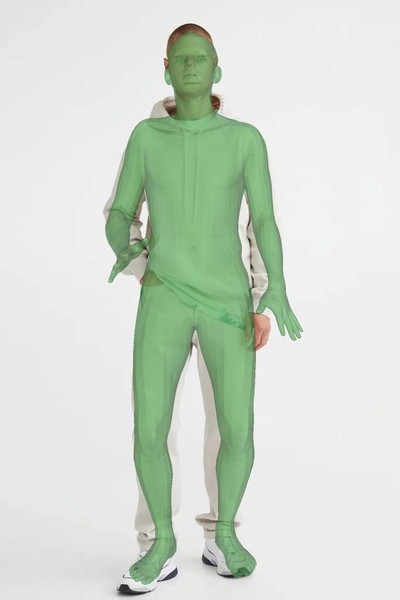}}
\subfloat{\includegraphics[height=0.24\textheight]{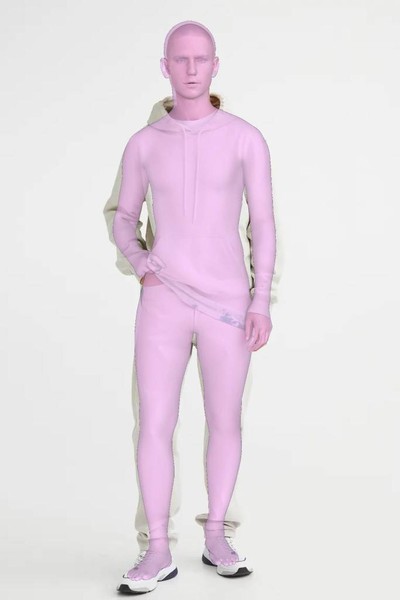}}

\subfloat{\includegraphics[height=0.24\textheight]{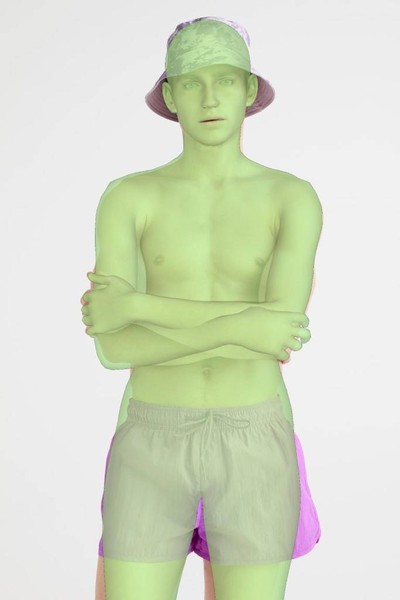}}
\subfloat{\includegraphics[height=0.24\textheight]{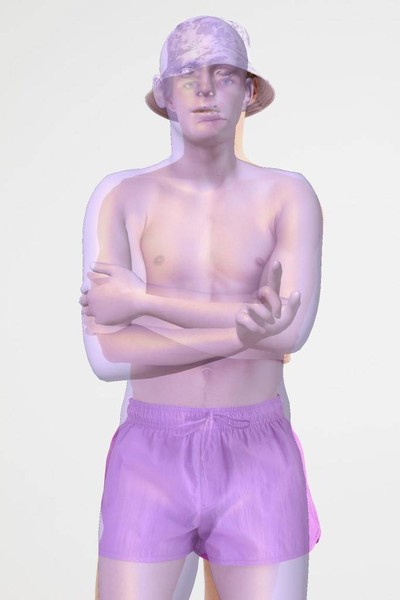}}
\subfloat{\includegraphics[height=0.24\textheight]{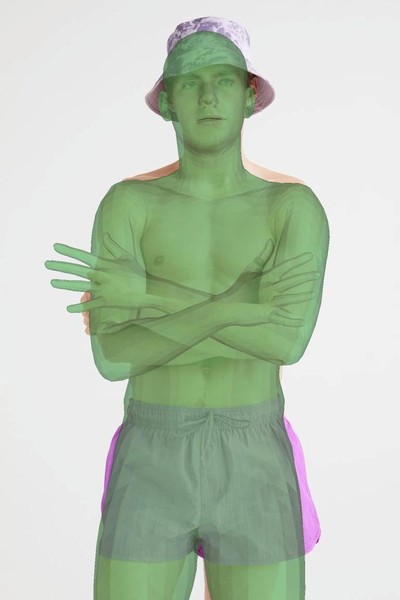}}
\subfloat{\includegraphics[height=0.24\textheight]{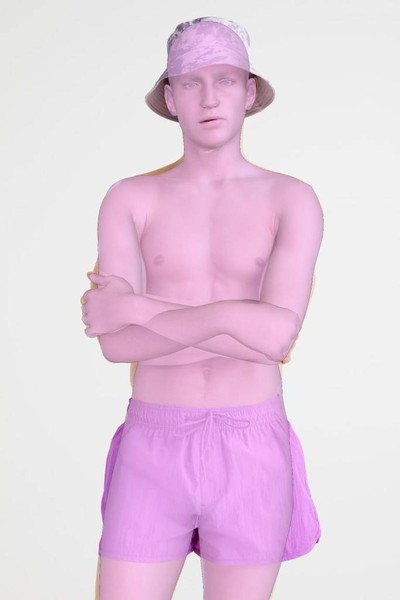}}

\subfloat[SMPLify-X \cite{pavlakos2019expressive}]
{\includegraphics[height=0.24\textheight]{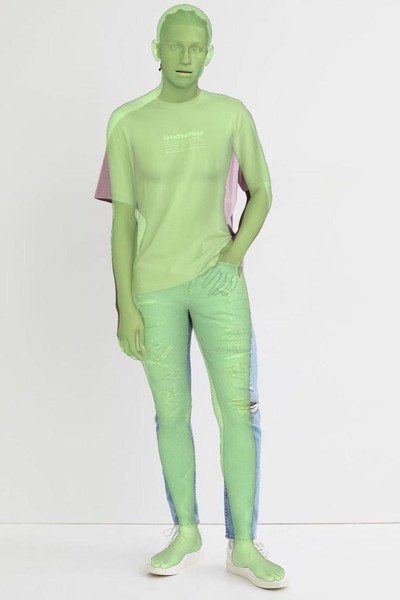}}
\subfloat[PyMAF-X \cite{pymafx2022}]{\includegraphics[height=0.24\textheight]{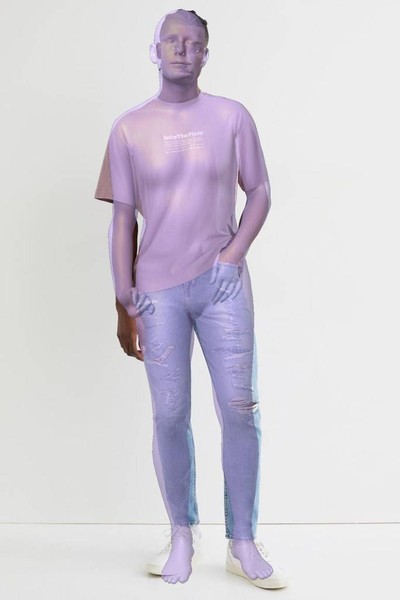}}
\subfloat[SHAPY \cite{choutas2022accurate}]
{\includegraphics[height=0.24\textheight]
{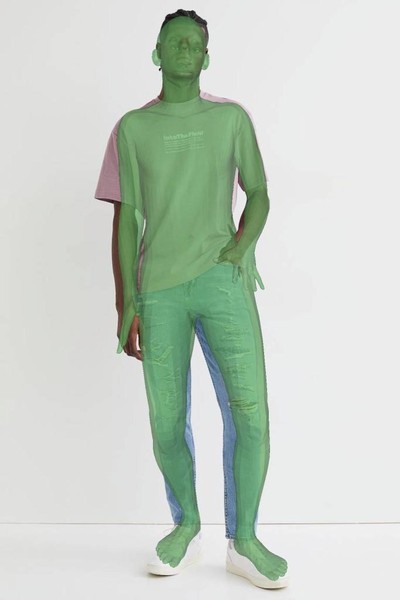}}
\subfloat[\KBody{-.1}{.035} (Ours)]{\includegraphics[height=0.24\textheight]{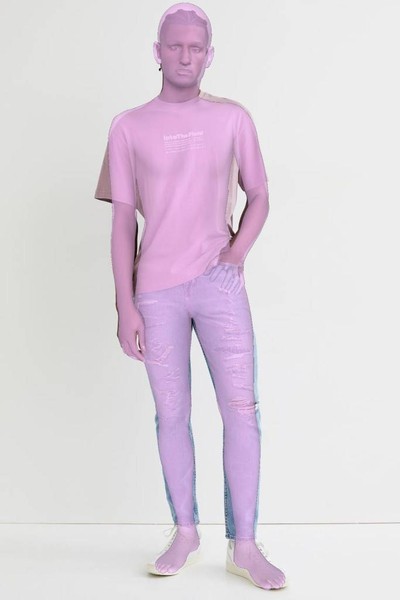}}

\caption{
Left-to-right: SMPLify-X \cite{pavlakos2019expressive} (\textcolor{caribbeangreen2}{light green}), PyMAF-X \cite{pymafx2022} (\textcolor{violet}{purple}), SHAPY \cite{choutas2022accurate} (\textcolor{jade}{green}) and KBody (\textcolor{candypink}{pink}).
}
\label{fig:hm1}
\end{figure*}

%% file: figures/supp/hm2.tex
\begin{figure*}[!htbp]
\captionsetup[subfigure]{position=bottom,labelformat=empty}

\centering

\subfloat{\includegraphics[height=0.24\textheight]{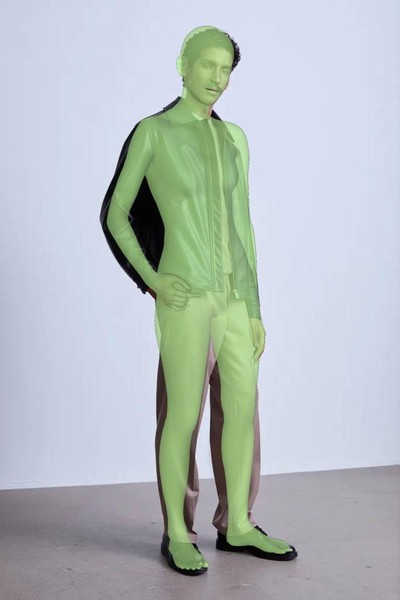}}
\subfloat{\includegraphics[height=0.24\textheight]{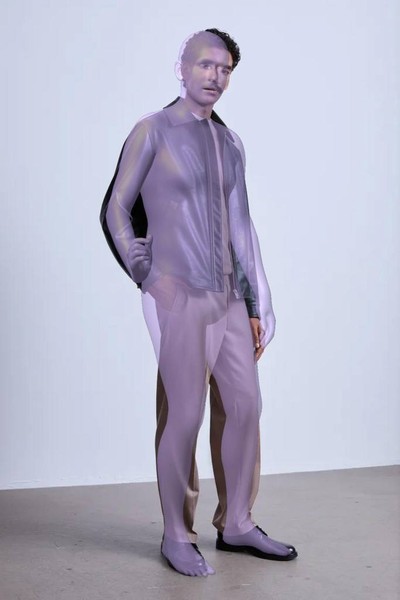}}
\subfloat{\includegraphics[height=0.24\textheight]{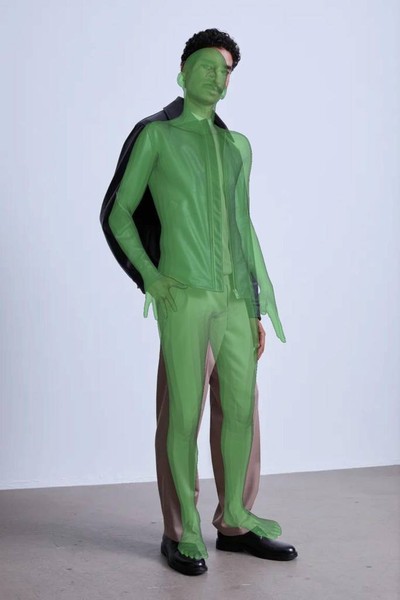}}
\subfloat{\includegraphics[height=0.24\textheight]{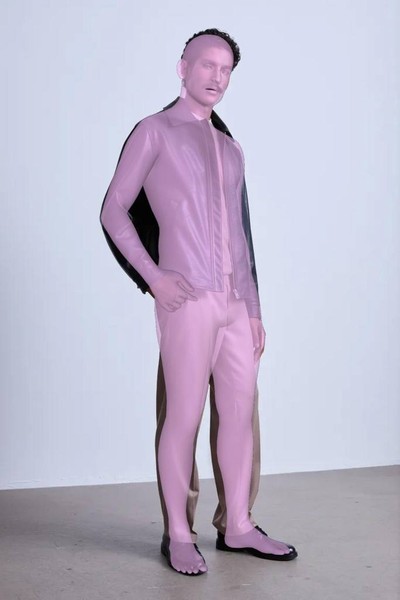}}

\subfloat{\includegraphics[height=0.24\textheight]{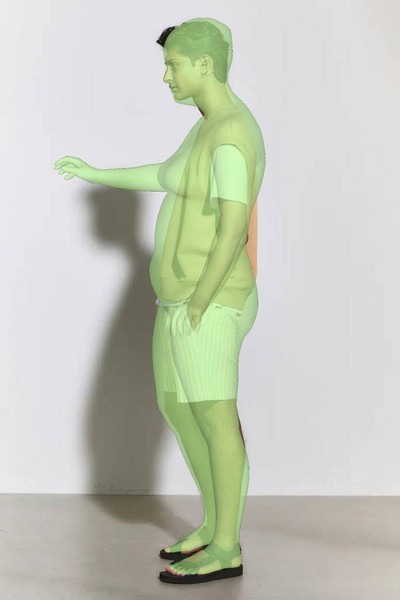}}
\subfloat{\includegraphics[height=0.24\textheight]{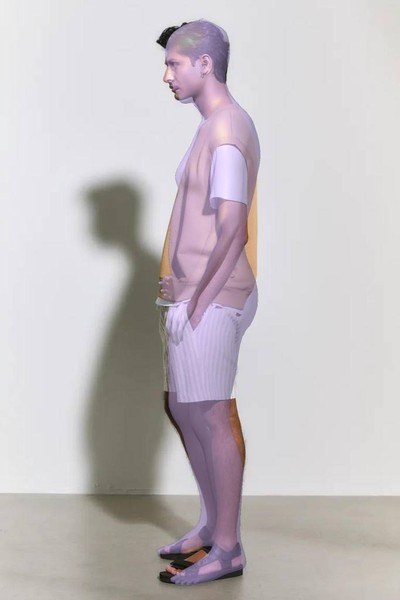}}
\subfloat{\includegraphics[height=0.24\textheight]{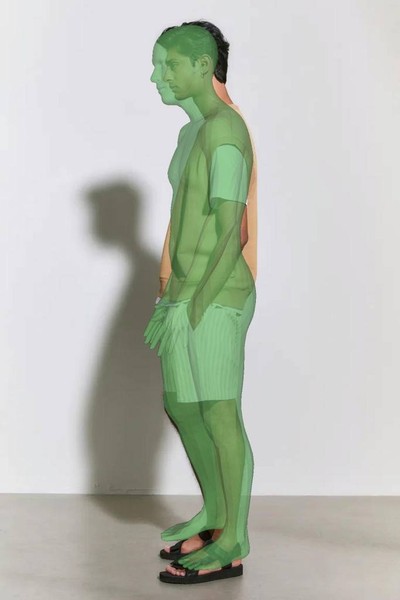}}
\subfloat{\includegraphics[height=0.24\textheight]{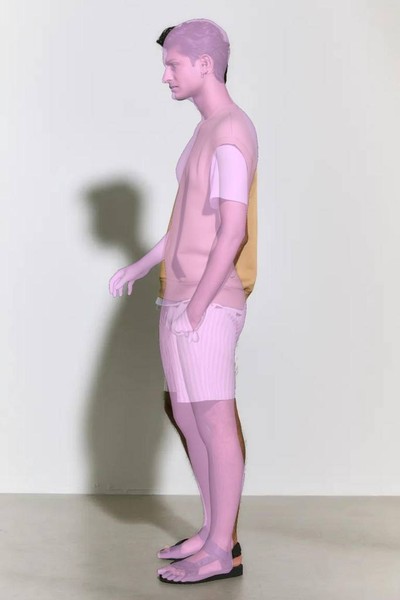}}

\subfloat{\includegraphics[height=0.24\textheight]{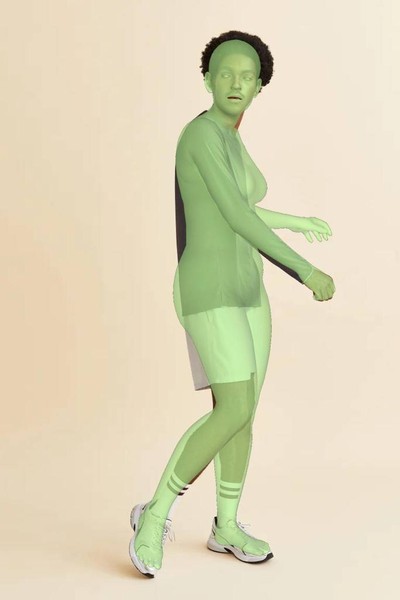}}
\subfloat{\includegraphics[height=0.24\textheight]{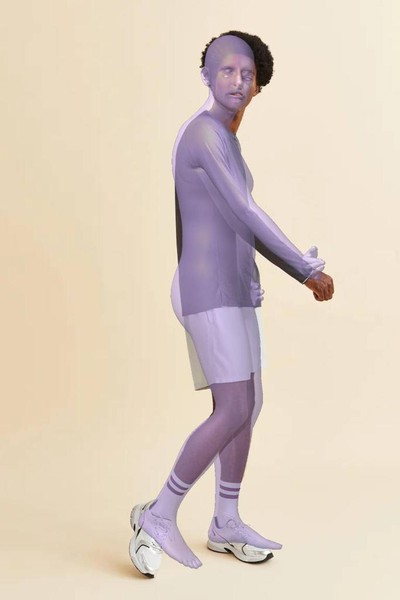}}
\subfloat{\includegraphics[height=0.24\textheight]{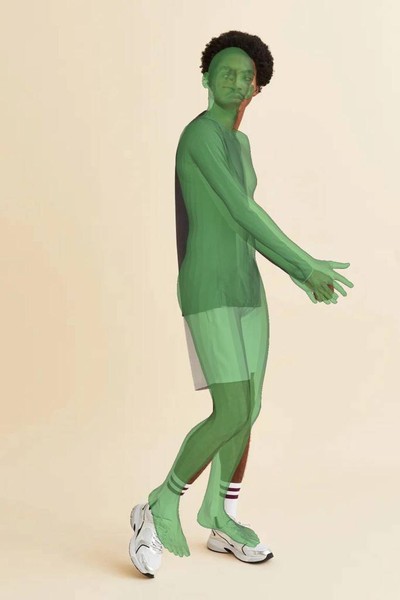}}
\subfloat{\includegraphics[height=0.24\textheight]{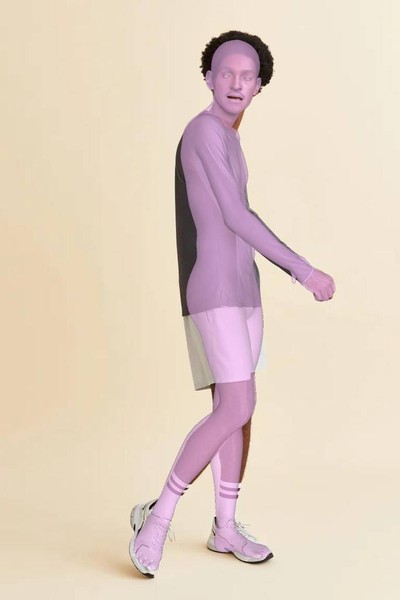}}

\subfloat[SMPLify-X \cite{pavlakos2019expressive}]
{\includegraphics[height=0.24\textheight]{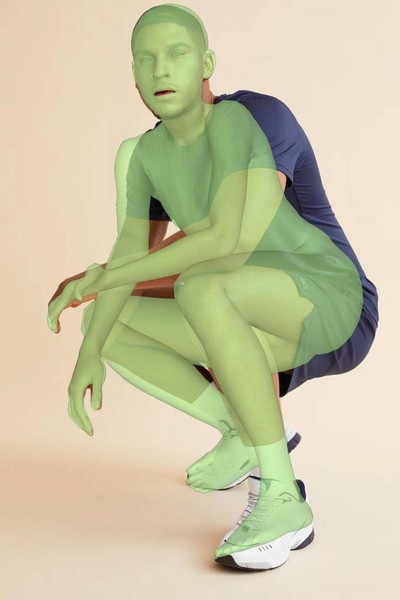}}
\subfloat[PyMAF-X \cite{pymafx2022}]{\includegraphics[height=0.24\textheight]{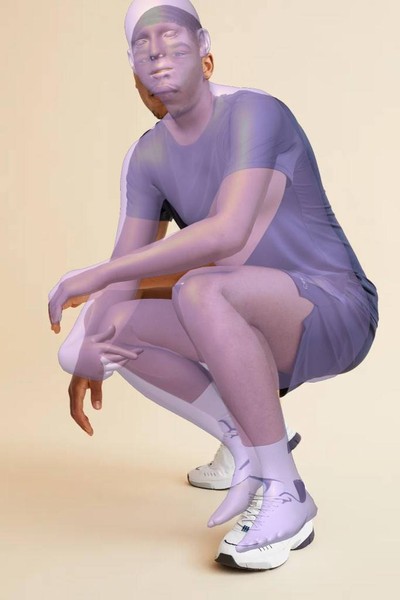}}
\subfloat[SHAPY \cite{choutas2022accurate}]
{\includegraphics[height=0.24\textheight]
{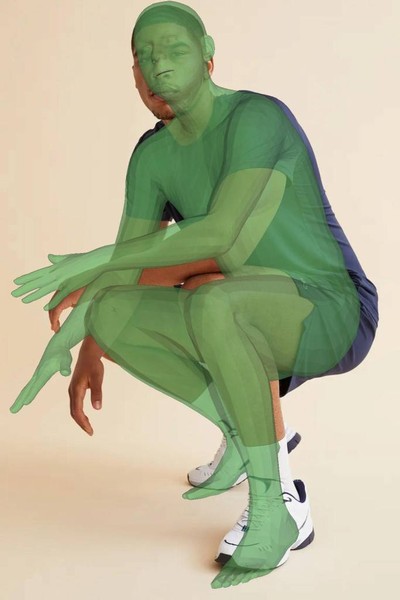}}
\subfloat[\KBody{-.1}{.035} (Ours)]{\includegraphics[height=0.24\textheight]{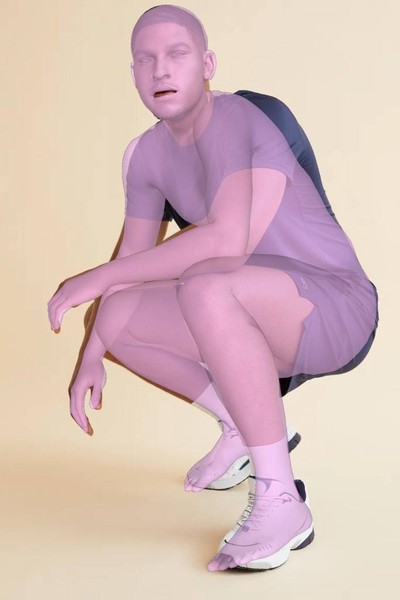}}

\caption{
Left-to-right: SMPLify-X \cite{pavlakos2019expressive} (\textcolor{caribbeangreen2}{light green}), PyMAF-X \cite{pymafx2022} (\textcolor{violet}{purple}), SHAPY \cite{choutas2022accurate} (\textcolor{jade}{green}) and KBody (\textcolor{candypink}{pink}).
}
\label{fig:hm2}
\end{figure*}

%% file: figures/supp/hm3.tex
\begin{figure*}[!htbp]
\captionsetup[subfigure]{position=bottom,labelformat=empty}

\centering

\subfloat{\includegraphics[height=0.24\textheight]{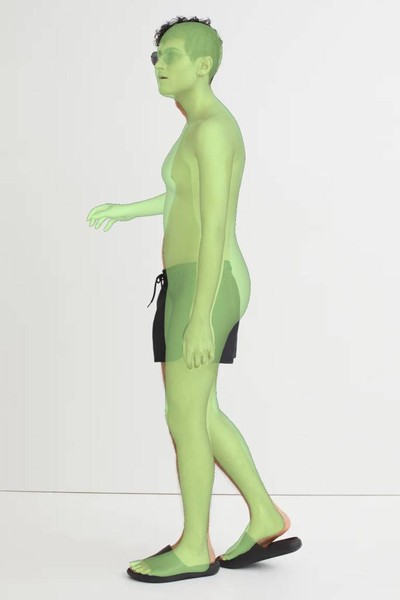}}
\subfloat{\includegraphics[height=0.24\textheight]{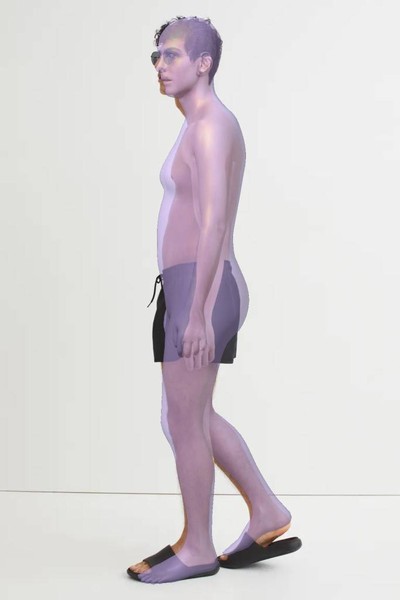}}
\subfloat{\includegraphics[height=0.24\textheight]{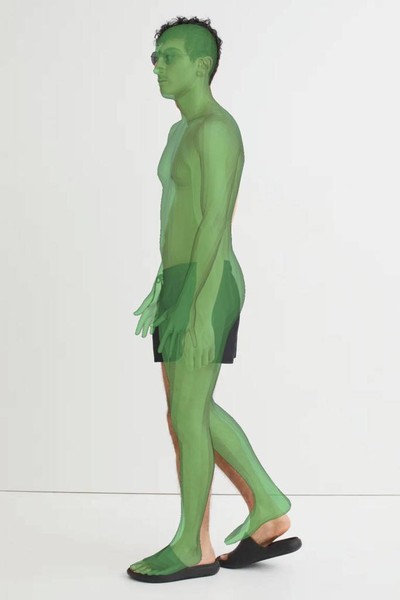}}
\subfloat{\includegraphics[height=0.24\textheight]{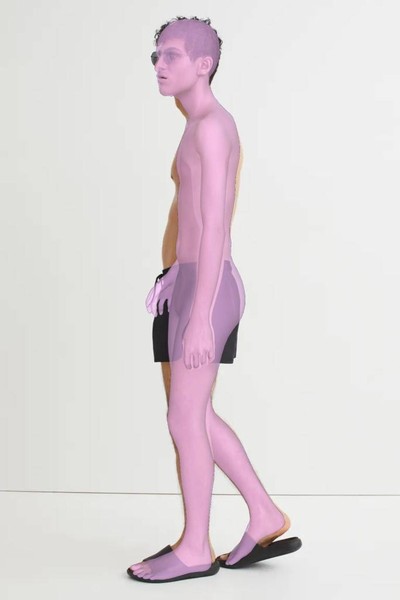}}

\subfloat{\includegraphics[height=0.24\textheight]{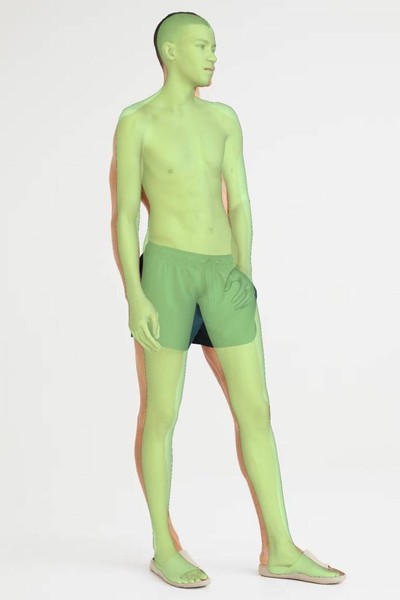}}
\subfloat{\includegraphics[height=0.24\textheight]{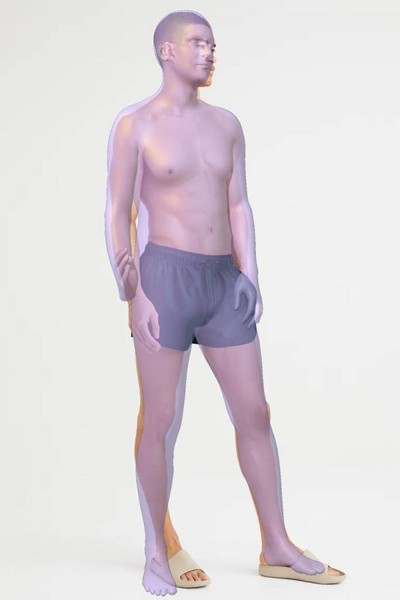}}
\subfloat{\includegraphics[height=0.24\textheight]{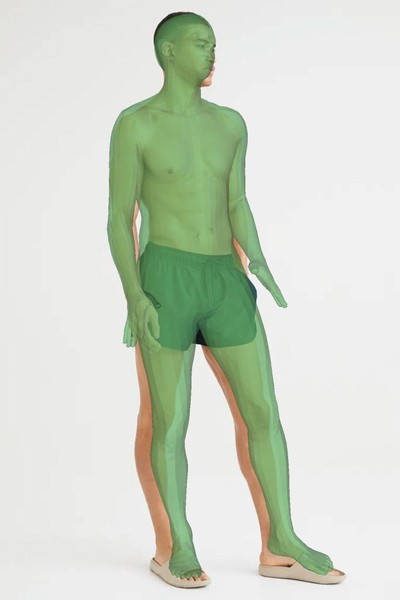}}
\subfloat{\includegraphics[height=0.24\textheight]{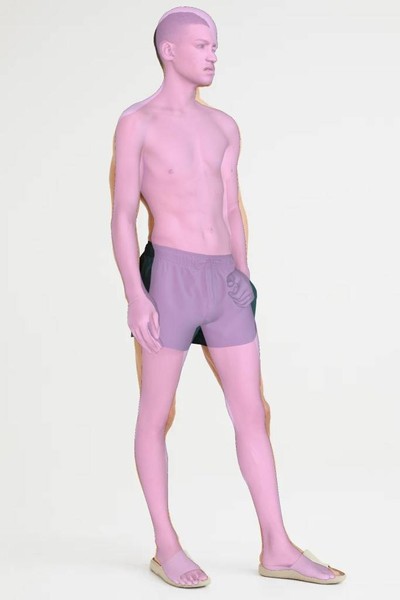}}

\subfloat{\includegraphics[height=0.24\textheight]{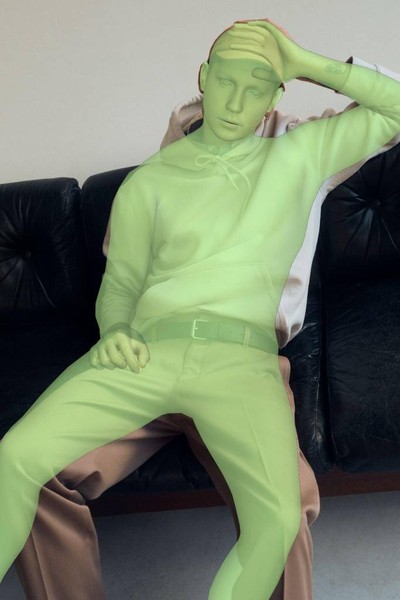}}
\subfloat{\includegraphics[height=0.24\textheight]{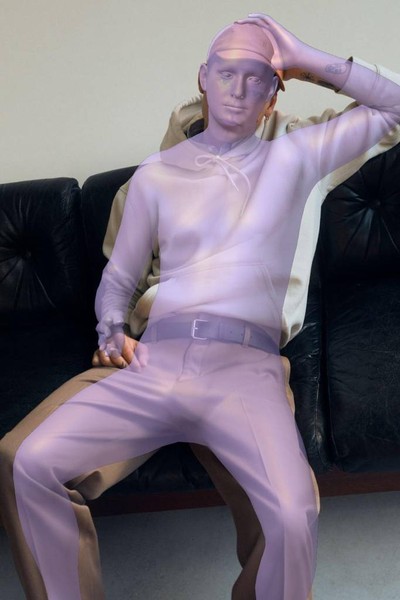}}
\subfloat{\includegraphics[height=0.24\textheight]{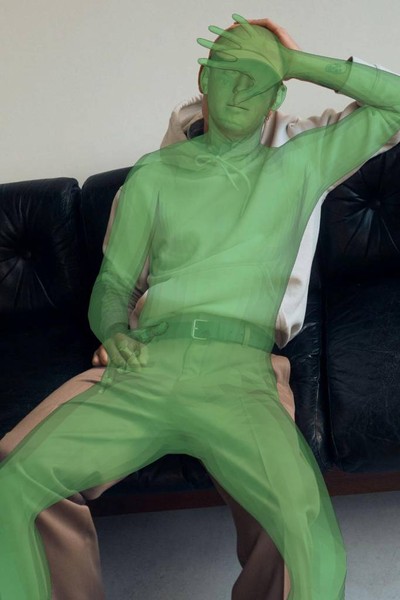}}
\subfloat{\includegraphics[height=0.24\textheight]{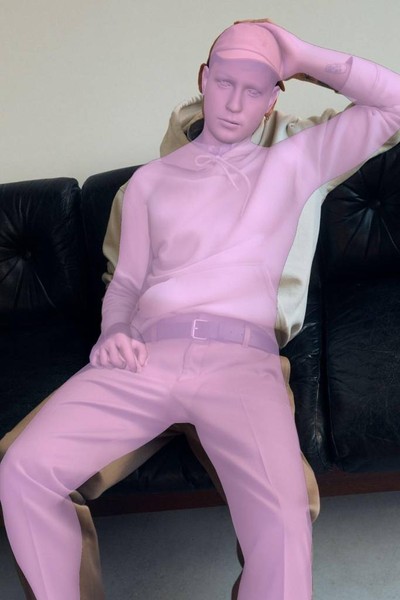}}

\subfloat[SMPLify-X \cite{pavlakos2019expressive}]
{\includegraphics[height=0.24\textheight]{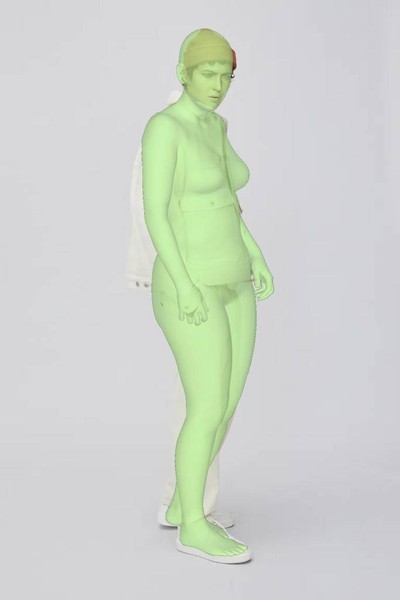}}
\subfloat[PyMAF-X \cite{pymafx2022}]{\includegraphics[height=0.24\textheight]{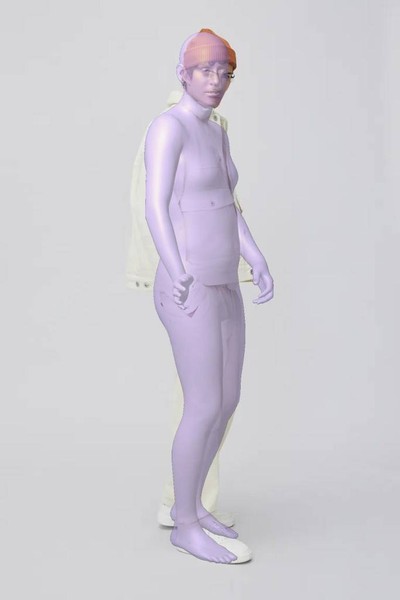}}
\subfloat[SHAPY \cite{choutas2022accurate}]
{\includegraphics[height=0.24\textheight]
{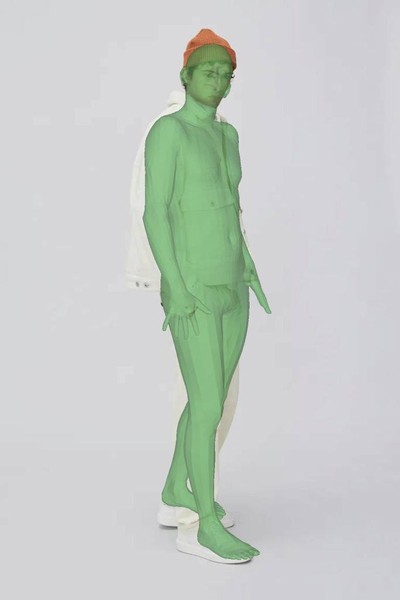}}
\subfloat[\KBody{-.1}{.035} (Ours)]{\includegraphics[height=0.24\textheight]{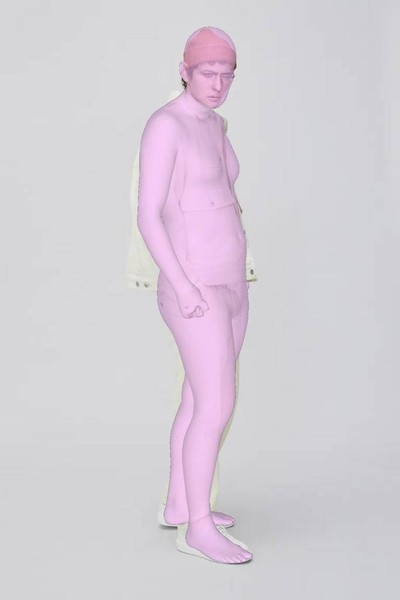}}

\caption{
Left-to-right: SMPLify-X \cite{pavlakos2019expressive} (\textcolor{caribbeangreen2}{light green}), PyMAF-X \cite{pymafx2022} (\textcolor{violet}{purple}), SHAPY \cite{choutas2022accurate} (\textcolor{jade}{green}) and KBody (\textcolor{candypink}{pink}).
}
\label{fig:hm3}
\end{figure*}

%% file: figures/supp/hm_splendid.tex
\begin{figure*}[!htbp]
\captionsetup[subfigure]{position=bottom,labelformat=empty}

\centering

\subfloat{\includegraphics[height=0.24\textheight]{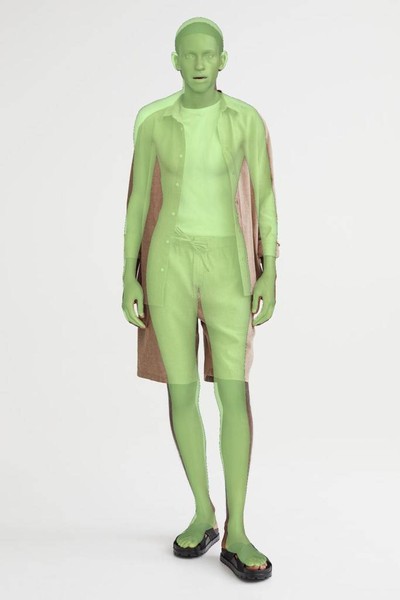}}
\subfloat{\includegraphics[height=0.24\textheight]{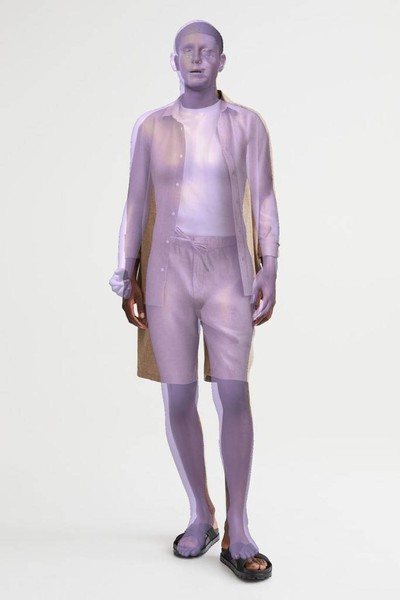}}
\subfloat{\includegraphics[height=0.24\textheight]{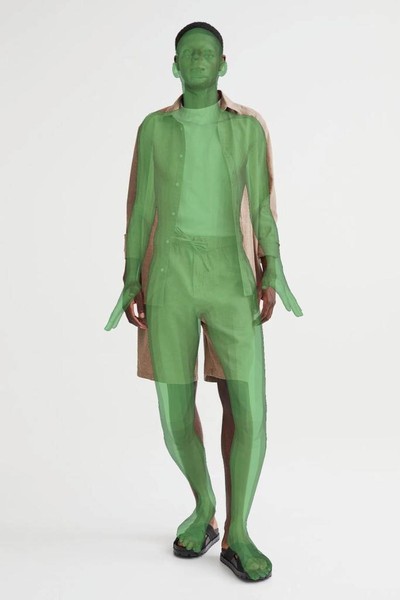}}
\subfloat{\includegraphics[height=0.24\textheight]{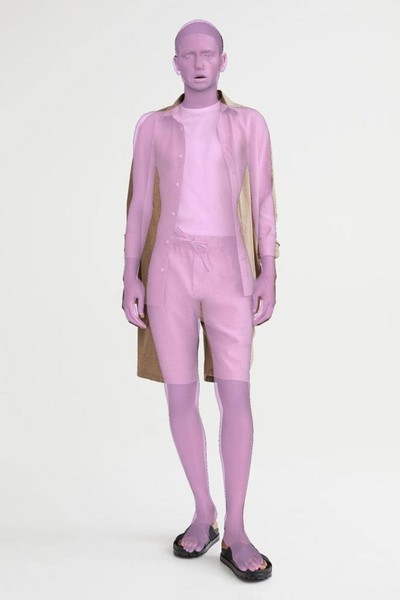}}

\subfloat{\includegraphics[height=0.24\textheight]{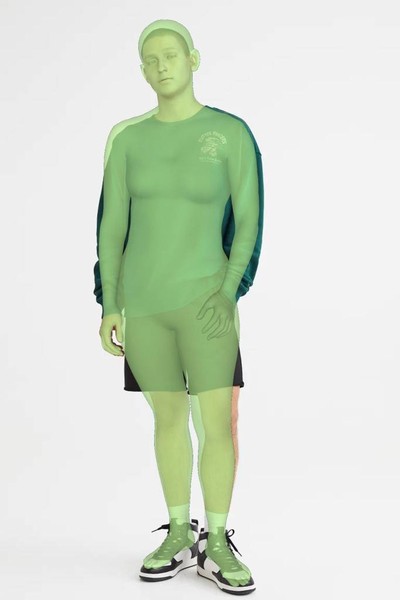}}
\subfloat{\includegraphics[height=0.24\textheight]{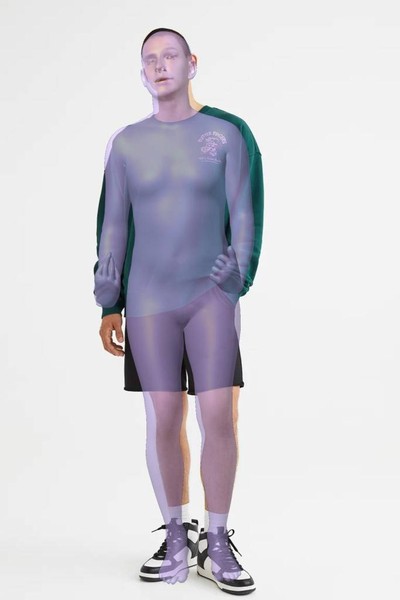}}
\subfloat{\includegraphics[height=0.24\textheight]{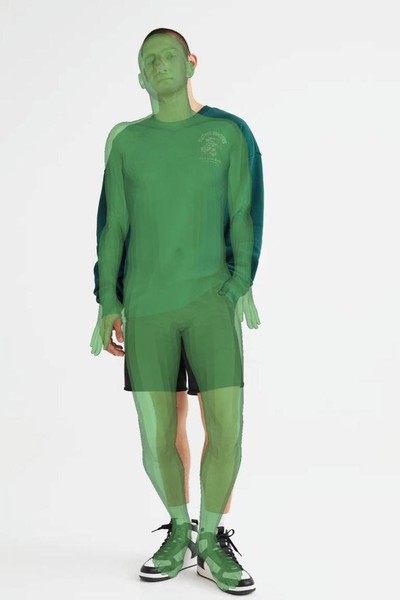}}
\subfloat{\includegraphics[height=0.24\textheight]{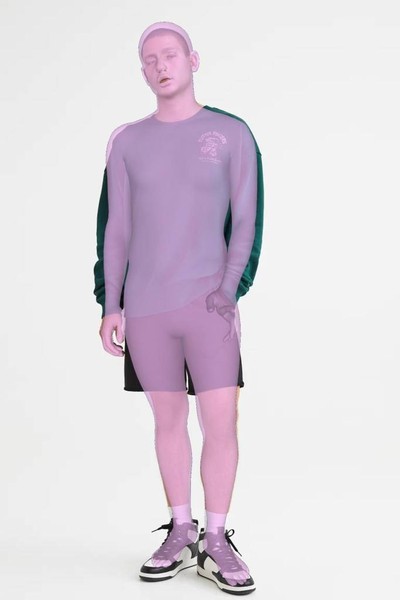}}

\subfloat{\includegraphics[height=0.2\textheight]{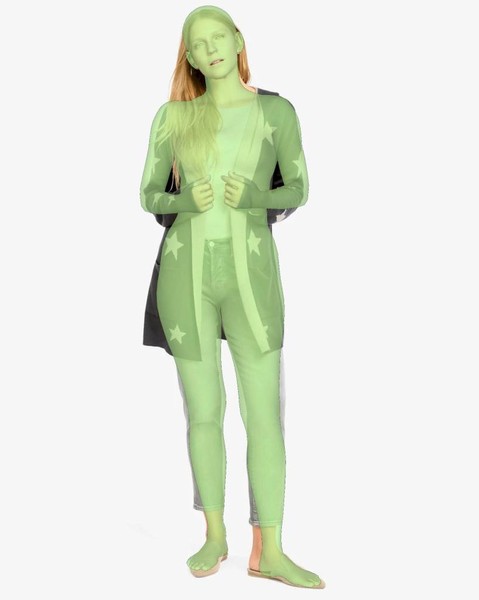}}
\subfloat{\includegraphics[height=0.2\textheight]{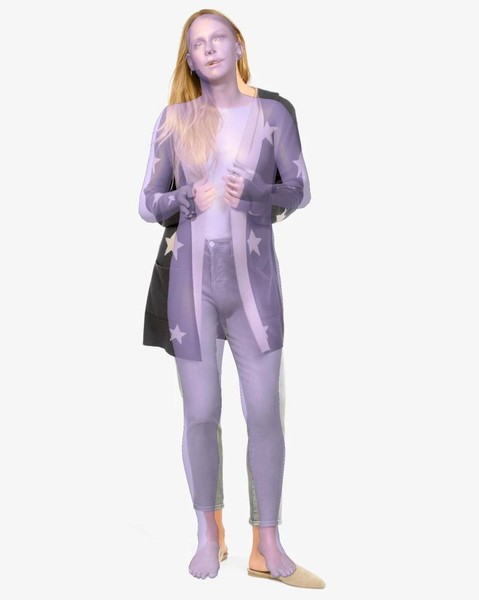}}
\subfloat{\includegraphics[height=0.2\textheight]{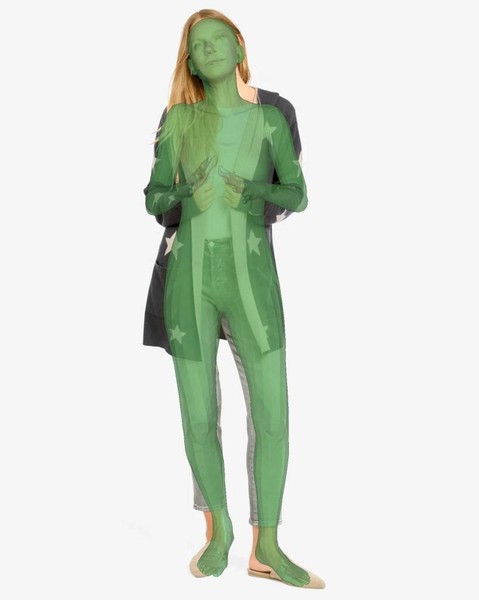}}
\subfloat{\includegraphics[height=0.2\textheight]{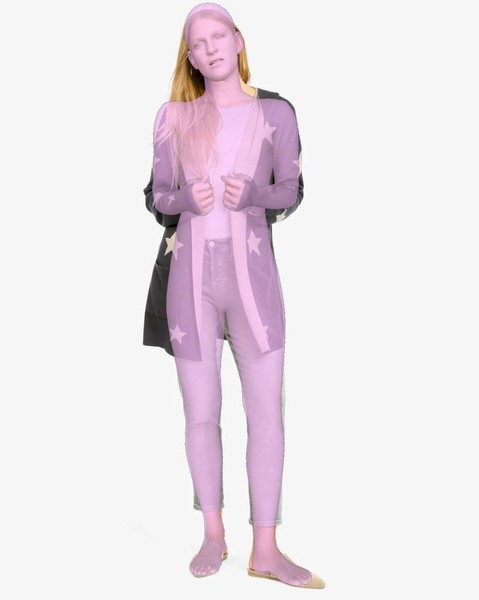}}

\subfloat[SMPLify-X \cite{pavlakos2019expressive}]
{\includegraphics[height=0.2\textheight]{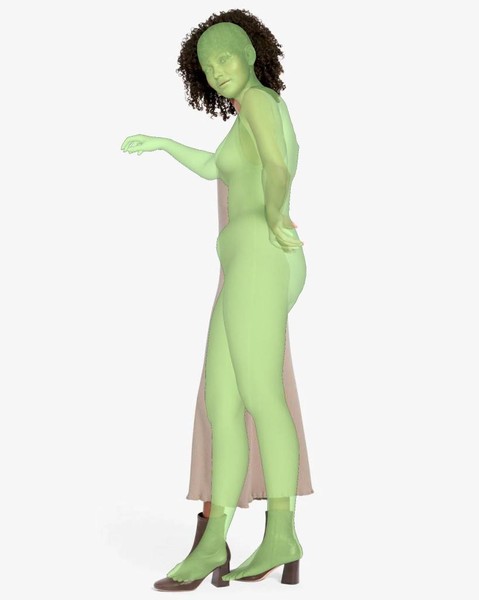}}
\subfloat[PyMAF-X \cite{pymafx2022}]{\includegraphics[height=0.2\textheight]{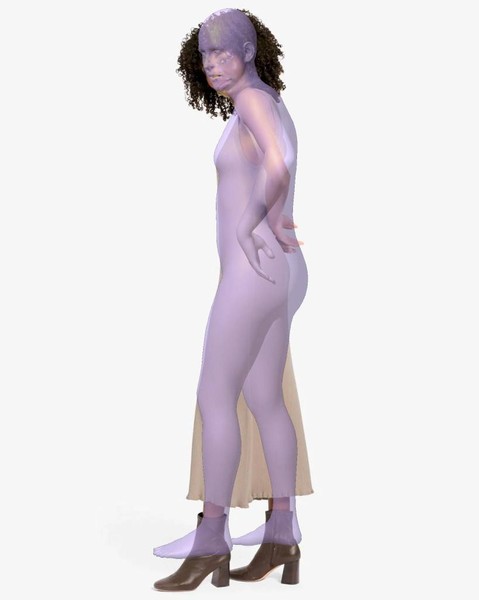}}
\subfloat[SHAPY \cite{choutas2022accurate}]
{\includegraphics[height=0.2\textheight]
{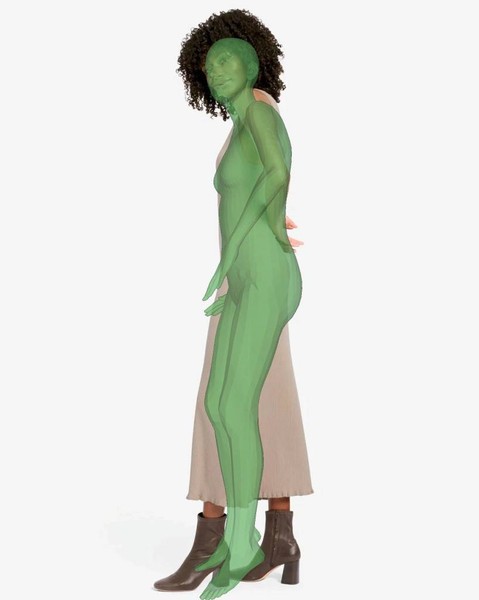}}
\subfloat[\KBody{-.1}{.035} (Ours)]{\includegraphics[height=0.2\textheight]{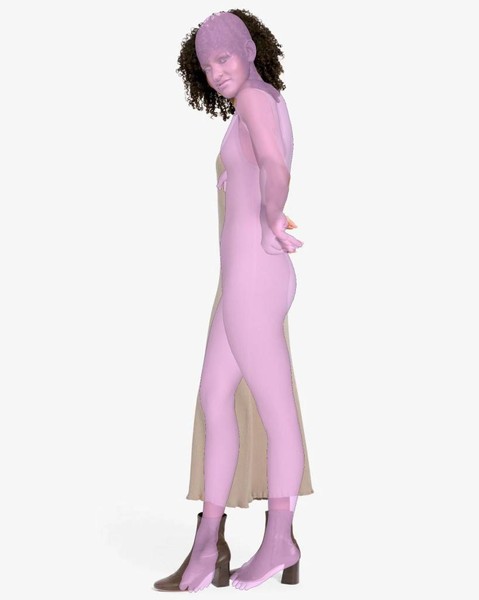}}

\caption{
Left-to-right: SMPLify-X \cite{pavlakos2019expressive} (\textcolor{caribbeangreen2}{light green}), PyMAF-X \cite{pymafx2022} (\textcolor{violet}{purple}), SHAPY \cite{choutas2022accurate} (\textcolor{jade}{green}) and KBody (\textcolor{candypink}{pink}).
}
\label{fig:hm_splendid}
\end{figure*}

%% file: figures/supp/splendid.tex
\begin{figure*}[!htbp]
\captionsetup[subfigure]{position=bottom,labelformat=empty}

\centering

\subfloat{\includegraphics[height=0.24\textheight]{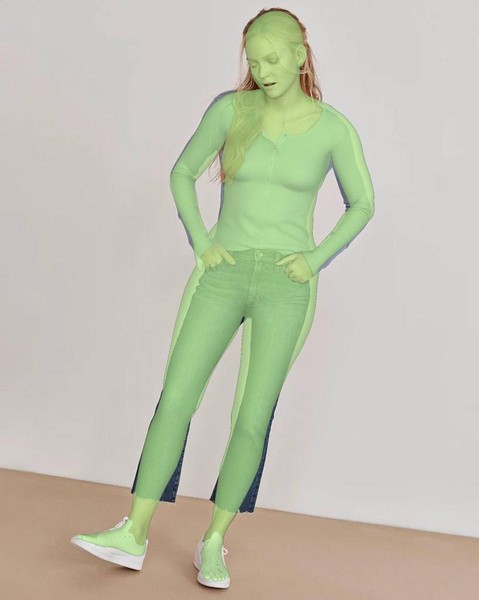}}
\subfloat{\includegraphics[height=0.24\textheight]{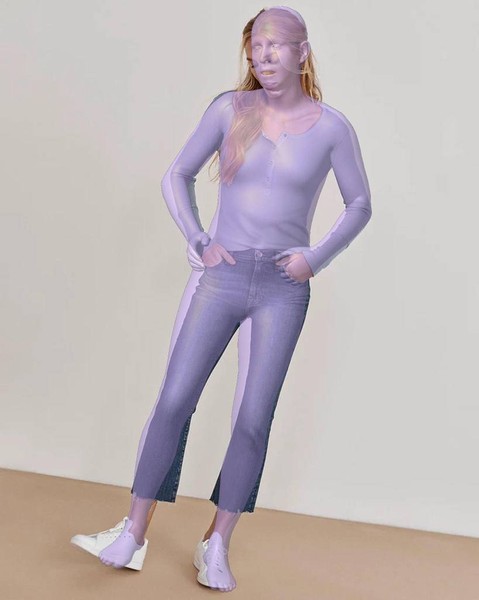}}
\subfloat{\includegraphics[height=0.24\textheight]{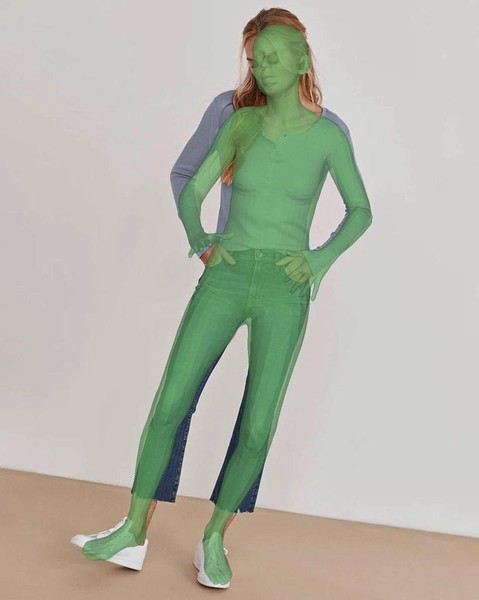}}
\subfloat{\includegraphics[height=0.24\textheight]{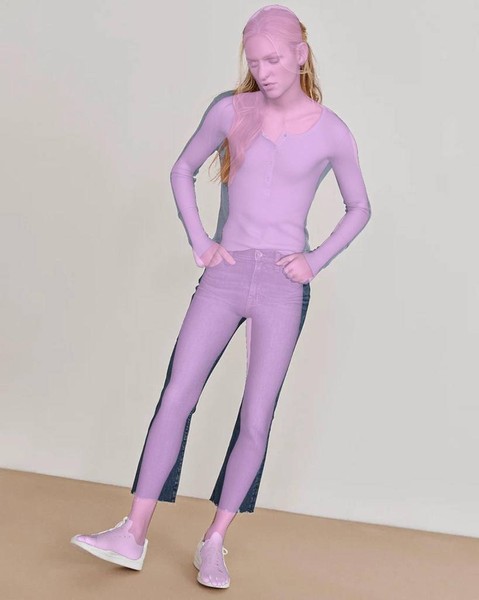}}

\subfloat{\includegraphics[height=0.24\textheight]{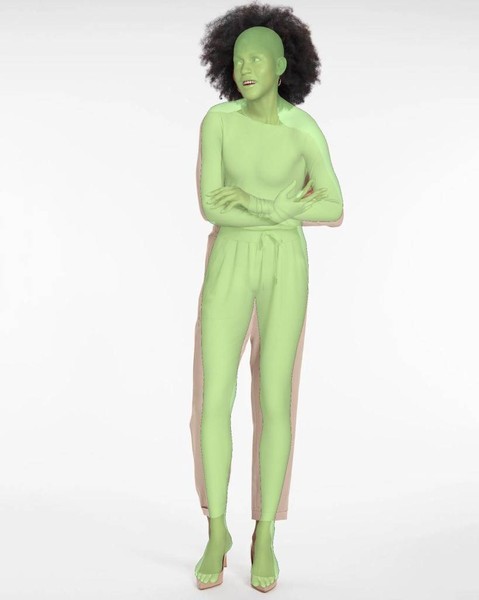}}
\subfloat{\includegraphics[height=0.24\textheight]{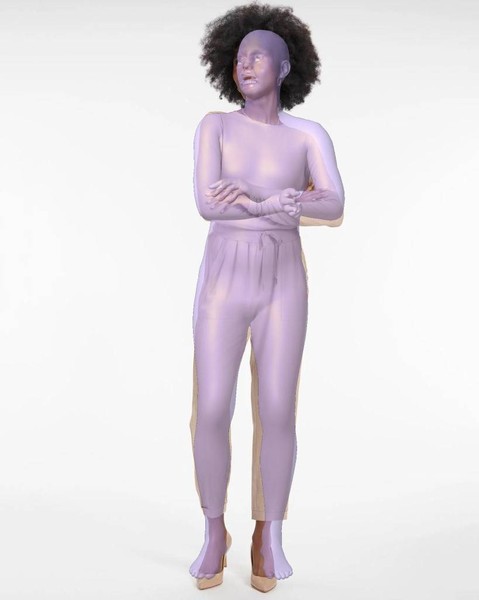}}
\subfloat{\includegraphics[height=0.24\textheight]{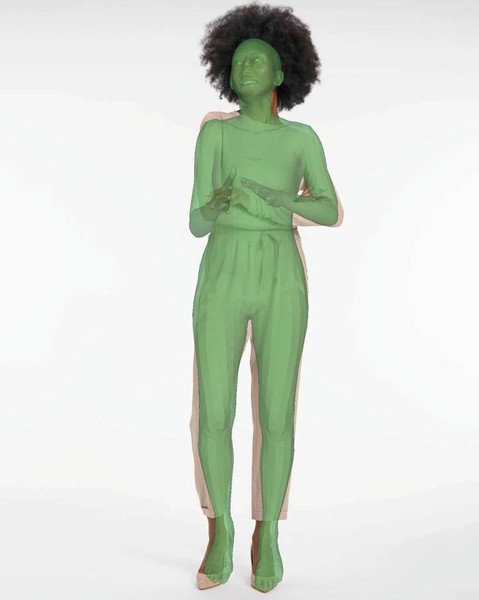}}
\subfloat{\includegraphics[height=0.24\textheight]{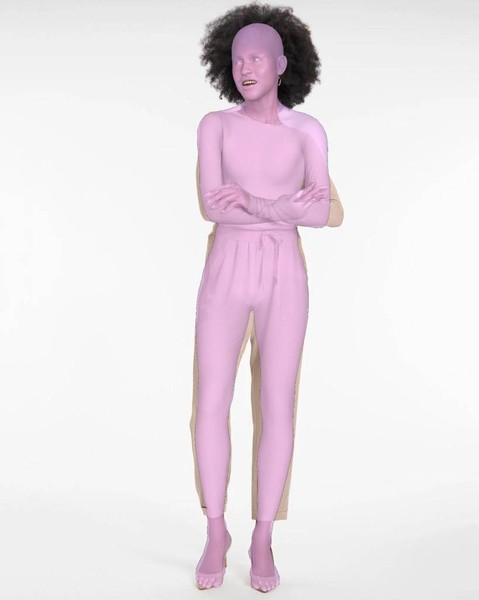}}

\subfloat{\includegraphics[height=0.24\textheight]{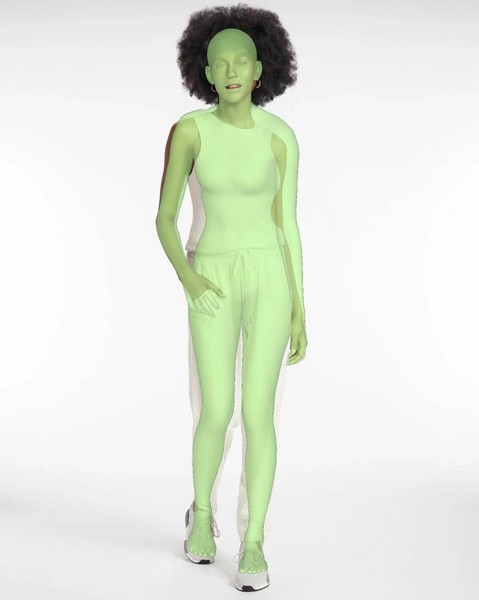}}
\subfloat{\includegraphics[height=0.24\textheight]{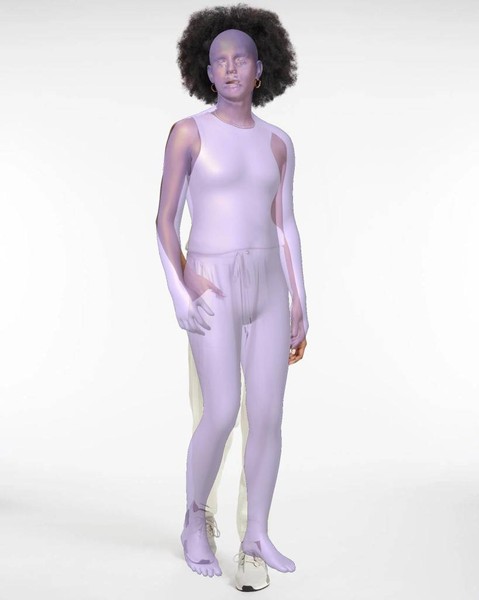}}
\subfloat{\includegraphics[height=0.24\textheight]{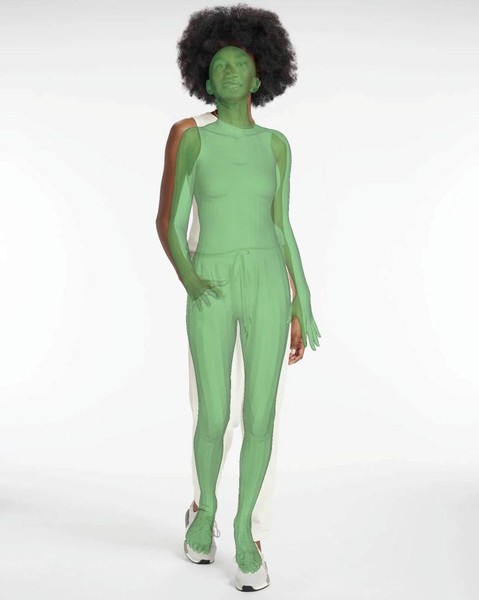}}
\subfloat{\includegraphics[height=0.24\textheight]{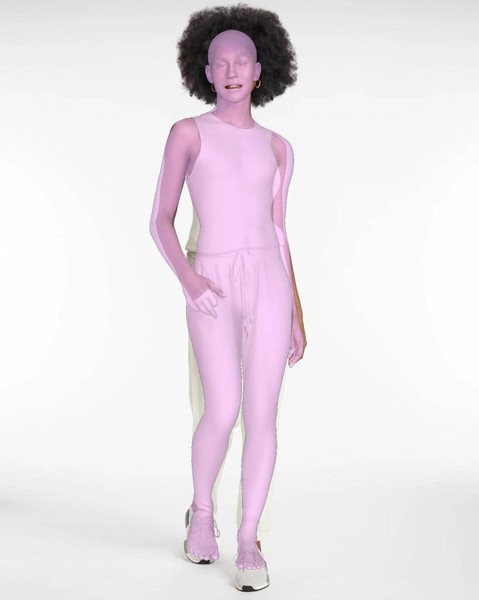}}

\subfloat[SMPLify-X \cite{pavlakos2019expressive}]
{\includegraphics[height=0.24\textheight]{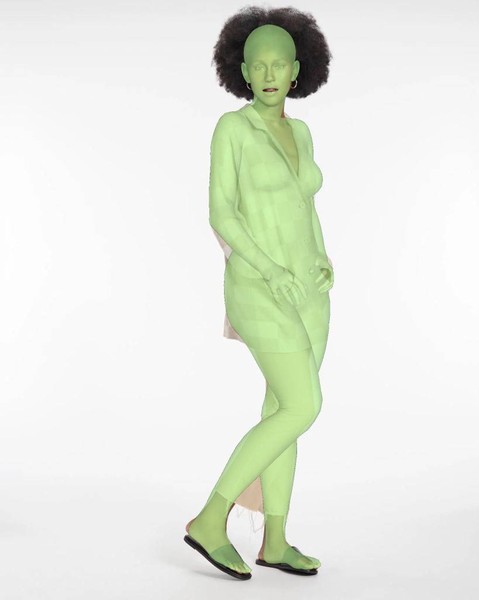}}
\subfloat[PyMAF-X \cite{pymafx2022}]{\includegraphics[height=0.24\textheight]{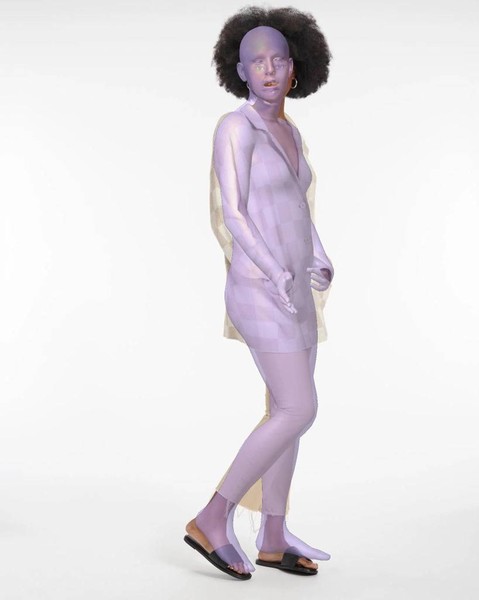}}
\subfloat[SHAPY \cite{choutas2022accurate}]
{\includegraphics[height=0.24\textheight]
{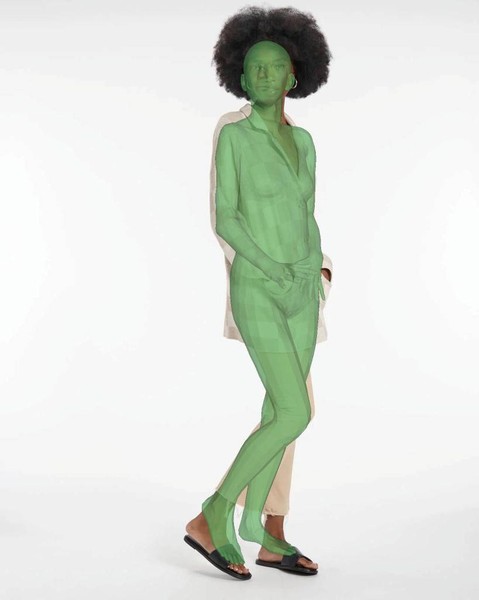}}
\subfloat[\KBody{-.1}{.035} (Ours)]{\includegraphics[height=0.24\textheight]{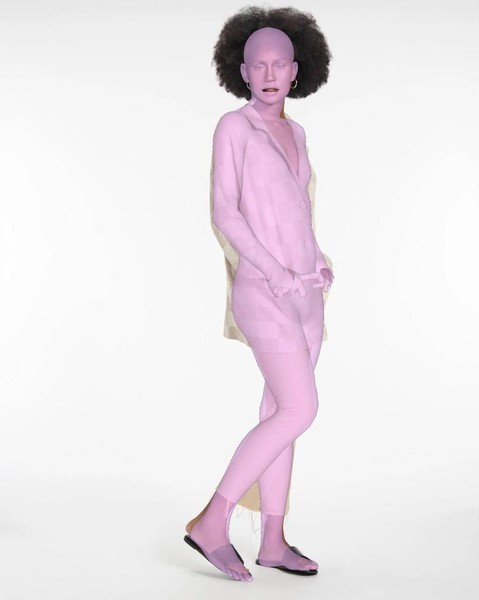}}

\caption{
Left-to-right: SMPLify-X \cite{pavlakos2019expressive} (\textcolor{caribbeangreen2}{light green}), PyMAF-X \cite{pymafx2022} (\textcolor{violet}{purple}), SHAPY \cite{choutas2022accurate} (\textcolor{jade}{green}) and KBody (\textcolor{candypink}{pink}).
}
\label{fig:splendid}
\end{figure*}

%% file: figures/supp/problem1.tex
\begin{figure*}[!htbp]
\captionsetup[subfigure]{position=bottom,labelformat=empty}

\centering

\subfloat{\includegraphics[height=0.24\textheight]{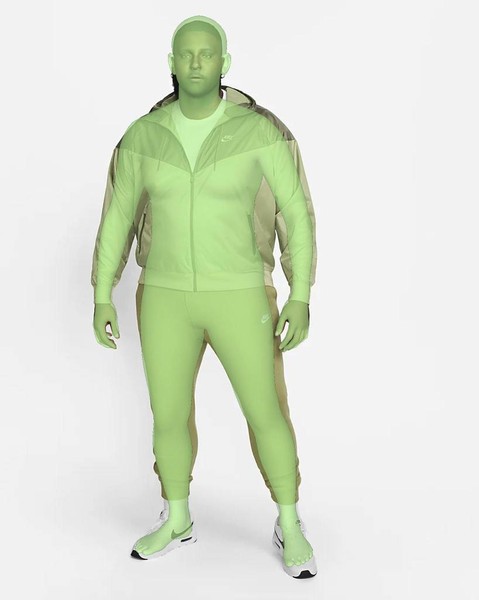}}
\subfloat{\includegraphics[height=0.24\textheight]{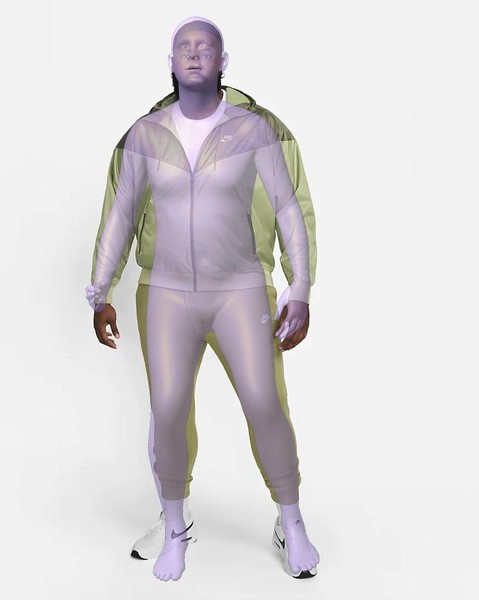}}
\subfloat{\includegraphics[height=0.24\textheight]{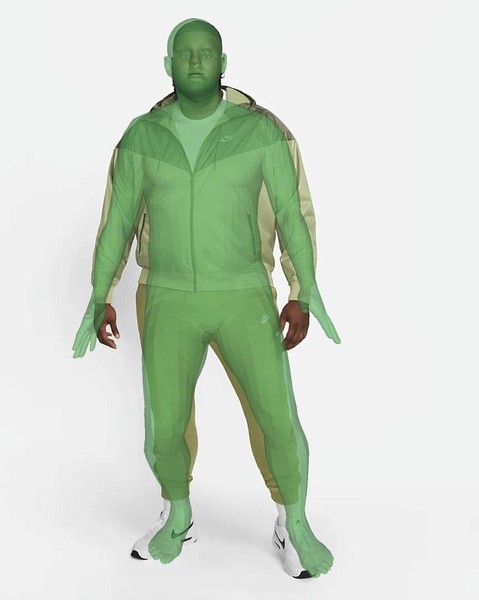}}
\subfloat{\includegraphics[height=0.24\textheight]{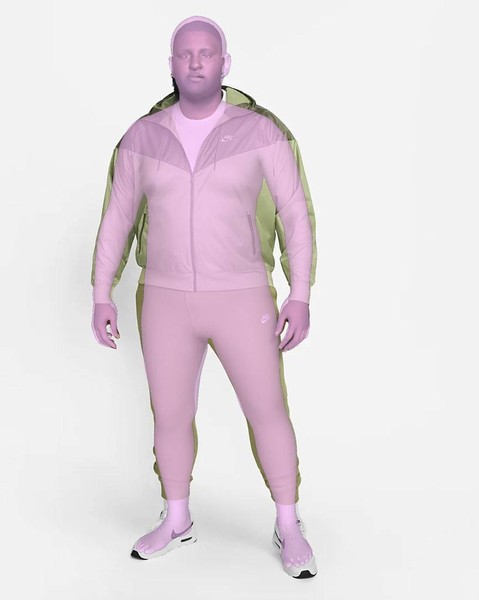}}

\subfloat{\includegraphics[height=0.24\textheight]{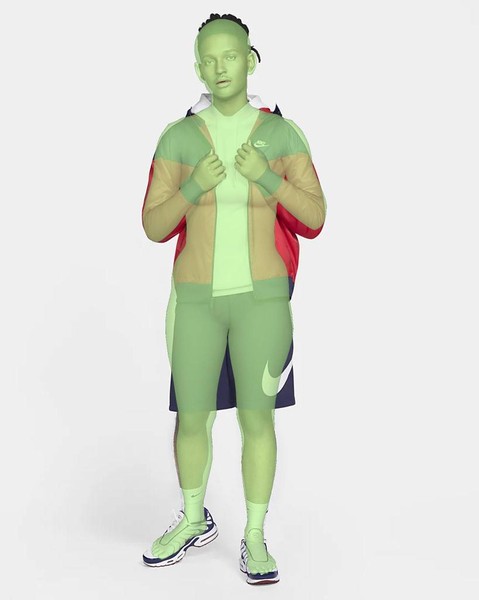}}
\subfloat{\includegraphics[height=0.24\textheight]{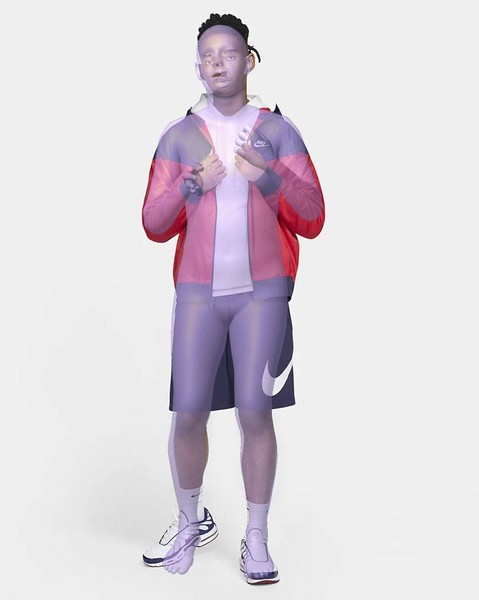}}
\subfloat{\includegraphics[height=0.24\textheight]{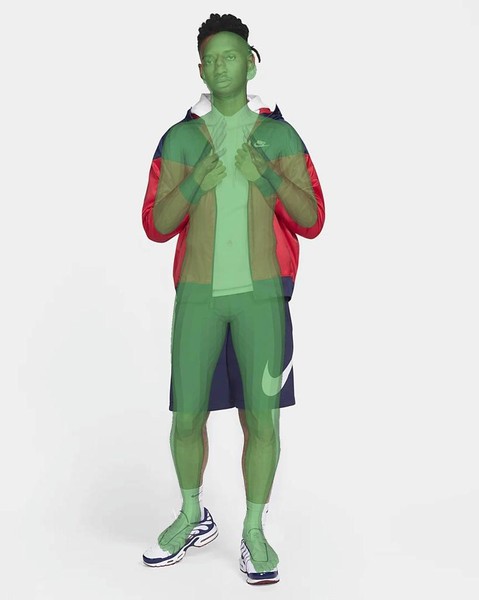}}
\subfloat{\includegraphics[height=0.24\textheight]{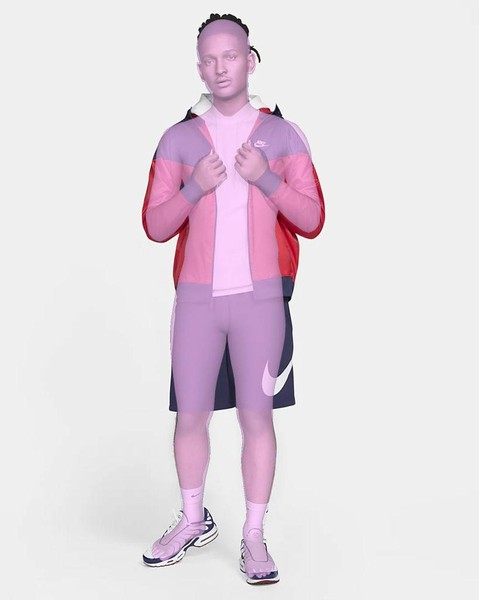}}

\subfloat{\includegraphics[height=0.24\textheight]{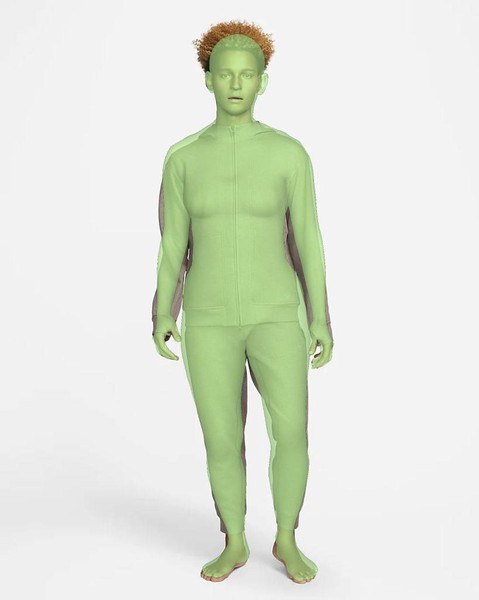}}
\subfloat{\includegraphics[height=0.24\textheight]{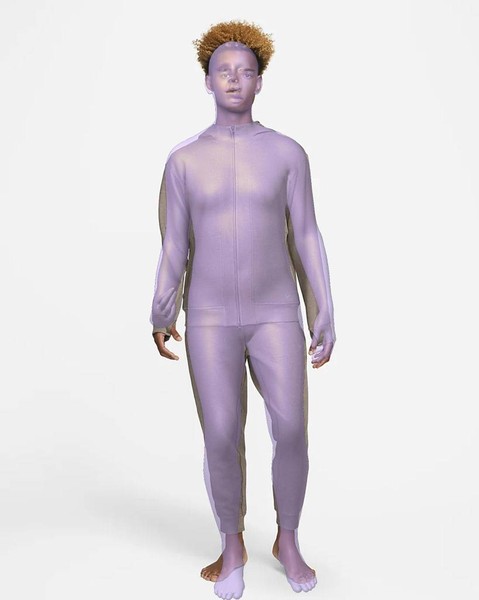}}
\subfloat{\includegraphics[height=0.24\textheight]{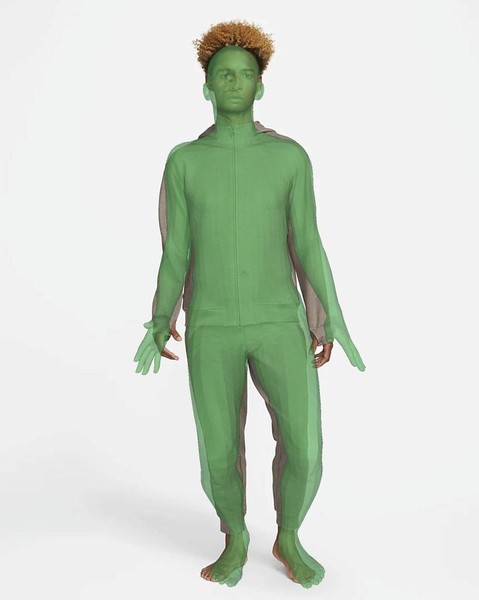}}
\subfloat{\includegraphics[height=0.24\textheight]{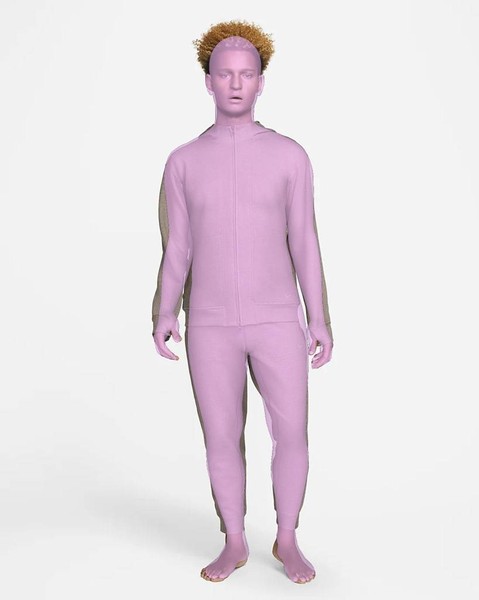}}

\subfloat[SMPLify-X \cite{pavlakos2019expressive}]
{\includegraphics[height=0.24\textheight]{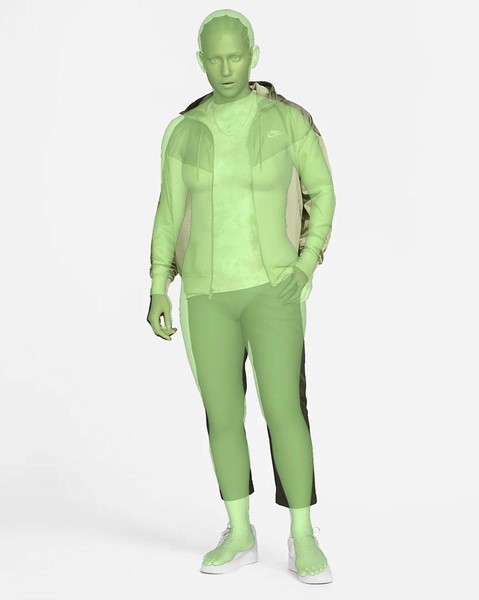}}
\subfloat[PyMAF-X \cite{pymafx2022}]{\includegraphics[height=0.24\textheight]{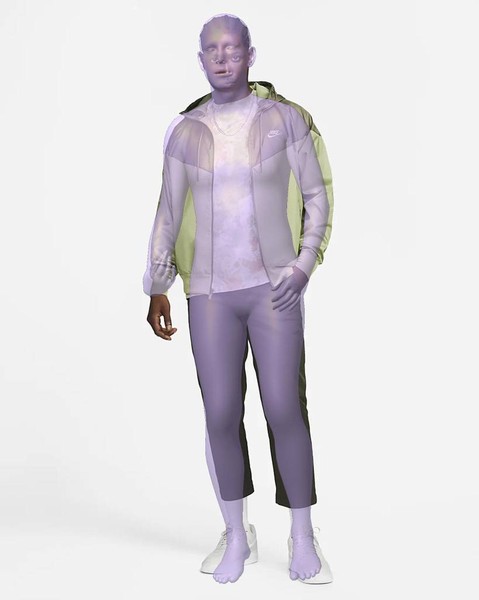}}
\subfloat[SHAPY \cite{choutas2022accurate}]
{\includegraphics[height=0.24\textheight]
{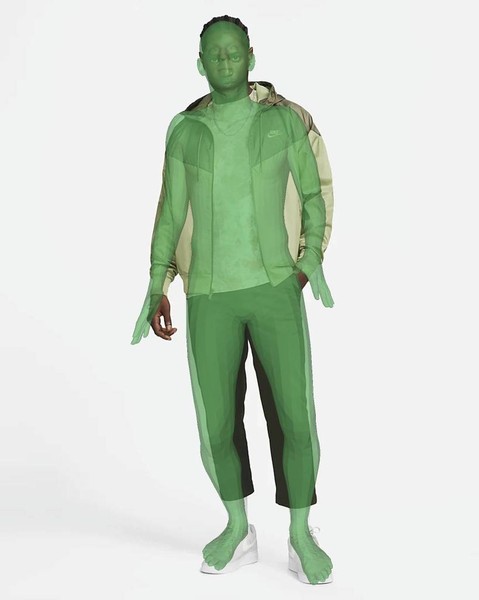}}
\subfloat[\KBody{-.1}{.035} (Ours)]{\includegraphics[height=0.24\textheight]{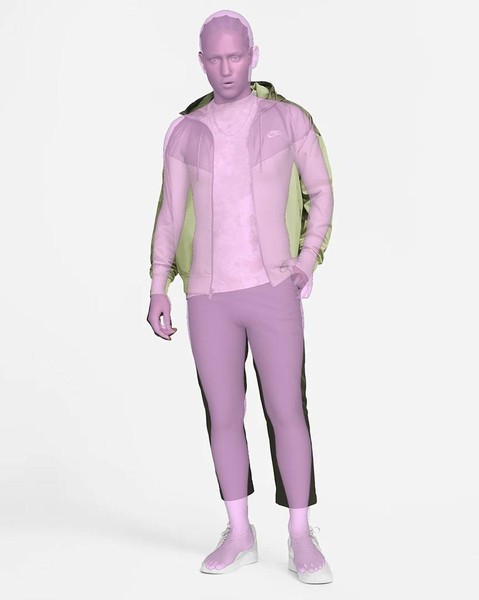}}

\caption{
Left-to-right: SMPLify-X \cite{pavlakos2019expressive} (\textcolor{caribbeangreen2}{light green}), PyMAF-X \cite{pymafx2022} (\textcolor{violet}{purple}), SHAPY \cite{choutas2022accurate} (\textcolor{jade}{green}) and KBody (\textcolor{candypink}{pink}).
}
\label{fig:p1}
\end{figure*}

%% file: figures/supp/problem2.tex
\begin{figure*}[!htbp]
\captionsetup[subfigure]{position=bottom,labelformat=empty}

\centering

\subfloat{\includegraphics[height=0.24\textheight]{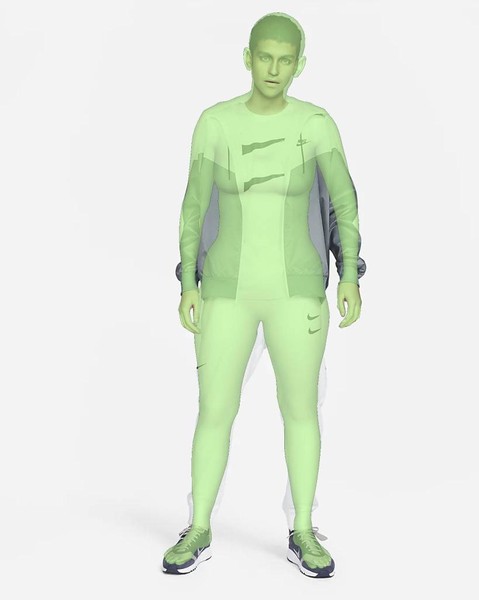}}
\subfloat{\includegraphics[height=0.24\textheight]{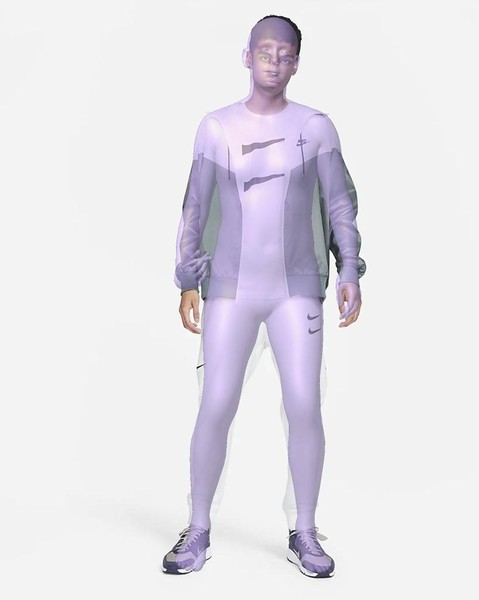}}
\subfloat{\includegraphics[height=0.24\textheight]{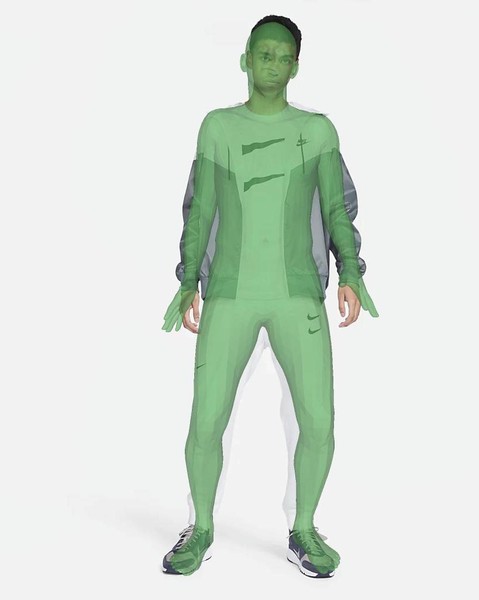}}
\subfloat{\includegraphics[height=0.24\textheight]{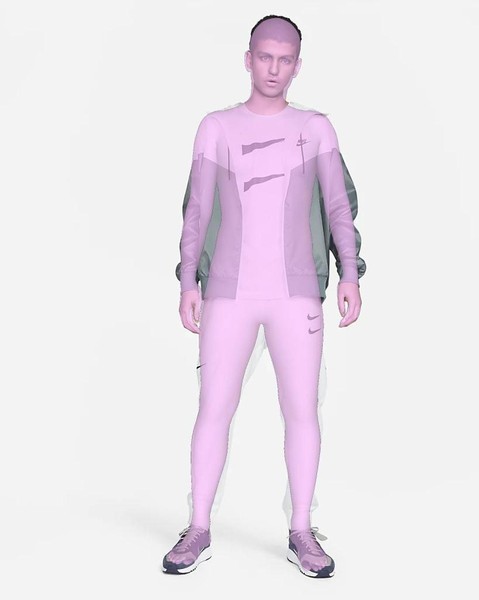}}

\subfloat{\includegraphics[height=0.24\textheight]{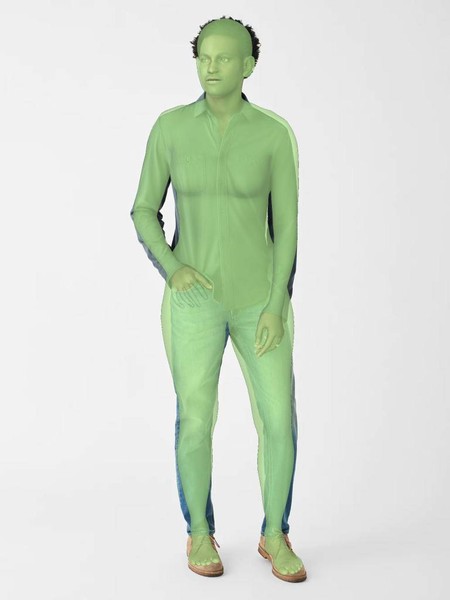}}
\subfloat{\includegraphics[height=0.24\textheight]{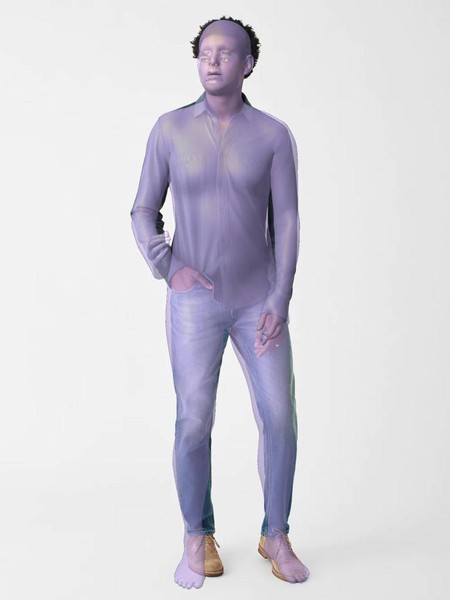}}
\subfloat{\includegraphics[height=0.24\textheight]{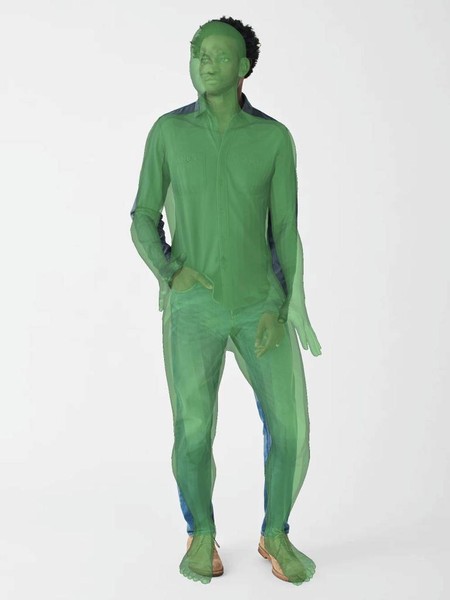}}
\subfloat{\includegraphics[height=0.24\textheight]{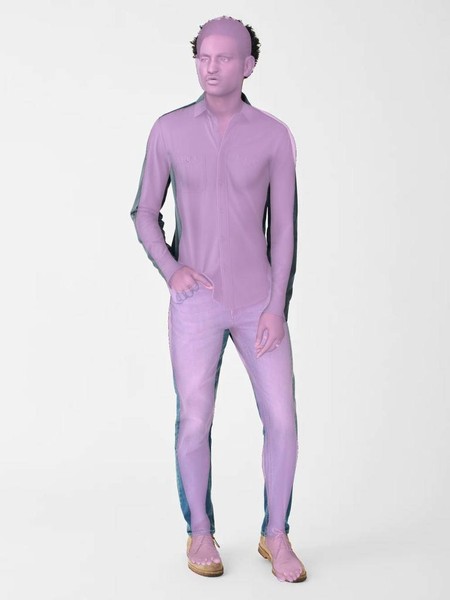}}

\subfloat{\includegraphics[height=0.24\textheight]{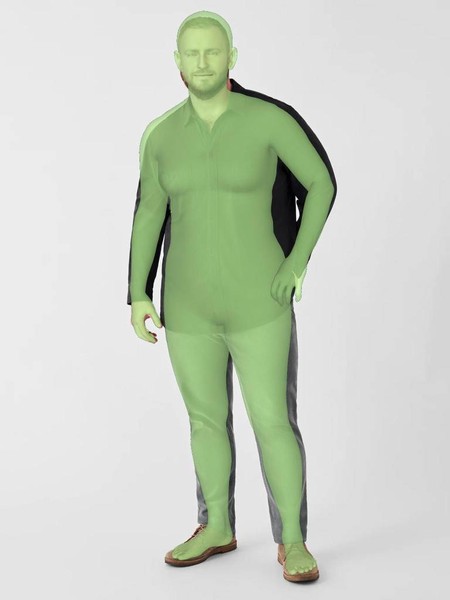}}
\subfloat{\includegraphics[height=0.24\textheight]{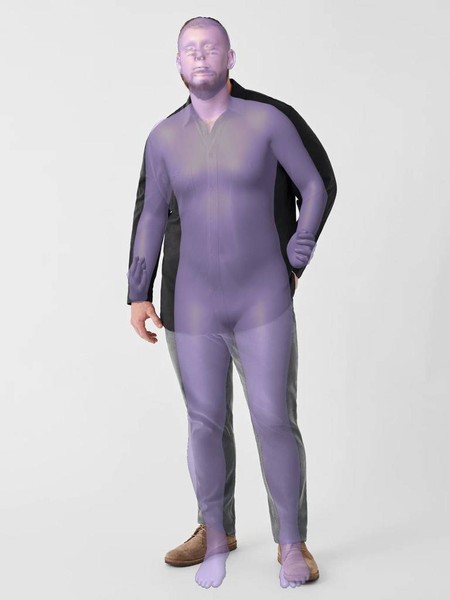}}
\subfloat{\includegraphics[height=0.24\textheight]{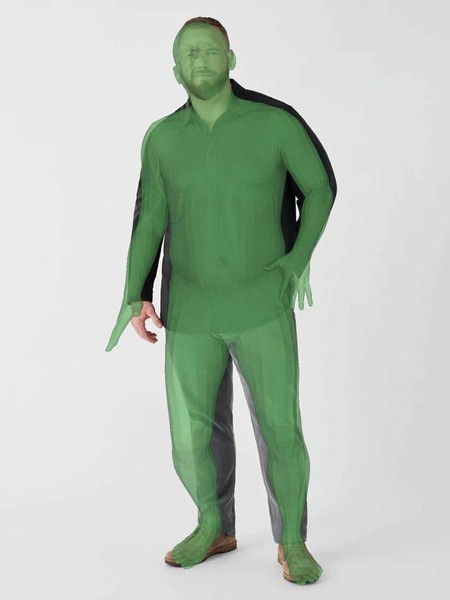}}
\subfloat{\includegraphics[height=0.24\textheight]{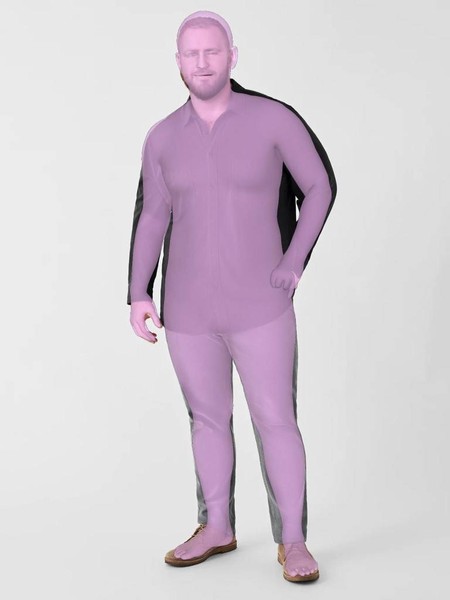}}

\subfloat[SMPLify-X \cite{pavlakos2019expressive}]
{\includegraphics[height=0.24\textheight]{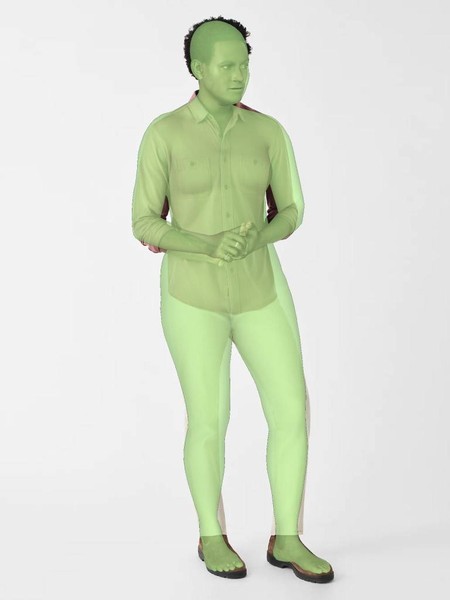}}
\subfloat[PyMAF-X \cite{pymafx2022}]{\includegraphics[height=0.24\textheight]{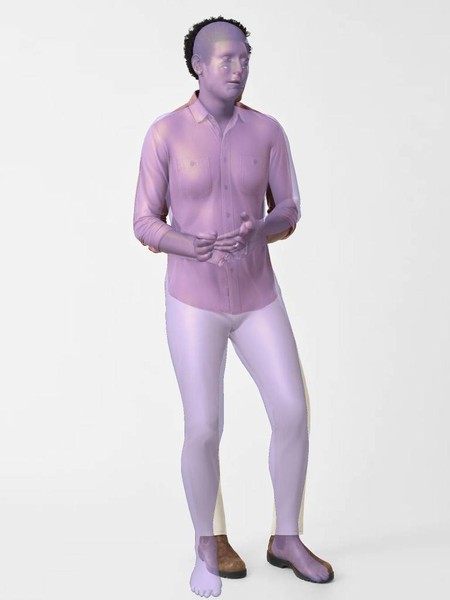}}
\subfloat[SHAPY \cite{choutas2022accurate}]
{\includegraphics[height=0.24\textheight]
{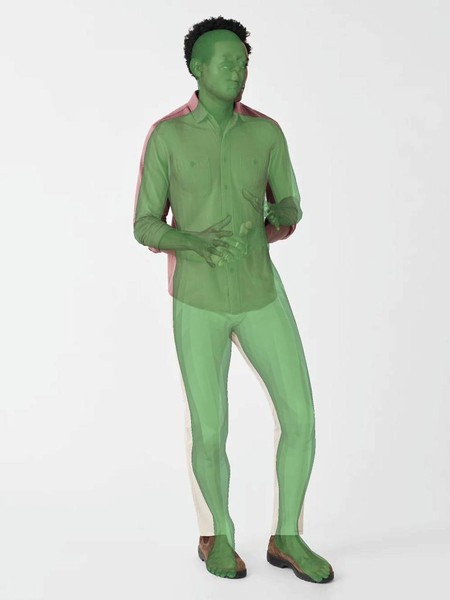}}
\subfloat[\KBody{-.1}{.035} (Ours)]{\includegraphics[height=0.24\textheight]{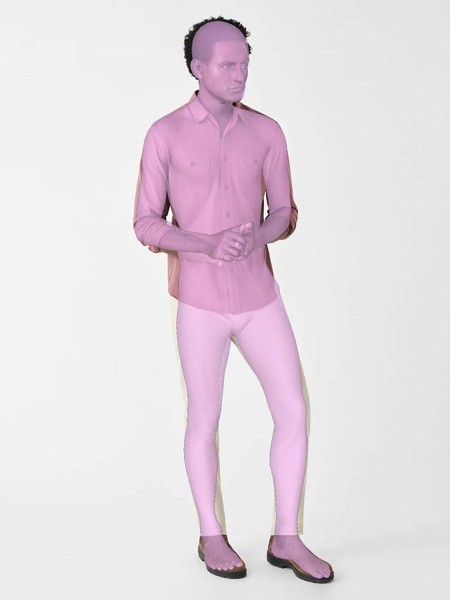}}

\caption{
Left-to-right: SMPLify-X \cite{pavlakos2019expressive} (\textcolor{caribbeangreen2}{light green}), PyMAF-X \cite{pymafx2022} (\textcolor{violet}{purple}), SHAPY \cite{choutas2022accurate} (\textcolor{jade}{green}) and KBody (\textcolor{candypink}{pink}).
}
\label{fig:p2}
\end{figure*}

%% file: figures/supp/problem3.tex
\begin{figure*}[!htbp]
\captionsetup[subfigure]{position=bottom,labelformat=empty}

\centering

\subfloat{\includegraphics[height=0.24\textheight]{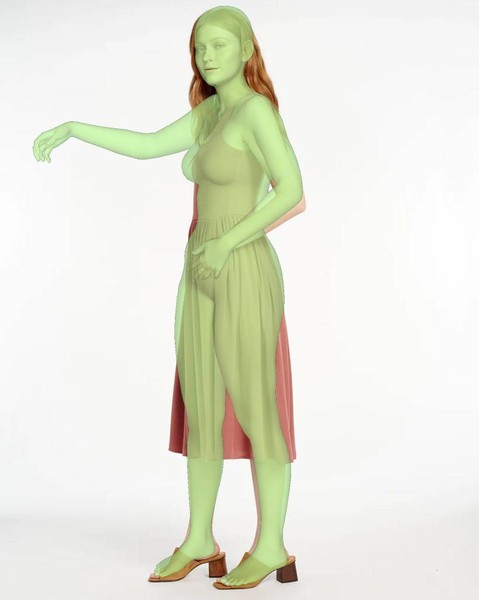}}
\subfloat{\includegraphics[height=0.24\textheight]{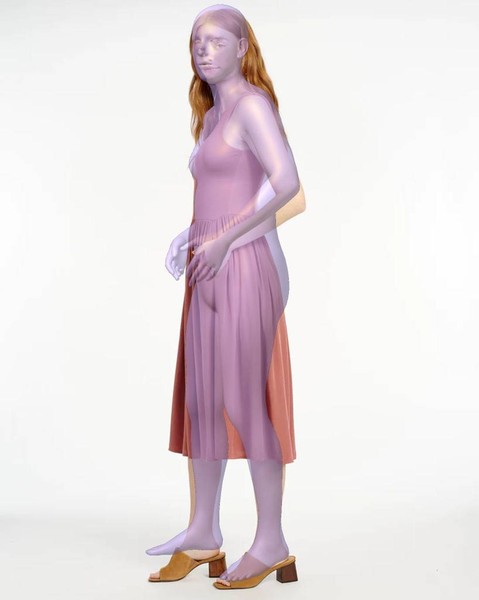}}
\subfloat{\includegraphics[height=0.24\textheight]{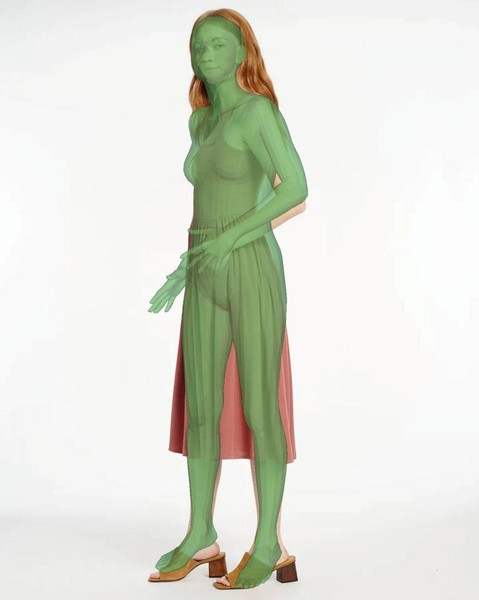}}
\subfloat{\includegraphics[height=0.24\textheight]{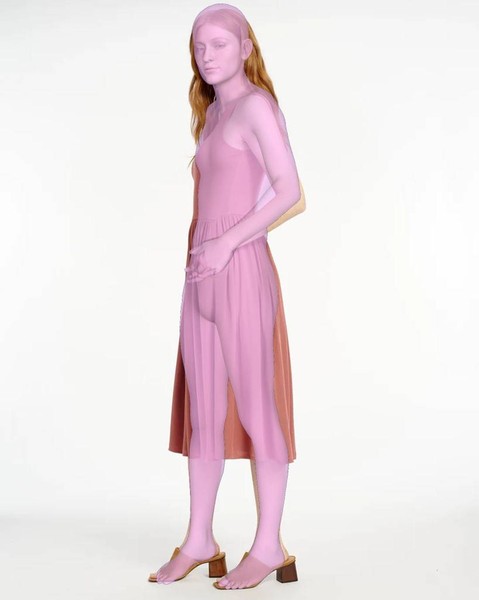}}

\subfloat{\includegraphics[height=0.24\textheight]{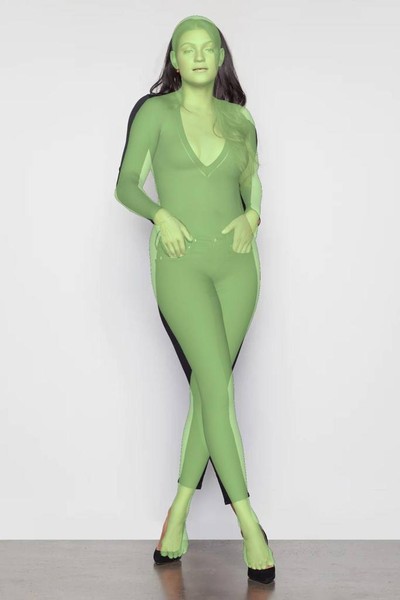}}
\subfloat{\includegraphics[height=0.24\textheight]{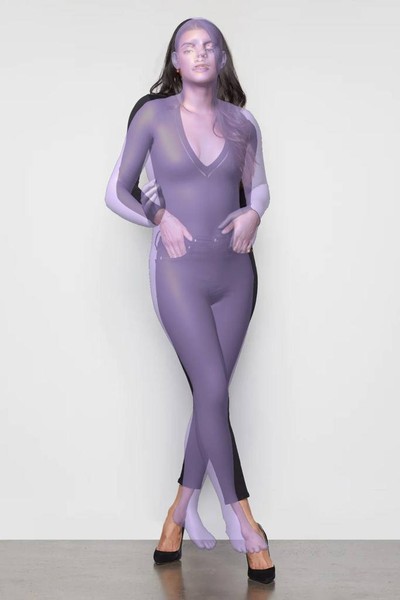}}
\subfloat{\includegraphics[height=0.24\textheight]{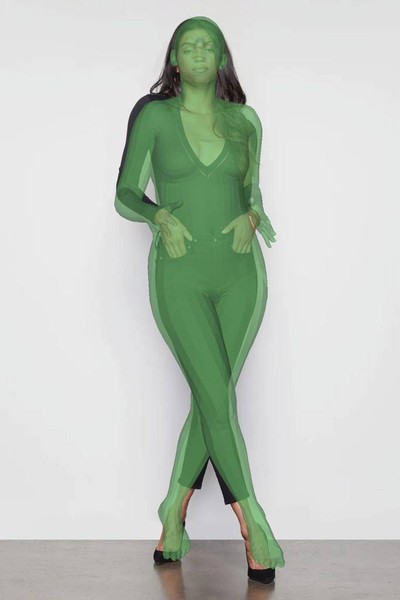}}
\subfloat{\includegraphics[height=0.24\textheight]{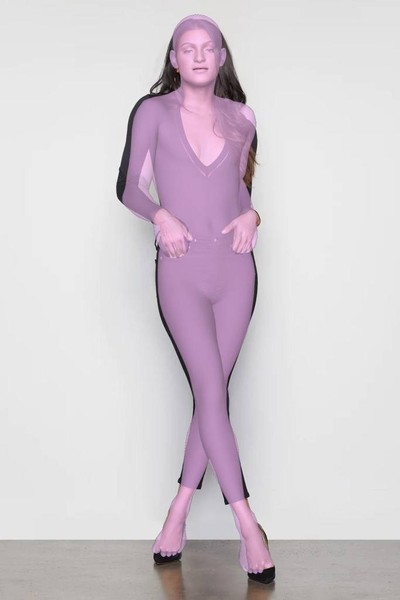}}

\subfloat{\includegraphics[height=0.24\textheight]{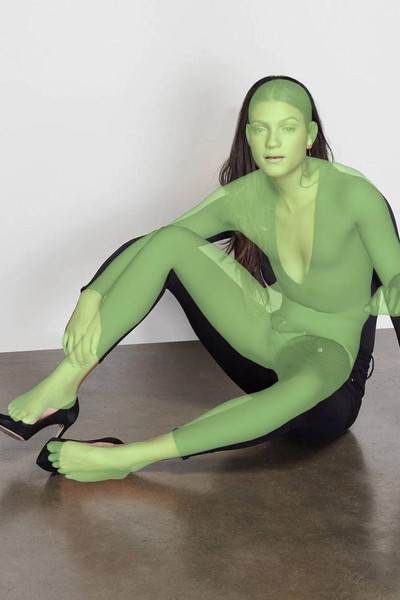}}
\subfloat{\includegraphics[height=0.24\textheight]{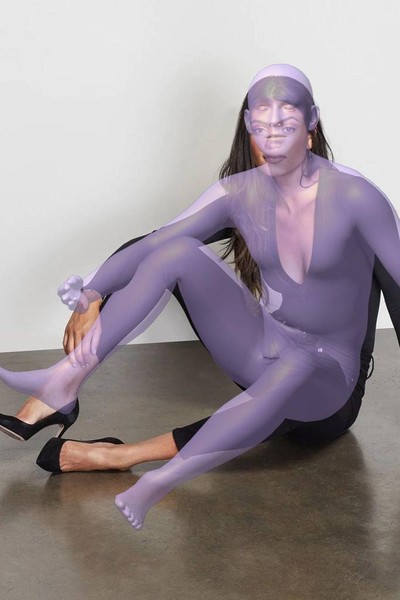}}
\subfloat{\includegraphics[height=0.24\textheight]{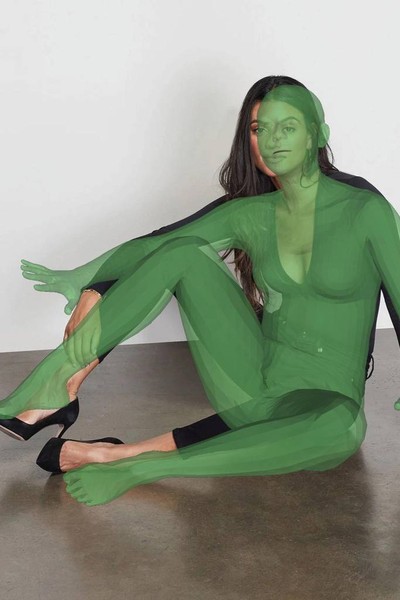}}
\subfloat{\includegraphics[height=0.24\textheight]{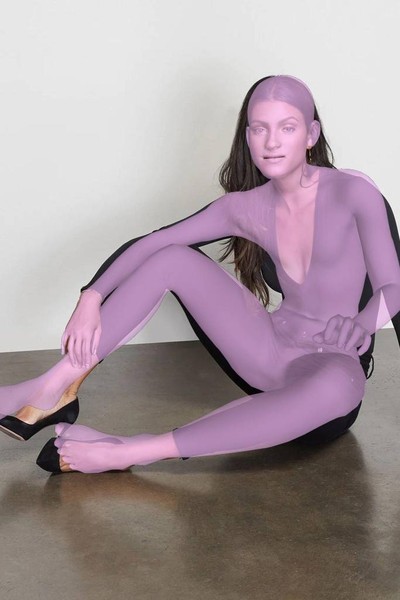}}

\subfloat[SMPLify-X \cite{pavlakos2019expressive}]
{\includegraphics[height=0.24\textheight]{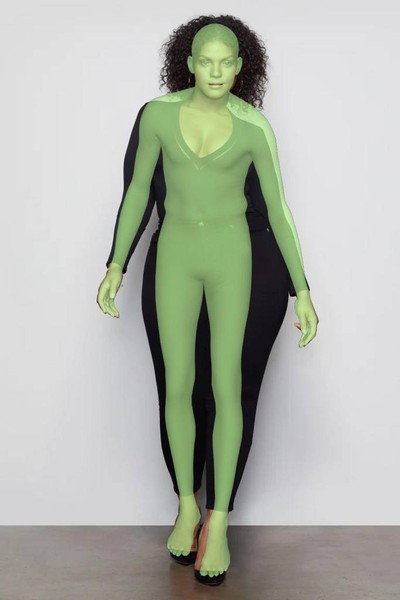}}
\subfloat[PyMAF-X \cite{pymafx2022}]{\includegraphics[height=0.24\textheight]{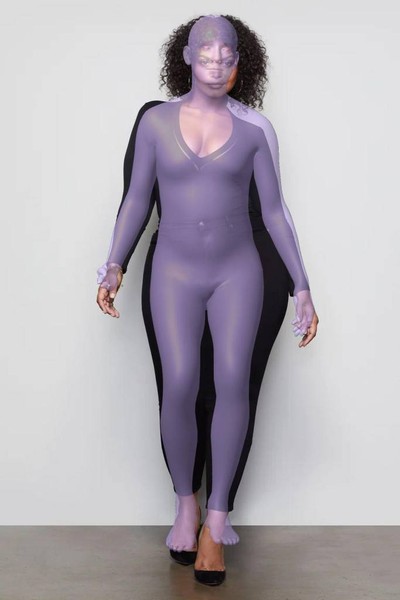}}
\subfloat[SHAPY \cite{choutas2022accurate}]
{\includegraphics[height=0.24\textheight]
{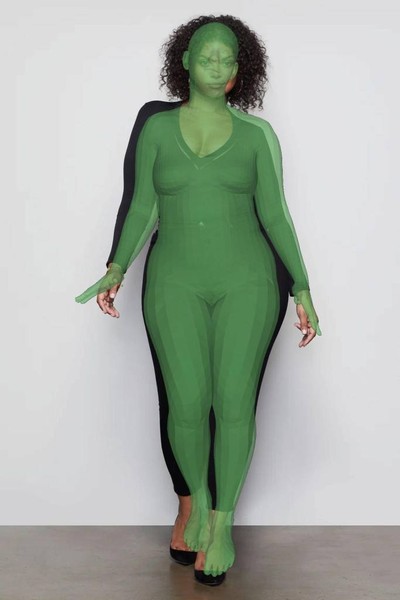}}
\subfloat[\KBody{-.1}{.035} (Ours)]{\includegraphics[height=0.24\textheight]{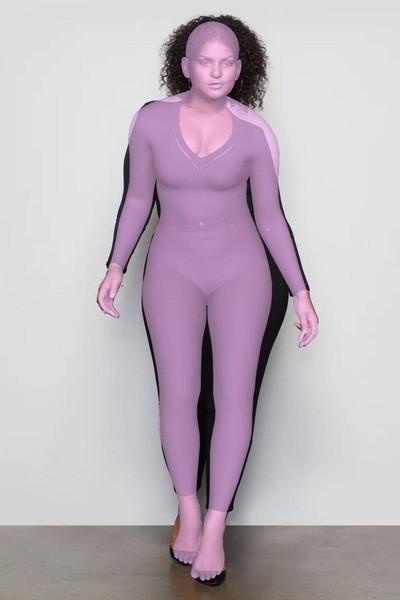}}

\caption{
Left-to-right: SMPLify-X \cite{pavlakos2019expressive} (\textcolor{caribbeangreen2}{light green}), PyMAF-X \cite{pymafx2022} (\textcolor{violet}{purple}), SHAPY \cite{choutas2022accurate} (\textcolor{jade}{green}) and KBody (\textcolor{candypink}{pink}).
}
\label{fig:p3}
\end{figure*}

%% file: figures/supp/problem4.tex
\begin{figure*}[!htbp]
\captionsetup[subfigure]{position=bottom,labelformat=empty}

\centering

\subfloat{\includegraphics[height=0.24\textheight]{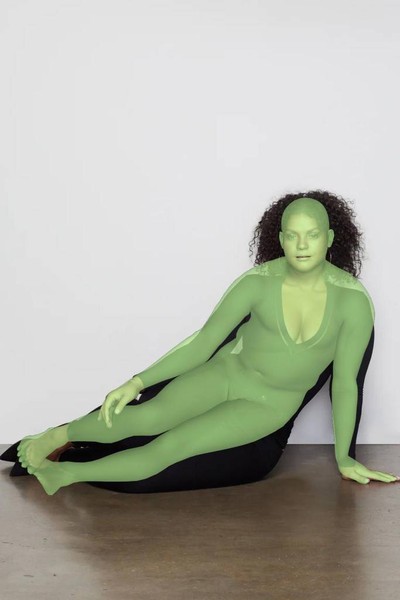}}
\subfloat{\includegraphics[height=0.24\textheight]{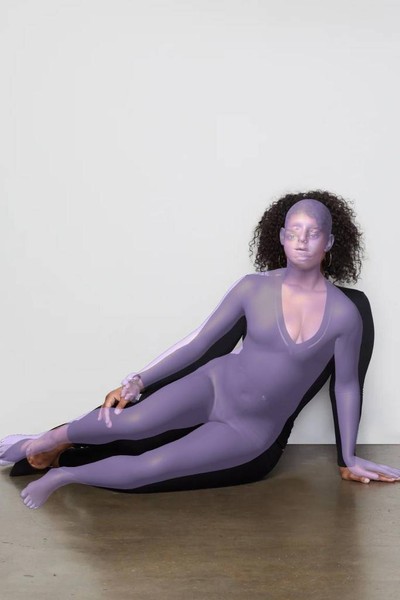}}
\subfloat{\includegraphics[height=0.24\textheight]{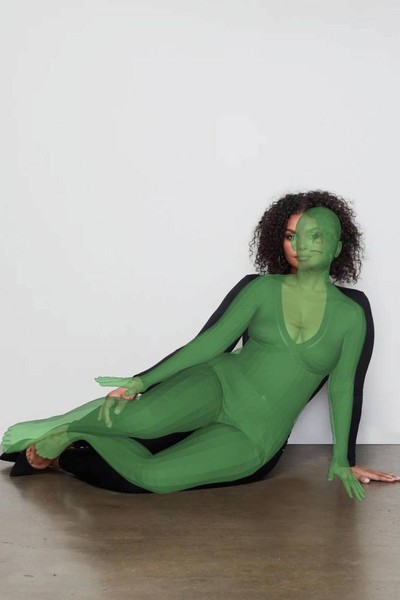}}
\subfloat{\includegraphics[height=0.24\textheight]{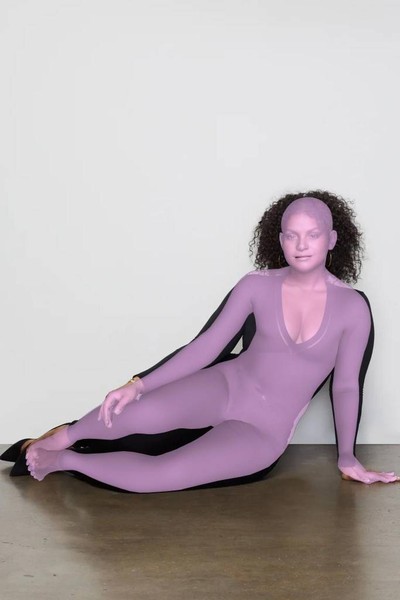}}

\subfloat{\includegraphics[height=0.24\textheight]{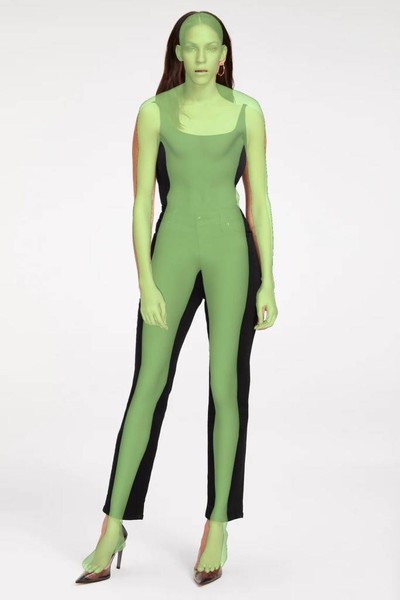}}
\subfloat{\includegraphics[height=0.24\textheight]{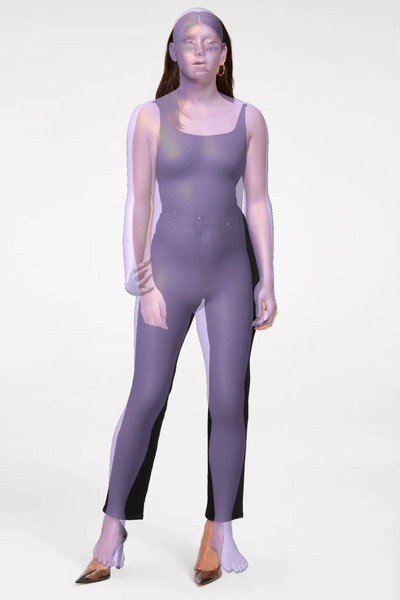}}
\subfloat{\includegraphics[height=0.24\textheight]{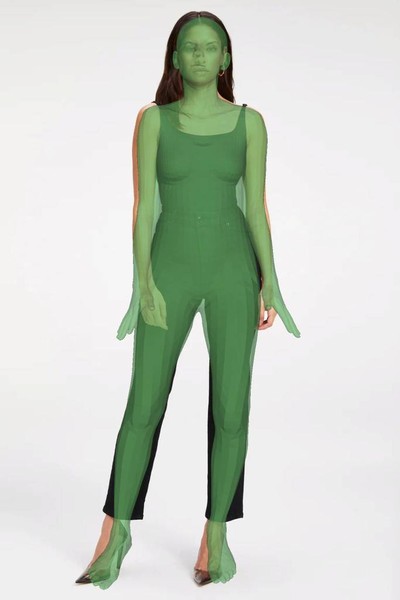}}
\subfloat{\includegraphics[height=0.24\textheight]{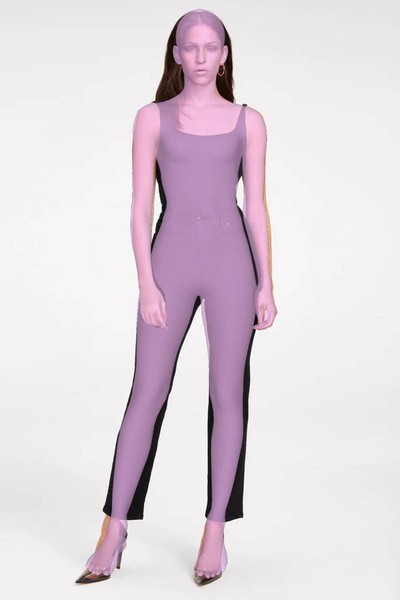}}

\subfloat{\includegraphics[height=0.24\textheight]{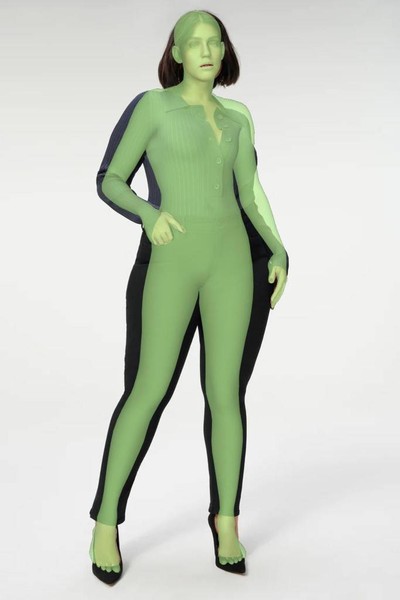}}
\subfloat{\includegraphics[height=0.24\textheight]{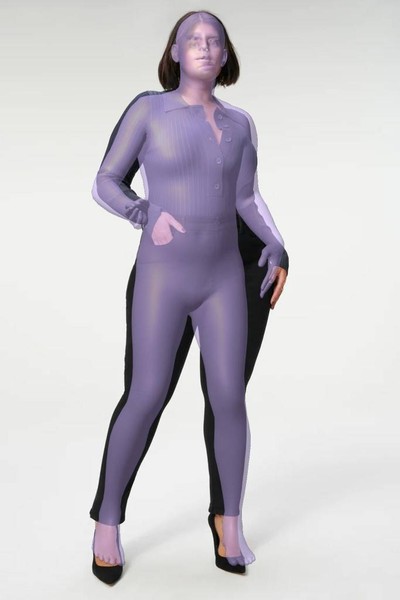}}
\subfloat{\includegraphics[height=0.24\textheight]{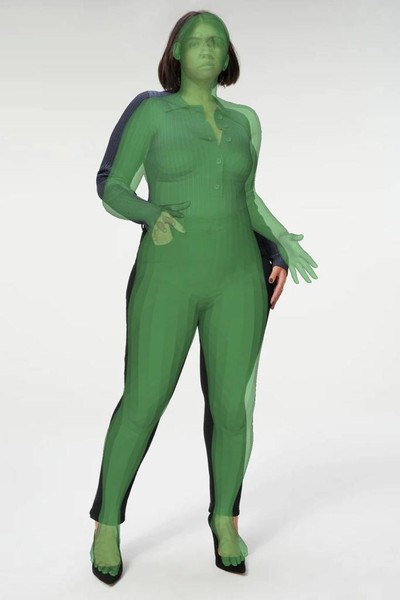}}
\subfloat{\includegraphics[height=0.24\textheight]{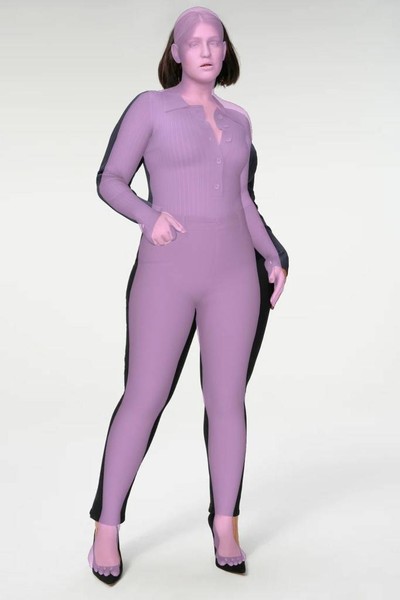}}

\subfloat[SMPLify-X \cite{pavlakos2019expressive}]
{\includegraphics[height=0.24\textheight]{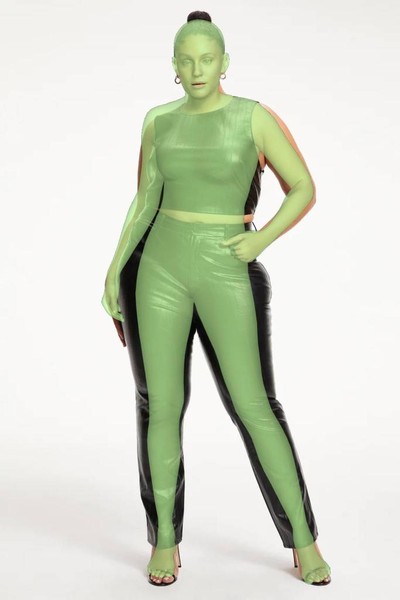}}
\subfloat[PyMAF-X \cite{pymafx2022}]{\includegraphics[height=0.24\textheight]{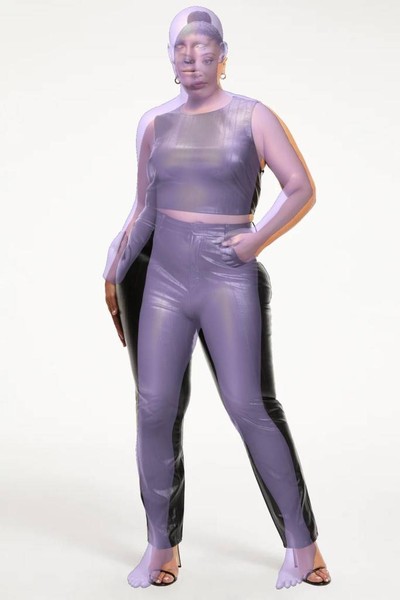}}
\subfloat[SHAPY \cite{choutas2022accurate}]
{\includegraphics[height=0.24\textheight]
{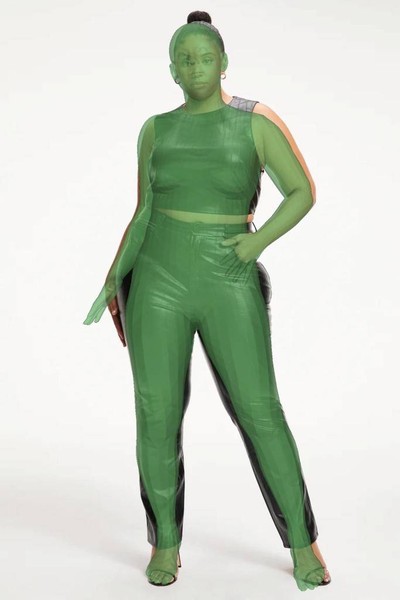}}
\subfloat[\KBody{-.1}{.035} (Ours)]{\includegraphics[height=0.24\textheight]{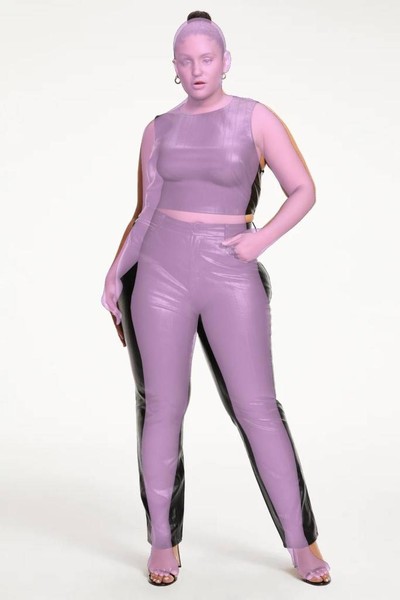}}

\caption{
Left-to-right: SMPLify-X \cite{pavlakos2019expressive} (\textcolor{caribbeangreen2}{light green}), PyMAF-X \cite{pymafx2022} (\textcolor{violet}{purple}), SHAPY \cite{choutas2022accurate} (\textcolor{jade}{green}) and KBody (\textcolor{candypink}{pink}).
}
\label{fig:p4}
\end{figure*}

%% file: figures/supp/problem5.tex
\begin{figure*}[!htbp]
\captionsetup[subfigure]{position=bottom,labelformat=empty}

\centering

\subfloat{\includegraphics[height=0.24\textheight]{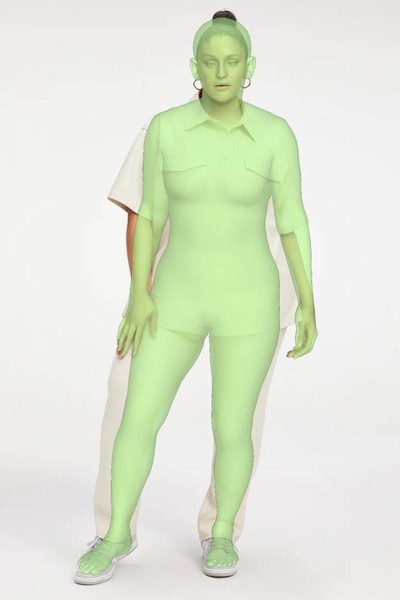}}
\subfloat{\includegraphics[height=0.24\textheight]{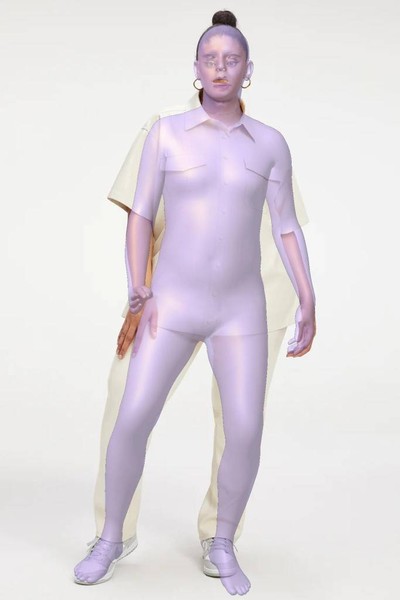}}
\subfloat{\includegraphics[height=0.24\textheight]{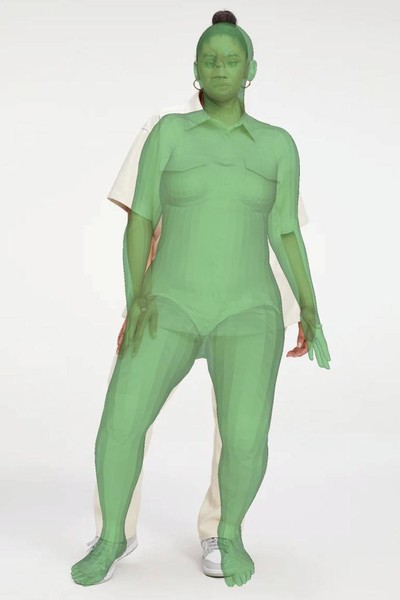}}
\subfloat{\includegraphics[height=0.24\textheight]{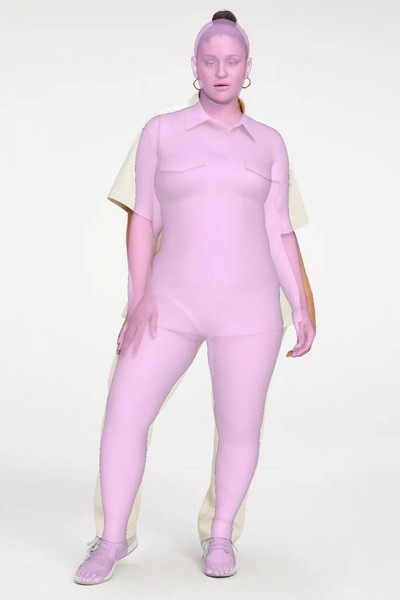}}

\subfloat{\includegraphics[height=0.24\textheight]{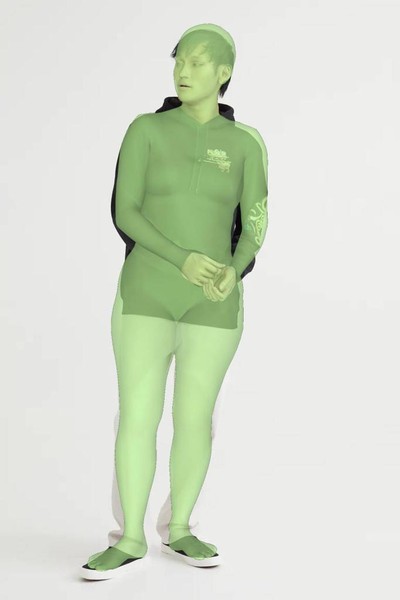}}
\subfloat{\includegraphics[height=0.24\textheight]{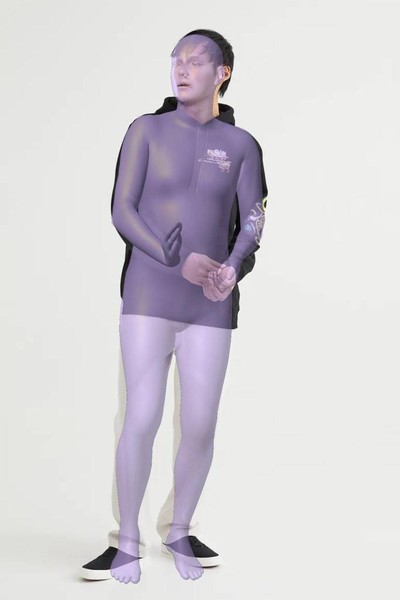}}
\subfloat{\includegraphics[height=0.24\textheight]{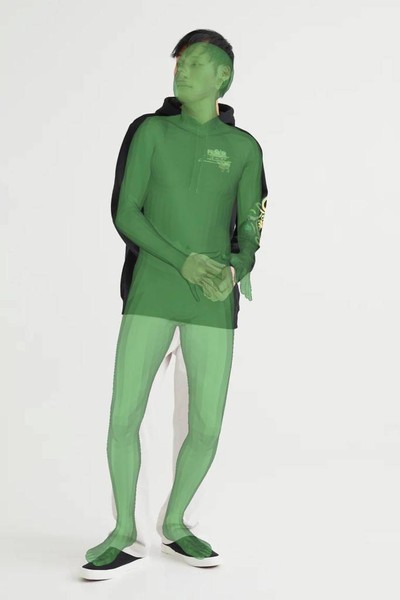}}
\subfloat{\includegraphics[height=0.24\textheight]{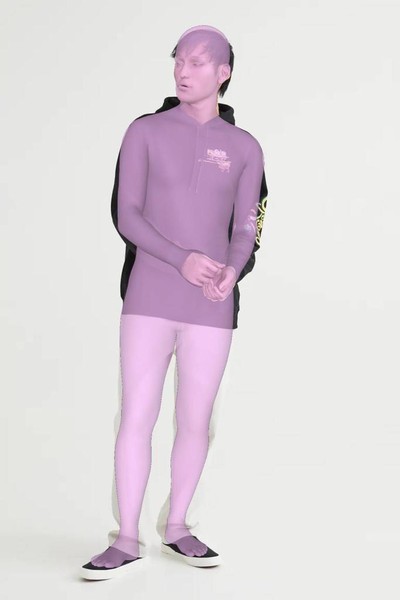}}

\subfloat{\includegraphics[height=0.24\textheight]{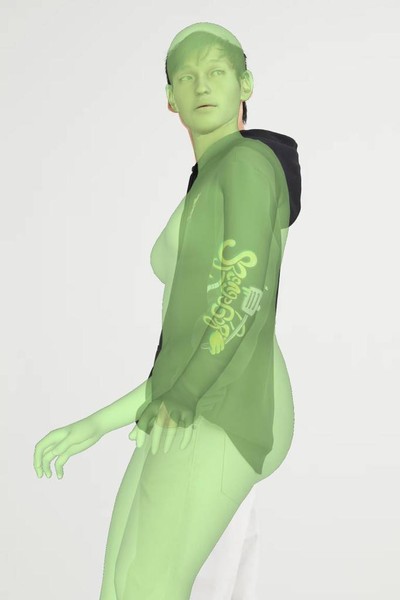}}
\subfloat{\includegraphics[height=0.24\textheight]{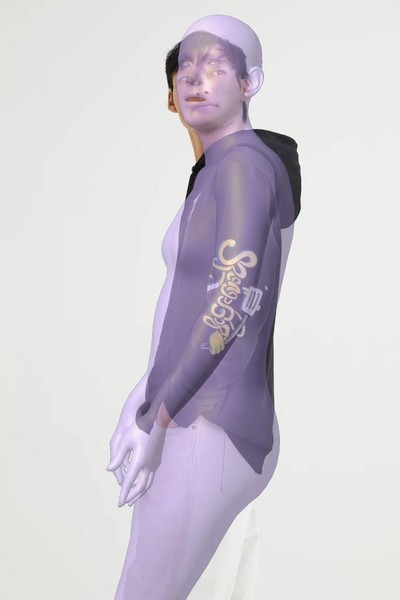}}
\subfloat{\includegraphics[height=0.24\textheight]{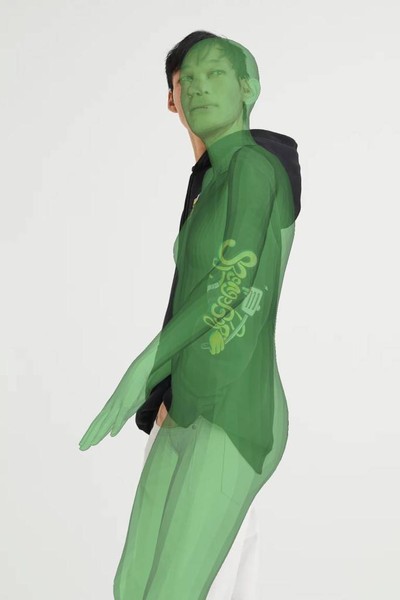}}
\subfloat{\includegraphics[height=0.24\textheight]{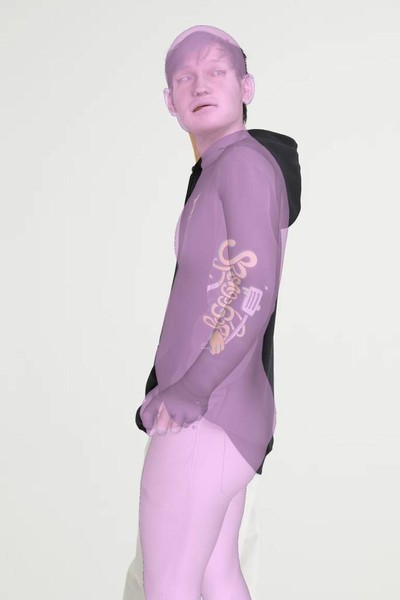}}

\subfloat[SMPLify-X \cite{pavlakos2019expressive}]
{\includegraphics[height=0.24\textheight]{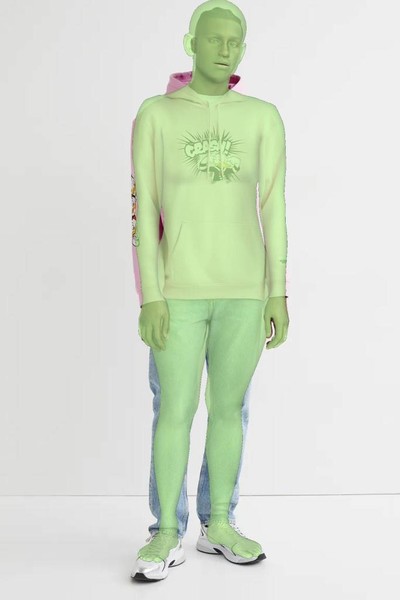}}
\subfloat[PyMAF-X \cite{pymafx2022}]{\includegraphics[height=0.24\textheight]{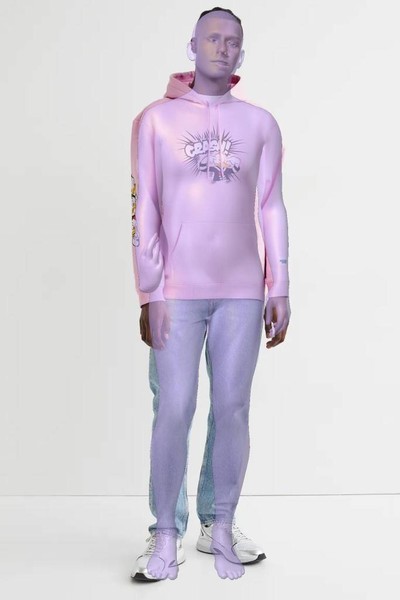}}
\subfloat[SHAPY \cite{choutas2022accurate}]
{\includegraphics[height=0.24\textheight]
{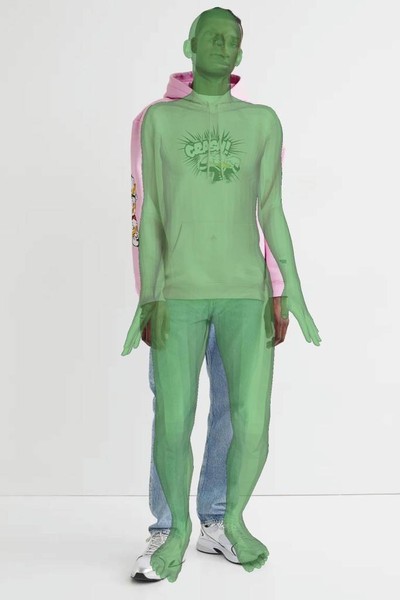}}
\subfloat[\KBody{-.1}{.035} (Ours)]{\includegraphics[height=0.24\textheight]{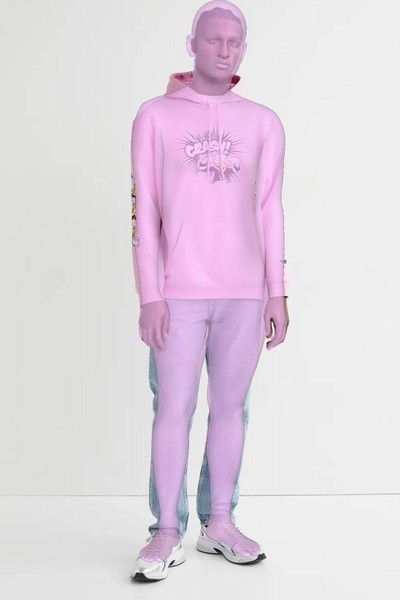}}

\caption{
Left-to-right: SMPLify-X \cite{pavlakos2019expressive} (\textcolor{caribbeangreen2}{light green}), PyMAF-X \cite{pymafx2022} (\textcolor{violet}{purple}), SHAPY \cite{choutas2022accurate} (\textcolor{jade}{green}) and KBody (\textcolor{candypink}{pink}).
}
\label{fig:p5}
\end{figure*}

%% file: figures/supp/oversized1.tex
\begin{figure*}[!htbp]
\captionsetup[subfigure]{position=bottom,labelformat=empty}

\centering

\subfloat{\includegraphics[height=0.24\textheight]{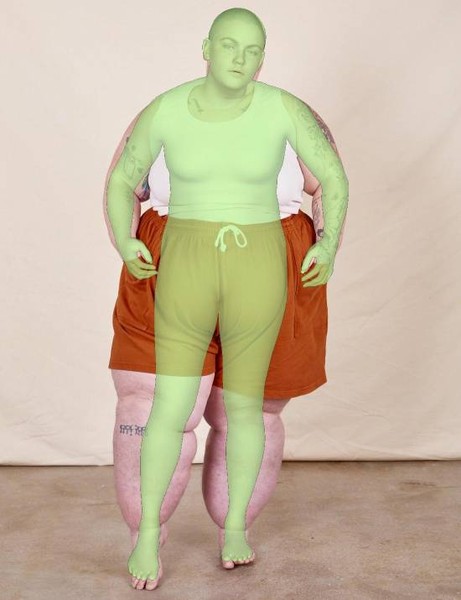}}
\subfloat{\includegraphics[height=0.24\textheight]{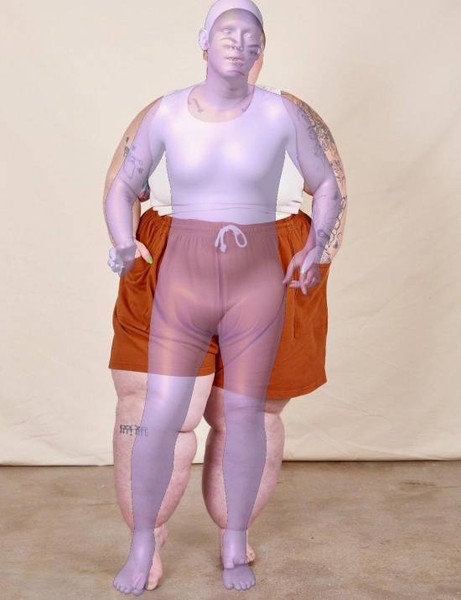}}
\subfloat{\includegraphics[height=0.24\textheight]{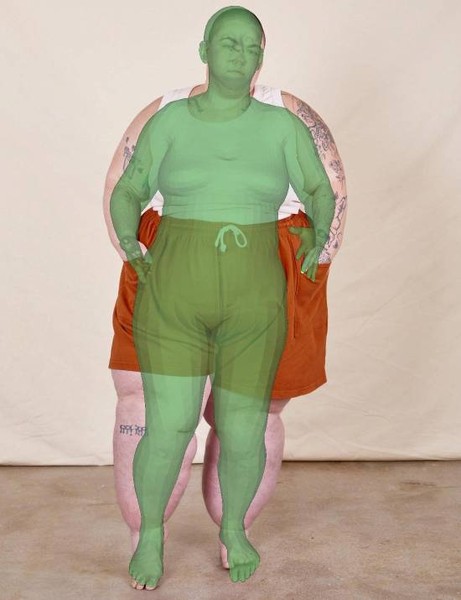}}
\subfloat{\includegraphics[height=0.24\textheight]{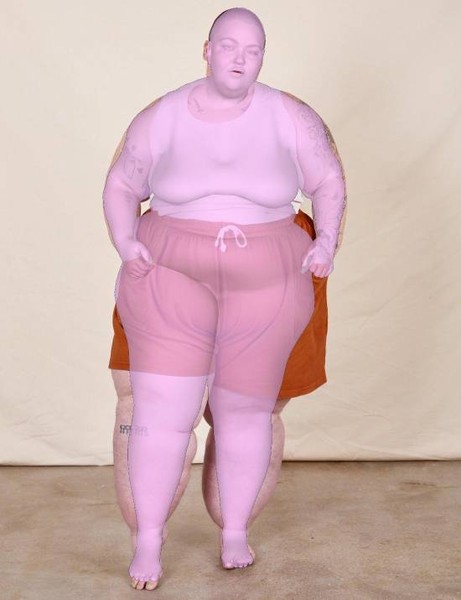}}

\subfloat{\includegraphics[height=0.24\textheight]{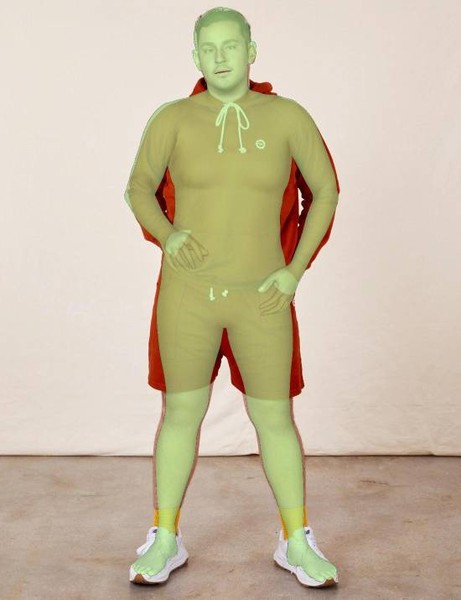}}
\subfloat{\includegraphics[height=0.24\textheight]{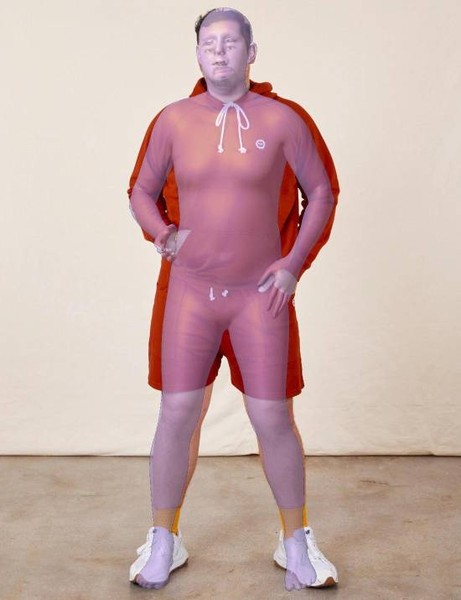}}
\subfloat{\includegraphics[height=0.24\textheight]{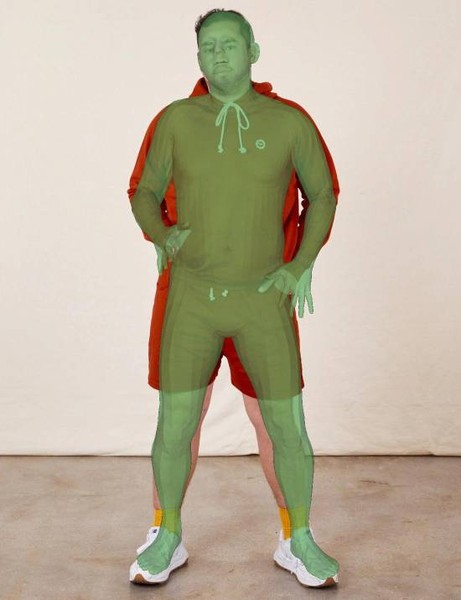}}
\subfloat{\includegraphics[height=0.24\textheight]{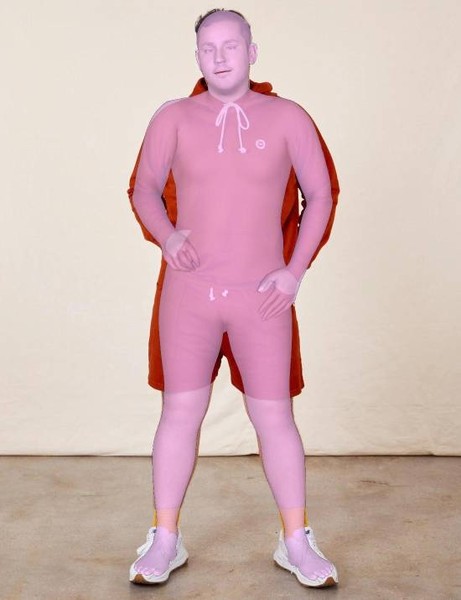}}

\subfloat{\includegraphics[height=0.24\textheight]{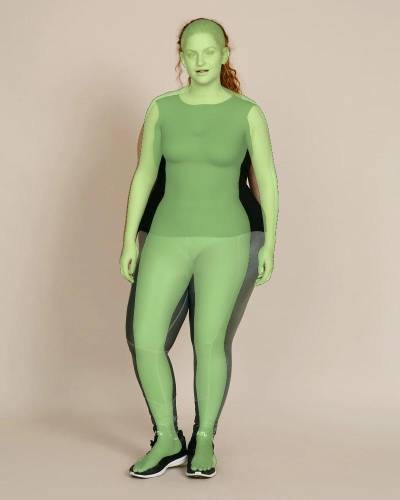}}
\subfloat{\includegraphics[height=0.24\textheight]{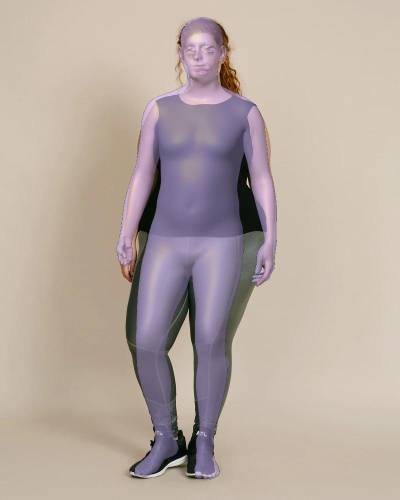}}
\subfloat{\includegraphics[height=0.24\textheight]{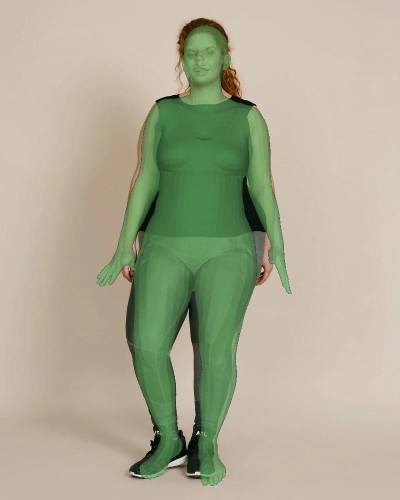}}
\subfloat{\includegraphics[height=0.24\textheight]{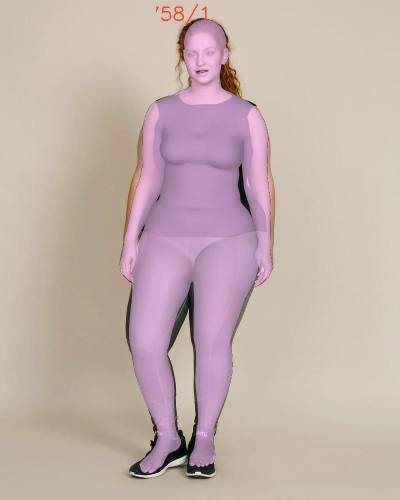}}

\subfloat[SMPLify-X \cite{pavlakos2019expressive}]
{\includegraphics[height=0.24\textheight]{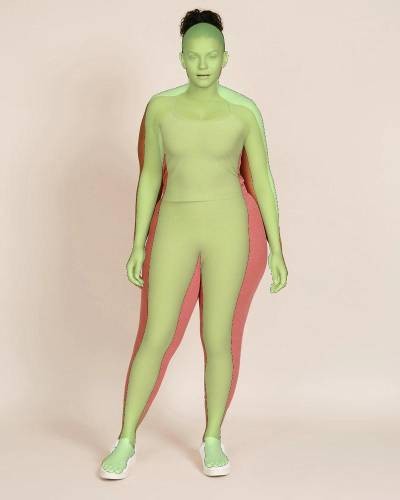}}
\subfloat[PyMAF-X \cite{pymafx2022}]{\includegraphics[height=0.24\textheight]{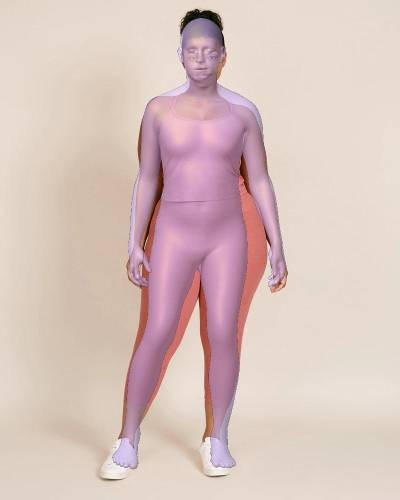}}
\subfloat[SHAPY \cite{choutas2022accurate}]
{\includegraphics[height=0.24\textheight]
{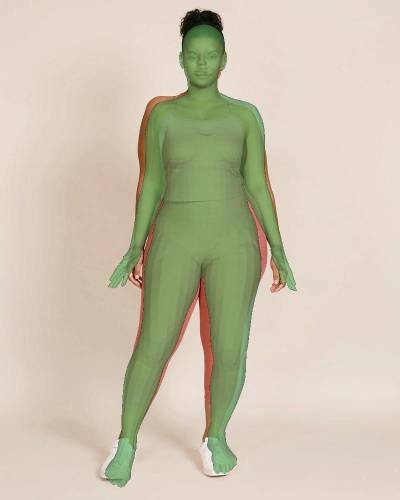}}
\subfloat[\KBody{-.1}{.035} (Ours)]{\includegraphics[height=0.24\textheight]{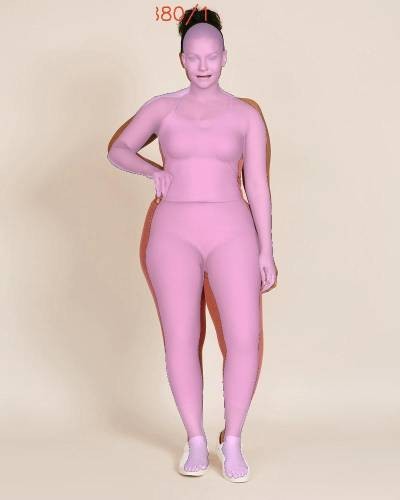}}

\caption{
Left-to-right: SMPLify-X \cite{pavlakos2019expressive} (\textcolor{caribbeangreen2}{light green}), PyMAF-X \cite{pymafx2022} (\textcolor{violet}{purple}), SHAPY \cite{choutas2022accurate} (\textcolor{jade}{green}) and KBody (\textcolor{candypink}{pink}).
}
\label{fig:ov1}
\end{figure*}

%% file: figures/supp/oversized2.tex
\begin{figure*}[!htbp]
\captionsetup[subfigure]{position=bottom,labelformat=empty}

\centering

\subfloat{\includegraphics[height=0.24\textheight]{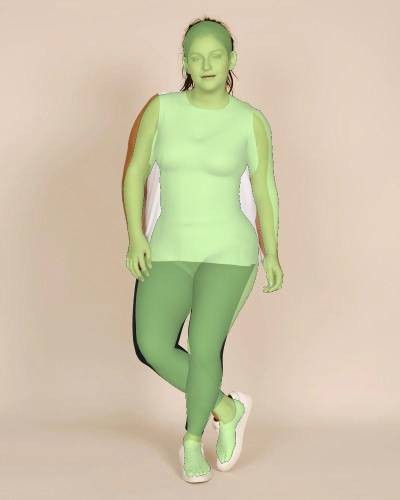}}
\subfloat{\includegraphics[height=0.24\textheight]{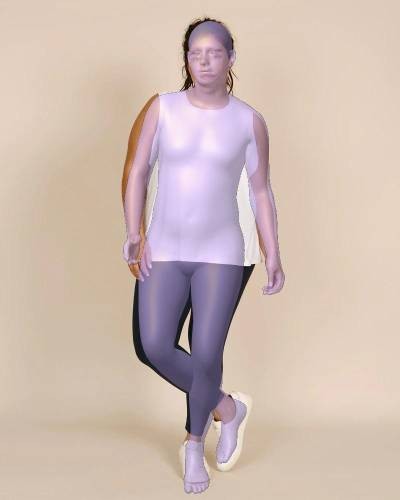}}
\subfloat{\includegraphics[height=0.24\textheight]{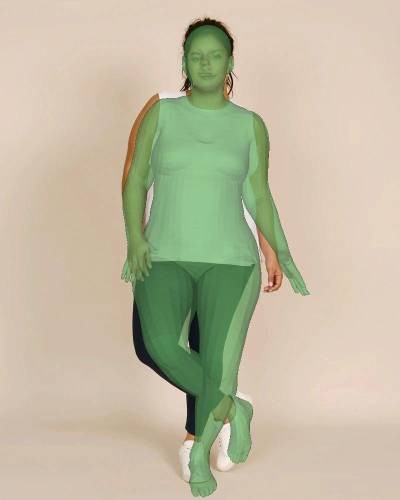}}
\subfloat{\includegraphics[height=0.24\textheight]{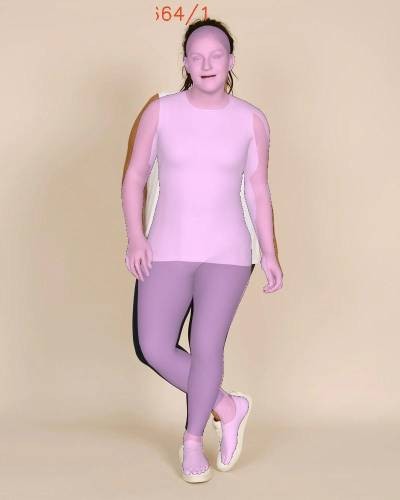}}

\subfloat{\includegraphics[height=0.24\textheight]{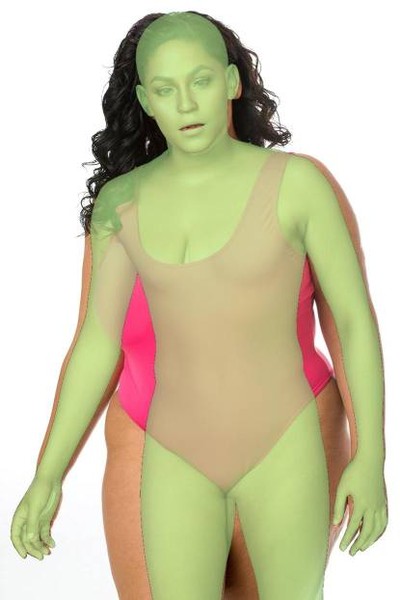}}
\subfloat{\includegraphics[height=0.24\textheight]{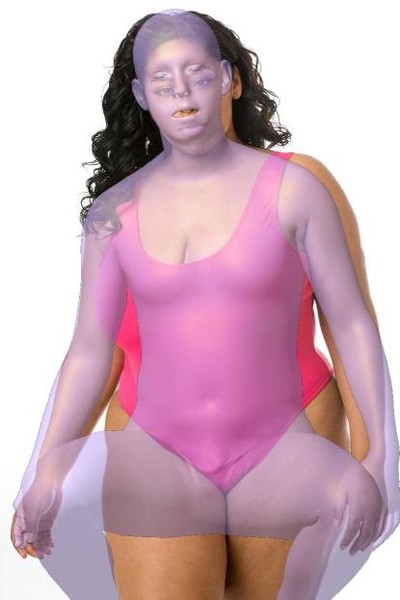}}
\subfloat{\includegraphics[height=0.24\textheight]{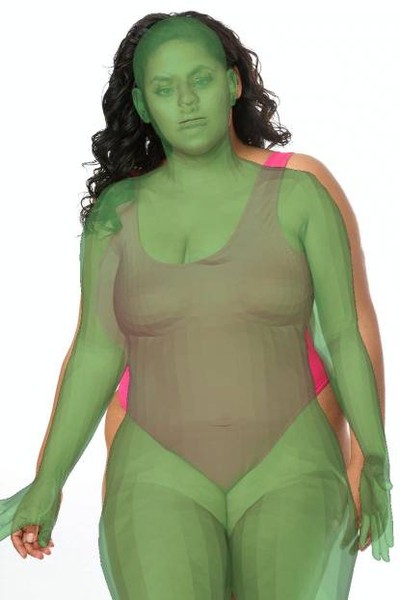}}
\subfloat{\includegraphics[height=0.24\textheight]{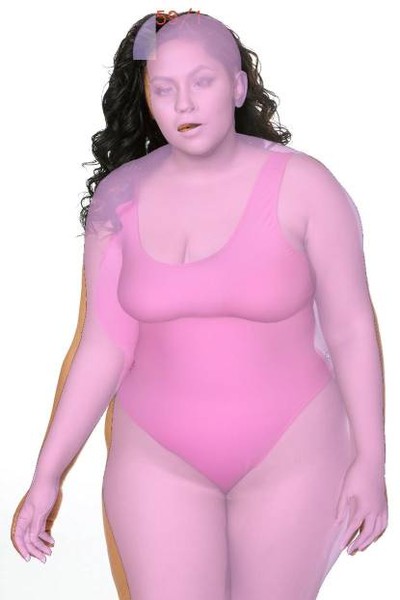}}

\subfloat{\includegraphics[height=0.24\textheight]{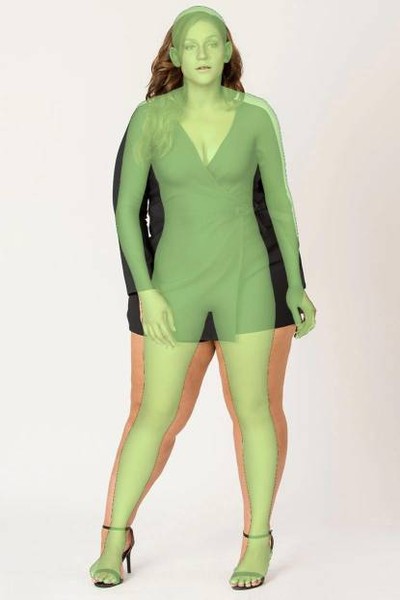}}
\subfloat{\includegraphics[height=0.24\textheight]{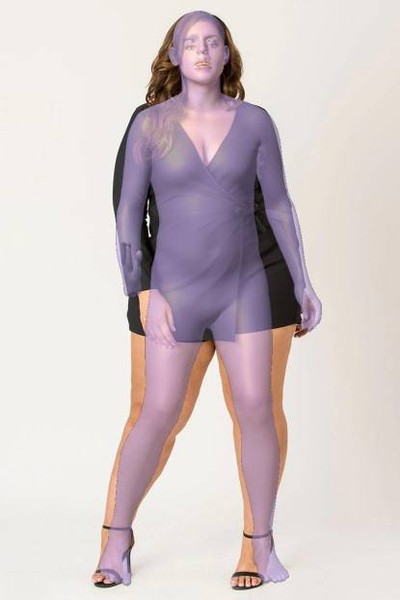}}
\subfloat{\includegraphics[height=0.24\textheight]{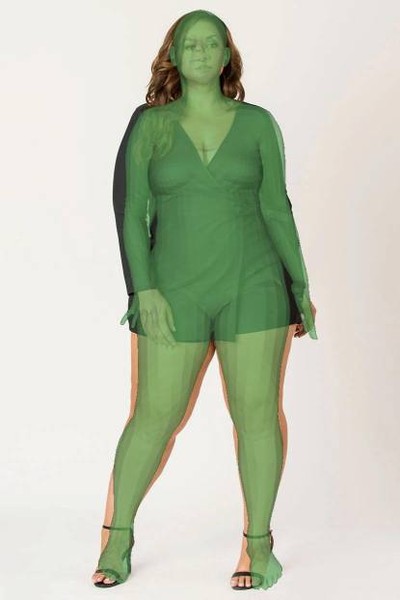}}
\subfloat{\includegraphics[height=0.24\textheight]{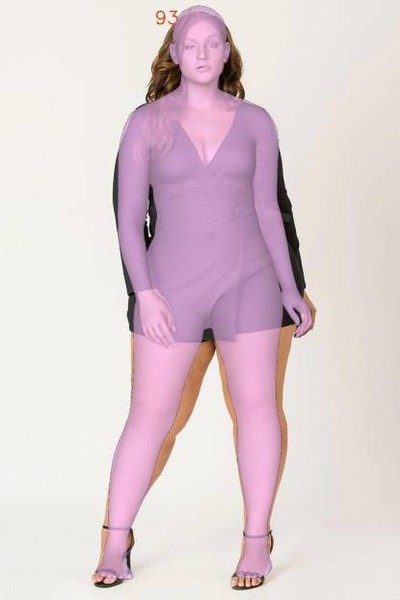}}

\subfloat[SMPLify-X \cite{pavlakos2019expressive}]
{\includegraphics[height=0.24\textheight]{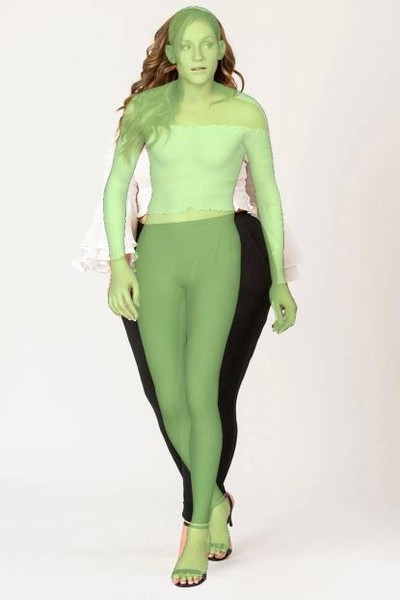}}
\subfloat[PyMAF-X \cite{pymafx2022}]{\includegraphics[height=0.24\textheight]{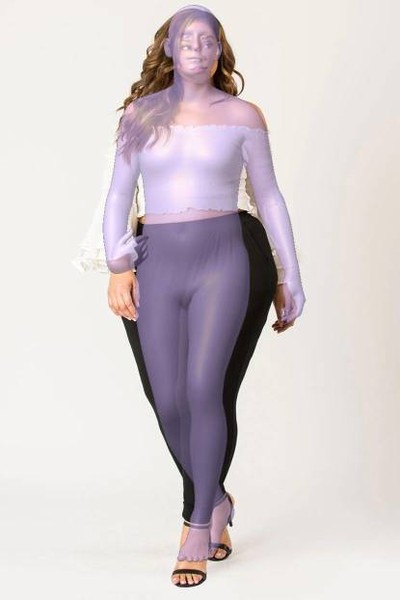}}
\subfloat[SHAPY \cite{choutas2022accurate}]
{\includegraphics[height=0.24\textheight]
{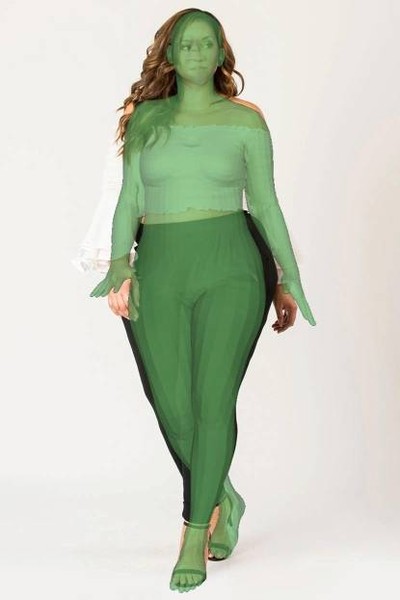}}
\subfloat[\KBody{-.1}{.035} (Ours)]{\includegraphics[height=0.24\textheight]{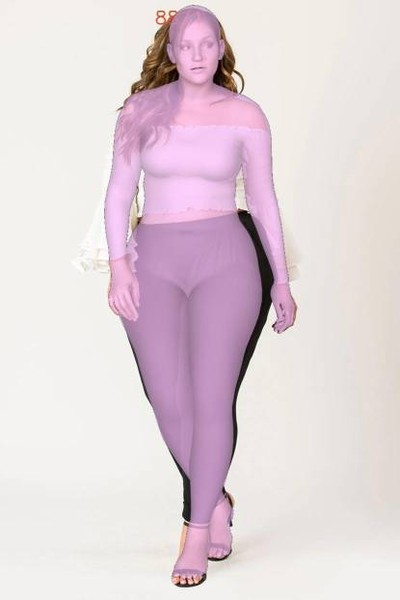}}

\caption{
Left-to-right: SMPLify-X \cite{pavlakos2019expressive} (\textcolor{caribbeangreen2}{light green}), PyMAF-X \cite{pymafx2022} (\textcolor{violet}{purple}), SHAPY \cite{choutas2022accurate} (\textcolor{jade}{green}) and KBody (\textcolor{candypink}{pink}).
}
\label{fig:ov2}
\end{figure*}

%% file: figures/supp/oversized3.tex
\begin{figure*}[!htbp]
\captionsetup[subfigure]{position=bottom,labelformat=empty}

\centering

\subfloat{\includegraphics[height=0.24\textheight]{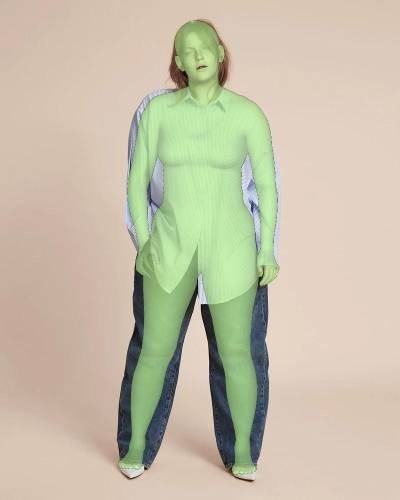}}
\subfloat{\includegraphics[height=0.24\textheight]{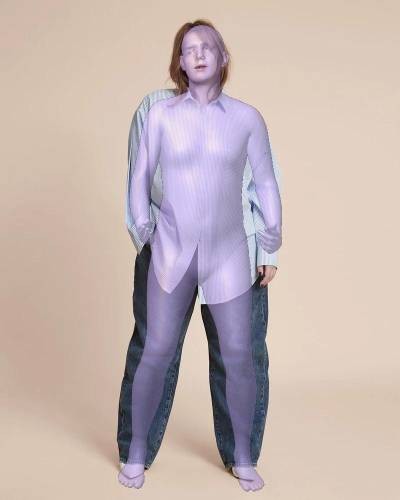}}
\subfloat{\includegraphics[height=0.24\textheight]{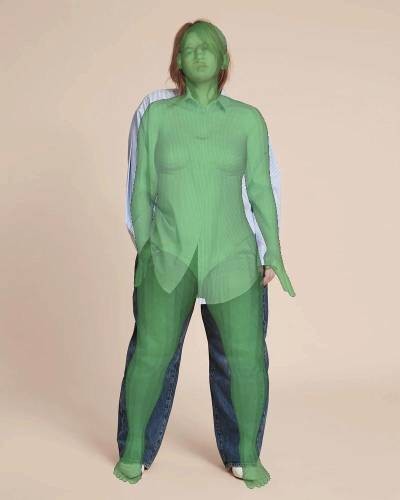}}
\subfloat{\includegraphics[height=0.24\textheight]{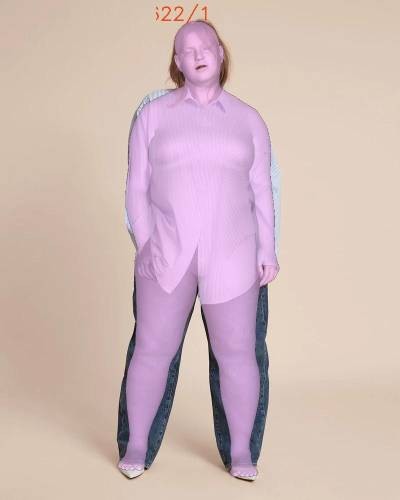}}

\subfloat{\includegraphics[height=0.24\textheight]{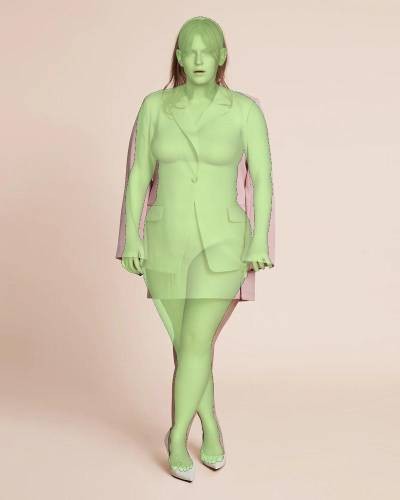}}
\subfloat{\includegraphics[height=0.24\textheight]{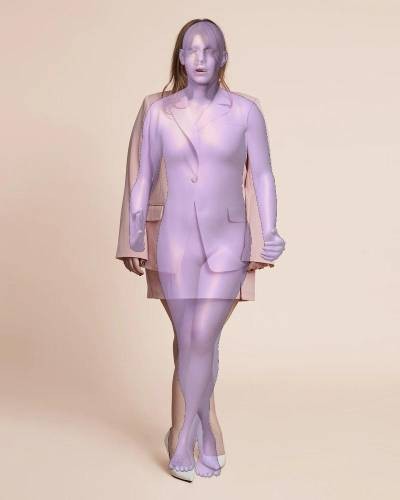}}
\subfloat{\includegraphics[height=0.24\textheight]{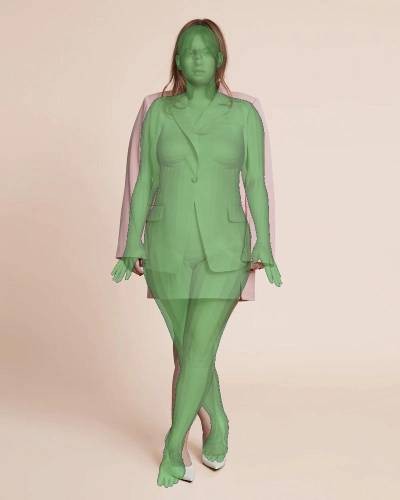}}
\subfloat{\includegraphics[height=0.24\textheight]{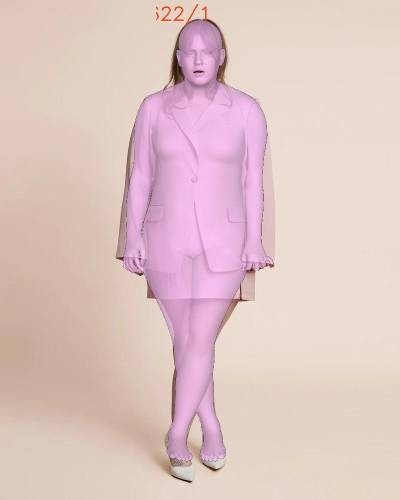}}

\subfloat{\includegraphics[height=0.24\textheight]{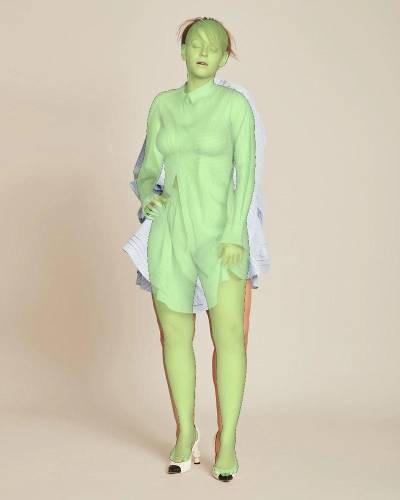}}
\subfloat{\includegraphics[height=0.24\textheight]{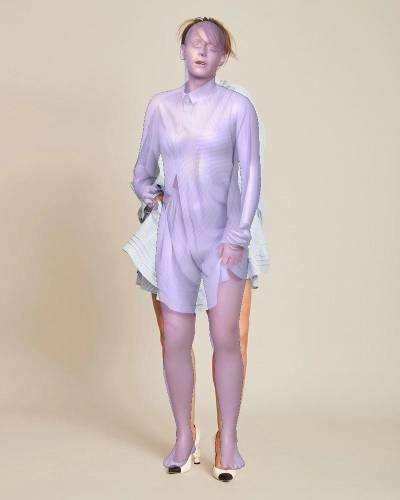}}
\subfloat{\includegraphics[height=0.24\textheight]{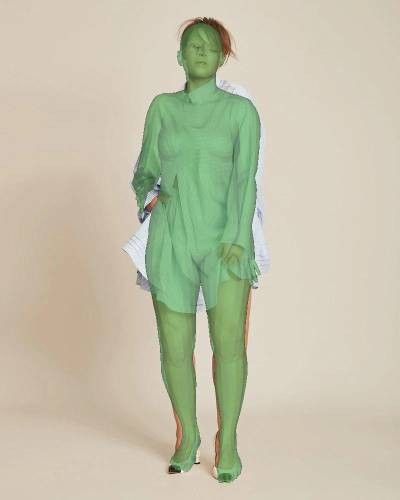}}
\subfloat{\includegraphics[height=0.24\textheight]{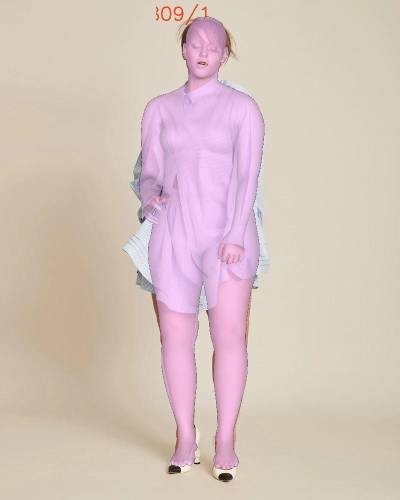}}

\subfloat[SMPLify-X \cite{pavlakos2019expressive}]
{\includegraphics[height=0.24\textheight]{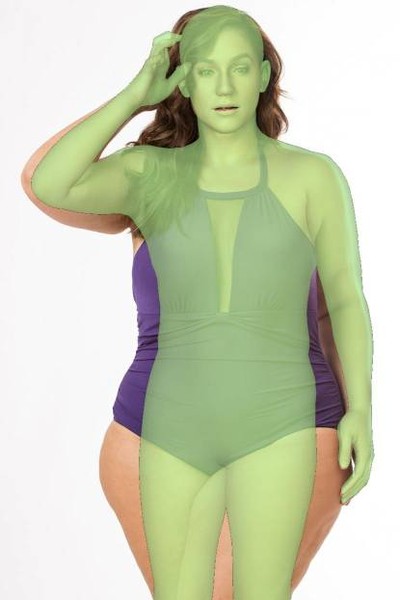}}
\subfloat[PyMAF-X \cite{pymafx2022}]{\includegraphics[height=0.24\textheight]{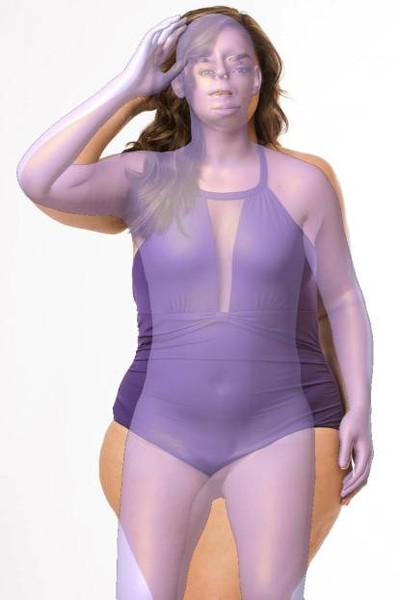}}
\subfloat[SHAPY \cite{choutas2022accurate}]
{\includegraphics[height=0.24\textheight]
{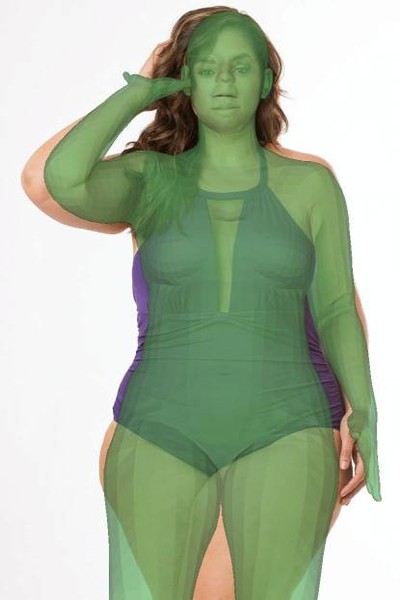}}
\subfloat[\KBody{-.1}{.035} (Ours)]{\includegraphics[height=0.24\textheight]{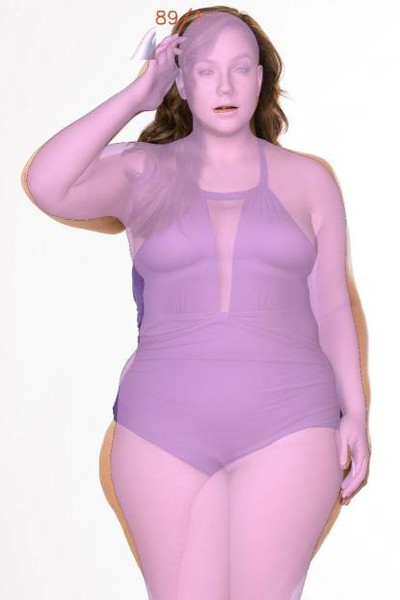}}

\caption{
Left-to-right: SMPLify-X \cite{pavlakos2019expressive} (\textcolor{caribbeangreen2}{light green}), PyMAF-X \cite{pymafx2022} (\textcolor{violet}{purple}), SHAPY \cite{choutas2022accurate} (\textcolor{jade}{green}) and KBody (\textcolor{candypink}{pink}).
}
\label{fig:ov3}
\end{figure*}

%% file: figures/supp/veronica1.tex
\begin{figure*}[!htbp]
\captionsetup[subfigure]{position=bottom,labelformat=empty}

\centering

\subfloat{\includegraphics[height=0.24\textheight]{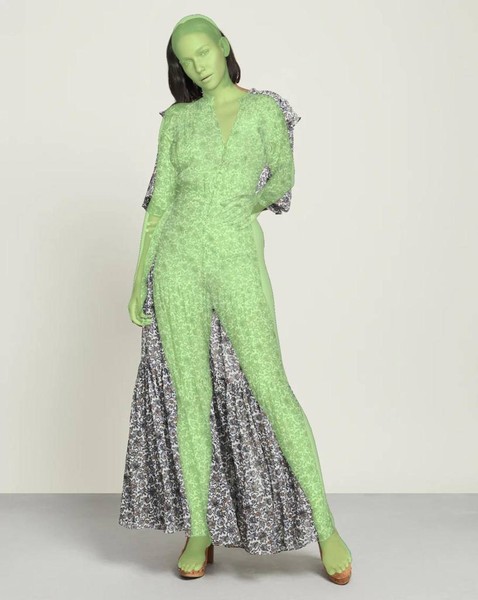}}
\subfloat{\includegraphics[height=0.24\textheight]{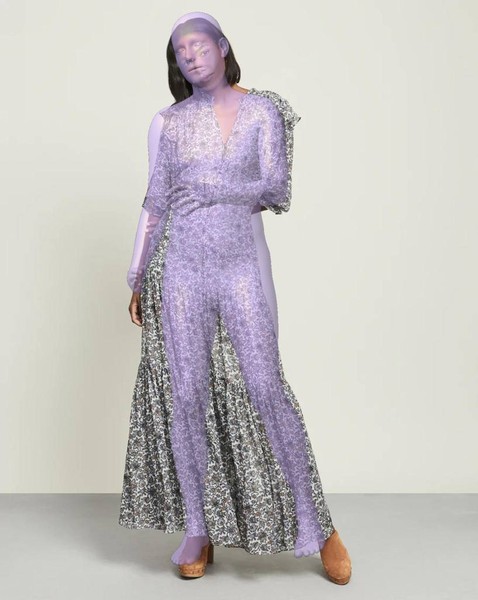}}
\subfloat{\includegraphics[height=0.24\textheight]{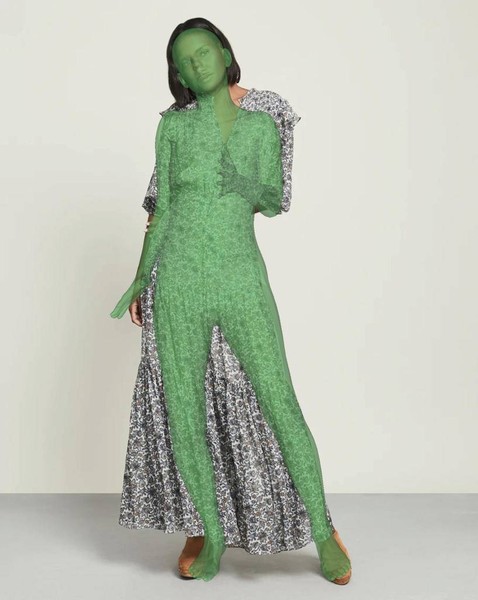}}
\subfloat{\includegraphics[height=0.24\textheight]{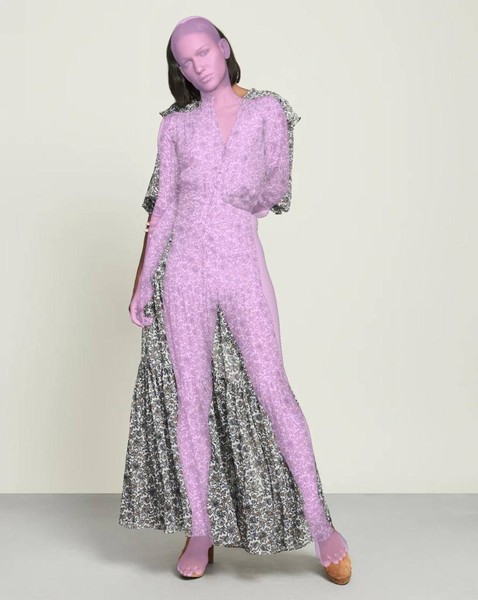}}

\subfloat{\includegraphics[height=0.24\textheight]{images/results/full/veronica/smplifyx_001.jpg}}
\subfloat{\includegraphics[height=0.24\textheight]{images/results/full/veronica/pymafx_001.jpg}}
\subfloat{\includegraphics[height=0.24\textheight]{images/results/full/veronica/shapy_001.jpg}}
\subfloat{\includegraphics[height=0.24\textheight]{images/results/full/veronica/kbody_001.jpg}}

\subfloat{\includegraphics[height=0.24\textheight]{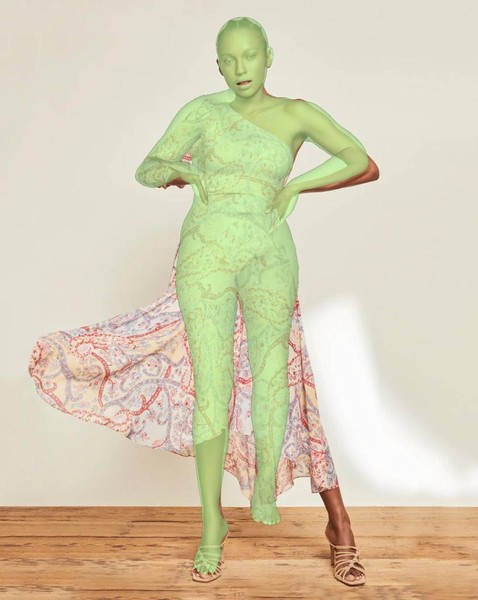}}
\subfloat{\includegraphics[height=0.24\textheight]{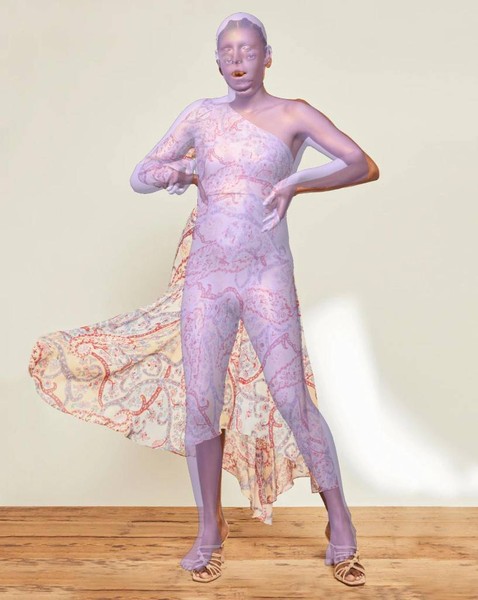}}
\subfloat{\includegraphics[height=0.24\textheight]{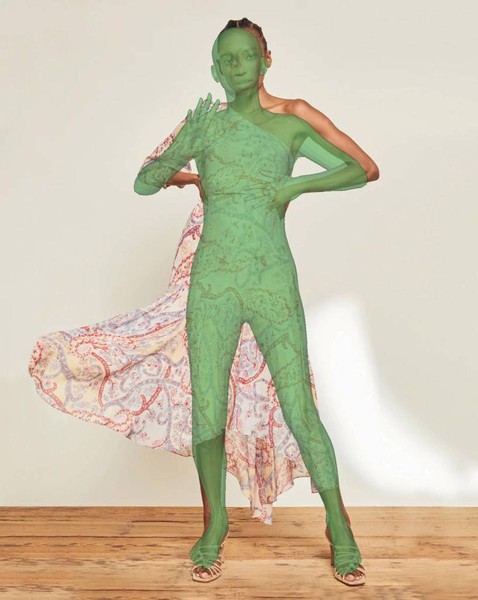}}
\subfloat{\includegraphics[height=0.24\textheight]{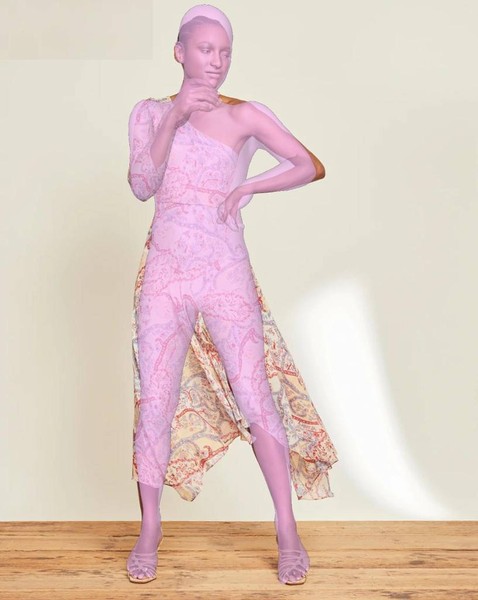}}

\subfloat[SMPLify-X \cite{pavlakos2019expressive}]
{\includegraphics[height=0.24\textheight]{images/results/full/veronica/smplifyx_003.jpg}}
\subfloat[PyMAF-X \cite{pymafx2022}]{\includegraphics[height=0.24\textheight]{images/results/full/veronica/pymafx_003.jpg}}
\subfloat[SHAPY \cite{choutas2022accurate}]
{\includegraphics[height=0.24\textheight]
{images/results/full/veronica/shapy_003.jpg}}
\subfloat[\KBody{-.1}{.035} (Ours)]{\includegraphics[height=0.24\textheight]{images/results/full/veronica/kbody_003.jpg}}

\caption{
Left-to-right: SMPLify-X \cite{pavlakos2019expressive} (\textcolor{caribbeangreen2}{light green}), PyMAF-X \cite{pymafx2022} (\textcolor{violet}{purple}), SHAPY \cite{choutas2022accurate} (\textcolor{jade}{green}) and KBody (\textcolor{candypink}{pink}).
}
\label{fig:v1}
\end{figure*}

%% file: figures/supp/veronica2.tex
\begin{figure*}[!htbp]
\captionsetup[subfigure]{position=bottom,labelformat=empty}

\centering

\subfloat{\includegraphics[height=0.24\textheight]{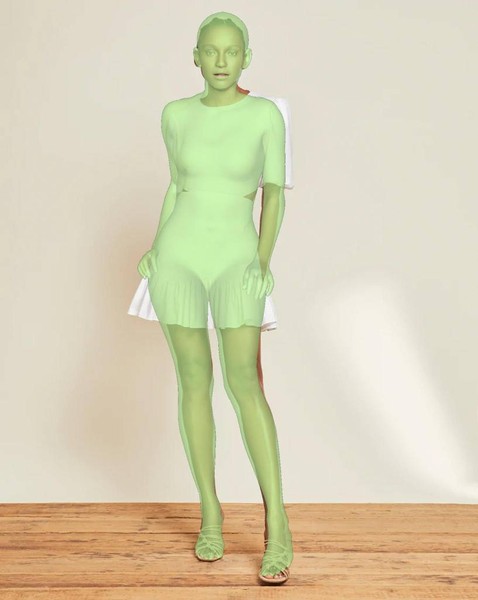}}
\subfloat{\includegraphics[height=0.24\textheight]{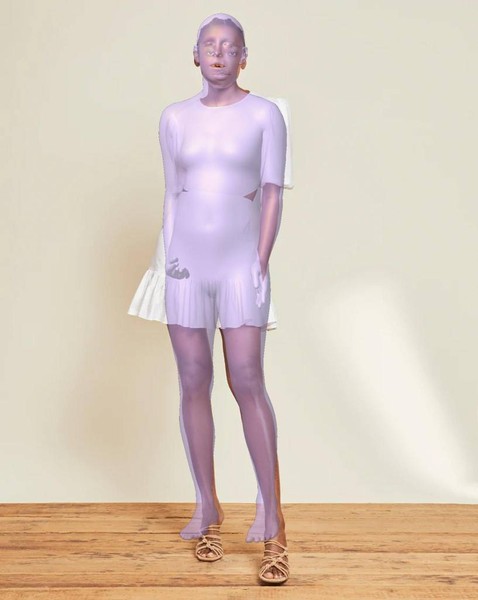}}
\subfloat{\includegraphics[height=0.24\textheight]{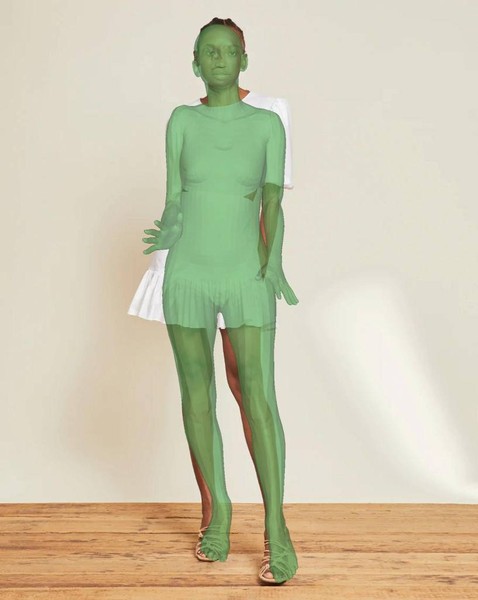}}
\subfloat{\includegraphics[height=0.24\textheight]{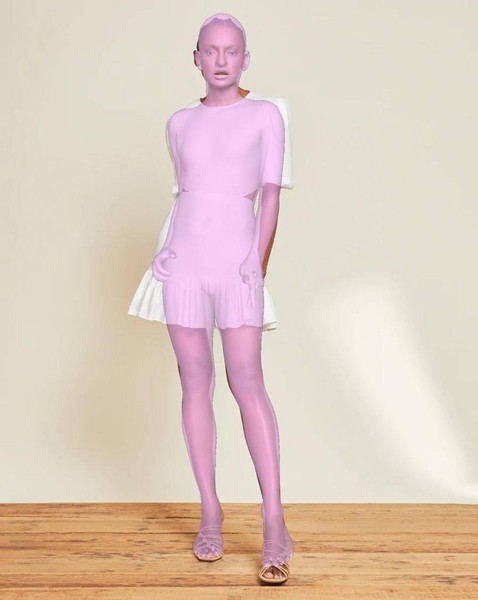}}

\subfloat{\includegraphics[height=0.24\textheight]{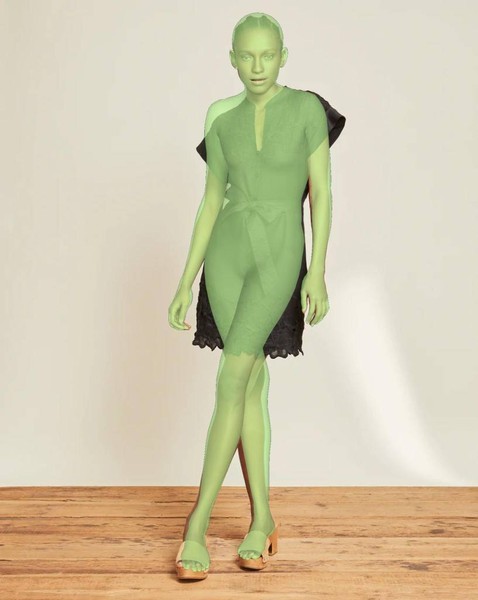}}
\subfloat{\includegraphics[height=0.24\textheight]{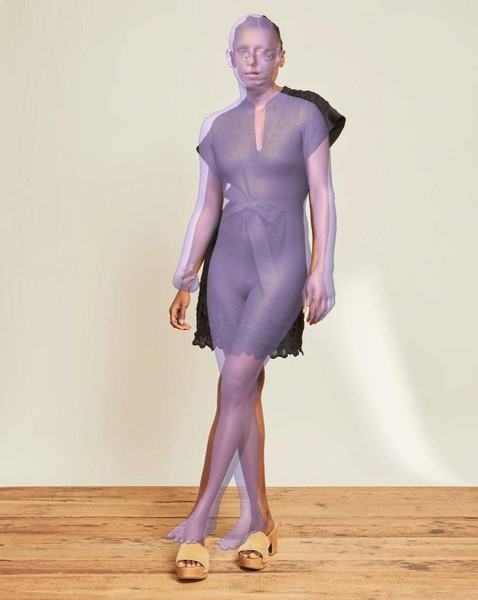}}
\subfloat{\includegraphics[height=0.24\textheight]{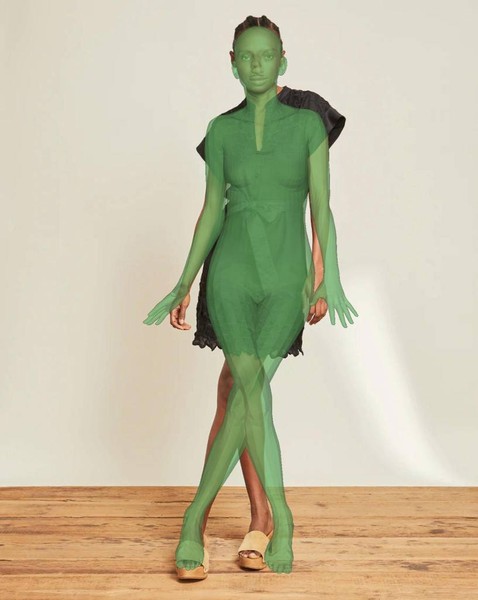}}
\subfloat{\includegraphics[height=0.24\textheight]{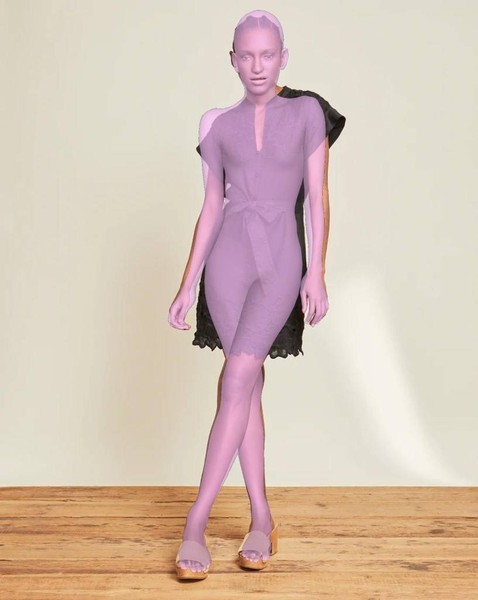}}

\subfloat{\includegraphics[height=0.24\textheight]{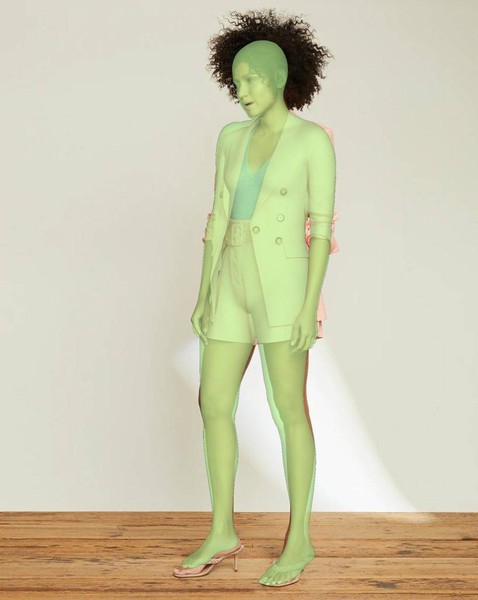}}
\subfloat{\includegraphics[height=0.24\textheight]{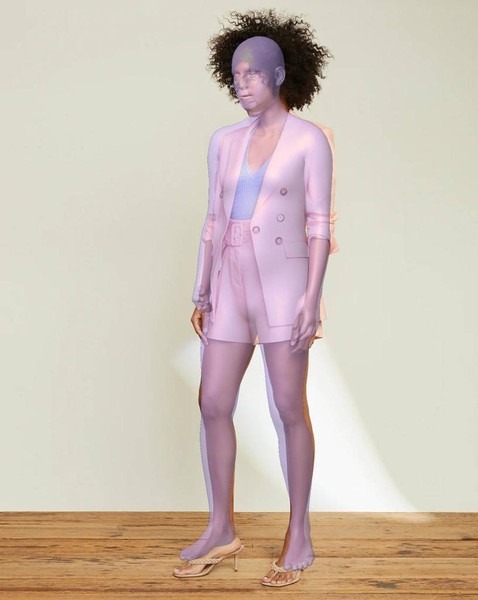}}
\subfloat{\includegraphics[height=0.24\textheight]{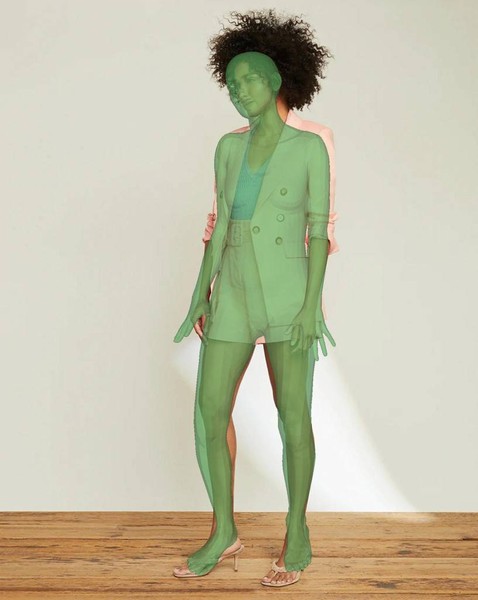}}
\subfloat{\includegraphics[height=0.24\textheight]{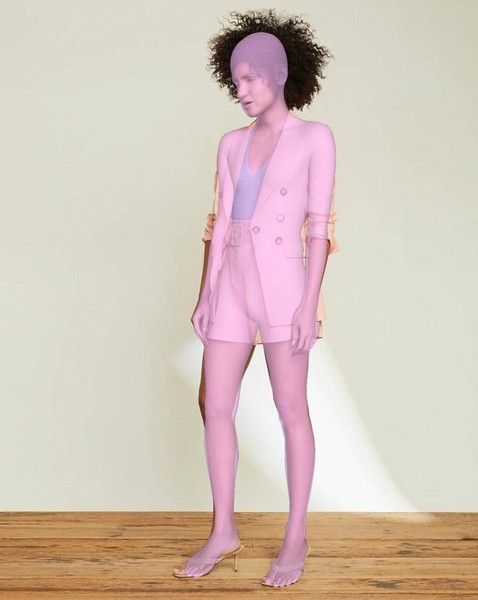}}

\subfloat[SMPLify-X \cite{pavlakos2019expressive}]
{\includegraphics[height=0.24\textheight]{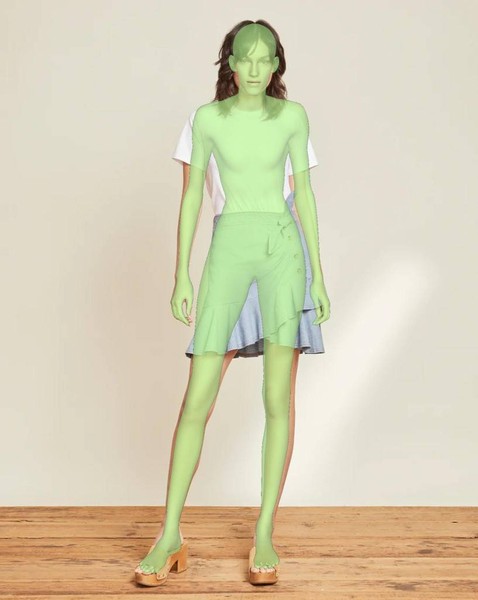}}
\subfloat[PyMAF-X \cite{pymafx2022}]{\includegraphics[height=0.24\textheight]{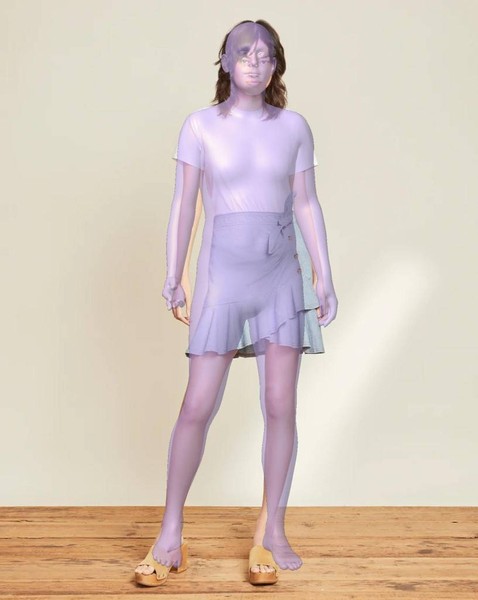}}
\subfloat[SHAPY \cite{choutas2022accurate}]
{\includegraphics[height=0.24\textheight]
{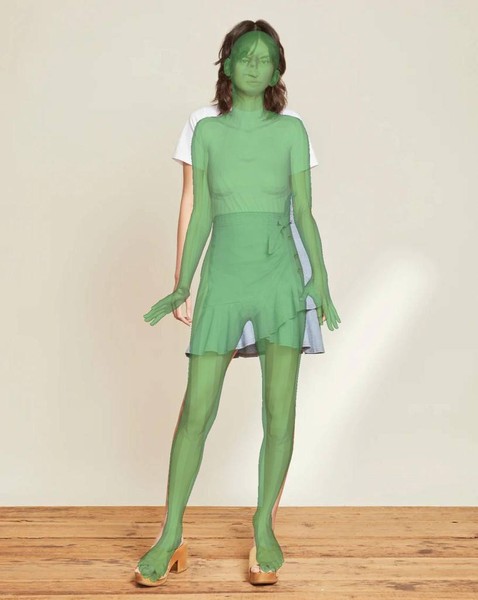}}
\subfloat[\KBody{-.1}{.035} (Ours)]{\includegraphics[height=0.24\textheight]{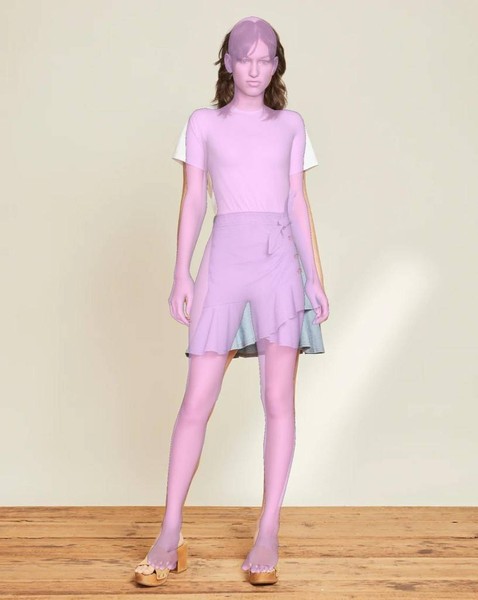}}

\caption{
Left-to-right: SMPLify-X \cite{pavlakos2019expressive} (\textcolor{caribbeangreen2}{light green}), PyMAF-X \cite{pymafx2022} (\textcolor{violet}{purple}), SHAPY \cite{choutas2022accurate} (\textcolor{jade}{green}) and KBody (\textcolor{candypink}{pink}).
}
\label{fig:v2}
\end{figure*}

%% file: figures/supp/veronica3.tex
\begin{figure*}[!htbp]
\captionsetup[subfigure]{position=bottom,labelformat=empty}

\centering

\subfloat{\includegraphics[height=0.24\textheight]{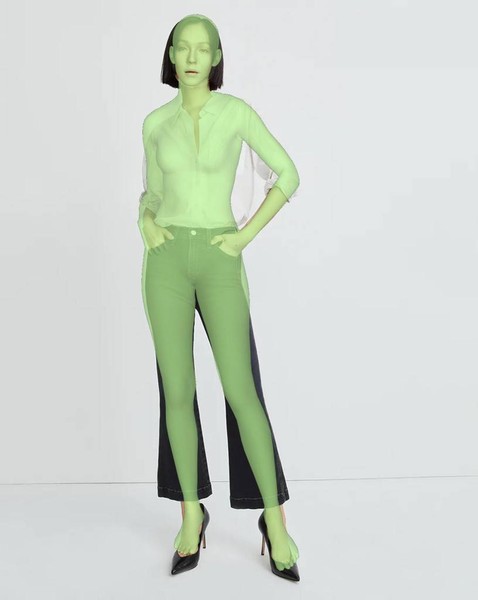}}
\subfloat{\includegraphics[height=0.24\textheight]{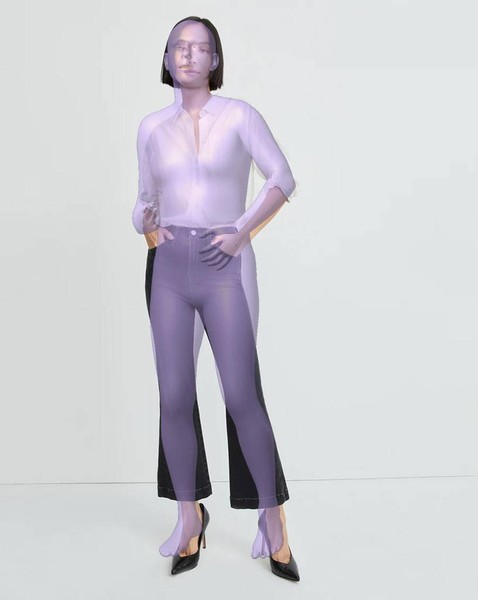}}
\subfloat{\includegraphics[height=0.24\textheight]{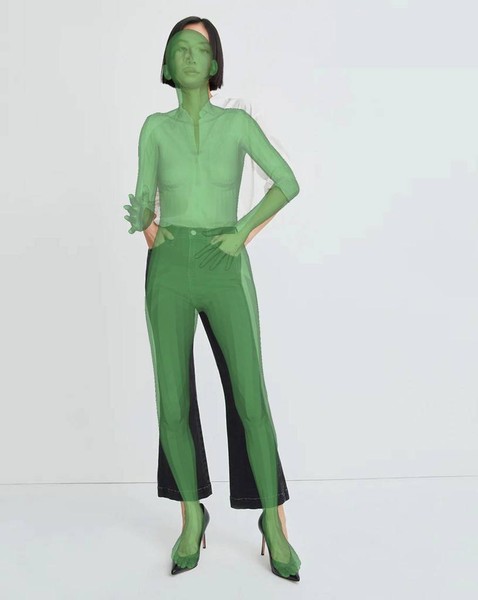}}
\subfloat{\includegraphics[height=0.24\textheight]{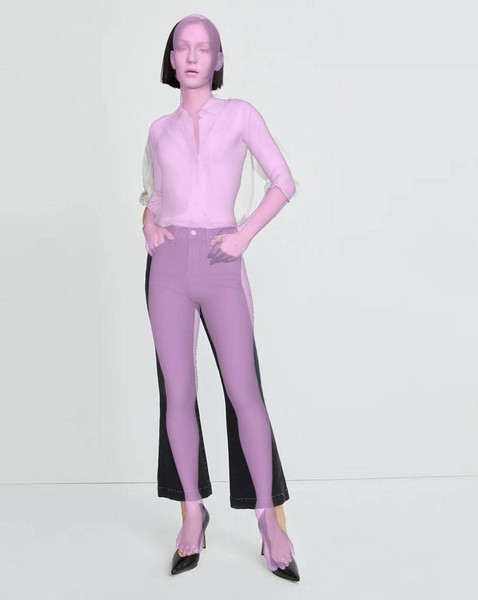}}

\subfloat{\includegraphics[height=0.24\textheight]{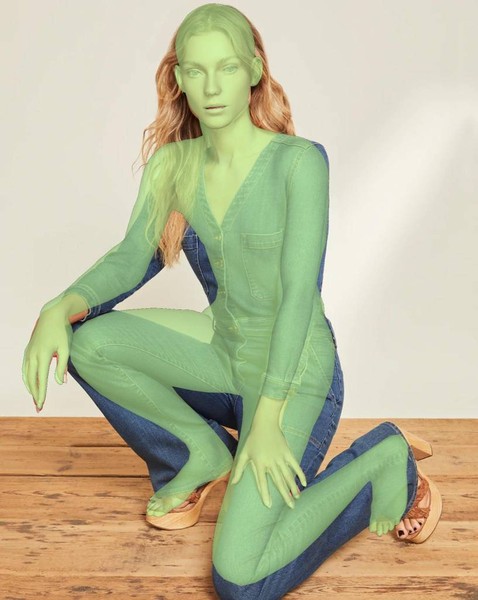}}
\subfloat{\includegraphics[height=0.24\textheight]{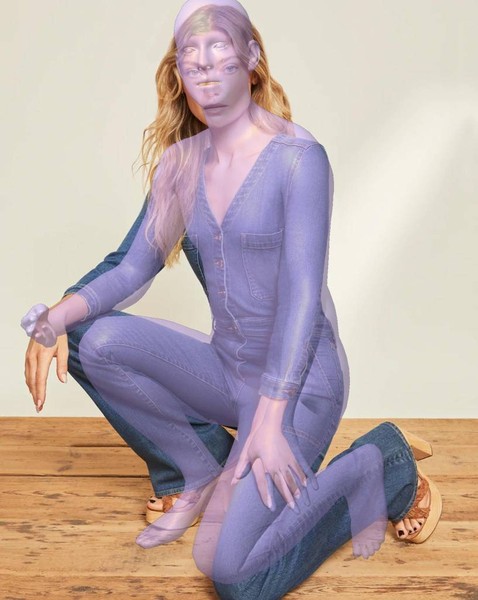}}
\subfloat{\includegraphics[height=0.24\textheight]{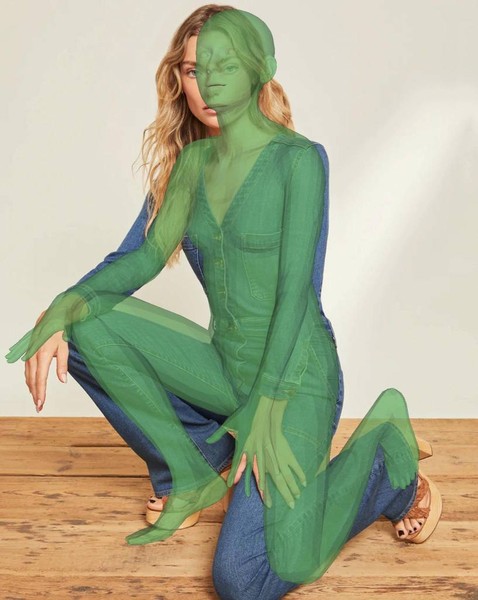}}
\subfloat{\includegraphics[height=0.24\textheight]{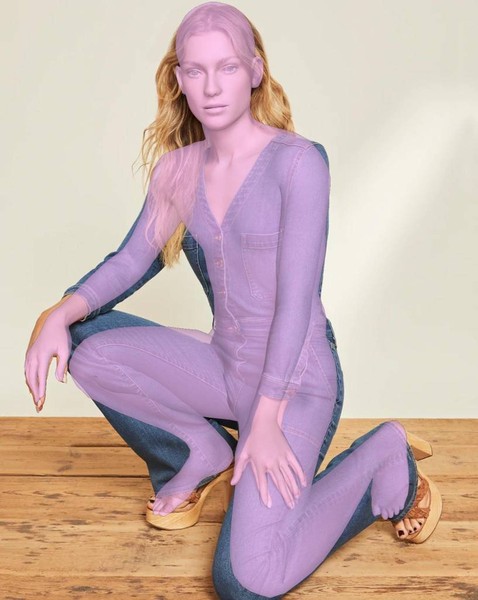}}

\subfloat{\includegraphics[height=0.24\textheight]{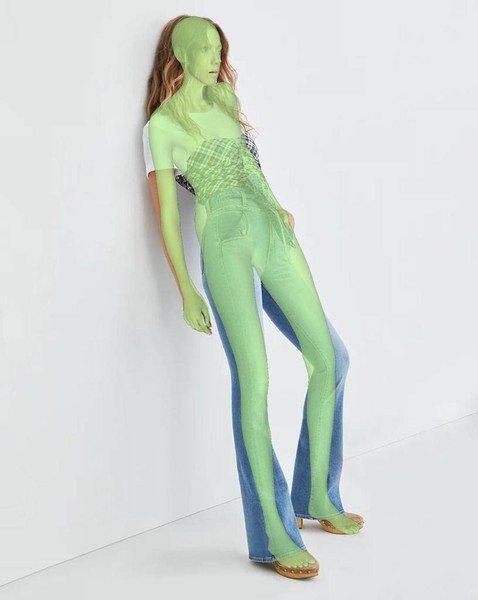}}
\subfloat{\includegraphics[height=0.24\textheight]{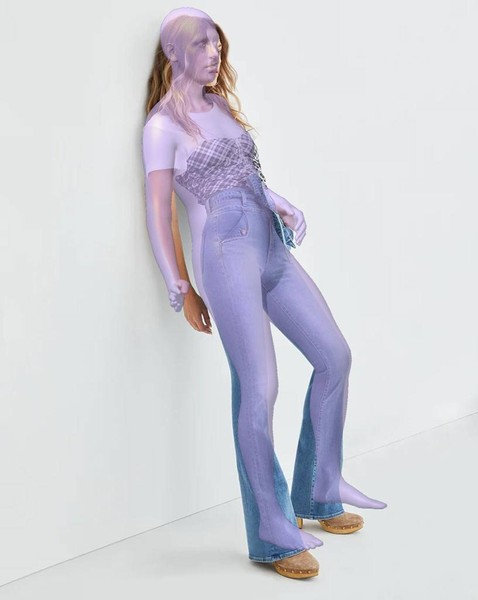}}
\subfloat{\includegraphics[height=0.24\textheight]{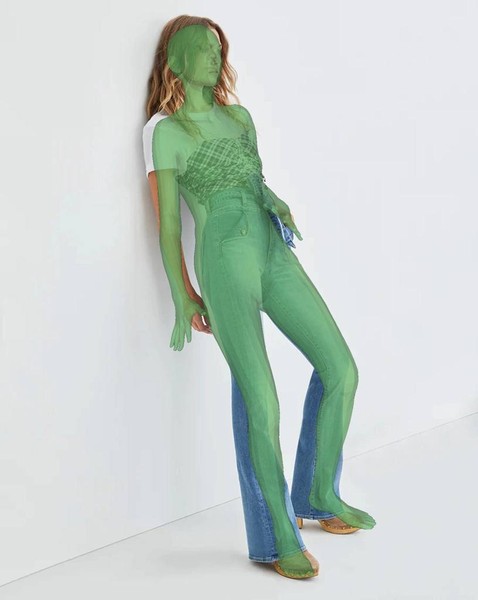}}
\subfloat{\includegraphics[height=0.24\textheight]{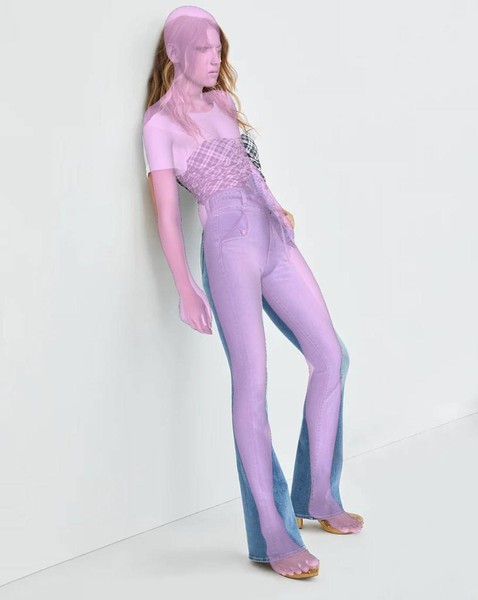}}

\subfloat[SMPLify-X \cite{pavlakos2019expressive}]
{\includegraphics[height=0.24\textheight]{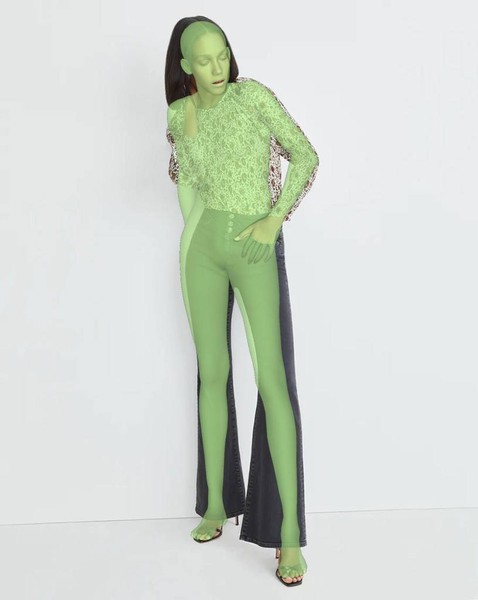}}
\subfloat[PyMAF-X \cite{pymafx2022}]{\includegraphics[height=0.24\textheight]{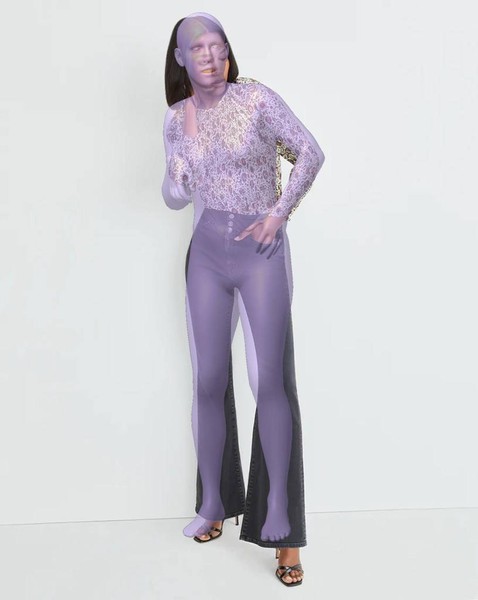}}
\subfloat[SHAPY \cite{choutas2022accurate}]
{\includegraphics[height=0.24\textheight]
{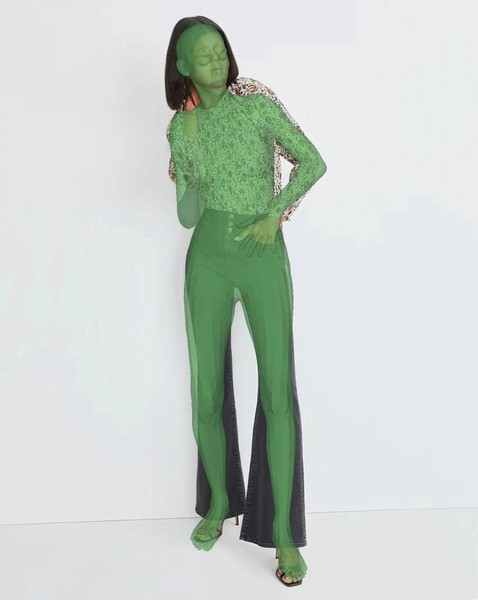}}
\subfloat[\KBody{-.1}{.035} (Ours)]{\includegraphics[height=0.24\textheight]{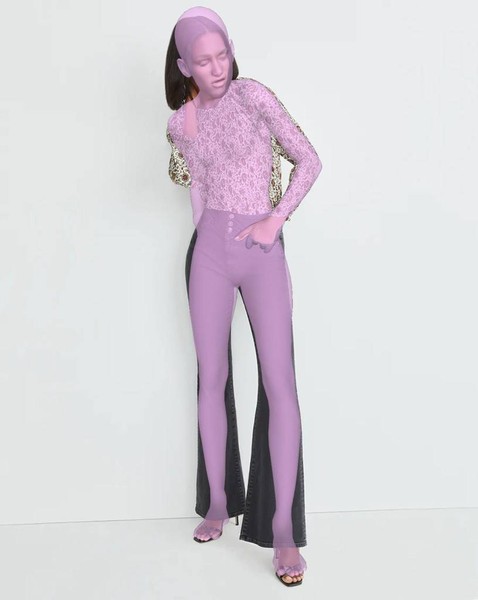}}

\caption{
Left-to-right: SMPLify-X \cite{pavlakos2019expressive} (\textcolor{caribbeangreen2}{light green}), PyMAF-X \cite{pymafx2022} (\textcolor{violet}{purple}), SHAPY \cite{choutas2022accurate} (\textcolor{jade}{green}) and KBody (\textcolor{candypink}{pink}).
}
\label{fig:v3}
\end{figure*}

%% file: figures/supp/womens_plus_size1.tex
\begin{figure*}[!htbp]
\captionsetup[subfigure]{position=bottom,labelformat=empty}

\centering

\subfloat{\includegraphics[height=0.24\textheight]{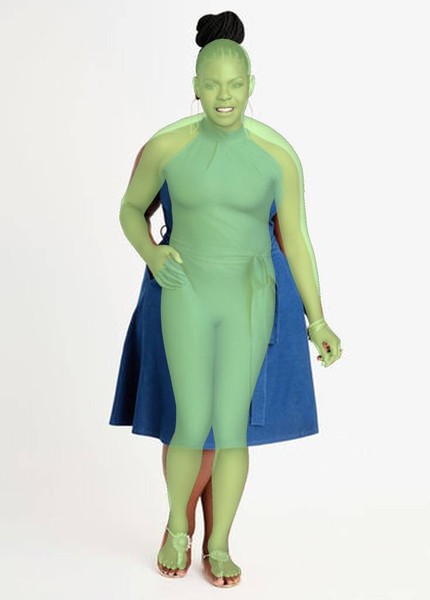}}
\subfloat{\includegraphics[height=0.24\textheight]{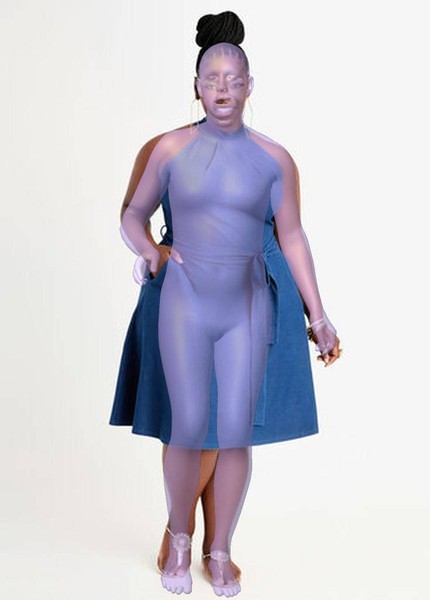}}
\subfloat{\includegraphics[height=0.24\textheight]{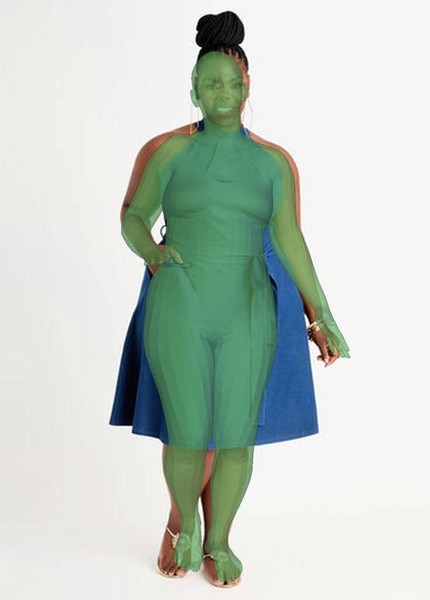}}
\subfloat{\includegraphics[height=0.24\textheight]{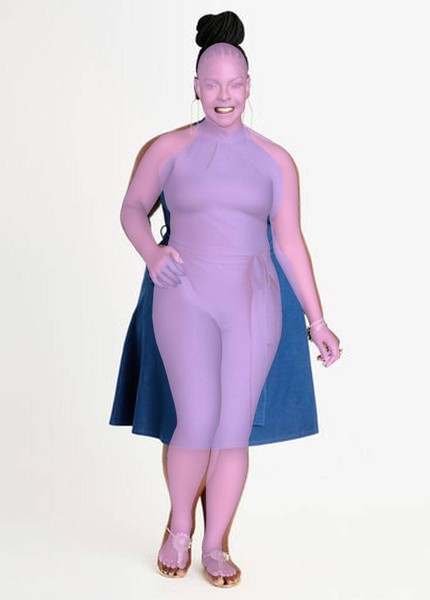}}

\subfloat{\includegraphics[height=0.24\textheight]{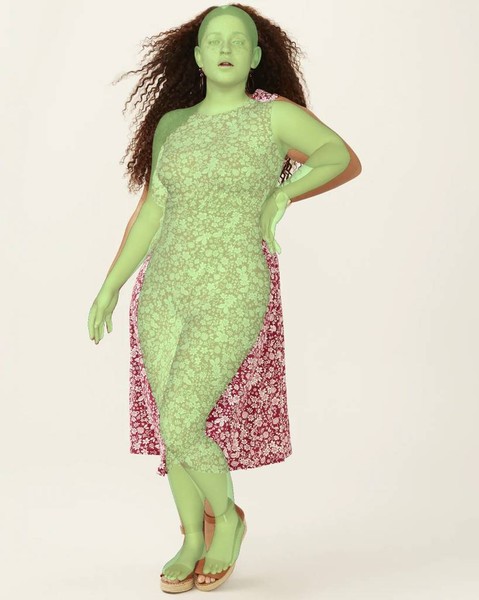}}
\subfloat{\includegraphics[height=0.24\textheight]{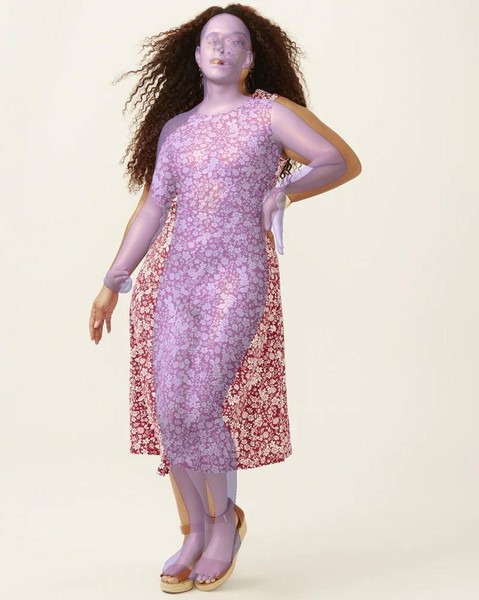}}
\subfloat{\includegraphics[height=0.24\textheight]{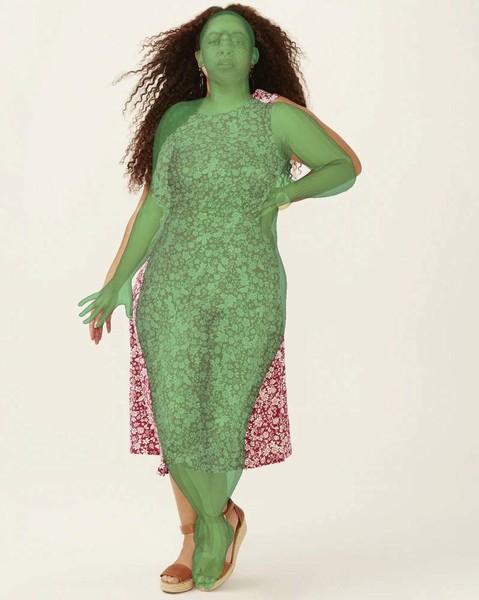}}
\subfloat{\includegraphics[height=0.24\textheight]{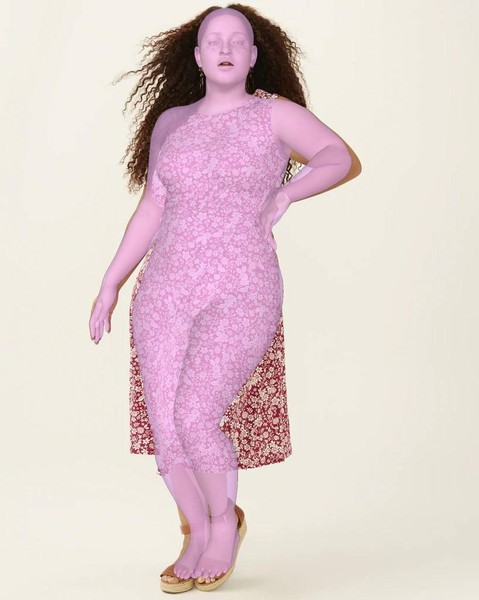}}

\subfloat{\includegraphics[height=0.24\textheight]{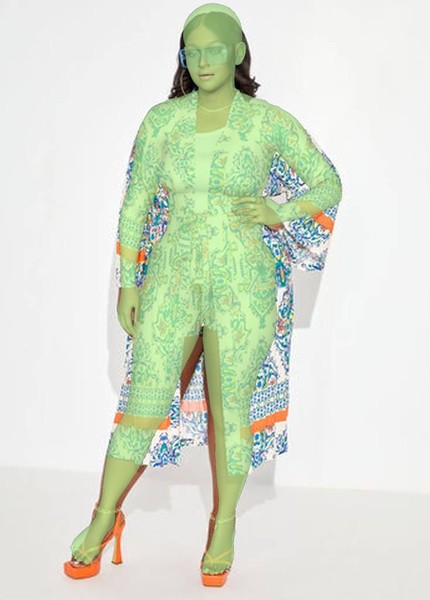}}
\subfloat{\includegraphics[height=0.24\textheight]{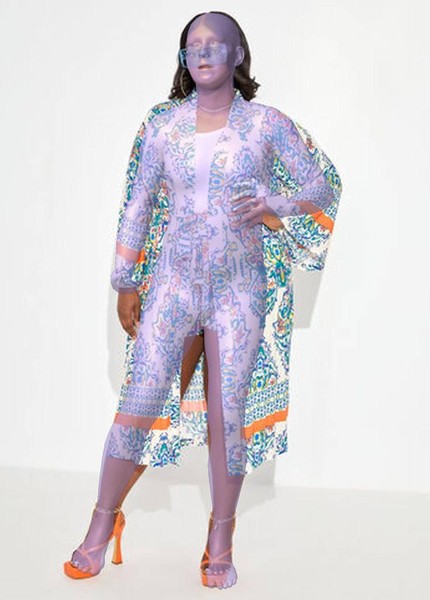}}
\subfloat{\includegraphics[height=0.24\textheight]{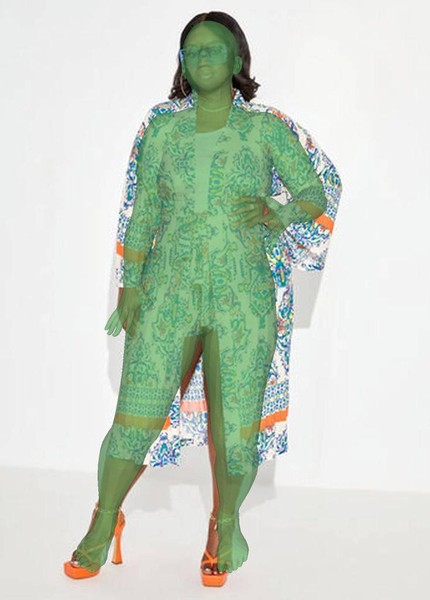}}
\subfloat{\includegraphics[height=0.24\textheight]{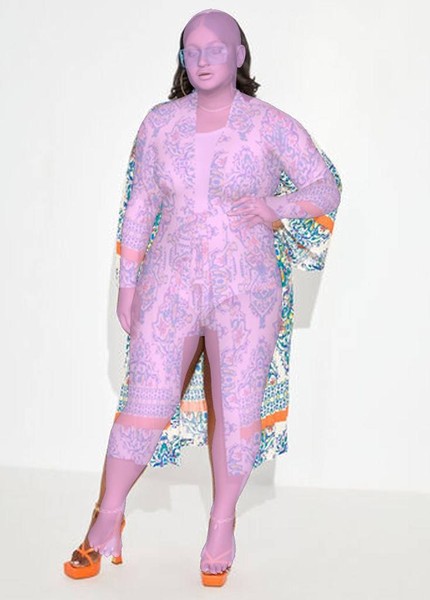}}

\subfloat[SMPLify-X \cite{pavlakos2019expressive}]
{\includegraphics[height=0.24\textheight]{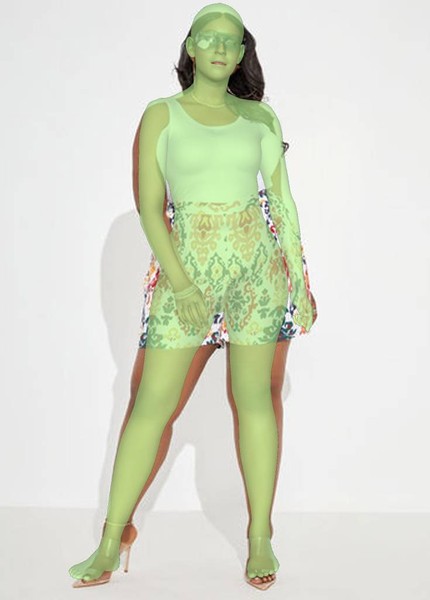}}
\subfloat[PyMAF-X \cite{pymafx2022}]{\includegraphics[height=0.24\textheight]{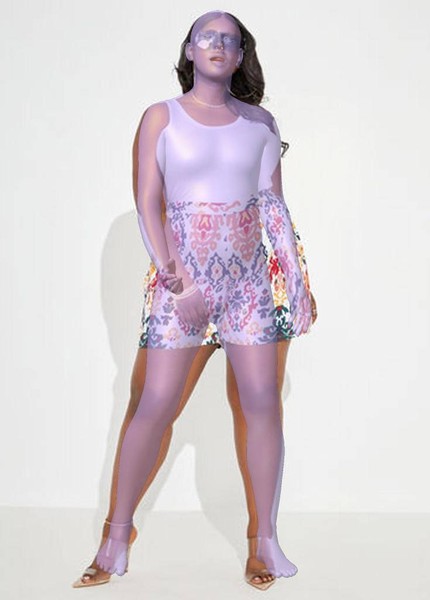}}
\subfloat[SHAPY \cite{choutas2022accurate}]
{\includegraphics[height=0.24\textheight]
{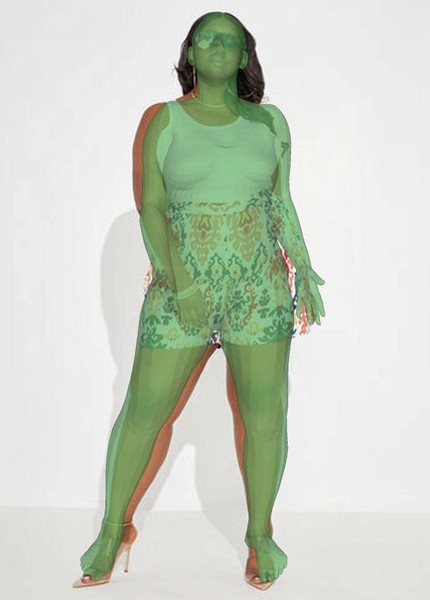}}
\subfloat[\KBody{-.1}{.035} (Ours)]{\includegraphics[height=0.24\textheight]{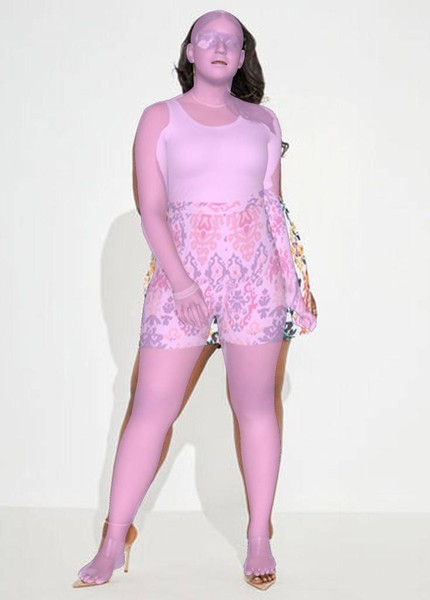}}

\caption{
Left-to-right: SMPLify-X \cite{pavlakos2019expressive} (\textcolor{caribbeangreen2}{light green}), PyMAF-X \cite{pymafx2022} (\textcolor{violet}{purple}), SHAPY \cite{choutas2022accurate} (\textcolor{jade}{green}) and KBody (\textcolor{candypink}{pink}).
}
\label{fig:w1}
\end{figure*}

%% file: figures/supp/womens_plus_size2.tex
\begin{figure*}[!htbp]
\captionsetup[subfigure]{position=bottom,labelformat=empty}

\centering

\subfloat{\includegraphics[height=0.24\textheight]{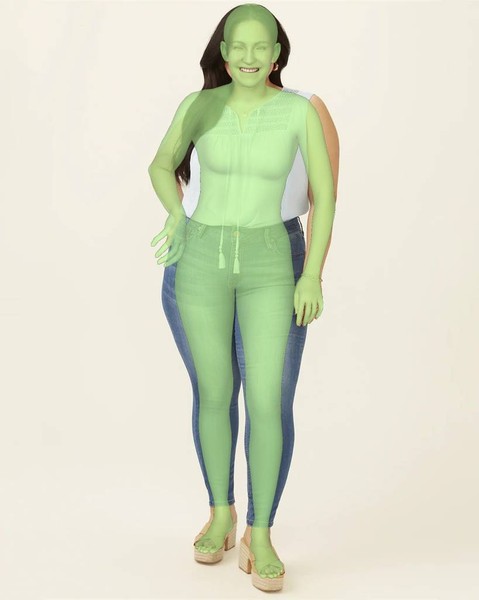}}
\subfloat{\includegraphics[height=0.24\textheight]{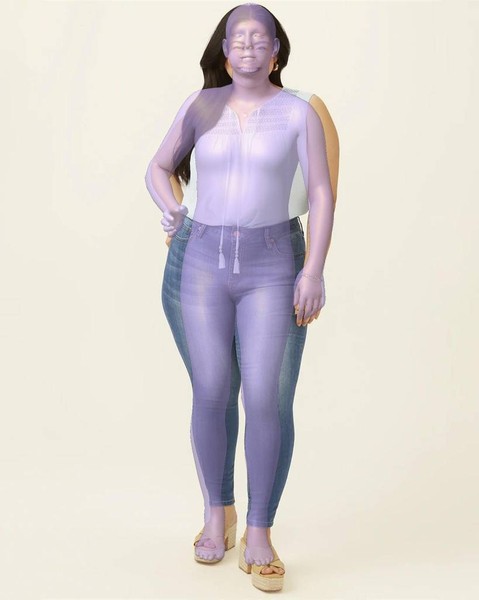}}
\subfloat{\includegraphics[height=0.24\textheight]{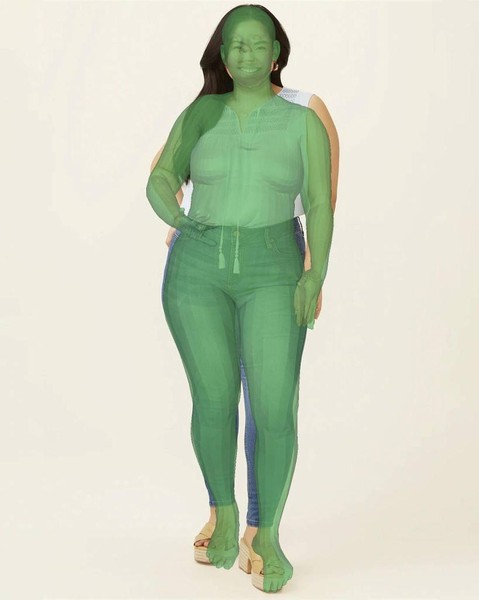}}
\subfloat{\includegraphics[height=0.24\textheight]{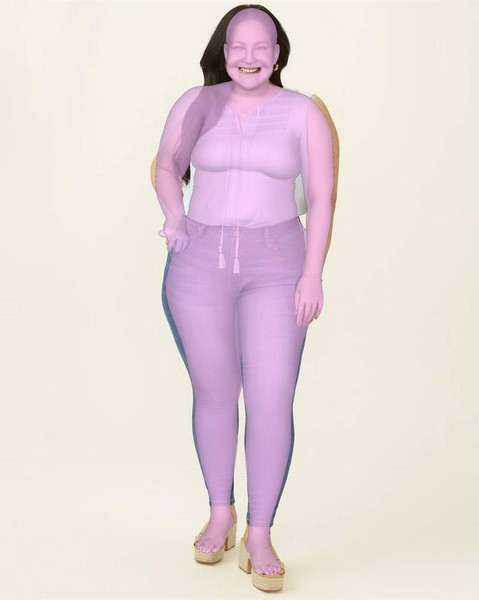}}

\subfloat{\includegraphics[height=0.24\textheight]{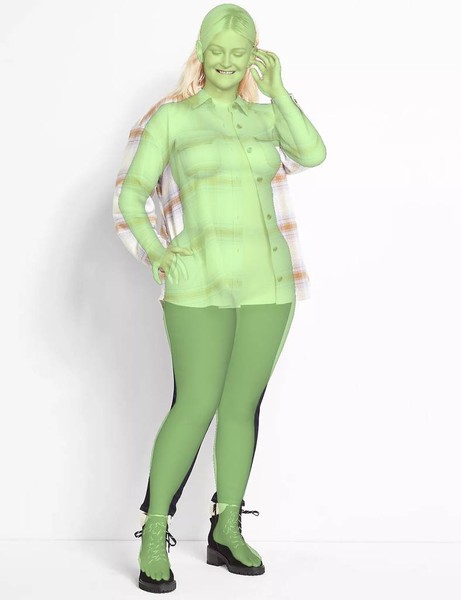}}
\subfloat{\includegraphics[height=0.24\textheight]{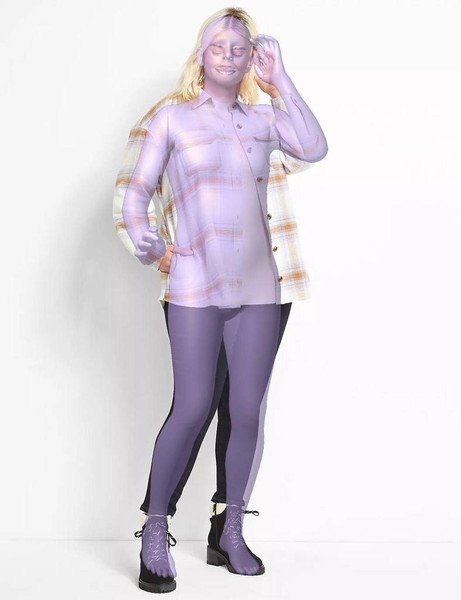}}
\subfloat{\includegraphics[height=0.24\textheight]{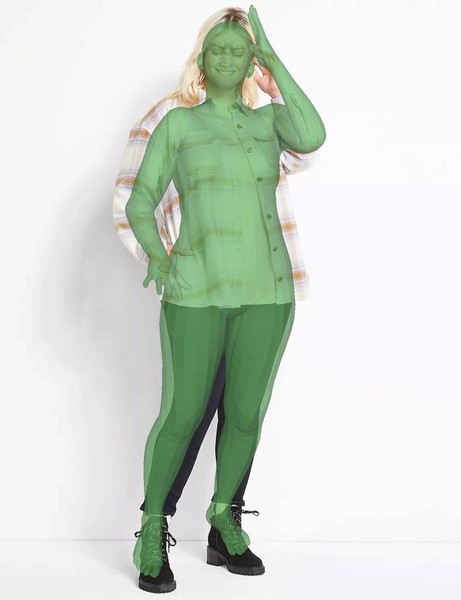}}
\subfloat{\includegraphics[height=0.24\textheight]{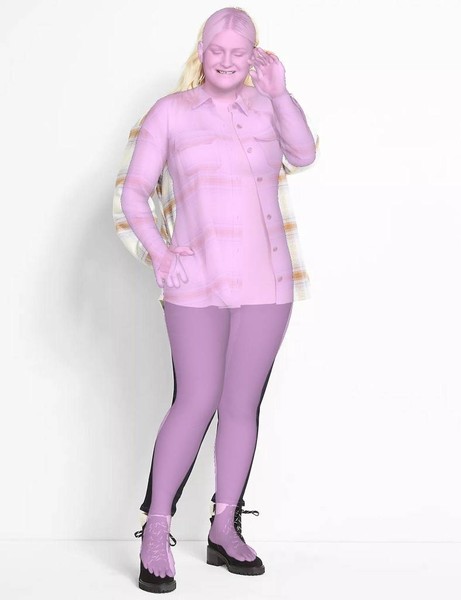}}

\subfloat{\includegraphics[height=0.24\textheight]{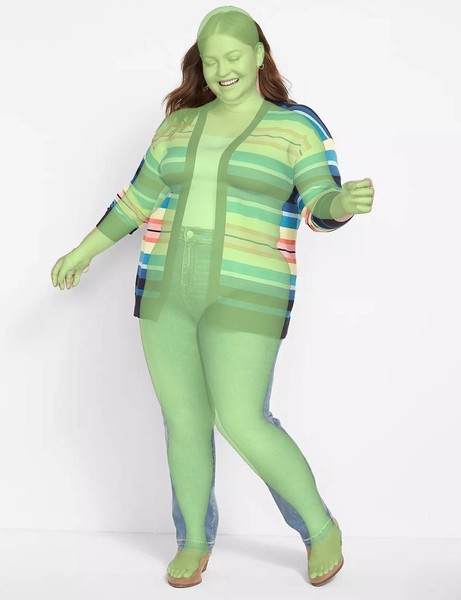}}
\subfloat{\includegraphics[height=0.24\textheight]{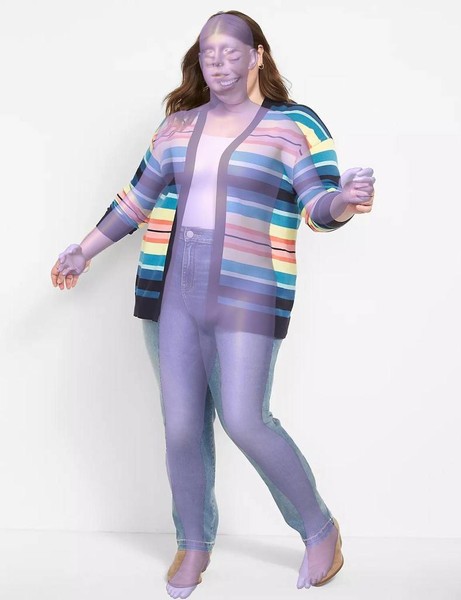}}
\subfloat{\includegraphics[height=0.24\textheight]{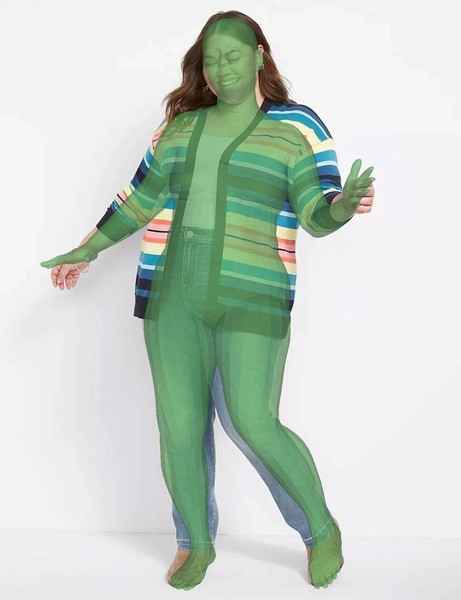}}
\subfloat{\includegraphics[height=0.24\textheight]{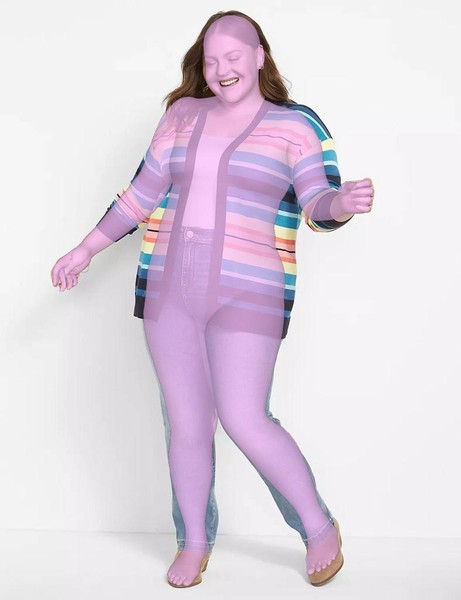}}

\subfloat[SMPLify-X \cite{pavlakos2019expressive}]
{\includegraphics[height=0.24\textheight]{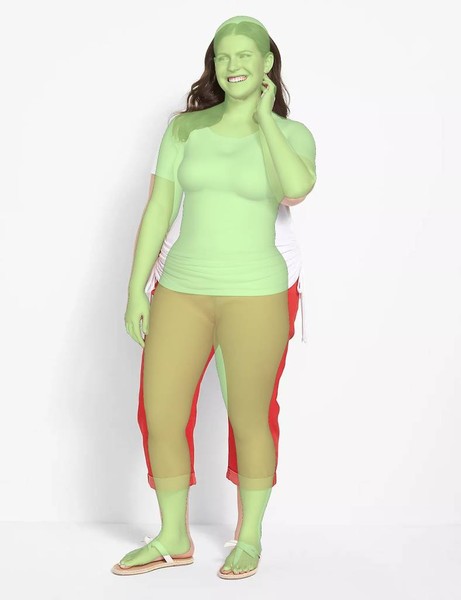}}
\subfloat[PyMAF-X \cite{pymafx2022}]{\includegraphics[height=0.24\textheight]{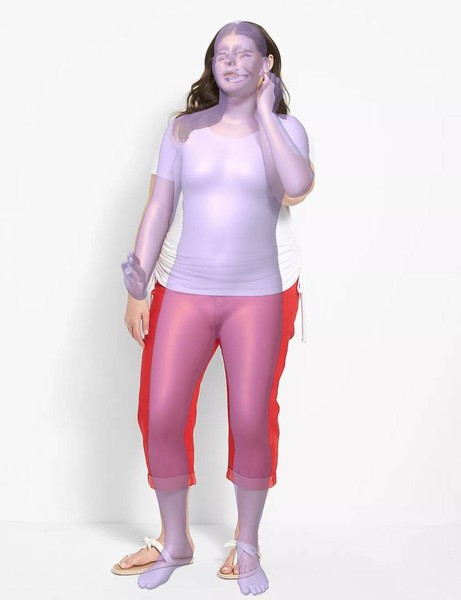}}
\subfloat[SHAPY \cite{choutas2022accurate}]
{\includegraphics[height=0.24\textheight]
{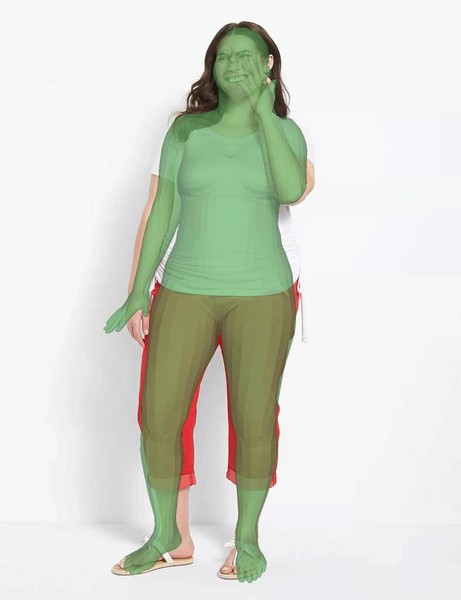}}
\subfloat[\KBody{-.1}{.035} (Ours)]{\includegraphics[height=0.24\textheight]{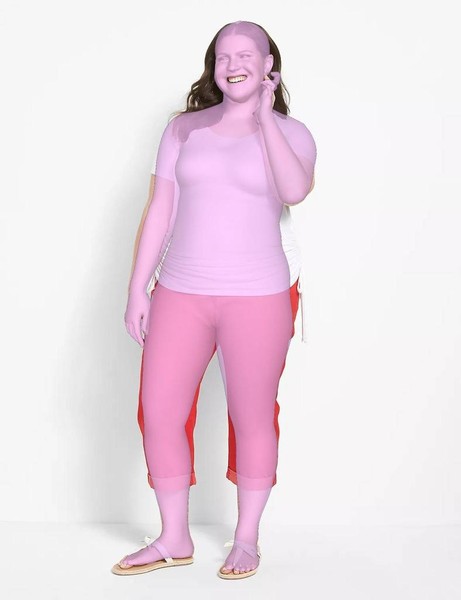}}

\caption{
Left-to-right: SMPLify-X \cite{pavlakos2019expressive} (\textcolor{caribbeangreen2}{light green}), PyMAF-X \cite{pymafx2022} (\textcolor{violet}{purple}), SHAPY \cite{choutas2022accurate} (\textcolor{jade}{green}) and KBody (\textcolor{candypink}{pink}).
}
\label{fig:w2}
\end{figure*}

%% file: figures/supp/womens_plus_size3.tex
\begin{figure*}[!htbp]
\captionsetup[subfigure]{position=bottom,labelformat=empty}

\centering

\subfloat{\includegraphics[height=0.24\textheight]{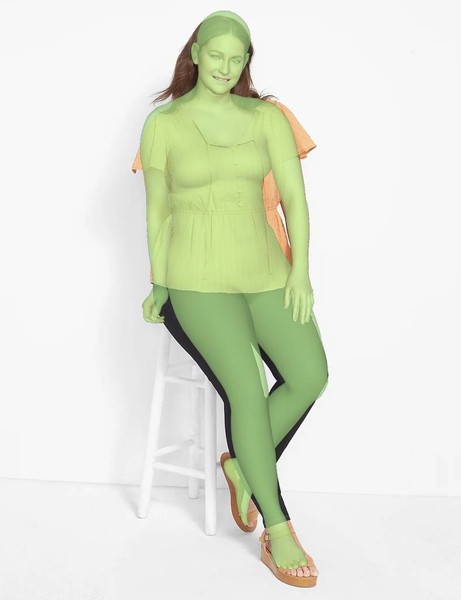}}
\subfloat{\includegraphics[height=0.24\textheight]{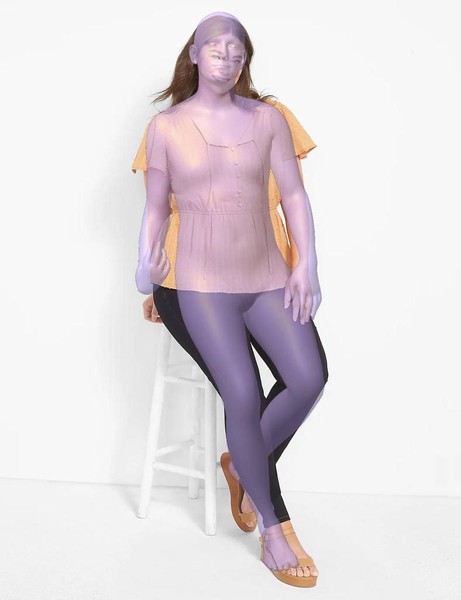}}
\subfloat{\includegraphics[height=0.24\textheight]{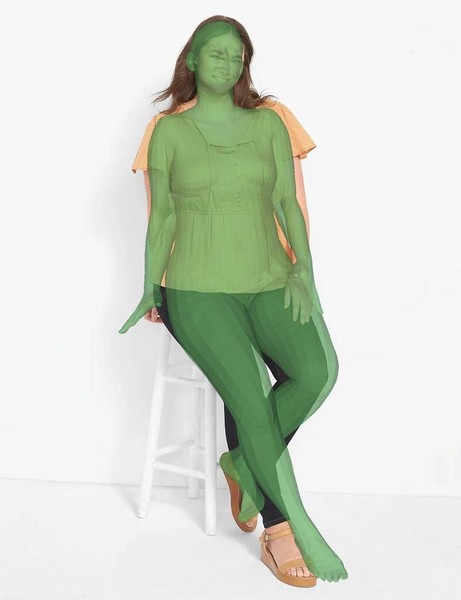}}
\subfloat{\includegraphics[height=0.24\textheight]{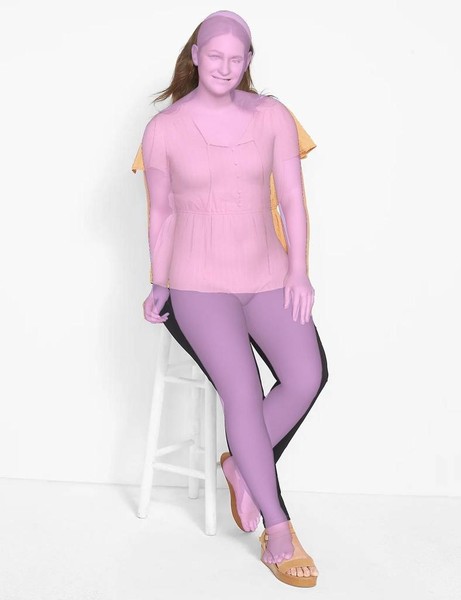}}

\subfloat{\includegraphics[height=0.24\textheight]{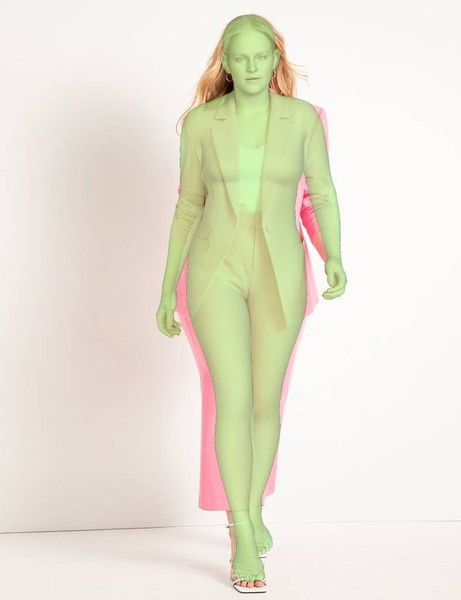}}
\subfloat{\includegraphics[height=0.24\textheight]{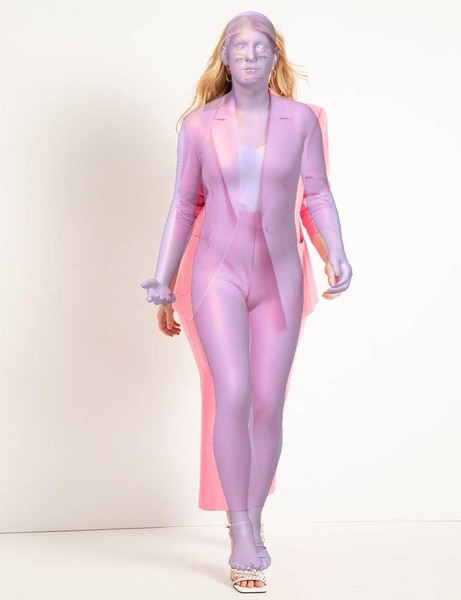}}
\subfloat{\includegraphics[height=0.24\textheight]{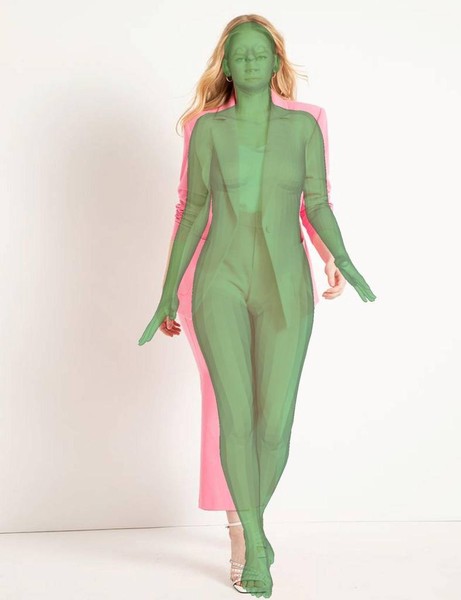}}
\subfloat{\includegraphics[height=0.24\textheight]{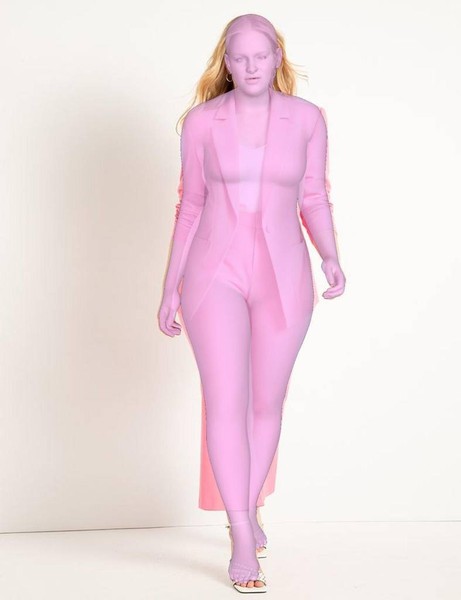}}

\subfloat{\includegraphics[height=0.24\textheight]{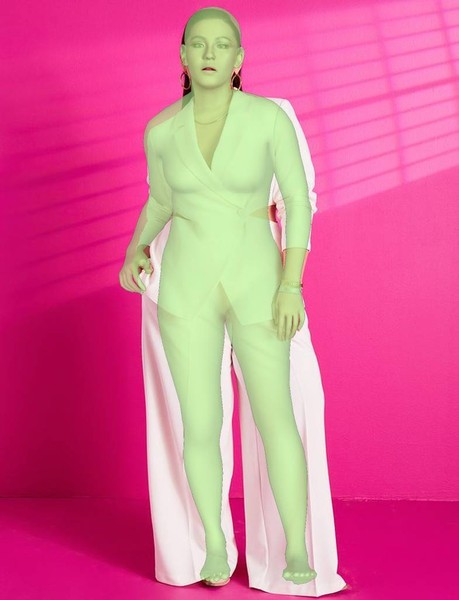}}
\subfloat{\includegraphics[height=0.24\textheight]{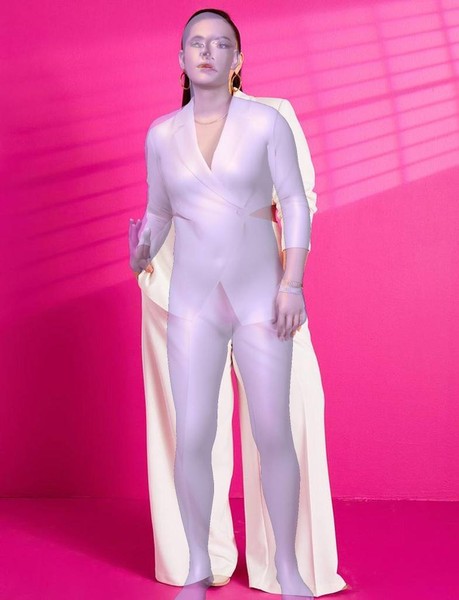}}
\subfloat{\includegraphics[height=0.24\textheight]{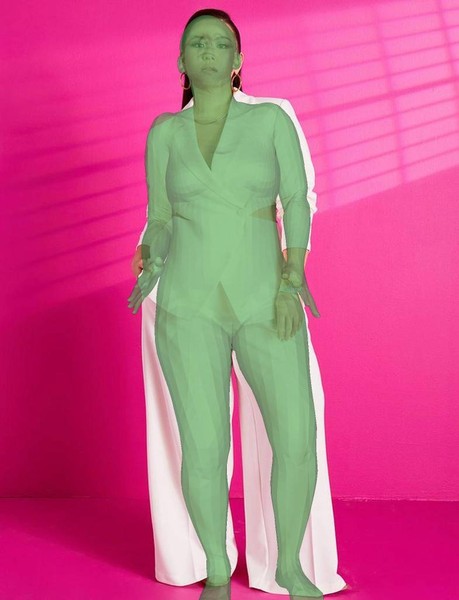}}
\subfloat{\includegraphics[height=0.24\textheight]{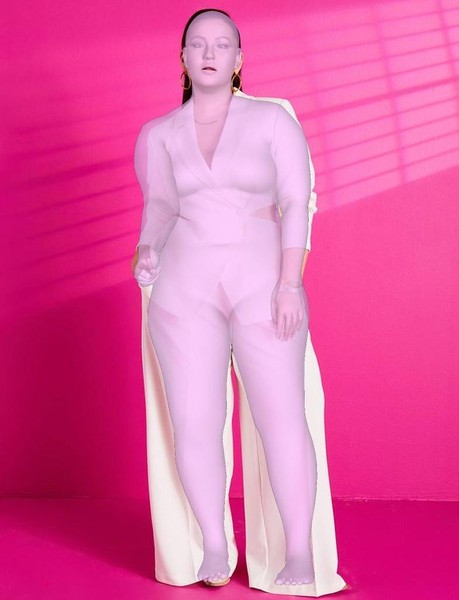}}

\subfloat[SMPLify-X \cite{pavlakos2019expressive}]
{\includegraphics[height=0.24\textheight]{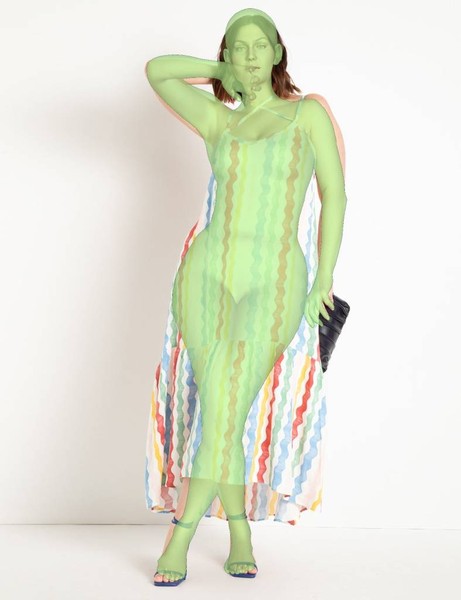}}
\subfloat[PyMAF-X \cite{pymafx2022}]{\includegraphics[height=0.24\textheight]{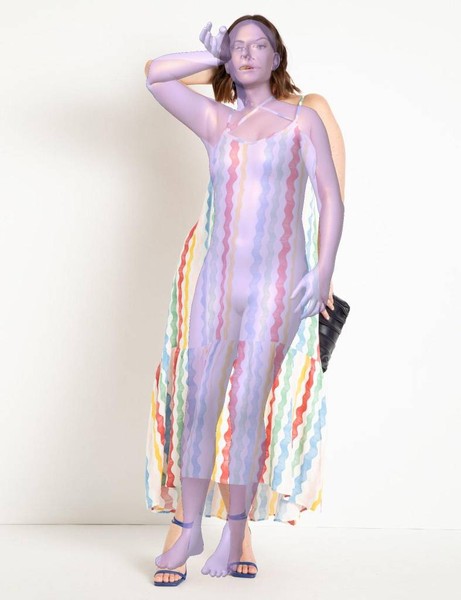}}
\subfloat[SHAPY \cite{choutas2022accurate}]
{\includegraphics[height=0.24\textheight]
{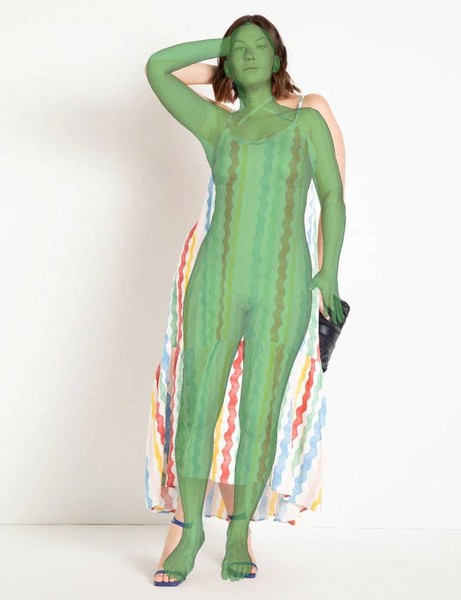}}
\subfloat[\KBody{-.1}{.035} (Ours)]{\includegraphics[height=0.24\textheight]{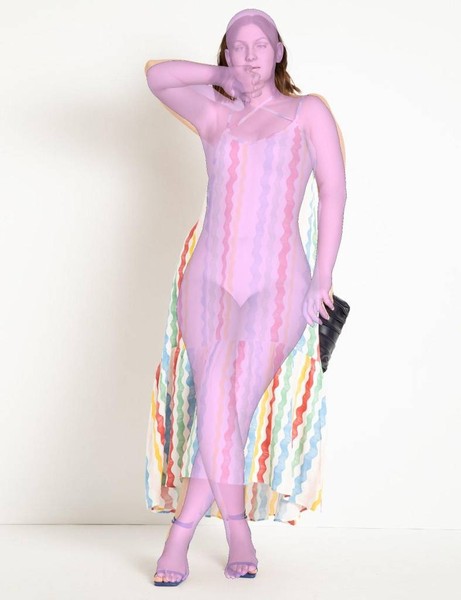}}

\caption{
Left-to-right: SMPLify-X \cite{pavlakos2019expressive} (\textcolor{caribbeangreen2}{light green}), PyMAF-X \cite{pymafx2022} (\textcolor{violet}{purple}), SHAPY \cite{choutas2022accurate} (\textcolor{jade}{green}) and KBody (\textcolor{candypink}{pink}).
}
\label{fig:w3}
\end{figure*}

%% file: figures/supp/womens_plus_size4.tex
\begin{figure*}[!htbp]
\captionsetup[subfigure]{position=bottom,labelformat=empty}

\centering

\subfloat{\includegraphics[height=0.24\textheight]{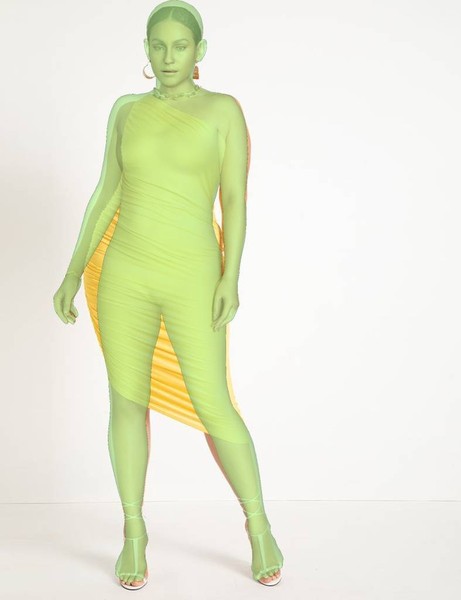}}
\subfloat{\includegraphics[height=0.24\textheight]{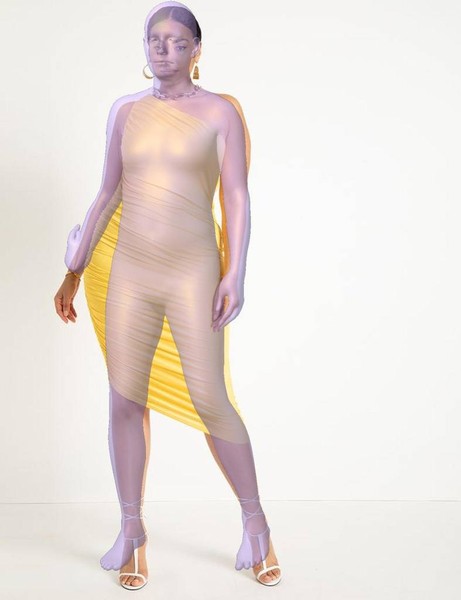}}
\subfloat{\includegraphics[height=0.24\textheight]{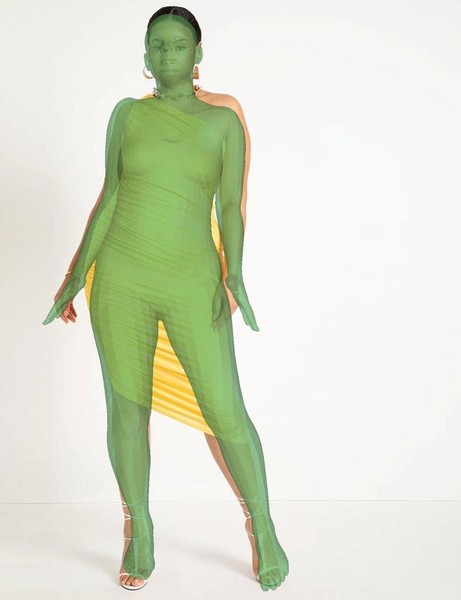}}
\subfloat{\includegraphics[height=0.24\textheight]{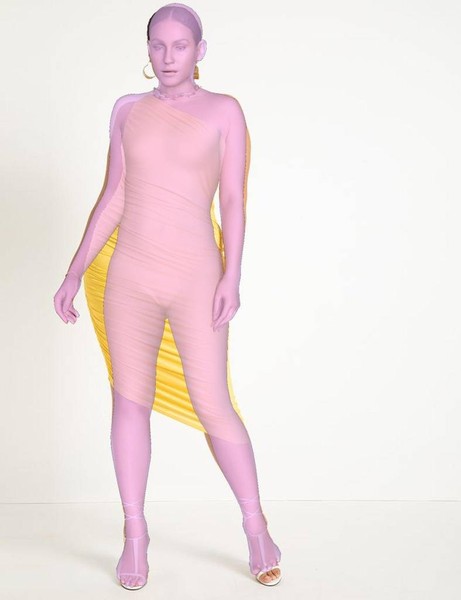}}

\subfloat{\includegraphics[height=0.24\textheight]{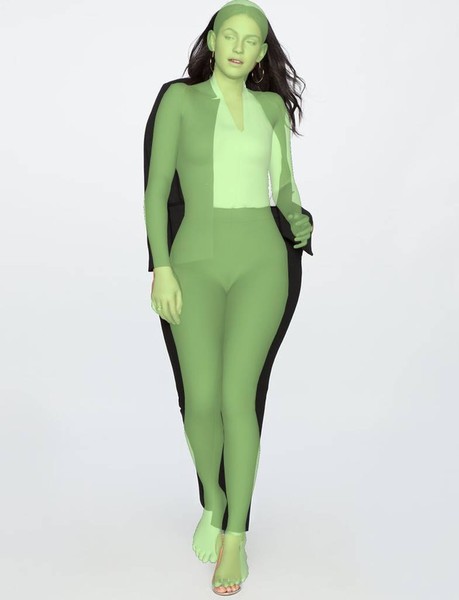}}
\subfloat{\includegraphics[height=0.24\textheight]{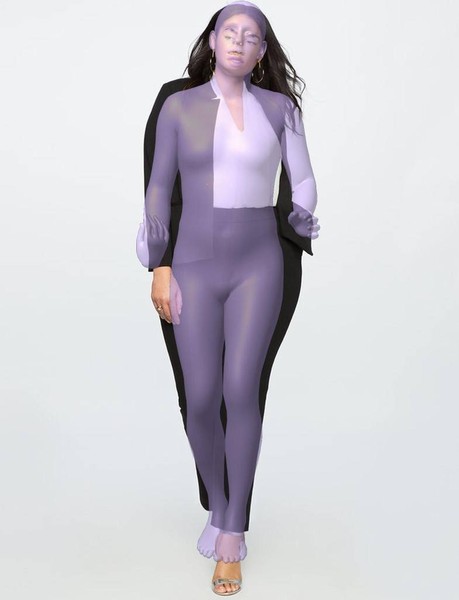}}
\subfloat{\includegraphics[height=0.24\textheight]{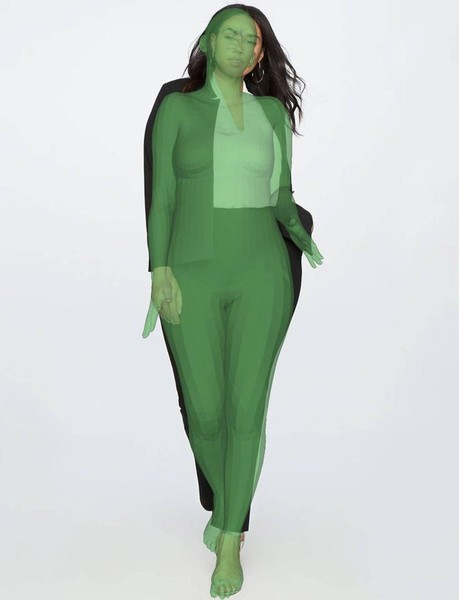}}
\subfloat{\includegraphics[height=0.24\textheight]{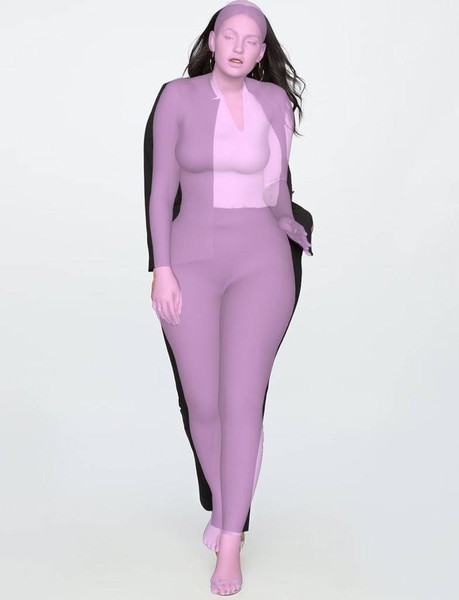}}

\subfloat{\includegraphics[height=0.24\textheight]{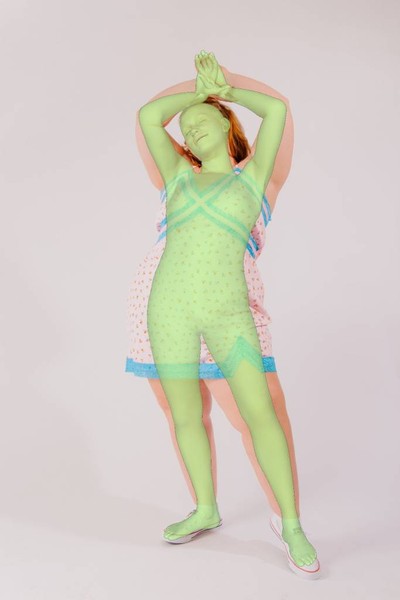}}
\subfloat{\includegraphics[height=0.24\textheight]{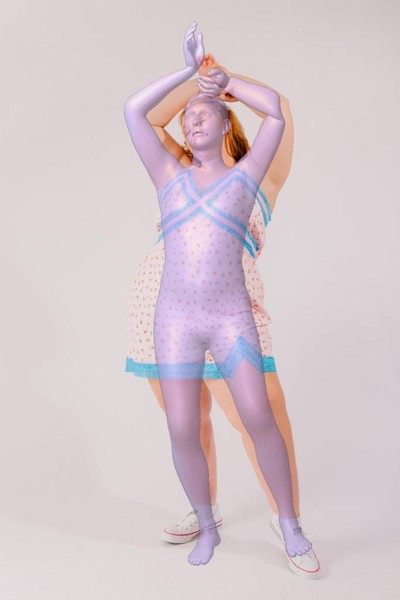}}
\subfloat{\includegraphics[height=0.24\textheight]{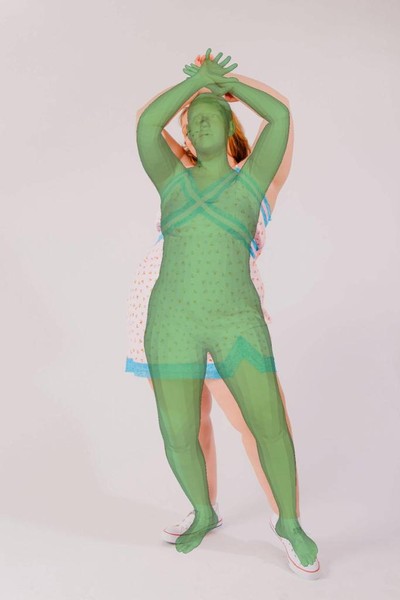}}
\subfloat{\includegraphics[height=0.24\textheight]{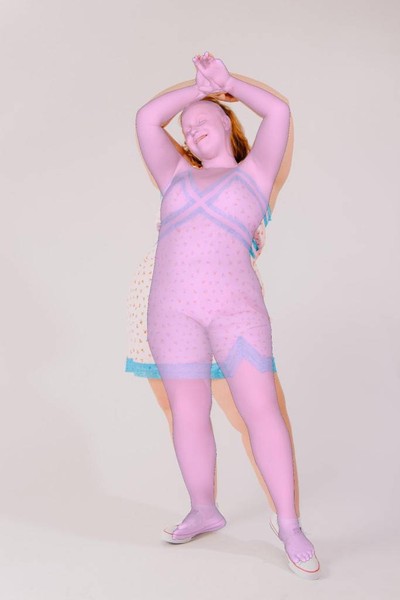}}

\subfloat[SMPLify-X \cite{pavlakos2019expressive}]
{\includegraphics[height=0.24\textheight]{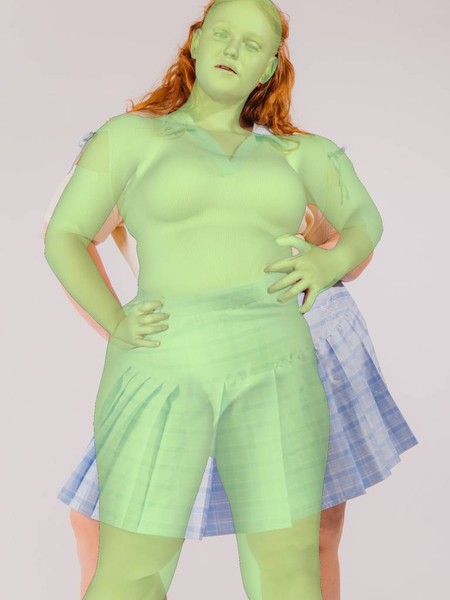}}
\subfloat[PyMAF-X \cite{pymafx2022}]{\includegraphics[height=0.24\textheight]{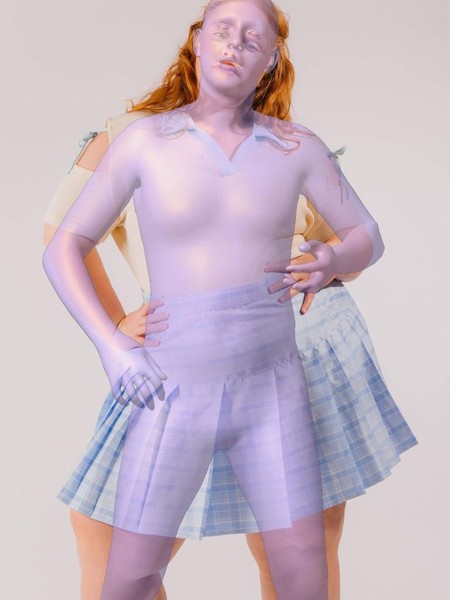}}
\subfloat[SHAPY \cite{choutas2022accurate}]
{\includegraphics[height=0.24\textheight]
{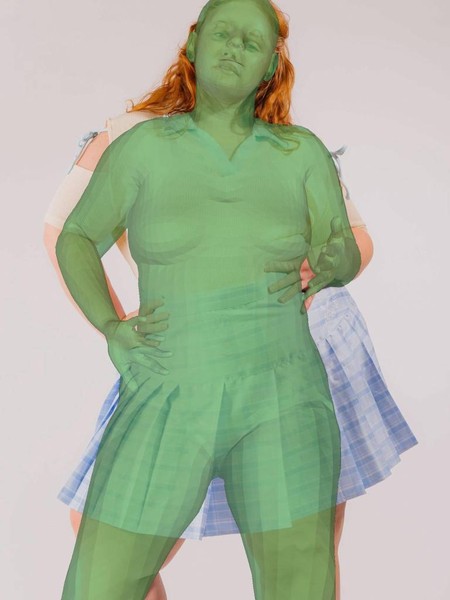}}
\subfloat[\KBody{-.1}{.035} (Ours)]{\includegraphics[height=0.24\textheight]{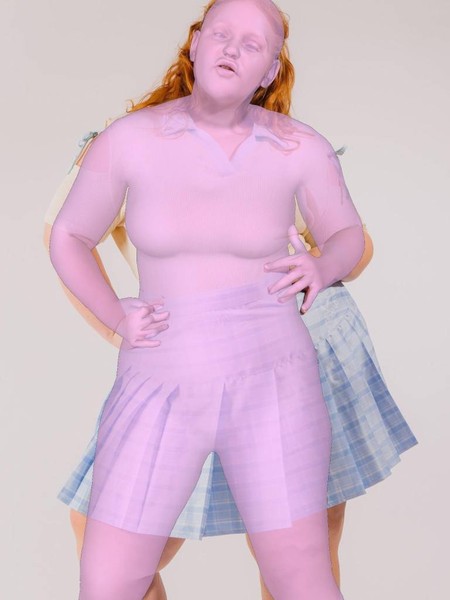}}

\caption{
Left-to-right: SMPLify-X \cite{pavlakos2019expressive} (\textcolor{caribbeangreen2}{light green}), PyMAF-X \cite{pymafx2022} (\textcolor{violet}{purple}), SHAPY \cite{choutas2022accurate} (\textcolor{jade}{green}) and KBody (\textcolor{candypink}{pink}).
}
\label{fig:w4}
\end{figure*}

%% file: figures/supp/womens_plus_size5.tex
\begin{figure*}[!htbp]
\captionsetup[subfigure]{position=bottom,labelformat=empty}

\centering

\subfloat{\includegraphics[height=0.24\textheight]{images/results/full/womens_plus_size/smplifyx_016.jpg}}
\subfloat{\includegraphics[height=0.24\textheight]{images/results/full/womens_plus_size/pymafx_016.jpg}}
\subfloat{\includegraphics[height=0.24\textheight]{images/results/full/womens_plus_size/shapy_016.jpg}}
\subfloat{\includegraphics[height=0.24\textheight]{images/results/full/womens_plus_size/kbody_016.jpg}}

\subfloat{\includegraphics[height=0.24\textheight]{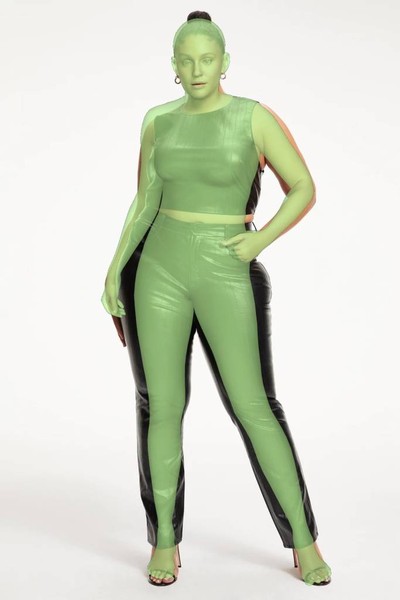}}
\subfloat{\includegraphics[height=0.24\textheight]{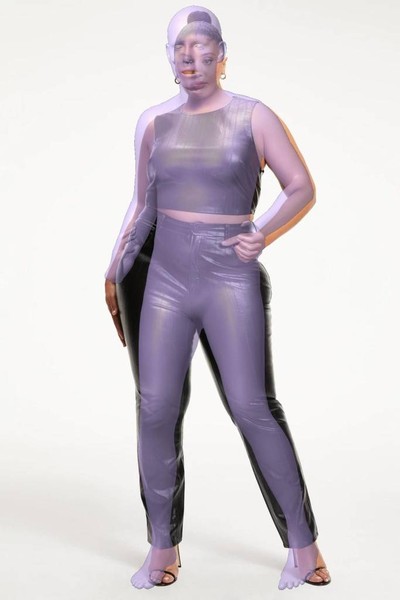}}
\subfloat{\includegraphics[height=0.24\textheight]{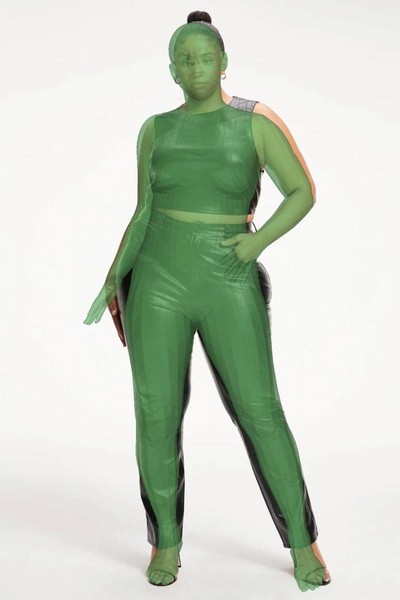}}
\subfloat{\includegraphics[height=0.24\textheight]{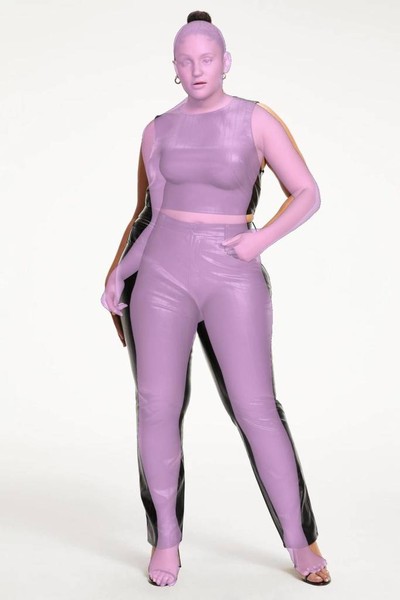}}

\subfloat{\includegraphics[height=0.24\textheight]{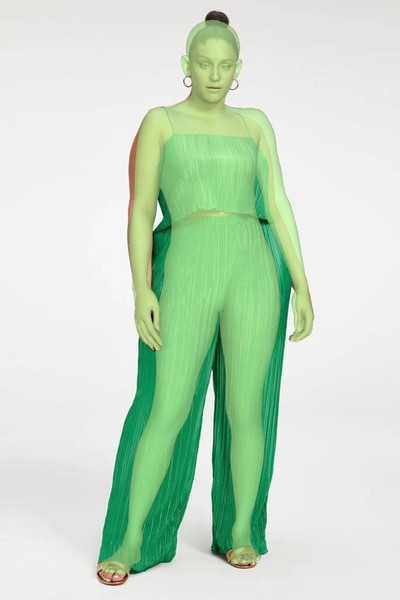}}
\subfloat{\includegraphics[height=0.24\textheight]{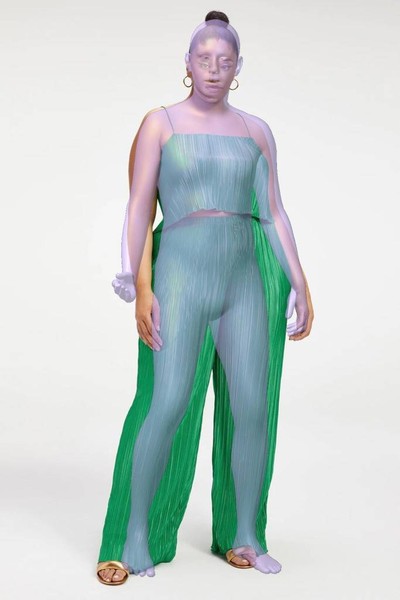}}
\subfloat{\includegraphics[height=0.24\textheight]{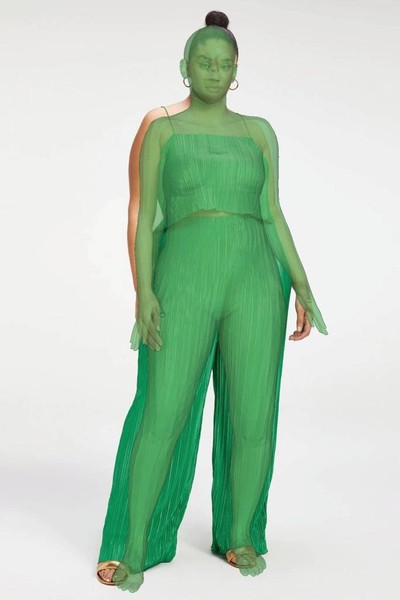}}
\subfloat{\includegraphics[height=0.24\textheight]{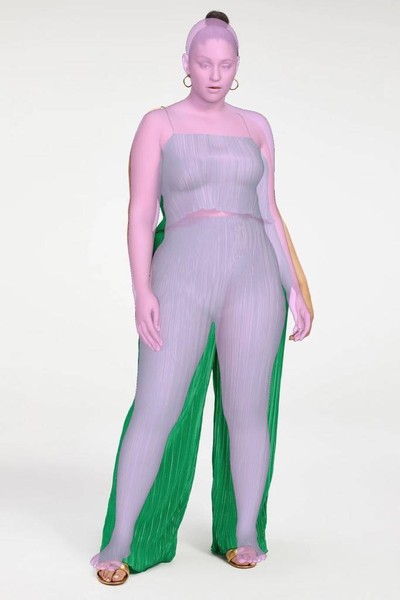}}

\subfloat[SMPLify-X \cite{pavlakos2019expressive}]
{\includegraphics[height=0.24\textheight]{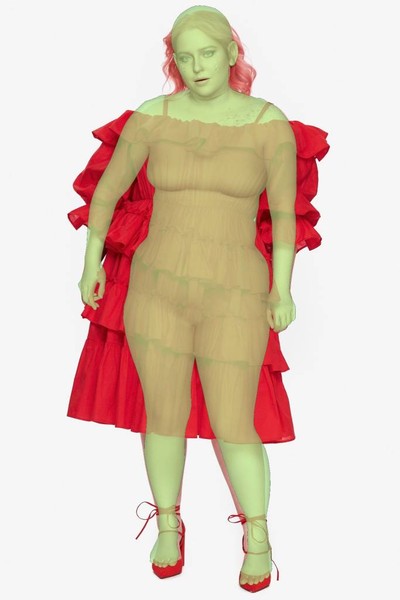}}
\subfloat[PyMAF-X \cite{pymafx2022}]{\includegraphics[height=0.24\textheight]{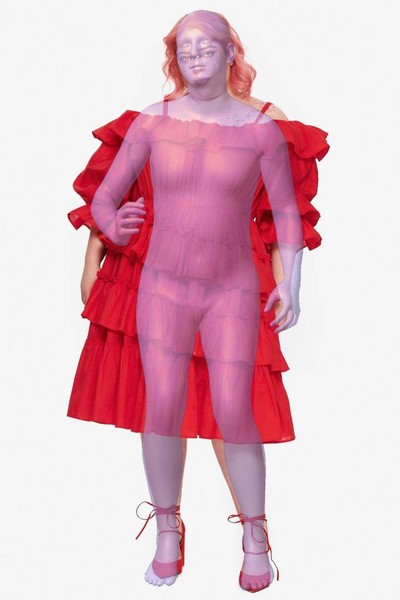}}
\subfloat[SHAPY \cite{choutas2022accurate}]
{\includegraphics[height=0.24\textheight]
{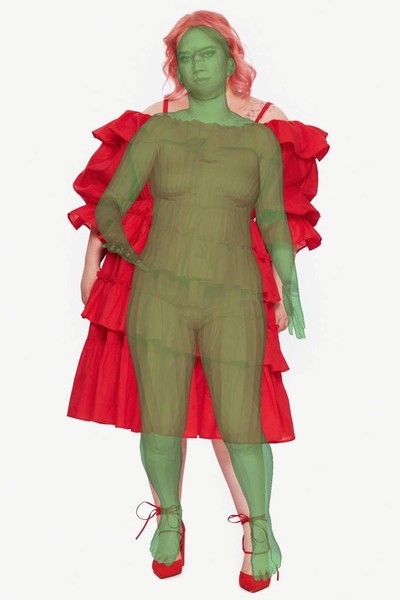}}
\subfloat[\KBody{-.1}{.035} (Ours)]{\includegraphics[height=0.24\textheight]{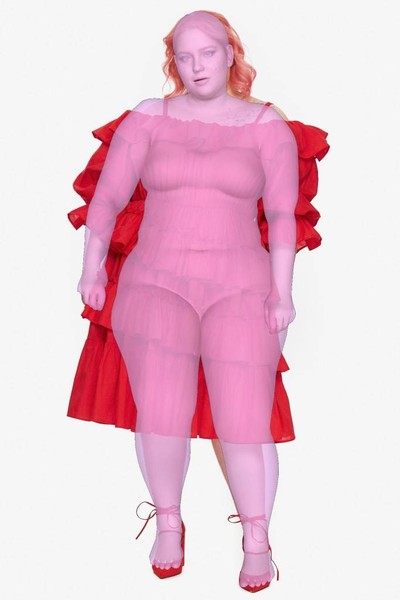}}

\caption{
Left-to-right: SMPLify-X \cite{pavlakos2019expressive} (\textcolor{caribbeangreen2}{light green}), PyMAF-X \cite{pymafx2022} (\textcolor{violet}{purple}), SHAPY \cite{choutas2022accurate} (\textcolor{jade}{green}) and KBody (\textcolor{candypink}{pink}).
}
\label{fig:w5}
\end{figure*}

%% file: figures/supp/womens_plus_size6.tex
\begin{figure*}[!htbp]
\captionsetup[subfigure]{position=bottom,labelformat=empty}

\centering

\subfloat{\includegraphics[height=0.24\textheight]{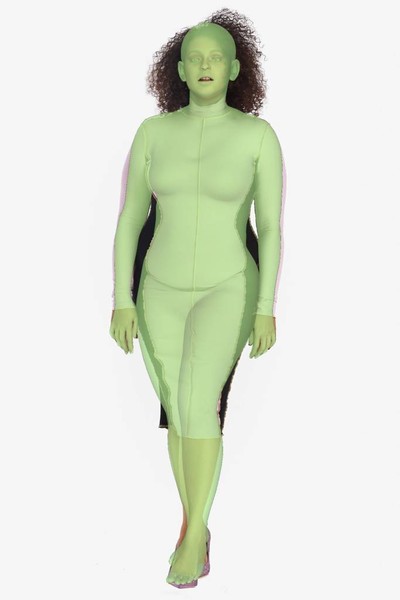}}
\subfloat{\includegraphics[height=0.24\textheight]{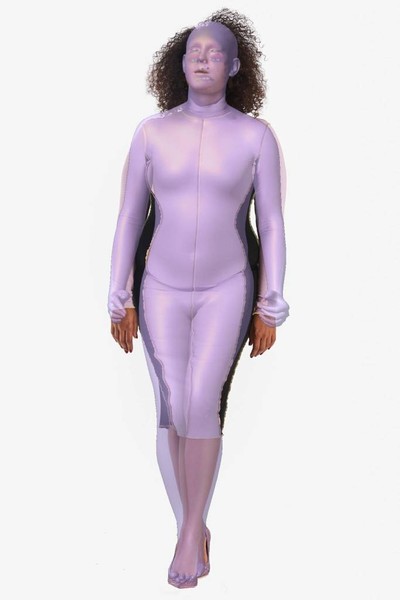}}
\subfloat{\includegraphics[height=0.24\textheight]{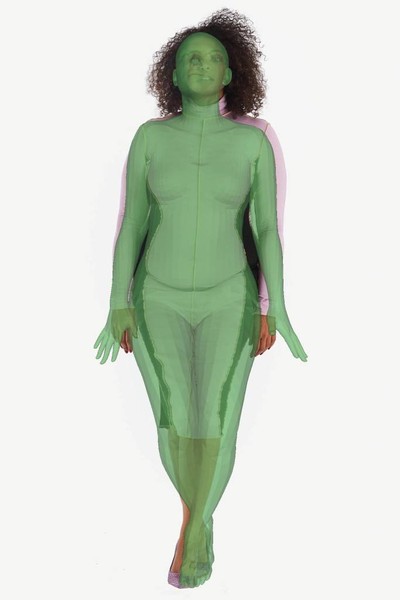}}
\subfloat{\includegraphics[height=0.24\textheight]{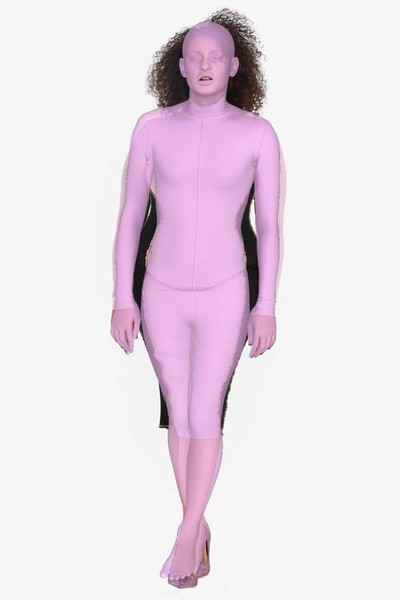}}

\subfloat{\includegraphics[height=0.24\textheight]{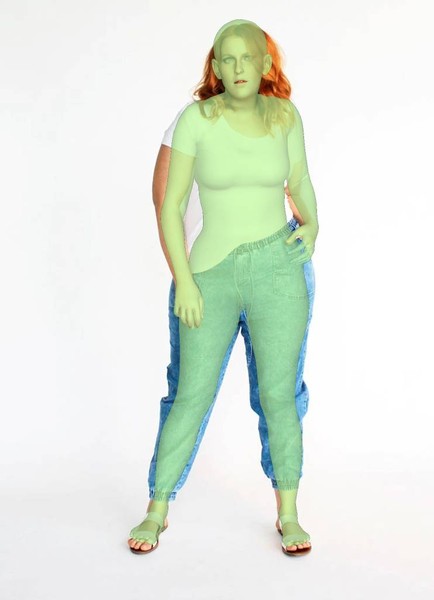}}
\subfloat{\includegraphics[height=0.24\textheight]{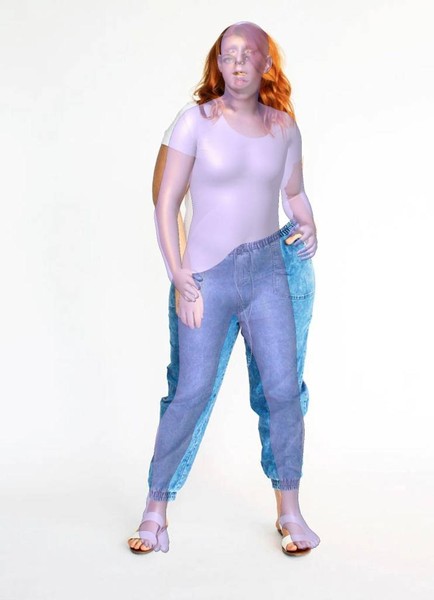}}
\subfloat{\includegraphics[height=0.24\textheight]{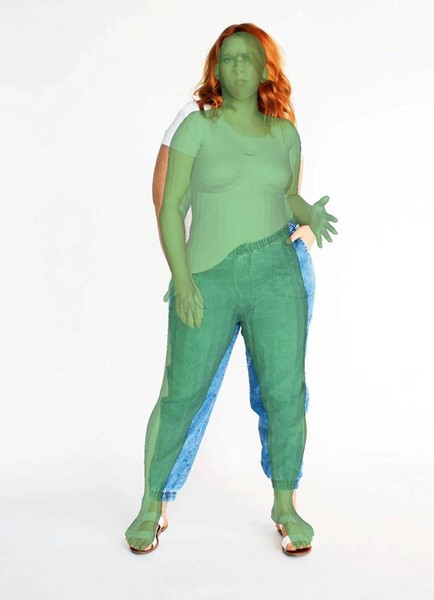}}
\subfloat{\includegraphics[height=0.24\textheight]{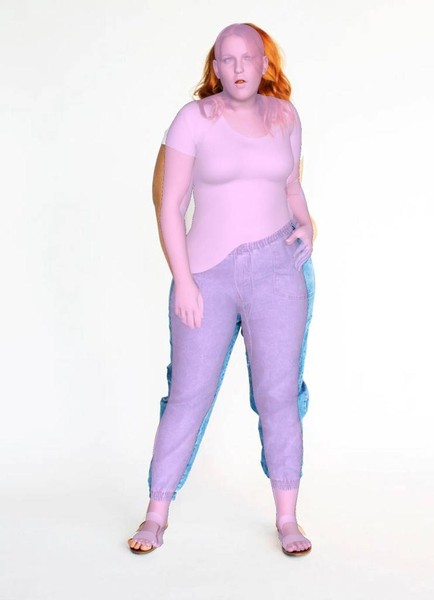}}

\subfloat{\includegraphics[height=0.24\textheight]{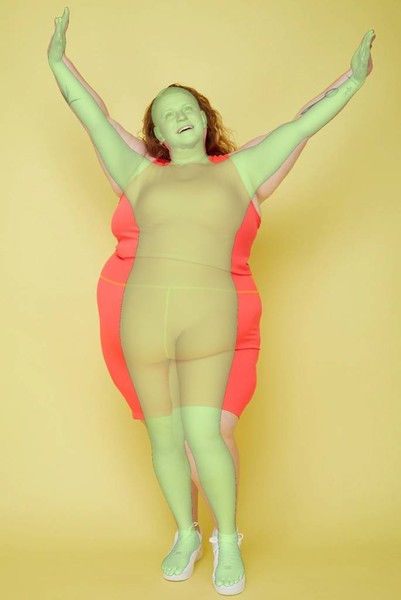}}
\subfloat{\includegraphics[height=0.24\textheight]{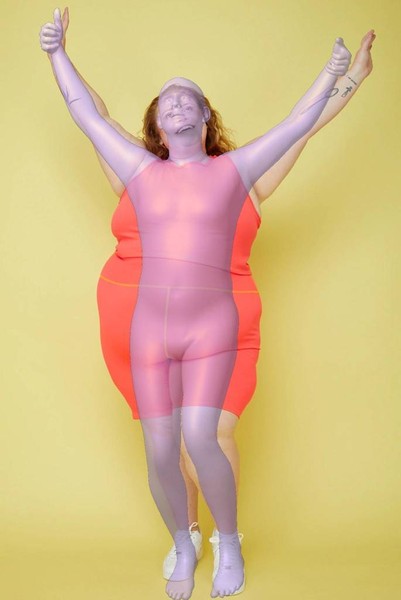}}
\subfloat{\includegraphics[height=0.24\textheight]{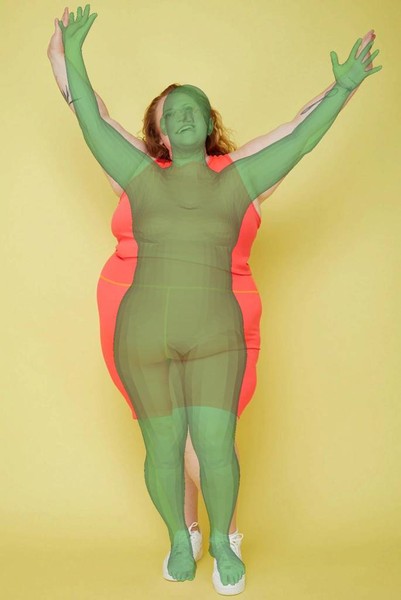}}
\subfloat{\includegraphics[height=0.24\textheight]{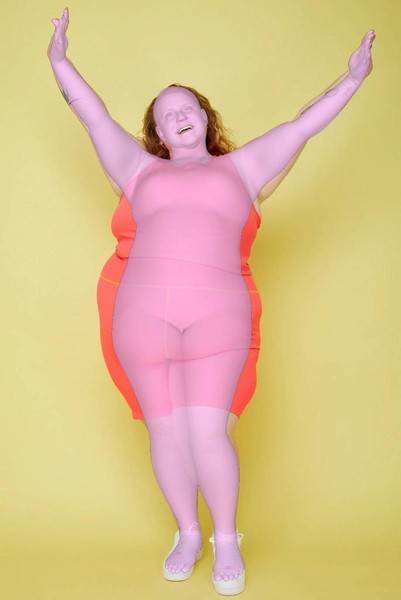}}

\caption{
Left-to-right: SMPLify-X \cite{pavlakos2019expressive} (\textcolor{caribbeangreen2}{light green}), PyMAF-X \cite{pymafx2022} (\textcolor{violet}{purple}), SHAPY \cite{choutas2022accurate} (\textcolor{jade}{green}) and KBody (\textcolor{candypink}{pink}).
}
\label{fig:w6}
\end{figure*}

%% file: figures/supp/adf_hm1.tex
\begin{figure*}[!htbp]
\captionsetup[subfigure]{position=bottom,labelformat=empty}

\centering

\raisebox{6.0\normalbaselineskip}[0pt][0pt]{\rotatebox[origin=c]{90}{\textcolor{red}{w/o} ADF}}
\subfloat[]{\includegraphics[height=0.24\textheight]
{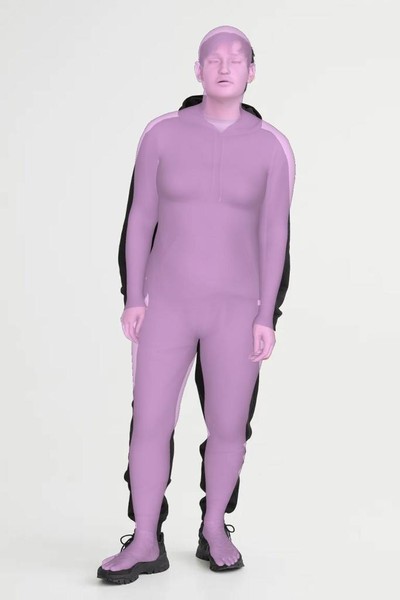}}
\subfloat[]{\includegraphics[height=0.24\textheight]
{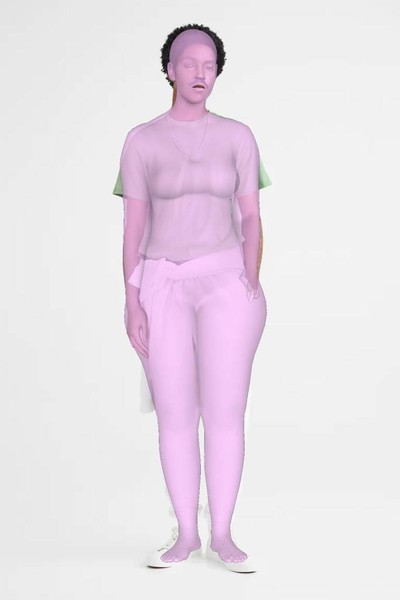}}
\subfloat[]{\includegraphics[height=0.24\textheight]
{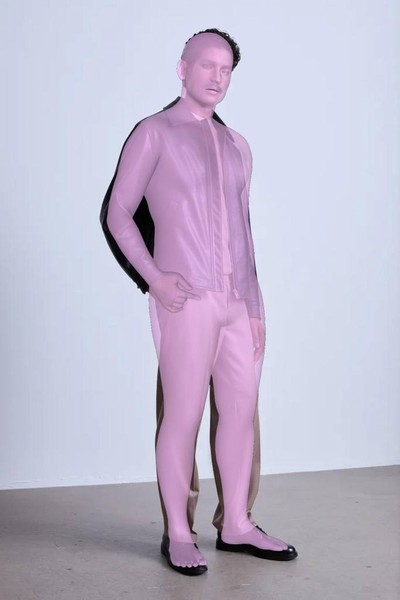}}
\subfloat[]{\includegraphics[height=0.24\textheight]
{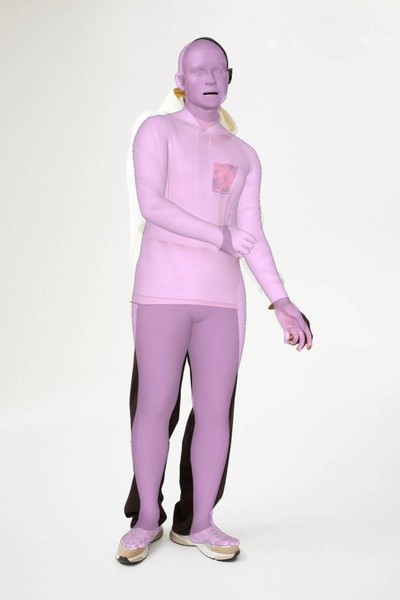}}\\

\raisebox{6.0\normalbaselineskip}[0pt][0pt]{\rotatebox[origin=c]{90}{\textcolor{green}{with} ADF}}
\subfloat[]{\includegraphics[height=0.24\textheight]{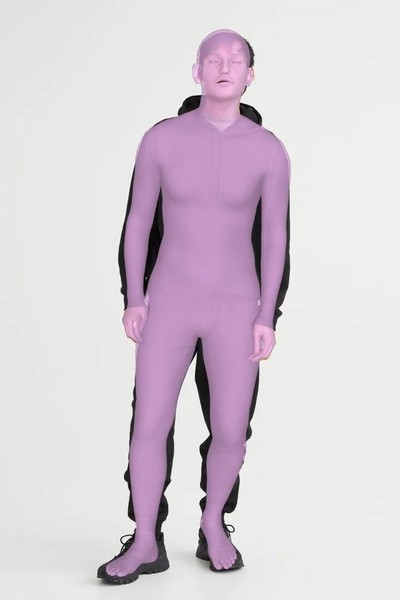}}
\subfloat[]{\includegraphics[height=0.24\textheight]{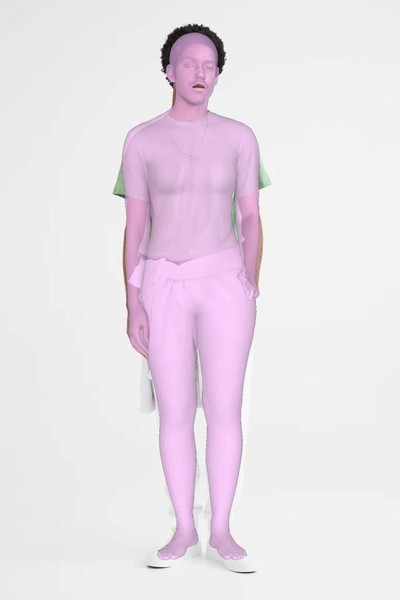}}
\subfloat[]{\includegraphics[height=0.24\textheight]{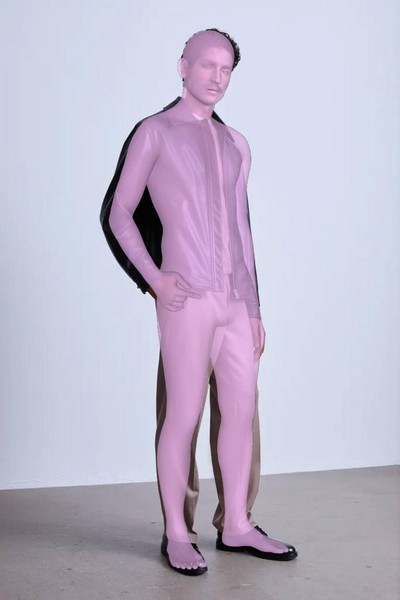}}
\subfloat[]{\includegraphics[height=0.24\textheight]{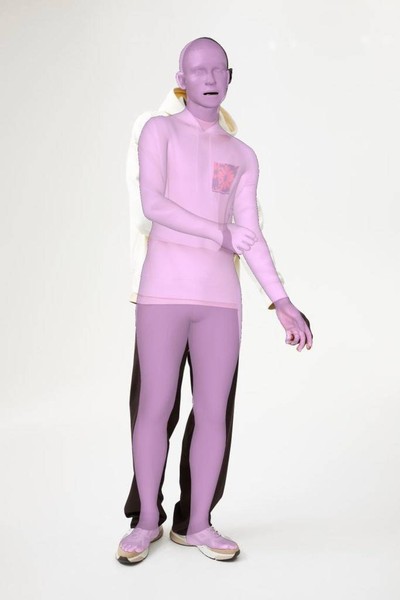}}

\raisebox{6.0\normalbaselineskip}[0pt][0pt]{\rotatebox[origin=c]{90}{\textcolor{red}{w/o} ADF}}
\subfloat[]{\includegraphics[height=0.24\textheight]
{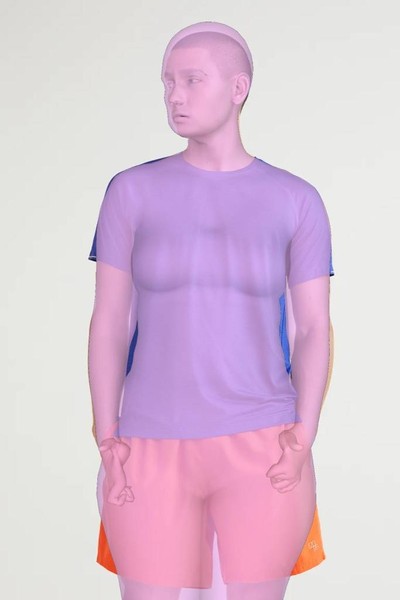}}
\subfloat[]{\includegraphics[height=0.24\textheight]
{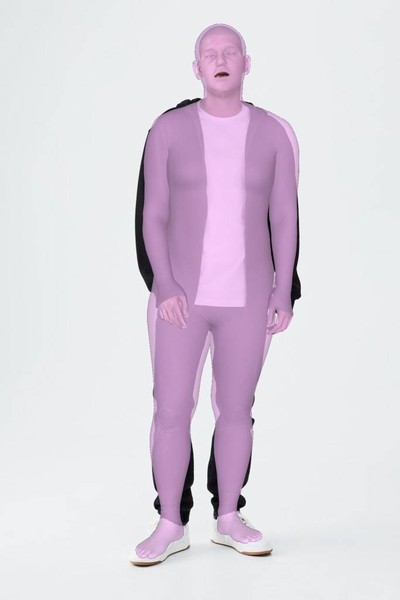}}
\subfloat[]{\includegraphics[height=0.24\textheight]
{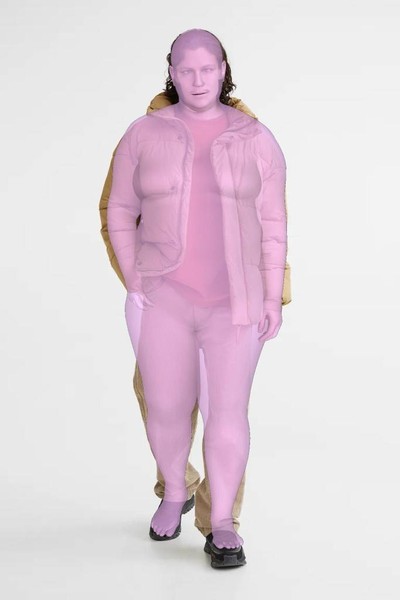}}
\subfloat[]{\includegraphics[height=0.24\textheight]
{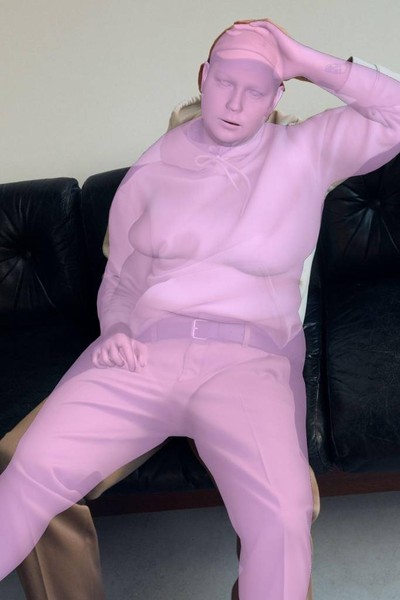}}\\

\raisebox{6.0\normalbaselineskip}[0pt][0pt]{\rotatebox[origin=c]{90}{\textcolor{green}{with} ADF}}
\subfloat[]{\includegraphics[height=0.24\textheight]{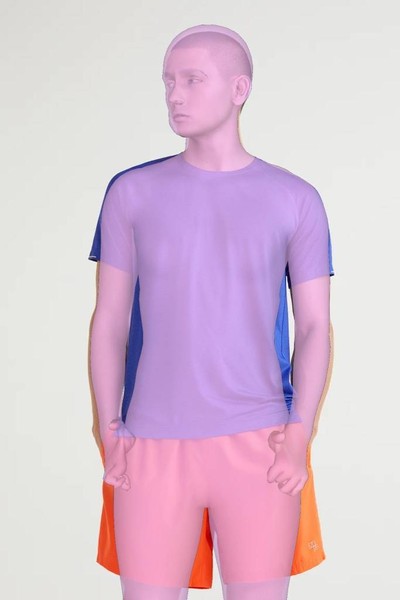}}
\subfloat[]{\includegraphics[height=0.24\textheight]{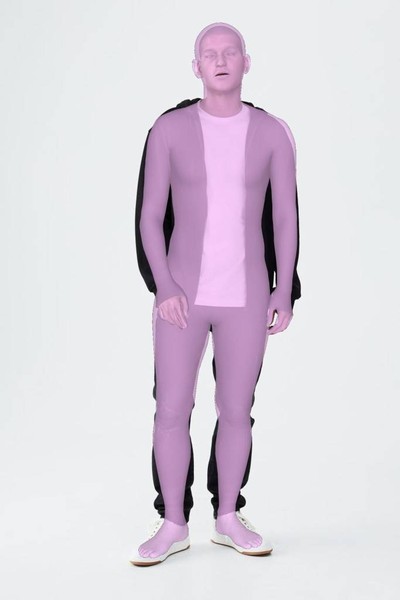}}
\subfloat[]{\includegraphics[height=0.24\textheight]{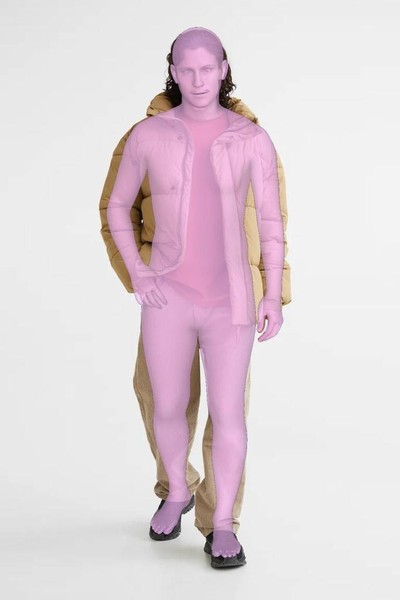}}
\subfloat[]{\includegraphics[height=0.24\textheight]{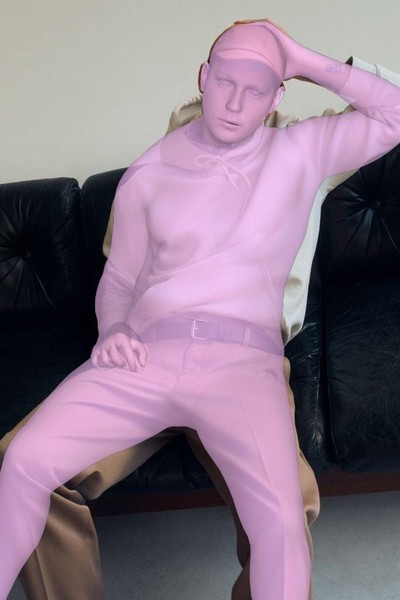}}

\caption{
\KBody{}{} fitting results \textcolor{red}{without} ADF on each even row and \textcolor{green}{with} ADF on each odd row.
}
\label{fig:adf_hm1}
\end{figure*}

%% file: figures/supp/adf_hm2.tex
\begin{figure*}[!htbp]
\captionsetup[subfigure]{position=bottom,labelformat=empty}

\centering

\raisebox{6.0\normalbaselineskip}[0pt][0pt]{\rotatebox[origin=c]{90}{\textcolor{red}{w/o} ADF}}
\subfloat[]{\includegraphics[height=0.24\textheight]{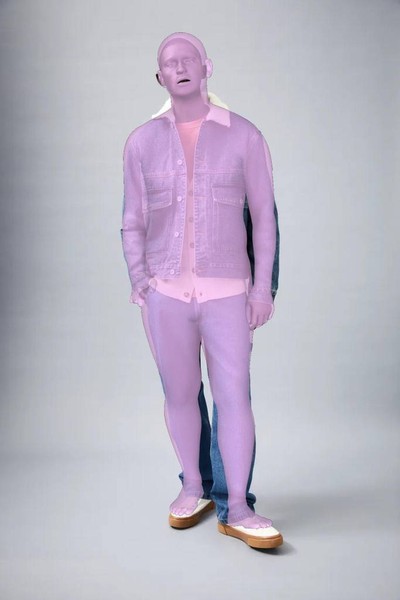}}
\subfloat[]{\includegraphics[height=0.24\textheight]{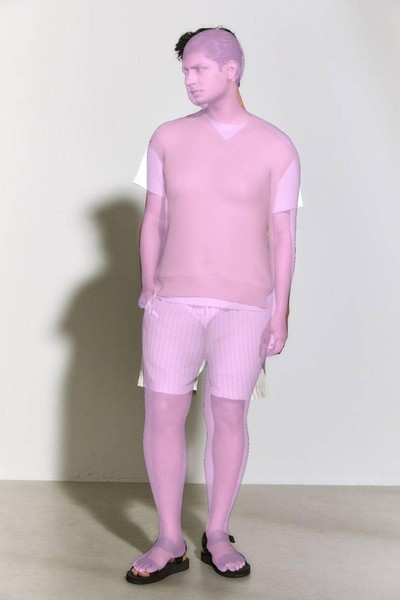}}
\subfloat[]{\includegraphics[height=0.24\textheight]{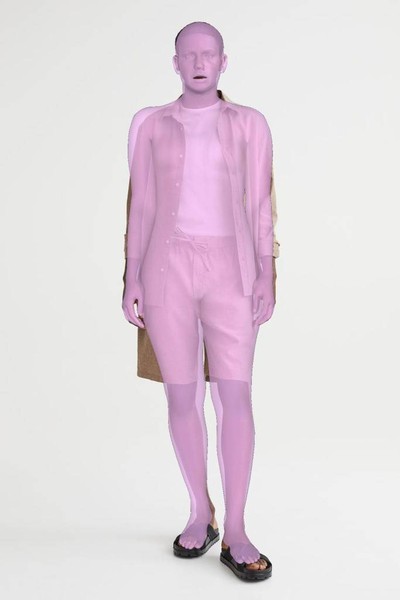}}
\subfloat[]{\includegraphics[height=0.24\textheight]{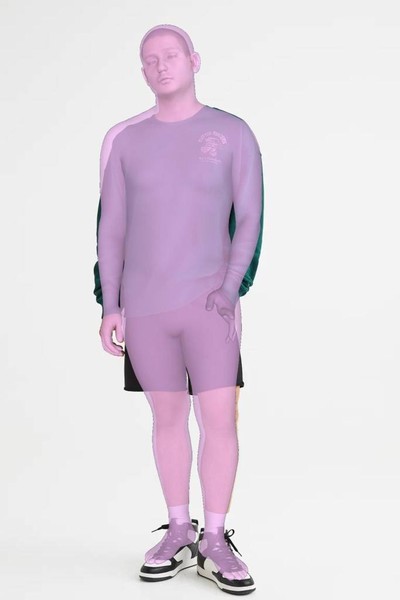}}\\

\raisebox{6.0\normalbaselineskip}[0pt][0pt]{\rotatebox[origin=c]{90}{\textcolor{green}{with} ADF}}
\subfloat[]{\includegraphics[height=0.24\textheight]{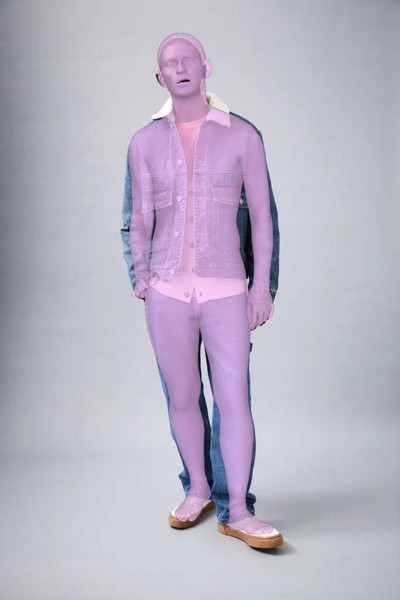}}
\subfloat[]{\includegraphics[height=0.24\textheight]{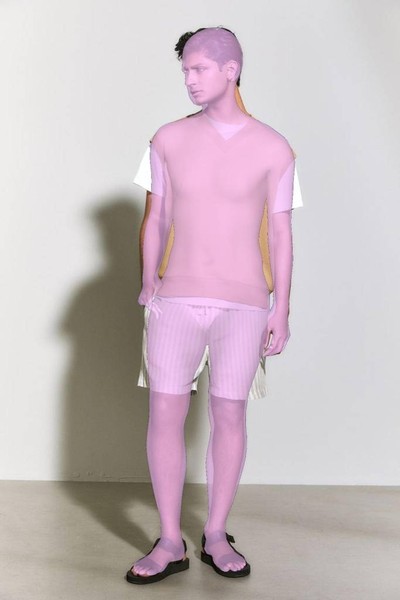}}
\subfloat[]{\includegraphics[height=0.24\textheight]{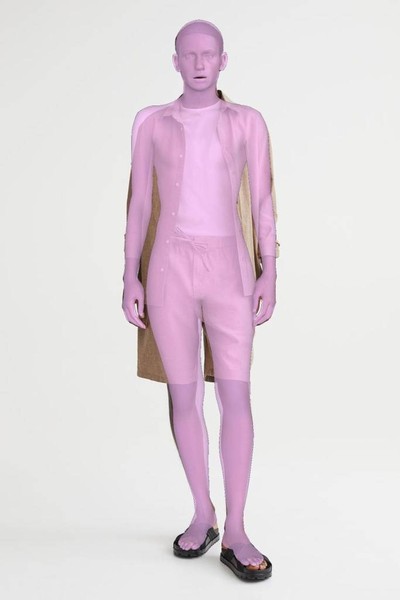}}
\subfloat[]{\includegraphics[height=0.24\textheight]{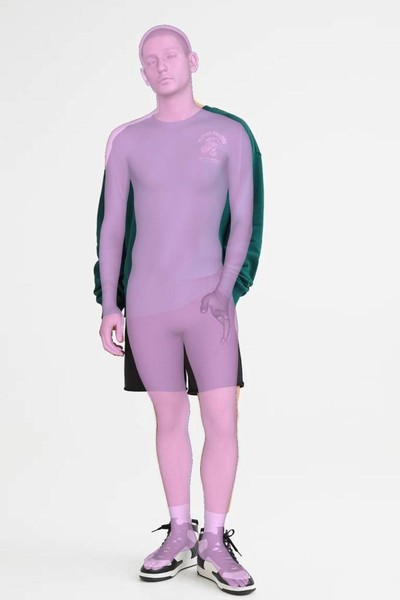}}

\raisebox{6.0\normalbaselineskip}[0pt][0pt]{\rotatebox[origin=c]{90}{\textcolor{red}{w/o} ADF}}
\subfloat[]{\includegraphics[height=0.24\textheight]{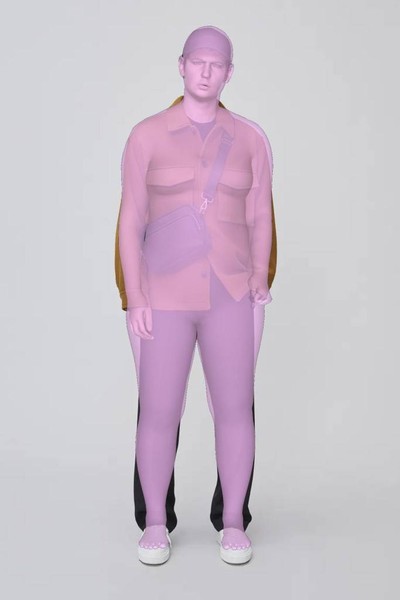}}
\subfloat[]{\includegraphics[height=0.24\textheight]{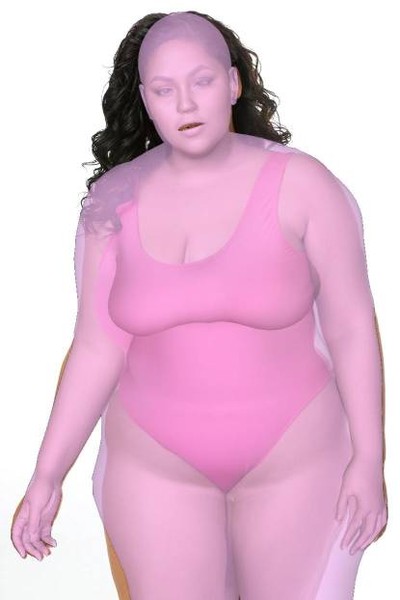}}
\subfloat[]{\includegraphics[height=0.24\textheight]
{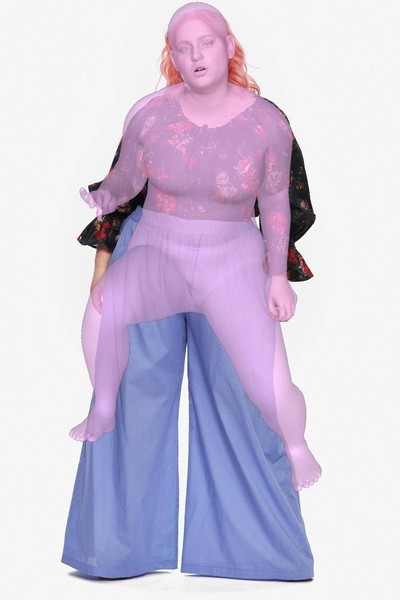}}
\subfloat[]{\includegraphics[height=0.24\textheight]
{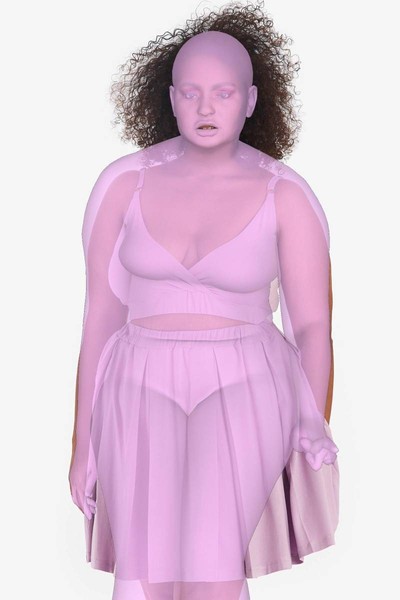}}\\

\raisebox{6.0\normalbaselineskip}[0pt][0pt]{\rotatebox[origin=c]{90}{\textcolor{green}{with} ADF}}
\subfloat[]{\includegraphics[height=0.24\textheight]{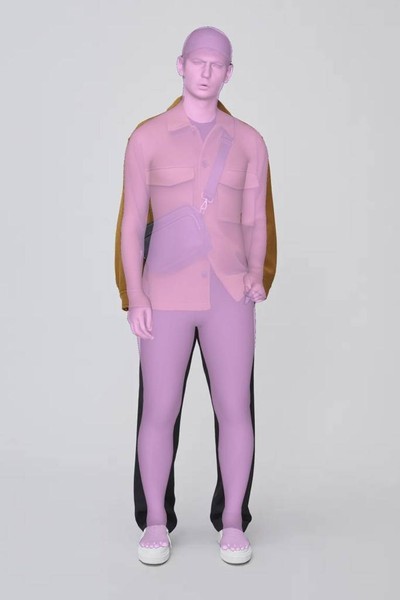}}
\raisebox{6.0\normalbaselineskip}[0pt][0pt]{\rotatebox[origin=c]{90}{\textcolor{green}{with} ADF}}
\subfloat[]{\includegraphics[height=0.24\textheight]{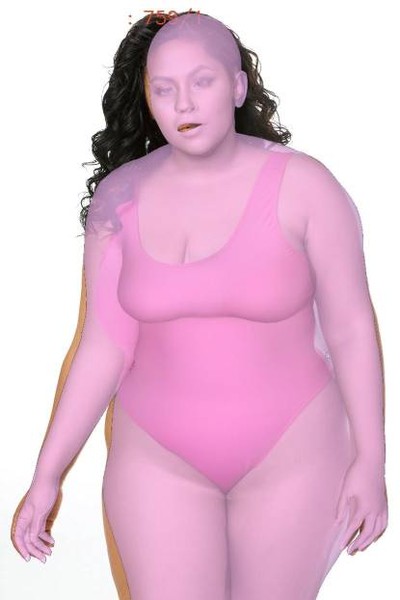}}
\subfloat[]{\includegraphics[height=0.24\textheight]{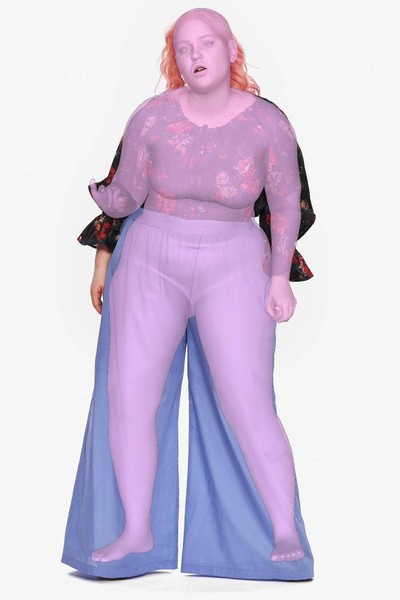}}
\subfloat[]{\includegraphics[height=0.24\textheight]{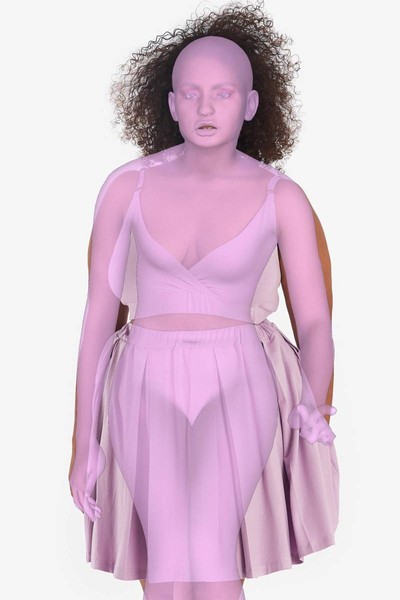}}

\caption{
\KBody{}{} fitting results \textcolor{red}{without} ADF on each even row and \textcolor{green}{with} ADF on each odd row.
}
\label{fig:adf_hm2}
\end{figure*}

%% file: figures/supp/adf_hm_mens_big.tex
\begin{figure*}[!htbp]
\captionsetup[subfigure]{position=bottom,labelformat=empty}

\centering

\raisebox{6.0\normalbaselineskip}[0pt][0pt]{\rotatebox[origin=c]{90}{\textcolor{red}{w/o} ADF}}
\subfloat[]{\includegraphics[height=0.24\textheight]{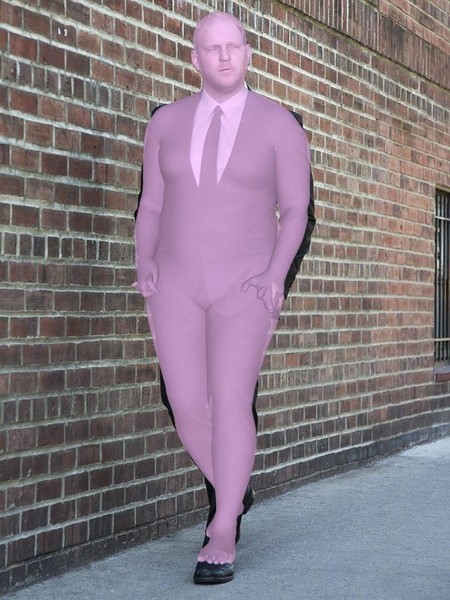}}
\subfloat[]{\includegraphics[height=0.24\textheight]{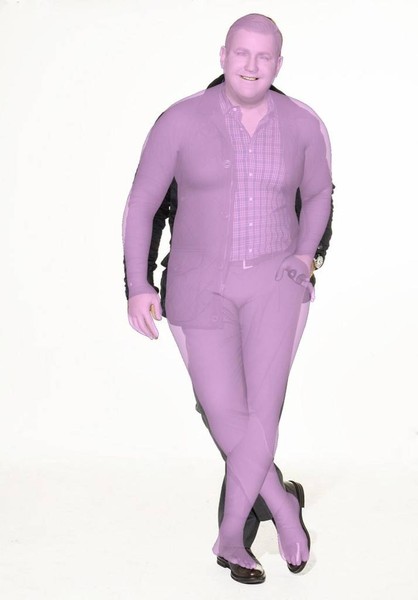}}
\subfloat[]{\includegraphics[height=0.24\textheight]{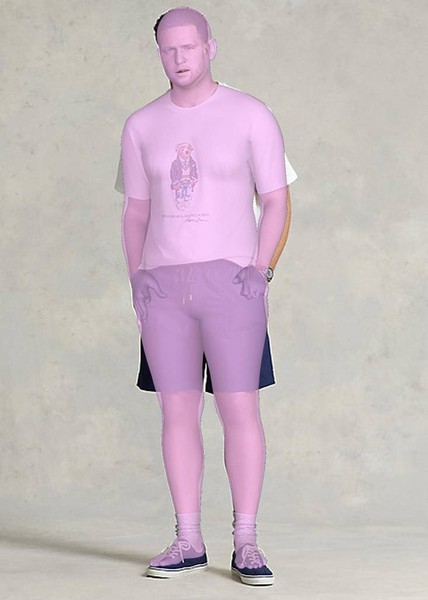}}
\subfloat[]{\includegraphics[height=0.24\textheight]{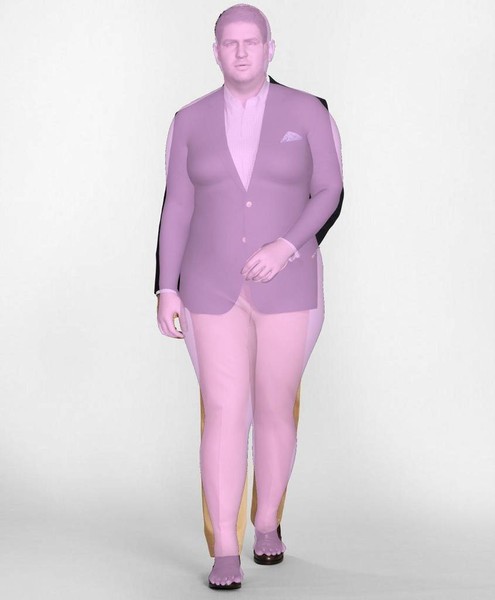}}\\

\raisebox{6.0\normalbaselineskip}[0pt][0pt]{\rotatebox[origin=c]{90}{\textcolor{green}{with} ADF}}
\subfloat[]{\includegraphics[height=0.24\textheight]{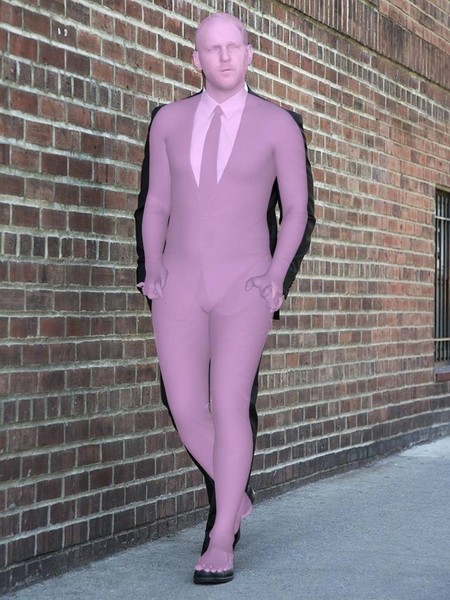}}
\subfloat[]{\includegraphics[height=0.24\textheight]{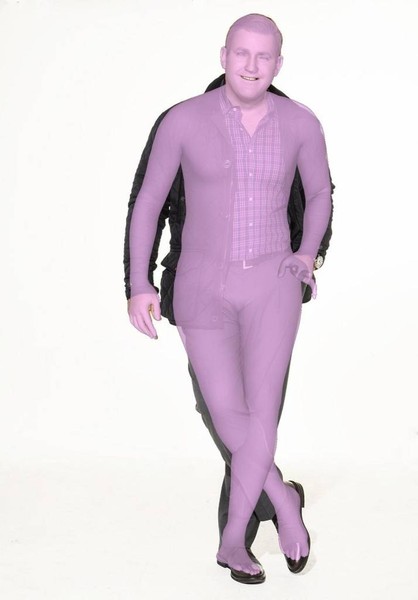}}
\subfloat[]{\includegraphics[height=0.24\textheight]{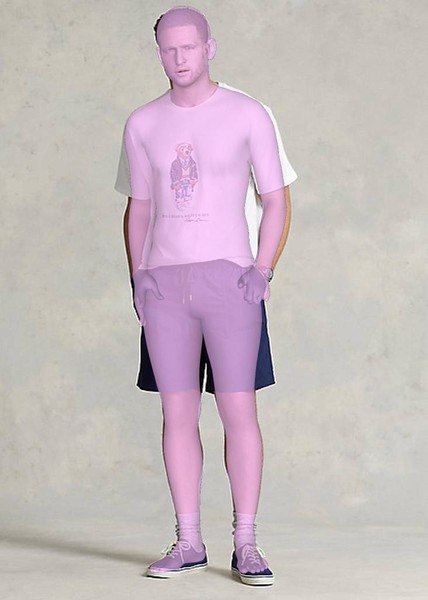}}
\subfloat[]{\includegraphics[height=0.24\textheight]{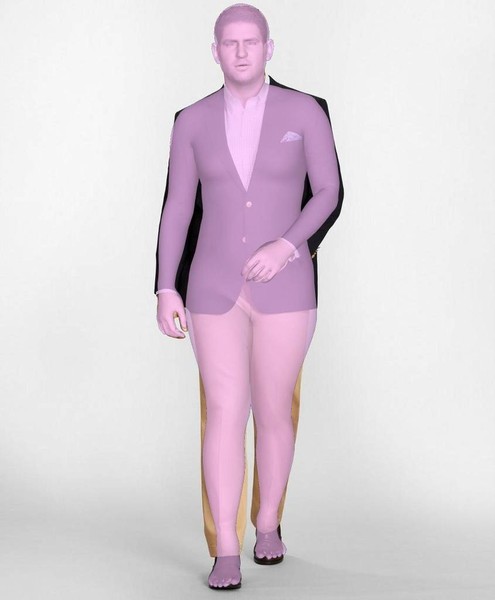}}\\

\raisebox{6.0\normalbaselineskip}[0pt][0pt]{\rotatebox[origin=c]{90}{\textcolor{red}{w/o} ADF}}
\subfloat[]{\includegraphics[height=0.24\textheight]{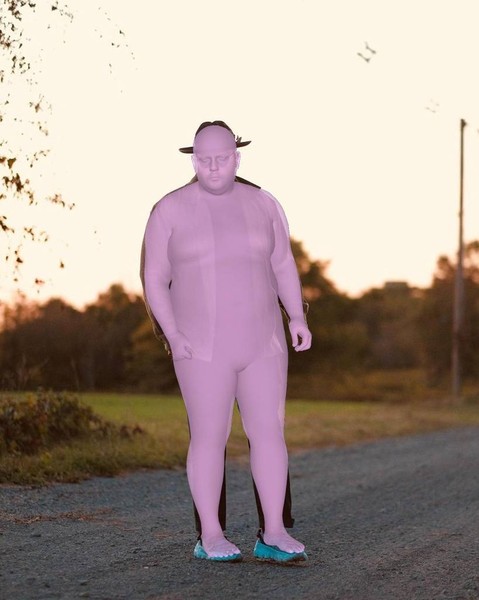}}
\subfloat[]{\includegraphics[height=0.24\textheight]{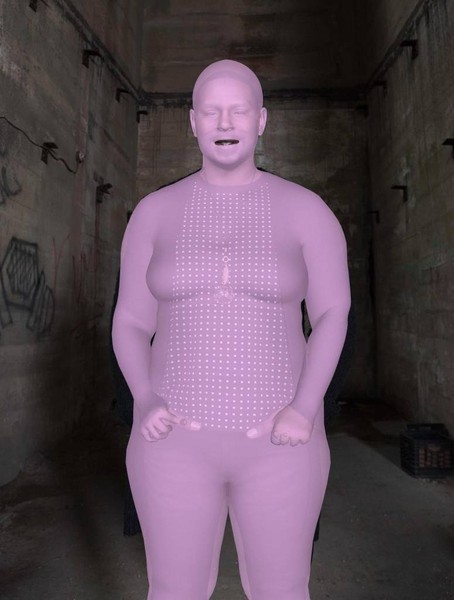}}
\subfloat[]{\includegraphics[height=0.24\textheight]{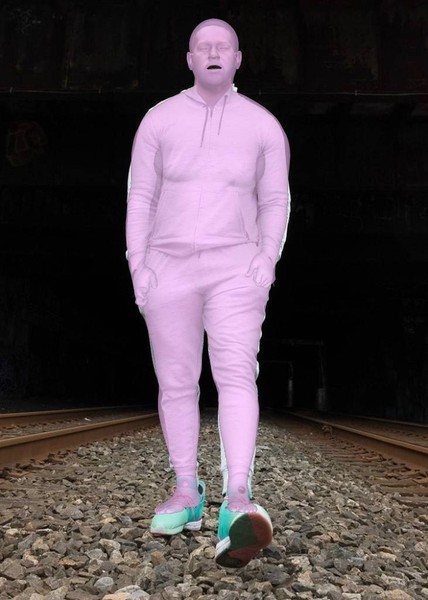}}
\subfloat[]{\includegraphics[height=0.24\textheight]{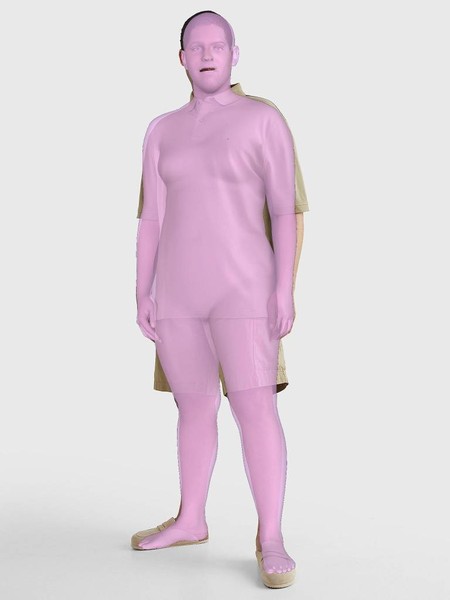}}\\

\raisebox{6.0\normalbaselineskip}[0pt][0pt]{\rotatebox[origin=c]{90}{\textcolor{green}{with} ADF}}
\subfloat[]{\includegraphics[height=0.24\textheight]{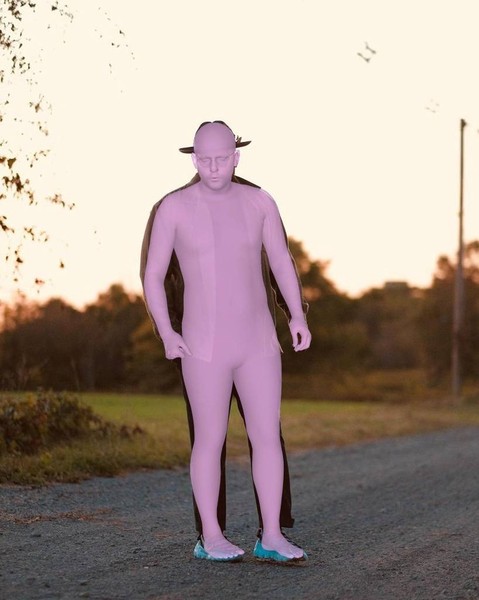}}
\subfloat[]{\includegraphics[height=0.24\textheight]{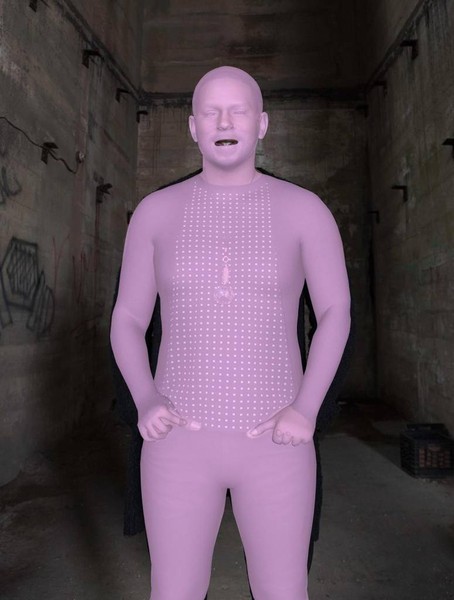}}
\subfloat[]{\includegraphics[height=0.24\textheight]{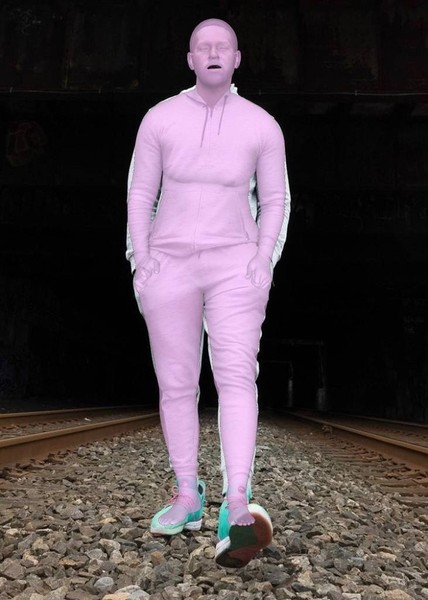}}
\subfloat[]{\includegraphics[height=0.24\textheight]{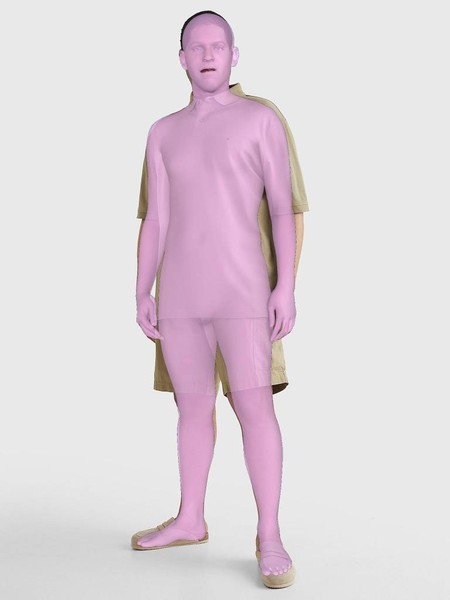}}

\caption{
\KBody{}{} fitting results \textcolor{red}{without} ADF on each even row and \textcolor{green}{with} ADF on each odd row.
}
\label{fig:adf_hm_mens_big}
\end{figure*}

%% file: figures/supp/adf_mens_big1.tex
\begin{figure*}[!htbp]
\captionsetup[subfigure]{position=bottom,labelformat=empty}

\centering

\raisebox{6.0\normalbaselineskip}[0pt][0pt]{\rotatebox[origin=c]{90}{\textcolor{red}{w/o} ADF}}
\subfloat[]{\includegraphics[height=0.24\textheight]
{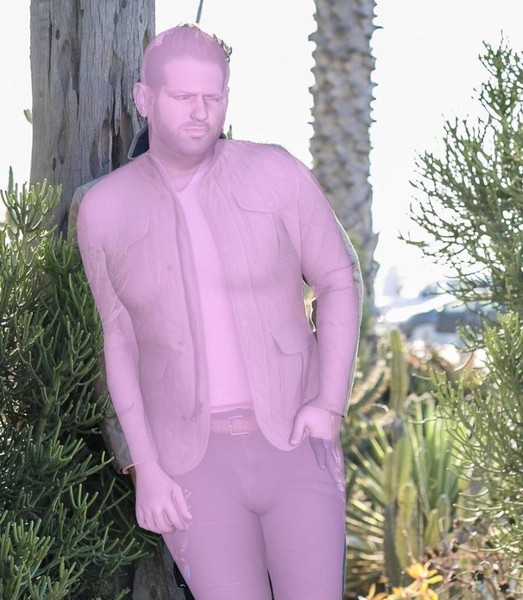}}
\subfloat[]{\includegraphics[height=0.24\textheight]
{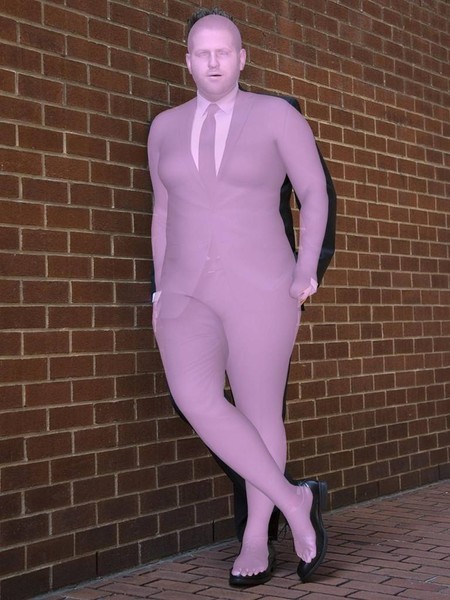}}
\subfloat[]{\includegraphics[height=0.24\textheight]
{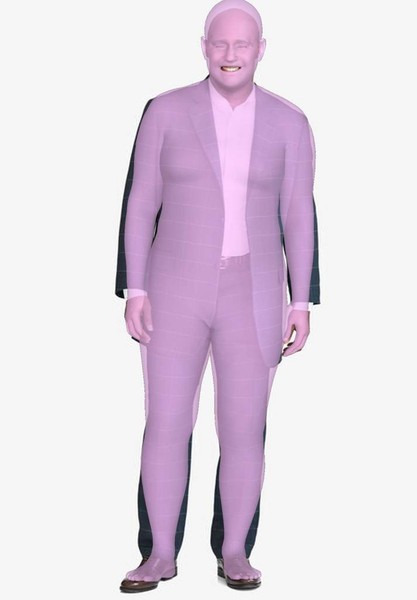}}
\subfloat[]{\includegraphics[height=0.24\textheight]
{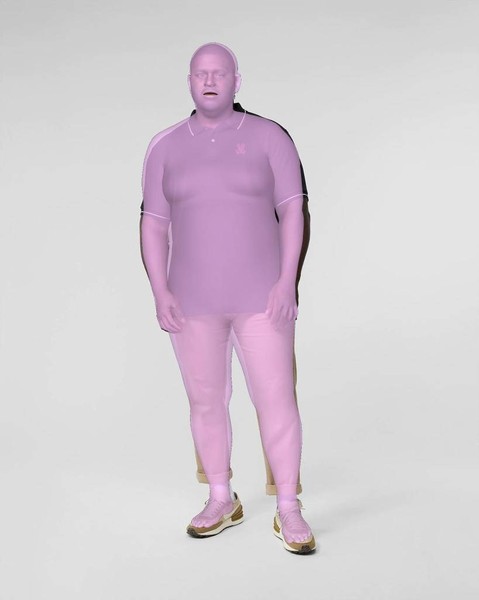}}\\

\raisebox{6.0\normalbaselineskip}[0pt][0pt]{\rotatebox[origin=c]{90}{\textcolor{green}{with} ADF}}
\subfloat[]{\includegraphics[height=0.24\textheight]{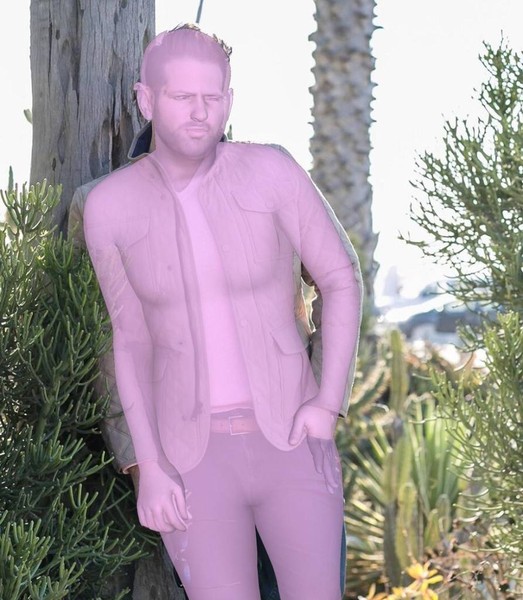}}
\subfloat[]{\includegraphics[height=0.24\textheight]{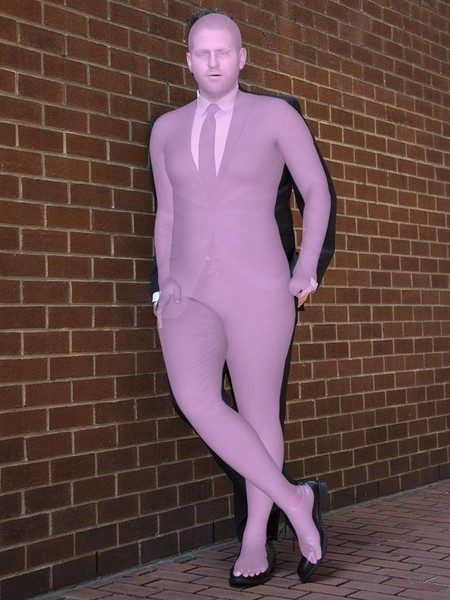}}
\subfloat[]{\includegraphics[height=0.24\textheight]{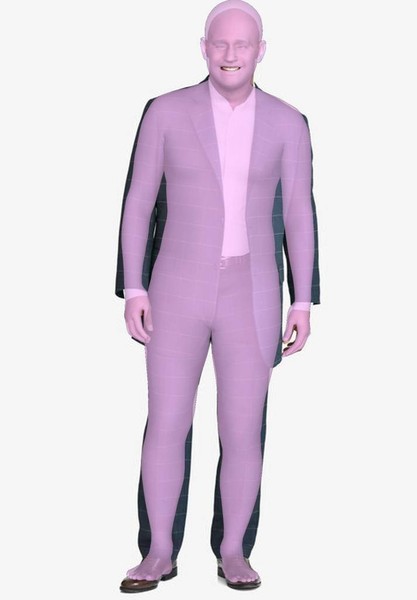}}
\subfloat[]{\includegraphics[height=0.24\textheight]{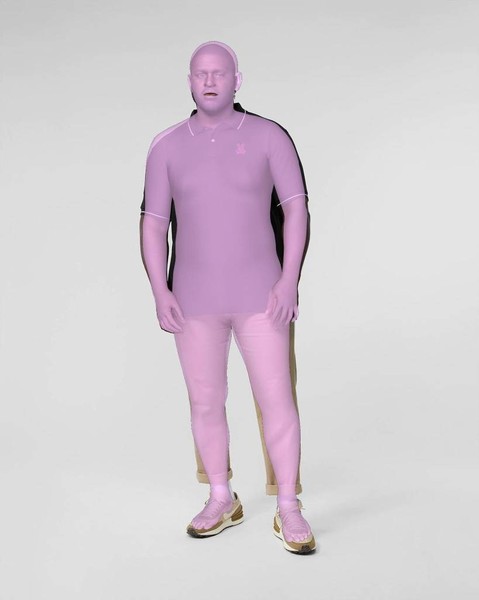}}\\

\raisebox{6.0\normalbaselineskip}[0pt][0pt]{\rotatebox[origin=c]{90}{\textcolor{red}{w/o} ADF}}
\subfloat[]{\includegraphics[height=0.2\textheight]
{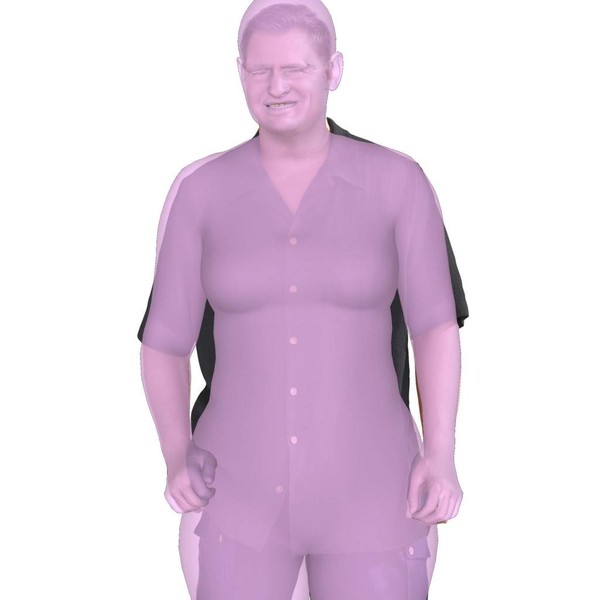}}
\subfloat[]{\includegraphics[height=0.2\textheight]
{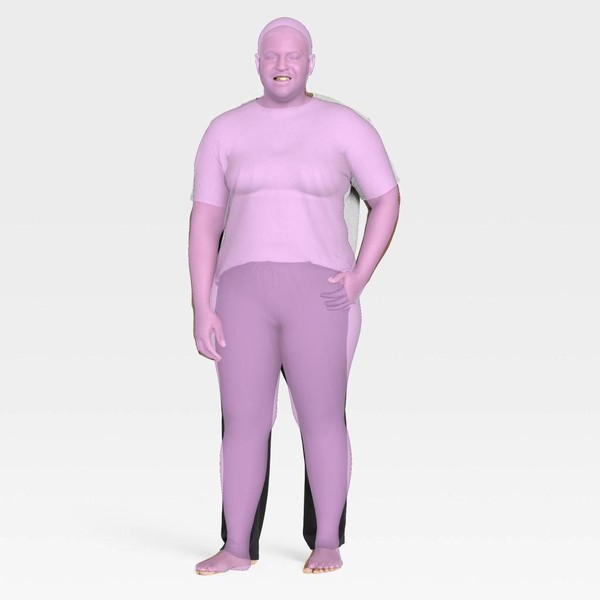}}
\subfloat[]{\includegraphics[height=0.2\textheight]
{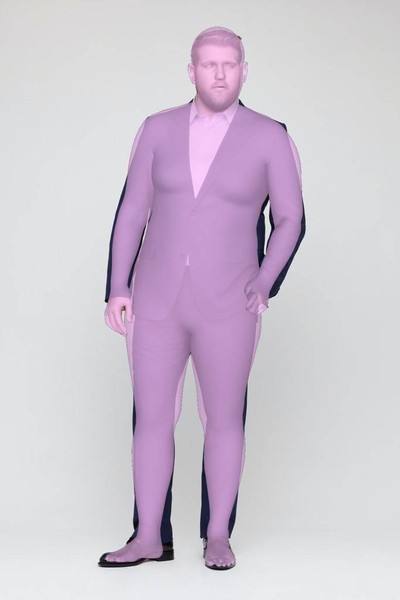}}
\subfloat[]{\includegraphics[height=0.2\textheight]
{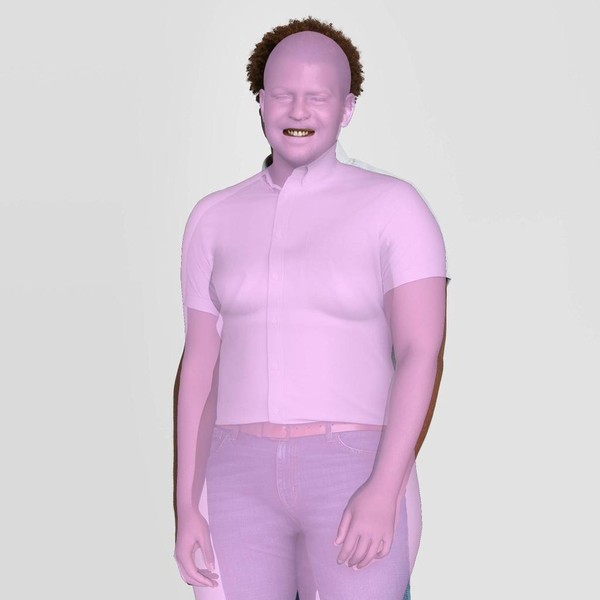}}\\

\raisebox{6.0\normalbaselineskip}[0pt][0pt]{\rotatebox[origin=c]{90}{\textcolor{green}{with} ADF}}
\subfloat[]{\includegraphics[height=0.2\textheight]{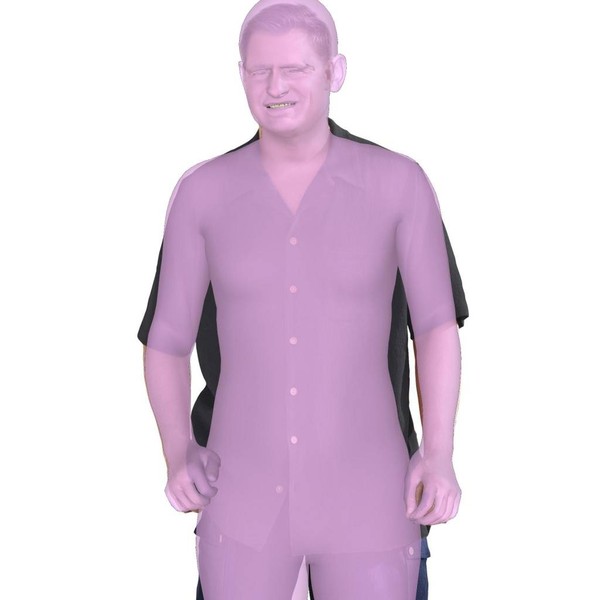}}
\subfloat[]{\includegraphics[height=0.2\textheight]{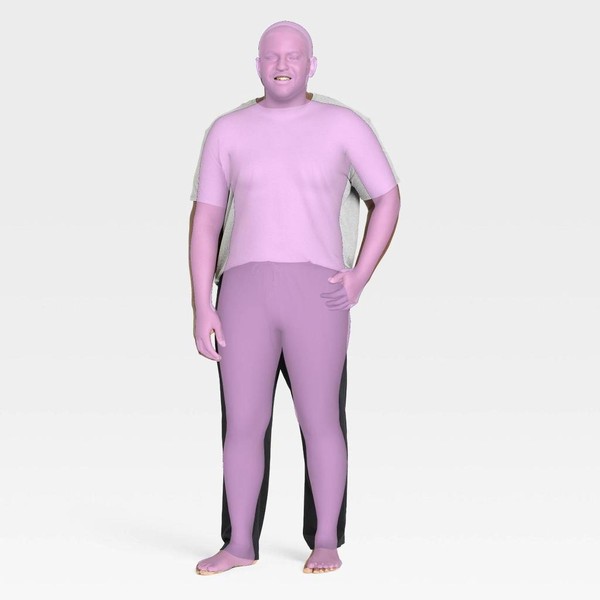}}
\subfloat[]{\includegraphics[height=0.2\textheight]{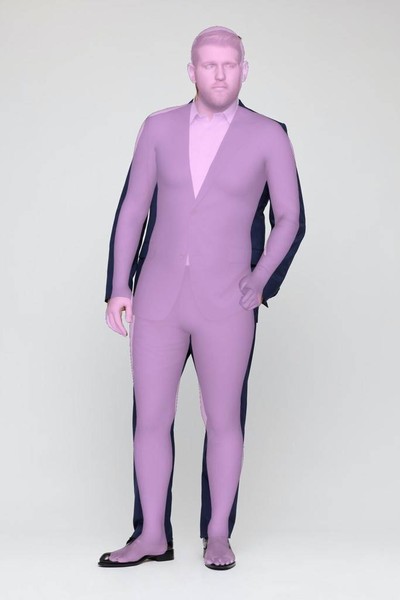}}
\subfloat[]{\includegraphics[height=0.2\textheight]{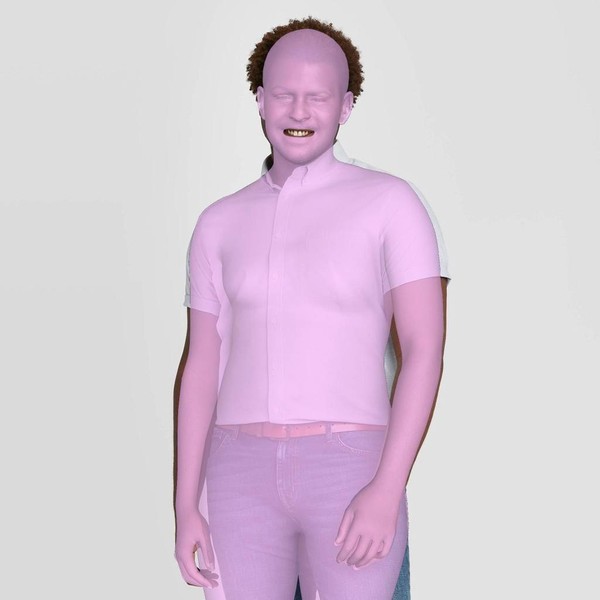}}

\caption{
\KBody{}{} fitting results \textcolor{red}{without} ADF on each even row and \textcolor{green}{with} ADF on each odd row.
}
\label{fig:adf_mens_big1}
\end{figure*}

%% file: figures/supp/partial_faherty1.tex
\begin{figure*}[!htbp]
\captionsetup[subfigure]{position=bottom,labelformat=empty}

\centering

\subfloat{\includegraphics[height=0.24\textheight]{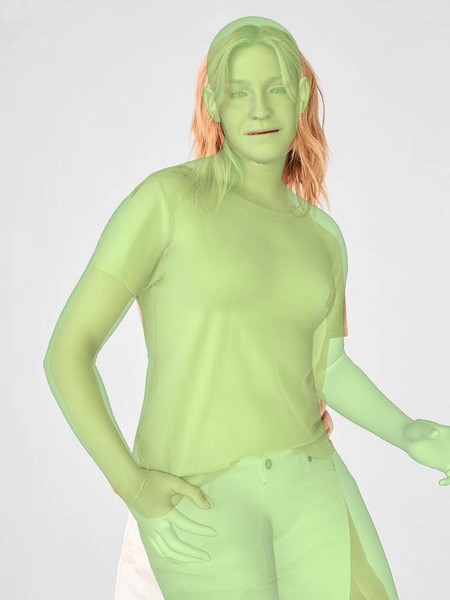}}
\subfloat{\includegraphics[height=0.24\textheight]{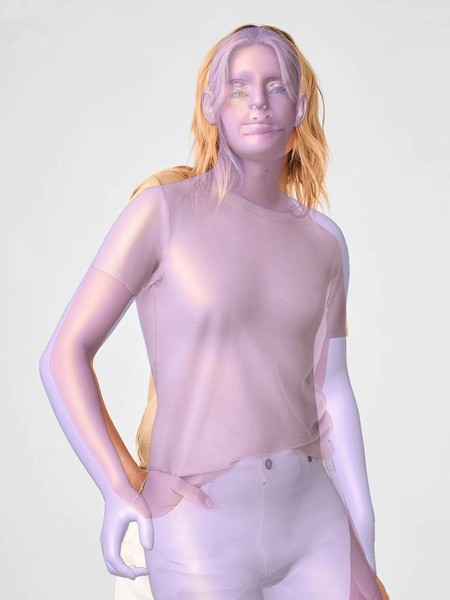}}
\subfloat{\includegraphics[height=0.24\textheight]{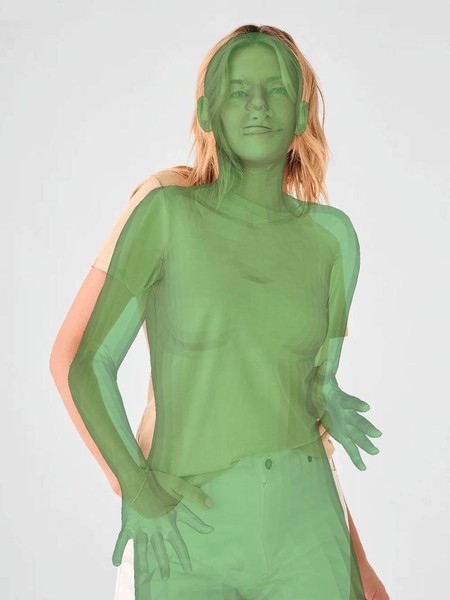}}
\subfloat{\includegraphics[height=0.24\textheight]{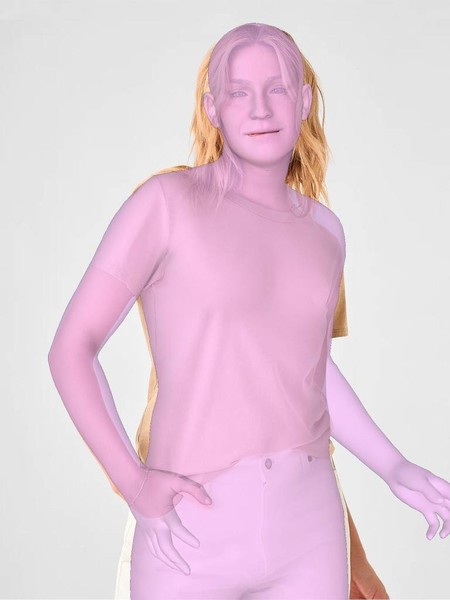}}\\

\subfloat{\includegraphics[height=0.24\textheight]{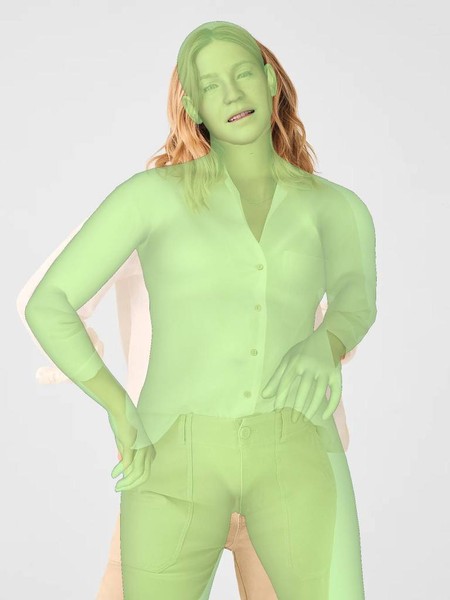}}
\subfloat{\includegraphics[height=0.24\textheight]{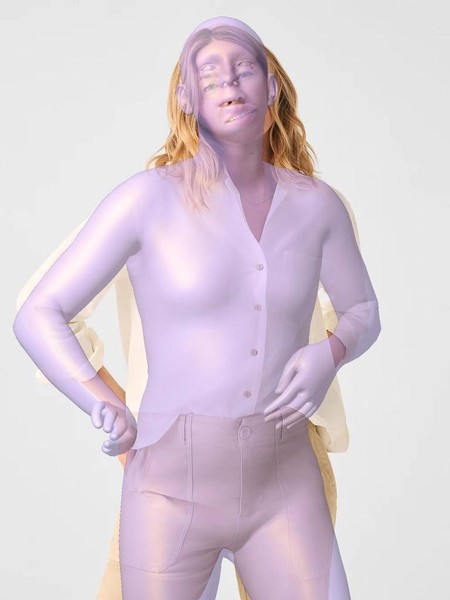}}
\subfloat{\includegraphics[height=0.24\textheight]{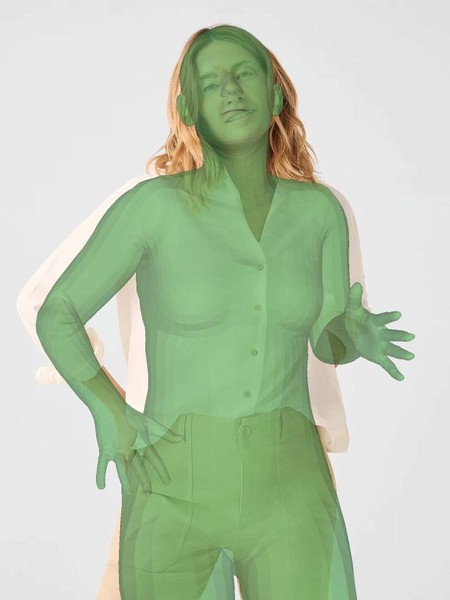}}
\subfloat{\includegraphics[height=0.24\textheight]{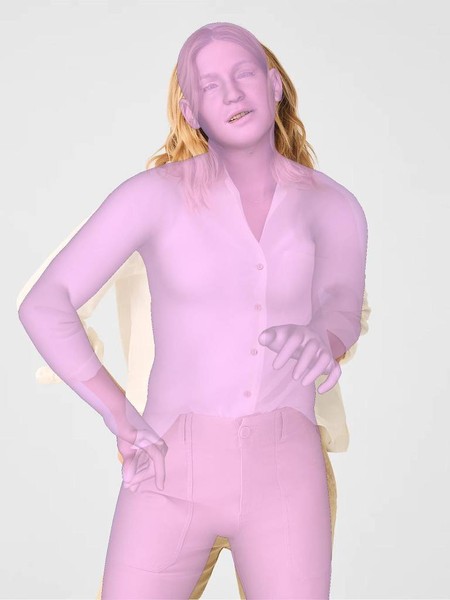}}\\

\subfloat[]{\includegraphics[height=0.24\textheight]{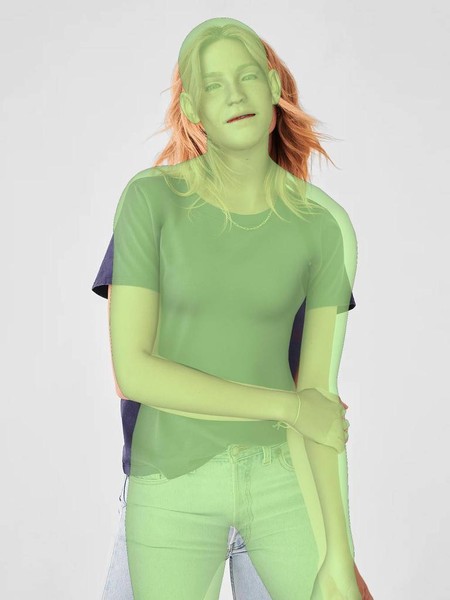}}
\subfloat[]{\includegraphics[height=0.24\textheight]{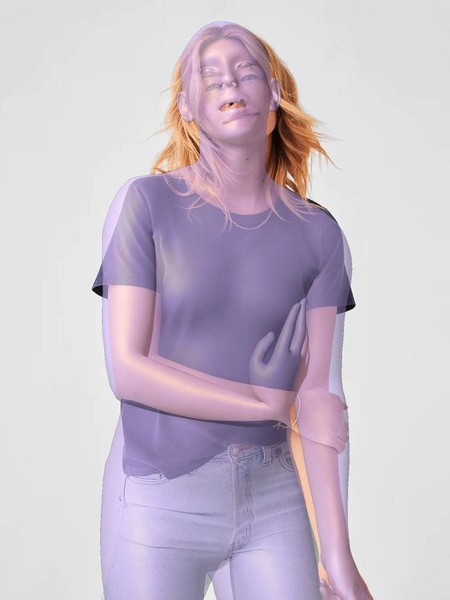}}
\subfloat[]{\includegraphics[height=0.24\textheight]{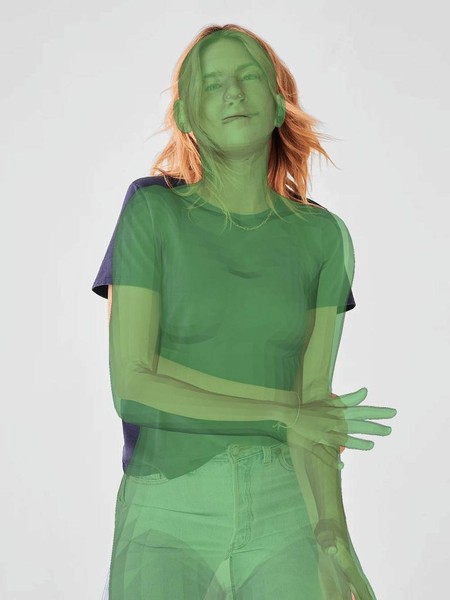}}
\subfloat[]{\includegraphics[height=0.24\textheight]{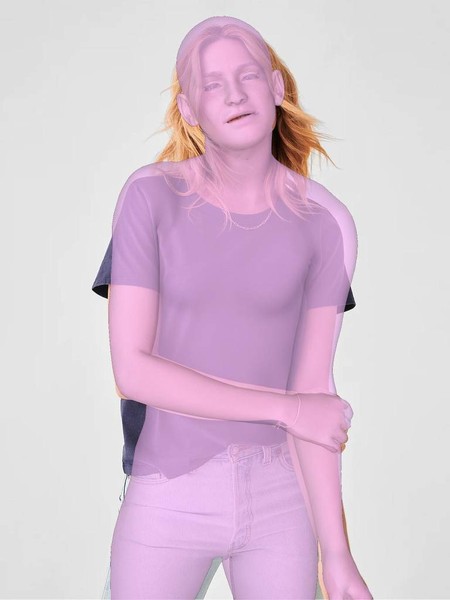}}\\

\vspace{-10pt}

\subfloat[SMPLify-X \cite{pavlakos2019expressive}]
{\includegraphics[height=0.24\textheight]{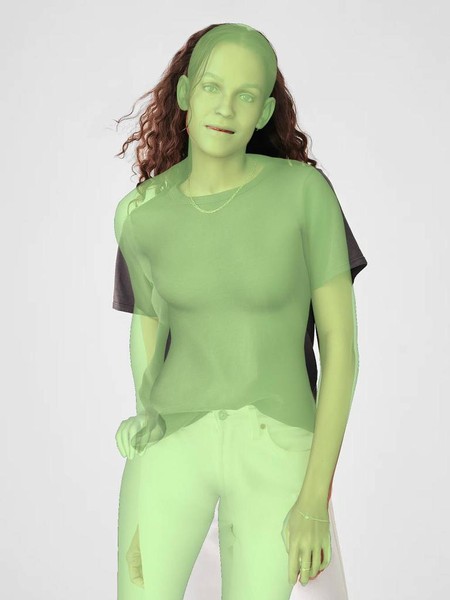}}
\subfloat[PyMAF-X \cite{pymafx2022}]{\includegraphics[height=0.24\textheight]{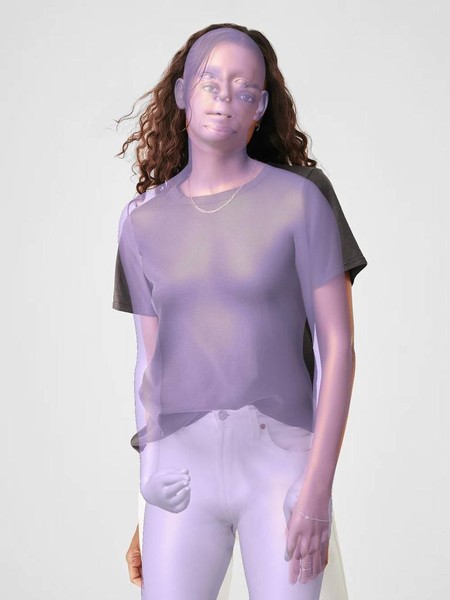}}
\subfloat[SHAPY \cite{choutas2022accurate}]
{\includegraphics[height=0.24\textheight]{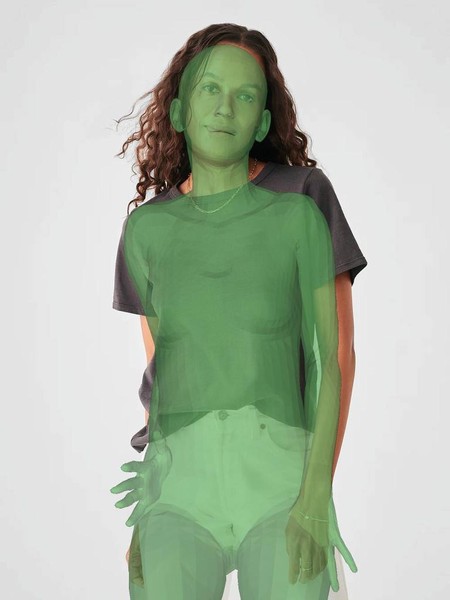}}
\subfloat[\KBody{-.1}{.035} (Ours)]{\includegraphics[height=0.24\textheight]{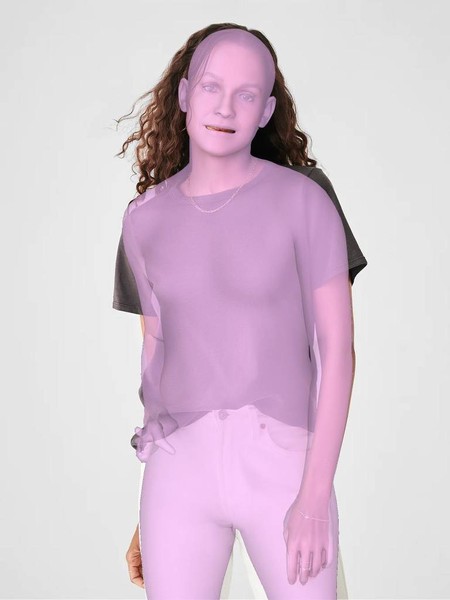}}

\caption{
Left-to-right: SMPLify-X \cite{pavlakos2019expressive} (\textcolor{caribbeangreen2}{light green}), PyMAF-X \cite{pymafx2022} (\textcolor{violet}{purple}), SHAPY \cite{choutas2022accurate} (\textcolor{jade}{green}) and KBody (\textcolor{candypink}{pink}).
}
\label{fig:partial_fh1}
\end{figure*}

%% file: figures/supp/partial_faherty2.tex
\begin{figure*}[!htbp]
\captionsetup[subfigure]{position=bottom,labelformat=empty}

\centering

\subfloat{\includegraphics[height=0.24\textheight]{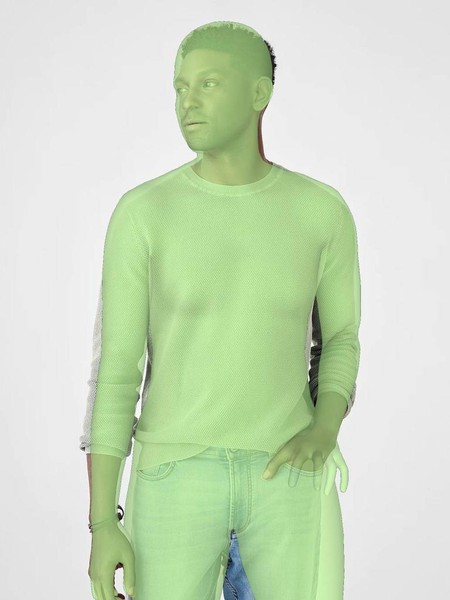}}
\subfloat{\includegraphics[height=0.24\textheight]{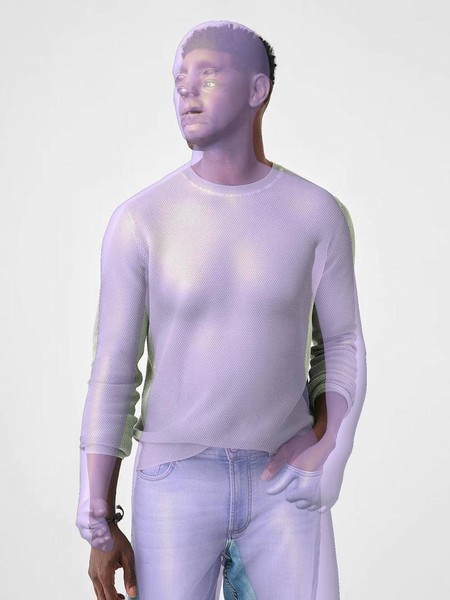}}
\subfloat{\includegraphics[height=0.24\textheight]{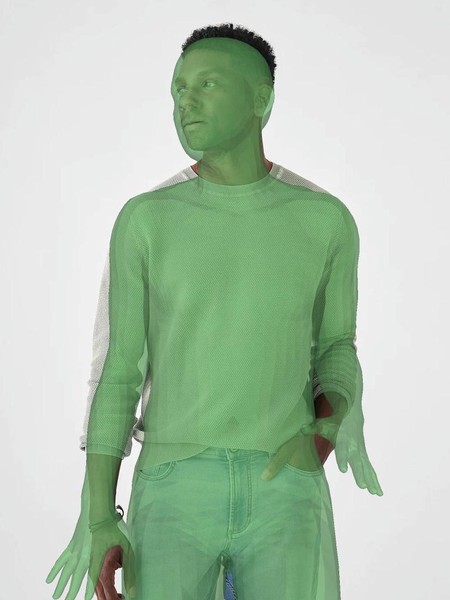}}
\subfloat{\includegraphics[height=0.24\textheight]{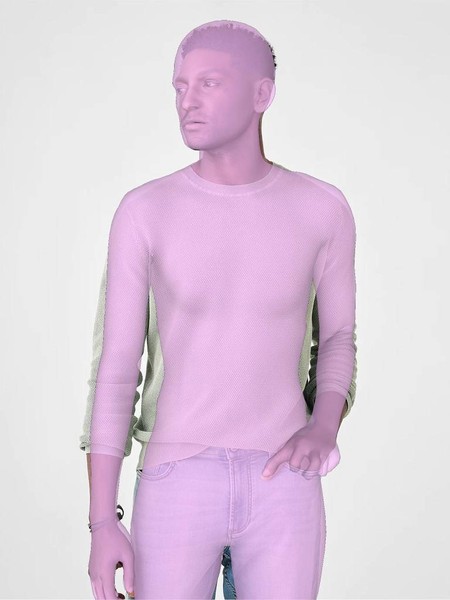}}\\

\subfloat{\includegraphics[height=0.24\textheight]{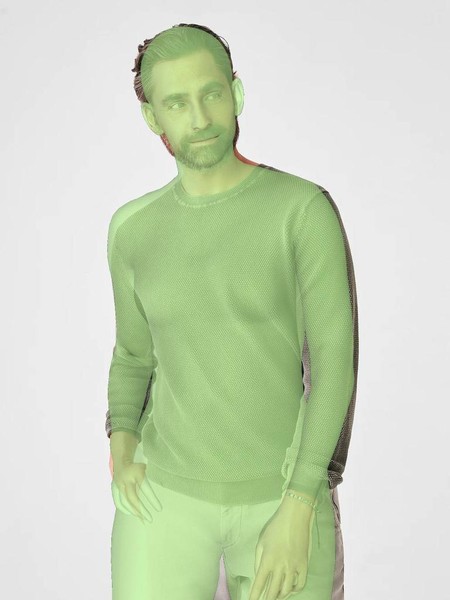}}
\subfloat{\includegraphics[height=0.24\textheight]{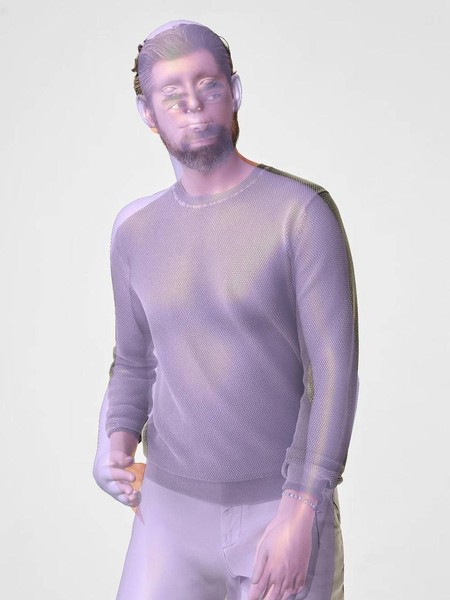}}
\subfloat{\includegraphics[height=0.24\textheight]{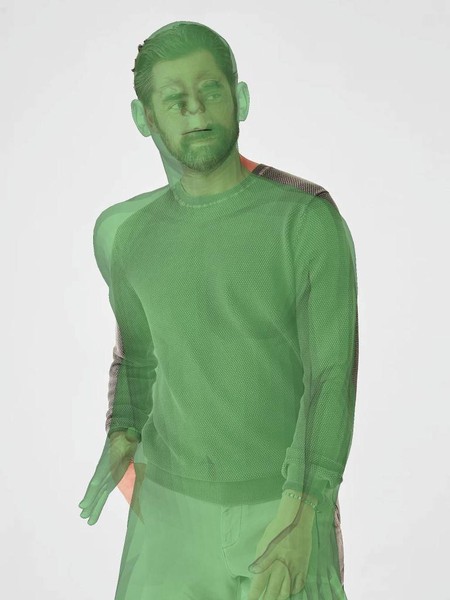}}
\subfloat{\includegraphics[height=0.24\textheight]{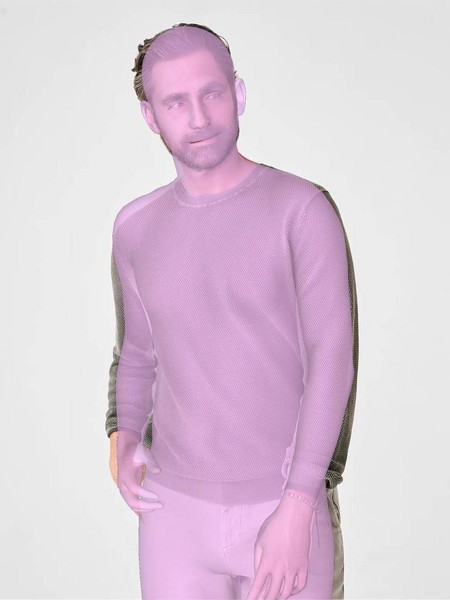}}\\

\subfloat[]{\includegraphics[height=0.24\textheight]{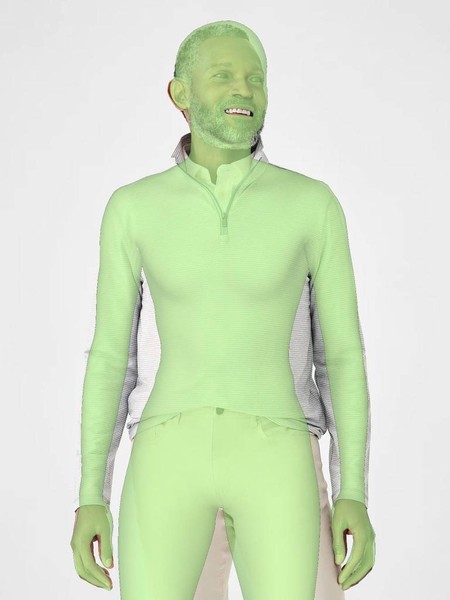}}
\subfloat[]{\includegraphics[height=0.24\textheight]{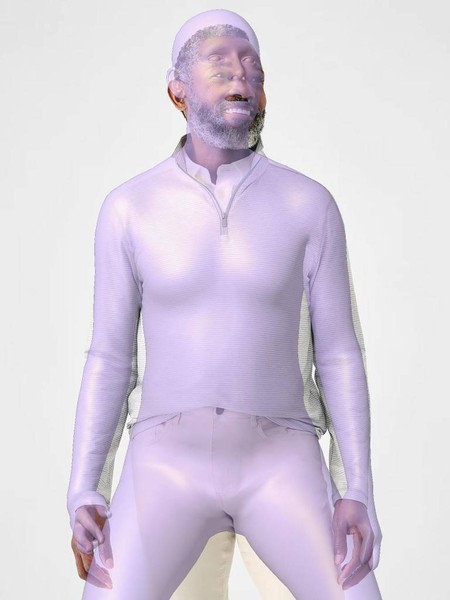}}
\subfloat[]{\includegraphics[height=0.24\textheight]{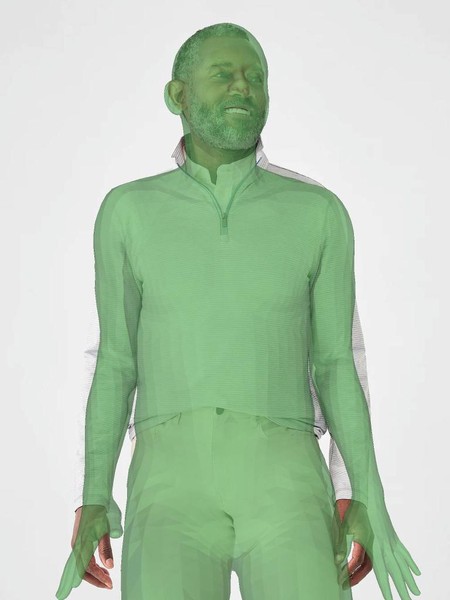}}
\subfloat[]{\includegraphics[height=0.24\textheight]{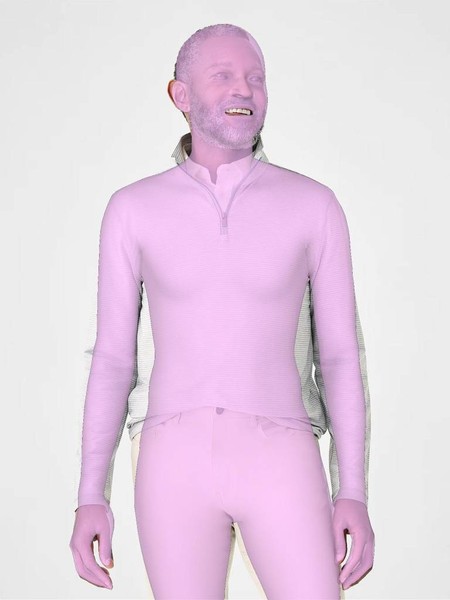}}\\

\vspace{-10pt}

\subfloat[SMPLify-X \cite{pavlakos2019expressive}]
{\includegraphics[height=0.24\textheight]{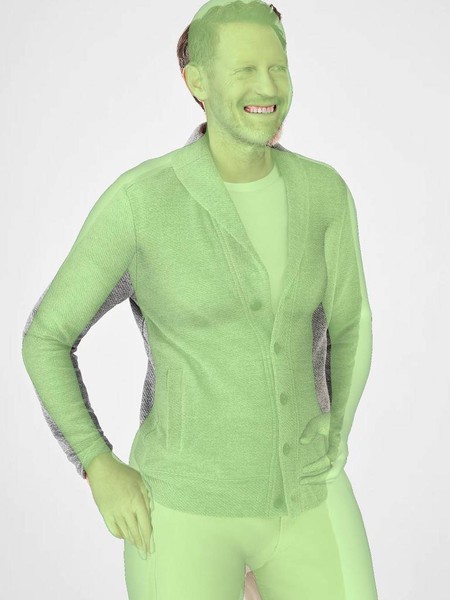}}
\subfloat[PyMAF-X \cite{pymafx2022}]{\includegraphics[height=0.24\textheight]{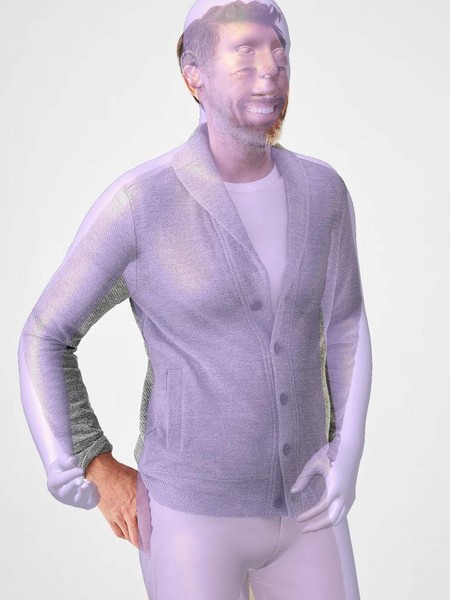}}
\subfloat[SHAPY \cite{choutas2022accurate}]
{\includegraphics[height=0.24\textheight]{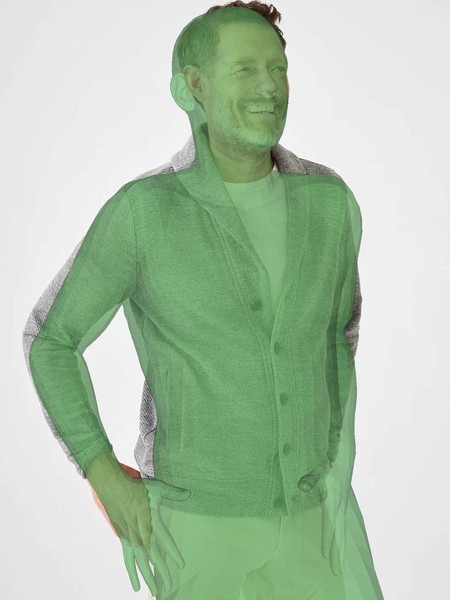}}
\subfloat[\KBody{-.1}{.035} (Ours)]{\includegraphics[height=0.24\textheight]{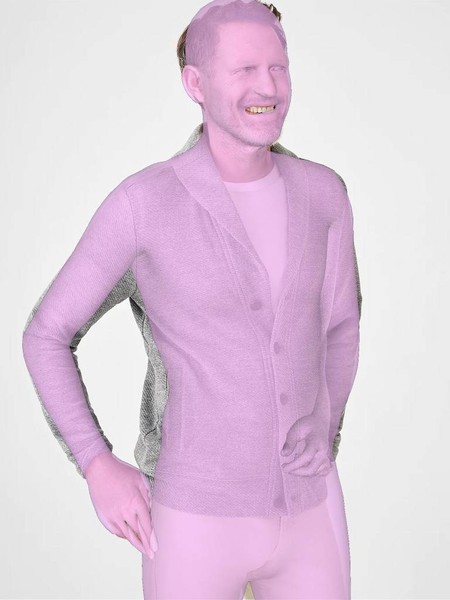}}

\caption{
Left-to-right: SMPLify-X \cite{pavlakos2019expressive} (\textcolor{caribbeangreen2}{light green}), PyMAF-X \cite{pymafx2022} (\textcolor{violet}{purple}), SHAPY \cite{choutas2022accurate} (\textcolor{jade}{green}) and KBody (\textcolor{candypink}{pink}).
}
\label{fig:partial_fh2}
\end{figure*}

%% file: figures/supp/partial_faherty3.tex
\begin{figure*}[!htbp]
\captionsetup[subfigure]{position=bottom,labelformat=empty}

\centering

\subfloat{\includegraphics[height=0.24\textheight]{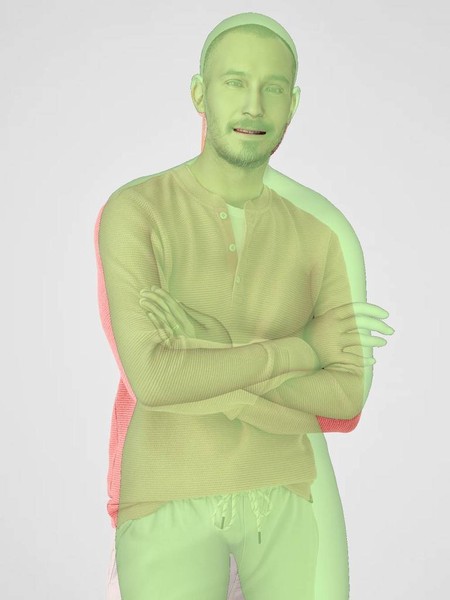}}
\subfloat{\includegraphics[height=0.24\textheight]{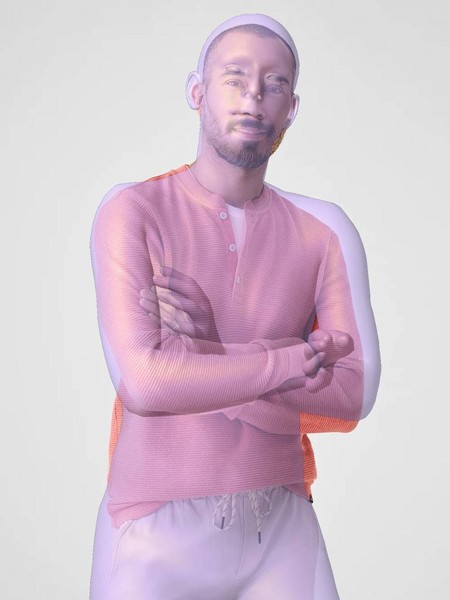}}
\subfloat{\includegraphics[height=0.24\textheight]{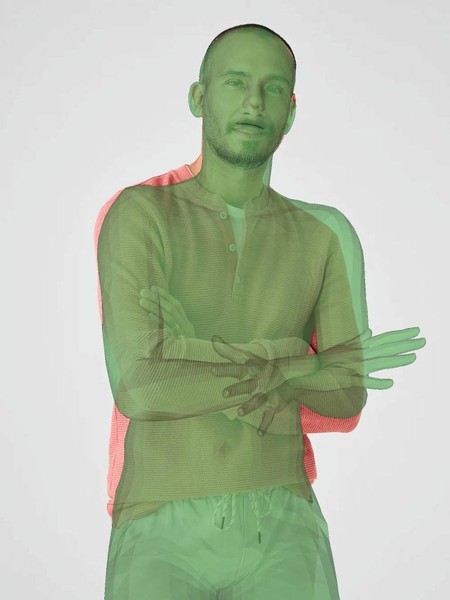}}
\subfloat{\includegraphics[height=0.24\textheight]{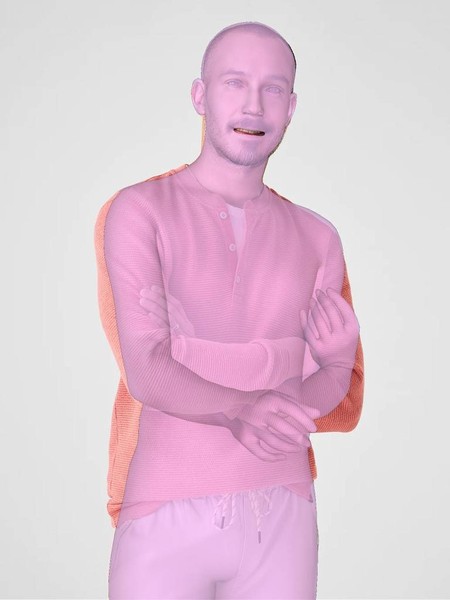}}\\

\subfloat{\includegraphics[height=0.24\textheight]{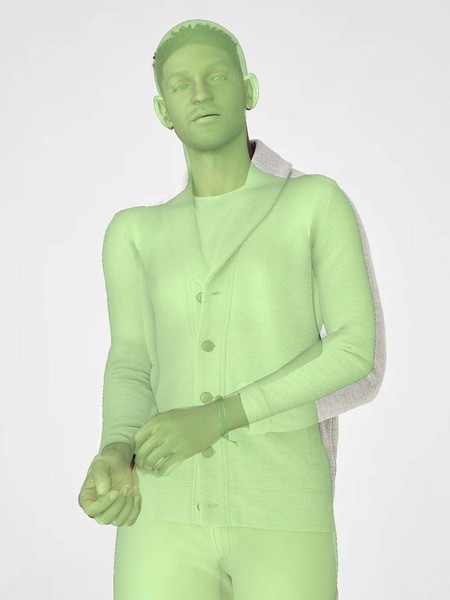}}
\subfloat{\includegraphics[height=0.24\textheight]{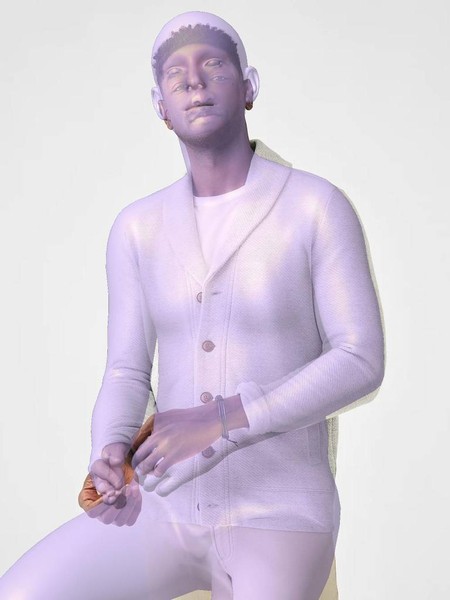}}
\subfloat{\includegraphics[height=0.24\textheight]{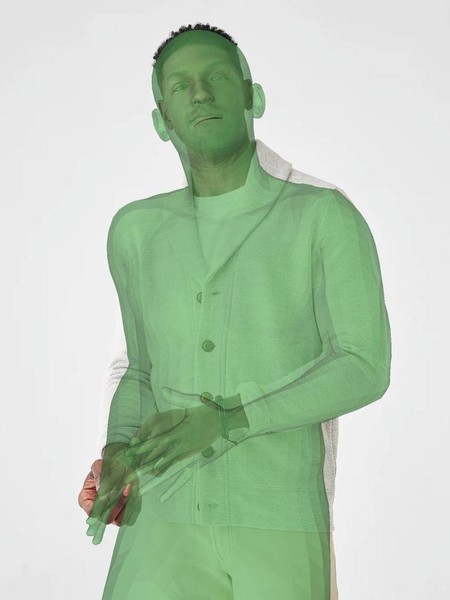}}
\subfloat{\includegraphics[height=0.24\textheight]{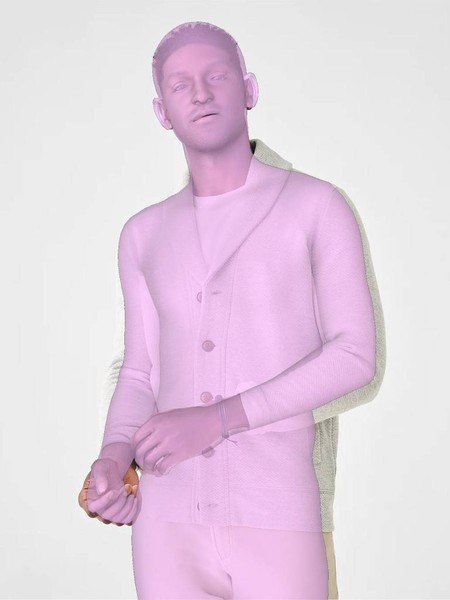}}\\

\subfloat[]{\includegraphics[height=0.24\textheight]{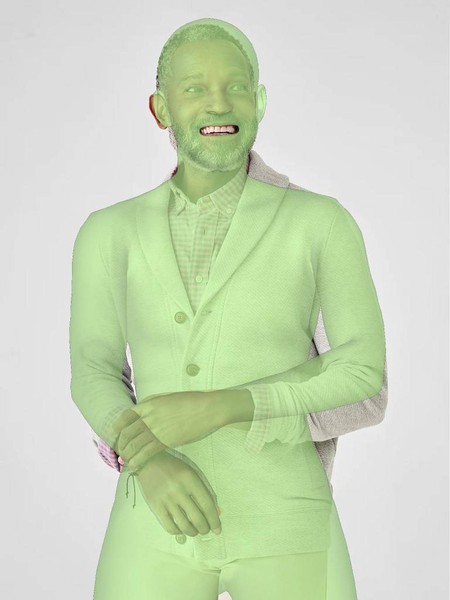}}
\subfloat[]{\includegraphics[height=0.24\textheight]{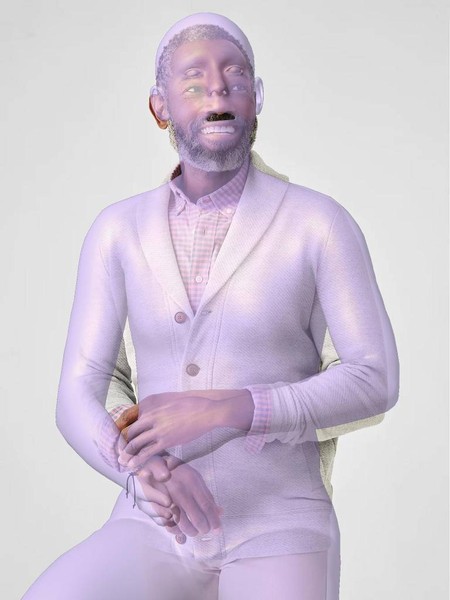}}
\subfloat[]{\includegraphics[height=0.24\textheight]{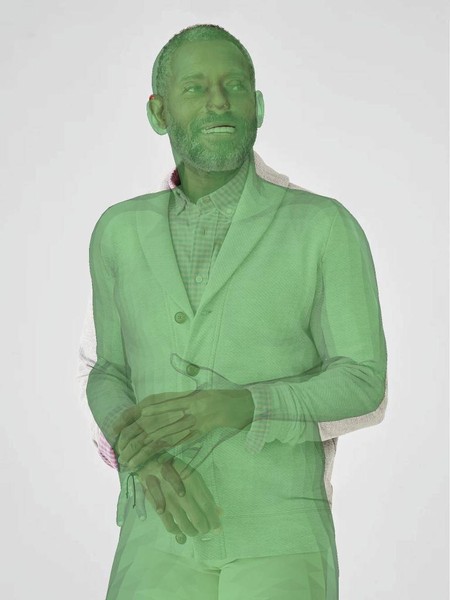}}
\subfloat[]{\includegraphics[height=0.24\textheight]{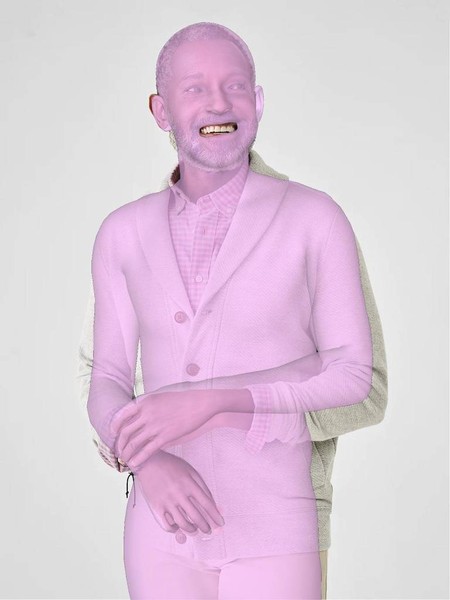}}\\

\vspace{-10pt}

\subfloat[SMPLify-X \cite{pavlakos2019expressive}]
{\includegraphics[height=0.24\textheight]{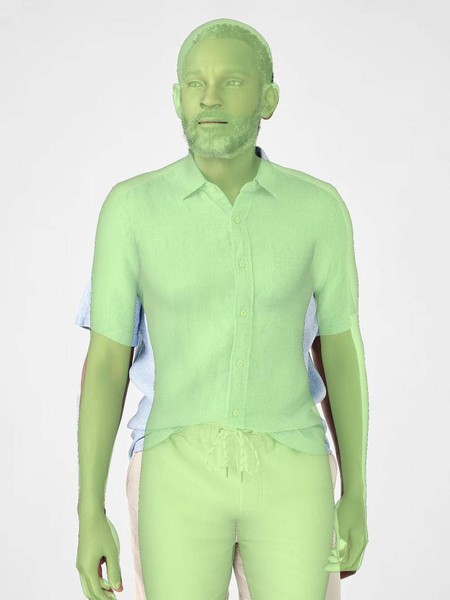}}
\subfloat[PyMAF-X \cite{pymafx2022}]{\includegraphics[height=0.24\textheight]{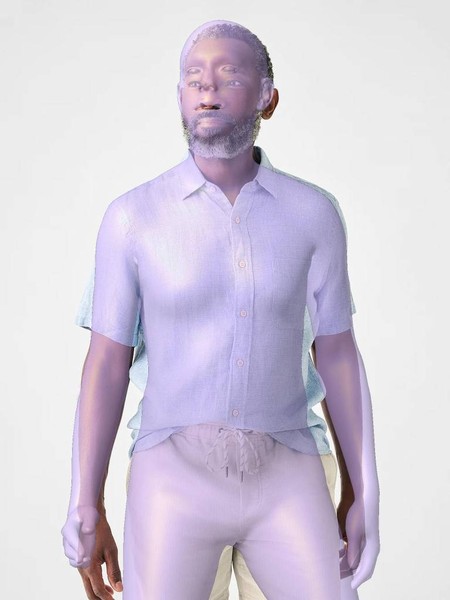}}
\subfloat[SHAPY \cite{choutas2022accurate}]
{\includegraphics[height=0.24\textheight]{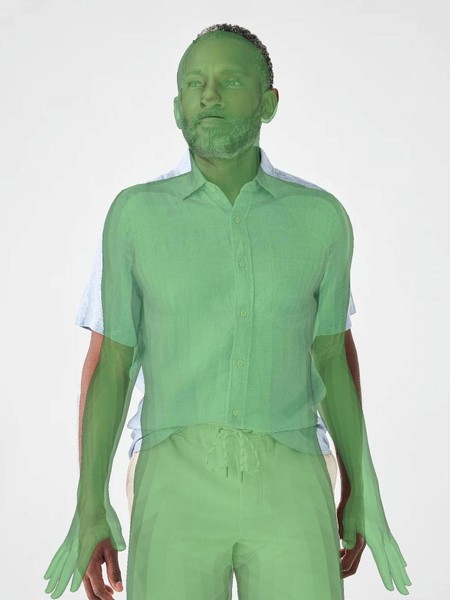}}
\subfloat[\KBody{-.1}{.035} (Ours)]{\includegraphics[height=0.24\textheight]{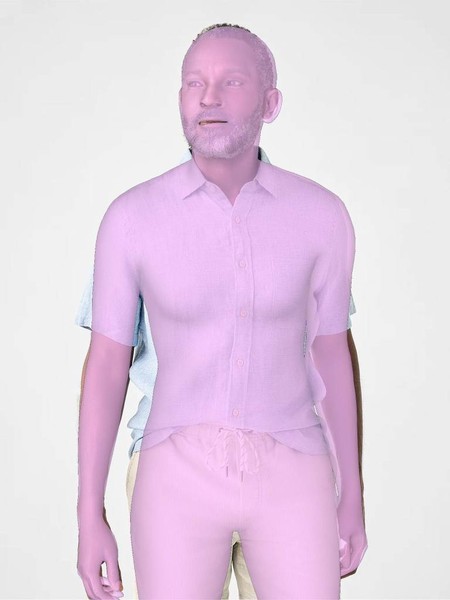}}

\caption{
Left-to-right: SMPLify-X \cite{pavlakos2019expressive} (\textcolor{caribbeangreen2}{light green}), PyMAF-X \cite{pymafx2022} (\textcolor{violet}{purple}), SHAPY \cite{choutas2022accurate} (\textcolor{jade}{green}) and KBody (\textcolor{candypink}{pink}).
}
\label{fig:partial_fh3}
\end{figure*}

%% file: figures/supp/partial_faherty4.tex
\begin{figure*}[!htbp]
\captionsetup[subfigure]{position=bottom,labelformat=empty}

\centering

\subfloat{\includegraphics[height=0.24\textheight]{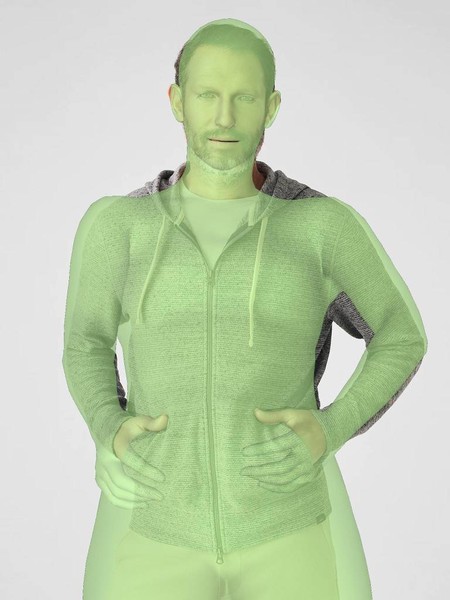}}
\subfloat{\includegraphics[height=0.24\textheight]{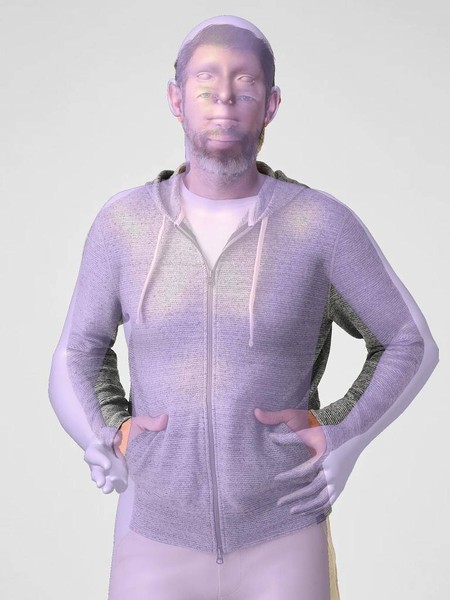}}
\subfloat{\includegraphics[height=0.24\textheight]{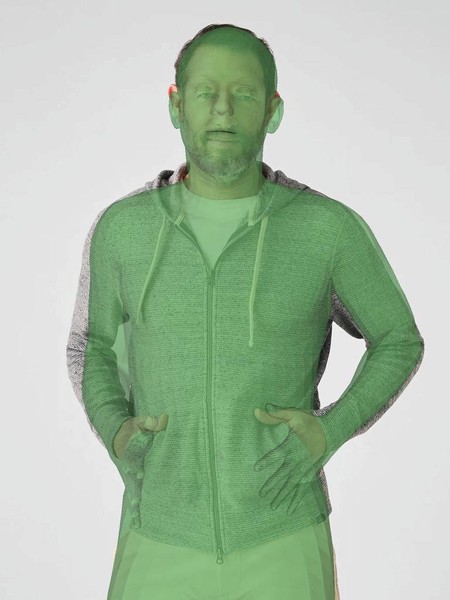}}
\subfloat{\includegraphics[height=0.24\textheight]{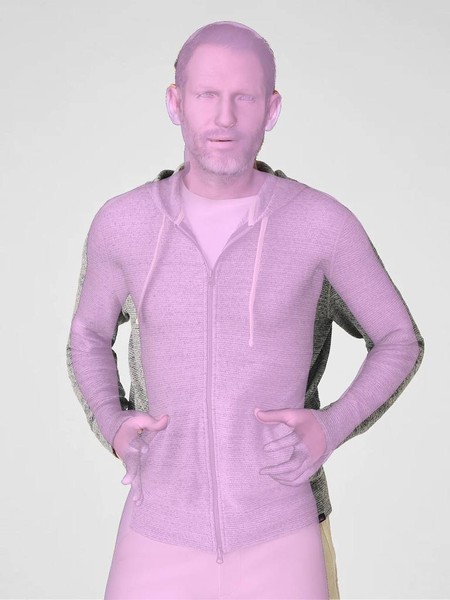}}\\

\subfloat{\includegraphics[height=0.24\textheight]{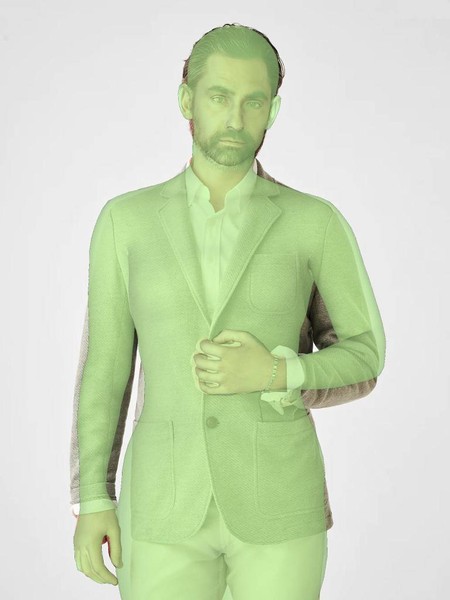}}
\subfloat{\includegraphics[height=0.24\textheight]{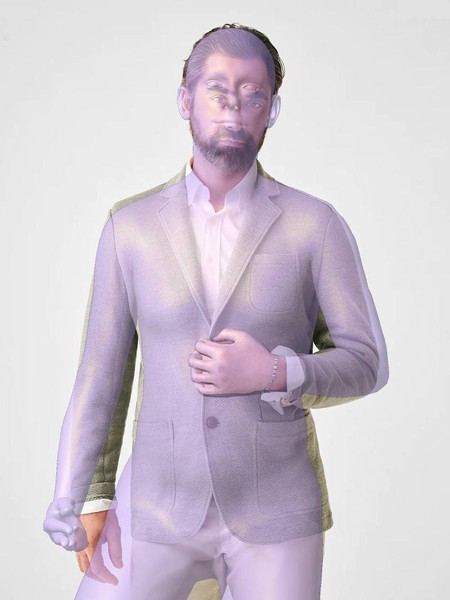}}
\subfloat{\includegraphics[height=0.24\textheight]{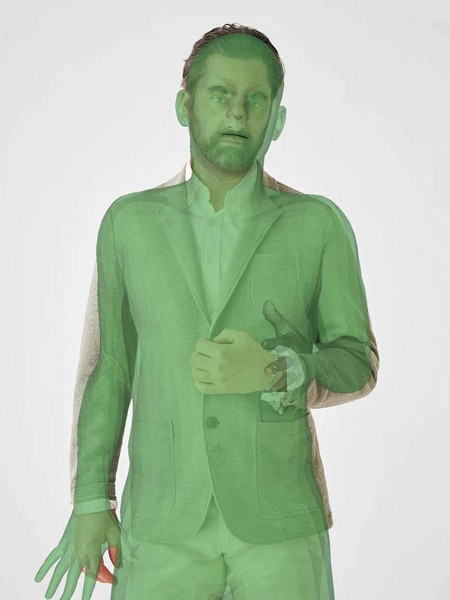}}
\subfloat{\includegraphics[height=0.24\textheight]{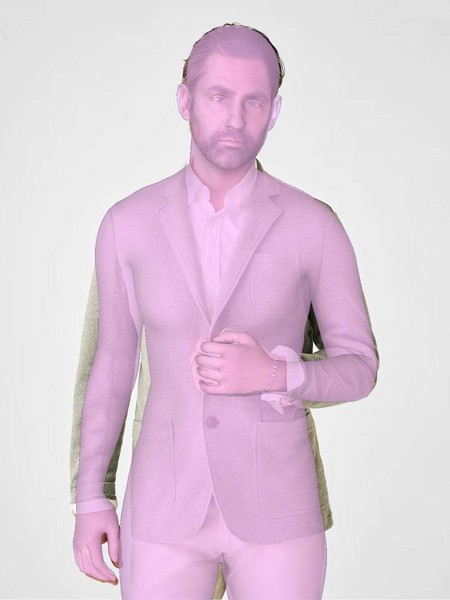}}\\

\subfloat[]{\includegraphics[height=0.24\textheight]{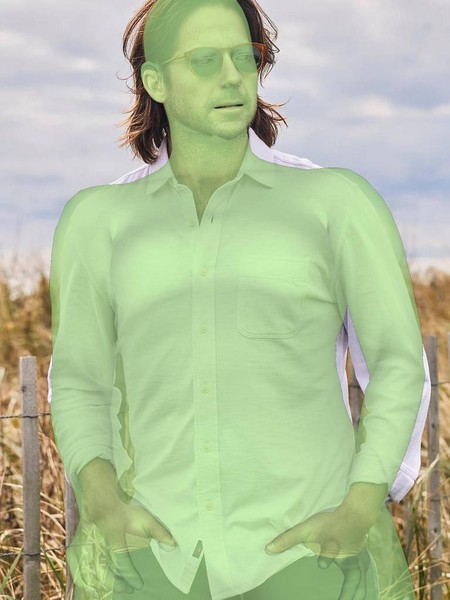}}
\subfloat[]{\includegraphics[height=0.24\textheight]{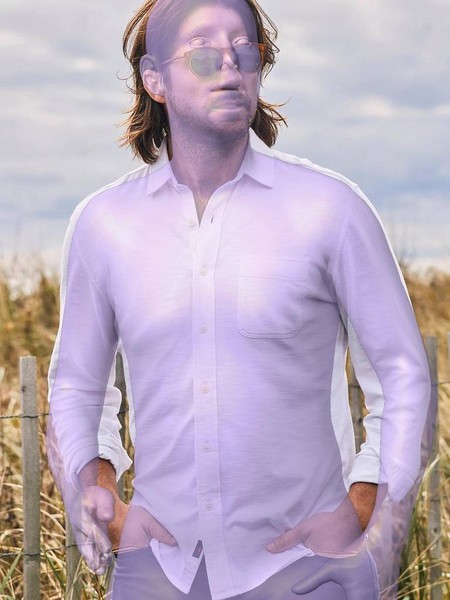}}
\subfloat[]{\includegraphics[height=0.24\textheight]{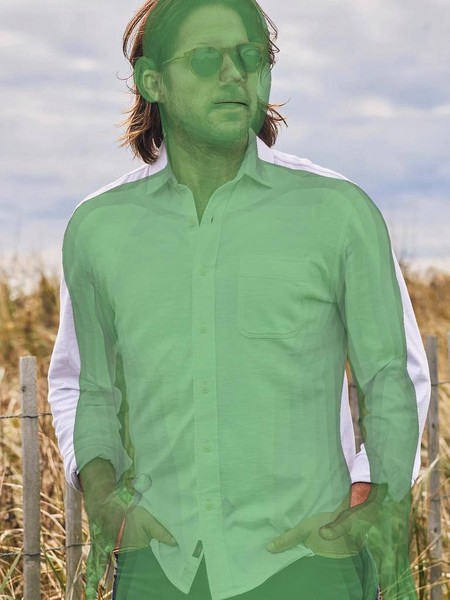}}
\subfloat[]{\includegraphics[height=0.24\textheight]{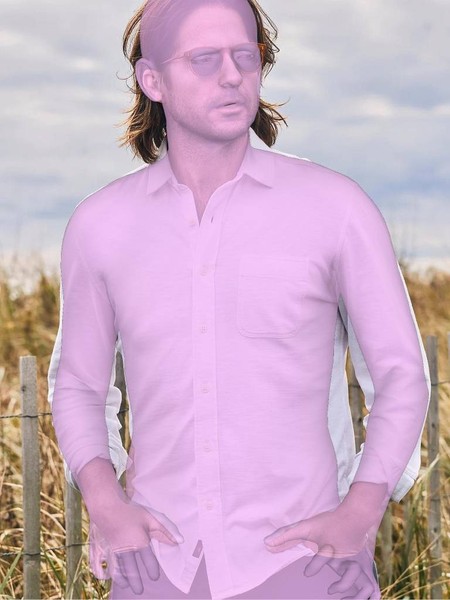}}\\

\vspace{-10pt}

\subfloat[SMPLify-X \cite{pavlakos2019expressive}]
{\includegraphics[height=0.24\textheight]{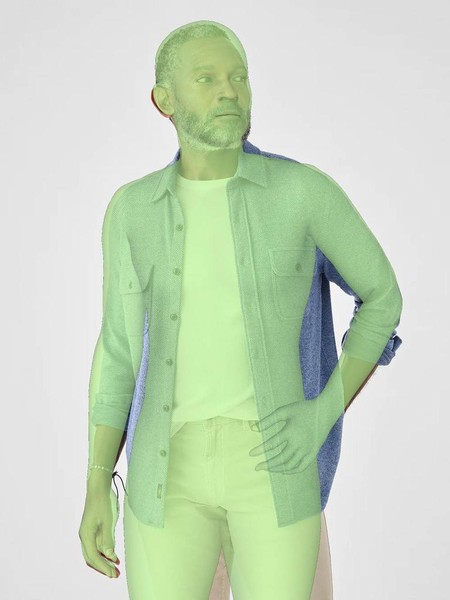}}
\subfloat[PyMAF-X \cite{pymafx2022}]{\includegraphics[height=0.24\textheight]{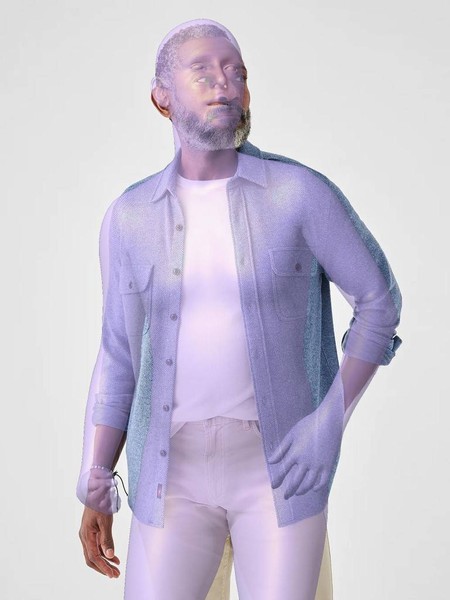}}
\subfloat[SHAPY \cite{choutas2022accurate}]
{\includegraphics[height=0.24\textheight]{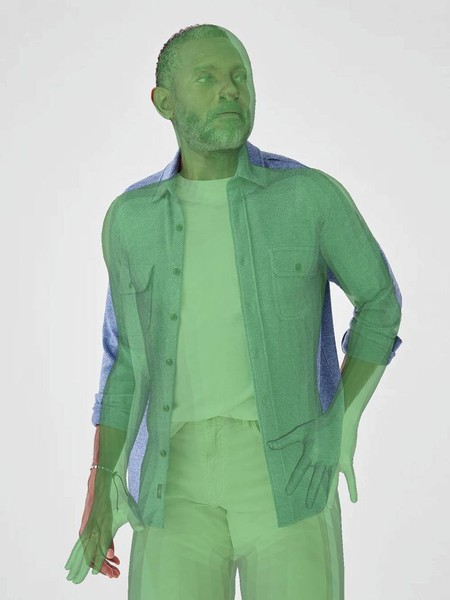}}
\subfloat[\KBody{-.1}{.035} (Ours)]{\includegraphics[height=0.24\textheight]{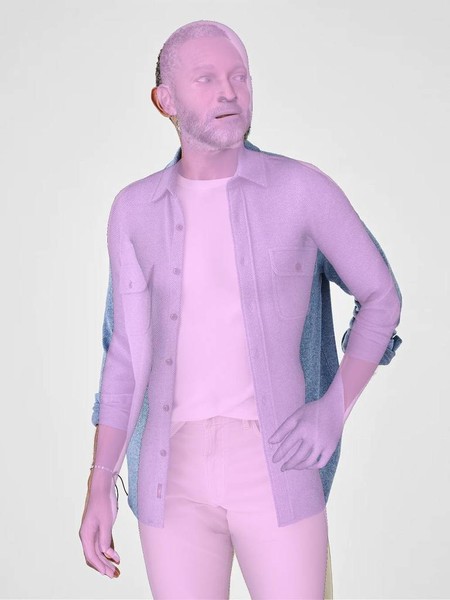}}

\caption{
Left-to-right: SMPLify-X \cite{pavlakos2019expressive} (\textcolor{caribbeangreen2}{light green}), PyMAF-X \cite{pymafx2022} (\textcolor{violet}{purple}), SHAPY \cite{choutas2022accurate} (\textcolor{jade}{green}) and KBody (\textcolor{candypink}{pink}).
}
\label{fig:partial_fh4}
\end{figure*}

%% file: figures/supp/partial_lululemone1.tex
\begin{figure*}[!htbp]
\captionsetup[subfigure]{position=bottom,labelformat=empty}

\centering

\subfloat{\includegraphics[height=0.24\textheight]{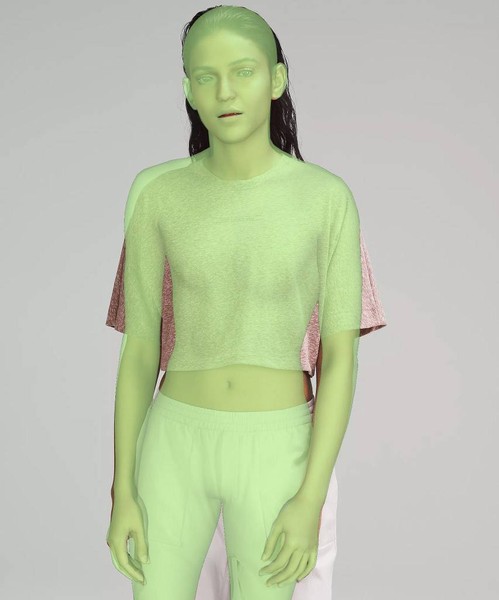}}
\subfloat{\includegraphics[height=0.24\textheight]{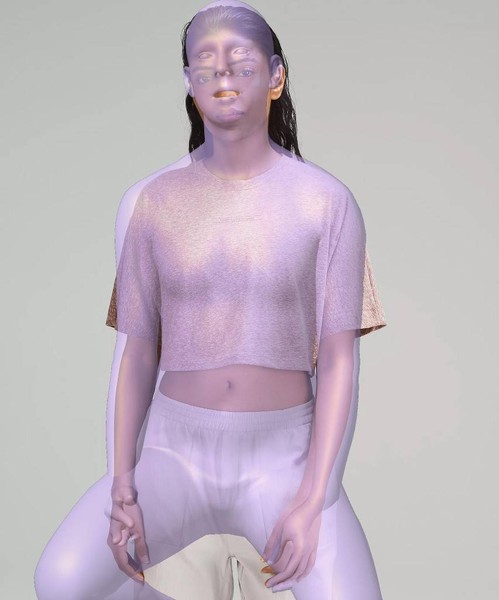}}
\subfloat{\includegraphics[height=0.24\textheight]{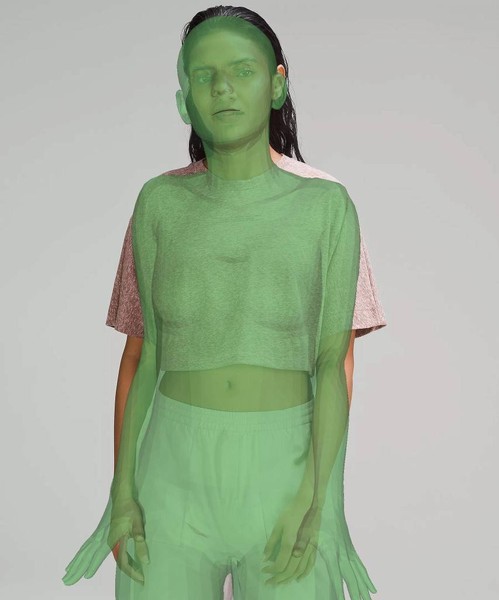}}
\subfloat{\includegraphics[height=0.24\textheight]{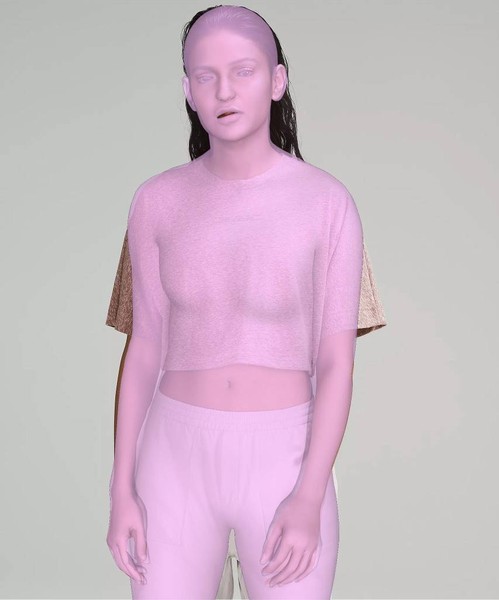}}\\

\subfloat{\includegraphics[height=0.24\textheight]{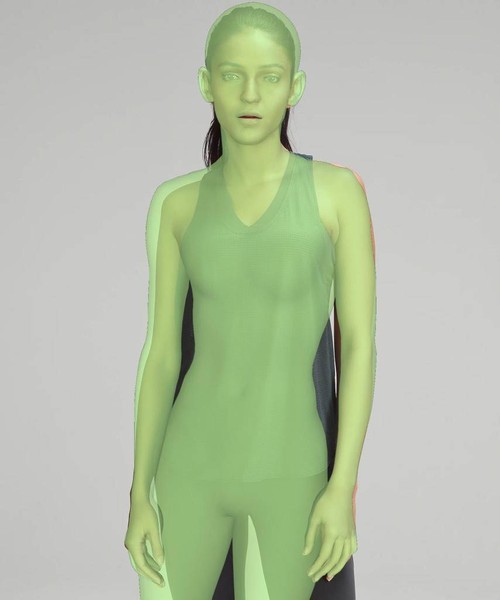}}
\subfloat{\includegraphics[height=0.24\textheight]{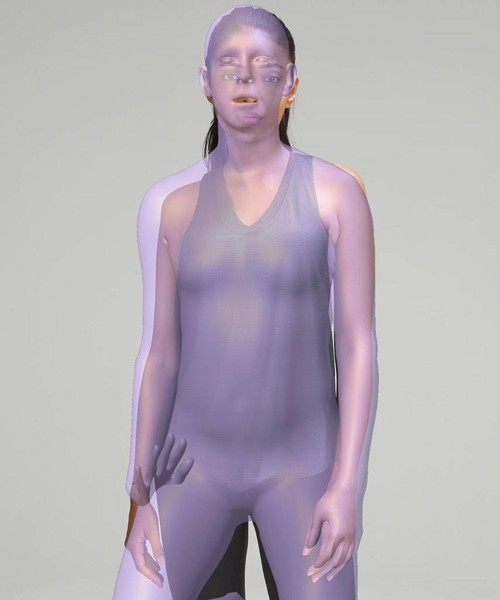}}
\subfloat{\includegraphics[height=0.24\textheight]{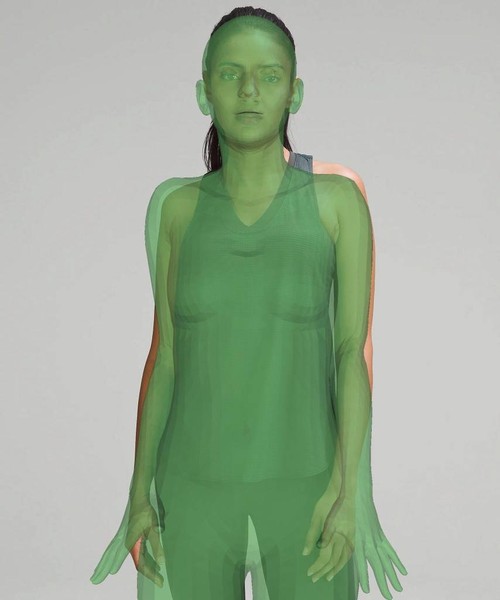}}
\subfloat{\includegraphics[height=0.24\textheight]{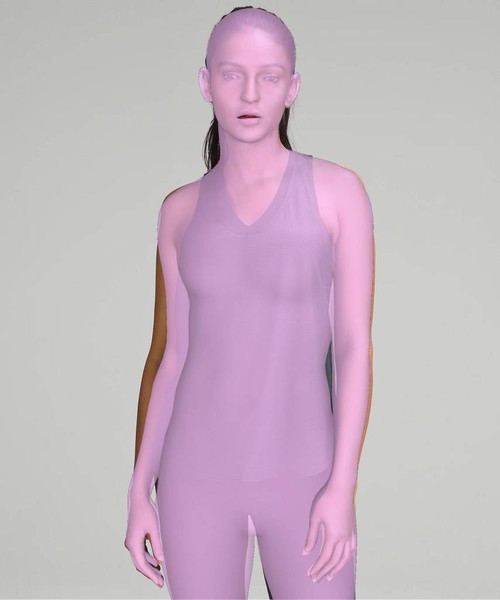}}\\

\subfloat[]{\includegraphics[height=0.24\textheight]{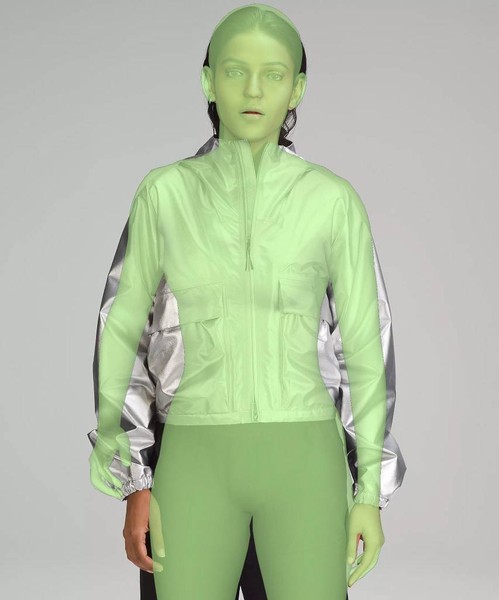}}
\subfloat[]{\includegraphics[height=0.24\textheight]{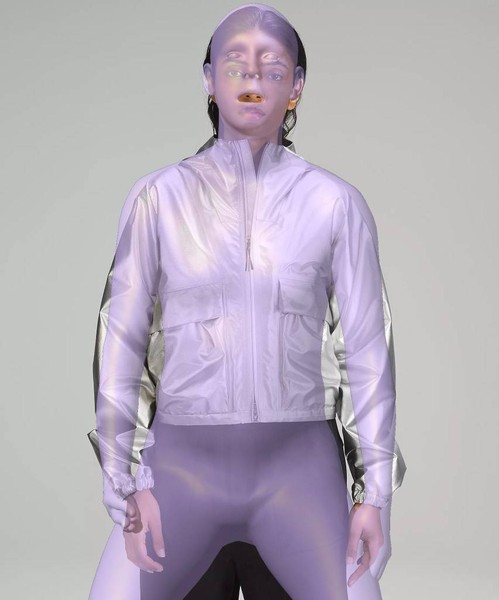}}
\subfloat[]{\includegraphics[height=0.24\textheight]{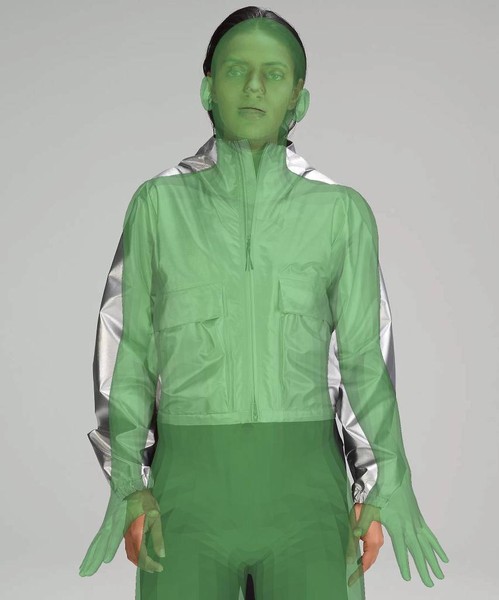}}
\subfloat[]{\includegraphics[height=0.24\textheight]{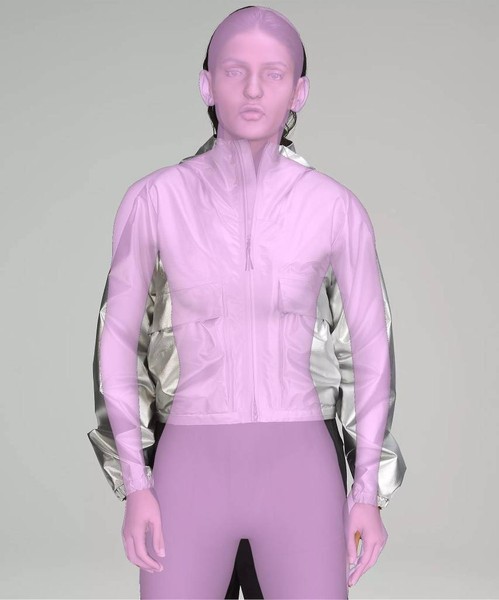}}\\

\vspace{-10pt}

\subfloat[SMPLify-X \cite{pavlakos2019expressive}]
{\includegraphics[height=0.24\textheight]{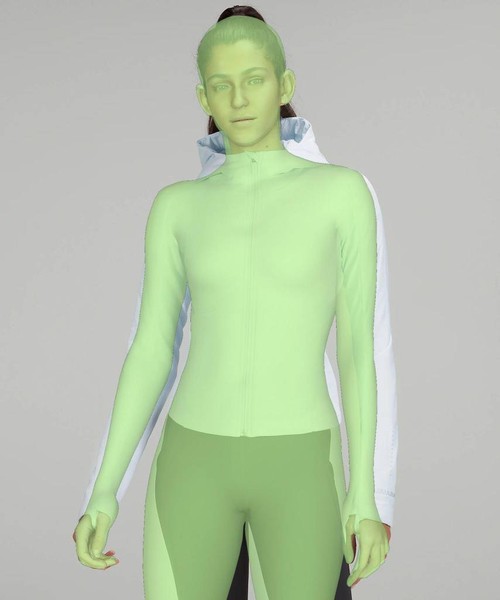}}
\subfloat[PyMAF-X \cite{pymafx2022}]{\includegraphics[height=0.24\textheight]{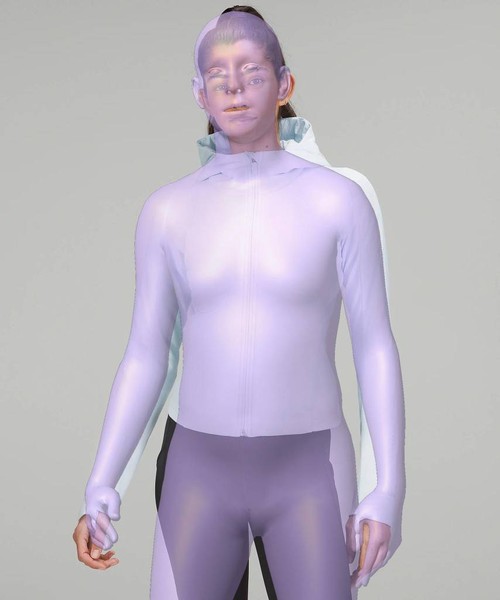}}
\subfloat[SHAPY \cite{choutas2022accurate}]
{\includegraphics[height=0.24\textheight]{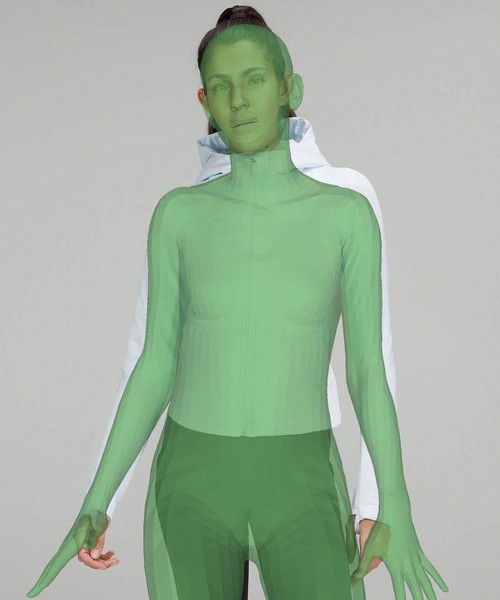}}
\subfloat[\KBody{-.1}{.035} (Ours)]{\includegraphics[height=0.24\textheight]{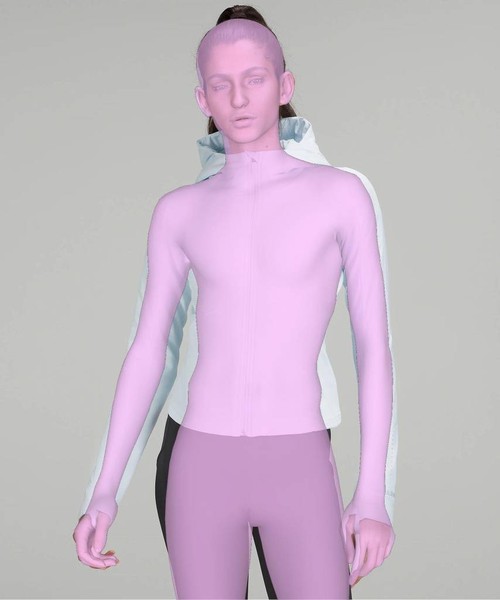}}

\caption{
Left-to-right: SMPLify-X \cite{pavlakos2019expressive} (\textcolor{caribbeangreen2}{light green}), PyMAF-X \cite{pymafx2022} (\textcolor{violet}{purple}), SHAPY \cite{choutas2022accurate} (\textcolor{jade}{green}) and KBody (\textcolor{candypink}{pink}).
}
\label{fig:partial_ll1}
\end{figure*}

%% file: figures/supp/partial_lululemone2.tex
\begin{figure*}[!htbp]
\captionsetup[subfigure]{position=bottom,labelformat=empty}

\centering

\subfloat{\includegraphics[height=0.24\textheight]{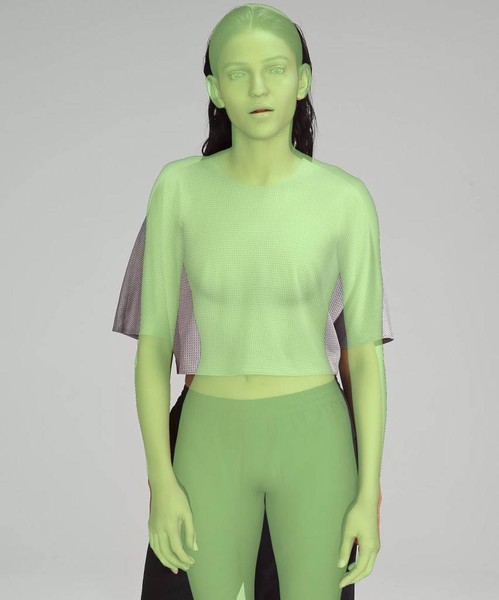}}
\subfloat{\includegraphics[height=0.24\textheight]{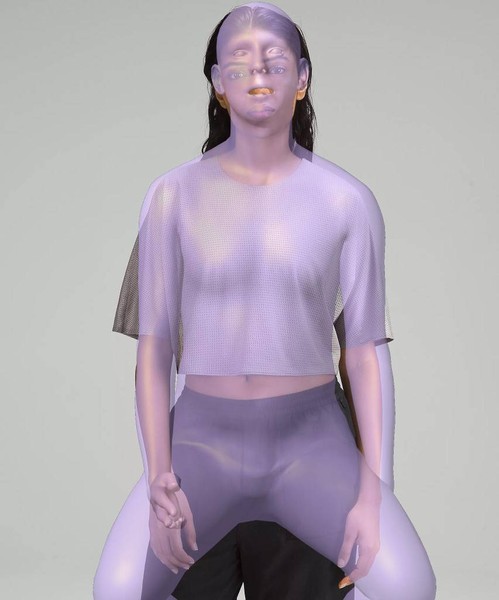}}
\subfloat{\includegraphics[height=0.24\textheight]{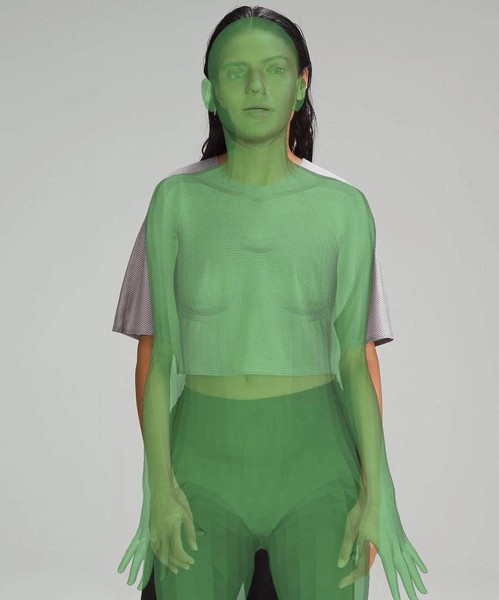}}
\subfloat{\includegraphics[height=0.24\textheight]{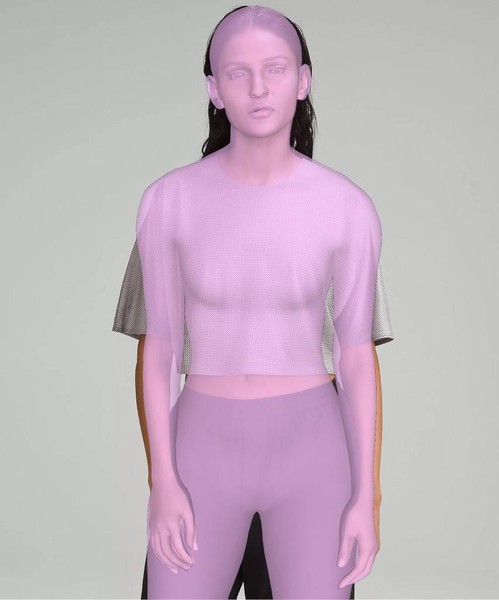}}\\

\subfloat{\includegraphics[height=0.24\textheight]{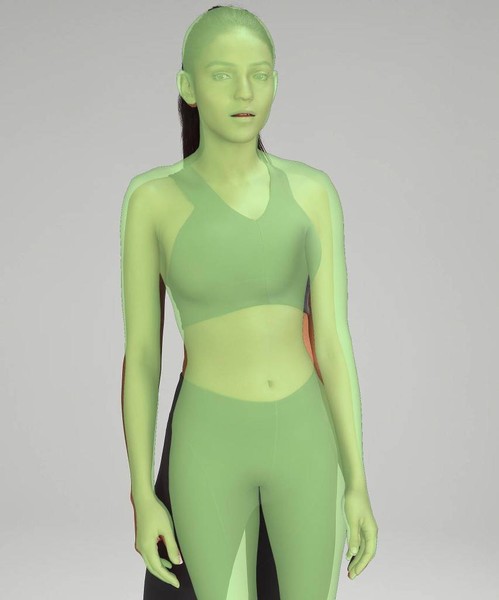}}
\subfloat{\includegraphics[height=0.24\textheight]{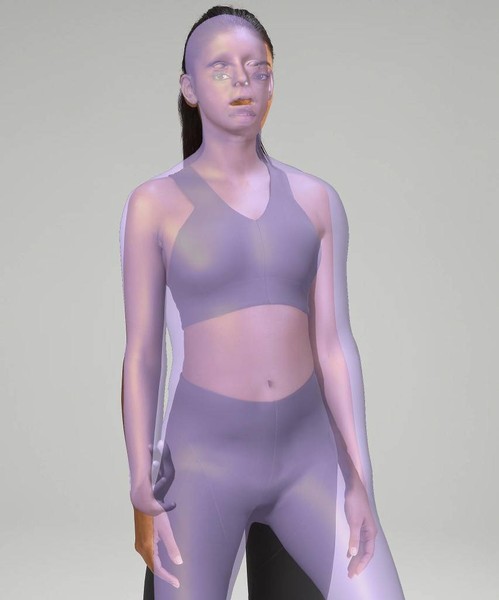}}
\subfloat{\includegraphics[height=0.24\textheight]{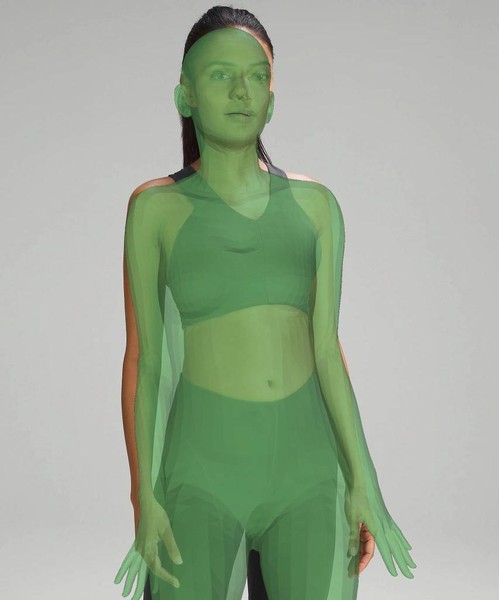}}
\subfloat{\includegraphics[height=0.24\textheight]{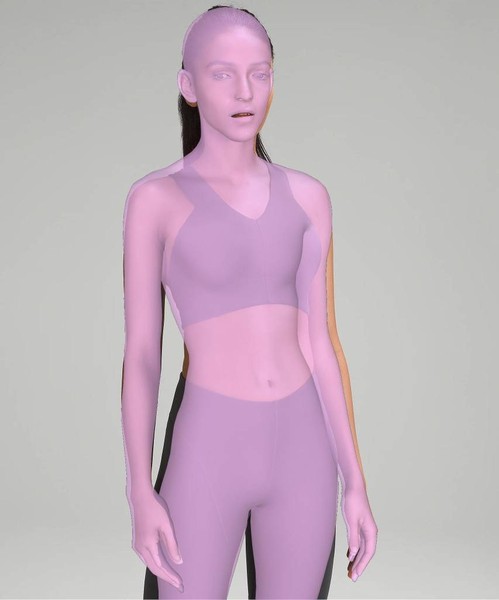}}\\

\subfloat[]{\includegraphics[height=0.24\textheight]{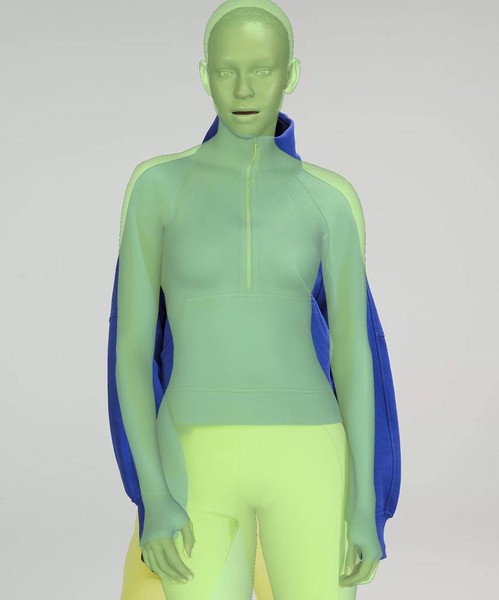}}
\subfloat[]{\includegraphics[height=0.24\textheight]{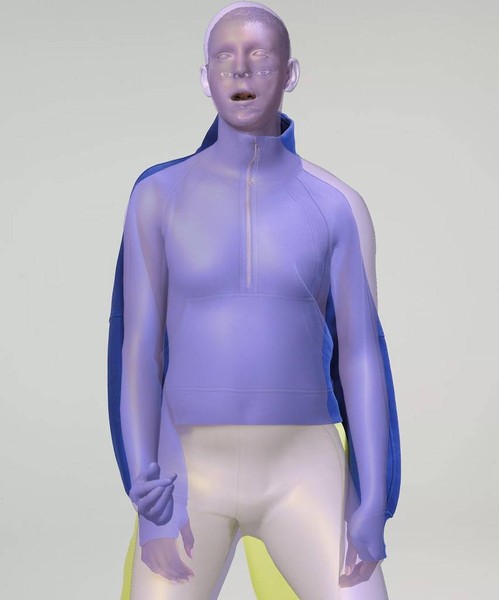}}
\subfloat[]{\includegraphics[height=0.24\textheight]{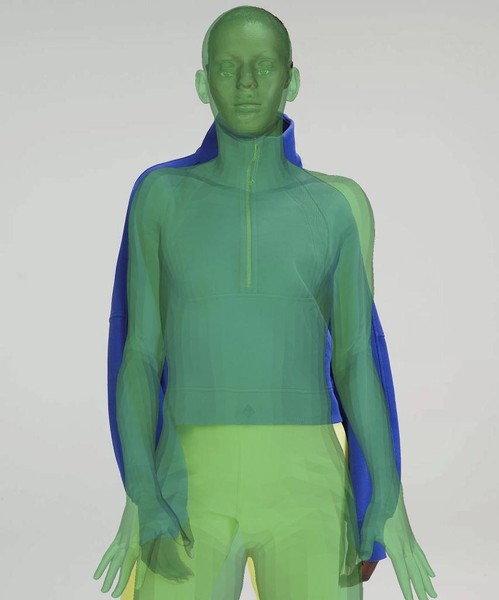}}
\subfloat[]{\includegraphics[height=0.24\textheight]{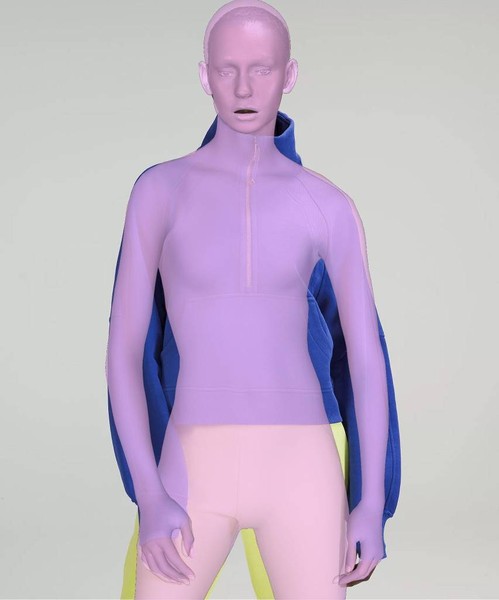}}\\

\vspace{-10pt}

\subfloat[SMPLify-X \cite{pavlakos2019expressive}]
{\includegraphics[height=0.24\textheight]{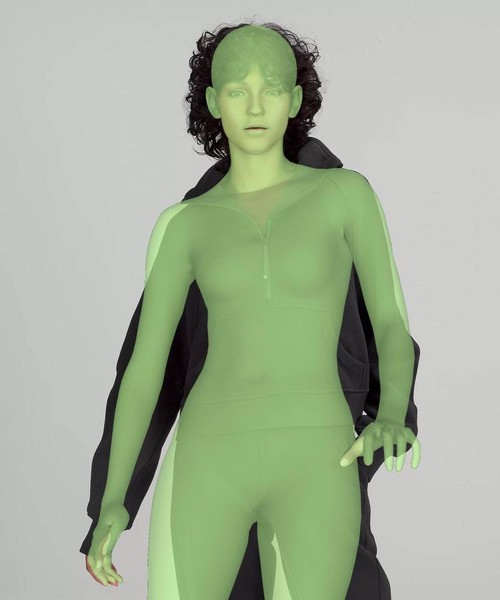}}
\subfloat[PyMAF-X \cite{pymafx2022}]{\includegraphics[height=0.24\textheight]{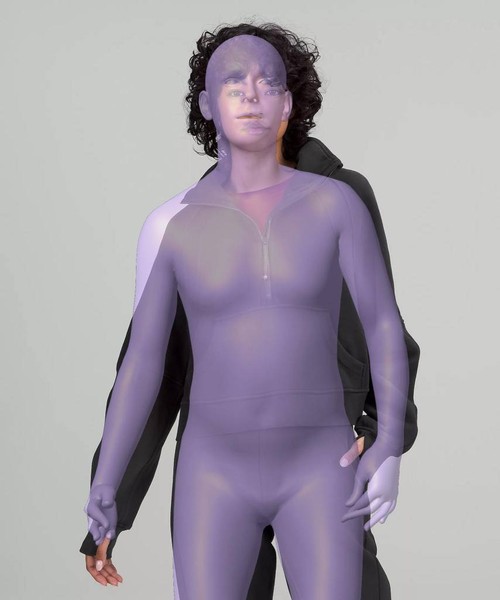}}
\subfloat[SHAPY \cite{choutas2022accurate}]
{\includegraphics[height=0.24\textheight]{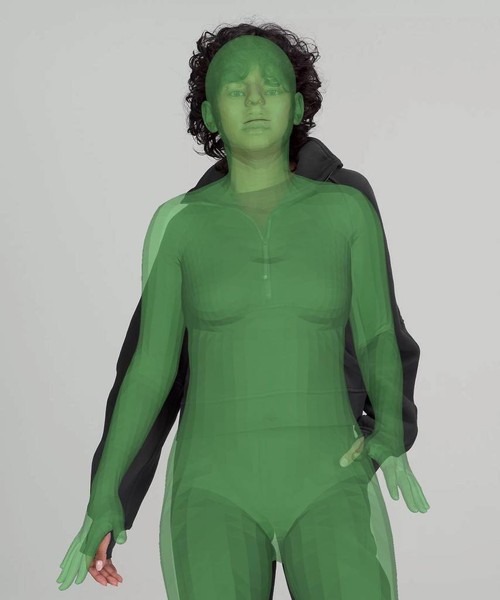}}
\subfloat[\KBody{-.1}{.035} (Ours)]{\includegraphics[height=0.24\textheight]{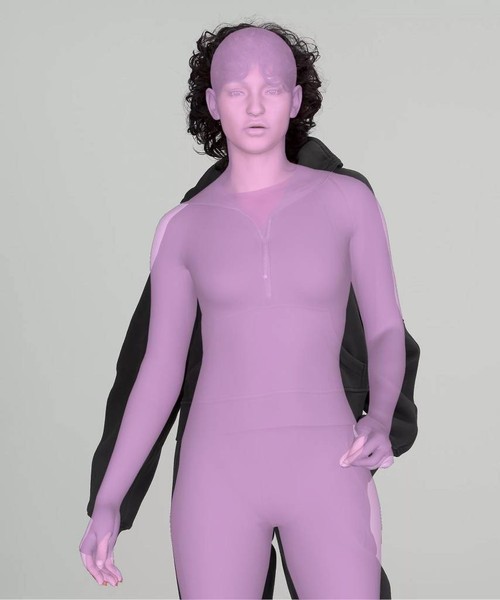}}

\caption{
Left-to-right: SMPLify-X \cite{pavlakos2019expressive} (\textcolor{caribbeangreen2}{light green}), PyMAF-X \cite{pymafx2022} (\textcolor{violet}{purple}), SHAPY \cite{choutas2022accurate} (\textcolor{jade}{green}) and KBody (\textcolor{candypink}{pink}).
}
\label{fig:partial_ll2}
\end{figure*}

%% file: figures/supp/partial_splendid1.tex
\begin{figure*}[!htbp]
\captionsetup[subfigure]{position=bottom,labelformat=empty}

\centering

\subfloat{\includegraphics[height=0.24\textheight]{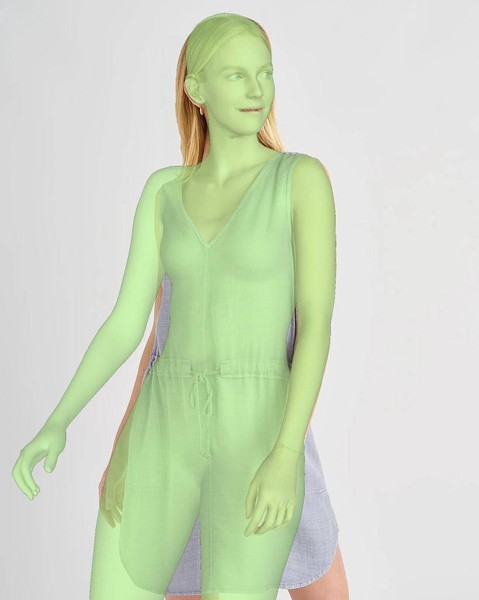}}
\subfloat{\includegraphics[height=0.24\textheight]{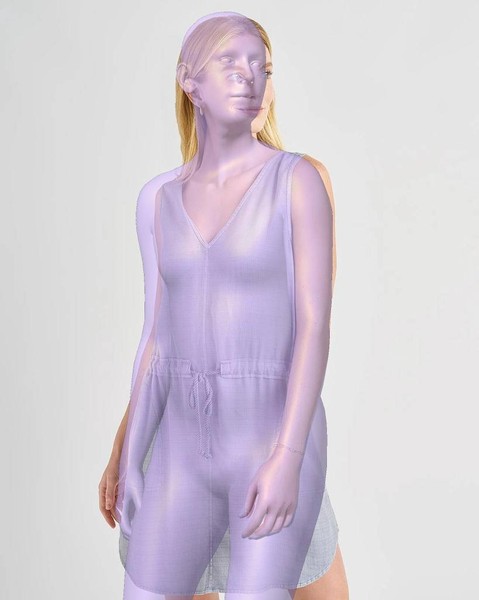}}
\subfloat{\includegraphics[height=0.24\textheight]{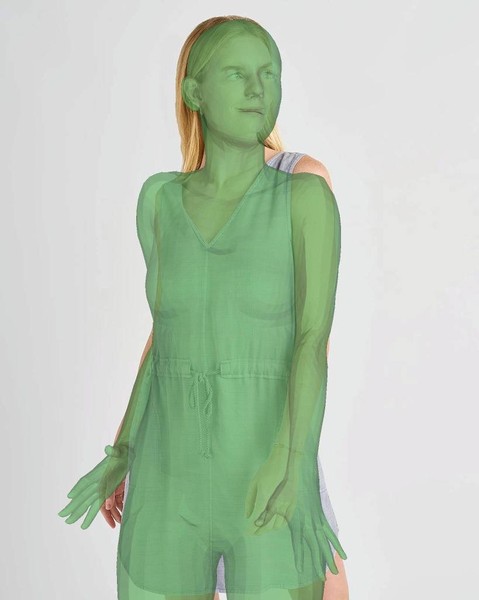}}
\subfloat{\includegraphics[height=0.24\textheight]{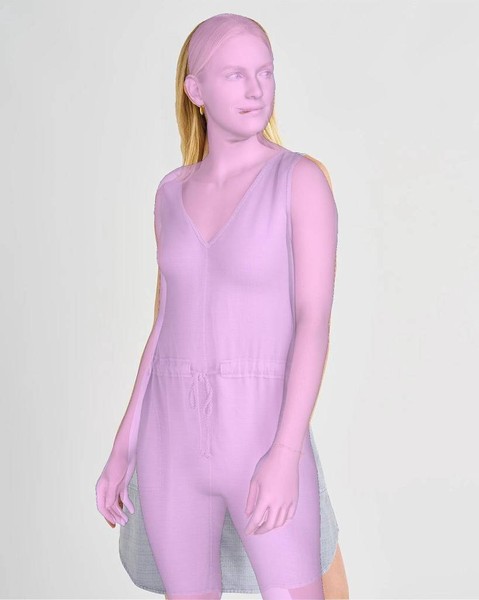}}\\

\subfloat{\includegraphics[height=0.24\textheight]{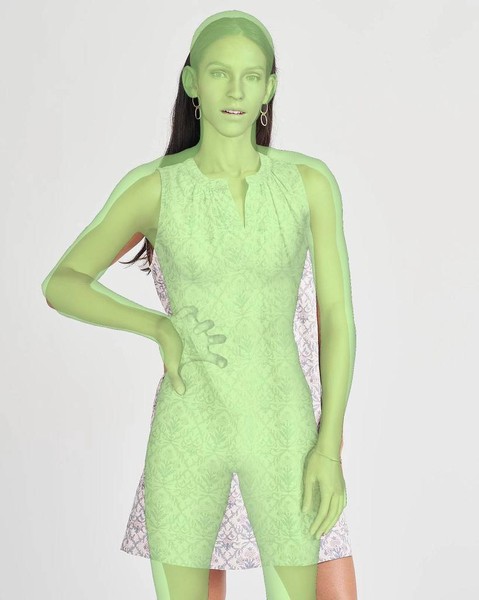}}
\subfloat{\includegraphics[height=0.24\textheight]{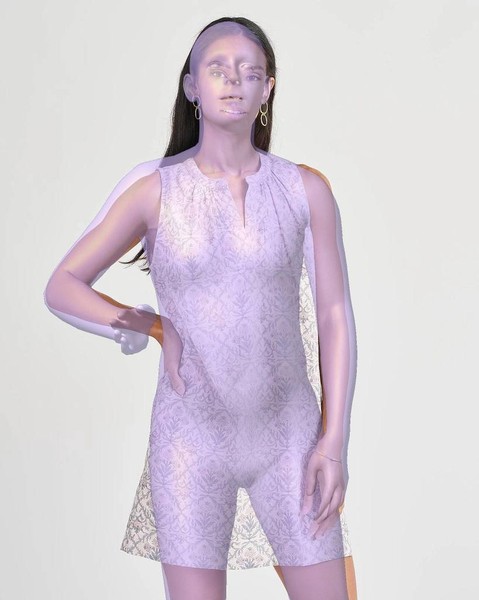}}
\subfloat{\includegraphics[height=0.24\textheight]{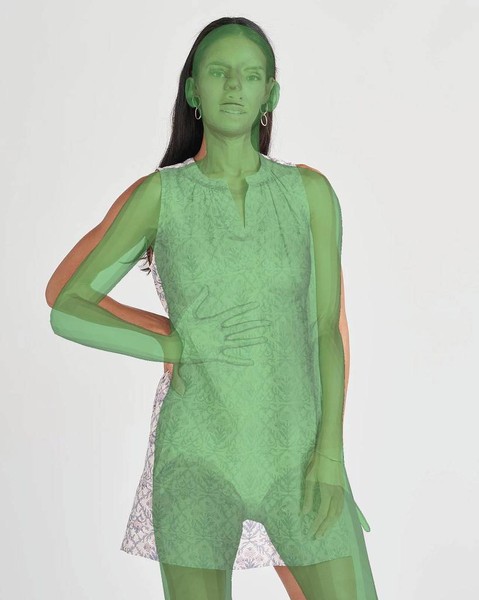}}
\subfloat{\includegraphics[height=0.24\textheight]{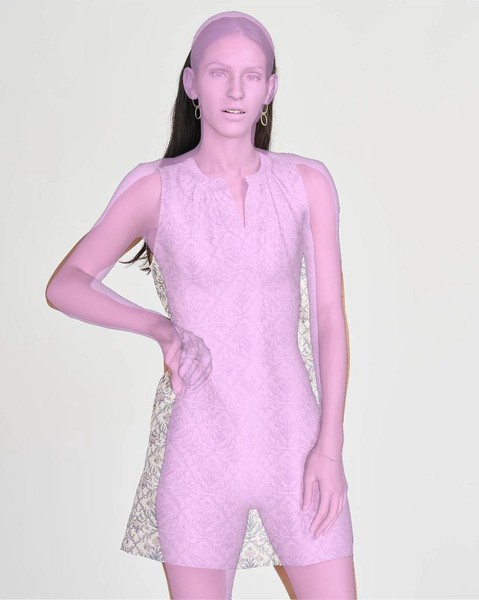}}\\

\subfloat[]{\includegraphics[height=0.24\textheight]{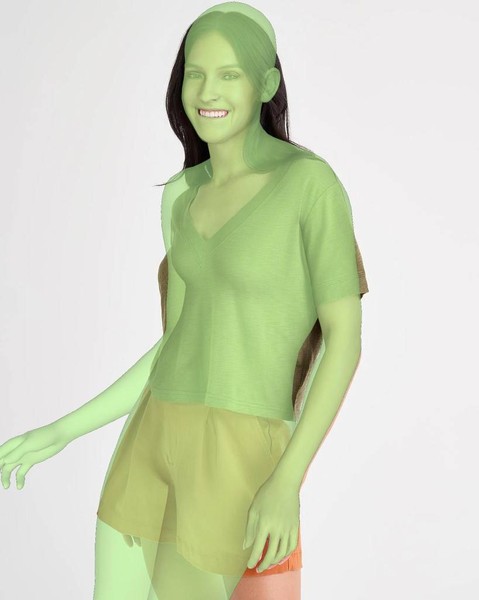}}
\subfloat[]{\includegraphics[height=0.24\textheight]{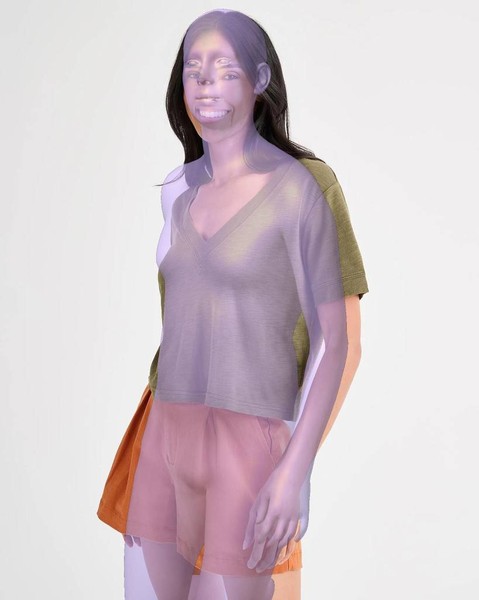}}
\subfloat[]{\includegraphics[height=0.24\textheight]{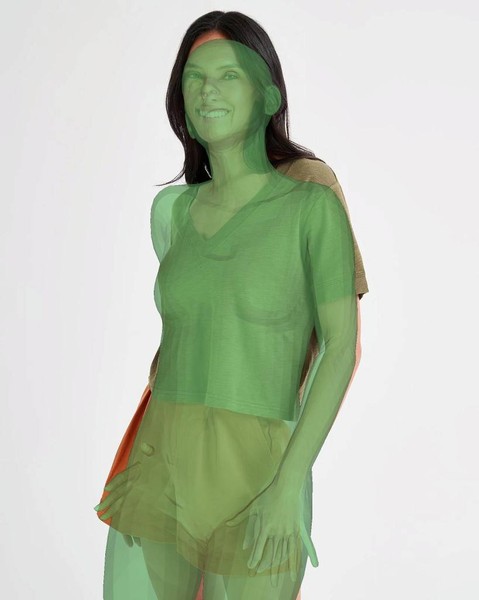}}
\subfloat[]{\includegraphics[height=0.24\textheight]{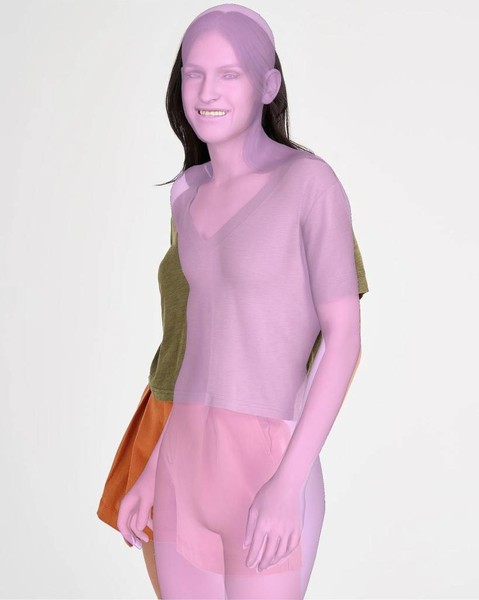}}\\

\vspace{-10pt}

\caption{
Left-to-right: SMPLify-X \cite{pavlakos2019expressive} (\textcolor{caribbeangreen2}{light green}), PyMAF-X \cite{pymafx2022} (\textcolor{violet}{purple}), SHAPY \cite{choutas2022accurate} (\textcolor{jade}{green}) and KBody (\textcolor{candypink}{pink}).
}
\label{fig:partial_sp1}
\end{figure*}

%% file: figures/supp/partial_splendid2.tex
\begin{figure*}[!htbp]
\captionsetup[subfigure]{position=bottom,labelformat=empty}

\centering

\subfloat{\includegraphics[height=0.24\textheight]{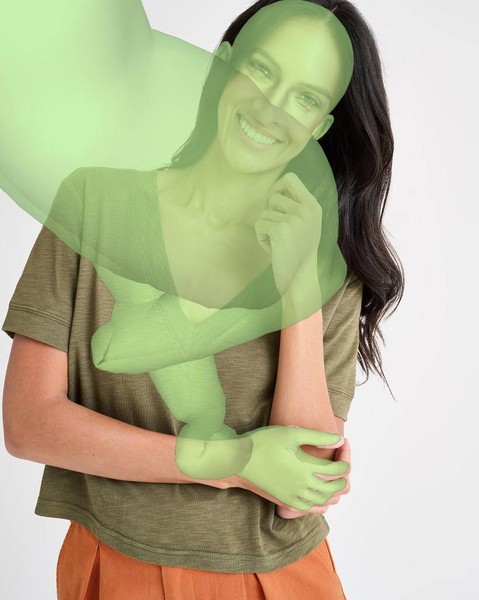}}
\subfloat{\includegraphics[height=0.24\textheight]{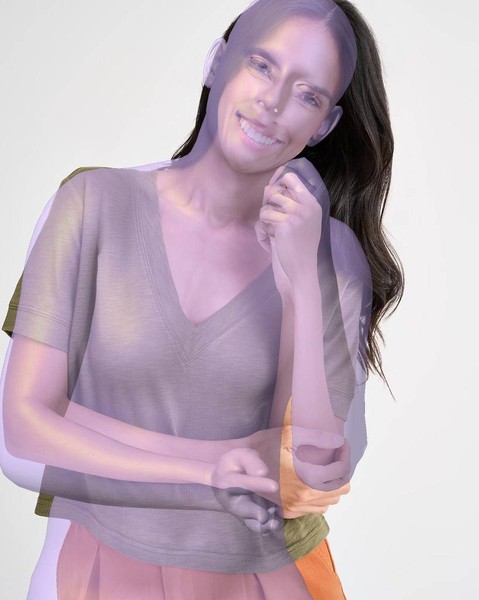}}
\subfloat{\includegraphics[height=0.24\textheight]{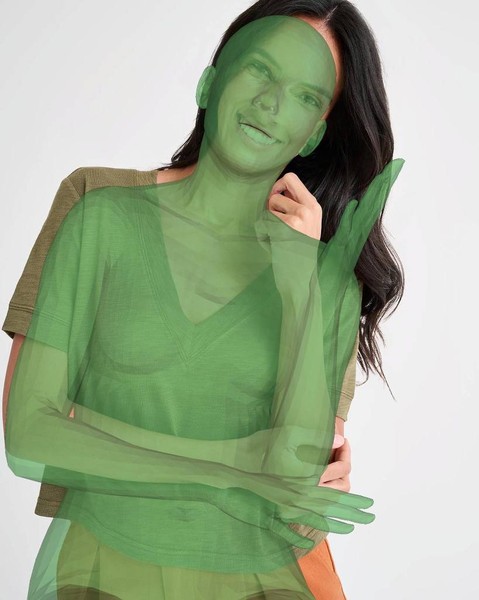}}
\subfloat{\includegraphics[height=0.24\textheight]{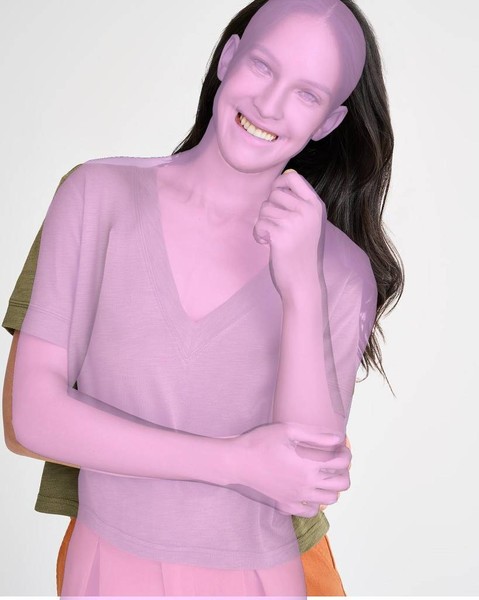}}\\

\subfloat{\includegraphics[height=0.24\textheight]{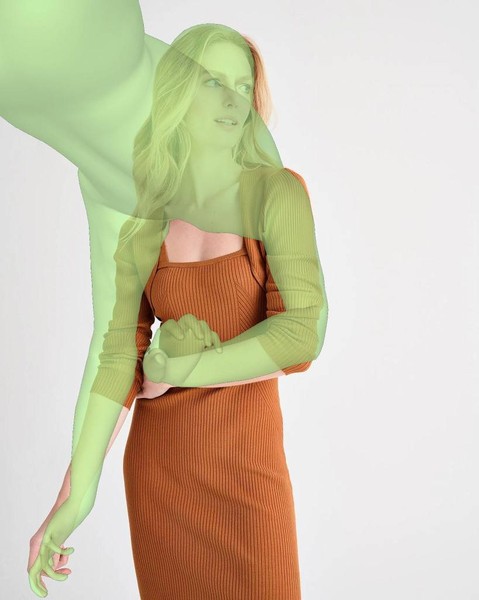}}
\subfloat{\includegraphics[height=0.24\textheight]{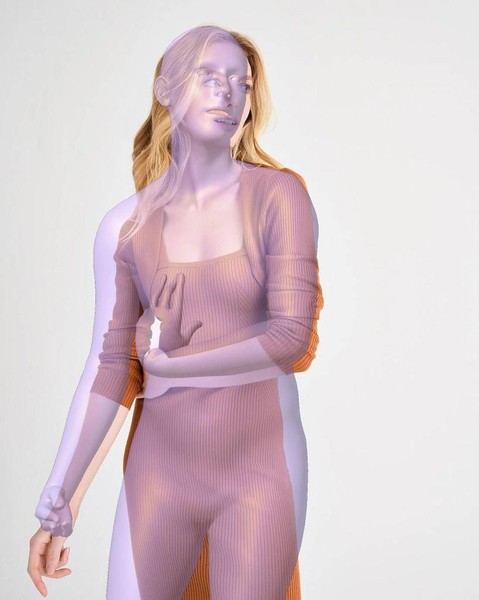}}
\subfloat{\includegraphics[height=0.24\textheight]{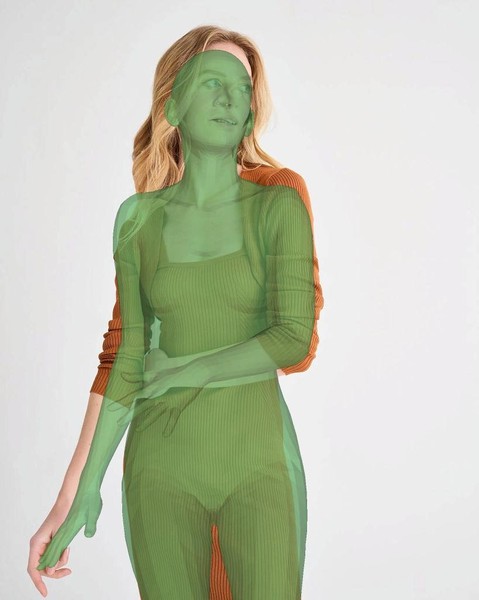}}
\subfloat{\includegraphics[height=0.24\textheight]{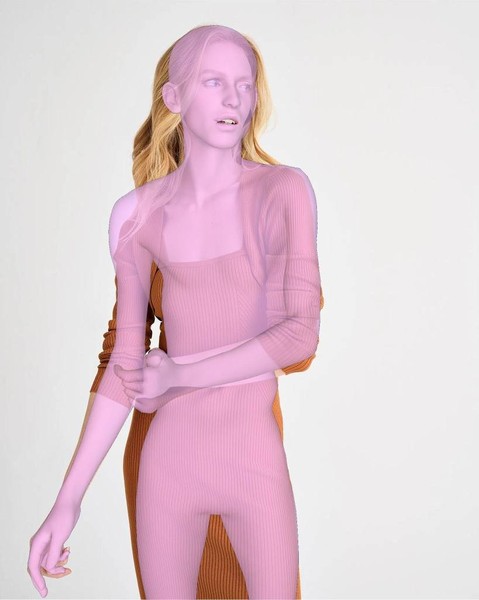}}\\

\subfloat[]{\includegraphics[height=0.24\textheight]{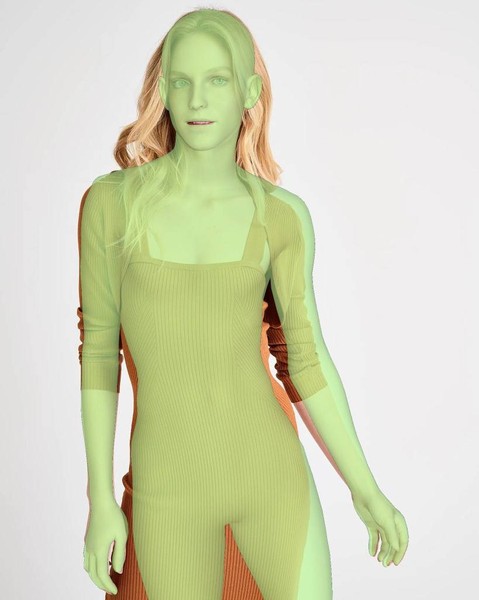}}
\subfloat[]{\includegraphics[height=0.24\textheight]{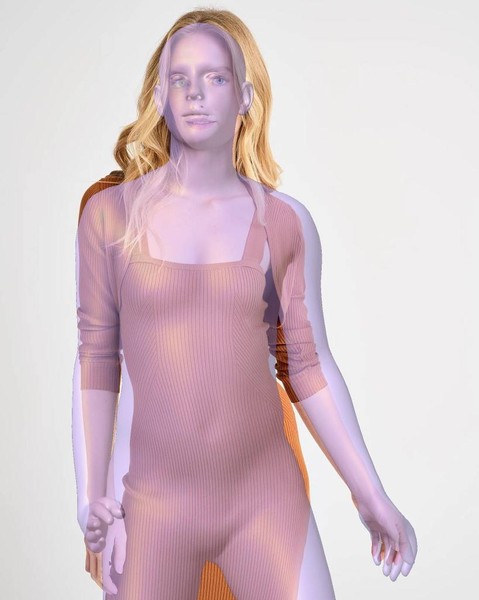}}
\subfloat[]{\includegraphics[height=0.24\textheight]{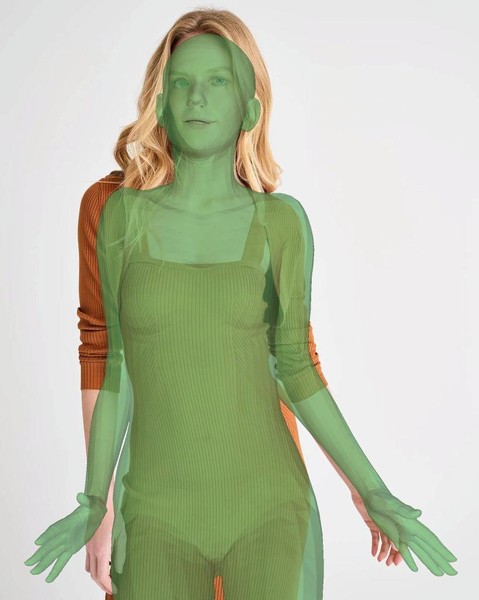}}
\subfloat[]{\includegraphics[height=0.24\textheight]{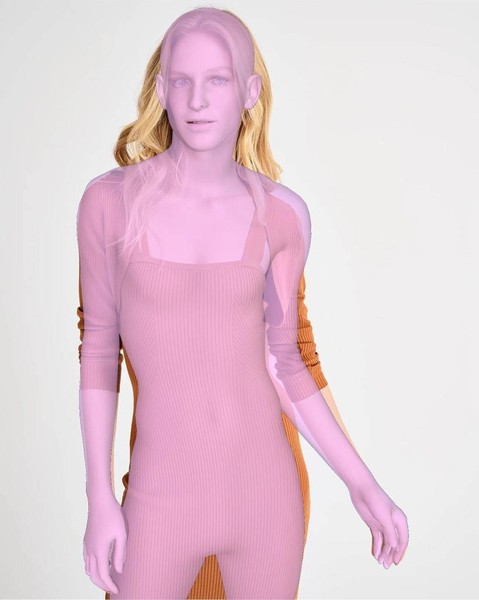}}\\

\vspace{-10pt}

\subfloat[SMPLify-X \cite{pavlakos2019expressive}]
{\includegraphics[height=0.24\textheight]{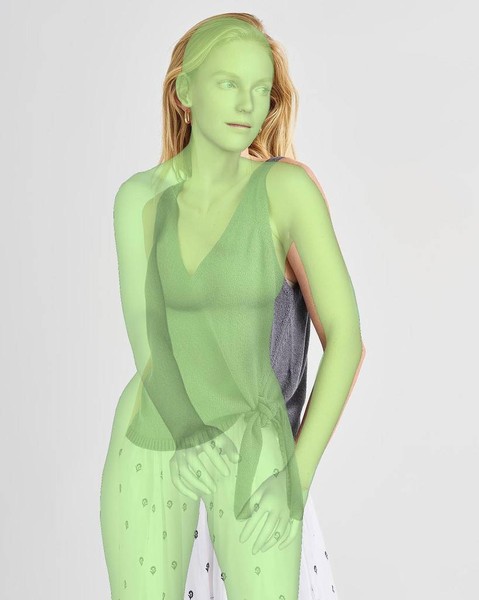}}
\subfloat[PyMAF-X \cite{pymafx2022}]{\includegraphics[height=0.24\textheight]{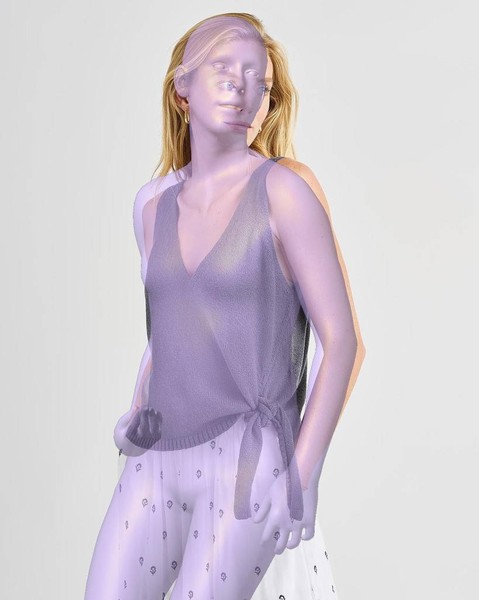}}
\subfloat[SHAPY \cite{choutas2022accurate}]
{\includegraphics[height=0.24\textheight]{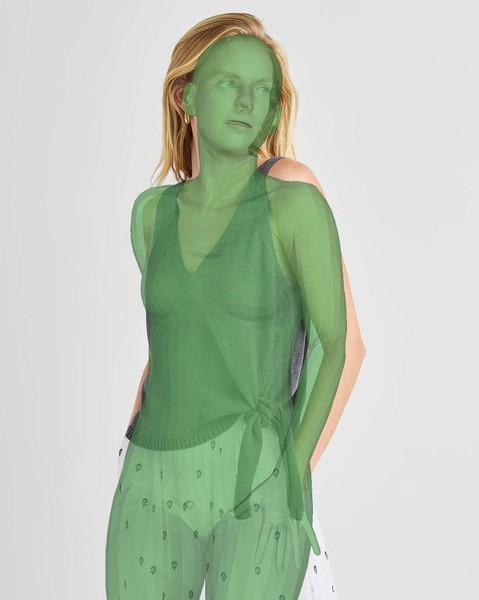}}
\subfloat[\KBody{-.1}{.035} (Ours)]{\includegraphics[height=0.24\textheight]{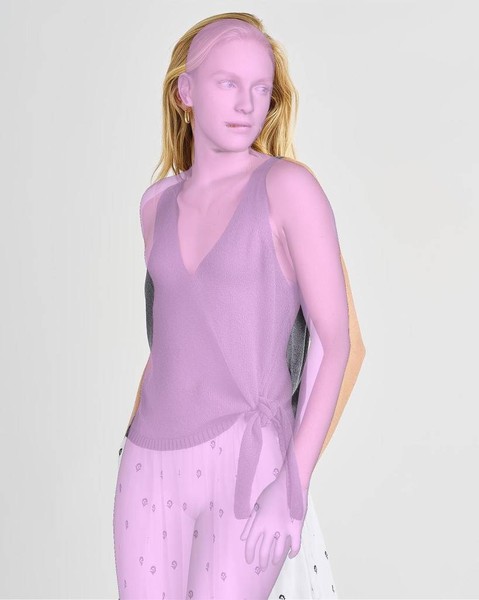}}

\caption{
Left-to-right: SMPLify-X \cite{pavlakos2019expressive} (\textcolor{caribbeangreen2}{light green}), PyMAF-X \cite{pymafx2022} (\textcolor{violet}{purple}), SHAPY \cite{choutas2022accurate} (\textcolor{jade}{green}) and KBody (\textcolor{candypink}{pink}).
}
\label{fig:partial_sp2}
\end{figure*}

%% file: figures/supp/partial_splendid3.tex
\begin{figure*}[!htbp]
\captionsetup[subfigure]{position=bottom,labelformat=empty}

\centering

\subfloat{\includegraphics[height=0.24\textheight]{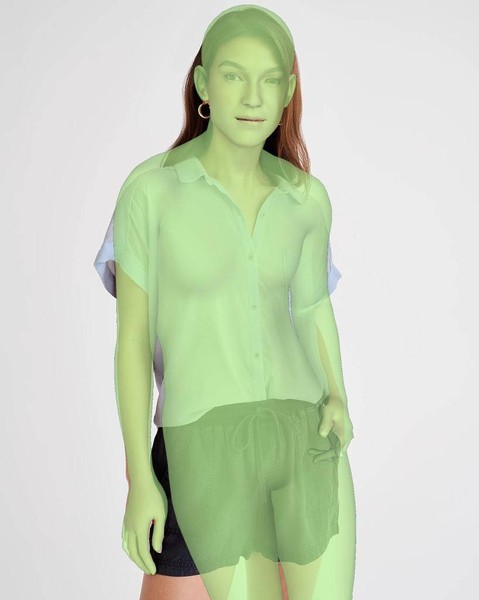}}
\subfloat{\includegraphics[height=0.24\textheight]{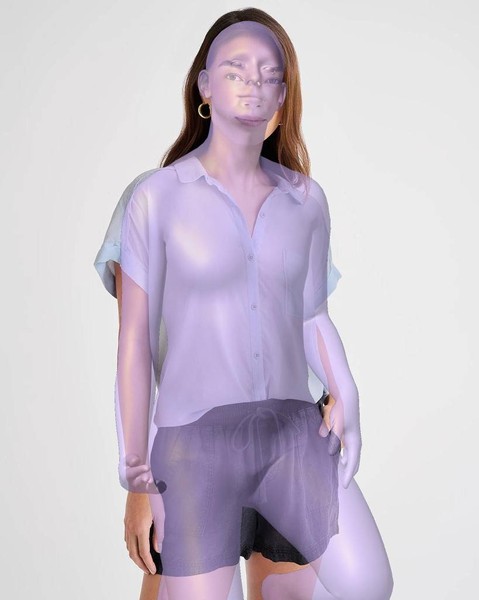}}
\subfloat{\includegraphics[height=0.24\textheight]{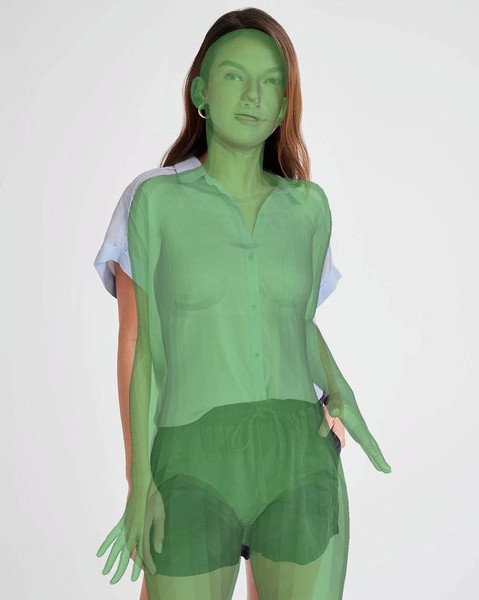}}
\subfloat{\includegraphics[height=0.24\textheight]{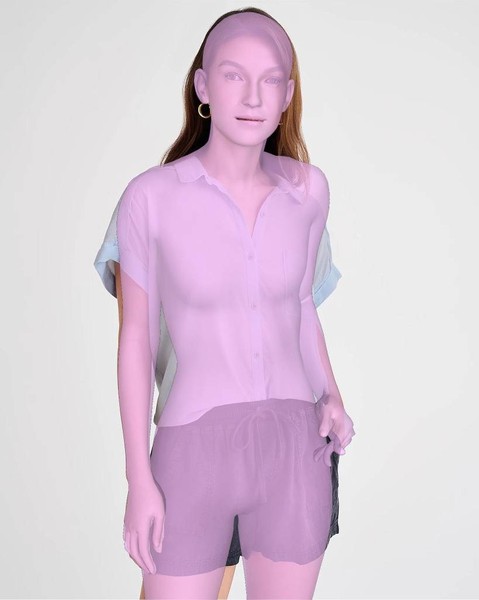}}\\

\subfloat{\includegraphics[height=0.24\textheight]{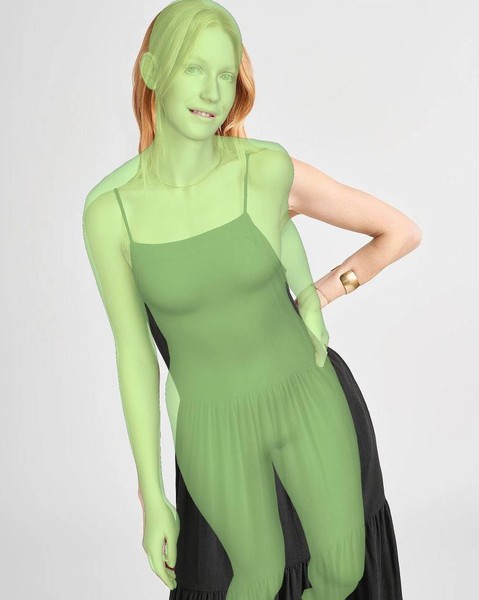}}
\subfloat{\includegraphics[height=0.24\textheight]{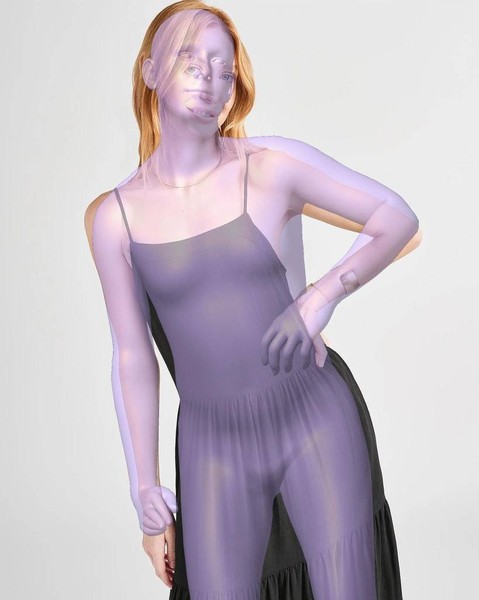}}
\subfloat{\includegraphics[height=0.24\textheight]{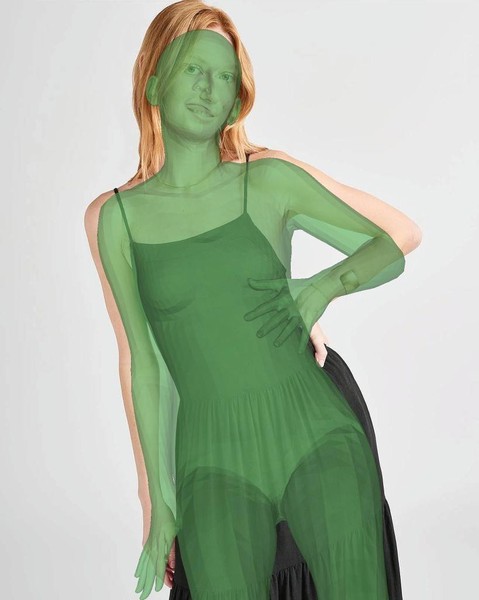}}
\subfloat{\includegraphics[height=0.24\textheight]{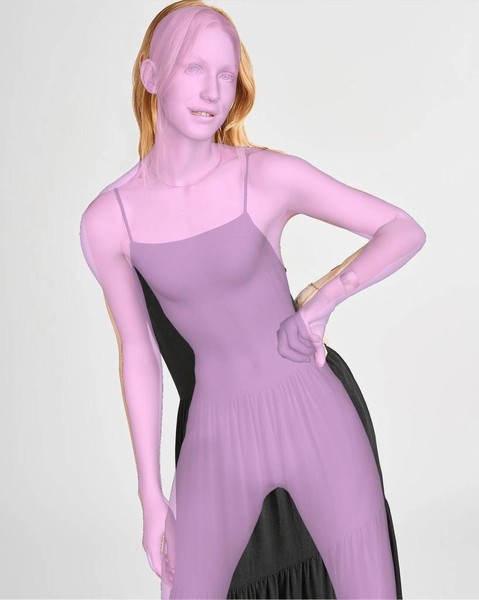}}\\

\subfloat[]{\includegraphics[height=0.24\textheight]{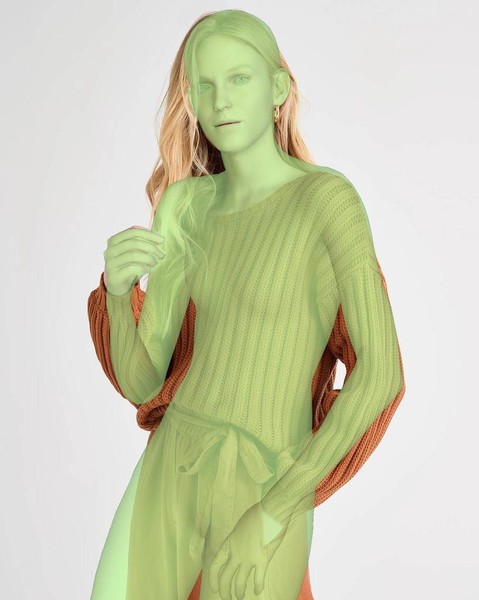}}
\subfloat[]{\includegraphics[height=0.24\textheight]{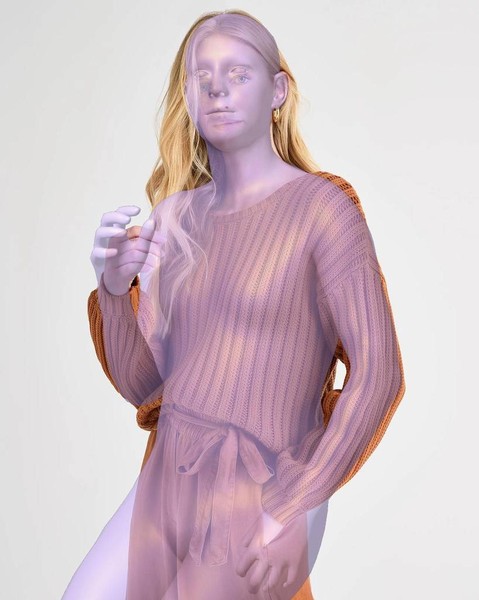}}
\subfloat[]{\includegraphics[height=0.24\textheight]{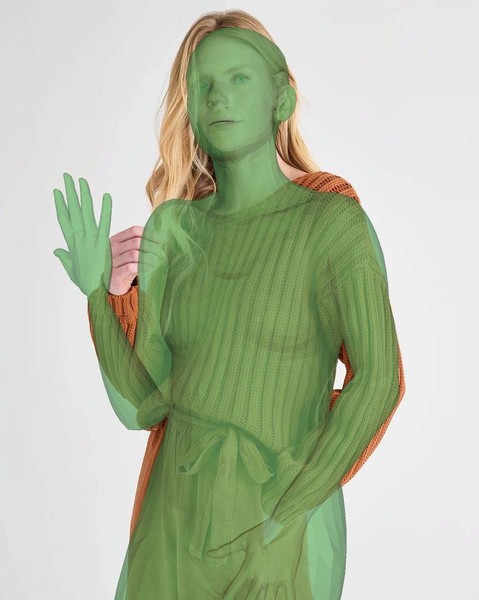}}
\subfloat[]{\includegraphics[height=0.24\textheight]{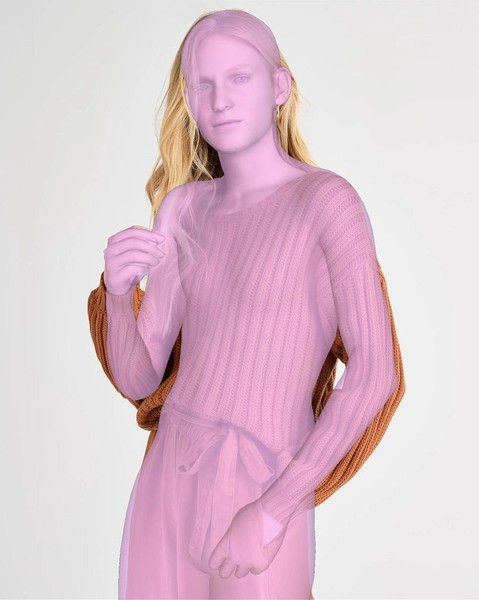}}\\

\vspace{-10pt}

\subfloat[SMPLify-X \cite{pavlakos2019expressive}]
{\includegraphics[height=0.24\textheight]{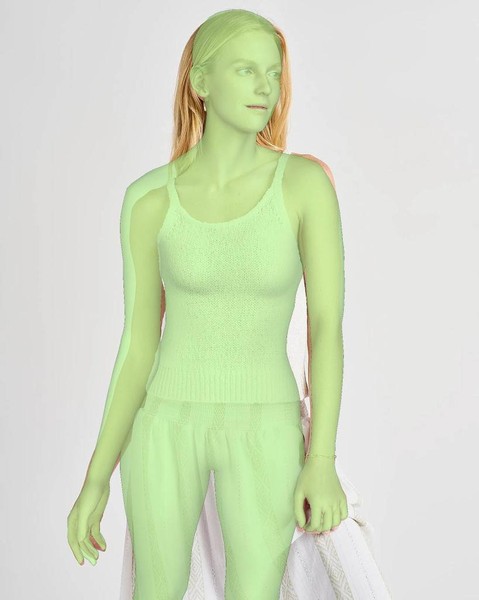}}
\subfloat[PyMAF-X \cite{pymafx2022}]{\includegraphics[height=0.24\textheight]{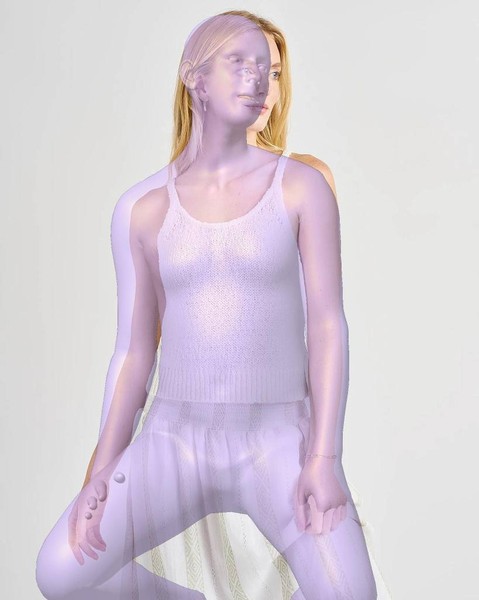}}
\subfloat[SHAPY \cite{choutas2022accurate}]
{\includegraphics[height=0.24\textheight]{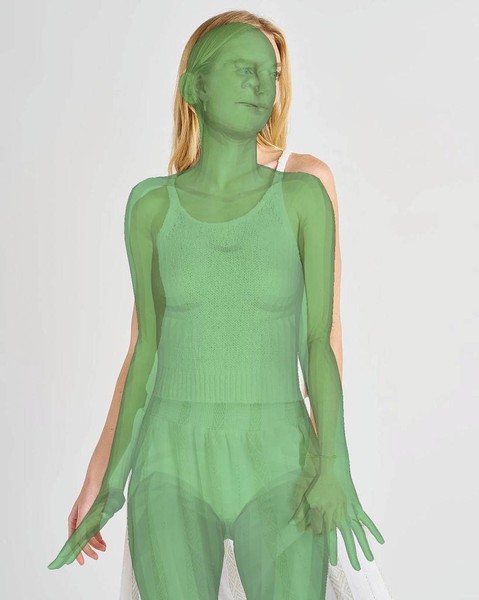}}
\subfloat[\KBody{-.1}{.035} (Ours)]{\includegraphics[height=0.24\textheight]{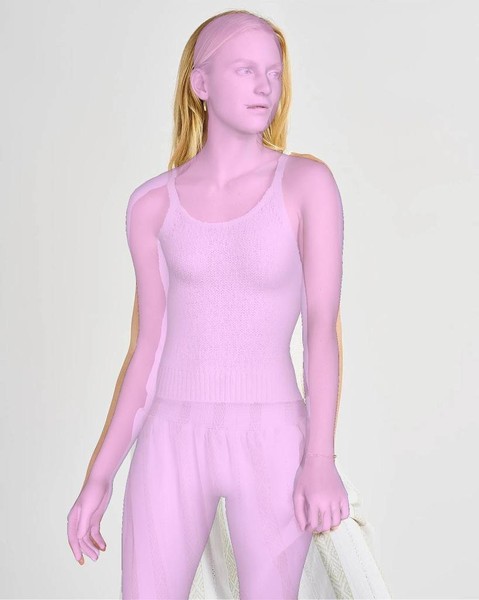}}

\caption{
Left-to-right: SMPLify-X \cite{pavlakos2019expressive} (\textcolor{caribbeangreen2}{light green}), PyMAF-X \cite{pymafx2022} (\textcolor{violet}{purple}), SHAPY \cite{choutas2022accurate} (\textcolor{jade}{green}) and KBody (\textcolor{candypink}{pink}).
}
\label{fig:partial_sp3}
\end{figure*}

%% file: figures/supp/partial_splendid5.tex
\begin{figure*}[!htbp]
\captionsetup[subfigure]{position=bottom,labelformat=empty}

\centering

\subfloat{\includegraphics[height=0.24\textheight]{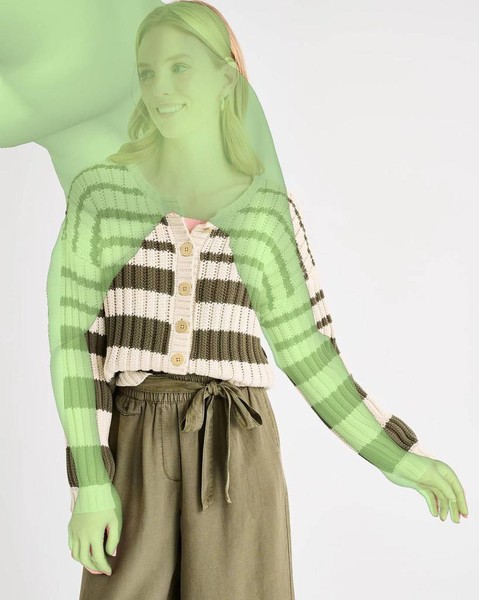}}
\subfloat{\includegraphics[height=0.24\textheight]{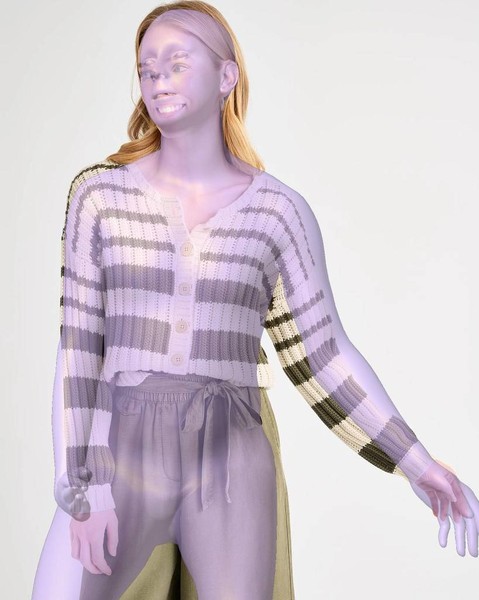}}
\subfloat{\includegraphics[height=0.24\textheight]{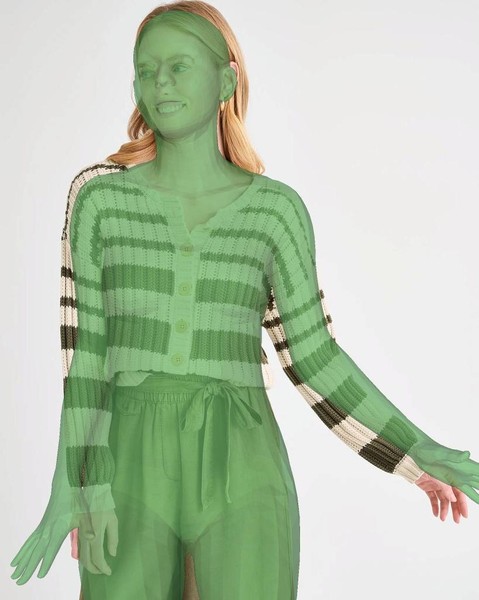}}
\subfloat{\includegraphics[height=0.24\textheight]{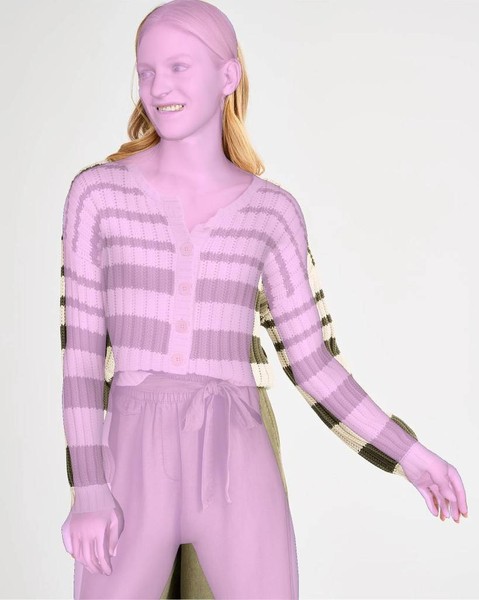}}\\

\subfloat[]{\includegraphics[height=0.24\textheight]{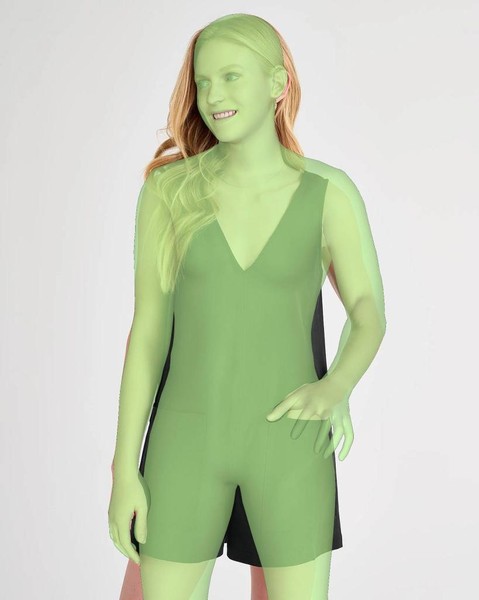}}
\subfloat[]{\includegraphics[height=0.24\textheight]{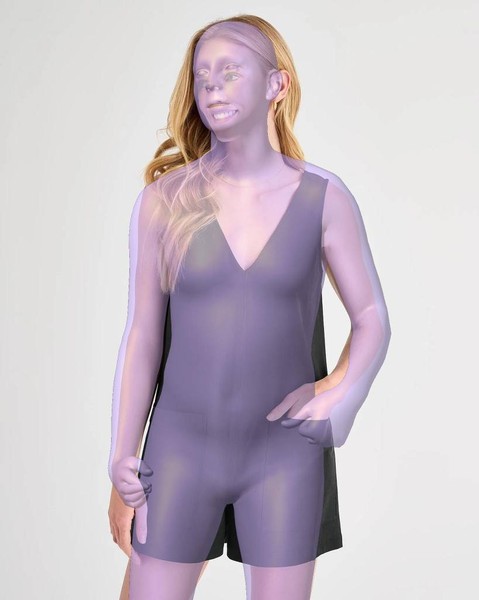}}
\subfloat[]{\includegraphics[height=0.24\textheight]{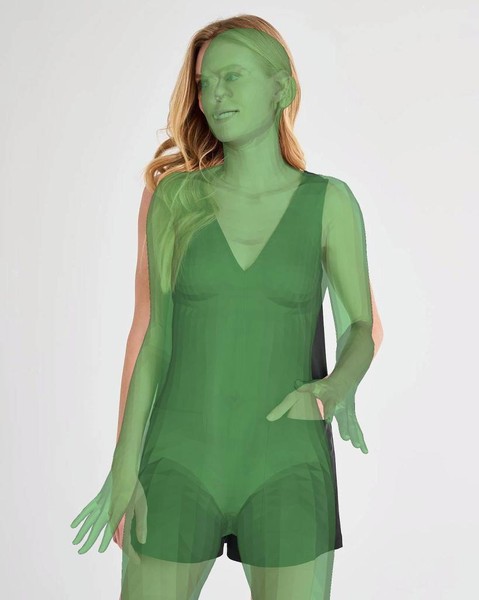}}
\subfloat[]{\includegraphics[height=0.24\textheight]{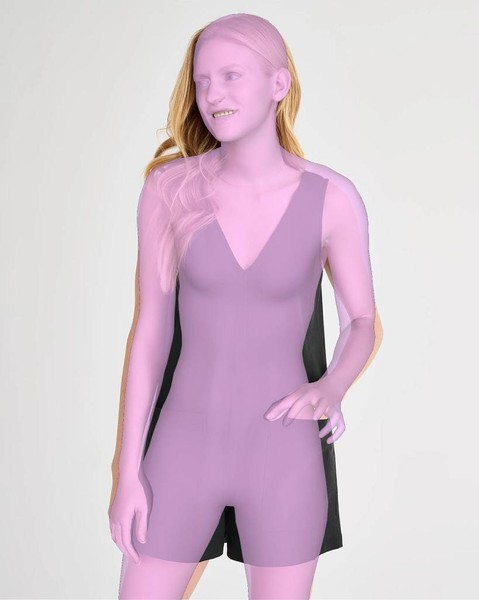}}\\

\vspace{-10pt}

\subfloat[]{\includegraphics[height=0.24\textheight]{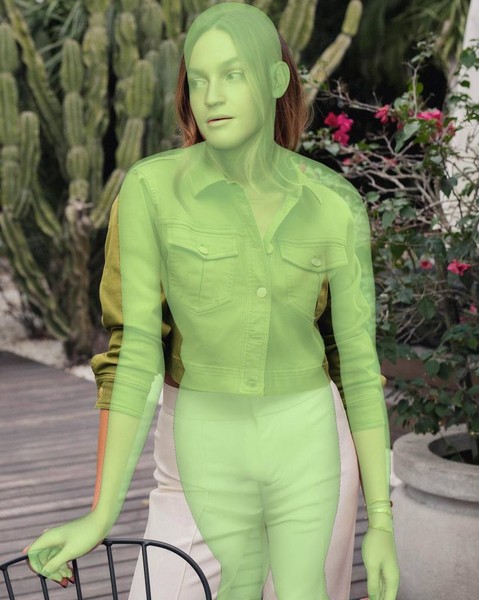}}
\subfloat[]{\includegraphics[height=0.24\textheight]{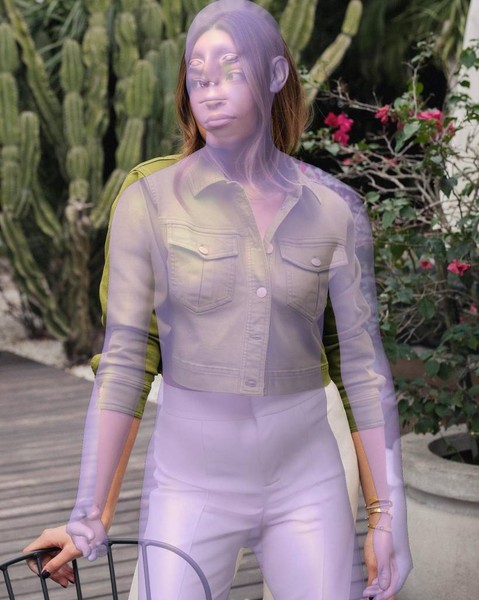}}
\subfloat[]{\includegraphics[height=0.24\textheight]{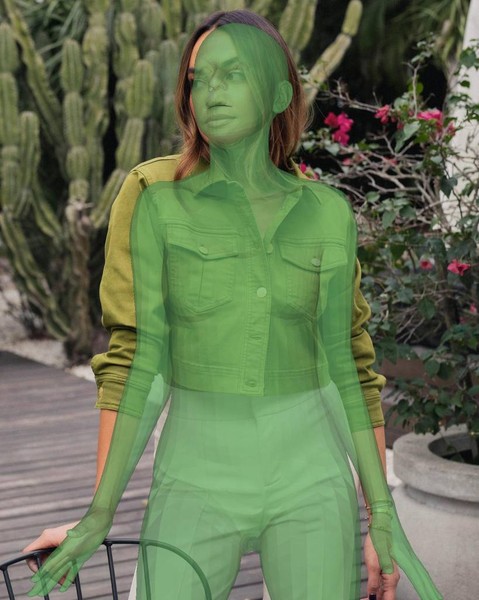}}
\subfloat[]{\includegraphics[height=0.24\textheight]{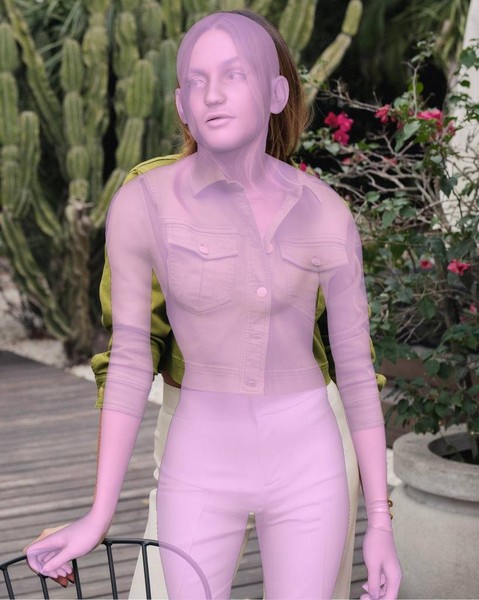}}\\

\vspace{-10pt}

\subfloat[SMPLify-X \cite{pavlakos2019expressive}]
{\includegraphics[height=0.24\textheight]{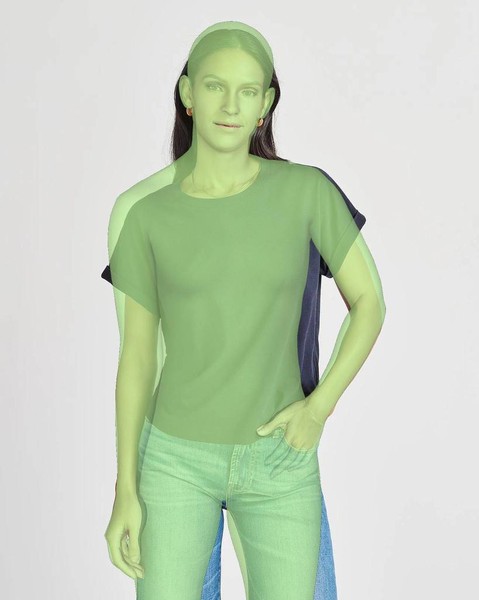}}
\subfloat[PyMAF-X \cite{pymafx2022}]{\includegraphics[height=0.24\textheight]{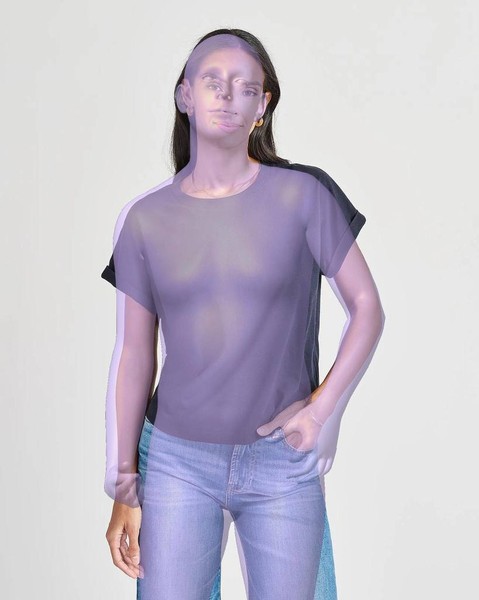}}
\subfloat[SHAPY \cite{choutas2022accurate}]
{\includegraphics[height=0.24\textheight]{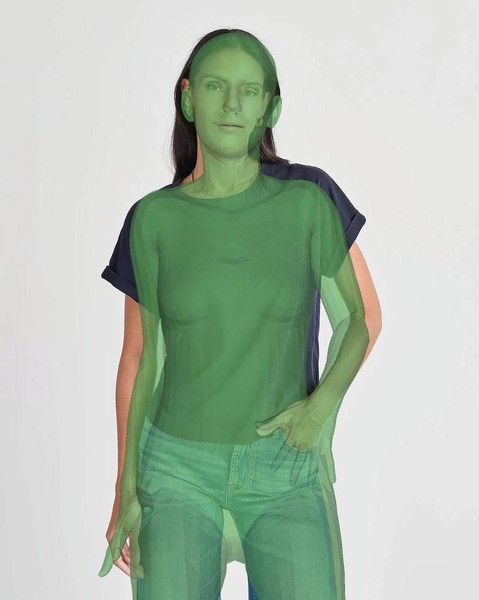}}
\subfloat[\KBody{-.1}{.035} (Ours)]{\includegraphics[height=0.24\textheight]{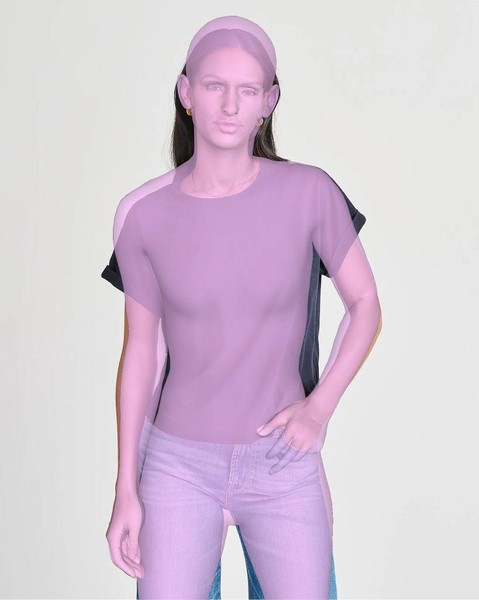}}

\caption{
Left-to-right: SMPLify-X \cite{pavlakos2019expressive} (\textcolor{caribbeangreen2}{light green}), PyMAF-X \cite{pymafx2022} (\textcolor{violet}{purple}), SHAPY \cite{choutas2022accurate} (\textcolor{jade}{green}) and KBody (\textcolor{candypink}{pink}).
}
\label{fig:partial_sp5}
\end{figure*}

%% file: figures/supp/partial_splendid7.tex
\begin{figure*}[!htbp]
\captionsetup[subfigure]{position=bottom,labelformat=empty}

\centering

\subfloat{\includegraphics[height=0.24\textheight]{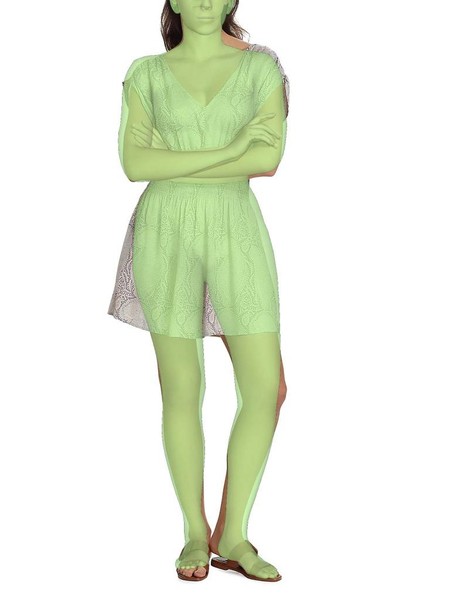}}
\subfloat{\includegraphics[height=0.24\textheight]{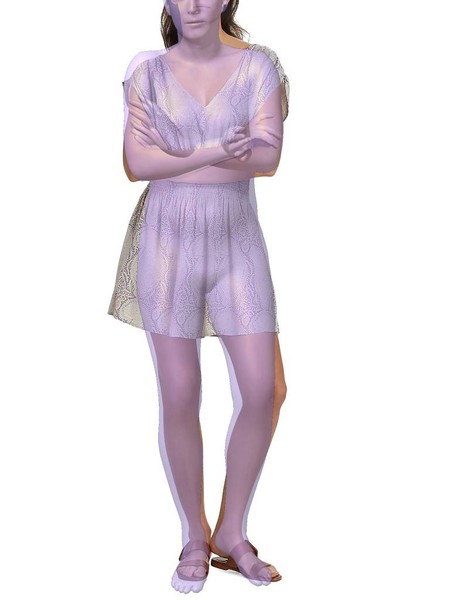}}
\subfloat{\includegraphics[height=0.24\textheight]{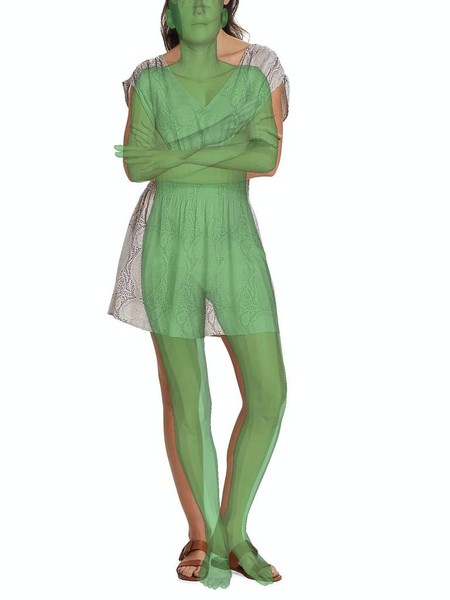}}
\subfloat{\includegraphics[height=0.24\textheight]{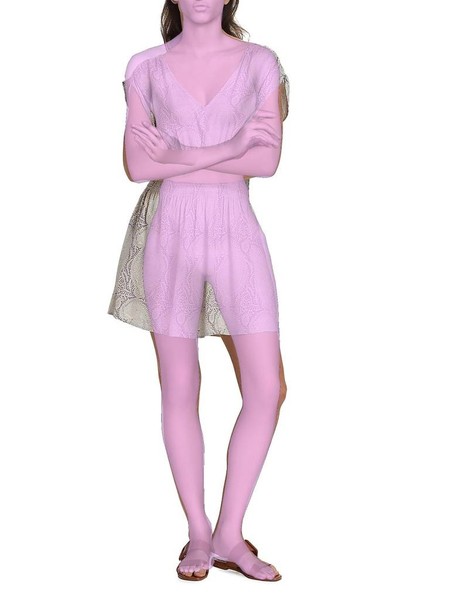}}\\

\subfloat[]{\includegraphics[height=0.24\textheight]{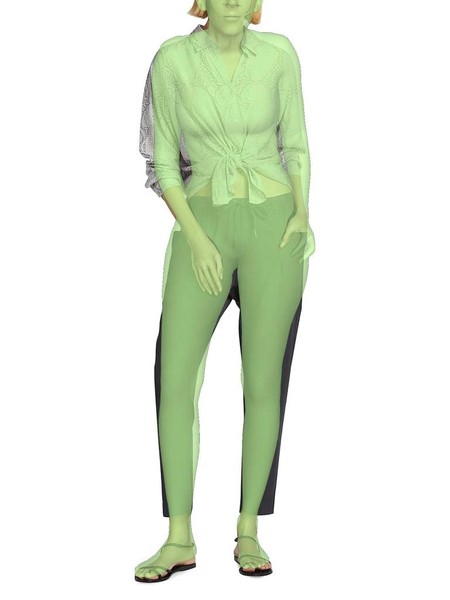}}
\subfloat[]{\includegraphics[height=0.24\textheight]{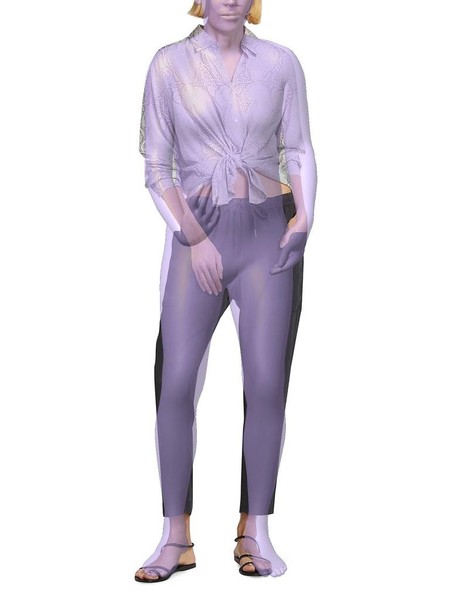}}
\subfloat[]{\includegraphics[height=0.24\textheight]{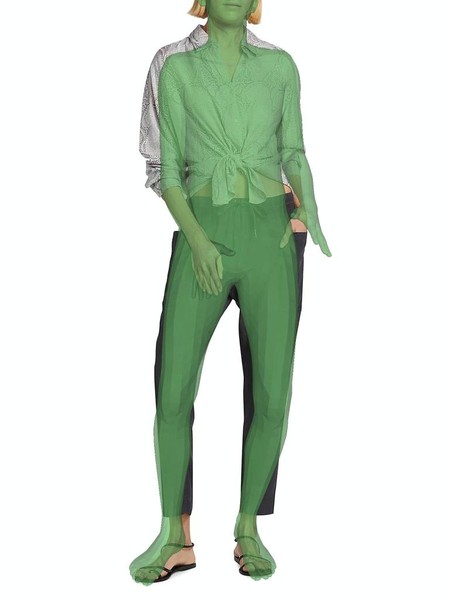}}
\subfloat[]{\includegraphics[height=0.24\textheight]{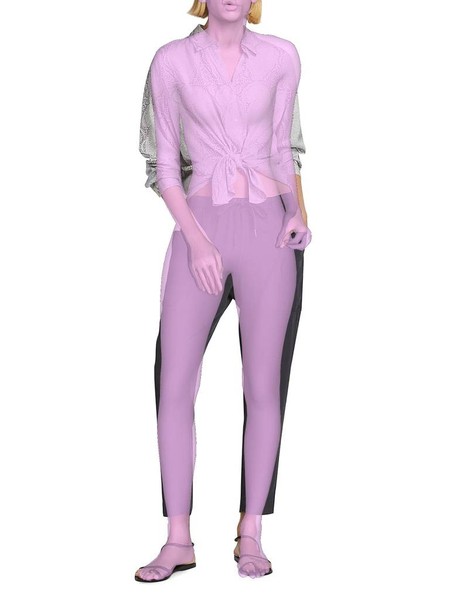}}\\

\vspace{-10pt}

\caption{
Left-to-right: SMPLify-X \cite{pavlakos2019expressive} (\textcolor{caribbeangreen2}{light green}), PyMAF-X \cite{pymafx2022} (\textcolor{violet}{purple}), SHAPY \cite{choutas2022accurate} (\textcolor{jade}{green}) and KBody (\textcolor{candypink}{pink}).
}
\label{fig:partial_sp7}
\end{figure*}

%% file: figures/supp/partial_splendid8.tex
\begin{figure*}[!htbp]
\captionsetup[subfigure]{position=bottom,labelformat=empty}

\centering

\subfloat{\includegraphics[height=0.24\textheight]{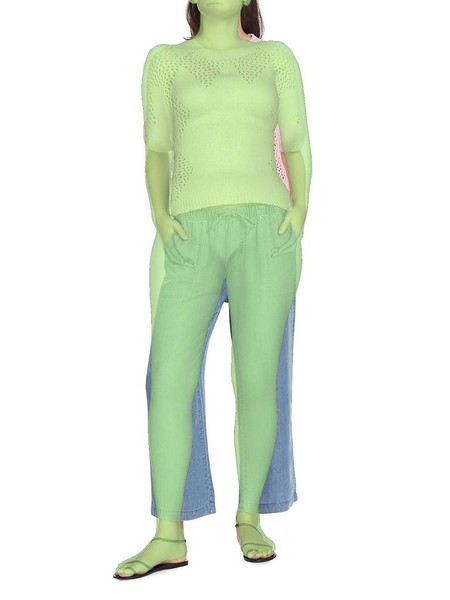}}
\subfloat{\includegraphics[height=0.24\textheight]{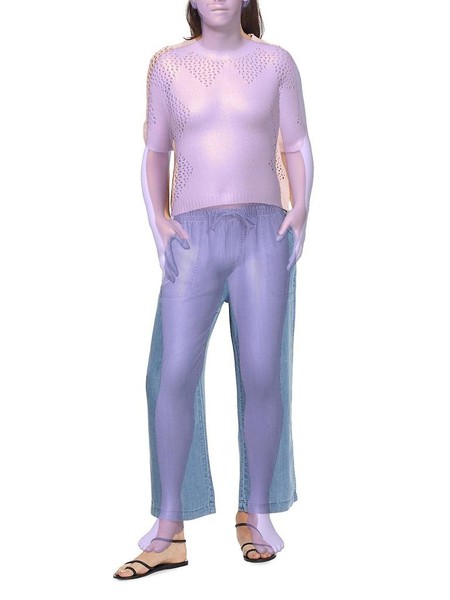}}
\subfloat{\includegraphics[height=0.24\textheight]{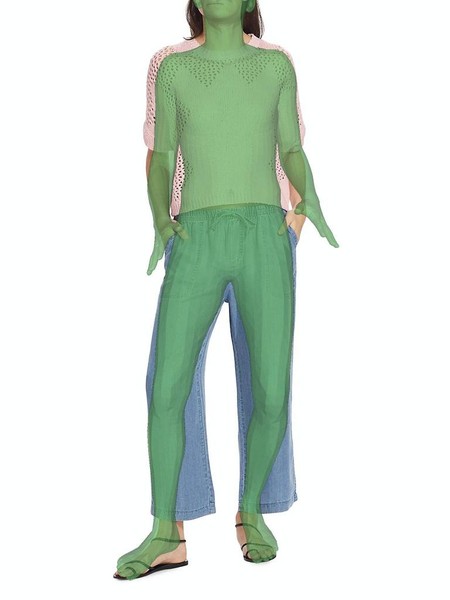}}
\subfloat{\includegraphics[height=0.24\textheight]{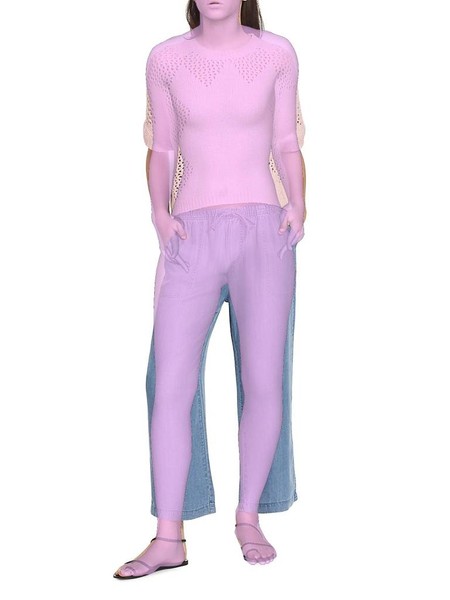}}\\

\subfloat{\includegraphics[height=0.24\textheight]{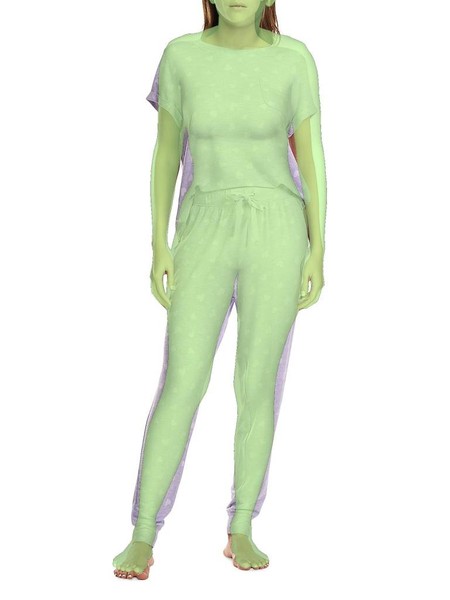}}
\subfloat{\includegraphics[height=0.24\textheight]{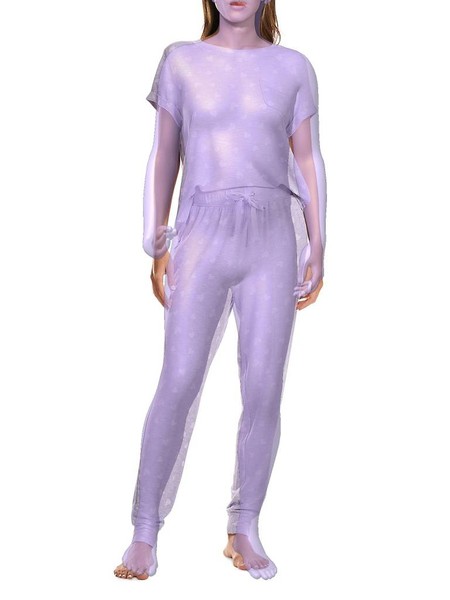}}
\subfloat{\includegraphics[height=0.24\textheight]{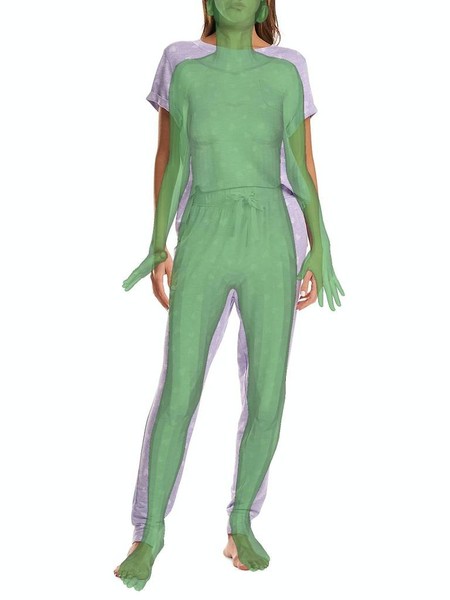}}
\subfloat{\includegraphics[height=0.24\textheight]{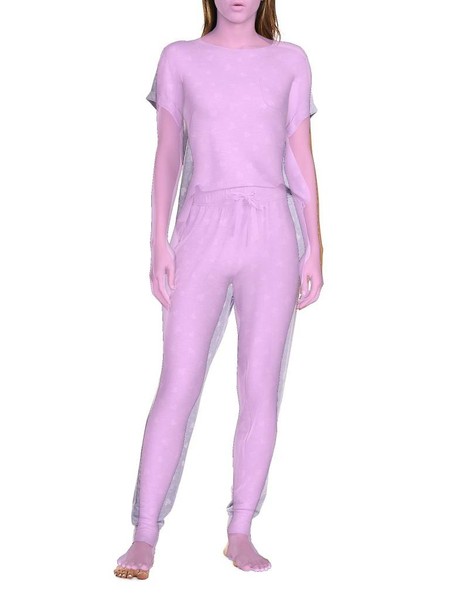}}\\

\subfloat[]{\includegraphics[height=0.24\textheight]{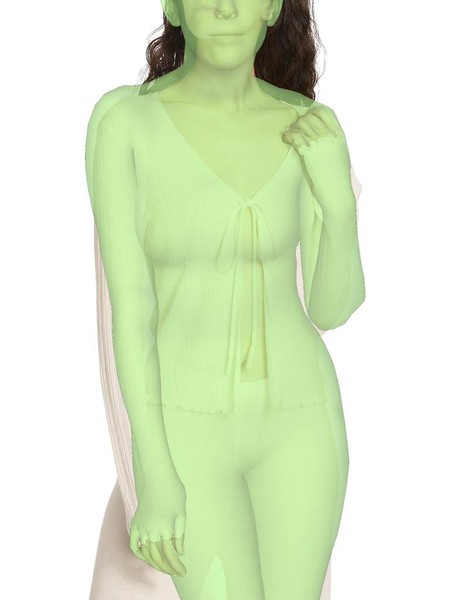}}
\subfloat[]{\includegraphics[height=0.24\textheight]{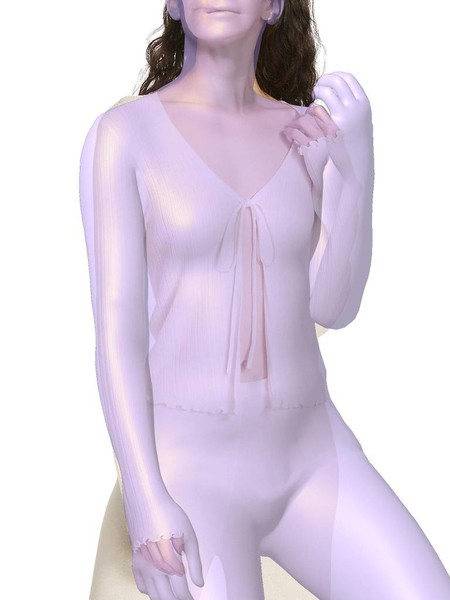}}
\subfloat[]{\includegraphics[height=0.24\textheight]{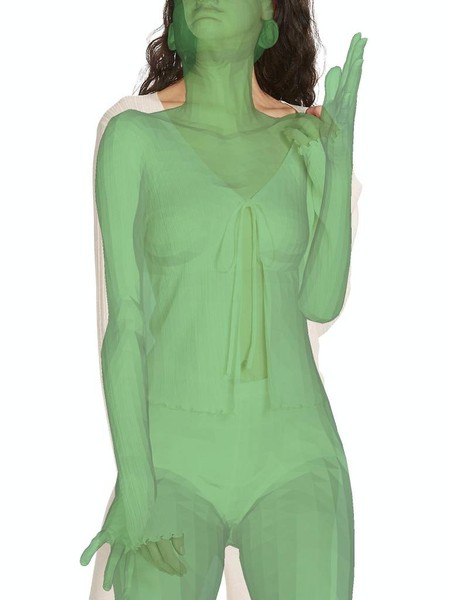}}
\subfloat[]{\includegraphics[height=0.24\textheight]{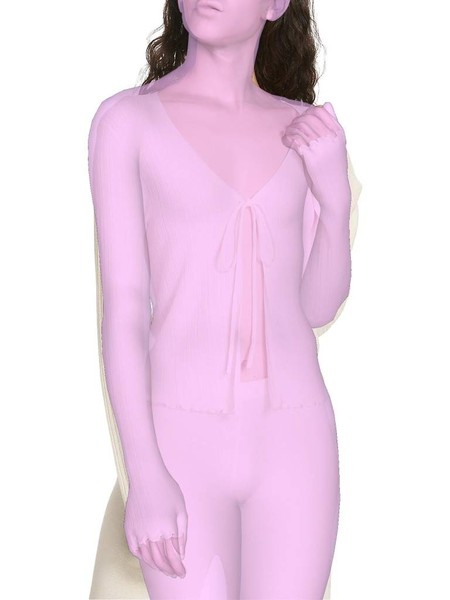}}\\

\vspace{-10pt}

\subfloat[SMPLify-X \cite{pavlakos2019expressive}]
{\includegraphics[height=0.24\textheight]{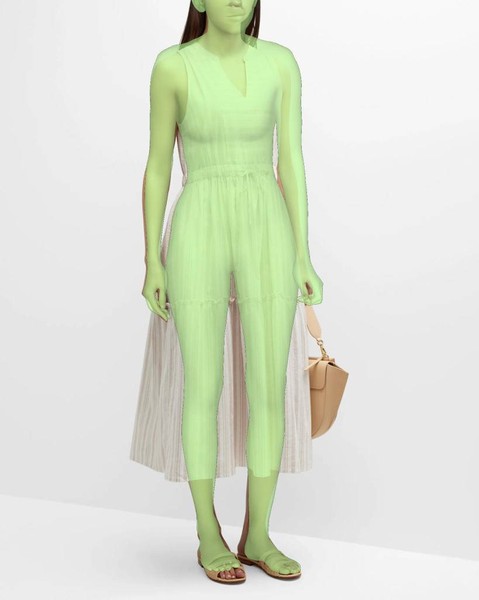}}
\subfloat[PyMAF-X \cite{pymafx2022}]{\includegraphics[height=0.24\textheight]{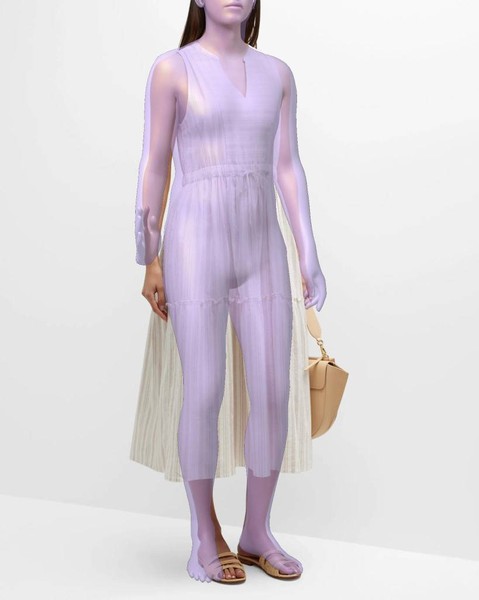}}
\subfloat[SHAPY \cite{choutas2022accurate}]
{\includegraphics[height=0.24\textheight]{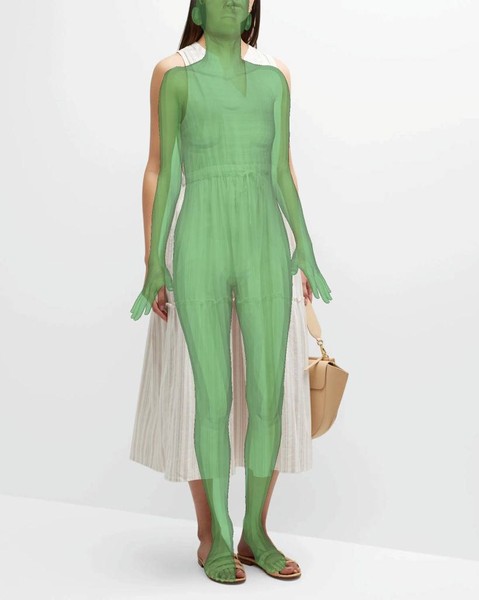}}
\subfloat[\KBody{-.1}{.035} (Ours)]{\includegraphics[height=0.24\textheight]{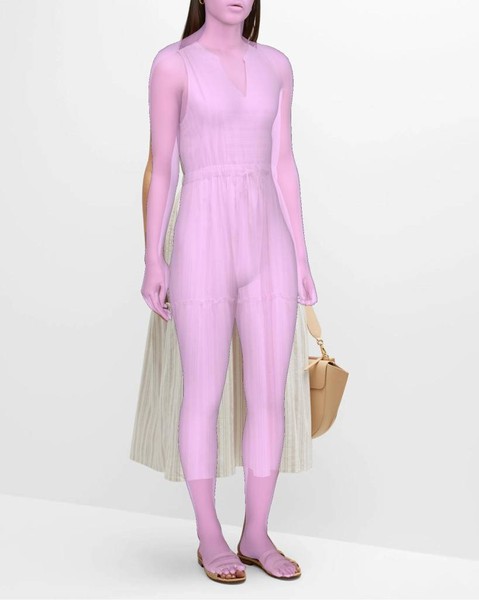}}

\caption{
Left-to-right: SMPLify-X \cite{pavlakos2019expressive} (\textcolor{caribbeangreen2}{light green}), PyMAF-X \cite{pymafx2022} (\textcolor{violet}{purple}), SHAPY \cite{choutas2022accurate} (\textcolor{jade}{green}) and KBody (\textcolor{candypink}{pink}).
}
\label{fig:partial_sp8}
\end{figure*}

%% file: figures/supp/partial_splendid9.tex
\begin{figure*}[!htbp]
\captionsetup[subfigure]{position=bottom,labelformat=empty}

\centering

\subfloat{\includegraphics[height=0.24\textheight]{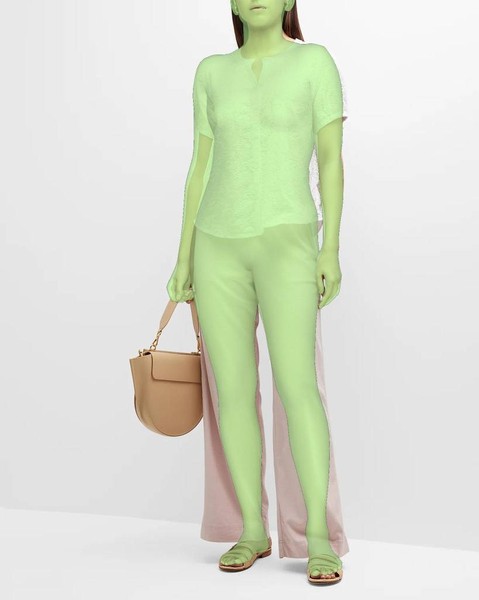}}
\subfloat{\includegraphics[height=0.24\textheight]{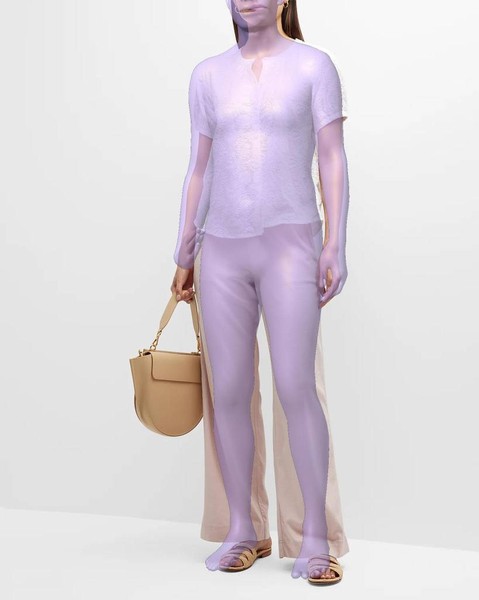}}
\subfloat{\includegraphics[height=0.24\textheight]{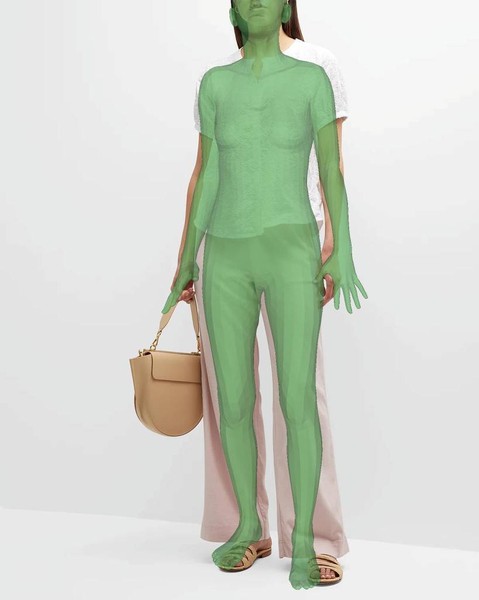}}
\subfloat{\includegraphics[height=0.24\textheight]{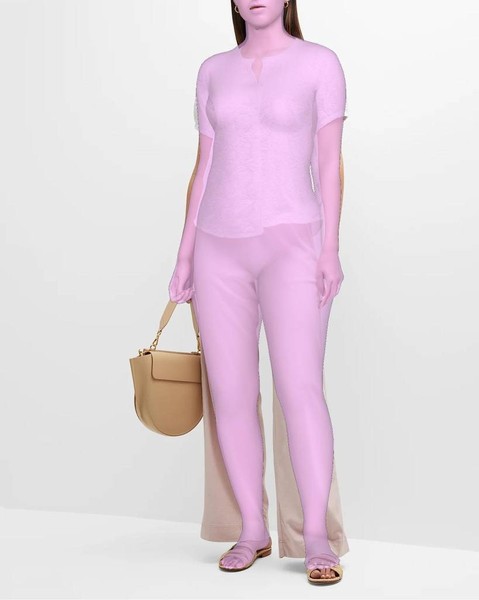}}\\

\subfloat{\includegraphics[height=0.24\textheight]{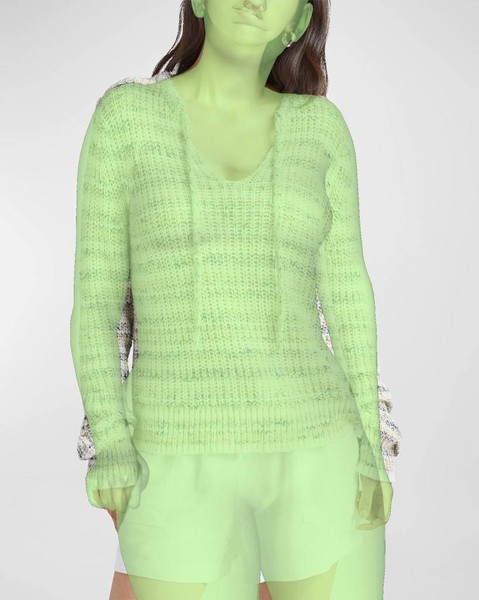}}
\subfloat{\includegraphics[height=0.24\textheight]{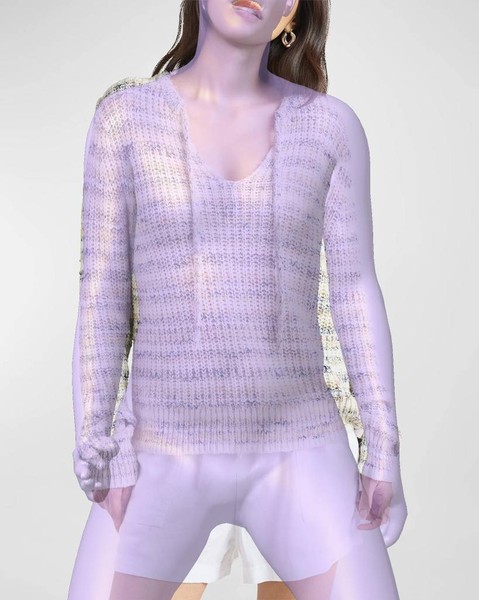}}
\subfloat{\includegraphics[height=0.24\textheight]{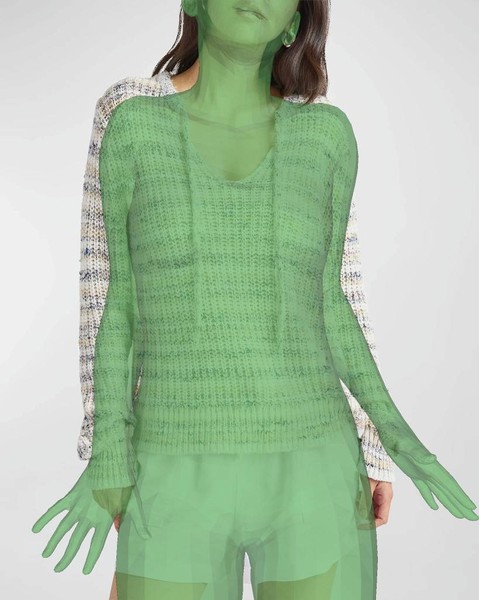}}
\subfloat{\includegraphics[height=0.24\textheight]{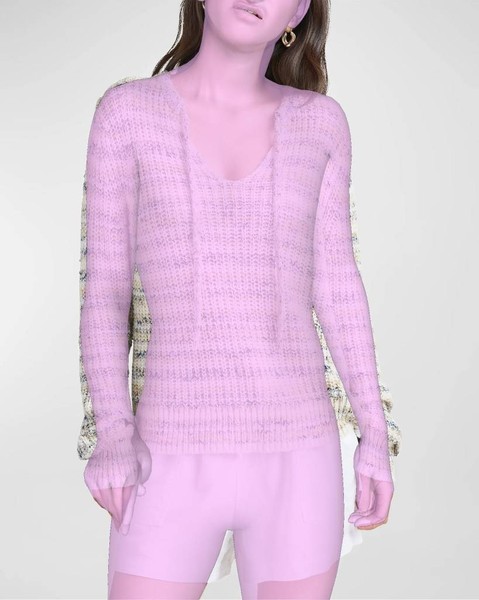}}\\

\subfloat[]{\includegraphics[height=0.24\textheight]{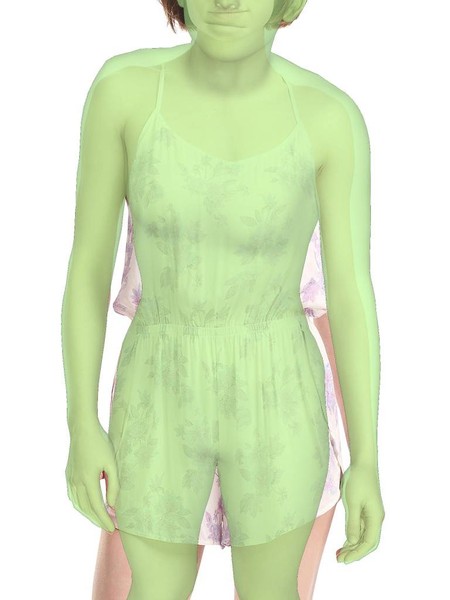}}
\subfloat[]{\includegraphics[height=0.24\textheight]{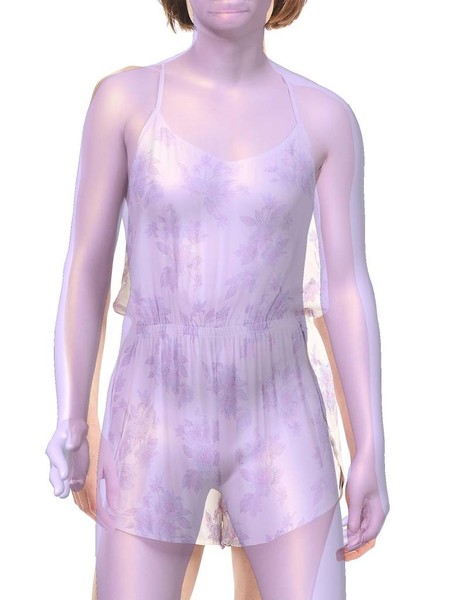}}
\subfloat[]{\includegraphics[height=0.24\textheight]{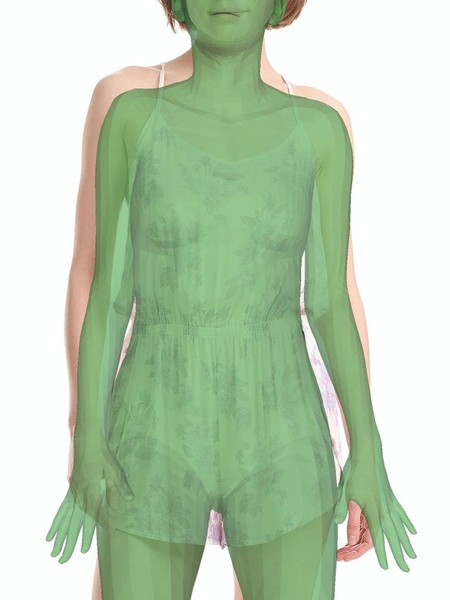}}
\subfloat[]{\includegraphics[height=0.24\textheight]{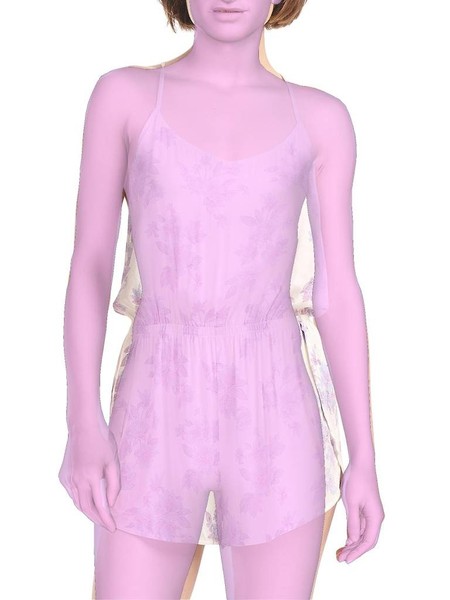}}\\

\vspace{-10pt}

\subfloat[SMPLify-X \cite{pavlakos2019expressive}]
{\includegraphics[height=0.24\textheight]{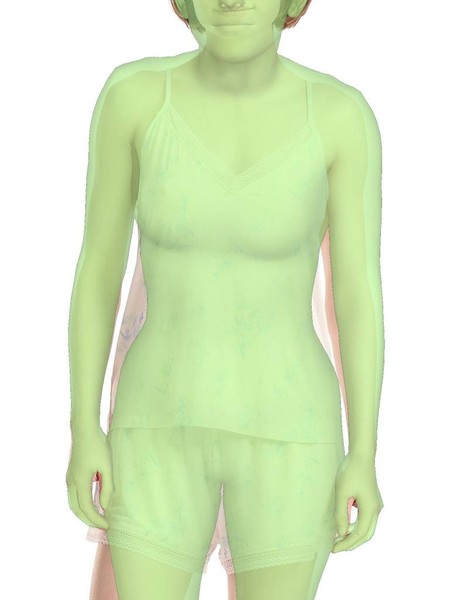}}
\subfloat[PyMAF-X \cite{pymafx2022}]{\includegraphics[height=0.24\textheight]{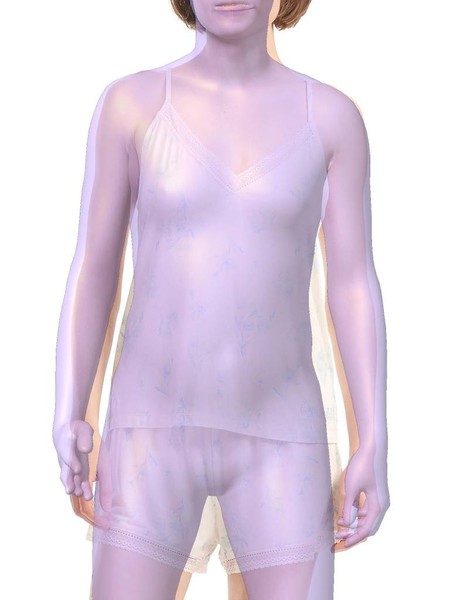}}
\subfloat[SHAPY \cite{choutas2022accurate}]
{\includegraphics[height=0.24\textheight]{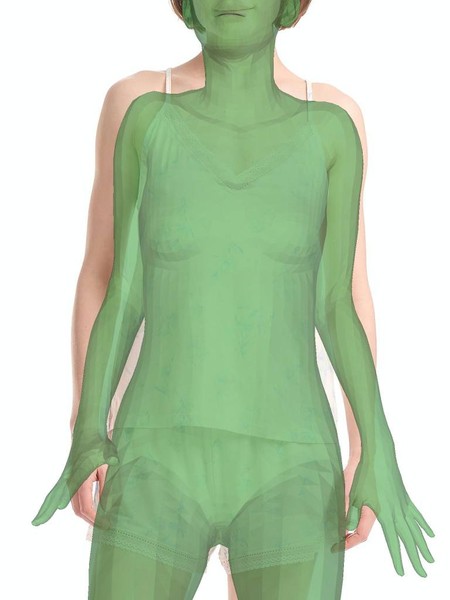}}
\subfloat[\KBody{-.1}{.035} (Ours)]{\includegraphics[height=0.24\textheight]{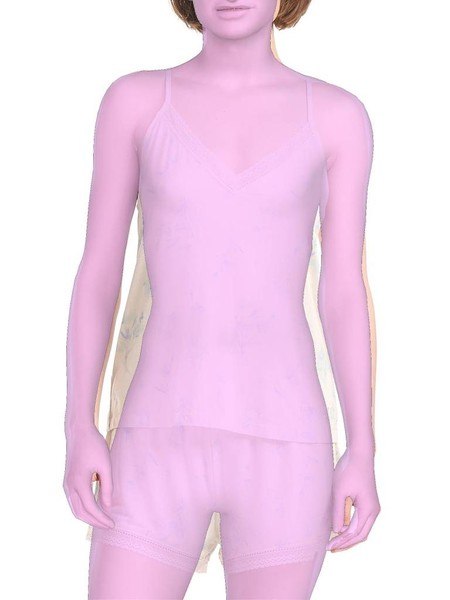}}

\caption{
Left-to-right: SMPLify-X \cite{pavlakos2019expressive} (\textcolor{caribbeangreen2}{light green}), PyMAF-X \cite{pymafx2022} (\textcolor{violet}{purple}), SHAPY \cite{choutas2022accurate} (\textcolor{jade}{green}) and KBody (\textcolor{candypink}{pink}).
}
\label{fig:partial_sp9}
\end{figure*}

%% file: figures/supp/partial_splendid_good_american.tex
\begin{figure*}[!htbp]
\captionsetup[subfigure]{position=bottom,labelformat=empty}

\centering

\subfloat{\includegraphics[height=0.24\textheight]{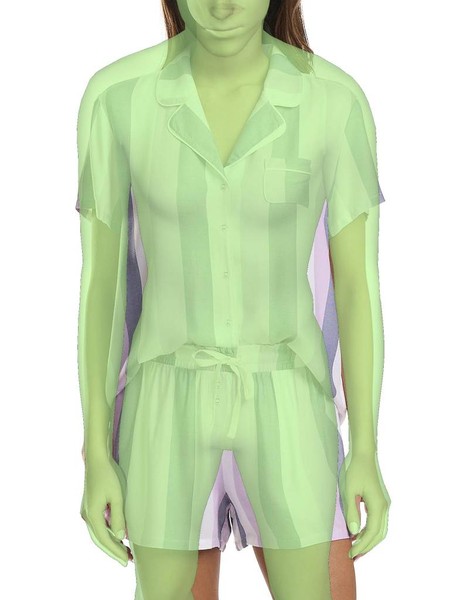}}
\subfloat{\includegraphics[height=0.24\textheight]{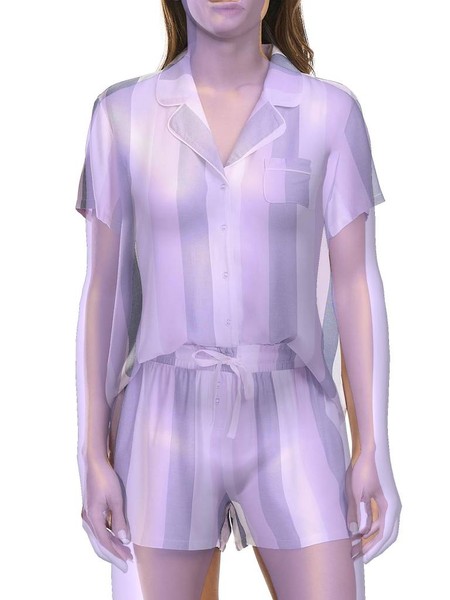}}
\subfloat{\includegraphics[height=0.24\textheight]{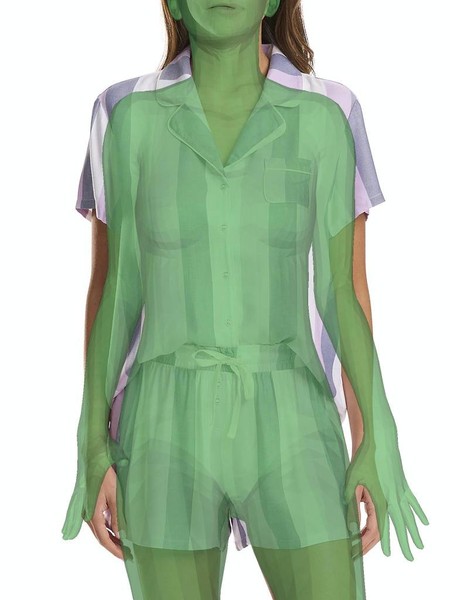}}
\subfloat{\includegraphics[height=0.24\textheight]{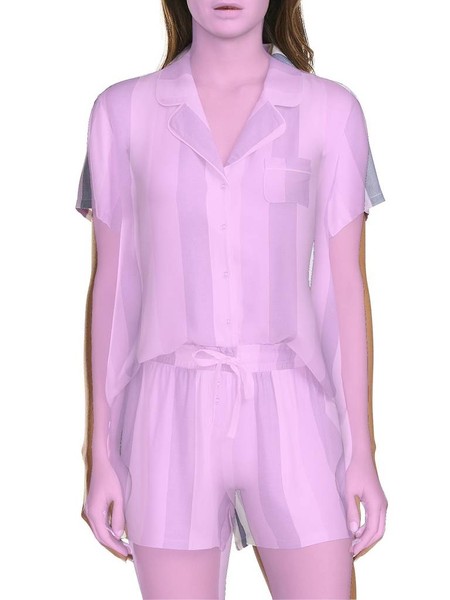}}\\

\subfloat{\includegraphics[height=0.24\textheight]{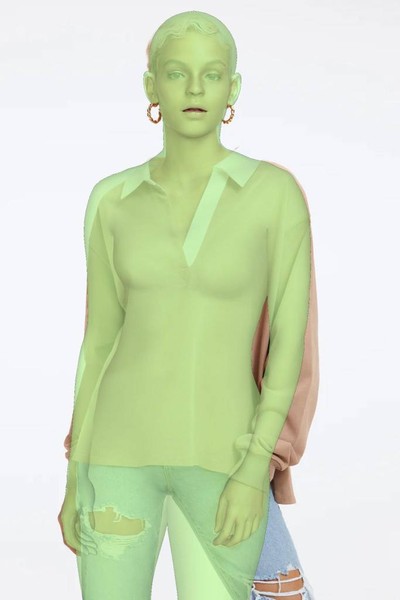}}
\subfloat{\includegraphics[height=0.24\textheight]{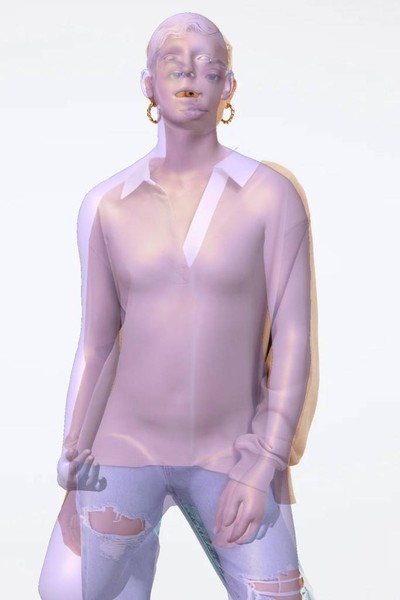}}
\subfloat{\includegraphics[height=0.24\textheight]{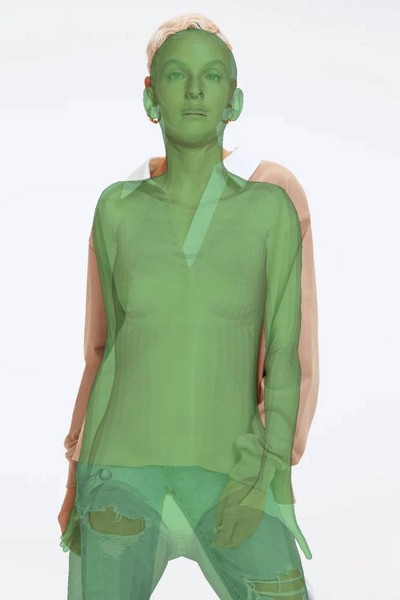}}
\subfloat{\includegraphics[height=0.24\textheight]{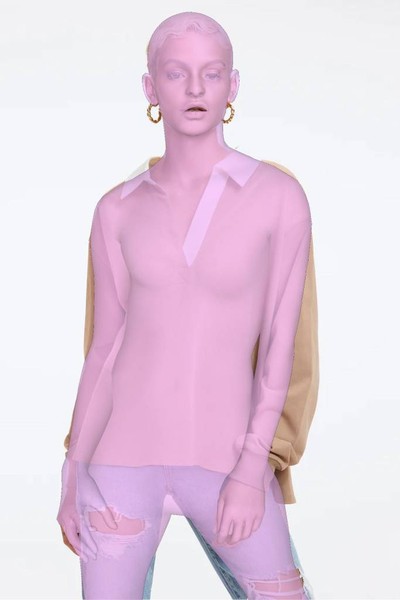}}\\

\subfloat[]{\includegraphics[height=0.24\textheight]{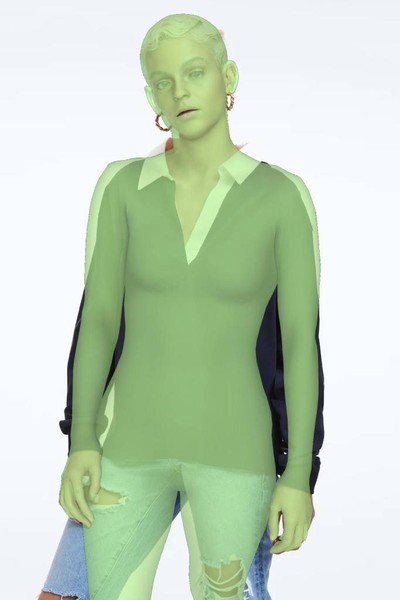}}
\subfloat[]{\includegraphics[height=0.24\textheight]{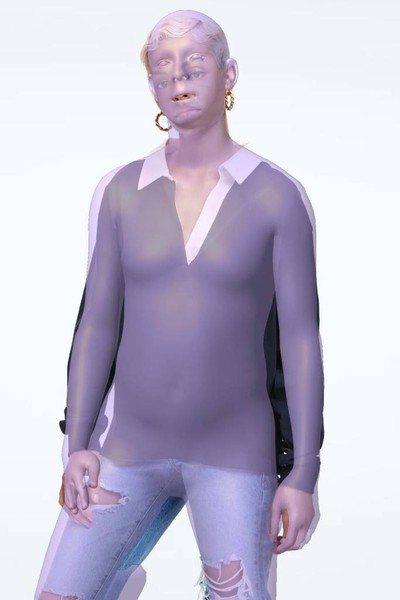}}
\subfloat[]{\includegraphics[height=0.24\textheight]{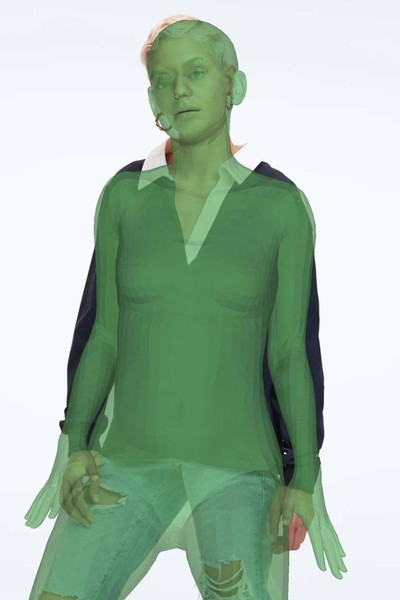}}
\subfloat[]{\includegraphics[height=0.24\textheight]{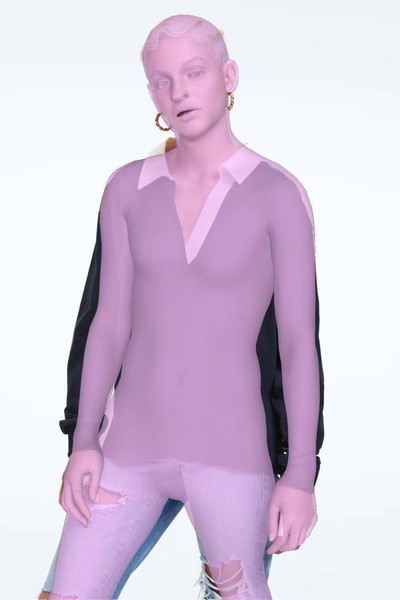}}\\

\vspace{-10pt}

\subfloat[SMPLify-X \cite{pavlakos2019expressive}]
{\includegraphics[height=0.24\textheight]{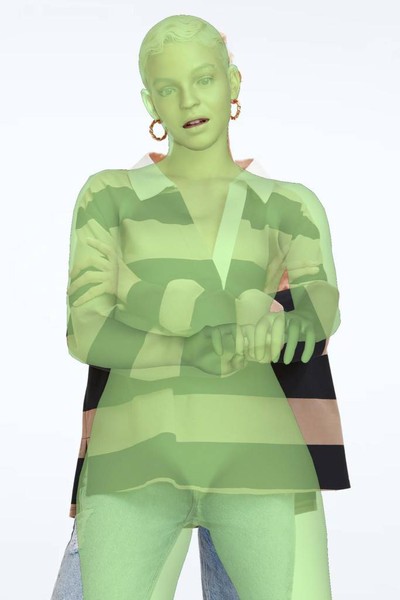}}
\subfloat[PyMAF-X \cite{pymafx2022}]{\includegraphics[height=0.24\textheight]{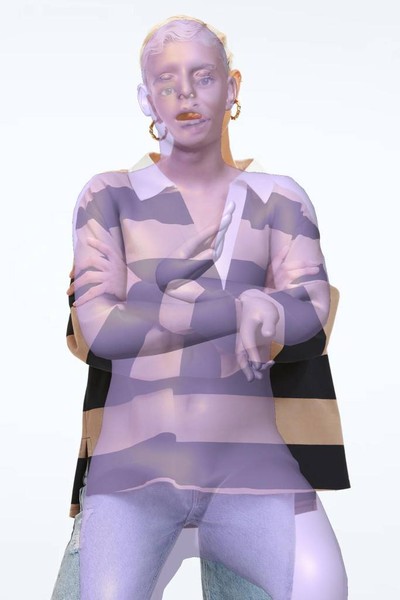}}
\subfloat[SHAPY \cite{choutas2022accurate}]
{\includegraphics[height=0.24\textheight]{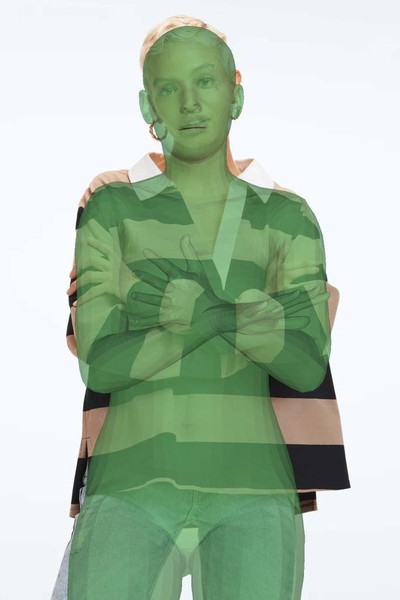}}
\subfloat[\KBody{-.1}{.035} (Ours)]{\includegraphics[height=0.24\textheight]{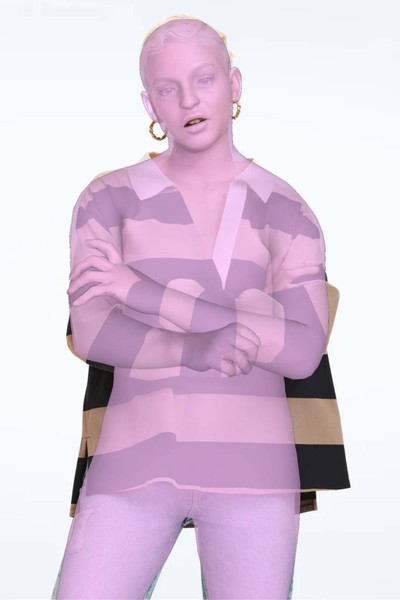}}

\caption{
Left-to-right: SMPLify-X \cite{pavlakos2019expressive} (\textcolor{caribbeangreen2}{light green}), PyMAF-X \cite{pymafx2022} (\textcolor{violet}{purple}), SHAPY \cite{choutas2022accurate} (\textcolor{jade}{green}) and KBody (\textcolor{candypink}{pink}).
}
\label{fig:partial_sp_good}
\end{figure*}

%% file: figures/supp/partial_good_american.tex
\begin{figure*}[!htbp]
\captionsetup[subfigure]{position=bottom,labelformat=empty}

\centering

\subfloat{\includegraphics[height=0.24\textheight]{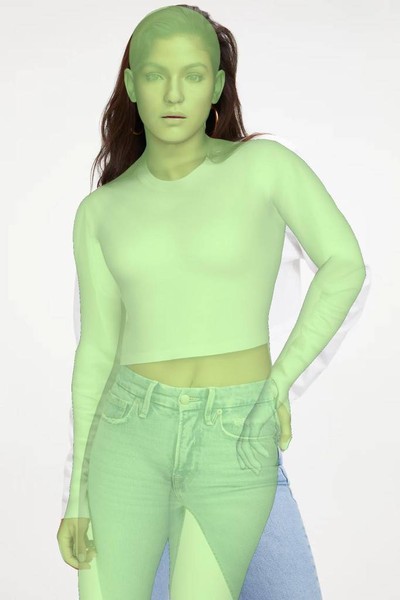}}
\subfloat{\includegraphics[height=0.24\textheight]{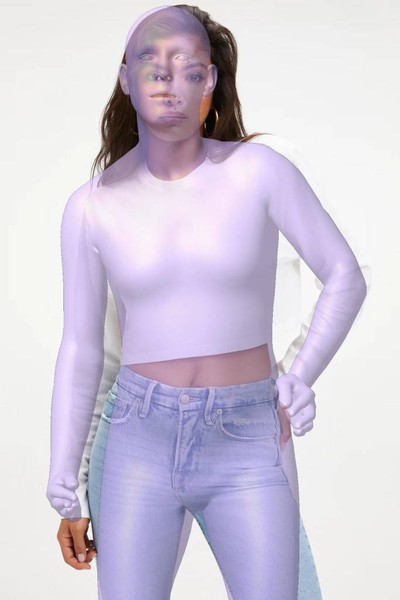}}
\subfloat{\includegraphics[height=0.24\textheight]{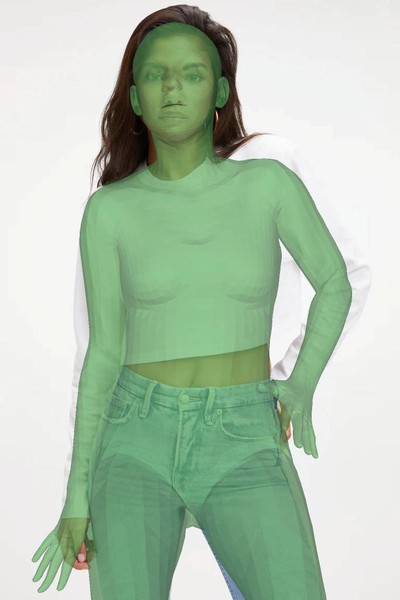}}
\subfloat{\includegraphics[height=0.24\textheight]{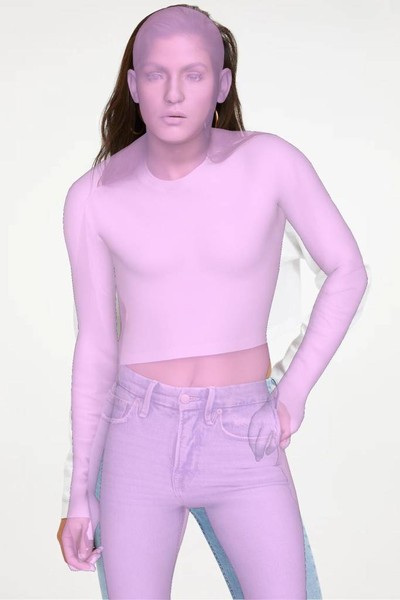}}\\

\subfloat[]{\includegraphics[height=0.24\textheight]{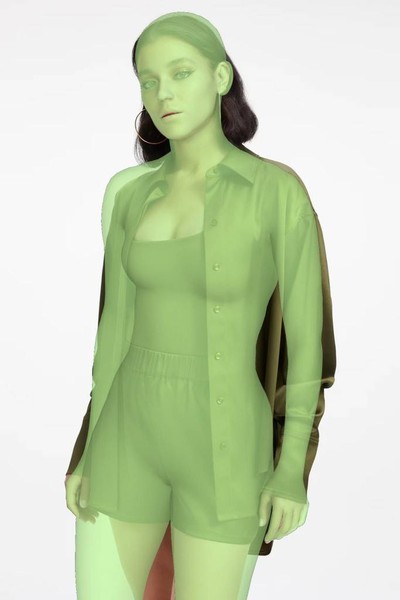}}
\subfloat[]{\includegraphics[height=0.24\textheight]{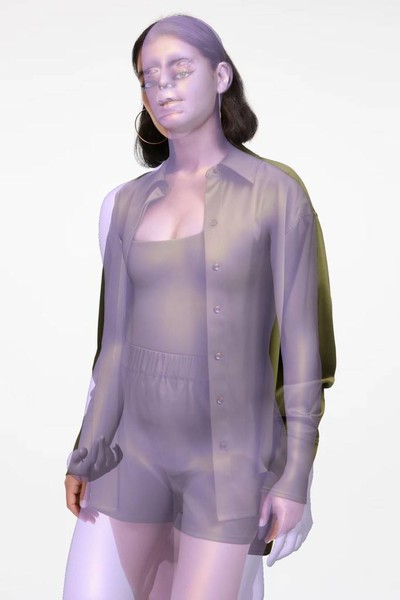}}
\subfloat[]{\includegraphics[height=0.24\textheight]{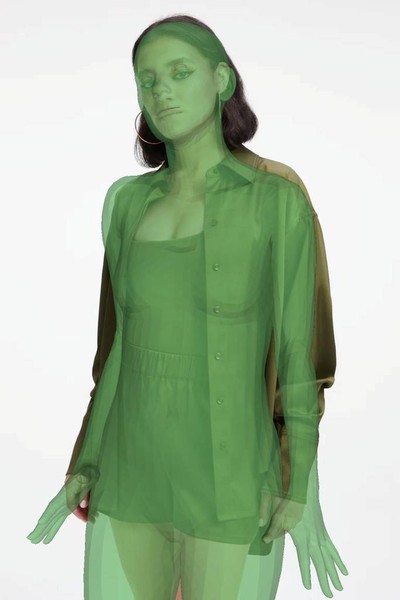}}
\subfloat[]{\includegraphics[height=0.24\textheight]{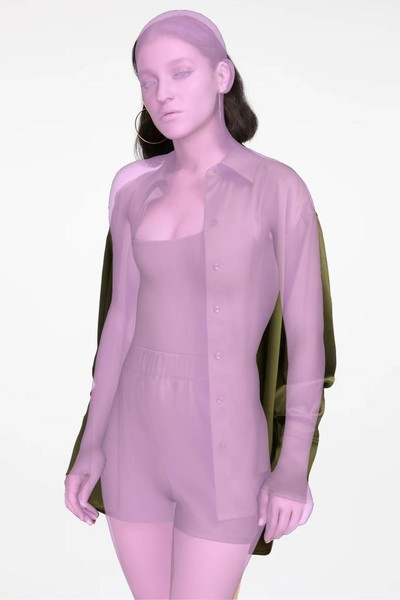}}\\

\vspace{-10pt}

\subfloat[SMPLify-X \cite{pavlakos2019expressive}]
{\includegraphics[height=0.24\textheight]{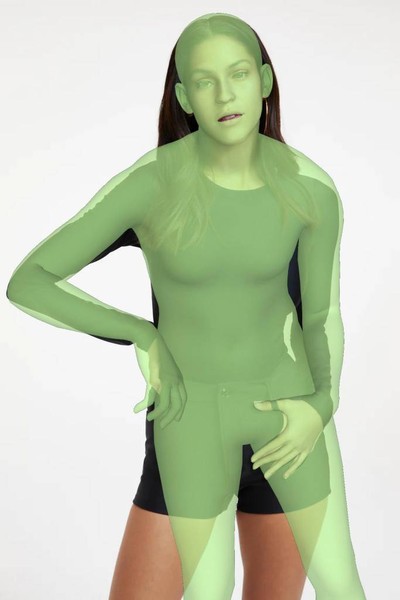}}
\subfloat[PyMAF-X \cite{pymafx2022}]{\includegraphics[height=0.24\textheight]{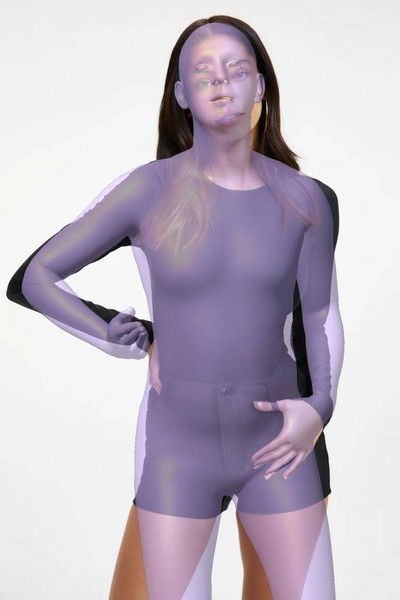}}
\subfloat[SHAPY \cite{choutas2022accurate}]
{\includegraphics[height=0.24\textheight]{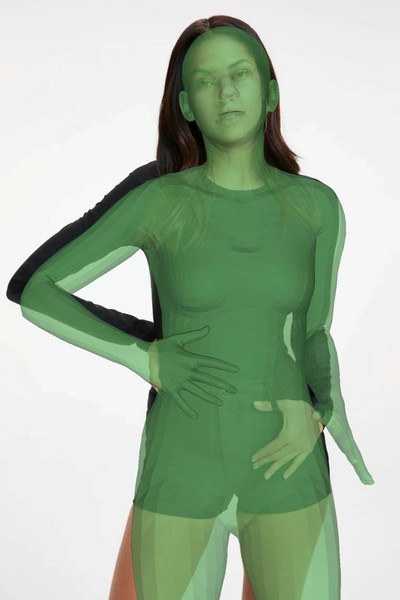}}
\subfloat[\KBody{-.1}{.035} (Ours)]{\includegraphics[height=0.24\textheight]{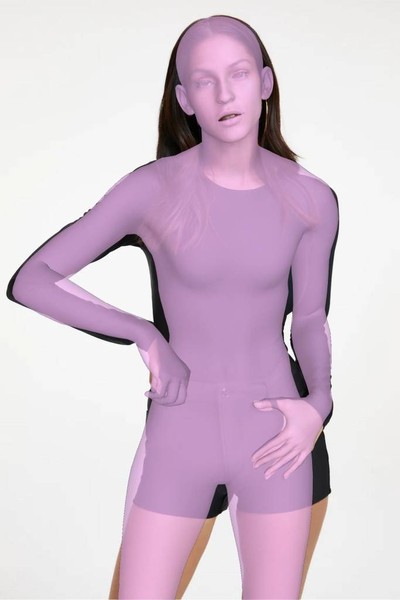}}

\caption{
Left-to-right: SMPLify-X \cite{pavlakos2019expressive} (\textcolor{caribbeangreen2}{light green}), PyMAF-X \cite{pymafx2022} (\textcolor{violet}{purple}), SHAPY \cite{choutas2022accurate} (\textcolor{jade}{green}) and KBody (\textcolor{candypink}{pink}).
}
\label{fig:partial_good}
\end{figure*}

%% file: figures/supp/partial_rl1.tex
\begin{figure*}[!htbp]
\captionsetup[subfigure]{position=bottom,labelformat=empty}

\centering

\subfloat{\includegraphics[height=0.24\textheight]{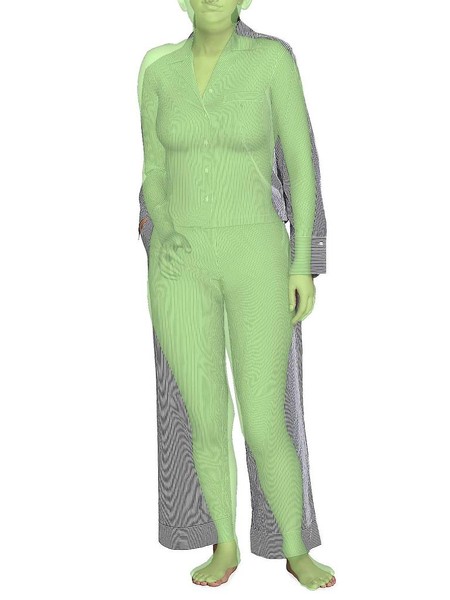}}
\subfloat{\includegraphics[height=0.24\textheight]{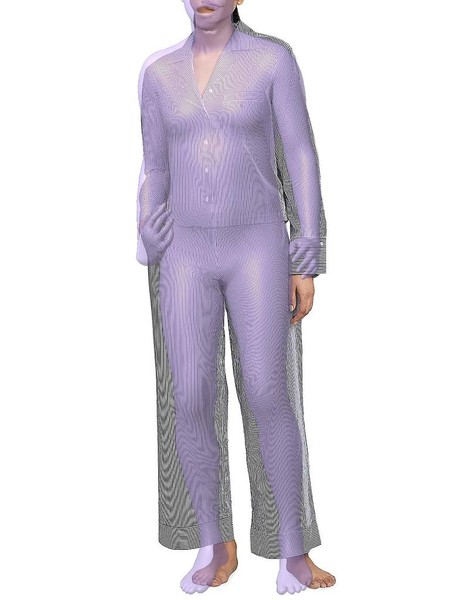}}
\subfloat{\includegraphics[height=0.24\textheight]{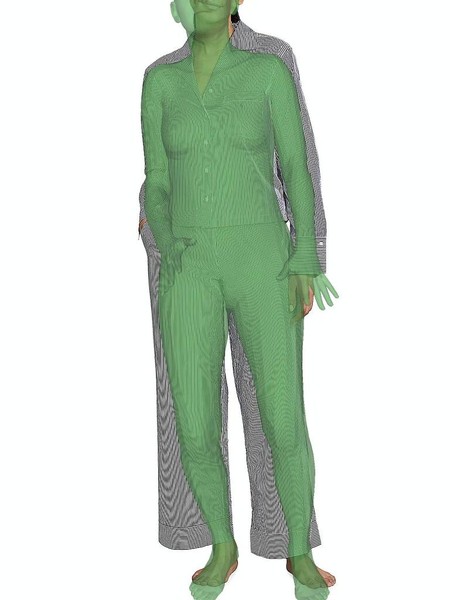}}
\subfloat{\includegraphics[height=0.24\textheight]{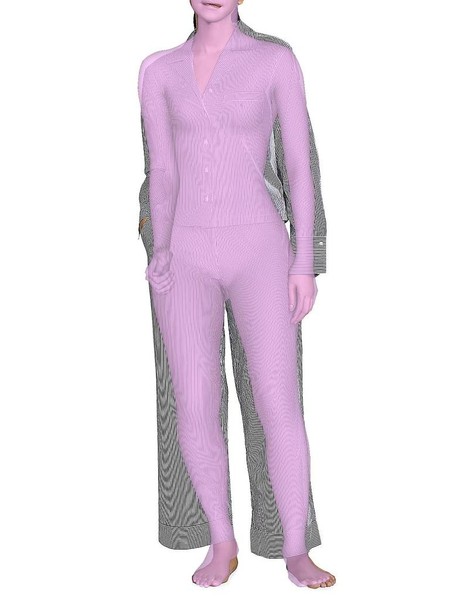}}\\

\subfloat{\includegraphics[height=0.24\textheight]{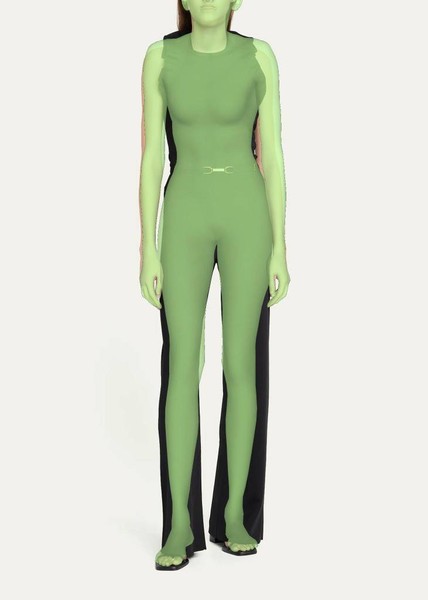}}
\subfloat{\includegraphics[height=0.24\textheight]{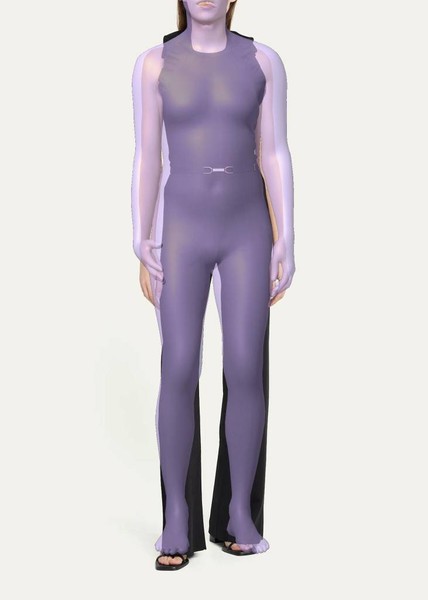}}
\subfloat{\includegraphics[height=0.24\textheight]{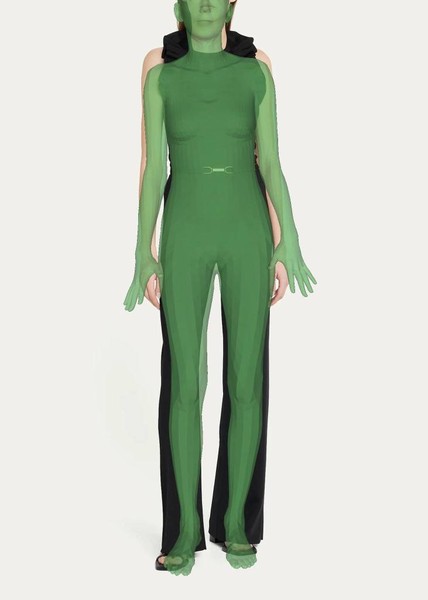}}
\subfloat{\includegraphics[height=0.24\textheight]{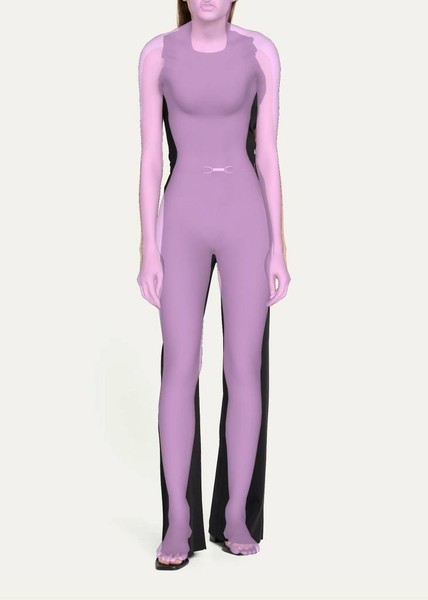}}\\

\subfloat[]{\includegraphics[height=0.24\textheight]{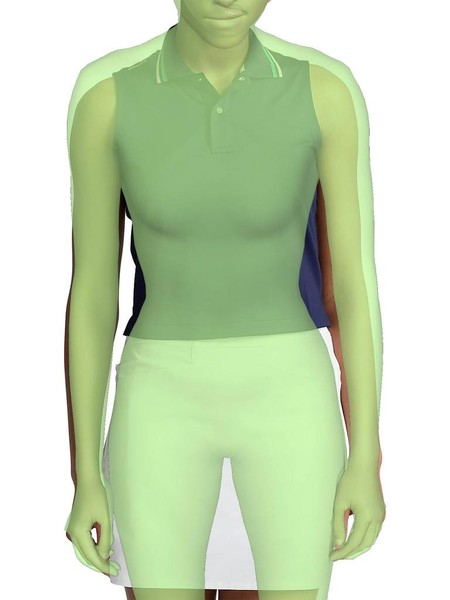}}
\subfloat[]{\includegraphics[height=0.24\textheight]{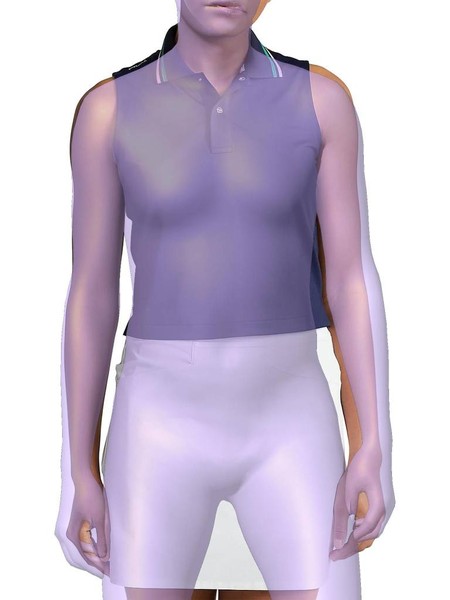}}
\subfloat[]{\includegraphics[height=0.24\textheight]{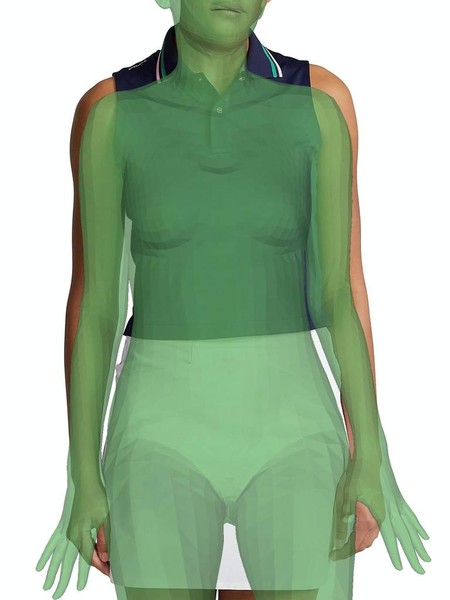}}
\subfloat[]{\includegraphics[height=0.24\textheight]{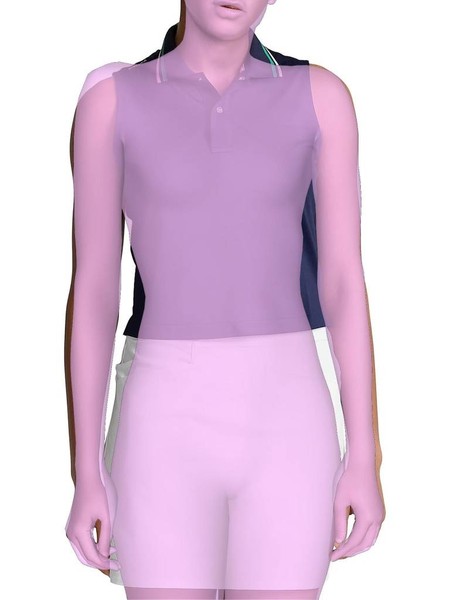}}\\

\vspace{-10pt}

\subfloat[SMPLify-X \cite{pavlakos2019expressive}]
{\includegraphics[height=0.24\textheight]{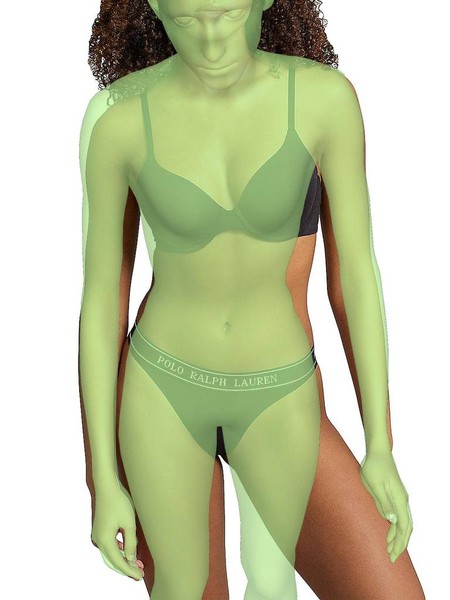}}
\subfloat[PyMAF-X \cite{pymafx2022}]{\includegraphics[height=0.24\textheight]{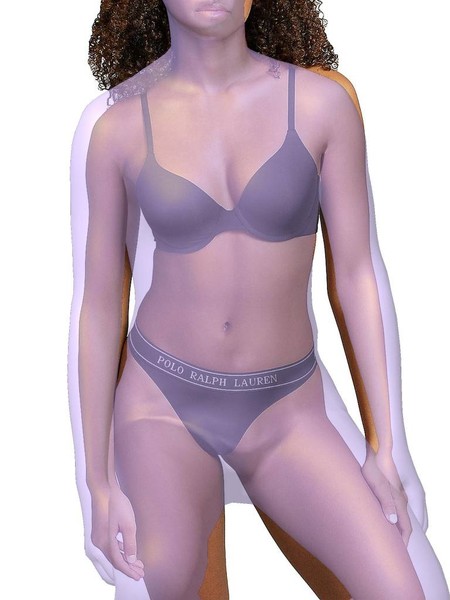}}
\subfloat[SHAPY \cite{choutas2022accurate}]
{\includegraphics[height=0.24\textheight]
{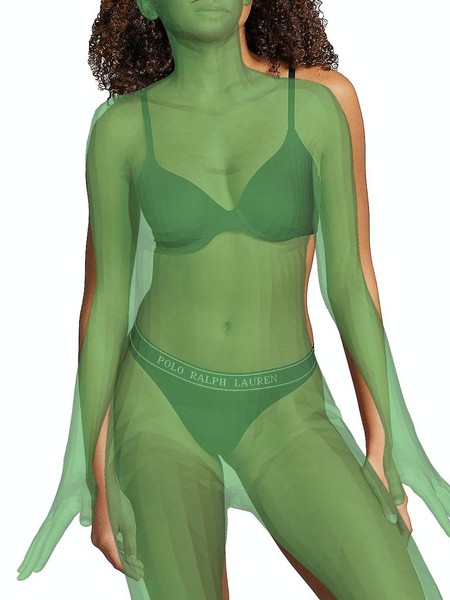}}
\subfloat[\KBody{-.1}{.035} (Ours)]{\includegraphics[height=0.24\textheight]{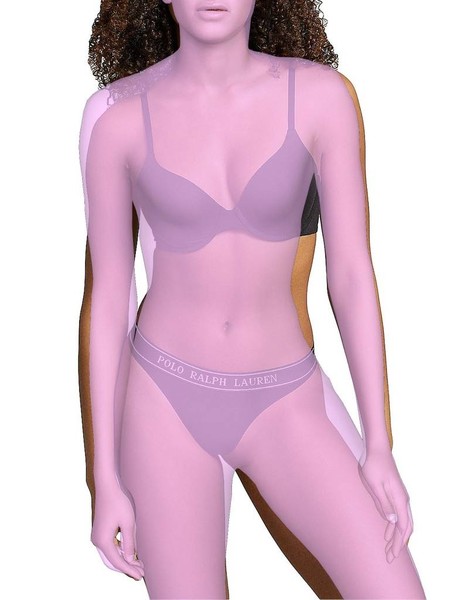}}

\caption{
Left-to-right: SMPLify-X \cite{pavlakos2019expressive} (\textcolor{caribbeangreen2}{light green}), PyMAF-X \cite{pymafx2022} (\textcolor{violet}{purple}), SHAPY \cite{choutas2022accurate} (\textcolor{jade}{green}) and KBody (\textcolor{candypink}{pink}).
}
\label{fig:partial_rl1}
\end{figure*}

%% file: figures/supp/partial_rl2.tex
\begin{figure*}[!htbp]
\captionsetup[subfigure]{position=bottom,labelformat=empty}

\centering

\subfloat{\includegraphics[height=0.24\textheight]{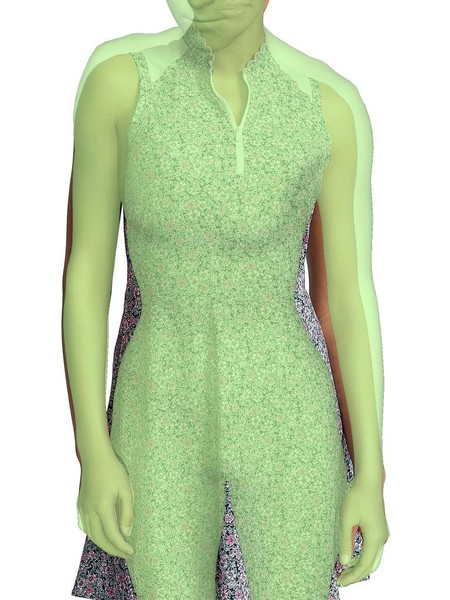}}
\subfloat{\includegraphics[height=0.24\textheight]{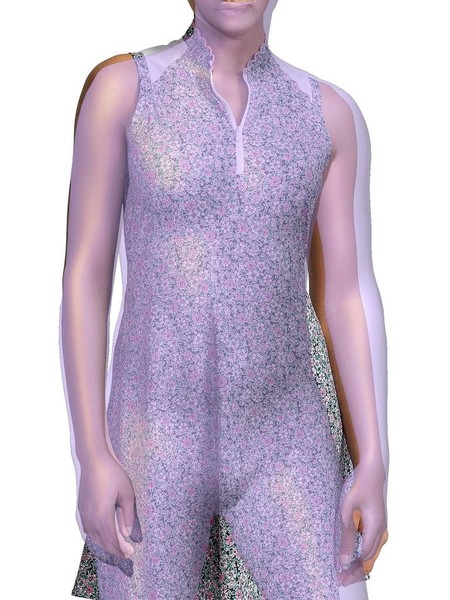}}
\subfloat{\includegraphics[height=0.24\textheight]{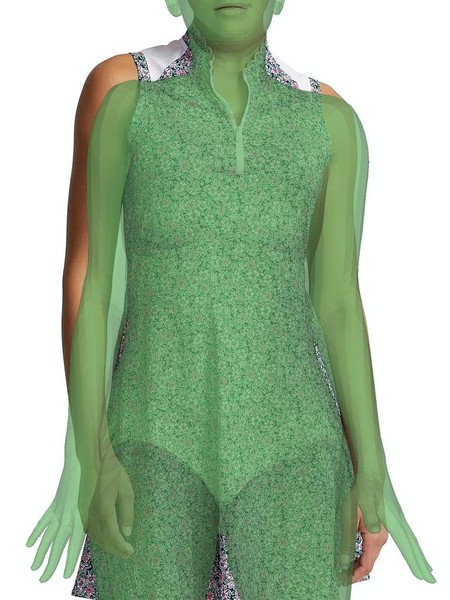}}
\subfloat{\includegraphics[height=0.24\textheight]{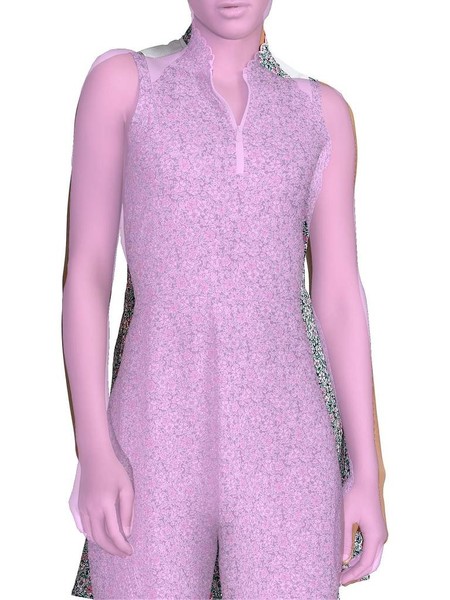}}\\

\subfloat{\includegraphics[height=0.24\textheight]{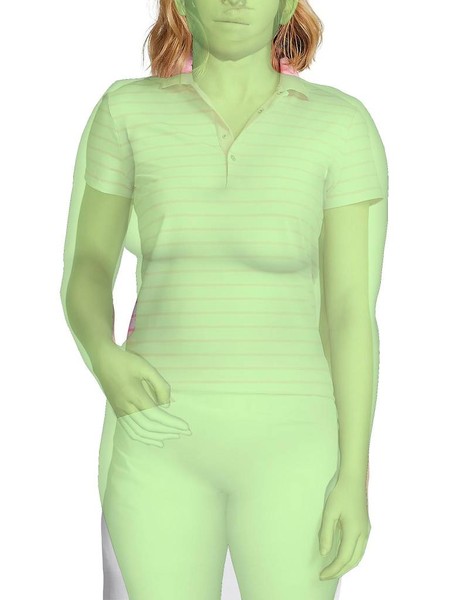}}
\subfloat{\includegraphics[height=0.24\textheight]{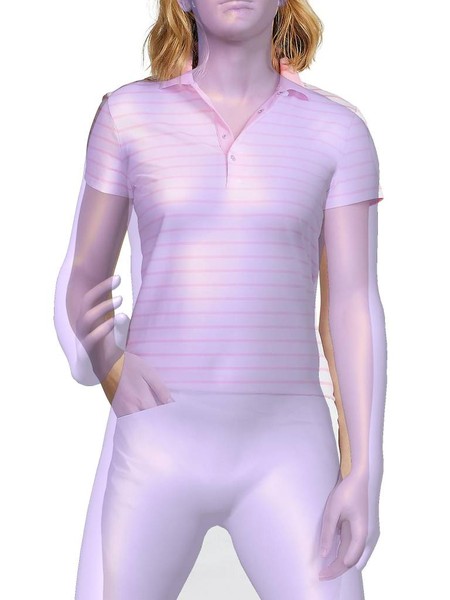}}
\subfloat{\includegraphics[height=0.24\textheight]{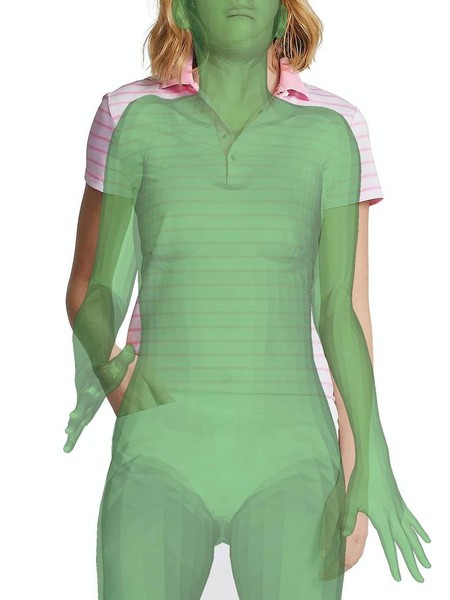}}
\subfloat{\includegraphics[height=0.24\textheight]{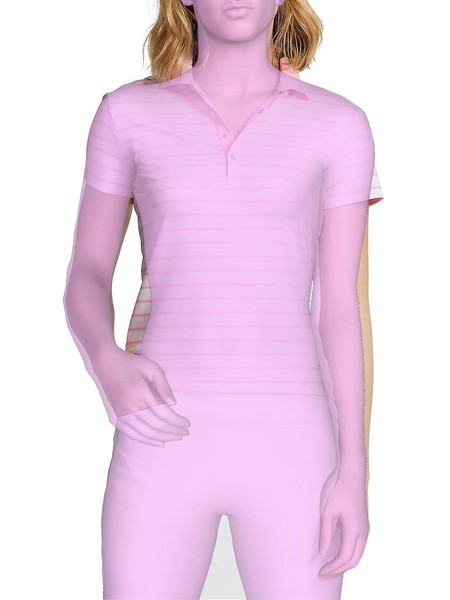}}\\

\subfloat[]{\includegraphics[height=0.24\textheight]{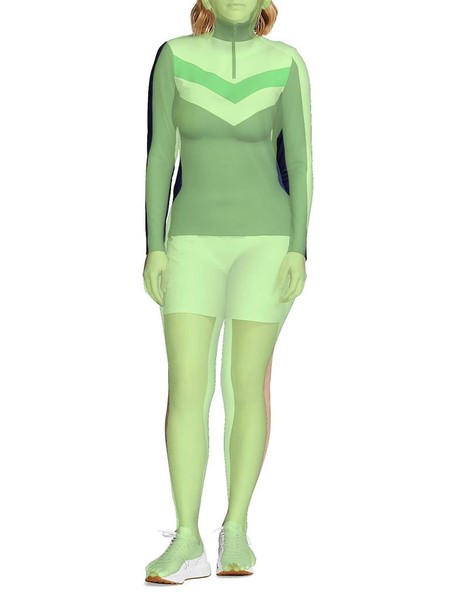}}
\subfloat[]{\includegraphics[height=0.24\textheight]{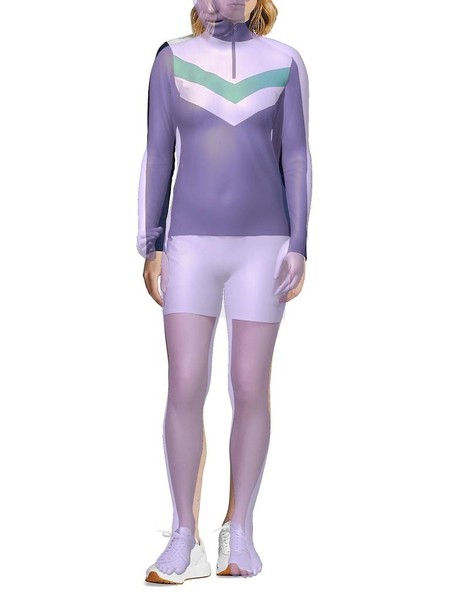}}
\subfloat[]{\includegraphics[height=0.24\textheight]{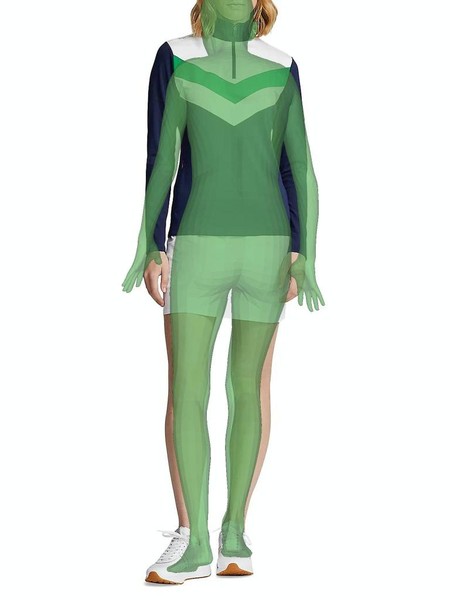}}
\subfloat[]{\includegraphics[height=0.24\textheight]{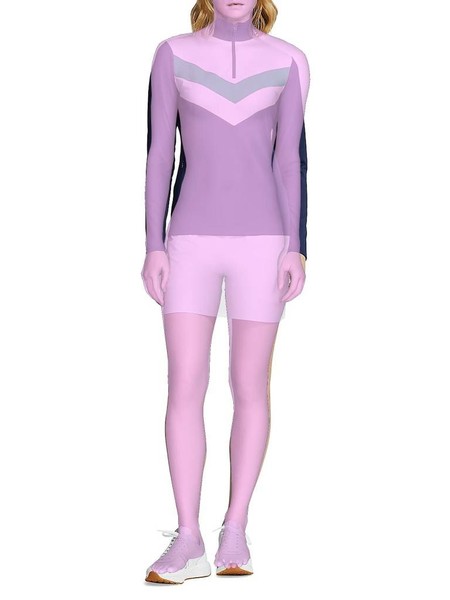}}\\

\vspace{-10pt}

\subfloat[SMPLify-X \cite{pavlakos2019expressive}]
{\includegraphics[height=0.24\textheight]{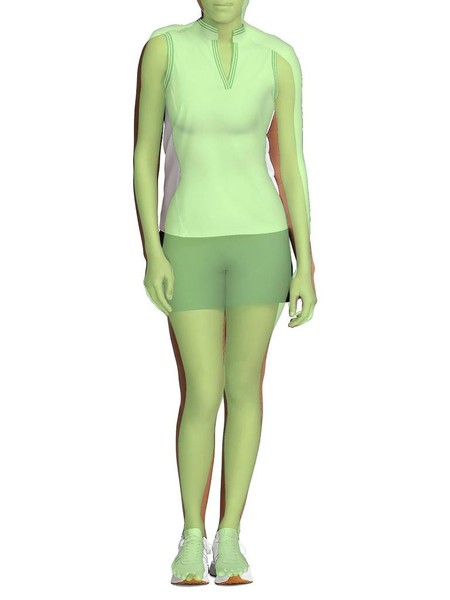}}
\subfloat[PyMAF-X \cite{pymafx2022}]{\includegraphics[height=0.24\textheight]{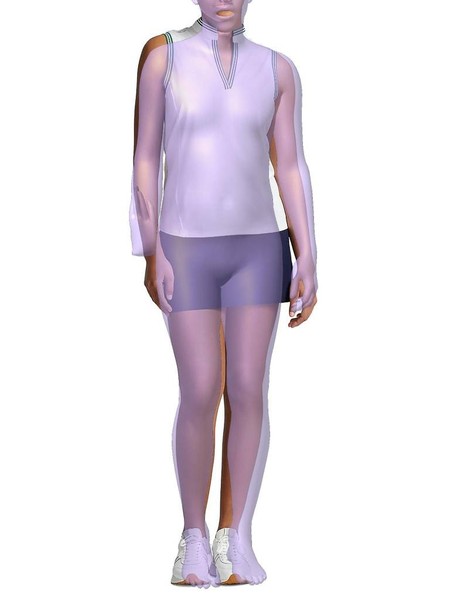}}
\subfloat[SHAPY \cite{choutas2022accurate}]
{\includegraphics[height=0.24\textheight]
{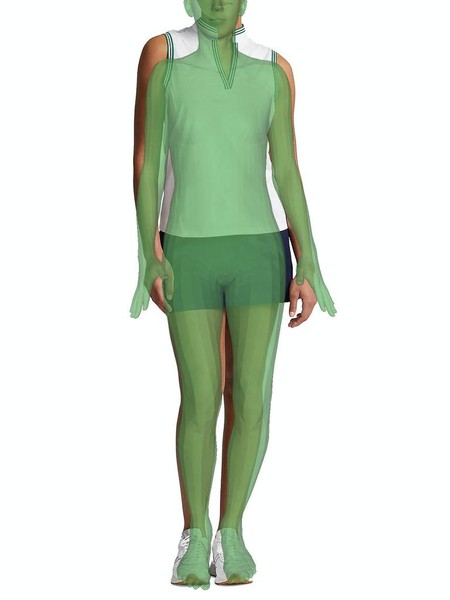}}
\subfloat[\KBody{-.1}{.035} (Ours)]{\includegraphics[height=0.24\textheight]{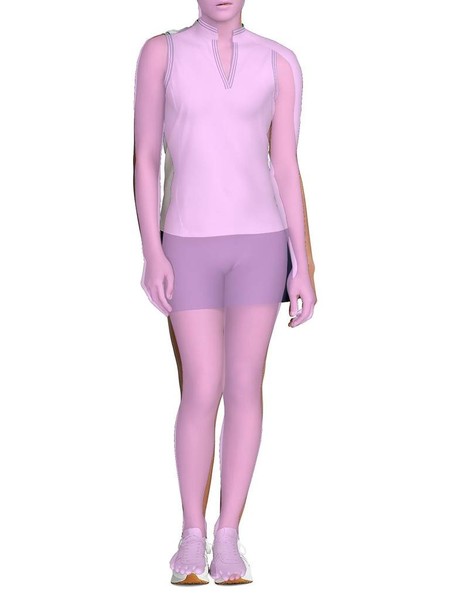}}

\caption{
Left-to-right: SMPLify-X \cite{pavlakos2019expressive} (\textcolor{caribbeangreen2}{light green}), PyMAF-X \cite{pymafx2022} (\textcolor{violet}{purple}), SHAPY \cite{choutas2022accurate} (\textcolor{jade}{green}) and KBody (\textcolor{candypink}{pink}).
}
\label{fig:partial_rl2}
\end{figure*}

%% file: figures/supp/partial_rl3.tex
\begin{figure*}[!htbp]
\captionsetup[subfigure]{position=bottom,labelformat=empty}

\centering

\subfloat{\includegraphics[height=0.24\textheight]{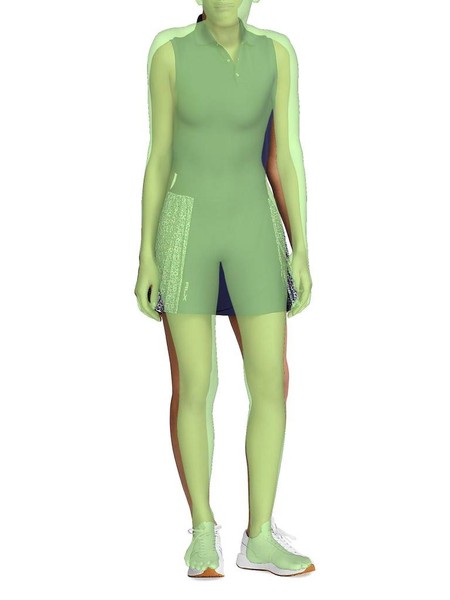}}
\subfloat{\includegraphics[height=0.24\textheight]{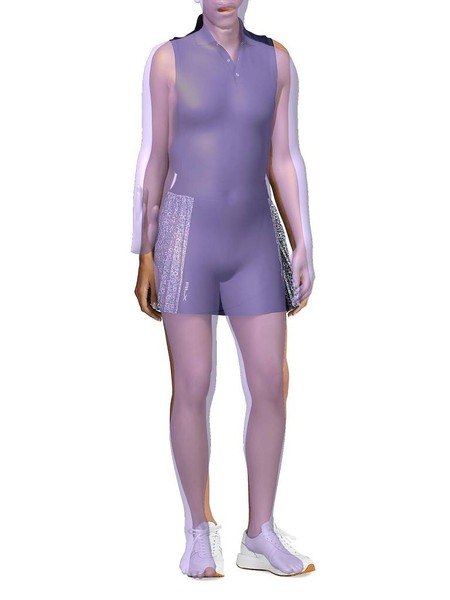}}
\subfloat{\includegraphics[height=0.24\textheight]{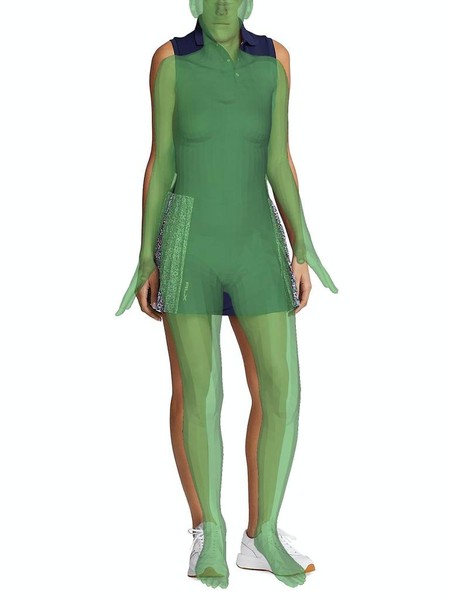}}
\subfloat{\includegraphics[height=0.24\textheight]{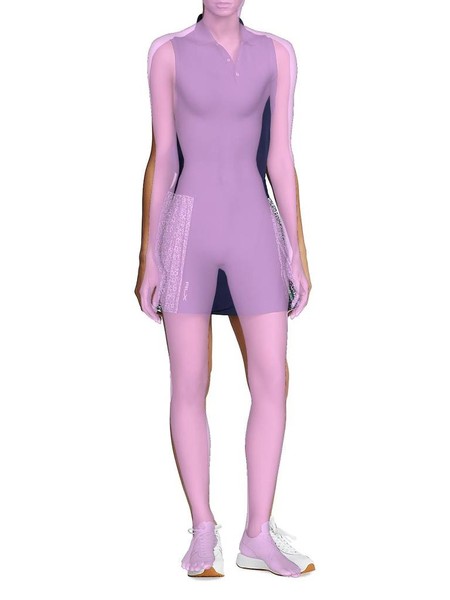}}\\

\subfloat{\includegraphics[height=0.24\textheight]{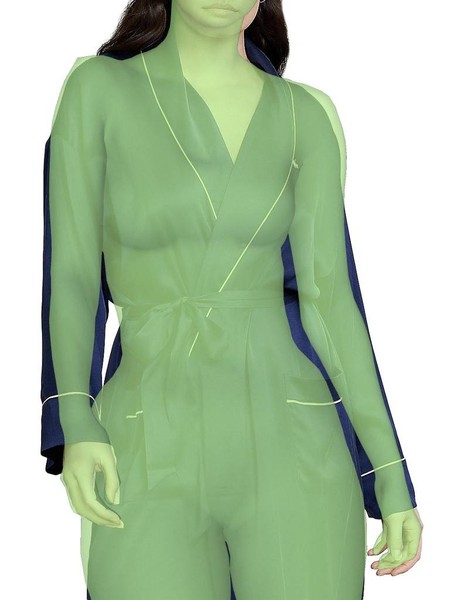}}
\subfloat{\includegraphics[height=0.24\textheight]{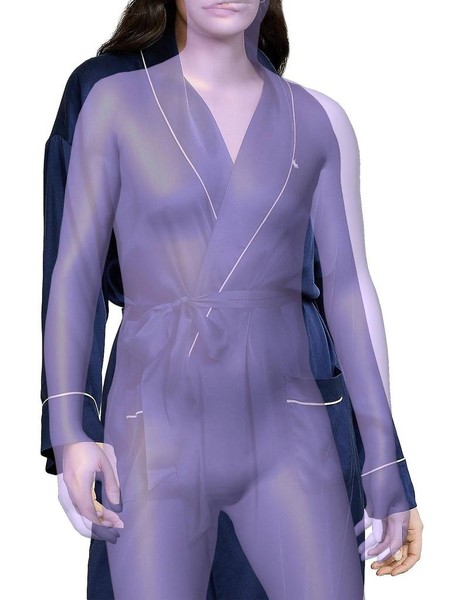}}
\subfloat{\includegraphics[height=0.24\textheight]{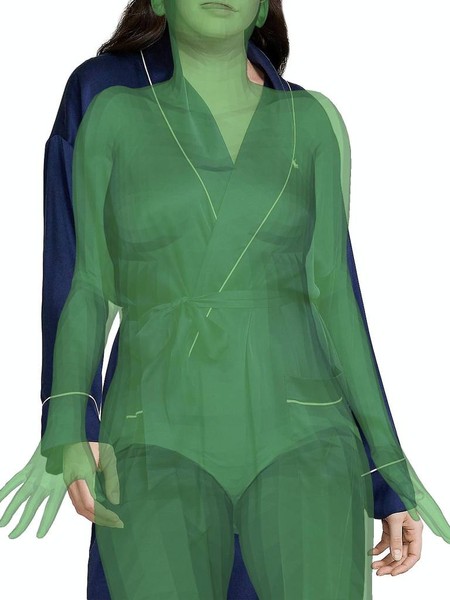}}
\subfloat{\includegraphics[height=0.24\textheight]{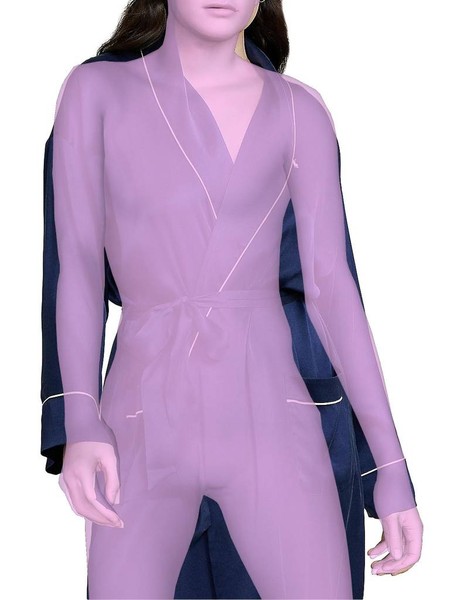}}\\

\vspace{-10pt}

\subfloat[SMPLify-X \cite{pavlakos2019expressive}]
{\includegraphics[height=0.24\textheight]{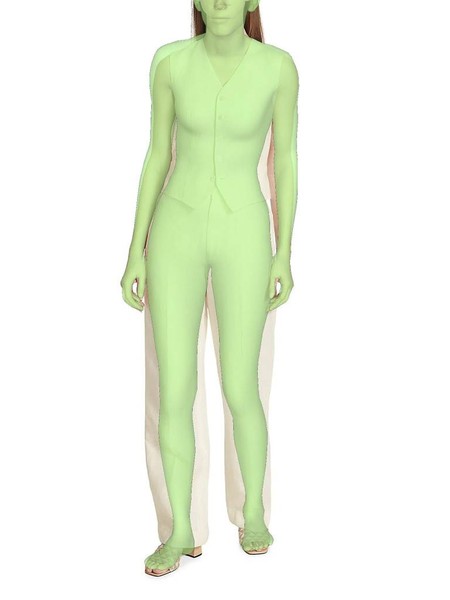}}
\subfloat[PyMAF-X \cite{pymafx2022}]{\includegraphics[height=0.24\textheight]{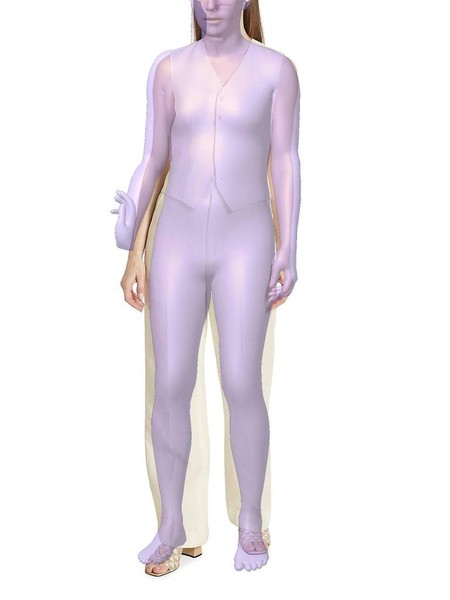}}
\subfloat[SHAPY \cite{choutas2022accurate}]
{\includegraphics[height=0.24\textheight]
{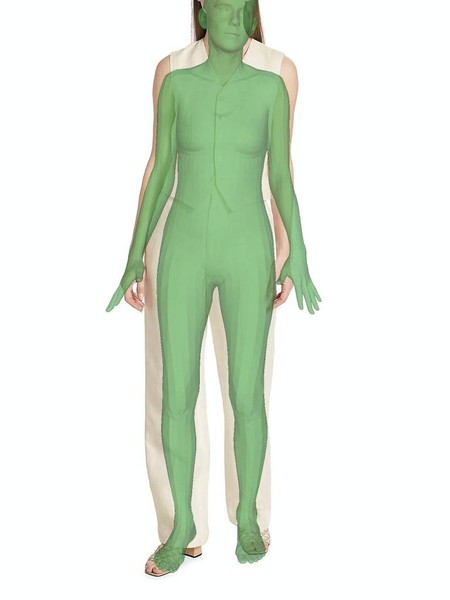}}
\subfloat[\KBody{-.1}{.035} (Ours)]{\includegraphics[height=0.24\textheight]{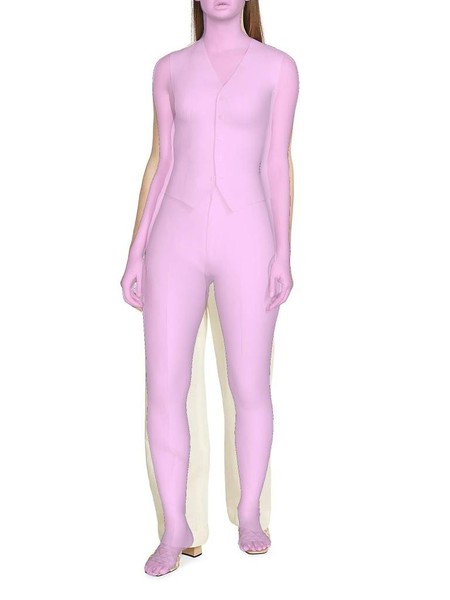}}

\caption{
Left-to-right: SMPLify-X \cite{pavlakos2019expressive} (\textcolor{caribbeangreen2}{light green}), PyMAF-X \cite{pymafx2022} (\textcolor{violet}{purple}), SHAPY \cite{choutas2022accurate} (\textcolor{jade}{green}) and KBody (\textcolor{candypink}{pink}).
}
\label{fig:partial_rl3}
\end{figure*}

%% file: figures/supp/partial_rl4.tex
\begin{figure*}[!htbp]
\captionsetup[subfigure]{position=bottom,labelformat=empty}

\centering

\subfloat{\includegraphics[height=0.24\textheight]{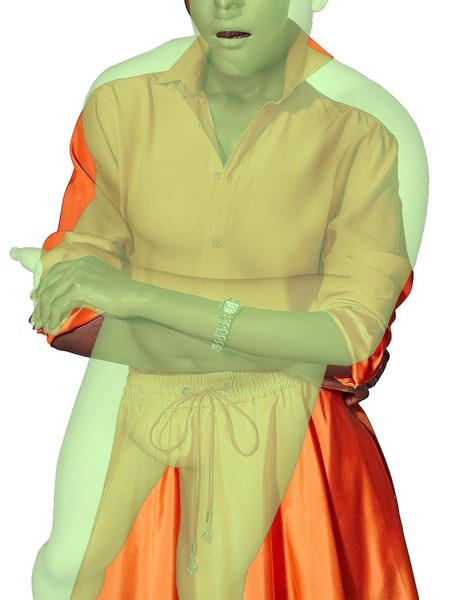}}
\subfloat{\includegraphics[height=0.24\textheight]{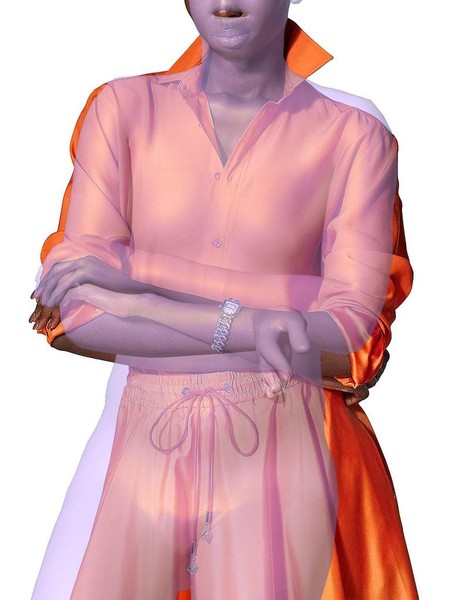}}
\subfloat{\includegraphics[height=0.24\textheight]{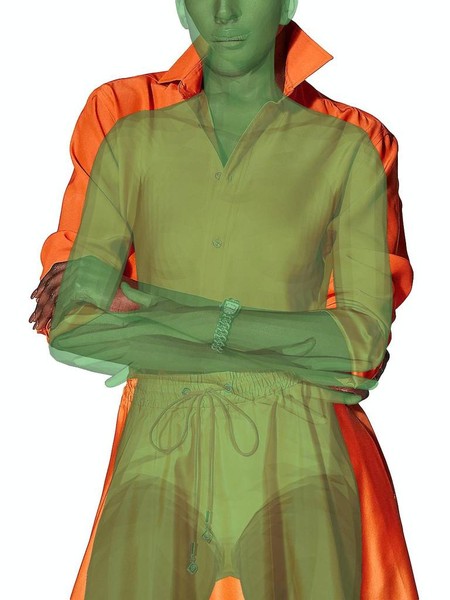}}
\subfloat{\includegraphics[height=0.24\textheight]{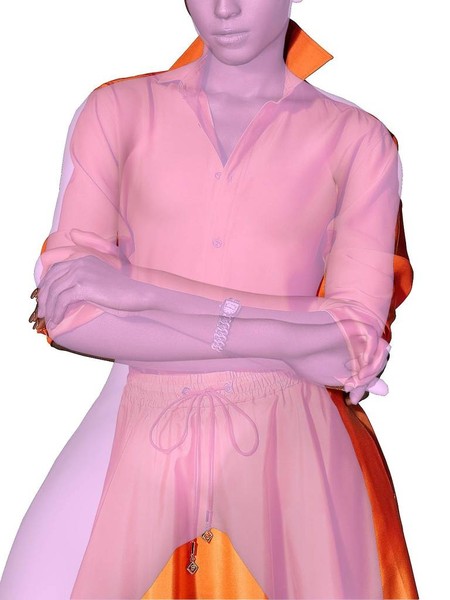}}\\

\subfloat{\includegraphics[height=0.24\textheight]{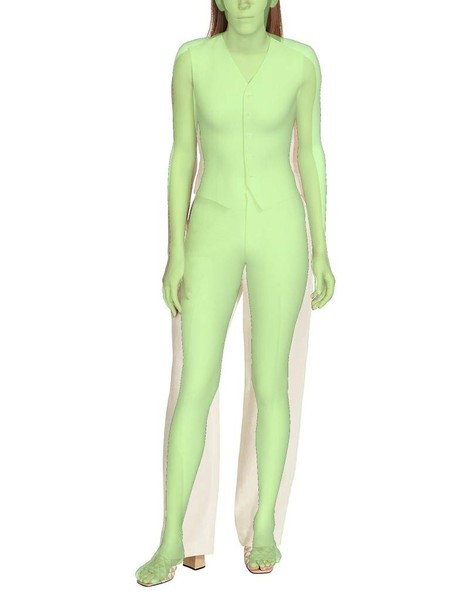}}
\subfloat{\includegraphics[height=0.24\textheight]{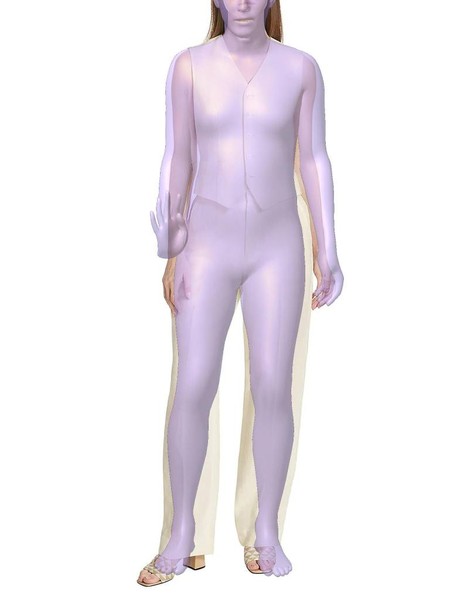}}
\subfloat{\includegraphics[height=0.24\textheight]{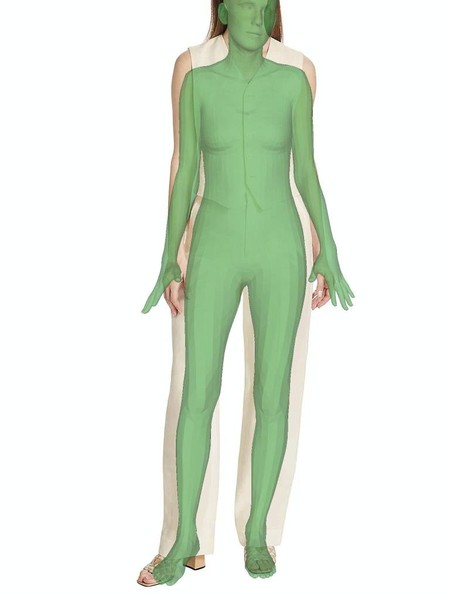}}
\subfloat{\includegraphics[height=0.24\textheight]{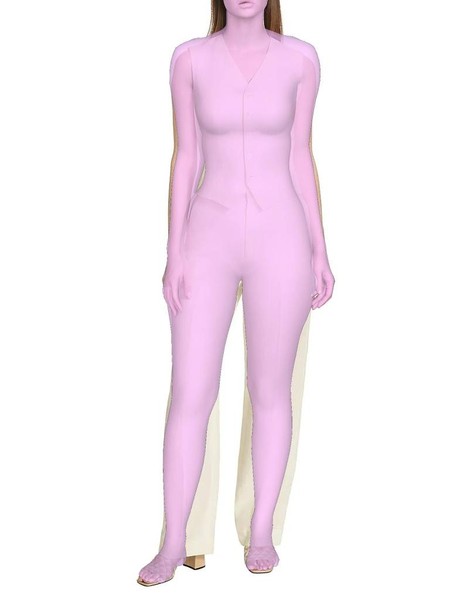}}\\

\subfloat[]{\includegraphics[height=0.24\textheight]{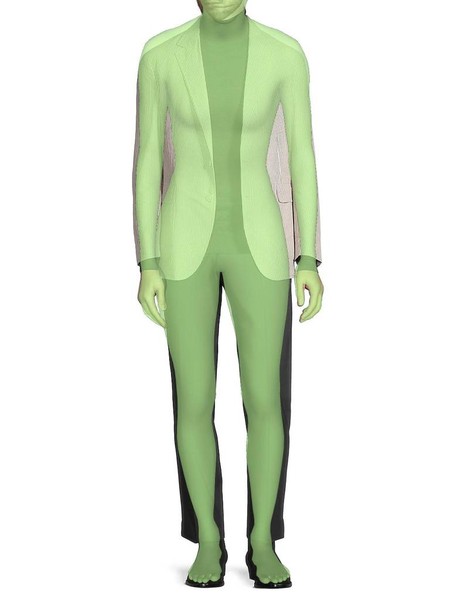}}
\subfloat[]{\includegraphics[height=0.24\textheight]{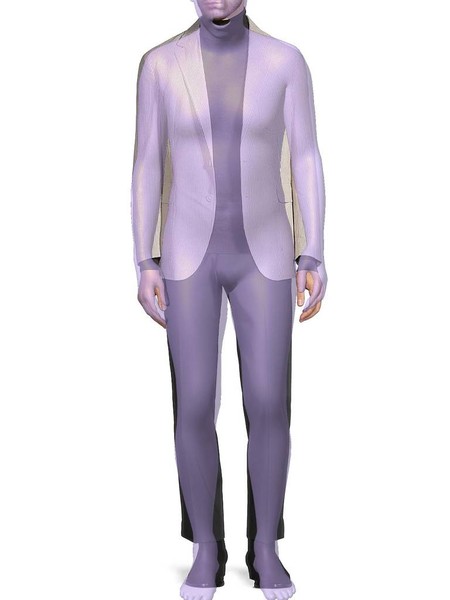}}
\subfloat[]{\includegraphics[height=0.24\textheight]{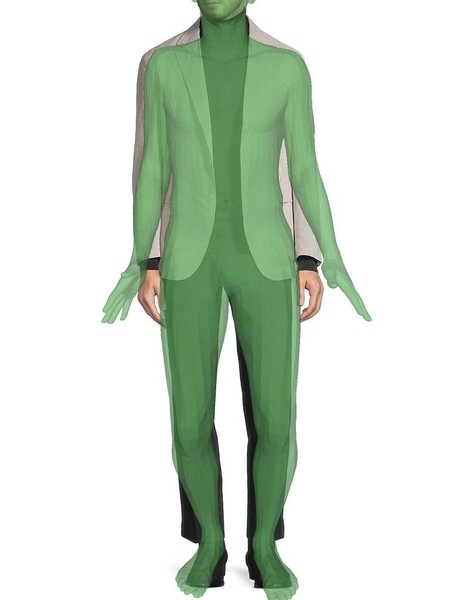}}
\subfloat[]{\includegraphics[height=0.24\textheight]{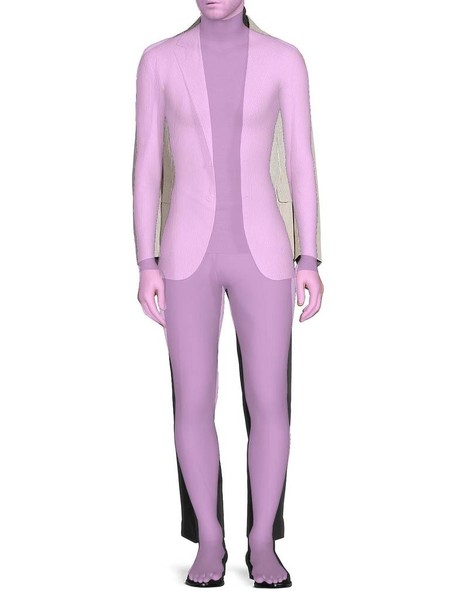}}\\

\vspace{-10pt}

\subfloat[SMPLify-X \cite{pavlakos2019expressive}]
{\includegraphics[height=0.24\textheight]{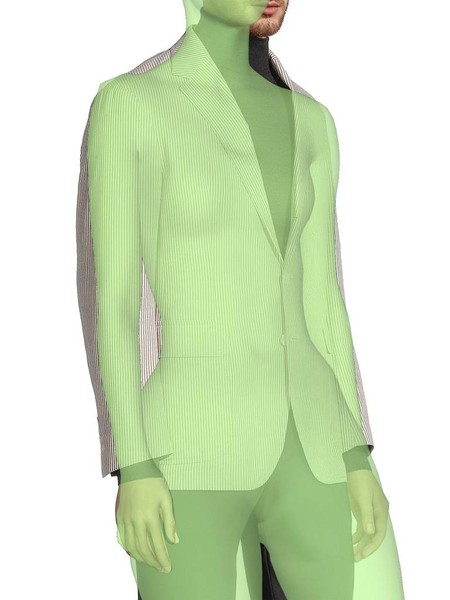}}
\subfloat[PyMAF-X \cite{pymafx2022}]{\includegraphics[height=0.24\textheight]{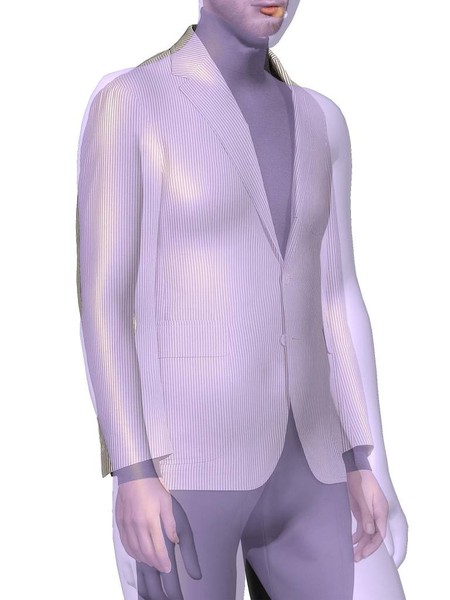}}
\subfloat[SHAPY \cite{choutas2022accurate}]
{\includegraphics[height=0.24\textheight]
{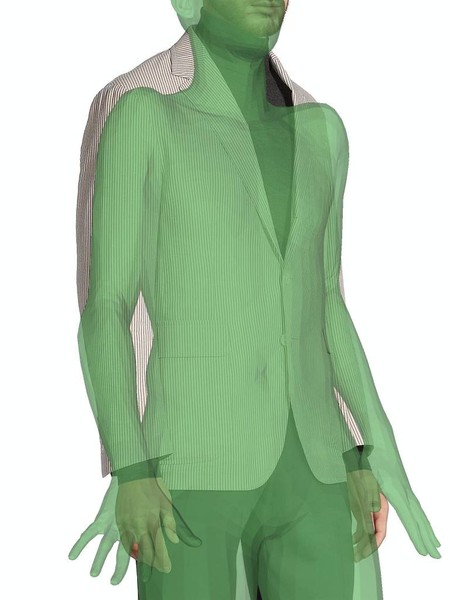}}
\subfloat[\KBody{-.1}{.035} (Ours)]{\includegraphics[height=0.24\textheight]{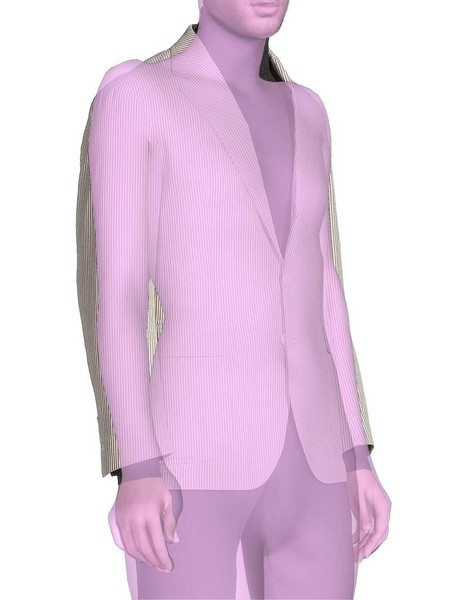}}

\caption{
Left-to-right: SMPLify-X \cite{pavlakos2019expressive} (\textcolor{caribbeangreen2}{light green}), PyMAF-X \cite{pymafx2022} (\textcolor{violet}{purple}), SHAPY \cite{choutas2022accurate} (\textcolor{jade}{green}) and KBody (\textcolor{candypink}{pink}).
}
\label{fig:partial_rl4}
\end{figure*}

%% file: figures/supp/partial_rl5.tex
\begin{figure*}[!htbp]
\captionsetup[subfigure]{position=bottom,labelformat=empty}

\centering

\subfloat{\includegraphics[height=0.24\textheight]{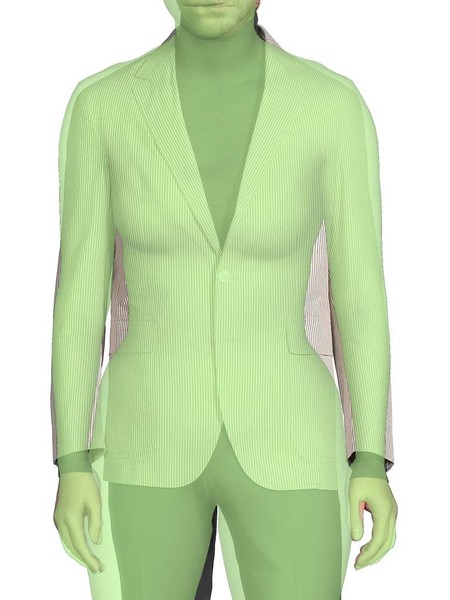}}
\subfloat{\includegraphics[height=0.24\textheight]{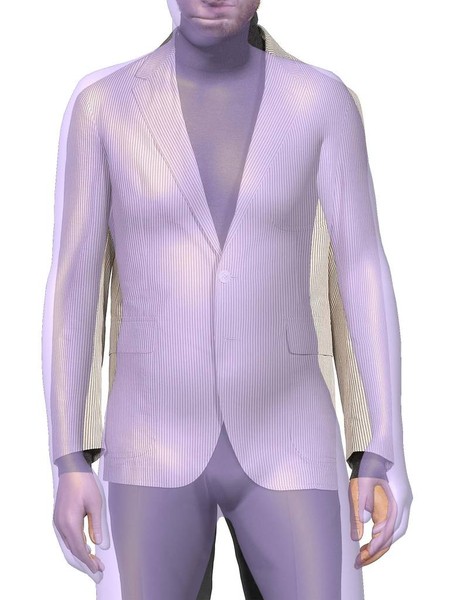}}
\subfloat{\includegraphics[height=0.24\textheight]{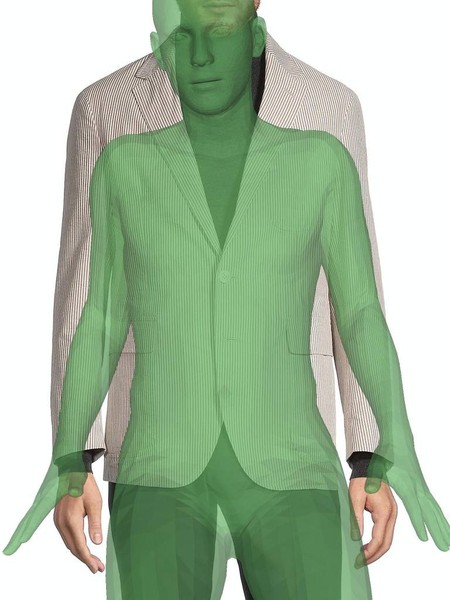}}
\subfloat{\includegraphics[height=0.24\textheight]{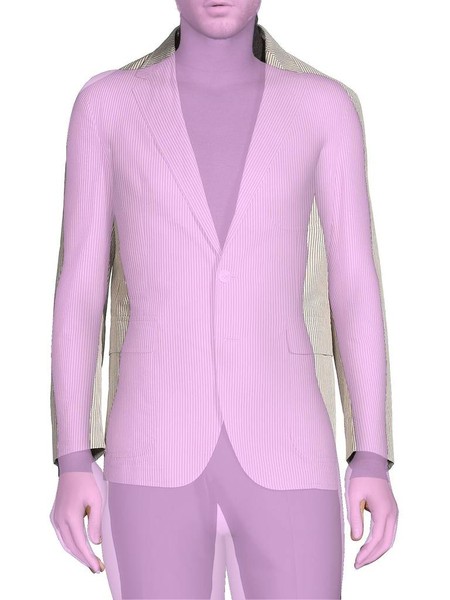}}\\

\subfloat{\includegraphics[height=0.24\textheight]{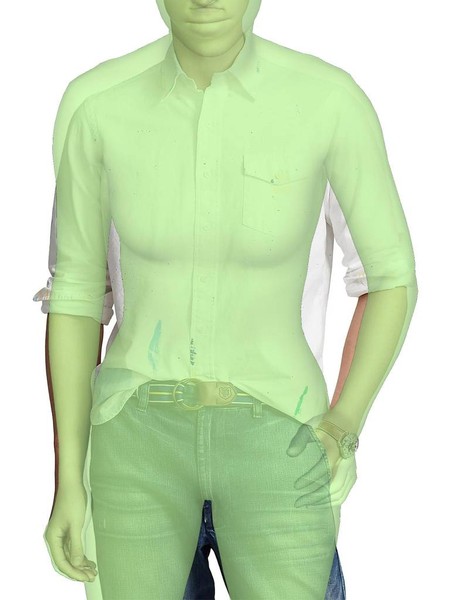}}
\subfloat{\includegraphics[height=0.24\textheight]{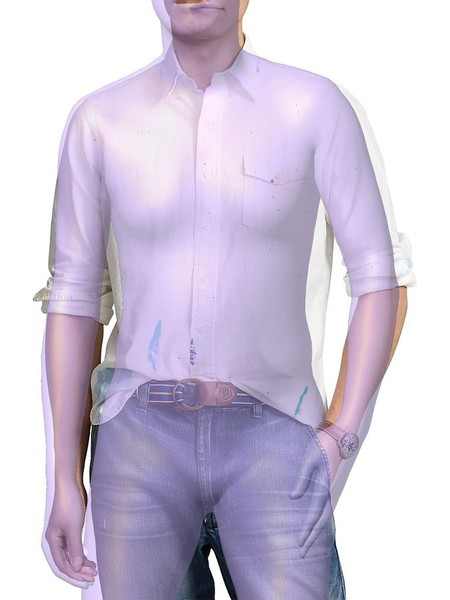}}
\subfloat{\includegraphics[height=0.24\textheight]{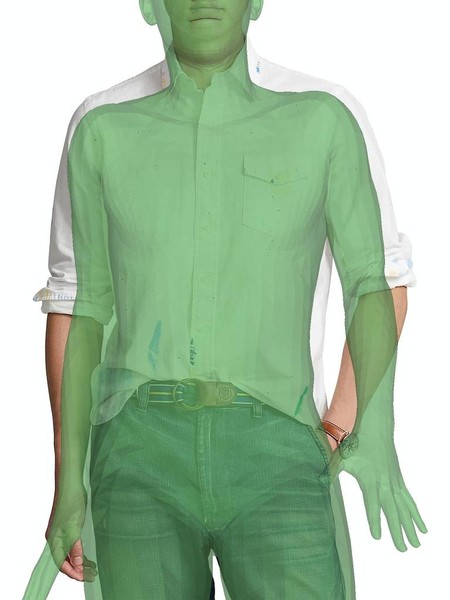}}
\subfloat{\includegraphics[height=0.24\textheight]{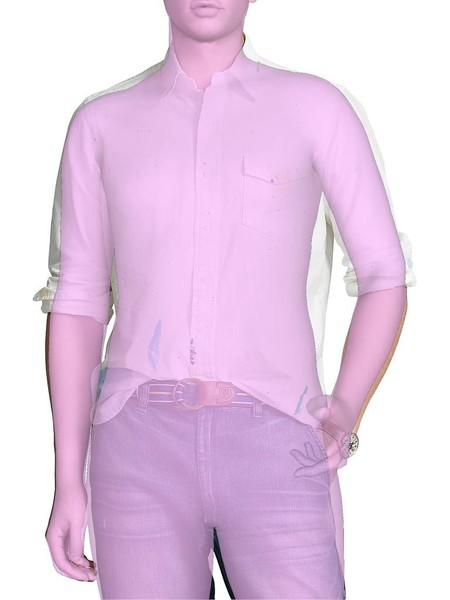}}\\

\subfloat[]{\includegraphics[height=0.24\textheight]{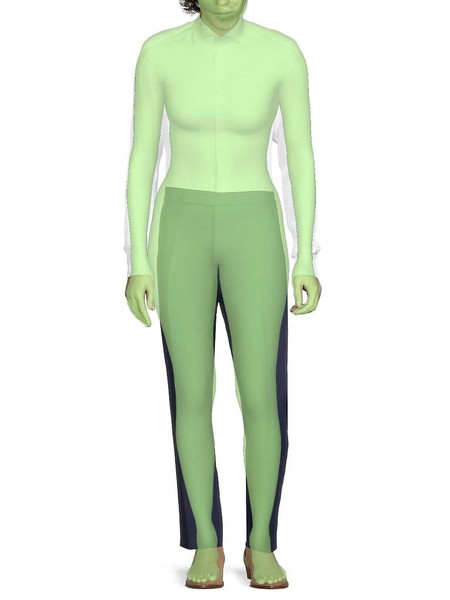}}
\subfloat[]{\includegraphics[height=0.24\textheight]{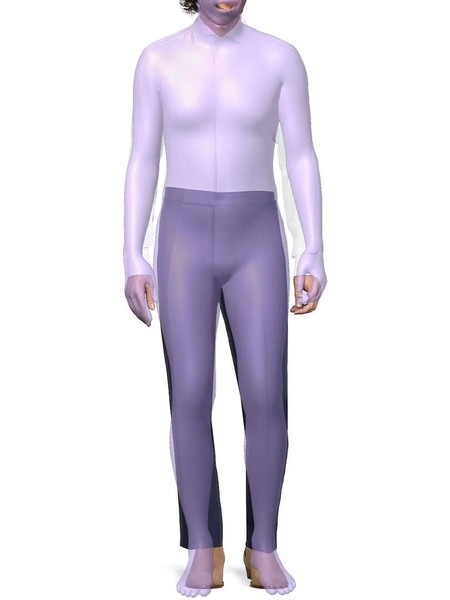}}
\subfloat[]{\includegraphics[height=0.24\textheight]{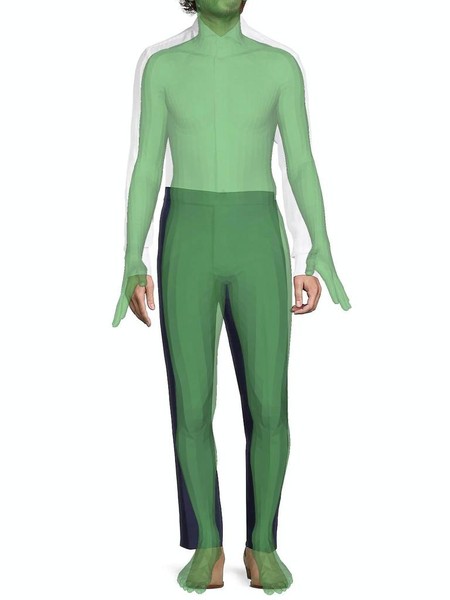}}
\subfloat[]{\includegraphics[height=0.24\textheight]{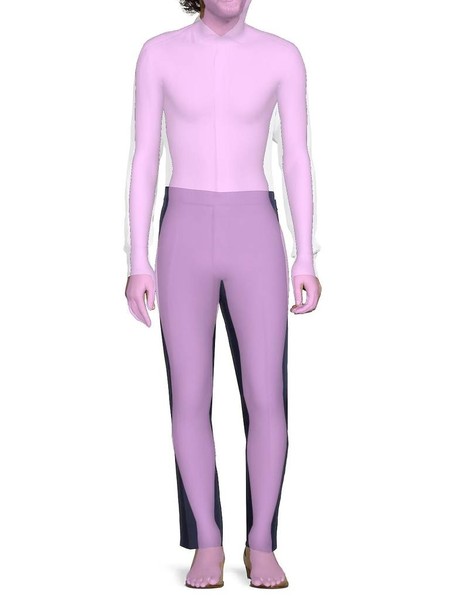}}\\

\vspace{-10pt}

\subfloat[SMPLify-X \cite{pavlakos2019expressive}]
{\includegraphics[height=0.24\textheight]{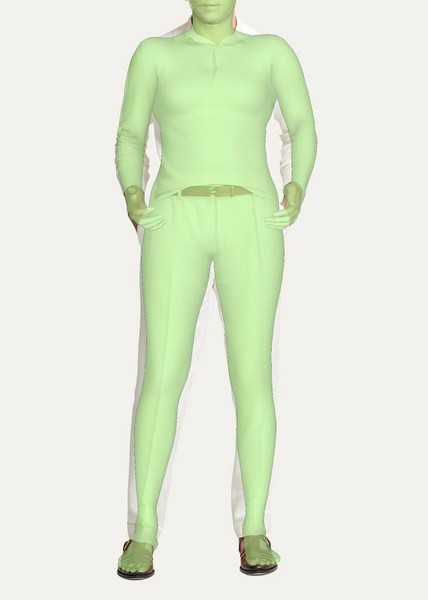}}
\subfloat[PyMAF-X \cite{pymafx2022}]{\includegraphics[height=0.24\textheight]{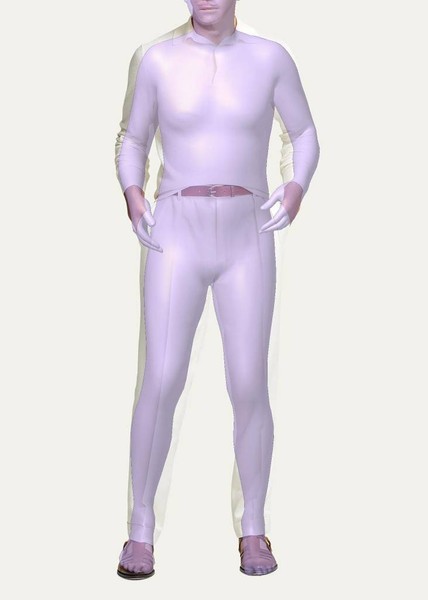}}
\subfloat[SHAPY \cite{choutas2022accurate}]
{\includegraphics[height=0.24\textheight]
{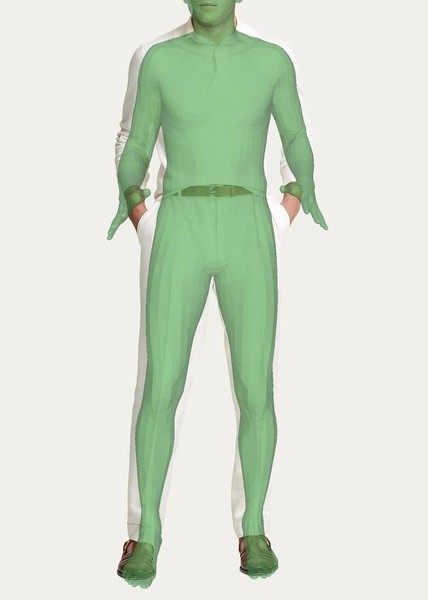}}
\subfloat[\KBody{-.1}{.035} (Ours)]{\includegraphics[height=0.24\textheight]{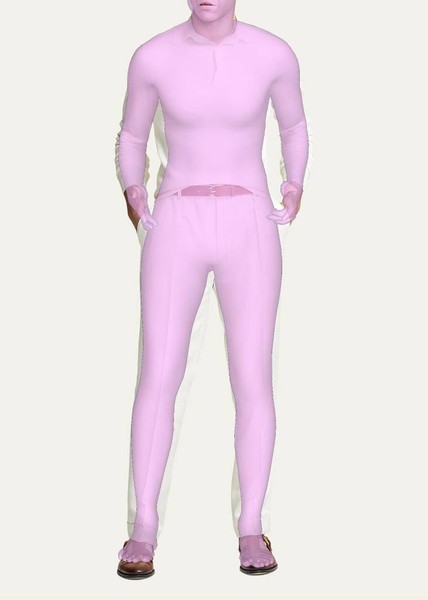}}

\caption{
Left-to-right: SMPLify-X \cite{pavlakos2019expressive} (\textcolor{caribbeangreen2}{light green}), PyMAF-X \cite{pymafx2022} (\textcolor{violet}{purple}), SHAPY \cite{choutas2022accurate} (\textcolor{jade}{green}) and KBody (\textcolor{candypink}{pink}).
}
\label{fig:partial_rl5}
\end{figure*}